\documentclass[11pt]{article}
\usepackage[a4paper,margin=1in]{geometry}
\usepackage[T1]{fontenc}
\usepackage{hyperref}
\usepackage[
  backend=biber,
  style=authoryear-comp,
  date=year,
  natbib=true,
  hyperref=true,
  backref=false,
  maxcitenames=2,
  mincitenames=1,
  maxbibnames=99,
  giveninits=true,
  uniquename=init,
  uniquelist=false,
  doi=false,
  isbn=false,
  eprint=true,
  url=false,
  dashed=false,
  sorting=nyt,
]{biblatex}

\AtEveryBibitem{\ifentrytype{inproceedings}{\clearname{editor}}{}}

\DeclareFieldFormat{urldate}{}

\let\cite\parencite
\let\citealp\barecite

\DeclareFieldFormat{citehyperref}{%
  \DeclareFieldAlias{bibhyperref}{noformat}%
  \bibhyperref{#1}}

\DeclareFieldFormat{textcitehyperref}{%
  \DeclareFieldAlias{bibhyperref}{noformat}%
  \bibhyperref{%
    #1%
    \ifbool{cbx:parens}
      {\bibcloseparen\global\boolfalse{cbx:parens}}
      {}}}

\savebibmacro{cite}
\savebibmacro{textcite}

\renewbibmacro*{cite}{%
  \printtext[citehyperref]{%
    \restorebibmacro{cite}%
    \usebibmacro{cite}}}

\renewbibmacro*{textcite}{%
  \ifboolexpr{
    ( not test {\iffieldundef{prefixnumber}} and
      test {\ifdefequal{\cbx@textcite@prefix}{\cbx@lasthash}} )
    or
    test {\iffieldequalstr{namehash}{\cbx@lasthash}}
  }
    {\global\boolfalse{cbx:parens}}
    {}%
  \printtext[textcitehyperref]{%
    \restorebibmacro{textcite}%
    \usebibmacro{textcite}}}

\let\cite\parencite

\DeclareNameAlias{default}{family-given}
\DeclareNameAlias{sortname}{family-given}   

\usepackage{microtype}
\hypersetup{hypertexnames=false}
\usepackage{graphicx}
\usepackage{adjustbox}
\usepackage{multirow}
\usepackage{amsmath,amssymb,amsfonts}
\usepackage{tabularx}
\usepackage{array}
\usepackage{tikz}
\usetikzlibrary{shapes,arrows,positioning,arrows.meta,decorations.pathreplacing,fit,backgrounds,calc,shadows}
\usepackage{pgfplots}
\pgfplotsset{compat=1.17}
\usepackage{algorithm}
\usepackage{algorithmicx}
\usepackage{algpseudocode}
\usepackage{booktabs}
\usepackage{pifont}
\usepackage{xcolor}
\usepackage{etoolbox}
\usepackage{colortbl}  
\usepackage{alltt}
\usepackage{subcaption}
\definecolor{mcInput}{HTML}{E8EEF2}
\definecolor{mcInputBdr}{HTML}{7B9BAF}
\definecolor{mcEncoder}{HTML}{D4E6F1}
\definecolor{mcEncoderBdr}{HTML}{2980B9}
\definecolor{mcEmbed}{HTML}{D5F5E3}
\definecolor{mcEmbedBdr}{HTML}{27AE60}
\definecolor{mcFusion}{HTML}{FADBD8}
\definecolor{mcFusionBdr}{HTML}{E74C3C}
\definecolor{mcOutput}{HTML}{F9E79F}
\definecolor{mcOutputBdr}{HTML}{F39C12}
\definecolor{mcText}{HTML}{E8DAEF}
\definecolor{mcTextBdr}{HTML}{8E44AD}
\definecolor{mcTemporal}{HTML}{FDEBD0}
\definecolor{mcTemporalBdr}{HTML}{E67E22}
\definecolor{mcHead}{HTML}{D7BDE2}
\definecolor{mcHeadBdr}{HTML}{7D3C98}
\definecolor{mcArrow}{HTML}{566573}
\definecolor{mcLabel}{HTML}{2C3E50}

\tikzset{
  mcbase/.style={rectangle, rounded corners=5pt, draw, thick,
    text centered, font=\sffamily\footnotesize,
    drop shadow={shadow xshift=0.5pt, shadow yshift=-0.5pt,
      fill=black!15, opacity=0.4}},
  mcinput/.style={mcbase, fill=mcInput, draw=mcInputBdr,
    text width=2.2cm, minimum height=1.3cm},
  mcencoder/.style={mcbase, fill=mcEncoder, draw=mcEncoderBdr,
    text width=2.4cm, minimum height=1.5cm},
  mcembed/.style={mcbase, fill=mcEmbed, draw=mcEmbedBdr,
    text width=2.2cm, minimum height=1.0cm},
  mcfusion/.style={mcbase, fill=mcFusion, draw=mcFusionBdr,
    text width=2.6cm, minimum height=1.8cm},
  mchead/.style={mcbase, fill=mcHead, draw=mcHeadBdr,
    text width=2.6cm, minimum height=1.8cm},
  mcoutput/.style={mcbase, fill=mcOutput, draw=mcOutputBdr,
    text width=2.4cm, minimum height=1.2cm},
  mctext/.style={mcbase, fill=mcText, draw=mcTextBdr,
    text width=2.2cm, minimum height=1.3cm},
  mctemporal/.style={mcbase, fill=mcTemporal, draw=mcTemporalBdr},
  mcarrow/.style={-{Stealth[length=5pt,width=4pt]}, semithick, mcArrow},
  mcdimlabel/.style={font=\sffamily\scriptsize, mcLabel, midway},
  mcstage/.style={font=\sffamily\bfseries\normalsize, mcLabel},
}
\usepackage{textcomp}
\usepackage{placeins}
\usepackage{needspace}
\newsavebox{\ResultTableBox}
\newcommand{\fullwidthresulttable}[2]{%
  \sbox{\ResultTableBox}{\begin{tabular}{@{}#1@{}}#2\end{tabular}}%
  \ifdim\wd\ResultTableBox>\linewidth
    \resizebox{\linewidth}{!}{\usebox{\ResultTableBox}}%
  \else
    \begin{tabular*}{\linewidth}{@{\extracolsep{\fill}}#1@{}}#2\end{tabular*}%
  \fi
}
\usepackage{float}  
\usepackage{makecell}
\usepackage{multicol}
\usepackage{enumitem}
\usepackage[title]{appendix}

\usepackage[framemethod=tikz]{mdframed}

\mdfdefinestyle{DashedBox}{
  linecolor=black,
  linewidth=0.5pt,
  tikzsetting={dash pattern=on 3pt off 2pt},
  roundcorner=3pt,
  backgroundcolor=gray!5,
  innertopmargin=6pt,
  innerbottommargin=6pt,
  innerleftmargin=8pt,
  innerrightmargin=8pt
}

\usepackage{fancyvrb}
\usepackage{fvextra}

\newcommand{\cmark}{\ding{51}}%
\newcommand{\xmark}{\ding{55}}%

\DeclareRobustCommand{\resultvalue}[1]{\ifcsname mcresult@#1\endcsname\csname mcresult@#1\endcsname\else\PackageError{MultiCyclone}{Unknown result key: #1}{Regenerate manuscript assets.}\fi}
\expandafter\def\csname mcresult@bdeck/all_bdeck_common/bdeck/DSHP/TD_mae\endcsname{8.95}
\expandafter\def\csname mcresult@bdeck/all_bdeck_common/bdeck/DSHP/TD_minus_aid_mae\endcsname{-0.30}
\expandafter\def\csname mcresult@bdeck/all_bdeck_common/bdeck/DSHP/TID_mae\endcsname{8.99}
\expandafter\def\csname mcresult@bdeck/all_bdeck_common/bdeck/DSHP/T_mae\endcsname{9.92}
\expandafter\def\csname mcresult@bdeck/all_bdeck_common/bdeck/DSHP/aid_mae\endcsname{9.26}
\expandafter\def\csname mcresult@bdeck/all_bdeck_common/bdeck/DSHP/changed_future_truth_rows\endcsname{41}
\expandafter\def\csname mcresult@bdeck/all_bdeck_common/bdeck/DSHP/changed_initial_anchor_rows\endcsname{33}
\expandafter\def\csname mcresult@bdeck/all_bdeck_common/bdeck/DSHP/n_obs\endcsname{1{,}121}
\expandafter\def\csname mcresult@bdeck/all_bdeck_common/bdeck/DSHP/n_storms\endcsname{66}
\expandafter\def\csname mcresult@bdeck/all_bdeck_common/bdeck/DSHP/storm_ci_high\endcsname{0.19}
\expandafter\def\csname mcresult@bdeck/all_bdeck_common/bdeck/DSHP/storm_ci_low\endcsname{-0.84}
\expandafter\def\csname mcresult@bdeck/all_bdeck_common/bdeck/HF3A/TD_mae\endcsname{8.28}
\expandafter\def\csname mcresult@bdeck/all_bdeck_common/bdeck/HF3A/TD_minus_aid_mae\endcsname{-0.57}
\expandafter\def\csname mcresult@bdeck/all_bdeck_common/bdeck/HF3A/TID_mae\endcsname{8.26}
\expandafter\def\csname mcresult@bdeck/all_bdeck_common/bdeck/HF3A/T_mae\endcsname{11.77}
\expandafter\def\csname mcresult@bdeck/all_bdeck_common/bdeck/HF3A/aid_mae\endcsname{8.85}
\expandafter\def\csname mcresult@bdeck/all_bdeck_common/bdeck/HF3A/changed_future_truth_rows\endcsname{0}
\expandafter\def\csname mcresult@bdeck/all_bdeck_common/bdeck/HF3A/changed_initial_anchor_rows\endcsname{0}
\expandafter\def\csname mcresult@bdeck/all_bdeck_common/bdeck/HF3A/n_obs\endcsname{158}
\expandafter\def\csname mcresult@bdeck/all_bdeck_common/bdeck/HF3A/n_storms\endcsname{13}
\expandafter\def\csname mcresult@bdeck/all_bdeck_common/bdeck/HF3A/storm_ci_high\endcsname{2.35}
\expandafter\def\csname mcresult@bdeck/all_bdeck_common/bdeck/HF3A/storm_ci_low\endcsname{-3.35}
\expandafter\def\csname mcresult@bdeck/all_bdeck_common/bdeck/HFAI/TD_mae\endcsname{9.10}
\expandafter\def\csname mcresult@bdeck/all_bdeck_common/bdeck/HFAI/TD_minus_aid_mae\endcsname{-0.61}
\expandafter\def\csname mcresult@bdeck/all_bdeck_common/bdeck/HFAI/TID_mae\endcsname{9.10}
\expandafter\def\csname mcresult@bdeck/all_bdeck_common/bdeck/HFAI/T_mae\endcsname{9.75}
\expandafter\def\csname mcresult@bdeck/all_bdeck_common/bdeck/HFAI/aid_mae\endcsname{9.72}
\expandafter\def\csname mcresult@bdeck/all_bdeck_common/bdeck/HFAI/changed_future_truth_rows\endcsname{41}
\expandafter\def\csname mcresult@bdeck/all_bdeck_common/bdeck/HFAI/changed_initial_anchor_rows\endcsname{33}
\expandafter\def\csname mcresult@bdeck/all_bdeck_common/bdeck/HFAI/n_obs\endcsname{641}
\expandafter\def\csname mcresult@bdeck/all_bdeck_common/bdeck/HFAI/n_storms\endcsname{35}
\expandafter\def\csname mcresult@bdeck/all_bdeck_common/bdeck/HFAI/storm_ci_high\endcsname{0.53}
\expandafter\def\csname mcresult@bdeck/all_bdeck_common/bdeck/HFAI/storm_ci_low\endcsname{-1.79}
\expandafter\def\csname mcresult@bdeck/all_bdeck_common/bdeck/HFBI/TD_mae\endcsname{9.10}
\expandafter\def\csname mcresult@bdeck/all_bdeck_common/bdeck/HFBI/TD_minus_aid_mae\endcsname{-0.53}
\expandafter\def\csname mcresult@bdeck/all_bdeck_common/bdeck/HFBI/TID_mae\endcsname{9.10}
\expandafter\def\csname mcresult@bdeck/all_bdeck_common/bdeck/HFBI/T_mae\endcsname{9.75}
\expandafter\def\csname mcresult@bdeck/all_bdeck_common/bdeck/HFBI/aid_mae\endcsname{9.63}
\expandafter\def\csname mcresult@bdeck/all_bdeck_common/bdeck/HFBI/changed_future_truth_rows\endcsname{41}
\expandafter\def\csname mcresult@bdeck/all_bdeck_common/bdeck/HFBI/changed_initial_anchor_rows\endcsname{33}
\expandafter\def\csname mcresult@bdeck/all_bdeck_common/bdeck/HFBI/n_obs\endcsname{641}
\expandafter\def\csname mcresult@bdeck/all_bdeck_common/bdeck/HFBI/n_storms\endcsname{35}
\expandafter\def\csname mcresult@bdeck/all_bdeck_common/bdeck/HFBI/storm_ci_high\endcsname{0.36}
\expandafter\def\csname mcresult@bdeck/all_bdeck_common/bdeck/HFBI/storm_ci_low\endcsname{-1.41}
\expandafter\def\csname mcresult@bdeck/all_bdeck_common/bdeck/HFSA/TD_mae\endcsname{9.10}
\expandafter\def\csname mcresult@bdeck/all_bdeck_common/bdeck/HFSA/TD_minus_aid_mae\endcsname{-1.73}
\expandafter\def\csname mcresult@bdeck/all_bdeck_common/bdeck/HFSA/TID_mae\endcsname{9.10}
\expandafter\def\csname mcresult@bdeck/all_bdeck_common/bdeck/HFSA/T_mae\endcsname{9.74}
\expandafter\def\csname mcresult@bdeck/all_bdeck_common/bdeck/HFSA/aid_mae\endcsname{10.83}
\expandafter\def\csname mcresult@bdeck/all_bdeck_common/bdeck/HFSA/changed_future_truth_rows\endcsname{41}
\expandafter\def\csname mcresult@bdeck/all_bdeck_common/bdeck/HFSA/changed_initial_anchor_rows\endcsname{33}
\expandafter\def\csname mcresult@bdeck/all_bdeck_common/bdeck/HFSA/n_obs\endcsname{646}
\expandafter\def\csname mcresult@bdeck/all_bdeck_common/bdeck/HFSA/n_storms\endcsname{35}
\expandafter\def\csname mcresult@bdeck/all_bdeck_common/bdeck/HFSA/storm_ci_high\endcsname{-0.78}
\expandafter\def\csname mcresult@bdeck/all_bdeck_common/bdeck/HFSA/storm_ci_low\endcsname{-2.68}
\expandafter\def\csname mcresult@bdeck/all_bdeck_common/bdeck/HFSB/TD_mae\endcsname{9.10}
\expandafter\def\csname mcresult@bdeck/all_bdeck_common/bdeck/HFSB/TD_minus_aid_mae\endcsname{-1.96}
\expandafter\def\csname mcresult@bdeck/all_bdeck_common/bdeck/HFSB/TID_mae\endcsname{9.10}
\expandafter\def\csname mcresult@bdeck/all_bdeck_common/bdeck/HFSB/T_mae\endcsname{9.74}
\expandafter\def\csname mcresult@bdeck/all_bdeck_common/bdeck/HFSB/aid_mae\endcsname{11.06}
\expandafter\def\csname mcresult@bdeck/all_bdeck_common/bdeck/HFSB/changed_future_truth_rows\endcsname{41}
\expandafter\def\csname mcresult@bdeck/all_bdeck_common/bdeck/HFSB/changed_initial_anchor_rows\endcsname{33}
\expandafter\def\csname mcresult@bdeck/all_bdeck_common/bdeck/HFSB/n_obs\endcsname{646}
\expandafter\def\csname mcresult@bdeck/all_bdeck_common/bdeck/HFSB/n_storms\endcsname{35}
\expandafter\def\csname mcresult@bdeck/all_bdeck_common/bdeck/HFSB/storm_ci_high\endcsname{-1.20}
\expandafter\def\csname mcresult@bdeck/all_bdeck_common/bdeck/HFSB/storm_ci_low\endcsname{-2.65}
\expandafter\def\csname mcresult@bdeck/all_bdeck_common/bdeck/HWFI/TD_mae\endcsname{8.97}
\expandafter\def\csname mcresult@bdeck/all_bdeck_common/bdeck/HWFI/TD_minus_aid_mae\endcsname{0.05}
\expandafter\def\csname mcresult@bdeck/all_bdeck_common/bdeck/HWFI/TID_mae\endcsname{8.99}
\expandafter\def\csname mcresult@bdeck/all_bdeck_common/bdeck/HWFI/T_mae\endcsname{9.98}
\expandafter\def\csname mcresult@bdeck/all_bdeck_common/bdeck/HWFI/aid_mae\endcsname{8.92}
\expandafter\def\csname mcresult@bdeck/all_bdeck_common/bdeck/HWFI/changed_future_truth_rows\endcsname{40}
\expandafter\def\csname mcresult@bdeck/all_bdeck_common/bdeck/HWFI/changed_initial_anchor_rows\endcsname{32}
\expandafter\def\csname mcresult@bdeck/all_bdeck_common/bdeck/HWFI/n_obs\endcsname{1{,}065}
\expandafter\def\csname mcresult@bdeck/all_bdeck_common/bdeck/HWFI/n_storms\endcsname{64}
\expandafter\def\csname mcresult@bdeck/all_bdeck_common/bdeck/HWFI/storm_ci_high\endcsname{0.71}
\expandafter\def\csname mcresult@bdeck/all_bdeck_common/bdeck/HWFI/storm_ci_low\endcsname{-0.64}
\expandafter\def\csname mcresult@bdeck/all_bdeck_common/bdeck/HWRF/TD_mae\endcsname{9.02}
\expandafter\def\csname mcresult@bdeck/all_bdeck_common/bdeck/HWRF/TD_minus_aid_mae\endcsname{-0.73}
\expandafter\def\csname mcresult@bdeck/all_bdeck_common/bdeck/HWRF/TID_mae\endcsname{9.06}
\expandafter\def\csname mcresult@bdeck/all_bdeck_common/bdeck/HWRF/T_mae\endcsname{9.98}
\expandafter\def\csname mcresult@bdeck/all_bdeck_common/bdeck/HWRF/aid_mae\endcsname{9.75}
\expandafter\def\csname mcresult@bdeck/all_bdeck_common/bdeck/HWRF/changed_future_truth_rows\endcsname{41}
\expandafter\def\csname mcresult@bdeck/all_bdeck_common/bdeck/HWRF/changed_initial_anchor_rows\endcsname{33}
\expandafter\def\csname mcresult@bdeck/all_bdeck_common/bdeck/HWRF/n_obs\endcsname{1{,}103}
\expandafter\def\csname mcresult@bdeck/all_bdeck_common/bdeck/HWRF/n_storms\endcsname{64}
\expandafter\def\csname mcresult@bdeck/all_bdeck_common/bdeck/HWRF/storm_ci_high\endcsname{0.19}
\expandafter\def\csname mcresult@bdeck/all_bdeck_common/bdeck/HWRF/storm_ci_low\endcsname{-1.57}
\expandafter\def\csname mcresult@bdeck/all_bdeck_common/bdeck/LGEM/TD_mae\endcsname{8.95}
\expandafter\def\csname mcresult@bdeck/all_bdeck_common/bdeck/LGEM/TD_minus_aid_mae\endcsname{-0.32}
\expandafter\def\csname mcresult@bdeck/all_bdeck_common/bdeck/LGEM/TID_mae\endcsname{8.99}
\expandafter\def\csname mcresult@bdeck/all_bdeck_common/bdeck/LGEM/T_mae\endcsname{9.92}
\expandafter\def\csname mcresult@bdeck/all_bdeck_common/bdeck/LGEM/aid_mae\endcsname{9.27}
\expandafter\def\csname mcresult@bdeck/all_bdeck_common/bdeck/LGEM/changed_future_truth_rows\endcsname{41}
\expandafter\def\csname mcresult@bdeck/all_bdeck_common/bdeck/LGEM/changed_initial_anchor_rows\endcsname{33}
\expandafter\def\csname mcresult@bdeck/all_bdeck_common/bdeck/LGEM/n_obs\endcsname{1{,}121}
\expandafter\def\csname mcresult@bdeck/all_bdeck_common/bdeck/LGEM/n_storms\endcsname{66}
\expandafter\def\csname mcresult@bdeck/all_bdeck_common/bdeck/LGEM/storm_ci_high\endcsname{0.24}
\expandafter\def\csname mcresult@bdeck/all_bdeck_common/bdeck/LGEM/storm_ci_low\endcsname{-0.93}
\expandafter\def\csname mcresult@bdeck/all_bdeck_common/bdeck/OFCI/TD_mae\endcsname{9.15}
\expandafter\def\csname mcresult@bdeck/all_bdeck_common/bdeck/OFCI/TD_minus_aid_mae\endcsname{0.70}
\expandafter\def\csname mcresult@bdeck/all_bdeck_common/bdeck/OFCI/TID_mae\endcsname{9.18}
\expandafter\def\csname mcresult@bdeck/all_bdeck_common/bdeck/OFCI/T_mae\endcsname{10.17}
\expandafter\def\csname mcresult@bdeck/all_bdeck_common/bdeck/OFCI/aid_mae\endcsname{8.45}
\expandafter\def\csname mcresult@bdeck/all_bdeck_common/bdeck/OFCI/changed_future_truth_rows\endcsname{41}
\expandafter\def\csname mcresult@bdeck/all_bdeck_common/bdeck/OFCI/changed_initial_anchor_rows\endcsname{32}
\expandafter\def\csname mcresult@bdeck/all_bdeck_common/bdeck/OFCI/n_obs\endcsname{1{,}017}
\expandafter\def\csname mcresult@bdeck/all_bdeck_common/bdeck/OFCI/n_storms\endcsname{64}
\expandafter\def\csname mcresult@bdeck/all_bdeck_common/bdeck/OFCI/storm_ci_high\endcsname{1.16}
\expandafter\def\csname mcresult@bdeck/all_bdeck_common/bdeck/OFCI/storm_ci_low\endcsname{0.21}
\expandafter\def\csname mcresult@bdeck/all_bdeck_common/bdeck/OFCL/TD_mae\endcsname{9.02}
\expandafter\def\csname mcresult@bdeck/all_bdeck_common/bdeck/OFCL/TD_minus_aid_mae\endcsname{1.34}
\expandafter\def\csname mcresult@bdeck/all_bdeck_common/bdeck/OFCL/TID_mae\endcsname{9.05}
\expandafter\def\csname mcresult@bdeck/all_bdeck_common/bdeck/OFCL/T_mae\endcsname{10.08}
\expandafter\def\csname mcresult@bdeck/all_bdeck_common/bdeck/OFCL/aid_mae\endcsname{7.68}
\expandafter\def\csname mcresult@bdeck/all_bdeck_common/bdeck/OFCL/changed_future_truth_rows\endcsname{41}
\expandafter\def\csname mcresult@bdeck/all_bdeck_common/bdeck/OFCL/changed_initial_anchor_rows\endcsname{33}
\expandafter\def\csname mcresult@bdeck/all_bdeck_common/bdeck/OFCL/n_obs\endcsname{1{,}077}
\expandafter\def\csname mcresult@bdeck/all_bdeck_common/bdeck/OFCL/n_storms\endcsname{64}
\expandafter\def\csname mcresult@bdeck/all_bdeck_common/bdeck/OFCL/storm_ci_high\endcsname{1.88}
\expandafter\def\csname mcresult@bdeck/all_bdeck_common/bdeck/OFCL/storm_ci_low\endcsname{0.78}
\expandafter\def\csname mcresult@bdeck/all_bdeck_common/canonical/DSHP/TD_mae\endcsname{8.90}
\expandafter\def\csname mcresult@bdeck/all_bdeck_common/canonical/DSHP/TD_minus_aid_mae\endcsname{-0.32}
\expandafter\def\csname mcresult@bdeck/all_bdeck_common/canonical/DSHP/TID_mae\endcsname{8.93}
\expandafter\def\csname mcresult@bdeck/all_bdeck_common/canonical/DSHP/T_mae\endcsname{9.91}
\expandafter\def\csname mcresult@bdeck/all_bdeck_common/canonical/DSHP/aid_mae\endcsname{9.22}
\expandafter\def\csname mcresult@bdeck/all_bdeck_common/canonical/DSHP/changed_future_truth_rows\endcsname{41}
\expandafter\def\csname mcresult@bdeck/all_bdeck_common/canonical/DSHP/changed_initial_anchor_rows\endcsname{33}
\expandafter\def\csname mcresult@bdeck/all_bdeck_common/canonical/DSHP/n_obs\endcsname{1{,}121}
\expandafter\def\csname mcresult@bdeck/all_bdeck_common/canonical/DSHP/n_storms\endcsname{66}
\expandafter\def\csname mcresult@bdeck/all_bdeck_common/canonical/DSHP/storm_ci_high\endcsname{0.17}
\expandafter\def\csname mcresult@bdeck/all_bdeck_common/canonical/DSHP/storm_ci_low\endcsname{-0.85}
\expandafter\def\csname mcresult@bdeck/all_bdeck_common/canonical/HF3A/TD_mae\endcsname{8.28}
\expandafter\def\csname mcresult@bdeck/all_bdeck_common/canonical/HF3A/TD_minus_aid_mae\endcsname{-0.57}
\expandafter\def\csname mcresult@bdeck/all_bdeck_common/canonical/HF3A/TID_mae\endcsname{8.26}
\expandafter\def\csname mcresult@bdeck/all_bdeck_common/canonical/HF3A/T_mae\endcsname{11.77}
\expandafter\def\csname mcresult@bdeck/all_bdeck_common/canonical/HF3A/aid_mae\endcsname{8.85}
\expandafter\def\csname mcresult@bdeck/all_bdeck_common/canonical/HF3A/changed_future_truth_rows\endcsname{0}
\expandafter\def\csname mcresult@bdeck/all_bdeck_common/canonical/HF3A/changed_initial_anchor_rows\endcsname{0}
\expandafter\def\csname mcresult@bdeck/all_bdeck_common/canonical/HF3A/n_obs\endcsname{158}
\expandafter\def\csname mcresult@bdeck/all_bdeck_common/canonical/HF3A/n_storms\endcsname{13}
\expandafter\def\csname mcresult@bdeck/all_bdeck_common/canonical/HF3A/storm_ci_high\endcsname{2.35}
\expandafter\def\csname mcresult@bdeck/all_bdeck_common/canonical/HF3A/storm_ci_low\endcsname{-3.35}
\expandafter\def\csname mcresult@bdeck/all_bdeck_common/canonical/HFAI/TD_mae\endcsname{9.01}
\expandafter\def\csname mcresult@bdeck/all_bdeck_common/canonical/HFAI/TD_minus_aid_mae\endcsname{-0.58}
\expandafter\def\csname mcresult@bdeck/all_bdeck_common/canonical/HFAI/TID_mae\endcsname{9.00}
\expandafter\def\csname mcresult@bdeck/all_bdeck_common/canonical/HFAI/T_mae\endcsname{9.75}
\expandafter\def\csname mcresult@bdeck/all_bdeck_common/canonical/HFAI/aid_mae\endcsname{9.59}
\expandafter\def\csname mcresult@bdeck/all_bdeck_common/canonical/HFAI/changed_future_truth_rows\endcsname{41}
\expandafter\def\csname mcresult@bdeck/all_bdeck_common/canonical/HFAI/changed_initial_anchor_rows\endcsname{33}
\expandafter\def\csname mcresult@bdeck/all_bdeck_common/canonical/HFAI/n_obs\endcsname{641}
\expandafter\def\csname mcresult@bdeck/all_bdeck_common/canonical/HFAI/n_storms\endcsname{35}
\expandafter\def\csname mcresult@bdeck/all_bdeck_common/canonical/HFAI/storm_ci_high\endcsname{0.58}
\expandafter\def\csname mcresult@bdeck/all_bdeck_common/canonical/HFAI/storm_ci_low\endcsname{-1.77}
\expandafter\def\csname mcresult@bdeck/all_bdeck_common/canonical/HFBI/TD_mae\endcsname{9.01}
\expandafter\def\csname mcresult@bdeck/all_bdeck_common/canonical/HFBI/TD_minus_aid_mae\endcsname{-0.52}
\expandafter\def\csname mcresult@bdeck/all_bdeck_common/canonical/HFBI/TID_mae\endcsname{9.00}
\expandafter\def\csname mcresult@bdeck/all_bdeck_common/canonical/HFBI/T_mae\endcsname{9.75}
\expandafter\def\csname mcresult@bdeck/all_bdeck_common/canonical/HFBI/aid_mae\endcsname{9.53}
\expandafter\def\csname mcresult@bdeck/all_bdeck_common/canonical/HFBI/changed_future_truth_rows\endcsname{41}
\expandafter\def\csname mcresult@bdeck/all_bdeck_common/canonical/HFBI/changed_initial_anchor_rows\endcsname{33}
\expandafter\def\csname mcresult@bdeck/all_bdeck_common/canonical/HFBI/n_obs\endcsname{641}
\expandafter\def\csname mcresult@bdeck/all_bdeck_common/canonical/HFBI/n_storms\endcsname{35}
\expandafter\def\csname mcresult@bdeck/all_bdeck_common/canonical/HFBI/storm_ci_high\endcsname{0.37}
\expandafter\def\csname mcresult@bdeck/all_bdeck_common/canonical/HFBI/storm_ci_low\endcsname{-1.40}
\expandafter\def\csname mcresult@bdeck/all_bdeck_common/canonical/HFSA/TD_mae\endcsname{9.01}
\expandafter\def\csname mcresult@bdeck/all_bdeck_common/canonical/HFSA/TD_minus_aid_mae\endcsname{-1.68}
\expandafter\def\csname mcresult@bdeck/all_bdeck_common/canonical/HFSA/TID_mae\endcsname{9.00}
\expandafter\def\csname mcresult@bdeck/all_bdeck_common/canonical/HFSA/T_mae\endcsname{9.73}
\expandafter\def\csname mcresult@bdeck/all_bdeck_common/canonical/HFSA/aid_mae\endcsname{10.69}
\expandafter\def\csname mcresult@bdeck/all_bdeck_common/canonical/HFSA/changed_future_truth_rows\endcsname{41}
\expandafter\def\csname mcresult@bdeck/all_bdeck_common/canonical/HFSA/changed_initial_anchor_rows\endcsname{33}
\expandafter\def\csname mcresult@bdeck/all_bdeck_common/canonical/HFSA/n_obs\endcsname{646}
\expandafter\def\csname mcresult@bdeck/all_bdeck_common/canonical/HFSA/n_storms\endcsname{35}
\expandafter\def\csname mcresult@bdeck/all_bdeck_common/canonical/HFSA/storm_ci_high\endcsname{-0.72}
\expandafter\def\csname mcresult@bdeck/all_bdeck_common/canonical/HFSA/storm_ci_low\endcsname{-2.66}
\expandafter\def\csname mcresult@bdeck/all_bdeck_common/canonical/HFSB/TD_mae\endcsname{9.01}
\expandafter\def\csname mcresult@bdeck/all_bdeck_common/canonical/HFSB/TD_minus_aid_mae\endcsname{-1.99}
\expandafter\def\csname mcresult@bdeck/all_bdeck_common/canonical/HFSB/TID_mae\endcsname{9.00}
\expandafter\def\csname mcresult@bdeck/all_bdeck_common/canonical/HFSB/T_mae\endcsname{9.73}
\expandafter\def\csname mcresult@bdeck/all_bdeck_common/canonical/HFSB/aid_mae\endcsname{11.00}
\expandafter\def\csname mcresult@bdeck/all_bdeck_common/canonical/HFSB/changed_future_truth_rows\endcsname{41}
\expandafter\def\csname mcresult@bdeck/all_bdeck_common/canonical/HFSB/changed_initial_anchor_rows\endcsname{33}
\expandafter\def\csname mcresult@bdeck/all_bdeck_common/canonical/HFSB/n_obs\endcsname{646}
\expandafter\def\csname mcresult@bdeck/all_bdeck_common/canonical/HFSB/n_storms\endcsname{35}
\expandafter\def\csname mcresult@bdeck/all_bdeck_common/canonical/HFSB/storm_ci_high\endcsname{-1.22}
\expandafter\def\csname mcresult@bdeck/all_bdeck_common/canonical/HFSB/storm_ci_low\endcsname{-2.69}
\expandafter\def\csname mcresult@bdeck/all_bdeck_common/canonical/HWFI/TD_mae\endcsname{8.91}
\expandafter\def\csname mcresult@bdeck/all_bdeck_common/canonical/HWFI/TD_minus_aid_mae\endcsname{0.06}
\expandafter\def\csname mcresult@bdeck/all_bdeck_common/canonical/HWFI/TID_mae\endcsname{8.92}
\expandafter\def\csname mcresult@bdeck/all_bdeck_common/canonical/HWFI/T_mae\endcsname{9.98}
\expandafter\def\csname mcresult@bdeck/all_bdeck_common/canonical/HWFI/aid_mae\endcsname{8.85}
\expandafter\def\csname mcresult@bdeck/all_bdeck_common/canonical/HWFI/changed_future_truth_rows\endcsname{40}
\expandafter\def\csname mcresult@bdeck/all_bdeck_common/canonical/HWFI/changed_initial_anchor_rows\endcsname{32}
\expandafter\def\csname mcresult@bdeck/all_bdeck_common/canonical/HWFI/n_obs\endcsname{1{,}065}
\expandafter\def\csname mcresult@bdeck/all_bdeck_common/canonical/HWFI/n_storms\endcsname{64}
\expandafter\def\csname mcresult@bdeck/all_bdeck_common/canonical/HWFI/storm_ci_high\endcsname{0.72}
\expandafter\def\csname mcresult@bdeck/all_bdeck_common/canonical/HWFI/storm_ci_low\endcsname{-0.63}
\expandafter\def\csname mcresult@bdeck/all_bdeck_common/canonical/HWRF/TD_mae\endcsname{8.97}
\expandafter\def\csname mcresult@bdeck/all_bdeck_common/canonical/HWRF/TD_minus_aid_mae\endcsname{-0.71}
\expandafter\def\csname mcresult@bdeck/all_bdeck_common/canonical/HWRF/TID_mae\endcsname{9.00}
\expandafter\def\csname mcresult@bdeck/all_bdeck_common/canonical/HWRF/T_mae\endcsname{9.98}
\expandafter\def\csname mcresult@bdeck/all_bdeck_common/canonical/HWRF/aid_mae\endcsname{9.68}
\expandafter\def\csname mcresult@bdeck/all_bdeck_common/canonical/HWRF/changed_future_truth_rows\endcsname{41}
\expandafter\def\csname mcresult@bdeck/all_bdeck_common/canonical/HWRF/changed_initial_anchor_rows\endcsname{33}
\expandafter\def\csname mcresult@bdeck/all_bdeck_common/canonical/HWRF/n_obs\endcsname{1{,}103}
\expandafter\def\csname mcresult@bdeck/all_bdeck_common/canonical/HWRF/n_storms\endcsname{64}
\expandafter\def\csname mcresult@bdeck/all_bdeck_common/canonical/HWRF/storm_ci_high\endcsname{0.20}
\expandafter\def\csname mcresult@bdeck/all_bdeck_common/canonical/HWRF/storm_ci_low\endcsname{-1.56}
\expandafter\def\csname mcresult@bdeck/all_bdeck_common/canonical/LGEM/TD_mae\endcsname{8.90}
\expandafter\def\csname mcresult@bdeck/all_bdeck_common/canonical/LGEM/TD_minus_aid_mae\endcsname{-0.34}
\expandafter\def\csname mcresult@bdeck/all_bdeck_common/canonical/LGEM/TID_mae\endcsname{8.93}
\expandafter\def\csname mcresult@bdeck/all_bdeck_common/canonical/LGEM/T_mae\endcsname{9.91}
\expandafter\def\csname mcresult@bdeck/all_bdeck_common/canonical/LGEM/aid_mae\endcsname{9.24}
\expandafter\def\csname mcresult@bdeck/all_bdeck_common/canonical/LGEM/changed_future_truth_rows\endcsname{41}
\expandafter\def\csname mcresult@bdeck/all_bdeck_common/canonical/LGEM/changed_initial_anchor_rows\endcsname{33}
\expandafter\def\csname mcresult@bdeck/all_bdeck_common/canonical/LGEM/n_obs\endcsname{1{,}121}
\expandafter\def\csname mcresult@bdeck/all_bdeck_common/canonical/LGEM/n_storms\endcsname{66}
\expandafter\def\csname mcresult@bdeck/all_bdeck_common/canonical/LGEM/storm_ci_high\endcsname{0.22}
\expandafter\def\csname mcresult@bdeck/all_bdeck_common/canonical/LGEM/storm_ci_low\endcsname{-0.94}
\expandafter\def\csname mcresult@bdeck/all_bdeck_common/canonical/OFCI/TD_mae\endcsname{9.09}
\expandafter\def\csname mcresult@bdeck/all_bdeck_common/canonical/OFCI/TD_minus_aid_mae\endcsname{0.66}
\expandafter\def\csname mcresult@bdeck/all_bdeck_common/canonical/OFCI/TID_mae\endcsname{9.12}
\expandafter\def\csname mcresult@bdeck/all_bdeck_common/canonical/OFCI/T_mae\endcsname{10.16}
\expandafter\def\csname mcresult@bdeck/all_bdeck_common/canonical/OFCI/aid_mae\endcsname{8.43}
\expandafter\def\csname mcresult@bdeck/all_bdeck_common/canonical/OFCI/changed_future_truth_rows\endcsname{41}
\expandafter\def\csname mcresult@bdeck/all_bdeck_common/canonical/OFCI/changed_initial_anchor_rows\endcsname{32}
\expandafter\def\csname mcresult@bdeck/all_bdeck_common/canonical/OFCI/n_obs\endcsname{1{,}017}
\expandafter\def\csname mcresult@bdeck/all_bdeck_common/canonical/OFCI/n_storms\endcsname{64}
\expandafter\def\csname mcresult@bdeck/all_bdeck_common/canonical/OFCI/storm_ci_high\endcsname{1.12}
\expandafter\def\csname mcresult@bdeck/all_bdeck_common/canonical/OFCI/storm_ci_low\endcsname{0.17}
\expandafter\def\csname mcresult@bdeck/all_bdeck_common/canonical/OFCL/TD_mae\endcsname{8.97}
\expandafter\def\csname mcresult@bdeck/all_bdeck_common/canonical/OFCL/TD_minus_aid_mae\endcsname{1.28}
\expandafter\def\csname mcresult@bdeck/all_bdeck_common/canonical/OFCL/TID_mae\endcsname{8.99}
\expandafter\def\csname mcresult@bdeck/all_bdeck_common/canonical/OFCL/T_mae\endcsname{10.07}
\expandafter\def\csname mcresult@bdeck/all_bdeck_common/canonical/OFCL/aid_mae\endcsname{7.69}
\expandafter\def\csname mcresult@bdeck/all_bdeck_common/canonical/OFCL/changed_future_truth_rows\endcsname{41}
\expandafter\def\csname mcresult@bdeck/all_bdeck_common/canonical/OFCL/changed_initial_anchor_rows\endcsname{33}
\expandafter\def\csname mcresult@bdeck/all_bdeck_common/canonical/OFCL/n_obs\endcsname{1{,}077}
\expandafter\def\csname mcresult@bdeck/all_bdeck_common/canonical/OFCL/n_storms\endcsname{64}
\expandafter\def\csname mcresult@bdeck/all_bdeck_common/canonical/OFCL/storm_ci_high\endcsname{1.82}
\expandafter\def\csname mcresult@bdeck/all_bdeck_common/canonical/OFCL/storm_ci_low\endcsname{0.70}
\expandafter\def\csname mcresult@bdeck/diagnostics/available_future_unique_initializations\endcsname{1{,}124}
\expandafter\def\csname mcresult@bdeck/diagnostics/changed_future_unique_initializations\endcsname{41}
\expandafter\def\csname mcresult@bdeck/diagnostics/dora_verified_distinct_timestamps\endcsname{67}
\expandafter\def\csname mcresult@bdeck/diagnostics/missing_future_unique_initializations\endcsname{1}
\expandafter\def\csname mcresult@bdeck/diagnostics/unique_initializations\endcsname{1{,}125}
\expandafter\def\csname mcresult@bdeck/nhc_tc_endpoints/bdeck/DSHP/TD_mae\endcsname{8.96}
\expandafter\def\csname mcresult@bdeck/nhc_tc_endpoints/bdeck/DSHP/TD_minus_aid_mae\endcsname{-0.31}
\expandafter\def\csname mcresult@bdeck/nhc_tc_endpoints/bdeck/DSHP/TID_mae\endcsname{9.00}
\expandafter\def\csname mcresult@bdeck/nhc_tc_endpoints/bdeck/DSHP/T_mae\endcsname{9.93}
\expandafter\def\csname mcresult@bdeck/nhc_tc_endpoints/bdeck/DSHP/aid_mae\endcsname{9.28}
\expandafter\def\csname mcresult@bdeck/nhc_tc_endpoints/bdeck/DSHP/changed_future_truth_rows\endcsname{40}
\expandafter\def\csname mcresult@bdeck/nhc_tc_endpoints/bdeck/DSHP/changed_initial_anchor_rows\endcsname{33}
\expandafter\def\csname mcresult@bdeck/nhc_tc_endpoints/bdeck/DSHP/n_obs\endcsname{1{,}116}
\expandafter\def\csname mcresult@bdeck/nhc_tc_endpoints/bdeck/DSHP/n_storms\endcsname{66}
\expandafter\def\csname mcresult@bdeck/nhc_tc_endpoints/bdeck/DSHP/storm_ci_high\endcsname{0.18}
\expandafter\def\csname mcresult@bdeck/nhc_tc_endpoints/bdeck/DSHP/storm_ci_low\endcsname{-0.85}
\expandafter\def\csname mcresult@bdeck/nhc_tc_endpoints/bdeck/HF3A/TD_mae\endcsname{8.28}
\expandafter\def\csname mcresult@bdeck/nhc_tc_endpoints/bdeck/HF3A/TD_minus_aid_mae\endcsname{-0.57}
\expandafter\def\csname mcresult@bdeck/nhc_tc_endpoints/bdeck/HF3A/TID_mae\endcsname{8.26}
\expandafter\def\csname mcresult@bdeck/nhc_tc_endpoints/bdeck/HF3A/T_mae\endcsname{11.77}
\expandafter\def\csname mcresult@bdeck/nhc_tc_endpoints/bdeck/HF3A/aid_mae\endcsname{8.85}
\expandafter\def\csname mcresult@bdeck/nhc_tc_endpoints/bdeck/HF3A/changed_future_truth_rows\endcsname{0}
\expandafter\def\csname mcresult@bdeck/nhc_tc_endpoints/bdeck/HF3A/changed_initial_anchor_rows\endcsname{0}
\expandafter\def\csname mcresult@bdeck/nhc_tc_endpoints/bdeck/HF3A/n_obs\endcsname{158}
\expandafter\def\csname mcresult@bdeck/nhc_tc_endpoints/bdeck/HF3A/n_storms\endcsname{13}
\expandafter\def\csname mcresult@bdeck/nhc_tc_endpoints/bdeck/HF3A/storm_ci_high\endcsname{2.35}
\expandafter\def\csname mcresult@bdeck/nhc_tc_endpoints/bdeck/HF3A/storm_ci_low\endcsname{-3.35}
\expandafter\def\csname mcresult@bdeck/nhc_tc_endpoints/bdeck/HFAI/TD_mae\endcsname{9.12}
\expandafter\def\csname mcresult@bdeck/nhc_tc_endpoints/bdeck/HFAI/TD_minus_aid_mae\endcsname{-0.66}
\expandafter\def\csname mcresult@bdeck/nhc_tc_endpoints/bdeck/HFAI/TID_mae\endcsname{9.12}
\expandafter\def\csname mcresult@bdeck/nhc_tc_endpoints/bdeck/HFAI/T_mae\endcsname{9.77}
\expandafter\def\csname mcresult@bdeck/nhc_tc_endpoints/bdeck/HFAI/aid_mae\endcsname{9.78}
\expandafter\def\csname mcresult@bdeck/nhc_tc_endpoints/bdeck/HFAI/changed_future_truth_rows\endcsname{40}
\expandafter\def\csname mcresult@bdeck/nhc_tc_endpoints/bdeck/HFAI/changed_initial_anchor_rows\endcsname{33}
\expandafter\def\csname mcresult@bdeck/nhc_tc_endpoints/bdeck/HFAI/n_obs\endcsname{636}
\expandafter\def\csname mcresult@bdeck/nhc_tc_endpoints/bdeck/HFAI/n_storms\endcsname{35}
\expandafter\def\csname mcresult@bdeck/nhc_tc_endpoints/bdeck/HFAI/storm_ci_high\endcsname{0.49}
\expandafter\def\csname mcresult@bdeck/nhc_tc_endpoints/bdeck/HFAI/storm_ci_low\endcsname{-1.84}
\expandafter\def\csname mcresult@bdeck/nhc_tc_endpoints/bdeck/HFBI/TD_mae\endcsname{9.12}
\expandafter\def\csname mcresult@bdeck/nhc_tc_endpoints/bdeck/HFBI/TD_minus_aid_mae\endcsname{-0.55}
\expandafter\def\csname mcresult@bdeck/nhc_tc_endpoints/bdeck/HFBI/TID_mae\endcsname{9.12}
\expandafter\def\csname mcresult@bdeck/nhc_tc_endpoints/bdeck/HFBI/T_mae\endcsname{9.77}
\expandafter\def\csname mcresult@bdeck/nhc_tc_endpoints/bdeck/HFBI/aid_mae\endcsname{9.67}
\expandafter\def\csname mcresult@bdeck/nhc_tc_endpoints/bdeck/HFBI/changed_future_truth_rows\endcsname{40}
\expandafter\def\csname mcresult@bdeck/nhc_tc_endpoints/bdeck/HFBI/changed_initial_anchor_rows\endcsname{33}
\expandafter\def\csname mcresult@bdeck/nhc_tc_endpoints/bdeck/HFBI/n_obs\endcsname{636}
\expandafter\def\csname mcresult@bdeck/nhc_tc_endpoints/bdeck/HFBI/n_storms\endcsname{35}
\expandafter\def\csname mcresult@bdeck/nhc_tc_endpoints/bdeck/HFBI/storm_ci_high\endcsname{0.34}
\expandafter\def\csname mcresult@bdeck/nhc_tc_endpoints/bdeck/HFBI/storm_ci_low\endcsname{-1.43}
\expandafter\def\csname mcresult@bdeck/nhc_tc_endpoints/bdeck/HFSA/TD_mae\endcsname{9.12}
\expandafter\def\csname mcresult@bdeck/nhc_tc_endpoints/bdeck/HFSA/TD_minus_aid_mae\endcsname{-1.74}
\expandafter\def\csname mcresult@bdeck/nhc_tc_endpoints/bdeck/HFSA/TID_mae\endcsname{9.12}
\expandafter\def\csname mcresult@bdeck/nhc_tc_endpoints/bdeck/HFSA/T_mae\endcsname{9.76}
\expandafter\def\csname mcresult@bdeck/nhc_tc_endpoints/bdeck/HFSA/aid_mae\endcsname{10.86}
\expandafter\def\csname mcresult@bdeck/nhc_tc_endpoints/bdeck/HFSA/changed_future_truth_rows\endcsname{40}
\expandafter\def\csname mcresult@bdeck/nhc_tc_endpoints/bdeck/HFSA/changed_initial_anchor_rows\endcsname{33}
\expandafter\def\csname mcresult@bdeck/nhc_tc_endpoints/bdeck/HFSA/n_obs\endcsname{641}
\expandafter\def\csname mcresult@bdeck/nhc_tc_endpoints/bdeck/HFSA/n_storms\endcsname{35}
\expandafter\def\csname mcresult@bdeck/nhc_tc_endpoints/bdeck/HFSA/storm_ci_high\endcsname{-0.78}
\expandafter\def\csname mcresult@bdeck/nhc_tc_endpoints/bdeck/HFSA/storm_ci_low\endcsname{-2.70}
\expandafter\def\csname mcresult@bdeck/nhc_tc_endpoints/bdeck/HFSB/TD_mae\endcsname{9.12}
\expandafter\def\csname mcresult@bdeck/nhc_tc_endpoints/bdeck/HFSB/TD_minus_aid_mae\endcsname{-1.98}
\expandafter\def\csname mcresult@bdeck/nhc_tc_endpoints/bdeck/HFSB/TID_mae\endcsname{9.12}
\expandafter\def\csname mcresult@bdeck/nhc_tc_endpoints/bdeck/HFSB/T_mae\endcsname{9.76}
\expandafter\def\csname mcresult@bdeck/nhc_tc_endpoints/bdeck/HFSB/aid_mae\endcsname{11.10}
\expandafter\def\csname mcresult@bdeck/nhc_tc_endpoints/bdeck/HFSB/changed_future_truth_rows\endcsname{40}
\expandafter\def\csname mcresult@bdeck/nhc_tc_endpoints/bdeck/HFSB/changed_initial_anchor_rows\endcsname{33}
\expandafter\def\csname mcresult@bdeck/nhc_tc_endpoints/bdeck/HFSB/n_obs\endcsname{641}
\expandafter\def\csname mcresult@bdeck/nhc_tc_endpoints/bdeck/HFSB/n_storms\endcsname{35}
\expandafter\def\csname mcresult@bdeck/nhc_tc_endpoints/bdeck/HFSB/storm_ci_high\endcsname{-1.21}
\expandafter\def\csname mcresult@bdeck/nhc_tc_endpoints/bdeck/HFSB/storm_ci_low\endcsname{-2.68}
\expandafter\def\csname mcresult@bdeck/nhc_tc_endpoints/bdeck/HWFI/TD_mae\endcsname{8.97}
\expandafter\def\csname mcresult@bdeck/nhc_tc_endpoints/bdeck/HWFI/TD_minus_aid_mae\endcsname{0.05}
\expandafter\def\csname mcresult@bdeck/nhc_tc_endpoints/bdeck/HWFI/TID_mae\endcsname{8.99}
\expandafter\def\csname mcresult@bdeck/nhc_tc_endpoints/bdeck/HWFI/T_mae\endcsname{9.99}
\expandafter\def\csname mcresult@bdeck/nhc_tc_endpoints/bdeck/HWFI/aid_mae\endcsname{8.92}
\expandafter\def\csname mcresult@bdeck/nhc_tc_endpoints/bdeck/HWFI/changed_future_truth_rows\endcsname{39}
\expandafter\def\csname mcresult@bdeck/nhc_tc_endpoints/bdeck/HWFI/changed_initial_anchor_rows\endcsname{32}
\expandafter\def\csname mcresult@bdeck/nhc_tc_endpoints/bdeck/HWFI/n_obs\endcsname{1{,}061}
\expandafter\def\csname mcresult@bdeck/nhc_tc_endpoints/bdeck/HWFI/n_storms\endcsname{64}
\expandafter\def\csname mcresult@bdeck/nhc_tc_endpoints/bdeck/HWFI/storm_ci_high\endcsname{0.72}
\expandafter\def\csname mcresult@bdeck/nhc_tc_endpoints/bdeck/HWFI/storm_ci_low\endcsname{-0.64}
\expandafter\def\csname mcresult@bdeck/nhc_tc_endpoints/bdeck/HWRF/TD_mae\endcsname{9.03}
\expandafter\def\csname mcresult@bdeck/nhc_tc_endpoints/bdeck/HWRF/TD_minus_aid_mae\endcsname{-0.74}
\expandafter\def\csname mcresult@bdeck/nhc_tc_endpoints/bdeck/HWRF/TID_mae\endcsname{9.07}
\expandafter\def\csname mcresult@bdeck/nhc_tc_endpoints/bdeck/HWRF/T_mae\endcsname{10.00}
\expandafter\def\csname mcresult@bdeck/nhc_tc_endpoints/bdeck/HWRF/aid_mae\endcsname{9.77}
\expandafter\def\csname mcresult@bdeck/nhc_tc_endpoints/bdeck/HWRF/changed_future_truth_rows\endcsname{40}
\expandafter\def\csname mcresult@bdeck/nhc_tc_endpoints/bdeck/HWRF/changed_initial_anchor_rows\endcsname{33}
\expandafter\def\csname mcresult@bdeck/nhc_tc_endpoints/bdeck/HWRF/n_obs\endcsname{1{,}098}
\expandafter\def\csname mcresult@bdeck/nhc_tc_endpoints/bdeck/HWRF/n_storms\endcsname{64}
\expandafter\def\csname mcresult@bdeck/nhc_tc_endpoints/bdeck/HWRF/storm_ci_high\endcsname{0.18}
\expandafter\def\csname mcresult@bdeck/nhc_tc_endpoints/bdeck/HWRF/storm_ci_low\endcsname{-1.58}
\expandafter\def\csname mcresult@bdeck/nhc_tc_endpoints/bdeck/LGEM/TD_mae\endcsname{8.96}
\expandafter\def\csname mcresult@bdeck/nhc_tc_endpoints/bdeck/LGEM/TD_minus_aid_mae\endcsname{-0.34}
\expandafter\def\csname mcresult@bdeck/nhc_tc_endpoints/bdeck/LGEM/TID_mae\endcsname{9.00}
\expandafter\def\csname mcresult@bdeck/nhc_tc_endpoints/bdeck/LGEM/T_mae\endcsname{9.93}
\expandafter\def\csname mcresult@bdeck/nhc_tc_endpoints/bdeck/LGEM/aid_mae\endcsname{9.30}
\expandafter\def\csname mcresult@bdeck/nhc_tc_endpoints/bdeck/LGEM/changed_future_truth_rows\endcsname{40}
\expandafter\def\csname mcresult@bdeck/nhc_tc_endpoints/bdeck/LGEM/changed_initial_anchor_rows\endcsname{33}
\expandafter\def\csname mcresult@bdeck/nhc_tc_endpoints/bdeck/LGEM/n_obs\endcsname{1{,}116}
\expandafter\def\csname mcresult@bdeck/nhc_tc_endpoints/bdeck/LGEM/n_storms\endcsname{66}
\expandafter\def\csname mcresult@bdeck/nhc_tc_endpoints/bdeck/LGEM/storm_ci_high\endcsname{0.22}
\expandafter\def\csname mcresult@bdeck/nhc_tc_endpoints/bdeck/LGEM/storm_ci_low\endcsname{-0.95}
\expandafter\def\csname mcresult@bdeck/nhc_tc_endpoints/bdeck/OFCI/TD_mae\endcsname{9.16}
\expandafter\def\csname mcresult@bdeck/nhc_tc_endpoints/bdeck/OFCI/TD_minus_aid_mae\endcsname{0.68}
\expandafter\def\csname mcresult@bdeck/nhc_tc_endpoints/bdeck/OFCI/TID_mae\endcsname{9.20}
\expandafter\def\csname mcresult@bdeck/nhc_tc_endpoints/bdeck/OFCI/T_mae\endcsname{10.18}
\expandafter\def\csname mcresult@bdeck/nhc_tc_endpoints/bdeck/OFCI/aid_mae\endcsname{8.48}
\expandafter\def\csname mcresult@bdeck/nhc_tc_endpoints/bdeck/OFCI/changed_future_truth_rows\endcsname{40}
\expandafter\def\csname mcresult@bdeck/nhc_tc_endpoints/bdeck/OFCI/changed_initial_anchor_rows\endcsname{32}
\expandafter\def\csname mcresult@bdeck/nhc_tc_endpoints/bdeck/OFCI/n_obs\endcsname{1{,}012}
\expandafter\def\csname mcresult@bdeck/nhc_tc_endpoints/bdeck/OFCI/n_storms\endcsname{64}
\expandafter\def\csname mcresult@bdeck/nhc_tc_endpoints/bdeck/OFCI/storm_ci_high\endcsname{1.14}
\expandafter\def\csname mcresult@bdeck/nhc_tc_endpoints/bdeck/OFCI/storm_ci_low\endcsname{0.19}
\expandafter\def\csname mcresult@bdeck/nhc_tc_endpoints/bdeck/OFCL/TD_mae\endcsname{9.03}
\expandafter\def\csname mcresult@bdeck/nhc_tc_endpoints/bdeck/OFCL/TD_minus_aid_mae\endcsname{1.33}
\expandafter\def\csname mcresult@bdeck/nhc_tc_endpoints/bdeck/OFCL/TID_mae\endcsname{9.06}
\expandafter\def\csname mcresult@bdeck/nhc_tc_endpoints/bdeck/OFCL/T_mae\endcsname{10.09}
\expandafter\def\csname mcresult@bdeck/nhc_tc_endpoints/bdeck/OFCL/aid_mae\endcsname{7.71}
\expandafter\def\csname mcresult@bdeck/nhc_tc_endpoints/bdeck/OFCL/changed_future_truth_rows\endcsname{40}
\expandafter\def\csname mcresult@bdeck/nhc_tc_endpoints/bdeck/OFCL/changed_initial_anchor_rows\endcsname{33}
\expandafter\def\csname mcresult@bdeck/nhc_tc_endpoints/bdeck/OFCL/n_obs\endcsname{1{,}072}
\expandafter\def\csname mcresult@bdeck/nhc_tc_endpoints/bdeck/OFCL/n_storms\endcsname{64}
\expandafter\def\csname mcresult@bdeck/nhc_tc_endpoints/bdeck/OFCL/storm_ci_high\endcsname{1.87}
\expandafter\def\csname mcresult@bdeck/nhc_tc_endpoints/bdeck/OFCL/storm_ci_low\endcsname{0.77}
\expandafter\def\csname mcresult@bdeck/nhc_tc_endpoints/canonical/DSHP/TD_mae\endcsname{8.91}
\expandafter\def\csname mcresult@bdeck/nhc_tc_endpoints/canonical/DSHP/TD_minus_aid_mae\endcsname{-0.33}
\expandafter\def\csname mcresult@bdeck/nhc_tc_endpoints/canonical/DSHP/TID_mae\endcsname{8.94}
\expandafter\def\csname mcresult@bdeck/nhc_tc_endpoints/canonical/DSHP/T_mae\endcsname{9.93}
\expandafter\def\csname mcresult@bdeck/nhc_tc_endpoints/canonical/DSHP/aid_mae\endcsname{9.25}
\expandafter\def\csname mcresult@bdeck/nhc_tc_endpoints/canonical/DSHP/changed_future_truth_rows\endcsname{40}
\expandafter\def\csname mcresult@bdeck/nhc_tc_endpoints/canonical/DSHP/changed_initial_anchor_rows\endcsname{33}
\expandafter\def\csname mcresult@bdeck/nhc_tc_endpoints/canonical/DSHP/n_obs\endcsname{1{,}116}
\expandafter\def\csname mcresult@bdeck/nhc_tc_endpoints/canonical/DSHP/n_storms\endcsname{66}
\expandafter\def\csname mcresult@bdeck/nhc_tc_endpoints/canonical/DSHP/storm_ci_high\endcsname{0.16}
\expandafter\def\csname mcresult@bdeck/nhc_tc_endpoints/canonical/DSHP/storm_ci_low\endcsname{-0.86}
\expandafter\def\csname mcresult@bdeck/nhc_tc_endpoints/canonical/HF3A/TD_mae\endcsname{8.28}
\expandafter\def\csname mcresult@bdeck/nhc_tc_endpoints/canonical/HF3A/TD_minus_aid_mae\endcsname{-0.57}
\expandafter\def\csname mcresult@bdeck/nhc_tc_endpoints/canonical/HF3A/TID_mae\endcsname{8.26}
\expandafter\def\csname mcresult@bdeck/nhc_tc_endpoints/canonical/HF3A/T_mae\endcsname{11.77}
\expandafter\def\csname mcresult@bdeck/nhc_tc_endpoints/canonical/HF3A/aid_mae\endcsname{8.85}
\expandafter\def\csname mcresult@bdeck/nhc_tc_endpoints/canonical/HF3A/changed_future_truth_rows\endcsname{0}
\expandafter\def\csname mcresult@bdeck/nhc_tc_endpoints/canonical/HF3A/changed_initial_anchor_rows\endcsname{0}
\expandafter\def\csname mcresult@bdeck/nhc_tc_endpoints/canonical/HF3A/n_obs\endcsname{158}
\expandafter\def\csname mcresult@bdeck/nhc_tc_endpoints/canonical/HF3A/n_storms\endcsname{13}
\expandafter\def\csname mcresult@bdeck/nhc_tc_endpoints/canonical/HF3A/storm_ci_high\endcsname{2.35}
\expandafter\def\csname mcresult@bdeck/nhc_tc_endpoints/canonical/HF3A/storm_ci_low\endcsname{-3.35}
\expandafter\def\csname mcresult@bdeck/nhc_tc_endpoints/canonical/HFAI/TD_mae\endcsname{9.03}
\expandafter\def\csname mcresult@bdeck/nhc_tc_endpoints/canonical/HFAI/TD_minus_aid_mae\endcsname{-0.62}
\expandafter\def\csname mcresult@bdeck/nhc_tc_endpoints/canonical/HFAI/TID_mae\endcsname{9.03}
\expandafter\def\csname mcresult@bdeck/nhc_tc_endpoints/canonical/HFAI/T_mae\endcsname{9.77}
\expandafter\def\csname mcresult@bdeck/nhc_tc_endpoints/canonical/HFAI/aid_mae\endcsname{9.66}
\expandafter\def\csname mcresult@bdeck/nhc_tc_endpoints/canonical/HFAI/changed_future_truth_rows\endcsname{40}
\expandafter\def\csname mcresult@bdeck/nhc_tc_endpoints/canonical/HFAI/changed_initial_anchor_rows\endcsname{33}
\expandafter\def\csname mcresult@bdeck/nhc_tc_endpoints/canonical/HFAI/n_obs\endcsname{636}
\expandafter\def\csname mcresult@bdeck/nhc_tc_endpoints/canonical/HFAI/n_storms\endcsname{35}
\expandafter\def\csname mcresult@bdeck/nhc_tc_endpoints/canonical/HFAI/storm_ci_high\endcsname{0.54}
\expandafter\def\csname mcresult@bdeck/nhc_tc_endpoints/canonical/HFAI/storm_ci_low\endcsname{-1.82}
\expandafter\def\csname mcresult@bdeck/nhc_tc_endpoints/canonical/HFBI/TD_mae\endcsname{9.03}
\expandafter\def\csname mcresult@bdeck/nhc_tc_endpoints/canonical/HFBI/TD_minus_aid_mae\endcsname{-0.54}
\expandafter\def\csname mcresult@bdeck/nhc_tc_endpoints/canonical/HFBI/TID_mae\endcsname{9.03}
\expandafter\def\csname mcresult@bdeck/nhc_tc_endpoints/canonical/HFBI/T_mae\endcsname{9.77}
\expandafter\def\csname mcresult@bdeck/nhc_tc_endpoints/canonical/HFBI/aid_mae\endcsname{9.57}
\expandafter\def\csname mcresult@bdeck/nhc_tc_endpoints/canonical/HFBI/changed_future_truth_rows\endcsname{40}
\expandafter\def\csname mcresult@bdeck/nhc_tc_endpoints/canonical/HFBI/changed_initial_anchor_rows\endcsname{33}
\expandafter\def\csname mcresult@bdeck/nhc_tc_endpoints/canonical/HFBI/n_obs\endcsname{636}
\expandafter\def\csname mcresult@bdeck/nhc_tc_endpoints/canonical/HFBI/n_storms\endcsname{35}
\expandafter\def\csname mcresult@bdeck/nhc_tc_endpoints/canonical/HFBI/storm_ci_high\endcsname{0.36}
\expandafter\def\csname mcresult@bdeck/nhc_tc_endpoints/canonical/HFBI/storm_ci_low\endcsname{-1.42}
\expandafter\def\csname mcresult@bdeck/nhc_tc_endpoints/canonical/HFSA/TD_mae\endcsname{9.03}
\expandafter\def\csname mcresult@bdeck/nhc_tc_endpoints/canonical/HFSA/TD_minus_aid_mae\endcsname{-1.69}
\expandafter\def\csname mcresult@bdeck/nhc_tc_endpoints/canonical/HFSA/TID_mae\endcsname{9.02}
\expandafter\def\csname mcresult@bdeck/nhc_tc_endpoints/canonical/HFSA/T_mae\endcsname{9.76}
\expandafter\def\csname mcresult@bdeck/nhc_tc_endpoints/canonical/HFSA/aid_mae\endcsname{10.72}
\expandafter\def\csname mcresult@bdeck/nhc_tc_endpoints/canonical/HFSA/changed_future_truth_rows\endcsname{40}
\expandafter\def\csname mcresult@bdeck/nhc_tc_endpoints/canonical/HFSA/changed_initial_anchor_rows\endcsname{33}
\expandafter\def\csname mcresult@bdeck/nhc_tc_endpoints/canonical/HFSA/n_obs\endcsname{641}
\expandafter\def\csname mcresult@bdeck/nhc_tc_endpoints/canonical/HFSA/n_storms\endcsname{35}
\expandafter\def\csname mcresult@bdeck/nhc_tc_endpoints/canonical/HFSA/storm_ci_high\endcsname{-0.71}
\expandafter\def\csname mcresult@bdeck/nhc_tc_endpoints/canonical/HFSA/storm_ci_low\endcsname{-2.68}
\expandafter\def\csname mcresult@bdeck/nhc_tc_endpoints/canonical/HFSB/TD_mae\endcsname{9.03}
\expandafter\def\csname mcresult@bdeck/nhc_tc_endpoints/canonical/HFSB/TD_minus_aid_mae\endcsname{-2.01}
\expandafter\def\csname mcresult@bdeck/nhc_tc_endpoints/canonical/HFSB/TID_mae\endcsname{9.02}
\expandafter\def\csname mcresult@bdeck/nhc_tc_endpoints/canonical/HFSB/T_mae\endcsname{9.76}
\expandafter\def\csname mcresult@bdeck/nhc_tc_endpoints/canonical/HFSB/aid_mae\endcsname{11.04}
\expandafter\def\csname mcresult@bdeck/nhc_tc_endpoints/canonical/HFSB/changed_future_truth_rows\endcsname{40}
\expandafter\def\csname mcresult@bdeck/nhc_tc_endpoints/canonical/HFSB/changed_initial_anchor_rows\endcsname{33}
\expandafter\def\csname mcresult@bdeck/nhc_tc_endpoints/canonical/HFSB/n_obs\endcsname{641}
\expandafter\def\csname mcresult@bdeck/nhc_tc_endpoints/canonical/HFSB/n_storms\endcsname{35}
\expandafter\def\csname mcresult@bdeck/nhc_tc_endpoints/canonical/HFSB/storm_ci_high\endcsname{-1.24}
\expandafter\def\csname mcresult@bdeck/nhc_tc_endpoints/canonical/HFSB/storm_ci_low\endcsname{-2.71}
\expandafter\def\csname mcresult@bdeck/nhc_tc_endpoints/canonical/HWFI/TD_mae\endcsname{8.92}
\expandafter\def\csname mcresult@bdeck/nhc_tc_endpoints/canonical/HWFI/TD_minus_aid_mae\endcsname{0.06}
\expandafter\def\csname mcresult@bdeck/nhc_tc_endpoints/canonical/HWFI/TID_mae\endcsname{8.93}
\expandafter\def\csname mcresult@bdeck/nhc_tc_endpoints/canonical/HWFI/T_mae\endcsname{9.99}
\expandafter\def\csname mcresult@bdeck/nhc_tc_endpoints/canonical/HWFI/aid_mae\endcsname{8.86}
\expandafter\def\csname mcresult@bdeck/nhc_tc_endpoints/canonical/HWFI/changed_future_truth_rows\endcsname{39}
\expandafter\def\csname mcresult@bdeck/nhc_tc_endpoints/canonical/HWFI/changed_initial_anchor_rows\endcsname{32}
\expandafter\def\csname mcresult@bdeck/nhc_tc_endpoints/canonical/HWFI/n_obs\endcsname{1{,}061}
\expandafter\def\csname mcresult@bdeck/nhc_tc_endpoints/canonical/HWFI/n_storms\endcsname{64}
\expandafter\def\csname mcresult@bdeck/nhc_tc_endpoints/canonical/HWFI/storm_ci_high\endcsname{0.73}
\expandafter\def\csname mcresult@bdeck/nhc_tc_endpoints/canonical/HWFI/storm_ci_low\endcsname{-0.63}
\expandafter\def\csname mcresult@bdeck/nhc_tc_endpoints/canonical/HWRF/TD_mae\endcsname{8.98}
\expandafter\def\csname mcresult@bdeck/nhc_tc_endpoints/canonical/HWRF/TD_minus_aid_mae\endcsname{-0.72}
\expandafter\def\csname mcresult@bdeck/nhc_tc_endpoints/canonical/HWRF/TID_mae\endcsname{9.01}
\expandafter\def\csname mcresult@bdeck/nhc_tc_endpoints/canonical/HWRF/T_mae\endcsname{10.00}
\expandafter\def\csname mcresult@bdeck/nhc_tc_endpoints/canonical/HWRF/aid_mae\endcsname{9.70}
\expandafter\def\csname mcresult@bdeck/nhc_tc_endpoints/canonical/HWRF/changed_future_truth_rows\endcsname{40}
\expandafter\def\csname mcresult@bdeck/nhc_tc_endpoints/canonical/HWRF/changed_initial_anchor_rows\endcsname{33}
\expandafter\def\csname mcresult@bdeck/nhc_tc_endpoints/canonical/HWRF/n_obs\endcsname{1{,}098}
\expandafter\def\csname mcresult@bdeck/nhc_tc_endpoints/canonical/HWRF/n_storms\endcsname{64}
\expandafter\def\csname mcresult@bdeck/nhc_tc_endpoints/canonical/HWRF/storm_ci_high\endcsname{0.20}
\expandafter\def\csname mcresult@bdeck/nhc_tc_endpoints/canonical/HWRF/storm_ci_low\endcsname{-1.58}
\expandafter\def\csname mcresult@bdeck/nhc_tc_endpoints/canonical/LGEM/TD_mae\endcsname{8.91}
\expandafter\def\csname mcresult@bdeck/nhc_tc_endpoints/canonical/LGEM/TD_minus_aid_mae\endcsname{-0.35}
\expandafter\def\csname mcresult@bdeck/nhc_tc_endpoints/canonical/LGEM/TID_mae\endcsname{8.94}
\expandafter\def\csname mcresult@bdeck/nhc_tc_endpoints/canonical/LGEM/T_mae\endcsname{9.93}
\expandafter\def\csname mcresult@bdeck/nhc_tc_endpoints/canonical/LGEM/aid_mae\endcsname{9.27}
\expandafter\def\csname mcresult@bdeck/nhc_tc_endpoints/canonical/LGEM/changed_future_truth_rows\endcsname{40}
\expandafter\def\csname mcresult@bdeck/nhc_tc_endpoints/canonical/LGEM/changed_initial_anchor_rows\endcsname{33}
\expandafter\def\csname mcresult@bdeck/nhc_tc_endpoints/canonical/LGEM/n_obs\endcsname{1{,}116}
\expandafter\def\csname mcresult@bdeck/nhc_tc_endpoints/canonical/LGEM/n_storms\endcsname{66}
\expandafter\def\csname mcresult@bdeck/nhc_tc_endpoints/canonical/LGEM/storm_ci_high\endcsname{0.21}
\expandafter\def\csname mcresult@bdeck/nhc_tc_endpoints/canonical/LGEM/storm_ci_low\endcsname{-0.96}
\expandafter\def\csname mcresult@bdeck/nhc_tc_endpoints/canonical/OFCI/TD_mae\endcsname{9.10}
\expandafter\def\csname mcresult@bdeck/nhc_tc_endpoints/canonical/OFCI/TD_minus_aid_mae\endcsname{0.64}
\expandafter\def\csname mcresult@bdeck/nhc_tc_endpoints/canonical/OFCI/TID_mae\endcsname{9.14}
\expandafter\def\csname mcresult@bdeck/nhc_tc_endpoints/canonical/OFCI/T_mae\endcsname{10.18}
\expandafter\def\csname mcresult@bdeck/nhc_tc_endpoints/canonical/OFCI/aid_mae\endcsname{8.46}
\expandafter\def\csname mcresult@bdeck/nhc_tc_endpoints/canonical/OFCI/changed_future_truth_rows\endcsname{40}
\expandafter\def\csname mcresult@bdeck/nhc_tc_endpoints/canonical/OFCI/changed_initial_anchor_rows\endcsname{32}
\expandafter\def\csname mcresult@bdeck/nhc_tc_endpoints/canonical/OFCI/n_obs\endcsname{1{,}012}
\expandafter\def\csname mcresult@bdeck/nhc_tc_endpoints/canonical/OFCI/n_storms\endcsname{64}
\expandafter\def\csname mcresult@bdeck/nhc_tc_endpoints/canonical/OFCI/storm_ci_high\endcsname{1.10}
\expandafter\def\csname mcresult@bdeck/nhc_tc_endpoints/canonical/OFCI/storm_ci_low\endcsname{0.15}
\expandafter\def\csname mcresult@bdeck/nhc_tc_endpoints/canonical/OFCL/TD_mae\endcsname{8.98}
\expandafter\def\csname mcresult@bdeck/nhc_tc_endpoints/canonical/OFCL/TD_minus_aid_mae\endcsname{1.26}
\expandafter\def\csname mcresult@bdeck/nhc_tc_endpoints/canonical/OFCL/TID_mae\endcsname{9.00}
\expandafter\def\csname mcresult@bdeck/nhc_tc_endpoints/canonical/OFCL/T_mae\endcsname{10.09}
\expandafter\def\csname mcresult@bdeck/nhc_tc_endpoints/canonical/OFCL/aid_mae\endcsname{7.72}
\expandafter\def\csname mcresult@bdeck/nhc_tc_endpoints/canonical/OFCL/changed_future_truth_rows\endcsname{40}
\expandafter\def\csname mcresult@bdeck/nhc_tc_endpoints/canonical/OFCL/changed_initial_anchor_rows\endcsname{33}
\expandafter\def\csname mcresult@bdeck/nhc_tc_endpoints/canonical/OFCL/n_obs\endcsname{1{,}072}
\expandafter\def\csname mcresult@bdeck/nhc_tc_endpoints/canonical/OFCL/n_storms\endcsname{64}
\expandafter\def\csname mcresult@bdeck/nhc_tc_endpoints/canonical/OFCL/storm_ci_high\endcsname{1.81}
\expandafter\def\csname mcresult@bdeck/nhc_tc_endpoints/canonical/OFCL/storm_ci_low\endcsname{0.69}
\expandafter\def\csname mcresult@case/AGATHA/current_vmax\endcsname{95}
\expandafter\def\csname mcresult@case/AGATHA/d_ensemble\endcsname{-10.6}
\expandafter\def\csname mcresult@case/AGATHA/d_seed1\endcsname{-10.8}
\expandafter\def\csname mcresult@case/AGATHA/gain\endcsname{+10.9}
\expandafter\def\csname mcresult@case/AGATHA/gain_rank\endcsname{104}
\expandafter\def\csname mcresult@case/AGATHA/tab_ensemble\endcsname{+2.5}
\expandafter\def\csname mcresult@case/AGATHA/td_ensemble\endcsname{-8.4}
\expandafter\def\csname mcresult@case/AGATHA/td_seed1\endcsname{-0.5}
\expandafter\def\csname mcresult@case/AGATHA/truth_delta\endcsname{-25}
\expandafter\def\csname mcresult@case/CALVIN/current_vmax\endcsname{65}
\expandafter\def\csname mcresult@case/CALVIN/d_ensemble\endcsname{-20.7}
\expandafter\def\csname mcresult@case/CALVIN/d_seed1\endcsname{-18.6}
\expandafter\def\csname mcresult@case/CALVIN/gain\endcsname{+7.7}
\expandafter\def\csname mcresult@case/CALVIN/gain_rank\endcsname{186}
\expandafter\def\csname mcresult@case/CALVIN/tab_ensemble\endcsname{-10.9}
\expandafter\def\csname mcresult@case/CALVIN/td_ensemble\endcsname{-21.3}
\expandafter\def\csname mcresult@case/CALVIN/td_seed1\endcsname{-21.5}
\expandafter\def\csname mcresult@case/CALVIN/truth_delta\endcsname{-20}
\expandafter\def\csname mcresult@case/DARBY/current_vmax\endcsname{65}
\expandafter\def\csname mcresult@case/DARBY/d_ensemble\endcsname{-23.6}
\expandafter\def\csname mcresult@case/DARBY/d_seed1\endcsname{-21.8}
\expandafter\def\csname mcresult@case/DARBY/gain\endcsname{-11.3}
\expandafter\def\csname mcresult@case/DARBY/gain_rank\endcsname{1{,}093}
\expandafter\def\csname mcresult@case/DARBY/tab_ensemble\endcsname{-26.5}
\expandafter\def\csname mcresult@case/DARBY/td_ensemble\endcsname{-37.8}
\expandafter\def\csname mcresult@case/DARBY/td_seed1\endcsname{-42.5}
\expandafter\def\csname mcresult@case/DARBY/truth_delta\endcsname{-25}
\expandafter\def\csname mcresult@case/DORA/current_vmax\endcsname{80}
\expandafter\def\csname mcresult@case/DORA/d_ensemble\endcsname{+39.9}
\expandafter\def\csname mcresult@case/DORA/d_seed1\endcsname{+31.9}
\expandafter\def\csname mcresult@case/DORA/gain\endcsname{+38.2}
\expandafter\def\csname mcresult@case/DORA/gain_rank\endcsname{3}
\expandafter\def\csname mcresult@case/DORA/tab_ensemble\endcsname{-5.0}
\expandafter\def\csname mcresult@case/DORA/td_ensemble\endcsname{+33.2}
\expandafter\def\csname mcresult@case/DORA/td_seed1\endcsname{+24.2}
\expandafter\def\csname mcresult@case/DORA/truth_delta\endcsname{+35}
\expandafter\def\csname mcresult@case/FIONA/current_vmax\endcsname{100}
\expandafter\def\csname mcresult@case/FIONA/d_ensemble\endcsname{+4.0}
\expandafter\def\csname mcresult@case/FIONA/d_seed1\endcsname{+6.5}
\expandafter\def\csname mcresult@case/FIONA/gain\endcsname{+1.7}
\expandafter\def\csname mcresult@case/FIONA/gain_rank\endcsname{505}
\expandafter\def\csname mcresult@case/FIONA/tab_ensemble\endcsname{+2.9}
\expandafter\def\csname mcresult@case/FIONA/td_ensemble\endcsname{+4.5}
\expandafter\def\csname mcresult@case/FIONA/td_seed1\endcsname{+8.4}
\expandafter\def\csname mcresult@case/FIONA/truth_delta\endcsname{+20}
\expandafter\def\csname mcresult@case/IAN/current_vmax\endcsname{40}
\expandafter\def\csname mcresult@case/IAN/d_ensemble\endcsname{+27.6}
\expandafter\def\csname mcresult@case/IAN/d_seed1\endcsname{+26.1}
\expandafter\def\csname mcresult@case/IAN/gain\endcsname{+14.7}
\expandafter\def\csname mcresult@case/IAN/gain_rank\endcsname{62}
\expandafter\def\csname mcresult@case/IAN/tab_ensemble\endcsname{+14.1}
\expandafter\def\csname mcresult@case/IAN/td_ensemble\endcsname{+31.2}
\expandafter\def\csname mcresult@case/IAN/td_seed1\endcsname{+28.1}
\expandafter\def\csname mcresult@case/IAN/truth_delta\endcsname{+30}
\expandafter\def\csname mcresult@case/IDALIA/current_vmax\endcsname{30}
\expandafter\def\csname mcresult@case/IDALIA/d_ensemble\endcsname{+11.2}
\expandafter\def\csname mcresult@case/IDALIA/d_seed1\endcsname{+14.0}
\expandafter\def\csname mcresult@case/IDALIA/gain\endcsname{+4.2}
\expandafter\def\csname mcresult@case/IDALIA/gain_rank\endcsname{334}
\expandafter\def\csname mcresult@case/IDALIA/tab_ensemble\endcsname{+10.1}
\expandafter\def\csname mcresult@case/IDALIA/td_ensemble\endcsname{+14.3}
\expandafter\def\csname mcresult@case/IDALIA/td_seed1\endcsname{+14.2}
\expandafter\def\csname mcresult@case/IDALIA/truth_delta\endcsname{+25}
\expandafter\def\csname mcresult@case/LEE/current_vmax\endcsname{85}
\expandafter\def\csname mcresult@case/LEE/d_ensemble\endcsname{+25.2}
\expandafter\def\csname mcresult@case/LEE/d_seed1\endcsname{+12.9}
\expandafter\def\csname mcresult@case/LEE/gain\endcsname{-2.4}
\expandafter\def\csname mcresult@case/LEE/gain_rank\endcsname{792}
\expandafter\def\csname mcresult@case/LEE/tab_ensemble\endcsname{+27.7}
\expandafter\def\csname mcresult@case/LEE/td_ensemble\endcsname{+25.3}
\expandafter\def\csname mcresult@case/LEE/td_seed1\endcsname{+15.4}
\expandafter\def\csname mcresult@case/LEE/truth_delta\endcsname{+50}
\expandafter\def\csname mcresult@case/MADELINE/current_vmax\endcsname{50}
\expandafter\def\csname mcresult@case/MADELINE/d_ensemble\endcsname{-1.9}
\expandafter\def\csname mcresult@case/MADELINE/d_seed1\endcsname{-1.6}
\expandafter\def\csname mcresult@case/MADELINE/gain\endcsname{-4.4}
\expandafter\def\csname mcresult@case/MADELINE/gain_rank\endcsname{914}
\expandafter\def\csname mcresult@case/MADELINE/tab_ensemble\endcsname{-7.8}
\expandafter\def\csname mcresult@case/MADELINE/td_ensemble\endcsname{-3.4}
\expandafter\def\csname mcresult@case/MADELINE/td_seed1\endcsname{-7.1}
\expandafter\def\csname mcresult@case/MADELINE/truth_delta\endcsname{-15}
\expandafter\def\csname mcresult@case/OTIS/current_vmax\endcsname{55}
\expandafter\def\csname mcresult@case/OTIS/d_ensemble\endcsname{+22.7}
\expandafter\def\csname mcresult@case/OTIS/d_seed1\endcsname{+23.8}
\expandafter\def\csname mcresult@case/OTIS/gain\endcsname{+19.7}
\expandafter\def\csname mcresult@case/OTIS/gain_rank\endcsname{25}
\expandafter\def\csname mcresult@case/OTIS/tab_ensemble\endcsname{+4.7}
\expandafter\def\csname mcresult@case/OTIS/td_ensemble\endcsname{+24.4}
\expandafter\def\csname mcresult@case/OTIS/td_seed1\endcsname{+24.0}
\expandafter\def\csname mcresult@case/OTIS/truth_delta\endcsname{+90}
\expandafter\def\csname mcresult@case/PHILIPPE/current_vmax\endcsname{45}
\expandafter\def\csname mcresult@case/PHILIPPE/d_ensemble\endcsname{+2.9}
\expandafter\def\csname mcresult@case/PHILIPPE/d_seed1\endcsname{+4.8}
\expandafter\def\csname mcresult@case/PHILIPPE/gain\endcsname{-2.2}
\expandafter\def\csname mcresult@case/PHILIPPE/gain_rank\endcsname{778}
\expandafter\def\csname mcresult@case/PHILIPPE/tab_ensemble\endcsname{+0.8}
\expandafter\def\csname mcresult@case/PHILIPPE/td_ensemble\endcsname{+3.0}
\expandafter\def\csname mcresult@case/PHILIPPE/td_seed1\endcsname{-0.9}
\expandafter\def\csname mcresult@case/PHILIPPE/truth_delta\endcsname{+0}
\expandafter\def\csname mcresult@cell/ablation_mae/disc_hybrid_24/MSLP/mae/mean\endcsname{7.26}
\expandafter\def\csname mcresult@cell/ablation_mae/disc_hybrid_24/MSLP/mae/sd\endcsname{0.05}
\expandafter\def\csname mcresult@cell/ablation_mae/disc_hybrid_24/VMAX/mae/mean\endcsname{10.79}
\expandafter\def\csname mcresult@cell/ablation_mae/disc_hybrid_24/VMAX/mae/sd\endcsname{0.07}
\expandafter\def\csname mcresult@cell/ablation_mae/disc_hybrid_48/MSLP/mae/mean\endcsname{12.19}
\expandafter\def\csname mcresult@cell/ablation_mae/disc_hybrid_48/MSLP/mae/sd\endcsname{0.10}
\expandafter\def\csname mcresult@cell/ablation_mae/disc_hybrid_48/VMAX/mae/mean\endcsname{17.04}
\expandafter\def\csname mcresult@cell/ablation_mae/disc_hybrid_48/VMAX/mae/sd\endcsname{0.18}
\expandafter\def\csname mcresult@cell/ablation_mae/disc_hybrid_72/MSLP/mae/mean\endcsname{14.37}
\expandafter\def\csname mcresult@cell/ablation_mae/disc_hybrid_72/MSLP/mae/sd\endcsname{0.22}
\expandafter\def\csname mcresult@cell/ablation_mae/disc_hybrid_72/VMAX/mae/mean\endcsname{19.84}
\expandafter\def\csname mcresult@cell/ablation_mae/disc_hybrid_72/VMAX/mae/sd\endcsname{0.17}
\expandafter\def\csname mcresult@cell/ablation_mae/disc_neural_24/MSLP/mae/mean\endcsname{6.56}
\expandafter\def\csname mcresult@cell/ablation_mae/disc_neural_24/MSLP/mae/sd\endcsname{0.17}
\expandafter\def\csname mcresult@cell/ablation_mae/disc_neural_24/VMAX/mae/mean\endcsname{9.90}
\expandafter\def\csname mcresult@cell/ablation_mae/disc_neural_24/VMAX/mae/sd\endcsname{0.20}
\expandafter\def\csname mcresult@cell/ablation_mae/disc_neural_48/MSLP/mae/mean\endcsname{10.26}
\expandafter\def\csname mcresult@cell/ablation_mae/disc_neural_48/MSLP/mae/sd\endcsname{0.19}
\expandafter\def\csname mcresult@cell/ablation_mae/disc_neural_48/VMAX/mae/mean\endcsname{14.32}
\expandafter\def\csname mcresult@cell/ablation_mae/disc_neural_48/VMAX/mae/sd\endcsname{0.17}
\expandafter\def\csname mcresult@cell/ablation_mae/disc_neural_72/MSLP/mae/mean\endcsname{12.94}
\expandafter\def\csname mcresult@cell/ablation_mae/disc_neural_72/MSLP/mae/sd\endcsname{0.46}
\expandafter\def\csname mcresult@cell/ablation_mae/disc_neural_72/VMAX/mae/mean\endcsname{17.63}
\expandafter\def\csname mcresult@cell/ablation_mae/disc_neural_72/VMAX/mae/sd\endcsname{0.80}
\expandafter\def\csname mcresult@cell/ablation_mae/full_hybrid_24/MSLP/mae/mean\endcsname{6.43}
\expandafter\def\csname mcresult@cell/ablation_mae/full_hybrid_24/MSLP/mae/sd\endcsname{0.18}
\expandafter\def\csname mcresult@cell/ablation_mae/full_hybrid_24/VMAX/mae/mean\endcsname{9.78}
\expandafter\def\csname mcresult@cell/ablation_mae/full_hybrid_24/VMAX/mae/sd\endcsname{0.16}
\expandafter\def\csname mcresult@cell/ablation_mae/full_hybrid_48/MSLP/mae/mean\endcsname{10.82}
\expandafter\def\csname mcresult@cell/ablation_mae/full_hybrid_48/MSLP/mae/sd\endcsname{0.20}
\expandafter\def\csname mcresult@cell/ablation_mae/full_hybrid_48/VMAX/mae/mean\endcsname{15.27}
\expandafter\def\csname mcresult@cell/ablation_mae/full_hybrid_48/VMAX/mae/sd\endcsname{0.16}
\expandafter\def\csname mcresult@cell/ablation_mae/full_hybrid_72/MSLP/mae/mean\endcsname{13.31}
\expandafter\def\csname mcresult@cell/ablation_mae/full_hybrid_72/MSLP/mae/sd\endcsname{0.26}
\expandafter\def\csname mcresult@cell/ablation_mae/full_hybrid_72/VMAX/mae/mean\endcsname{17.99}
\expandafter\def\csname mcresult@cell/ablation_mae/full_hybrid_72/VMAX/mae/sd\endcsname{0.20}
\expandafter\def\csname mcresult@cell/ablation_mae/full_neural_24/MSLP/mae/mean\endcsname{6.23}
\expandafter\def\csname mcresult@cell/ablation_mae/full_neural_24/MSLP/mae/sd\endcsname{0.08}
\expandafter\def\csname mcresult@cell/ablation_mae/full_neural_24/VMAX/mae/mean\endcsname{9.45}
\expandafter\def\csname mcresult@cell/ablation_mae/full_neural_24/VMAX/mae/sd\endcsname{0.13}
\expandafter\def\csname mcresult@cell/ablation_mae/full_neural_48/MSLP/mae/mean\endcsname{9.67}
\expandafter\def\csname mcresult@cell/ablation_mae/full_neural_48/MSLP/mae/sd\endcsname{0.27}
\expandafter\def\csname mcresult@cell/ablation_mae/full_neural_48/VMAX/mae/mean\endcsname{13.65}
\expandafter\def\csname mcresult@cell/ablation_mae/full_neural_48/VMAX/mae/sd\endcsname{0.22}
\expandafter\def\csname mcresult@cell/ablation_mae/full_neural_72/MSLP/mae/mean\endcsname{12.44}
\expandafter\def\csname mcresult@cell/ablation_mae/full_neural_72/MSLP/mae/sd\endcsname{0.49}
\expandafter\def\csname mcresult@cell/ablation_mae/full_neural_72/VMAX/mae/mean\endcsname{17.11}
\expandafter\def\csname mcresult@cell/ablation_mae/full_neural_72/VMAX/mae/sd\endcsname{0.61}
\expandafter\def\csname mcresult@cell/ablation_mae/img_disc_hybrid_24/MSLP/mae/mean\endcsname{7.12}
\expandafter\def\csname mcresult@cell/ablation_mae/img_disc_hybrid_24/MSLP/mae/sd\endcsname{0.07}
\expandafter\def\csname mcresult@cell/ablation_mae/img_disc_hybrid_24/VMAX/mae/mean\endcsname{10.72}
\expandafter\def\csname mcresult@cell/ablation_mae/img_disc_hybrid_24/VMAX/mae/sd\endcsname{0.15}
\expandafter\def\csname mcresult@cell/ablation_mae/img_disc_hybrid_48/MSLP/mae/mean\endcsname{11.70}
\expandafter\def\csname mcresult@cell/ablation_mae/img_disc_hybrid_48/MSLP/mae/sd\endcsname{0.17}
\expandafter\def\csname mcresult@cell/ablation_mae/img_disc_hybrid_48/VMAX/mae/mean\endcsname{16.32}
\expandafter\def\csname mcresult@cell/ablation_mae/img_disc_hybrid_48/VMAX/mae/sd\endcsname{0.20}
\expandafter\def\csname mcresult@cell/ablation_mae/img_disc_hybrid_72/MSLP/mae/mean\endcsname{13.94}
\expandafter\def\csname mcresult@cell/ablation_mae/img_disc_hybrid_72/MSLP/mae/sd\endcsname{0.19}
\expandafter\def\csname mcresult@cell/ablation_mae/img_disc_hybrid_72/VMAX/mae/mean\endcsname{18.94}
\expandafter\def\csname mcresult@cell/ablation_mae/img_disc_hybrid_72/VMAX/mae/sd\endcsname{0.37}
\expandafter\def\csname mcresult@cell/ablation_mae/img_disc_neural_24/MSLP/mae/mean\endcsname{6.45}
\expandafter\def\csname mcresult@cell/ablation_mae/img_disc_neural_24/MSLP/mae/sd\endcsname{0.18}
\expandafter\def\csname mcresult@cell/ablation_mae/img_disc_neural_24/VMAX/mae/mean\endcsname{9.74}
\expandafter\def\csname mcresult@cell/ablation_mae/img_disc_neural_24/VMAX/mae/sd\endcsname{0.27}
\expandafter\def\csname mcresult@cell/ablation_mae/img_disc_neural_48/MSLP/mae/mean\endcsname{10.11}
\expandafter\def\csname mcresult@cell/ablation_mae/img_disc_neural_48/MSLP/mae/sd\endcsname{0.15}
\expandafter\def\csname mcresult@cell/ablation_mae/img_disc_neural_48/VMAX/mae/mean\endcsname{14.38}
\expandafter\def\csname mcresult@cell/ablation_mae/img_disc_neural_48/VMAX/mae/sd\endcsname{0.54}
\expandafter\def\csname mcresult@cell/ablation_mae/img_disc_neural_72/MSLP/mae/mean\endcsname{12.93}
\expandafter\def\csname mcresult@cell/ablation_mae/img_disc_neural_72/MSLP/mae/sd\endcsname{0.59}
\expandafter\def\csname mcresult@cell/ablation_mae/img_disc_neural_72/VMAX/mae/mean\endcsname{18.08}
\expandafter\def\csname mcresult@cell/ablation_mae/img_disc_neural_72/VMAX/mae/sd\endcsname{1.48}
\expandafter\def\csname mcresult@cell/ablation_mae/img_hybrid_24/MSLP/mae/mean\endcsname{8.10}
\expandafter\def\csname mcresult@cell/ablation_mae/img_hybrid_24/MSLP/mae/sd\endcsname{0.06}
\expandafter\def\csname mcresult@cell/ablation_mae/img_hybrid_24/VMAX/mae/mean\endcsname{12.21}
\expandafter\def\csname mcresult@cell/ablation_mae/img_hybrid_24/VMAX/mae/sd\endcsname{0.14}
\expandafter\def\csname mcresult@cell/ablation_mae/img_hybrid_48/MSLP/mae/mean\endcsname{12.69}
\expandafter\def\csname mcresult@cell/ablation_mae/img_hybrid_48/MSLP/mae/sd\endcsname{0.08}
\expandafter\def\csname mcresult@cell/ablation_mae/img_hybrid_48/VMAX/mae/mean\endcsname{18.35}
\expandafter\def\csname mcresult@cell/ablation_mae/img_hybrid_48/VMAX/mae/sd\endcsname{0.30}
\expandafter\def\csname mcresult@cell/ablation_mae/img_hybrid_72/MSLP/mae/mean\endcsname{14.53}
\expandafter\def\csname mcresult@cell/ablation_mae/img_hybrid_72/MSLP/mae/sd\endcsname{0.13}
\expandafter\def\csname mcresult@cell/ablation_mae/img_hybrid_72/VMAX/mae/mean\endcsname{20.21}
\expandafter\def\csname mcresult@cell/ablation_mae/img_hybrid_72/VMAX/mae/sd\endcsname{0.19}
\expandafter\def\csname mcresult@cell/ablation_mae/img_neural_24/MSLP/mae/mean\endcsname{8.15}
\expandafter\def\csname mcresult@cell/ablation_mae/img_neural_24/MSLP/mae/sd\endcsname{0.13}
\expandafter\def\csname mcresult@cell/ablation_mae/img_neural_24/VMAX/mae/mean\endcsname{12.40}
\expandafter\def\csname mcresult@cell/ablation_mae/img_neural_24/VMAX/mae/sd\endcsname{0.16}
\expandafter\def\csname mcresult@cell/ablation_mae/img_neural_48/MSLP/mae/mean\endcsname{13.31}
\expandafter\def\csname mcresult@cell/ablation_mae/img_neural_48/MSLP/mae/sd\endcsname{0.21}
\expandafter\def\csname mcresult@cell/ablation_mae/img_neural_48/VMAX/mae/mean\endcsname{18.97}
\expandafter\def\csname mcresult@cell/ablation_mae/img_neural_48/VMAX/mae/sd\endcsname{0.37}
\expandafter\def\csname mcresult@cell/ablation_mae/img_neural_72/MSLP/mae/mean\endcsname{15.13}
\expandafter\def\csname mcresult@cell/ablation_mae/img_neural_72/MSLP/mae/sd\endcsname{0.52}
\expandafter\def\csname mcresult@cell/ablation_mae/img_neural_72/VMAX/mae/mean\endcsname{20.43}
\expandafter\def\csname mcresult@cell/ablation_mae/img_neural_72/VMAX/mae/sd\endcsname{0.75}
\expandafter\def\csname mcresult@cell/ablation_mae/tab_disc_hybrid_24/MSLP/mae/mean\endcsname{6.36}
\expandafter\def\csname mcresult@cell/ablation_mae/tab_disc_hybrid_24/MSLP/mae/sd\endcsname{0.08}
\expandafter\def\csname mcresult@cell/ablation_mae/tab_disc_hybrid_24/VMAX/mae/mean\endcsname{9.49}
\expandafter\def\csname mcresult@cell/ablation_mae/tab_disc_hybrid_24/VMAX/mae/sd\endcsname{0.08}
\expandafter\def\csname mcresult@cell/ablation_mae/tab_disc_hybrid_48/MSLP/mae/mean\endcsname{11.14}
\expandafter\def\csname mcresult@cell/ablation_mae/tab_disc_hybrid_48/MSLP/mae/sd\endcsname{0.09}
\expandafter\def\csname mcresult@cell/ablation_mae/tab_disc_hybrid_48/VMAX/mae/mean\endcsname{15.41}
\expandafter\def\csname mcresult@cell/ablation_mae/tab_disc_hybrid_48/VMAX/mae/sd\endcsname{0.09}
\expandafter\def\csname mcresult@cell/ablation_mae/tab_disc_hybrid_72/MSLP/mae/mean\endcsname{13.90}
\expandafter\def\csname mcresult@cell/ablation_mae/tab_disc_hybrid_72/MSLP/mae/sd\endcsname{0.08}
\expandafter\def\csname mcresult@cell/ablation_mae/tab_disc_hybrid_72/VMAX/mae/mean\endcsname{18.55}
\expandafter\def\csname mcresult@cell/ablation_mae/tab_disc_hybrid_72/VMAX/mae/sd\endcsname{0.27}
\expandafter\def\csname mcresult@cell/ablation_mae/tab_disc_neural_24/MSLP/mae/mean\endcsname{6.23}
\expandafter\def\csname mcresult@cell/ablation_mae/tab_disc_neural_24/MSLP/mae/sd\endcsname{0.08}
\expandafter\def\csname mcresult@cell/ablation_mae/tab_disc_neural_24/VMAX/mae/mean\endcsname{9.46}
\expandafter\def\csname mcresult@cell/ablation_mae/tab_disc_neural_24/VMAX/mae/sd\endcsname{0.16}
\expandafter\def\csname mcresult@cell/ablation_mae/tab_disc_neural_48/MSLP/mae/mean\endcsname{9.58}
\expandafter\def\csname mcresult@cell/ablation_mae/tab_disc_neural_48/MSLP/mae/sd\endcsname{0.30}
\expandafter\def\csname mcresult@cell/ablation_mae/tab_disc_neural_48/VMAX/mae/mean\endcsname{13.44}
\expandafter\def\csname mcresult@cell/ablation_mae/tab_disc_neural_48/VMAX/mae/sd\endcsname{0.23}
\expandafter\def\csname mcresult@cell/ablation_mae/tab_disc_neural_72/MSLP/mae/mean\endcsname{12.50}
\expandafter\def\csname mcresult@cell/ablation_mae/tab_disc_neural_72/MSLP/mae/sd\endcsname{0.32}
\expandafter\def\csname mcresult@cell/ablation_mae/tab_disc_neural_72/VMAX/mae/mean\endcsname{16.93}
\expandafter\def\csname mcresult@cell/ablation_mae/tab_disc_neural_72/VMAX/mae/sd\endcsname{0.67}
\expandafter\def\csname mcresult@cell/ablation_mae/tab_hybrid_24/MSLP/mae/mean\endcsname{6.55}
\expandafter\def\csname mcresult@cell/ablation_mae/tab_hybrid_24/MSLP/mae/sd\endcsname{0.02}
\expandafter\def\csname mcresult@cell/ablation_mae/tab_hybrid_24/VMAX/mae/mean\endcsname{9.88}
\expandafter\def\csname mcresult@cell/ablation_mae/tab_hybrid_24/VMAX/mae/sd\endcsname{0.06}
\expandafter\def\csname mcresult@cell/ablation_mae/tab_hybrid_48/MSLP/mae/mean\endcsname{11.04}
\expandafter\def\csname mcresult@cell/ablation_mae/tab_hybrid_48/MSLP/mae/sd\endcsname{0.13}
\expandafter\def\csname mcresult@cell/ablation_mae/tab_hybrid_48/VMAX/mae/mean\endcsname{15.29}
\expandafter\def\csname mcresult@cell/ablation_mae/tab_hybrid_48/VMAX/mae/sd\endcsname{0.18}
\expandafter\def\csname mcresult@cell/ablation_mae/tab_hybrid_72/MSLP/mae/mean\endcsname{13.57}
\expandafter\def\csname mcresult@cell/ablation_mae/tab_hybrid_72/MSLP/mae/sd\endcsname{0.08}
\expandafter\def\csname mcresult@cell/ablation_mae/tab_hybrid_72/VMAX/mae/mean\endcsname{18.41}
\expandafter\def\csname mcresult@cell/ablation_mae/tab_hybrid_72/VMAX/mae/sd\endcsname{0.07}
\expandafter\def\csname mcresult@cell/ablation_mae/tab_img_hybrid_24/MSLP/mae/mean\endcsname{6.90}
\expandafter\def\csname mcresult@cell/ablation_mae/tab_img_hybrid_24/MSLP/mae/sd\endcsname{0.13}
\expandafter\def\csname mcresult@cell/ablation_mae/tab_img_hybrid_24/VMAX/mae/mean\endcsname{10.55}
\expandafter\def\csname mcresult@cell/ablation_mae/tab_img_hybrid_24/VMAX/mae/sd\endcsname{0.18}
\expandafter\def\csname mcresult@cell/ablation_mae/tab_img_hybrid_48/MSLP/mae/mean\endcsname{10.97}
\expandafter\def\csname mcresult@cell/ablation_mae/tab_img_hybrid_48/MSLP/mae/sd\endcsname{0.24}
\expandafter\def\csname mcresult@cell/ablation_mae/tab_img_hybrid_48/VMAX/mae/mean\endcsname{15.53}
\expandafter\def\csname mcresult@cell/ablation_mae/tab_img_hybrid_48/VMAX/mae/sd\endcsname{0.26}
\expandafter\def\csname mcresult@cell/ablation_mae/tab_img_hybrid_72/MSLP/mae/mean\endcsname{13.08}
\expandafter\def\csname mcresult@cell/ablation_mae/tab_img_hybrid_72/MSLP/mae/sd\endcsname{0.15}
\expandafter\def\csname mcresult@cell/ablation_mae/tab_img_hybrid_72/VMAX/mae/mean\endcsname{17.98}
\expandafter\def\csname mcresult@cell/ablation_mae/tab_img_hybrid_72/VMAX/mae/sd\endcsname{0.12}
\expandafter\def\csname mcresult@cell/ablation_mae/tab_img_neural_24/MSLP/mae/mean\endcsname{6.72}
\expandafter\def\csname mcresult@cell/ablation_mae/tab_img_neural_24/MSLP/mae/sd\endcsname{0.05}
\expandafter\def\csname mcresult@cell/ablation_mae/tab_img_neural_24/VMAX/mae/mean\endcsname{10.27}
\expandafter\def\csname mcresult@cell/ablation_mae/tab_img_neural_24/VMAX/mae/sd\endcsname{0.06}
\expandafter\def\csname mcresult@cell/ablation_mae/tab_img_neural_48/MSLP/mae/mean\endcsname{10.80}
\expandafter\def\csname mcresult@cell/ablation_mae/tab_img_neural_48/MSLP/mae/sd\endcsname{0.31}
\expandafter\def\csname mcresult@cell/ablation_mae/tab_img_neural_48/VMAX/mae/mean\endcsname{15.49}
\expandafter\def\csname mcresult@cell/ablation_mae/tab_img_neural_48/VMAX/mae/sd\endcsname{0.54}
\expandafter\def\csname mcresult@cell/ablation_mae/tab_img_neural_72/MSLP/mae/mean\endcsname{13.95}
\expandafter\def\csname mcresult@cell/ablation_mae/tab_img_neural_72/MSLP/mae/sd\endcsname{0.42}
\expandafter\def\csname mcresult@cell/ablation_mae/tab_img_neural_72/VMAX/mae/mean\endcsname{18.34}
\expandafter\def\csname mcresult@cell/ablation_mae/tab_img_neural_72/VMAX/mae/sd\endcsname{0.44}
\expandafter\def\csname mcresult@cell/ablation_mae/tab_neural_24/MSLP/mae/mean\endcsname{6.60}
\expandafter\def\csname mcresult@cell/ablation_mae/tab_neural_24/MSLP/mae/sd\endcsname{0.08}
\expandafter\def\csname mcresult@cell/ablation_mae/tab_neural_24/VMAX/mae/mean\endcsname{10.13}
\expandafter\def\csname mcresult@cell/ablation_mae/tab_neural_24/VMAX/mae/sd\endcsname{0.05}
\expandafter\def\csname mcresult@cell/ablation_mae/tab_neural_48/MSLP/mae/mean\endcsname{10.77}
\expandafter\def\csname mcresult@cell/ablation_mae/tab_neural_48/MSLP/mae/sd\endcsname{0.20}
\expandafter\def\csname mcresult@cell/ablation_mae/tab_neural_48/VMAX/mae/mean\endcsname{15.32}
\expandafter\def\csname mcresult@cell/ablation_mae/tab_neural_48/VMAX/mae/sd\endcsname{0.38}
\expandafter\def\csname mcresult@cell/ablation_mae/tab_neural_72/MSLP/mae/mean\endcsname{13.82}
\expandafter\def\csname mcresult@cell/ablation_mae/tab_neural_72/MSLP/mae/sd\endcsname{0.42}
\expandafter\def\csname mcresult@cell/ablation_mae/tab_neural_72/VMAX/mae/mean\endcsname{18.77}
\expandafter\def\csname mcresult@cell/ablation_mae/tab_neural_72/VMAX/mae/sd\endcsname{0.58}
\expandafter\def\csname mcresult@cell/baselines/Ridge_Tab/MSLP/mae/mean\endcsname{8.12}
\expandafter\def\csname mcresult@cell/baselines/Ridge_Tab/MSLP/r2/mean\endcsname{0.24}
\expandafter\def\csname mcresult@cell/baselines/Ridge_Tab/MSLP/rmse/mean\endcsname{11.36}
\expandafter\def\csname mcresult@cell/baselines/Ridge_Tab/VMAX/mae/mean\endcsname{12.14}
\expandafter\def\csname mcresult@cell/baselines/Ridge_Tab/VMAX/r2/mean\endcsname{0.26}
\expandafter\def\csname mcresult@cell/baselines/Ridge_Tab/VMAX/rmse/mean\endcsname{16.36}
\expandafter\def\csname mcresult@cell/baselines/Ridge_Tab_Text/MSLP/mae/mean\endcsname{7.18}
\expandafter\def\csname mcresult@cell/baselines/Ridge_Tab_Text/MSLP/r2/mean\endcsname{0.41}
\expandafter\def\csname mcresult@cell/baselines/Ridge_Tab_Text/MSLP/rmse/mean\endcsname{10.04}
\expandafter\def\csname mcresult@cell/baselines/Ridge_Tab_Text/VMAX/mae/mean\endcsname{10.47}
\expandafter\def\csname mcresult@cell/baselines/Ridge_Tab_Text/VMAX/r2/mean\endcsname{0.45}
\expandafter\def\csname mcresult@cell/baselines/Ridge_Tab_Text/VMAX/rmse/mean\endcsname{14.11}
\expandafter\def\csname mcresult@cell/baselines/Ridge_Text/MSLP/mae/mean\endcsname{7.47}
\expandafter\def\csname mcresult@cell/baselines/Ridge_Text/MSLP/r2/mean\endcsname{0.36}
\expandafter\def\csname mcresult@cell/baselines/Ridge_Text/MSLP/rmse/mean\endcsname{10.41}
\expandafter\def\csname mcresult@cell/baselines/Ridge_Text/VMAX/mae/mean\endcsname{10.93}
\expandafter\def\csname mcresult@cell/baselines/Ridge_Text/VMAX/r2/mean\endcsname{0.40}
\expandafter\def\csname mcresult@cell/baselines/Ridge_Text/VMAX/rmse/mean\endcsname{14.74}
\expandafter\def\csname mcresult@cell/baselines/XGB_Tab/MSLP/mae/mean\endcsname{6.55}
\expandafter\def\csname mcresult@cell/baselines/XGB_Tab/MSLP/mae/sd\endcsname{0.02}
\expandafter\def\csname mcresult@cell/baselines/XGB_Tab/MSLP/r2/mean\endcsname{0.45}
\expandafter\def\csname mcresult@cell/baselines/XGB_Tab/MSLP/r2/sd\endcsname{0.00}
\expandafter\def\csname mcresult@cell/baselines/XGB_Tab/MSLP/rmse/mean\endcsname{9.69}
\expandafter\def\csname mcresult@cell/baselines/XGB_Tab/MSLP/rmse/sd\endcsname{0.03}
\expandafter\def\csname mcresult@cell/baselines/XGB_Tab/VMAX/mae/mean\endcsname{9.88}
\expandafter\def\csname mcresult@cell/baselines/XGB_Tab/VMAX/mae/sd\endcsname{0.06}
\expandafter\def\csname mcresult@cell/baselines/XGB_Tab/VMAX/r2/mean\endcsname{0.46}
\expandafter\def\csname mcresult@cell/baselines/XGB_Tab/VMAX/r2/sd\endcsname{0.00}
\expandafter\def\csname mcresult@cell/baselines/XGB_Tab/VMAX/rmse/mean\endcsname{14.04}
\expandafter\def\csname mcresult@cell/baselines/XGB_Tab/VMAX/rmse/sd\endcsname{0.06}
\expandafter\def\csname mcresult@cell/baselines/XGB_Tab_Text/MSLP/mae/mean\endcsname{6.36}
\expandafter\def\csname mcresult@cell/baselines/XGB_Tab_Text/MSLP/mae/sd\endcsname{0.08}
\expandafter\def\csname mcresult@cell/baselines/XGB_Tab_Text/MSLP/r2/mean\endcsname{0.51}
\expandafter\def\csname mcresult@cell/baselines/XGB_Tab_Text/MSLP/r2/sd\endcsname{0.01}
\expandafter\def\csname mcresult@cell/baselines/XGB_Tab_Text/MSLP/rmse/mean\endcsname{9.16}
\expandafter\def\csname mcresult@cell/baselines/XGB_Tab_Text/MSLP/rmse/sd\endcsname{0.08}
\expandafter\def\csname mcresult@cell/baselines/XGB_Tab_Text/VMAX/mae/mean\endcsname{9.49}
\expandafter\def\csname mcresult@cell/baselines/XGB_Tab_Text/VMAX/mae/sd\endcsname{0.08}
\expandafter\def\csname mcresult@cell/baselines/XGB_Tab_Text/VMAX/r2/mean\endcsname{0.51}
\expandafter\def\csname mcresult@cell/baselines/XGB_Tab_Text/VMAX/r2/sd\endcsname{0.01}
\expandafter\def\csname mcresult@cell/baselines/XGB_Tab_Text/VMAX/rmse/mean\endcsname{13.29}
\expandafter\def\csname mcresult@cell/baselines/XGB_Tab_Text/VMAX/rmse/sd\endcsname{0.13}
\expandafter\def\csname mcresult@cell/baselines/XGB_Text/MSLP/mae/mean\endcsname{7.26}
\expandafter\def\csname mcresult@cell/baselines/XGB_Text/MSLP/mae/sd\endcsname{0.05}
\expandafter\def\csname mcresult@cell/baselines/XGB_Text/MSLP/r2/mean\endcsname{0.35}
\expandafter\def\csname mcresult@cell/baselines/XGB_Text/MSLP/r2/sd\endcsname{0.01}
\expandafter\def\csname mcresult@cell/baselines/XGB_Text/MSLP/rmse/mean\endcsname{10.50}
\expandafter\def\csname mcresult@cell/baselines/XGB_Text/MSLP/rmse/sd\endcsname{0.05}
\expandafter\def\csname mcresult@cell/baselines/XGB_Text/VMAX/mae/mean\endcsname{10.79}
\expandafter\def\csname mcresult@cell/baselines/XGB_Text/VMAX/mae/sd\endcsname{0.07}
\expandafter\def\csname mcresult@cell/baselines/XGB_Text/VMAX/r2/mean\endcsname{0.39}
\expandafter\def\csname mcresult@cell/baselines/XGB_Text/VMAX/r2/sd\endcsname{0.01}
\expandafter\def\csname mcresult@cell/baselines/XGB_Text/VMAX/rmse/mean\endcsname{14.92}
\expandafter\def\csname mcresult@cell/baselines/XGB_Text/VMAX/rmse/sd\endcsname{0.10}
\expandafter\def\csname mcresult@cell/baselines/exact_h_hour_trend/MSLPmae/mean\endcsname{10.32}
\expandafter\def\csname mcresult@cell/baselines/exact_h_hour_trend/MSLPr2/mean\endcsname{-0.60}
\expandafter\def\csname mcresult@cell/baselines/exact_h_hour_trend/MSLPrmse/mean\endcsname{16.51}
\expandafter\def\csname mcresult@cell/baselines/exact_h_hour_trend/VMAXmae/mean\endcsname{16.12}
\expandafter\def\csname mcresult@cell/baselines/exact_h_hour_trend/VMAXr2/mean\endcsname{-0.54}
\expandafter\def\csname mcresult@cell/baselines/exact_h_hour_trend/VMAXrmse/mean\endcsname{23.63}
\expandafter\def\csname mcresult@cell/baselines/persistence/MSLPmae/mean\endcsname{8.86}
\expandafter\def\csname mcresult@cell/baselines/persistence/MSLPr2/mean\endcsname{-0.04}
\expandafter\def\csname mcresult@cell/baselines/persistence/MSLPrmse/mean\endcsname{13.29}
\expandafter\def\csname mcresult@cell/baselines/persistence/VMAXmae/mean\endcsname{13.78}
\expandafter\def\csname mcresult@cell/baselines/persistence/VMAXr2/mean\endcsname{-0.03}
\expandafter\def\csname mcresult@cell/baselines/persistence/VMAXrmse/mean\endcsname{19.33}
\expandafter\def\csname mcresult@cell/baselines/train_mean_climatology/MSLPmae/mean\endcsname{8.73}
\expandafter\def\csname mcresult@cell/baselines/train_mean_climatology/MSLPr2/mean\endcsname{-0.01}
\expandafter\def\csname mcresult@cell/baselines/train_mean_climatology/MSLPrmse/mean\endcsname{13.11}
\expandafter\def\csname mcresult@cell/baselines/train_mean_climatology/VMAXmae/mean\endcsname{13.76}
\expandafter\def\csname mcresult@cell/baselines/train_mean_climatology/VMAXr2/mean\endcsname{-0.01}
\expandafter\def\csname mcresult@cell/baselines/train_mean_climatology/VMAXrmse/mean\endcsname{19.15}
\expandafter\def\csname mcresult@cell/best_models_comparison/Hybrid_T/MSLP/mae/mean\endcsname{6.55}
\expandafter\def\csname mcresult@cell/best_models_comparison/Hybrid_T/MSLP/mae/sd\endcsname{0.02}
\expandafter\def\csname mcresult@cell/best_models_comparison/Hybrid_T/MSLP/r2/mean\endcsname{0.45}
\expandafter\def\csname mcresult@cell/best_models_comparison/Hybrid_T/MSLP/r2/sd\endcsname{0.00}
\expandafter\def\csname mcresult@cell/best_models_comparison/Hybrid_T/VMAX/dir_acc/mean\endcsname{0.700}
\expandafter\def\csname mcresult@cell/best_models_comparison/Hybrid_T/VMAX/dir_acc/sd\endcsname{0.004}
\expandafter\def\csname mcresult@cell/best_models_comparison/Hybrid_T/VMAX/mae/mean\endcsname{9.88}
\expandafter\def\csname mcresult@cell/best_models_comparison/Hybrid_T/VMAX/mae/sd\endcsname{0.06}
\expandafter\def\csname mcresult@cell/best_models_comparison/Hybrid_T/VMAX/r2/mean\endcsname{0.46}
\expandafter\def\csname mcresult@cell/best_models_comparison/Hybrid_T/VMAX/r2/sd\endcsname{0.00}
\expandafter\def\csname mcresult@cell/best_models_comparison/Hybrid_T_D/MSLP/mae/mean\endcsname{6.36}
\expandafter\def\csname mcresult@cell/best_models_comparison/Hybrid_T_D/MSLP/mae/sd\endcsname{0.08}
\expandafter\def\csname mcresult@cell/best_models_comparison/Hybrid_T_D/MSLP/r2/mean\endcsname{0.51}
\expandafter\def\csname mcresult@cell/best_models_comparison/Hybrid_T_D/MSLP/r2/sd\endcsname{0.01}
\expandafter\def\csname mcresult@cell/best_models_comparison/Hybrid_T_D/VMAX/dir_acc/mean\endcsname{0.714}
\expandafter\def\csname mcresult@cell/best_models_comparison/Hybrid_T_D/VMAX/dir_acc/sd\endcsname{0.006}
\expandafter\def\csname mcresult@cell/best_models_comparison/Hybrid_T_D/VMAX/mae/mean\endcsname{9.49}
\expandafter\def\csname mcresult@cell/best_models_comparison/Hybrid_T_D/VMAX/mae/sd\endcsname{0.08}
\expandafter\def\csname mcresult@cell/best_models_comparison/Hybrid_T_D/VMAX/r2/mean\endcsname{0.51}
\expandafter\def\csname mcresult@cell/best_models_comparison/Hybrid_T_D/VMAX/r2/sd\endcsname{0.01}
\expandafter\def\csname mcresult@cell/best_models_comparison/Hybrid_T_I_D/MSLP/mae/mean\endcsname{6.43}
\expandafter\def\csname mcresult@cell/best_models_comparison/Hybrid_T_I_D/MSLP/mae/sd\endcsname{0.18}
\expandafter\def\csname mcresult@cell/best_models_comparison/Hybrid_T_I_D/MSLP/r2/mean\endcsname{0.49}
\expandafter\def\csname mcresult@cell/best_models_comparison/Hybrid_T_I_D/MSLP/r2/sd\endcsname{0.03}
\expandafter\def\csname mcresult@cell/best_models_comparison/Hybrid_T_I_D/VMAX/dir_acc/mean\endcsname{0.708}
\expandafter\def\csname mcresult@cell/best_models_comparison/Hybrid_T_I_D/VMAX/dir_acc/sd\endcsname{0.009}
\expandafter\def\csname mcresult@cell/best_models_comparison/Hybrid_T_I_D/VMAX/mae/mean\endcsname{9.78}
\expandafter\def\csname mcresult@cell/best_models_comparison/Hybrid_T_I_D/VMAX/mae/sd\endcsname{0.16}
\expandafter\def\csname mcresult@cell/best_models_comparison/Hybrid_T_I_D/VMAX/r2/mean\endcsname{0.49}
\expandafter\def\csname mcresult@cell/best_models_comparison/Hybrid_T_I_D/VMAX/r2/sd\endcsname{0.02}
\expandafter\def\csname mcresult@cell/best_models_comparison/Neural_T/MSLP/mae/mean\endcsname{6.60}
\expandafter\def\csname mcresult@cell/best_models_comparison/Neural_T/MSLP/mae/sd\endcsname{0.08}
\expandafter\def\csname mcresult@cell/best_models_comparison/Neural_T/MSLP/r2/mean\endcsname{0.44}
\expandafter\def\csname mcresult@cell/best_models_comparison/Neural_T/MSLP/r2/sd\endcsname{0.01}
\expandafter\def\csname mcresult@cell/best_models_comparison/Neural_T/VMAX/dir_acc/mean\endcsname{0.684}
\expandafter\def\csname mcresult@cell/best_models_comparison/Neural_T/VMAX/dir_acc/sd\endcsname{0.006}
\expandafter\def\csname mcresult@cell/best_models_comparison/Neural_T/VMAX/mae/mean\endcsname{10.13}
\expandafter\def\csname mcresult@cell/best_models_comparison/Neural_T/VMAX/mae/sd\endcsname{0.05}
\expandafter\def\csname mcresult@cell/best_models_comparison/Neural_T/VMAX/r2/mean\endcsname{0.44}
\expandafter\def\csname mcresult@cell/best_models_comparison/Neural_T/VMAX/r2/sd\endcsname{0.01}
\expandafter\def\csname mcresult@cell/best_models_comparison/Neural_T_D/MSLP/mae/mean\endcsname{6.23}
\expandafter\def\csname mcresult@cell/best_models_comparison/Neural_T_D/MSLP/mae/sd\endcsname{0.08}
\expandafter\def\csname mcresult@cell/best_models_comparison/Neural_T_D/MSLP/r2/mean\endcsname{0.51}
\expandafter\def\csname mcresult@cell/best_models_comparison/Neural_T_D/MSLP/r2/sd\endcsname{0.01}
\expandafter\def\csname mcresult@cell/best_models_comparison/Neural_T_D/VMAX/dir_acc/mean\endcsname{0.723}
\expandafter\def\csname mcresult@cell/best_models_comparison/Neural_T_D/VMAX/dir_acc/sd\endcsname{0.009}
\expandafter\def\csname mcresult@cell/best_models_comparison/Neural_T_D/VMAX/mae/mean\endcsname{9.46}
\expandafter\def\csname mcresult@cell/best_models_comparison/Neural_T_D/VMAX/mae/sd\endcsname{0.16}
\expandafter\def\csname mcresult@cell/best_models_comparison/Neural_T_D/VMAX/r2/mean\endcsname{0.53}
\expandafter\def\csname mcresult@cell/best_models_comparison/Neural_T_D/VMAX/r2/sd\endcsname{0.01}
\expandafter\def\csname mcresult@cell/best_models_comparison/Neural_T_I_D/MSLP/mae/mean\endcsname{6.23}
\expandafter\def\csname mcresult@cell/best_models_comparison/Neural_T_I_D/MSLP/mae/sd\endcsname{0.08}
\expandafter\def\csname mcresult@cell/best_models_comparison/Neural_T_I_D/MSLP/r2/mean\endcsname{0.51}
\expandafter\def\csname mcresult@cell/best_models_comparison/Neural_T_I_D/MSLP/r2/sd\endcsname{0.01}
\expandafter\def\csname mcresult@cell/best_models_comparison/Neural_T_I_D/VMAX/dir_acc/mean\endcsname{0.724}
\expandafter\def\csname mcresult@cell/best_models_comparison/Neural_T_I_D/VMAX/dir_acc/sd\endcsname{0.006}
\expandafter\def\csname mcresult@cell/best_models_comparison/Neural_T_I_D/VMAX/mae/mean\endcsname{9.45}
\expandafter\def\csname mcresult@cell/best_models_comparison/Neural_T_I_D/VMAX/mae/sd\endcsname{0.13}
\expandafter\def\csname mcresult@cell/best_models_comparison/Neural_T_I_D/VMAX/r2/mean\endcsname{0.53}
\expandafter\def\csname mcresult@cell/best_models_comparison/Neural_T_I_D/VMAX/r2/sd\endcsname{0.01}
\expandafter\def\csname mcresult@cell/best_models_comparison/Temporal_T_I/MSLP/mae/mean\endcsname{7.05}
\expandafter\def\csname mcresult@cell/best_models_comparison/Temporal_T_I/MSLP/mae/sd\endcsname{0.10}
\expandafter\def\csname mcresult@cell/best_models_comparison/Temporal_T_I/MSLP/r2/mean\endcsname{0.37}
\expandafter\def\csname mcresult@cell/best_models_comparison/Temporal_T_I/MSLP/r2/sd\endcsname{0.03}
\expandafter\def\csname mcresult@cell/best_models_comparison/Temporal_T_I/VMAX/dir_acc/mean\endcsname{0.648}
\expandafter\def\csname mcresult@cell/best_models_comparison/Temporal_T_I/VMAX/dir_acc/sd\endcsname{0.004}
\expandafter\def\csname mcresult@cell/best_models_comparison/Temporal_T_I/VMAX/mae/mean\endcsname{10.56}
\expandafter\def\csname mcresult@cell/best_models_comparison/Temporal_T_I/VMAX/mae/sd\endcsname{0.11}
\expandafter\def\csname mcresult@cell/best_models_comparison/Temporal_T_I/VMAX/r2/mean\endcsname{0.36}
\expandafter\def\csname mcresult@cell/best_models_comparison/Temporal_T_I/VMAX/r2/sd\endcsname{0.01}
\expandafter\def\csname mcresult@cell/best_models_comparison/Temporal_T_I_D_GRU/MSLP/mae/mean\endcsname{6.63}
\expandafter\def\csname mcresult@cell/best_models_comparison/Temporal_T_I_D_GRU/MSLP/mae/sd\endcsname{0.13}
\expandafter\def\csname mcresult@cell/best_models_comparison/Temporal_T_I_D_GRU/MSLP/r2/mean\endcsname{0.45}
\expandafter\def\csname mcresult@cell/best_models_comparison/Temporal_T_I_D_GRU/MSLP/r2/sd\endcsname{0.01}
\expandafter\def\csname mcresult@cell/best_models_comparison/Temporal_T_I_D_GRU/VMAX/dir_acc/mean\endcsname{0.656}
\expandafter\def\csname mcresult@cell/best_models_comparison/Temporal_T_I_D_GRU/VMAX/dir_acc/sd\endcsname{0.015}
\expandafter\def\csname mcresult@cell/best_models_comparison/Temporal_T_I_D_GRU/VMAX/mae/mean\endcsname{9.92}
\expandafter\def\csname mcresult@cell/best_models_comparison/Temporal_T_I_D_GRU/VMAX/mae/sd\endcsname{0.27}
\expandafter\def\csname mcresult@cell/best_models_comparison/Temporal_T_I_D_GRU/VMAX/r2/mean\endcsname{0.44}
\expandafter\def\csname mcresult@cell/best_models_comparison/Temporal_T_I_D_GRU/VMAX/r2/sd\endcsname{0.02}
\expandafter\def\csname mcresult@cell/best_models_comparison/Temporal_T_I_D_GRU_XGB/MSLP/mae/mean\endcsname{6.41}
\expandafter\def\csname mcresult@cell/best_models_comparison/Temporal_T_I_D_GRU_XGB/MSLP/mae/sd\endcsname{0.17}
\expandafter\def\csname mcresult@cell/best_models_comparison/Temporal_T_I_D_GRU_XGB/MSLP/r2/mean\endcsname{0.47}
\expandafter\def\csname mcresult@cell/best_models_comparison/Temporal_T_I_D_GRU_XGB/MSLP/r2/sd\endcsname{0.02}
\expandafter\def\csname mcresult@cell/best_models_comparison/Temporal_T_I_D_GRU_XGB/VMAX/dir_acc/mean\endcsname{0.666}
\expandafter\def\csname mcresult@cell/best_models_comparison/Temporal_T_I_D_GRU_XGB/VMAX/dir_acc/sd\endcsname{0.010}
\expandafter\def\csname mcresult@cell/best_models_comparison/Temporal_T_I_D_GRU_XGB/VMAX/mae/mean\endcsname{9.68}
\expandafter\def\csname mcresult@cell/best_models_comparison/Temporal_T_I_D_GRU_XGB/VMAX/mae/sd\endcsname{0.21}
\expandafter\def\csname mcresult@cell/best_models_comparison/Temporal_T_I_D_GRU_XGB/VMAX/r2/mean\endcsname{0.46}
\expandafter\def\csname mcresult@cell/best_models_comparison/Temporal_T_I_D_GRU_XGB/VMAX/r2/sd\endcsname{0.02}
\expandafter\def\csname mcresult@cell/combined_ablation_24h/disc_hybrid_24/MSLP/r2/mean\endcsname{0.35}
\expandafter\def\csname mcresult@cell/combined_ablation_24h/disc_hybrid_24/MSLP/r2/sd\endcsname{0.01}
\expandafter\def\csname mcresult@cell/combined_ablation_24h/disc_hybrid_24/VMAX/r2/mean\endcsname{0.39}
\expandafter\def\csname mcresult@cell/combined_ablation_24h/disc_hybrid_24/VMAX/r2/sd\endcsname{0.01}
\expandafter\def\csname mcresult@cell/combined_ablation_24h/disc_hybrid_48/MSLP/r2/mean\endcsname{0.31}
\expandafter\def\csname mcresult@cell/combined_ablation_24h/disc_hybrid_48/MSLP/r2/sd\endcsname{0.02}
\expandafter\def\csname mcresult@cell/combined_ablation_24h/disc_hybrid_48/VMAX/r2/mean\endcsname{0.37}
\expandafter\def\csname mcresult@cell/combined_ablation_24h/disc_hybrid_48/VMAX/r2/sd\endcsname{0.01}
\expandafter\def\csname mcresult@cell/combined_ablation_24h/disc_hybrid_72/MSLP/r2/mean\endcsname{0.20}
\expandafter\def\csname mcresult@cell/combined_ablation_24h/disc_hybrid_72/MSLP/r2/sd\endcsname{0.01}
\expandafter\def\csname mcresult@cell/combined_ablation_24h/disc_hybrid_72/VMAX/r2/mean\endcsname{0.27}
\expandafter\def\csname mcresult@cell/combined_ablation_24h/disc_hybrid_72/VMAX/r2/sd\endcsname{0.01}
\expandafter\def\csname mcresult@cell/combined_ablation_24h/disc_neural_24/MSLP/r2/mean\endcsname{0.46}
\expandafter\def\csname mcresult@cell/combined_ablation_24h/disc_neural_24/MSLP/r2/sd\endcsname{0.03}
\expandafter\def\csname mcresult@cell/combined_ablation_24h/disc_neural_24/VMAX/r2/mean\endcsname{0.50}
\expandafter\def\csname mcresult@cell/combined_ablation_24h/disc_neural_24/VMAX/r2/sd\endcsname{0.01}
\expandafter\def\csname mcresult@cell/combined_ablation_24h/disc_neural_48/MSLP/r2/mean\endcsname{0.51}
\expandafter\def\csname mcresult@cell/combined_ablation_24h/disc_neural_48/MSLP/r2/sd\endcsname{0.01}
\expandafter\def\csname mcresult@cell/combined_ablation_24h/disc_neural_48/VMAX/r2/mean\endcsname{0.54}
\expandafter\def\csname mcresult@cell/combined_ablation_24h/disc_neural_48/VMAX/r2/sd\endcsname{0.01}
\expandafter\def\csname mcresult@cell/combined_ablation_24h/disc_neural_72/MSLP/r2/mean\endcsname{0.35}
\expandafter\def\csname mcresult@cell/combined_ablation_24h/disc_neural_72/MSLP/r2/sd\endcsname{0.06}
\expandafter\def\csname mcresult@cell/combined_ablation_24h/disc_neural_72/VMAX/r2/mean\endcsname{0.42}
\expandafter\def\csname mcresult@cell/combined_ablation_24h/disc_neural_72/VMAX/r2/sd\endcsname{0.06}
\expandafter\def\csname mcresult@cell/combined_ablation_24h/full_hybrid_24/MSLP/r2/mean\endcsname{0.49}
\expandafter\def\csname mcresult@cell/combined_ablation_24h/full_hybrid_24/MSLP/r2/sd\endcsname{0.03}
\expandafter\def\csname mcresult@cell/combined_ablation_24h/full_hybrid_24/VMAX/r2/mean\endcsname{0.49}
\expandafter\def\csname mcresult@cell/combined_ablation_24h/full_hybrid_24/VMAX/r2/sd\endcsname{0.02}
\expandafter\def\csname mcresult@cell/combined_ablation_24h/full_hybrid_48/MSLP/r2/mean\endcsname{0.47}
\expandafter\def\csname mcresult@cell/combined_ablation_24h/full_hybrid_48/MSLP/r2/sd\endcsname{0.02}
\expandafter\def\csname mcresult@cell/combined_ablation_24h/full_hybrid_48/VMAX/r2/mean\endcsname{0.49}
\expandafter\def\csname mcresult@cell/combined_ablation_24h/full_hybrid_48/VMAX/r2/sd\endcsname{0.01}
\expandafter\def\csname mcresult@cell/combined_ablation_24h/full_hybrid_72/MSLP/r2/mean\endcsname{0.40}
\expandafter\def\csname mcresult@cell/combined_ablation_24h/full_hybrid_72/MSLP/r2/sd\endcsname{0.02}
\expandafter\def\csname mcresult@cell/combined_ablation_24h/full_hybrid_72/VMAX/r2/mean\endcsname{0.44}
\expandafter\def\csname mcresult@cell/combined_ablation_24h/full_hybrid_72/VMAX/r2/sd\endcsname{0.01}
\expandafter\def\csname mcresult@cell/combined_ablation_24h/full_neural_24/MSLP/r2/mean\endcsname{0.51}
\expandafter\def\csname mcresult@cell/combined_ablation_24h/full_neural_24/MSLP/r2/sd\endcsname{0.01}
\expandafter\def\csname mcresult@cell/combined_ablation_24h/full_neural_24/VMAX/r2/mean\endcsname{0.53}
\expandafter\def\csname mcresult@cell/combined_ablation_24h/full_neural_24/VMAX/r2/sd\endcsname{0.01}
\expandafter\def\csname mcresult@cell/combined_ablation_24h/full_neural_48/MSLP/r2/mean\endcsname{0.54}
\expandafter\def\csname mcresult@cell/combined_ablation_24h/full_neural_48/MSLP/r2/sd\endcsname{0.02}
\expandafter\def\csname mcresult@cell/combined_ablation_24h/full_neural_48/VMAX/r2/mean\endcsname{0.57}
\expandafter\def\csname mcresult@cell/combined_ablation_24h/full_neural_48/VMAX/r2/sd\endcsname{0.01}
\expandafter\def\csname mcresult@cell/combined_ablation_24h/full_neural_72/MSLP/r2/mean\endcsname{0.39}
\expandafter\def\csname mcresult@cell/combined_ablation_24h/full_neural_72/MSLP/r2/sd\endcsname{0.06}
\expandafter\def\csname mcresult@cell/combined_ablation_24h/full_neural_72/VMAX/r2/mean\endcsname{0.45}
\expandafter\def\csname mcresult@cell/combined_ablation_24h/full_neural_72/VMAX/r2/sd\endcsname{0.04}
\expandafter\def\csname mcresult@cell/combined_ablation_24h/img_disc_hybrid_24/MSLP/r2/mean\endcsname{0.39}
\expandafter\def\csname mcresult@cell/combined_ablation_24h/img_disc_hybrid_24/MSLP/r2/sd\endcsname{0.01}
\expandafter\def\csname mcresult@cell/combined_ablation_24h/img_disc_hybrid_24/VMAX/r2/mean\endcsname{0.40}
\expandafter\def\csname mcresult@cell/combined_ablation_24h/img_disc_hybrid_24/VMAX/r2/sd\endcsname{0.02}
\expandafter\def\csname mcresult@cell/combined_ablation_24h/img_disc_hybrid_48/MSLP/r2/mean\endcsname{0.37}
\expandafter\def\csname mcresult@cell/combined_ablation_24h/img_disc_hybrid_48/MSLP/r2/sd\endcsname{0.03}
\expandafter\def\csname mcresult@cell/combined_ablation_24h/img_disc_hybrid_48/VMAX/r2/mean\endcsname{0.42}
\expandafter\def\csname mcresult@cell/combined_ablation_24h/img_disc_hybrid_48/VMAX/r2/sd\endcsname{0.01}
\expandafter\def\csname mcresult@cell/combined_ablation_24h/img_disc_hybrid_72/MSLP/r2/mean\endcsname{0.28}
\expandafter\def\csname mcresult@cell/combined_ablation_24h/img_disc_hybrid_72/MSLP/r2/sd\endcsname{0.02}
\expandafter\def\csname mcresult@cell/combined_ablation_24h/img_disc_hybrid_72/VMAX/r2/mean\endcsname{0.35}
\expandafter\def\csname mcresult@cell/combined_ablation_24h/img_disc_hybrid_72/VMAX/r2/sd\endcsname{0.03}
\expandafter\def\csname mcresult@cell/combined_ablation_24h/img_disc_neural_24/MSLP/r2/mean\endcsname{0.48}
\expandafter\def\csname mcresult@cell/combined_ablation_24h/img_disc_neural_24/MSLP/r2/sd\endcsname{0.02}
\expandafter\def\csname mcresult@cell/combined_ablation_24h/img_disc_neural_24/VMAX/r2/mean\endcsname{0.50}
\expandafter\def\csname mcresult@cell/combined_ablation_24h/img_disc_neural_24/VMAX/r2/sd\endcsname{0.01}
\expandafter\def\csname mcresult@cell/combined_ablation_24h/img_disc_neural_48/MSLP/r2/mean\endcsname{0.49}
\expandafter\def\csname mcresult@cell/combined_ablation_24h/img_disc_neural_48/MSLP/r2/sd\endcsname{0.02}
\expandafter\def\csname mcresult@cell/combined_ablation_24h/img_disc_neural_48/VMAX/r2/mean\endcsname{0.52}
\expandafter\def\csname mcresult@cell/combined_ablation_24h/img_disc_neural_48/VMAX/r2/sd\endcsname{0.03}
\expandafter\def\csname mcresult@cell/combined_ablation_24h/img_disc_neural_72/MSLP/r2/mean\endcsname{0.36}
\expandafter\def\csname mcresult@cell/combined_ablation_24h/img_disc_neural_72/MSLP/r2/sd\endcsname{0.05}
\expandafter\def\csname mcresult@cell/combined_ablation_24h/img_disc_neural_72/VMAX/r2/mean\endcsname{0.39}
\expandafter\def\csname mcresult@cell/combined_ablation_24h/img_disc_neural_72/VMAX/r2/sd\endcsname{0.10}
\expandafter\def\csname mcresult@cell/combined_ablation_24h/img_hybrid_24/MSLP/r2/mean\endcsname{0.19}
\expandafter\def\csname mcresult@cell/combined_ablation_24h/img_hybrid_24/MSLP/r2/sd\endcsname{0.01}
\expandafter\def\csname mcresult@cell/combined_ablation_24h/img_hybrid_24/VMAX/r2/mean\endcsname{0.23}
\expandafter\def\csname mcresult@cell/combined_ablation_24h/img_hybrid_24/VMAX/r2/sd\endcsname{0.02}
\expandafter\def\csname mcresult@cell/combined_ablation_24h/img_hybrid_48/MSLP/r2/mean\endcsname{0.23}
\expandafter\def\csname mcresult@cell/combined_ablation_24h/img_hybrid_48/MSLP/r2/sd\endcsname{0.02}
\expandafter\def\csname mcresult@cell/combined_ablation_24h/img_hybrid_48/VMAX/r2/mean\endcsname{0.26}
\expandafter\def\csname mcresult@cell/combined_ablation_24h/img_hybrid_48/VMAX/r2/sd\endcsname{0.02}
\expandafter\def\csname mcresult@cell/combined_ablation_24h/img_hybrid_72/MSLP/r2/mean\endcsname{0.22}
\expandafter\def\csname mcresult@cell/combined_ablation_24h/img_hybrid_72/MSLP/r2/sd\endcsname{0.01}
\expandafter\def\csname mcresult@cell/combined_ablation_24h/img_hybrid_72/VMAX/r2/mean\endcsname{0.27}
\expandafter\def\csname mcresult@cell/combined_ablation_24h/img_hybrid_72/VMAX/r2/sd\endcsname{0.01}
\expandafter\def\csname mcresult@cell/combined_ablation_24h/img_neural_24/MSLP/r2/mean\endcsname{0.20}
\expandafter\def\csname mcresult@cell/combined_ablation_24h/img_neural_24/MSLP/r2/sd\endcsname{0.02}
\expandafter\def\csname mcresult@cell/combined_ablation_24h/img_neural_24/VMAX/r2/mean\endcsname{0.22}
\expandafter\def\csname mcresult@cell/combined_ablation_24h/img_neural_24/VMAX/r2/sd\endcsname{0.02}
\expandafter\def\csname mcresult@cell/combined_ablation_24h/img_neural_48/MSLP/r2/mean\endcsname{0.12}
\expandafter\def\csname mcresult@cell/combined_ablation_24h/img_neural_48/MSLP/r2/sd\endcsname{0.06}
\expandafter\def\csname mcresult@cell/combined_ablation_24h/img_neural_48/VMAX/r2/mean\endcsname{0.16}
\expandafter\def\csname mcresult@cell/combined_ablation_24h/img_neural_48/VMAX/r2/sd\endcsname{0.06}
\expandafter\def\csname mcresult@cell/combined_ablation_24h/img_neural_72/MSLP/r2/mean\endcsname{0.15}
\expandafter\def\csname mcresult@cell/combined_ablation_24h/img_neural_72/MSLP/r2/sd\endcsname{0.08}
\expandafter\def\csname mcresult@cell/combined_ablation_24h/img_neural_72/VMAX/r2/mean\endcsname{0.24}
\expandafter\def\csname mcresult@cell/combined_ablation_24h/img_neural_72/VMAX/r2/sd\endcsname{0.06}
\expandafter\def\csname mcresult@cell/combined_ablation_24h/tab_disc_hybrid_24/MSLP/r2/mean\endcsname{0.51}
\expandafter\def\csname mcresult@cell/combined_ablation_24h/tab_disc_hybrid_24/MSLP/r2/sd\endcsname{0.01}
\expandafter\def\csname mcresult@cell/combined_ablation_24h/tab_disc_hybrid_24/VMAX/r2/mean\endcsname{0.51}
\expandafter\def\csname mcresult@cell/combined_ablation_24h/tab_disc_hybrid_24/VMAX/r2/sd\endcsname{0.01}
\expandafter\def\csname mcresult@cell/combined_ablation_24h/tab_disc_hybrid_48/MSLP/r2/mean\endcsname{0.48}
\expandafter\def\csname mcresult@cell/combined_ablation_24h/tab_disc_hybrid_48/MSLP/r2/sd\endcsname{0.01}
\expandafter\def\csname mcresult@cell/combined_ablation_24h/tab_disc_hybrid_48/VMAX/r2/mean\endcsname{0.51}
\expandafter\def\csname mcresult@cell/combined_ablation_24h/tab_disc_hybrid_48/VMAX/r2/sd\endcsname{0.01}
\expandafter\def\csname mcresult@cell/combined_ablation_24h/tab_disc_hybrid_72/MSLP/r2/mean\endcsname{0.37}
\expandafter\def\csname mcresult@cell/combined_ablation_24h/tab_disc_hybrid_72/MSLP/r2/sd\endcsname{0.01}
\expandafter\def\csname mcresult@cell/combined_ablation_24h/tab_disc_hybrid_72/VMAX/r2/mean\endcsname{0.42}
\expandafter\def\csname mcresult@cell/combined_ablation_24h/tab_disc_hybrid_72/VMAX/r2/sd\endcsname{0.01}
\expandafter\def\csname mcresult@cell/combined_ablation_24h/tab_disc_neural_24/MSLP/r2/mean\endcsname{0.51}
\expandafter\def\csname mcresult@cell/combined_ablation_24h/tab_disc_neural_24/MSLP/r2/sd\endcsname{0.01}
\expandafter\def\csname mcresult@cell/combined_ablation_24h/tab_disc_neural_24/VMAX/r2/mean\endcsname{0.53}
\expandafter\def\csname mcresult@cell/combined_ablation_24h/tab_disc_neural_24/VMAX/r2/sd\endcsname{0.01}
\expandafter\def\csname mcresult@cell/combined_ablation_24h/tab_disc_neural_48/MSLP/r2/mean\endcsname{0.55}
\expandafter\def\csname mcresult@cell/combined_ablation_24h/tab_disc_neural_48/MSLP/r2/sd\endcsname{0.02}
\expandafter\def\csname mcresult@cell/combined_ablation_24h/tab_disc_neural_48/VMAX/r2/mean\endcsname{0.58}
\expandafter\def\csname mcresult@cell/combined_ablation_24h/tab_disc_neural_48/VMAX/r2/sd\endcsname{0.02}
\expandafter\def\csname mcresult@cell/combined_ablation_24h/tab_disc_neural_72/MSLP/r2/mean\endcsname{0.39}
\expandafter\def\csname mcresult@cell/combined_ablation_24h/tab_disc_neural_72/MSLP/r2/sd\endcsname{0.06}
\expandafter\def\csname mcresult@cell/combined_ablation_24h/tab_disc_neural_72/VMAX/r2/mean\endcsname{0.46}
\expandafter\def\csname mcresult@cell/combined_ablation_24h/tab_disc_neural_72/VMAX/r2/sd\endcsname{0.06}
\expandafter\def\csname mcresult@cell/combined_ablation_24h/tab_hybrid_24/MSLP/r2/mean\endcsname{0.45}
\expandafter\def\csname mcresult@cell/combined_ablation_24h/tab_hybrid_24/MSLP/r2/sd\endcsname{0.00}
\expandafter\def\csname mcresult@cell/combined_ablation_24h/tab_hybrid_24/VMAX/r2/mean\endcsname{0.46}
\expandafter\def\csname mcresult@cell/combined_ablation_24h/tab_hybrid_24/VMAX/r2/sd\endcsname{0.00}
\expandafter\def\csname mcresult@cell/combined_ablation_24h/tab_hybrid_48/MSLP/r2/mean\endcsname{0.45}
\expandafter\def\csname mcresult@cell/combined_ablation_24h/tab_hybrid_48/MSLP/r2/sd\endcsname{0.01}
\expandafter\def\csname mcresult@cell/combined_ablation_24h/tab_hybrid_48/VMAX/r2/mean\endcsname{0.49}
\expandafter\def\csname mcresult@cell/combined_ablation_24h/tab_hybrid_48/VMAX/r2/sd\endcsname{0.01}
\expandafter\def\csname mcresult@cell/combined_ablation_24h/tab_hybrid_72/MSLP/r2/mean\endcsname{0.36}
\expandafter\def\csname mcresult@cell/combined_ablation_24h/tab_hybrid_72/MSLP/r2/sd\endcsname{0.01}
\expandafter\def\csname mcresult@cell/combined_ablation_24h/tab_hybrid_72/VMAX/r2/mean\endcsname{0.42}
\expandafter\def\csname mcresult@cell/combined_ablation_24h/tab_hybrid_72/VMAX/r2/sd\endcsname{0.01}
\expandafter\def\csname mcresult@cell/combined_ablation_24h/tab_img_hybrid_24/MSLP/r2/mean\endcsname{0.40}
\expandafter\def\csname mcresult@cell/combined_ablation_24h/tab_img_hybrid_24/MSLP/r2/sd\endcsname{0.02}
\expandafter\def\csname mcresult@cell/combined_ablation_24h/tab_img_hybrid_24/VMAX/r2/mean\endcsname{0.40}
\expandafter\def\csname mcresult@cell/combined_ablation_24h/tab_img_hybrid_24/VMAX/r2/sd\endcsname{0.02}
\expandafter\def\csname mcresult@cell/combined_ablation_24h/tab_img_hybrid_48/MSLP/r2/mean\endcsname{0.42}
\expandafter\def\csname mcresult@cell/combined_ablation_24h/tab_img_hybrid_48/MSLP/r2/sd\endcsname{0.02}
\expandafter\def\csname mcresult@cell/combined_ablation_24h/tab_img_hybrid_48/VMAX/r2/mean\endcsname{0.43}
\expandafter\def\csname mcresult@cell/combined_ablation_24h/tab_img_hybrid_48/VMAX/r2/sd\endcsname{0.02}
\expandafter\def\csname mcresult@cell/combined_ablation_24h/tab_img_hybrid_72/MSLP/r2/mean\endcsname{0.39}
\expandafter\def\csname mcresult@cell/combined_ablation_24h/tab_img_hybrid_72/MSLP/r2/sd\endcsname{0.01}
\expandafter\def\csname mcresult@cell/combined_ablation_24h/tab_img_hybrid_72/VMAX/r2/mean\endcsname{0.42}
\expandafter\def\csname mcresult@cell/combined_ablation_24h/tab_img_hybrid_72/VMAX/r2/sd\endcsname{0.01}
\expandafter\def\csname mcresult@cell/combined_ablation_24h/tab_img_neural_24/MSLP/r2/mean\endcsname{0.45}
\expandafter\def\csname mcresult@cell/combined_ablation_24h/tab_img_neural_24/MSLP/r2/sd\endcsname{0.01}
\expandafter\def\csname mcresult@cell/combined_ablation_24h/tab_img_neural_24/VMAX/r2/mean\endcsname{0.45}
\expandafter\def\csname mcresult@cell/combined_ablation_24h/tab_img_neural_24/VMAX/r2/sd\endcsname{0.01}
\expandafter\def\csname mcresult@cell/combined_ablation_24h/tab_img_neural_48/MSLP/r2/mean\endcsname{0.42}
\expandafter\def\csname mcresult@cell/combined_ablation_24h/tab_img_neural_48/MSLP/r2/sd\endcsname{0.03}
\expandafter\def\csname mcresult@cell/combined_ablation_24h/tab_img_neural_48/VMAX/r2/mean\endcsname{0.45}
\expandafter\def\csname mcresult@cell/combined_ablation_24h/tab_img_neural_48/VMAX/r2/sd\endcsname{0.02}
\expandafter\def\csname mcresult@cell/combined_ablation_24h/tab_img_neural_72/MSLP/r2/mean\endcsname{0.31}
\expandafter\def\csname mcresult@cell/combined_ablation_24h/tab_img_neural_72/MSLP/r2/sd\endcsname{0.10}
\expandafter\def\csname mcresult@cell/combined_ablation_24h/tab_img_neural_72/VMAX/r2/mean\endcsname{0.41}
\expandafter\def\csname mcresult@cell/combined_ablation_24h/tab_img_neural_72/VMAX/r2/sd\endcsname{0.03}
\expandafter\def\csname mcresult@cell/combined_ablation_24h/tab_neural_24/MSLP/r2/mean\endcsname{0.44}
\expandafter\def\csname mcresult@cell/combined_ablation_24h/tab_neural_24/MSLP/r2/sd\endcsname{0.01}
\expandafter\def\csname mcresult@cell/combined_ablation_24h/tab_neural_24/VMAX/r2/mean\endcsname{0.44}
\expandafter\def\csname mcresult@cell/combined_ablation_24h/tab_neural_24/VMAX/r2/sd\endcsname{0.01}
\expandafter\def\csname mcresult@cell/combined_ablation_24h/tab_neural_48/MSLP/r2/mean\endcsname{0.44}
\expandafter\def\csname mcresult@cell/combined_ablation_24h/tab_neural_48/MSLP/r2/sd\endcsname{0.02}
\expandafter\def\csname mcresult@cell/combined_ablation_24h/tab_neural_48/VMAX/r2/mean\endcsname{0.47}
\expandafter\def\csname mcresult@cell/combined_ablation_24h/tab_neural_48/VMAX/r2/sd\endcsname{0.02}
\expandafter\def\csname mcresult@cell/combined_ablation_24h/tab_neural_72/MSLP/r2/mean\endcsname{0.33}
\expandafter\def\csname mcresult@cell/combined_ablation_24h/tab_neural_72/MSLP/r2/sd\endcsname{0.03}
\expandafter\def\csname mcresult@cell/combined_ablation_24h/tab_neural_72/VMAX/r2/mean\endcsname{0.39}
\expandafter\def\csname mcresult@cell/combined_ablation_24h/tab_neural_72/VMAX/r2/sd\endcsname{0.02}
\expandafter\def\csname mcresult@cell/dacc_confusion/Balanced_accuracy/tab/mean\endcsname{0.524}
\expandafter\def\csname mcresult@cell/dacc_confusion/Balanced_accuracy/tabD/mean\endcsname{0.568}
\expandafter\def\csname mcresult@cell/dacc_confusion/Overall_accuracy/tab/mean\endcsname{0.690}
\expandafter\def\csname mcresult@cell/dacc_confusion/Overall_accuracy/tabD/mean\endcsname{0.737}
\expandafter\def\csname mcresult@cell/dacc_confusion/Steady/tabD_recall/mean\endcsname{0.000}
\expandafter\def\csname mcresult@cell/dacc_confusion/Steady/tab_recall/mean\endcsname{0.000}
\expandafter\def\csname mcresult@cell/dacc_confusion/Strengthening/tabD_recall/mean\endcsname{0.894}
\expandafter\def\csname mcresult@cell/dacc_confusion/Strengthening/tab_recall/mean\endcsname{0.889}
\expandafter\def\csname mcresult@cell/dacc_confusion/Weakening/tabD_recall/mean\endcsname{0.810}
\expandafter\def\csname mcresult@cell/dacc_confusion/Weakening/tab_recall/mean\endcsname{0.684}
\expandafter\def\csname mcresult@cell/dacc_tolerance_24h/Balanced_accuracy/tab/mean\endcsname{0.561}
\expandafter\def\csname mcresult@cell/dacc_tolerance_24h/Balanced_accuracy/tabD/mean\endcsname{0.602}
\expandafter\def\csname mcresult@cell/dacc_tolerance_24h/Overall_accuracy/tab/mean\endcsname{0.655}
\expandafter\def\csname mcresult@cell/dacc_tolerance_24h/Overall_accuracy/tabD/mean\endcsname{0.710}
\expandafter\def\csname mcresult@cell/dacc_tolerance_24h/Steady/tabD_recall/mean\endcsname{0.255}
\expandafter\def\csname mcresult@cell/dacc_tolerance_24h/Steady/tab_recall/mean\endcsname{0.291}
\expandafter\def\csname mcresult@cell/dacc_tolerance_24h/Strengthening/tabD_recall/mean\endcsname{0.833}
\expandafter\def\csname mcresult@cell/dacc_tolerance_24h/Strengthening/tab_recall/mean\endcsname{0.797}
\expandafter\def\csname mcresult@cell/dacc_tolerance_24h/Weakening/tabD_recall/mean\endcsname{0.719}
\expandafter\def\csname mcresult@cell/dacc_tolerance_24h/Weakening/tab_recall/mean\endcsname{0.596}
\expandafter\def\csname mcresult@cell/full_results_mslp/disc_120/MSLP/r2/mean\endcsname{0.00}
\expandafter\def\csname mcresult@cell/full_results_mslp/disc_120/MSLP/r2/sd\endcsname{0.15}
\expandafter\def\csname mcresult@cell/full_results_mslp/disc_120/MSLP/rmse/mean\endcsname{22.00}
\expandafter\def\csname mcresult@cell/full_results_mslp/disc_120/MSLP/rmse/sd\endcsname{1.56}
\expandafter\def\csname mcresult@cell/full_results_mslp/disc_120/MSLP/skill_score/mean\endcsname{0.00}
\expandafter\def\csname mcresult@cell/full_results_mslp/disc_120/MSLP/skill_score/sd\endcsname{0.07}
\expandafter\def\csname mcresult@cell/full_results_mslp/disc_24/MSLP/r2/mean\endcsname{0.48}
\expandafter\def\csname mcresult@cell/full_results_mslp/disc_24/MSLP/r2/sd\endcsname{0.02}
\expandafter\def\csname mcresult@cell/full_results_mslp/disc_24/MSLP/rmse/mean\endcsname{9.51}
\expandafter\def\csname mcresult@cell/full_results_mslp/disc_24/MSLP/rmse/sd\endcsname{0.16}
\expandafter\def\csname mcresult@cell/full_results_mslp/disc_24/MSLP/skill_score/mean\endcsname{0.28}
\expandafter\def\csname mcresult@cell/full_results_mslp/disc_24/MSLP/skill_score/sd\endcsname{0.01}
\expandafter\def\csname mcresult@cell/full_results_mslp/disc_48/MSLP/r2/mean\endcsname{0.51}
\expandafter\def\csname mcresult@cell/full_results_mslp/disc_48/MSLP/r2/sd\endcsname{0.01}
\expandafter\def\csname mcresult@cell/full_results_mslp/disc_48/MSLP/rmse/mean\endcsname{13.87}
\expandafter\def\csname mcresult@cell/full_results_mslp/disc_48/MSLP/rmse/sd\endcsname{0.20}
\expandafter\def\csname mcresult@cell/full_results_mslp/disc_48/MSLP/skill_score/mean\endcsname{0.30}
\expandafter\def\csname mcresult@cell/full_results_mslp/disc_48/MSLP/skill_score/sd\endcsname{0.01}
\expandafter\def\csname mcresult@cell/full_results_mslp/disc_72/MSLP/r2/mean\endcsname{0.35}
\expandafter\def\csname mcresult@cell/full_results_mslp/disc_72/MSLP/r2/sd\endcsname{0.06}
\expandafter\def\csname mcresult@cell/full_results_mslp/disc_72/MSLP/rmse/mean\endcsname{17.36}
\expandafter\def\csname mcresult@cell/full_results_mslp/disc_72/MSLP/rmse/sd\endcsname{0.80}
\expandafter\def\csname mcresult@cell/full_results_mslp/disc_72/MSLP/skill_score/mean\endcsname{0.19}
\expandafter\def\csname mcresult@cell/full_results_mslp/disc_72/MSLP/skill_score/sd\endcsname{0.04}
\expandafter\def\csname mcresult@cell/full_results_mslp/disc_96/MSLP/r2/mean\endcsname{0.14}
\expandafter\def\csname mcresult@cell/full_results_mslp/disc_96/MSLP/r2/sd\endcsname{0.14}
\expandafter\def\csname mcresult@cell/full_results_mslp/disc_96/MSLP/rmse/mean\endcsname{19.73}
\expandafter\def\csname mcresult@cell/full_results_mslp/disc_96/MSLP/rmse/sd\endcsname{1.57}
\expandafter\def\csname mcresult@cell/full_results_mslp/disc_96/MSLP/skill_score/mean\endcsname{0.07}
\expandafter\def\csname mcresult@cell/full_results_mslp/disc_96/MSLP/skill_score/sd\endcsname{0.07}
\expandafter\def\csname mcresult@cell/full_results_mslp/full_120/MSLP/r2/mean\endcsname{0.01}
\expandafter\def\csname mcresult@cell/full_results_mslp/full_120/MSLP/r2/sd\endcsname{0.10}
\expandafter\def\csname mcresult@cell/full_results_mslp/full_120/MSLP/rmse/mean\endcsname{21.74}
\expandafter\def\csname mcresult@cell/full_results_mslp/full_120/MSLP/rmse/sd\endcsname{1.09}
\expandafter\def\csname mcresult@cell/full_results_mslp/full_120/MSLP/skill_score/mean\endcsname{0.01}
\expandafter\def\csname mcresult@cell/full_results_mslp/full_120/MSLP/skill_score/sd\endcsname{0.05}
\expandafter\def\csname mcresult@cell/full_results_mslp/full_24/MSLP/r2/mean\endcsname{0.52}
\expandafter\def\csname mcresult@cell/full_results_mslp/full_24/MSLP/r2/sd\endcsname{0.03}
\expandafter\def\csname mcresult@cell/full_results_mslp/full_24/MSLP/rmse/mean\endcsname{9.03}
\expandafter\def\csname mcresult@cell/full_results_mslp/full_24/MSLP/rmse/sd\endcsname{0.31}
\expandafter\def\csname mcresult@cell/full_results_mslp/full_24/MSLP/skill_score/mean\endcsname{0.31}
\expandafter\def\csname mcresult@cell/full_results_mslp/full_24/MSLP/skill_score/sd\endcsname{0.02}
\expandafter\def\csname mcresult@cell/full_results_mslp/full_48/MSLP/r2/mean\endcsname{0.54}
\expandafter\def\csname mcresult@cell/full_results_mslp/full_48/MSLP/r2/sd\endcsname{0.02}
\expandafter\def\csname mcresult@cell/full_results_mslp/full_48/MSLP/rmse/mean\endcsname{13.34}
\expandafter\def\csname mcresult@cell/full_results_mslp/full_48/MSLP/rmse/sd\endcsname{0.26}
\expandafter\def\csname mcresult@cell/full_results_mslp/full_48/MSLP/skill_score/mean\endcsname{0.32}
\expandafter\def\csname mcresult@cell/full_results_mslp/full_48/MSLP/skill_score/sd\endcsname{0.01}
\expandafter\def\csname mcresult@cell/full_results_mslp/full_72/MSLP/r2/mean\endcsname{0.39}
\expandafter\def\csname mcresult@cell/full_results_mslp/full_72/MSLP/r2/sd\endcsname{0.06}
\expandafter\def\csname mcresult@cell/full_results_mslp/full_72/MSLP/rmse/mean\endcsname{16.66}
\expandafter\def\csname mcresult@cell/full_results_mslp/full_72/MSLP/rmse/sd\endcsname{0.89}
\expandafter\def\csname mcresult@cell/full_results_mslp/full_72/MSLP/skill_score/mean\endcsname{0.22}
\expandafter\def\csname mcresult@cell/full_results_mslp/full_72/MSLP/skill_score/sd\endcsname{0.04}
\expandafter\def\csname mcresult@cell/full_results_mslp/full_96/MSLP/r2/mean\endcsname{0.18}
\expandafter\def\csname mcresult@cell/full_results_mslp/full_96/MSLP/r2/sd\endcsname{0.08}
\expandafter\def\csname mcresult@cell/full_results_mslp/full_96/MSLP/rmse/mean\endcsname{19.12}
\expandafter\def\csname mcresult@cell/full_results_mslp/full_96/MSLP/rmse/sd\endcsname{0.94}
\expandafter\def\csname mcresult@cell/full_results_mslp/full_96/MSLP/skill_score/mean\endcsname{0.10}
\expandafter\def\csname mcresult@cell/full_results_mslp/full_96/MSLP/skill_score/sd\endcsname{0.04}
\expandafter\def\csname mcresult@cell/full_results_mslp/img_120/MSLP/r2/mean\endcsname{0.05}
\expandafter\def\csname mcresult@cell/full_results_mslp/img_120/MSLP/r2/sd\endcsname{0.03}
\expandafter\def\csname mcresult@cell/full_results_mslp/img_120/MSLP/rmse/mean\endcsname{21.33}
\expandafter\def\csname mcresult@cell/full_results_mslp/img_120/MSLP/rmse/sd\endcsname{0.33}
\expandafter\def\csname mcresult@cell/full_results_mslp/img_120/MSLP/skill_score/mean\endcsname{0.02}
\expandafter\def\csname mcresult@cell/full_results_mslp/img_120/MSLP/skill_score/sd\endcsname{0.01}
\expandafter\def\csname mcresult@cell/full_results_mslp/img_24/MSLP/r2/mean\endcsname{0.18}
\expandafter\def\csname mcresult@cell/full_results_mslp/img_24/MSLP/r2/sd\endcsname{0.03}
\expandafter\def\csname mcresult@cell/full_results_mslp/img_24/MSLP/rmse/mean\endcsname{11.82}
\expandafter\def\csname mcresult@cell/full_results_mslp/img_24/MSLP/rmse/sd\endcsname{0.22}
\expandafter\def\csname mcresult@cell/full_results_mslp/img_24/MSLP/skill_score/mean\endcsname{0.09}
\expandafter\def\csname mcresult@cell/full_results_mslp/img_24/MSLP/skill_score/sd\endcsname{0.02}
\expandafter\def\csname mcresult@cell/full_results_mslp/img_48/MSLP/r2/mean\endcsname{0.12}
\expandafter\def\csname mcresult@cell/full_results_mslp/img_48/MSLP/r2/sd\endcsname{0.06}
\expandafter\def\csname mcresult@cell/full_results_mslp/img_48/MSLP/rmse/mean\endcsname{18.49}
\expandafter\def\csname mcresult@cell/full_results_mslp/img_48/MSLP/rmse/sd\endcsname{0.65}
\expandafter\def\csname mcresult@cell/full_results_mslp/img_48/MSLP/skill_score/mean\endcsname{0.06}
\expandafter\def\csname mcresult@cell/full_results_mslp/img_48/MSLP/skill_score/sd\endcsname{0.03}
\expandafter\def\csname mcresult@cell/full_results_mslp/img_72/MSLP/r2/mean\endcsname{0.15}
\expandafter\def\csname mcresult@cell/full_results_mslp/img_72/MSLP/r2/sd\endcsname{0.08}
\expandafter\def\csname mcresult@cell/full_results_mslp/img_72/MSLP/rmse/mean\endcsname{19.67}
\expandafter\def\csname mcresult@cell/full_results_mslp/img_72/MSLP/rmse/sd\endcsname{0.89}
\expandafter\def\csname mcresult@cell/full_results_mslp/img_72/MSLP/skill_score/mean\endcsname{0.08}
\expandafter\def\csname mcresult@cell/full_results_mslp/img_72/MSLP/skill_score/sd\endcsname{0.04}
\expandafter\def\csname mcresult@cell/full_results_mslp/img_96/MSLP/r2/mean\endcsname{0.07}
\expandafter\def\csname mcresult@cell/full_results_mslp/img_96/MSLP/r2/sd\endcsname{0.09}
\expandafter\def\csname mcresult@cell/full_results_mslp/img_96/MSLP/rmse/mean\endcsname{20.30}
\expandafter\def\csname mcresult@cell/full_results_mslp/img_96/MSLP/rmse/sd\endcsname{0.94}
\expandafter\def\csname mcresult@cell/full_results_mslp/img_96/MSLP/skill_score/mean\endcsname{0.04}
\expandafter\def\csname mcresult@cell/full_results_mslp/img_96/MSLP/skill_score/sd\endcsname{0.04}
\expandafter\def\csname mcresult@cell/full_results_mslp/img_disc_120/MSLP/r2/mean\endcsname{0.02}
\expandafter\def\csname mcresult@cell/full_results_mslp/img_disc_120/MSLP/r2/sd\endcsname{0.15}
\expandafter\def\csname mcresult@cell/full_results_mslp/img_disc_120/MSLP/rmse/mean\endcsname{21.61}
\expandafter\def\csname mcresult@cell/full_results_mslp/img_disc_120/MSLP/rmse/sd\endcsname{1.72}
\expandafter\def\csname mcresult@cell/full_results_mslp/img_disc_120/MSLP/skill_score/mean\endcsname{0.01}
\expandafter\def\csname mcresult@cell/full_results_mslp/img_disc_120/MSLP/skill_score/sd\endcsname{0.08}
\expandafter\def\csname mcresult@cell/full_results_mslp/img_disc_24/MSLP/r2/mean\endcsname{0.47}
\expandafter\def\csname mcresult@cell/full_results_mslp/img_disc_24/MSLP/r2/sd\endcsname{0.03}
\expandafter\def\csname mcresult@cell/full_results_mslp/img_disc_24/MSLP/rmse/mean\endcsname{9.51}
\expandafter\def\csname mcresult@cell/full_results_mslp/img_disc_24/MSLP/rmse/sd\endcsname{0.23}
\expandafter\def\csname mcresult@cell/full_results_mslp/img_disc_24/MSLP/skill_score/mean\endcsname{0.27}
\expandafter\def\csname mcresult@cell/full_results_mslp/img_disc_24/MSLP/skill_score/sd\endcsname{0.02}
\expandafter\def\csname mcresult@cell/full_results_mslp/img_disc_48/MSLP/r2/mean\endcsname{0.49}
\expandafter\def\csname mcresult@cell/full_results_mslp/img_disc_48/MSLP/r2/sd\endcsname{0.02}
\expandafter\def\csname mcresult@cell/full_results_mslp/img_disc_48/MSLP/rmse/mean\endcsname{14.06}
\expandafter\def\csname mcresult@cell/full_results_mslp/img_disc_48/MSLP/rmse/sd\endcsname{0.31}
\expandafter\def\csname mcresult@cell/full_results_mslp/img_disc_48/MSLP/skill_score/mean\endcsname{0.29}
\expandafter\def\csname mcresult@cell/full_results_mslp/img_disc_48/MSLP/skill_score/sd\endcsname{0.02}
\expandafter\def\csname mcresult@cell/full_results_mslp/img_disc_72/MSLP/r2/mean\endcsname{0.36}
\expandafter\def\csname mcresult@cell/full_results_mslp/img_disc_72/MSLP/r2/sd\endcsname{0.05}
\expandafter\def\csname mcresult@cell/full_results_mslp/img_disc_72/MSLP/rmse/mean\endcsname{17.09}
\expandafter\def\csname mcresult@cell/full_results_mslp/img_disc_72/MSLP/rmse/sd\endcsname{0.66}
\expandafter\def\csname mcresult@cell/full_results_mslp/img_disc_72/MSLP/skill_score/mean\endcsname{0.20}
\expandafter\def\csname mcresult@cell/full_results_mslp/img_disc_72/MSLP/skill_score/sd\endcsname{0.03}
\expandafter\def\csname mcresult@cell/full_results_mslp/img_disc_96/MSLP/r2/mean\endcsname{0.03}
\expandafter\def\csname mcresult@cell/full_results_mslp/img_disc_96/MSLP/r2/sd\endcsname{0.22}
\expandafter\def\csname mcresult@cell/full_results_mslp/img_disc_96/MSLP/rmse/mean\endcsname{20.72}
\expandafter\def\csname mcresult@cell/full_results_mslp/img_disc_96/MSLP/rmse/sd\endcsname{2.34}
\expandafter\def\csname mcresult@cell/full_results_mslp/img_disc_96/MSLP/skill_score/mean\endcsname{0.02}
\expandafter\def\csname mcresult@cell/full_results_mslp/img_disc_96/MSLP/skill_score/sd\endcsname{0.11}
\expandafter\def\csname mcresult@cell/full_results_mslp/tab_120/MSLP/r2/mean\endcsname{0.12}
\expandafter\def\csname mcresult@cell/full_results_mslp/tab_120/MSLP/r2/sd\endcsname{0.05}
\expandafter\def\csname mcresult@cell/full_results_mslp/tab_120/MSLP/rmse/mean\endcsname{20.55}
\expandafter\def\csname mcresult@cell/full_results_mslp/tab_120/MSLP/rmse/sd\endcsname{0.59}
\expandafter\def\csname mcresult@cell/full_results_mslp/tab_120/MSLP/skill_score/mean\endcsname{0.06}
\expandafter\def\csname mcresult@cell/full_results_mslp/tab_120/MSLP/skill_score/sd\endcsname{0.03}
\expandafter\def\csname mcresult@cell/full_results_mslp/tab_24/MSLP/r2/mean\endcsname{0.44}
\expandafter\def\csname mcresult@cell/full_results_mslp/tab_24/MSLP/r2/sd\endcsname{0.01}
\expandafter\def\csname mcresult@cell/full_results_mslp/tab_24/MSLP/rmse/mean\endcsname{9.74}
\expandafter\def\csname mcresult@cell/full_results_mslp/tab_24/MSLP/rmse/sd\endcsname{0.09}
\expandafter\def\csname mcresult@cell/full_results_mslp/tab_24/MSLP/skill_score/mean\endcsname{0.25}
\expandafter\def\csname mcresult@cell/full_results_mslp/tab_24/MSLP/skill_score/sd\endcsname{0.01}
\expandafter\def\csname mcresult@cell/full_results_mslp/tab_48/MSLP/r2/mean\endcsname{0.44}
\expandafter\def\csname mcresult@cell/full_results_mslp/tab_48/MSLP/r2/sd\endcsname{0.02}
\expandafter\def\csname mcresult@cell/full_results_mslp/tab_48/MSLP/rmse/mean\endcsname{14.80}
\expandafter\def\csname mcresult@cell/full_results_mslp/tab_48/MSLP/rmse/sd\endcsname{0.24}
\expandafter\def\csname mcresult@cell/full_results_mslp/tab_48/MSLP/skill_score/mean\endcsname{0.25}
\expandafter\def\csname mcresult@cell/full_results_mslp/tab_48/MSLP/skill_score/sd\endcsname{0.01}
\expandafter\def\csname mcresult@cell/full_results_mslp/tab_72/MSLP/r2/mean\endcsname{0.33}
\expandafter\def\csname mcresult@cell/full_results_mslp/tab_72/MSLP/r2/sd\endcsname{0.03}
\expandafter\def\csname mcresult@cell/full_results_mslp/tab_72/MSLP/rmse/mean\endcsname{17.52}
\expandafter\def\csname mcresult@cell/full_results_mslp/tab_72/MSLP/rmse/sd\endcsname{0.39}
\expandafter\def\csname mcresult@cell/full_results_mslp/tab_72/MSLP/skill_score/mean\endcsname{0.18}
\expandafter\def\csname mcresult@cell/full_results_mslp/tab_72/MSLP/skill_score/sd\endcsname{0.02}
\expandafter\def\csname mcresult@cell/full_results_mslp/tab_96/MSLP/r2/mean\endcsname{0.18}
\expandafter\def\csname mcresult@cell/full_results_mslp/tab_96/MSLP/r2/sd\endcsname{0.05}
\expandafter\def\csname mcresult@cell/full_results_mslp/tab_96/MSLP/rmse/mean\endcsname{19.09}
\expandafter\def\csname mcresult@cell/full_results_mslp/tab_96/MSLP/rmse/sd\endcsname{0.57}
\expandafter\def\csname mcresult@cell/full_results_mslp/tab_96/MSLP/skill_score/mean\endcsname{0.10}
\expandafter\def\csname mcresult@cell/full_results_mslp/tab_96/MSLP/skill_score/sd\endcsname{0.03}
\expandafter\def\csname mcresult@cell/full_results_mslp/tab_disc_120/MSLP/r2/mean\endcsname{0.06}
\expandafter\def\csname mcresult@cell/full_results_mslp/tab_disc_120/MSLP/r2/sd\endcsname{0.07}
\expandafter\def\csname mcresult@cell/full_results_mslp/tab_disc_120/MSLP/rmse/mean\endcsname{21.17}
\expandafter\def\csname mcresult@cell/full_results_mslp/tab_disc_120/MSLP/rmse/sd\endcsname{0.74}
\expandafter\def\csname mcresult@cell/full_results_mslp/tab_disc_120/MSLP/skill_score/mean\endcsname{0.03}
\expandafter\def\csname mcresult@cell/full_results_mslp/tab_disc_120/MSLP/skill_score/sd\endcsname{0.03}
\expandafter\def\csname mcresult@cell/full_results_mslp/tab_disc_24/MSLP/r2/mean\endcsname{0.50}
\expandafter\def\csname mcresult@cell/full_results_mslp/tab_disc_24/MSLP/r2/sd\endcsname{0.04}
\expandafter\def\csname mcresult@cell/full_results_mslp/tab_disc_24/MSLP/rmse/mean\endcsname{9.21}
\expandafter\def\csname mcresult@cell/full_results_mslp/tab_disc_24/MSLP/rmse/sd\endcsname{0.36}
\expandafter\def\csname mcresult@cell/full_results_mslp/tab_disc_24/MSLP/skill_score/mean\endcsname{0.29}
\expandafter\def\csname mcresult@cell/full_results_mslp/tab_disc_24/MSLP/skill_score/sd\endcsname{0.03}
\expandafter\def\csname mcresult@cell/full_results_mslp/tab_disc_48/MSLP/r2/mean\endcsname{0.55}
\expandafter\def\csname mcresult@cell/full_results_mslp/tab_disc_48/MSLP/r2/sd\endcsname{0.02}
\expandafter\def\csname mcresult@cell/full_results_mslp/tab_disc_48/MSLP/rmse/mean\endcsname{13.15}
\expandafter\def\csname mcresult@cell/full_results_mslp/tab_disc_48/MSLP/rmse/sd\endcsname{0.32}
\expandafter\def\csname mcresult@cell/full_results_mslp/tab_disc_48/MSLP/skill_score/mean\endcsname{0.33}
\expandafter\def\csname mcresult@cell/full_results_mslp/tab_disc_48/MSLP/skill_score/sd\endcsname{0.02}
\expandafter\def\csname mcresult@cell/full_results_mslp/tab_disc_72/MSLP/r2/mean\endcsname{0.39}
\expandafter\def\csname mcresult@cell/full_results_mslp/tab_disc_72/MSLP/r2/sd\endcsname{0.06}
\expandafter\def\csname mcresult@cell/full_results_mslp/tab_disc_72/MSLP/rmse/mean\endcsname{16.66}
\expandafter\def\csname mcresult@cell/full_results_mslp/tab_disc_72/MSLP/rmse/sd\endcsname{0.80}
\expandafter\def\csname mcresult@cell/full_results_mslp/tab_disc_72/MSLP/skill_score/mean\endcsname{0.22}
\expandafter\def\csname mcresult@cell/full_results_mslp/tab_disc_72/MSLP/skill_score/sd\endcsname{0.04}
\expandafter\def\csname mcresult@cell/full_results_mslp/tab_disc_96/MSLP/r2/mean\endcsname{0.19}
\expandafter\def\csname mcresult@cell/full_results_mslp/tab_disc_96/MSLP/r2/sd\endcsname{0.09}
\expandafter\def\csname mcresult@cell/full_results_mslp/tab_disc_96/MSLP/rmse/mean\endcsname{19.02}
\expandafter\def\csname mcresult@cell/full_results_mslp/tab_disc_96/MSLP/rmse/sd\endcsname{1.08}
\expandafter\def\csname mcresult@cell/full_results_mslp/tab_disc_96/MSLP/skill_score/mean\endcsname{0.10}
\expandafter\def\csname mcresult@cell/full_results_mslp/tab_disc_96/MSLP/skill_score/sd\endcsname{0.05}
\expandafter\def\csname mcresult@cell/full_results_mslp/tab_img_120/MSLP/r2/mean\endcsname{0.05}
\expandafter\def\csname mcresult@cell/full_results_mslp/tab_img_120/MSLP/r2/sd\endcsname{0.10}
\expandafter\def\csname mcresult@cell/full_results_mslp/tab_img_120/MSLP/rmse/mean\endcsname{21.28}
\expandafter\def\csname mcresult@cell/full_results_mslp/tab_img_120/MSLP/rmse/sd\endcsname{1.08}
\expandafter\def\csname mcresult@cell/full_results_mslp/tab_img_120/MSLP/skill_score/mean\endcsname{0.03}
\expandafter\def\csname mcresult@cell/full_results_mslp/tab_img_120/MSLP/skill_score/sd\endcsname{0.05}
\expandafter\def\csname mcresult@cell/full_results_mslp/tab_img_24/MSLP/r2/mean\endcsname{0.46}
\expandafter\def\csname mcresult@cell/full_results_mslp/tab_img_24/MSLP/r2/sd\endcsname{0.03}
\expandafter\def\csname mcresult@cell/full_results_mslp/tab_img_24/MSLP/rmse/mean\endcsname{9.58}
\expandafter\def\csname mcresult@cell/full_results_mslp/tab_img_24/MSLP/rmse/sd\endcsname{0.22}
\expandafter\def\csname mcresult@cell/full_results_mslp/tab_img_24/MSLP/skill_score/mean\endcsname{0.27}
\expandafter\def\csname mcresult@cell/full_results_mslp/tab_img_24/MSLP/skill_score/sd\endcsname{0.02}
\expandafter\def\csname mcresult@cell/full_results_mslp/tab_img_48/MSLP/r2/mean\endcsname{0.42}
\expandafter\def\csname mcresult@cell/full_results_mslp/tab_img_48/MSLP/r2/sd\endcsname{0.03}
\expandafter\def\csname mcresult@cell/full_results_mslp/tab_img_48/MSLP/rmse/mean\endcsname{15.01}
\expandafter\def\csname mcresult@cell/full_results_mslp/tab_img_48/MSLP/rmse/sd\endcsname{0.40}
\expandafter\def\csname mcresult@cell/full_results_mslp/tab_img_48/MSLP/skill_score/mean\endcsname{0.24}
\expandafter\def\csname mcresult@cell/full_results_mslp/tab_img_48/MSLP/skill_score/sd\endcsname{0.02}
\expandafter\def\csname mcresult@cell/full_results_mslp/tab_img_72/MSLP/r2/mean\endcsname{0.31}
\expandafter\def\csname mcresult@cell/full_results_mslp/tab_img_72/MSLP/r2/sd\endcsname{0.10}
\expandafter\def\csname mcresult@cell/full_results_mslp/tab_img_72/MSLP/rmse/mean\endcsname{17.69}
\expandafter\def\csname mcresult@cell/full_results_mslp/tab_img_72/MSLP/rmse/sd\endcsname{1.21}
\expandafter\def\csname mcresult@cell/full_results_mslp/tab_img_72/MSLP/skill_score/mean\endcsname{0.17}
\expandafter\def\csname mcresult@cell/full_results_mslp/tab_img_72/MSLP/skill_score/sd\endcsname{0.06}
\expandafter\def\csname mcresult@cell/full_results_mslp/tab_img_96/MSLP/r2/mean\endcsname{0.20}
\expandafter\def\csname mcresult@cell/full_results_mslp/tab_img_96/MSLP/r2/sd\endcsname{0.06}
\expandafter\def\csname mcresult@cell/full_results_mslp/tab_img_96/MSLP/rmse/mean\endcsname{18.83}
\expandafter\def\csname mcresult@cell/full_results_mslp/tab_img_96/MSLP/rmse/sd\endcsname{0.75}
\expandafter\def\csname mcresult@cell/full_results_mslp/tab_img_96/MSLP/skill_score/mean\endcsname{0.11}
\expandafter\def\csname mcresult@cell/full_results_mslp/tab_img_96/MSLP/skill_score/sd\endcsname{0.04}
\expandafter\def\csname mcresult@cell/full_results_vmax/disc_120/VMAX/r2/mean\endcsname{0.00}
\expandafter\def\csname mcresult@cell/full_results_vmax/disc_120/VMAX/r2/sd\endcsname{0.15}
\expandafter\def\csname mcresult@cell/full_results_vmax/disc_120/VMAX/rmse/mean\endcsname{29.33}
\expandafter\def\csname mcresult@cell/full_results_vmax/disc_120/VMAX/rmse/sd\endcsname{2.15}
\expandafter\def\csname mcresult@cell/full_results_vmax/disc_120/VMAX/skill_score/mean\endcsname{0.00}
\expandafter\def\csname mcresult@cell/full_results_vmax/disc_120/VMAX/skill_score/sd\endcsname{0.07}
\expandafter\def\csname mcresult@cell/full_results_vmax/disc_24/VMAX/r2/mean\endcsname{0.53}
\expandafter\def\csname mcresult@cell/full_results_vmax/disc_24/VMAX/r2/sd\endcsname{0.01}
\expandafter\def\csname mcresult@cell/full_results_vmax/disc_24/VMAX/rmse/mean\endcsname{13.27}
\expandafter\def\csname mcresult@cell/full_results_vmax/disc_24/VMAX/rmse/sd\endcsname{0.17}
\expandafter\def\csname mcresult@cell/full_results_vmax/disc_24/VMAX/skill_score/mean\endcsname{0.31}
\expandafter\def\csname mcresult@cell/full_results_vmax/disc_24/VMAX/skill_score/sd\endcsname{0.01}
\expandafter\def\csname mcresult@cell/full_results_vmax/disc_48/VMAX/r2/mean\endcsname{0.54}
\expandafter\def\csname mcresult@cell/full_results_vmax/disc_48/VMAX/r2/sd\endcsname{0.01}
\expandafter\def\csname mcresult@cell/full_results_vmax/disc_48/VMAX/rmse/mean\endcsname{19.23}
\expandafter\def\csname mcresult@cell/full_results_vmax/disc_48/VMAX/rmse/sd\endcsname{0.25}
\expandafter\def\csname mcresult@cell/full_results_vmax/disc_48/VMAX/skill_score/mean\endcsname{0.32}
\expandafter\def\csname mcresult@cell/full_results_vmax/disc_48/VMAX/skill_score/sd\endcsname{0.01}
\expandafter\def\csname mcresult@cell/full_results_vmax/disc_72/VMAX/r2/mean\endcsname{0.42}
\expandafter\def\csname mcresult@cell/full_results_vmax/disc_72/VMAX/r2/sd\endcsname{0.06}
\expandafter\def\csname mcresult@cell/full_results_vmax/disc_72/VMAX/rmse/mean\endcsname{23.19}
\expandafter\def\csname mcresult@cell/full_results_vmax/disc_72/VMAX/rmse/sd\endcsname{1.12}
\expandafter\def\csname mcresult@cell/full_results_vmax/disc_72/VMAX/skill_score/mean\endcsname{0.24}
\expandafter\def\csname mcresult@cell/full_results_vmax/disc_72/VMAX/skill_score/sd\endcsname{0.04}
\expandafter\def\csname mcresult@cell/full_results_vmax/disc_96/VMAX/r2/mean\endcsname{0.19}
\expandafter\def\csname mcresult@cell/full_results_vmax/disc_96/VMAX/r2/sd\endcsname{0.11}
\expandafter\def\csname mcresult@cell/full_results_vmax/disc_96/VMAX/rmse/mean\endcsname{26.29}
\expandafter\def\csname mcresult@cell/full_results_vmax/disc_96/VMAX/rmse/sd\endcsname{1.76}
\expandafter\def\csname mcresult@cell/full_results_vmax/disc_96/VMAX/skill_score/mean\endcsname{0.10}
\expandafter\def\csname mcresult@cell/full_results_vmax/disc_96/VMAX/skill_score/sd\endcsname{0.06}
\expandafter\def\csname mcresult@cell/full_results_vmax/full_120/VMAX/r2/mean\endcsname{-0.00}
\expandafter\def\csname mcresult@cell/full_results_vmax/full_120/VMAX/r2/sd\endcsname{0.13}
\expandafter\def\csname mcresult@cell/full_results_vmax/full_120/VMAX/rmse/mean\endcsname{29.29}
\expandafter\def\csname mcresult@cell/full_results_vmax/full_120/VMAX/rmse/sd\endcsname{1.94}
\expandafter\def\csname mcresult@cell/full_results_vmax/full_120/VMAX/skill_score/mean\endcsname{0.00}
\expandafter\def\csname mcresult@cell/full_results_vmax/full_120/VMAX/skill_score/sd\endcsname{0.07}
\expandafter\def\csname mcresult@cell/full_results_vmax/full_24/VMAX/r2/mean\endcsname{0.55}
\expandafter\def\csname mcresult@cell/full_results_vmax/full_24/VMAX/r2/sd\endcsname{0.02}
\expandafter\def\csname mcresult@cell/full_results_vmax/full_24/VMAX/rmse/mean\endcsname{12.77}
\expandafter\def\csname mcresult@cell/full_results_vmax/full_24/VMAX/rmse/sd\endcsname{0.35}
\expandafter\def\csname mcresult@cell/full_results_vmax/full_24/VMAX/skill_score/mean\endcsname{0.33}
\expandafter\def\csname mcresult@cell/full_results_vmax/full_24/VMAX/skill_score/sd\endcsname{0.02}
\expandafter\def\csname mcresult@cell/full_results_vmax/full_48/VMAX/r2/mean\endcsname{0.57}
\expandafter\def\csname mcresult@cell/full_results_vmax/full_48/VMAX/r2/sd\endcsname{0.01}
\expandafter\def\csname mcresult@cell/full_results_vmax/full_48/VMAX/rmse/mean\endcsname{18.72}
\expandafter\def\csname mcresult@cell/full_results_vmax/full_48/VMAX/rmse/sd\endcsname{0.26}
\expandafter\def\csname mcresult@cell/full_results_vmax/full_48/VMAX/skill_score/mean\endcsname{0.34}
\expandafter\def\csname mcresult@cell/full_results_vmax/full_48/VMAX/skill_score/sd\endcsname{0.01}
\expandafter\def\csname mcresult@cell/full_results_vmax/full_72/VMAX/r2/mean\endcsname{0.45}
\expandafter\def\csname mcresult@cell/full_results_vmax/full_72/VMAX/r2/sd\endcsname{0.04}
\expandafter\def\csname mcresult@cell/full_results_vmax/full_72/VMAX/rmse/mean\endcsname{22.51}
\expandafter\def\csname mcresult@cell/full_results_vmax/full_72/VMAX/rmse/sd\endcsname{0.94}
\expandafter\def\csname mcresult@cell/full_results_vmax/full_72/VMAX/skill_score/mean\endcsname{0.26}
\expandafter\def\csname mcresult@cell/full_results_vmax/full_72/VMAX/skill_score/sd\endcsname{0.03}
\expandafter\def\csname mcresult@cell/full_results_vmax/full_96/VMAX/r2/mean\endcsname{0.19}
\expandafter\def\csname mcresult@cell/full_results_vmax/full_96/VMAX/r2/sd\endcsname{0.08}
\expandafter\def\csname mcresult@cell/full_results_vmax/full_96/VMAX/rmse/mean\endcsname{26.27}
\expandafter\def\csname mcresult@cell/full_results_vmax/full_96/VMAX/rmse/sd\endcsname{1.37}
\expandafter\def\csname mcresult@cell/full_results_vmax/full_96/VMAX/skill_score/mean\endcsname{0.10}
\expandafter\def\csname mcresult@cell/full_results_vmax/full_96/VMAX/skill_score/sd\endcsname{0.05}
\expandafter\def\csname mcresult@cell/full_results_vmax/img_120/VMAX/r2/mean\endcsname{0.15}
\expandafter\def\csname mcresult@cell/full_results_vmax/img_120/VMAX/r2/sd\endcsname{0.04}
\expandafter\def\csname mcresult@cell/full_results_vmax/img_120/VMAX/rmse/mean\endcsname{26.95}
\expandafter\def\csname mcresult@cell/full_results_vmax/img_120/VMAX/rmse/sd\endcsname{0.62}
\expandafter\def\csname mcresult@cell/full_results_vmax/img_120/VMAX/skill_score/mean\endcsname{0.08}
\expandafter\def\csname mcresult@cell/full_results_vmax/img_120/VMAX/skill_score/sd\endcsname{0.02}
\expandafter\def\csname mcresult@cell/full_results_vmax/img_24/VMAX/r2/mean\endcsname{0.21}
\expandafter\def\csname mcresult@cell/full_results_vmax/img_24/VMAX/r2/sd\endcsname{0.03}
\expandafter\def\csname mcresult@cell/full_results_vmax/img_24/VMAX/rmse/mean\endcsname{16.93}
\expandafter\def\csname mcresult@cell/full_results_vmax/img_24/VMAX/rmse/sd\endcsname{0.29}
\expandafter\def\csname mcresult@cell/full_results_vmax/img_24/VMAX/skill_score/mean\endcsname{0.11}
\expandafter\def\csname mcresult@cell/full_results_vmax/img_24/VMAX/skill_score/sd\endcsname{0.02}
\expandafter\def\csname mcresult@cell/full_results_vmax/img_48/VMAX/r2/mean\endcsname{0.16}
\expandafter\def\csname mcresult@cell/full_results_vmax/img_48/VMAX/r2/sd\endcsname{0.06}
\expandafter\def\csname mcresult@cell/full_results_vmax/img_48/VMAX/rmse/mean\endcsname{26.03}
\expandafter\def\csname mcresult@cell/full_results_vmax/img_48/VMAX/rmse/sd\endcsname{0.95}
\expandafter\def\csname mcresult@cell/full_results_vmax/img_48/VMAX/skill_score/mean\endcsname{0.09}
\expandafter\def\csname mcresult@cell/full_results_vmax/img_48/VMAX/skill_score/sd\endcsname{0.03}
\expandafter\def\csname mcresult@cell/full_results_vmax/img_72/VMAX/r2/mean\endcsname{0.24}
\expandafter\def\csname mcresult@cell/full_results_vmax/img_72/VMAX/r2/sd\endcsname{0.06}
\expandafter\def\csname mcresult@cell/full_results_vmax/img_72/VMAX/rmse/mean\endcsname{26.44}
\expandafter\def\csname mcresult@cell/full_results_vmax/img_72/VMAX/rmse/sd\endcsname{1.04}
\expandafter\def\csname mcresult@cell/full_results_vmax/img_72/VMAX/skill_score/mean\endcsname{0.13}
\expandafter\def\csname mcresult@cell/full_results_vmax/img_72/VMAX/skill_score/sd\endcsname{0.03}
\expandafter\def\csname mcresult@cell/full_results_vmax/img_96/VMAX/r2/mean\endcsname{0.18}
\expandafter\def\csname mcresult@cell/full_results_vmax/img_96/VMAX/r2/sd\endcsname{0.07}
\expandafter\def\csname mcresult@cell/full_results_vmax/img_96/VMAX/rmse/mean\endcsname{26.50}
\expandafter\def\csname mcresult@cell/full_results_vmax/img_96/VMAX/rmse/sd\endcsname{1.16}
\expandafter\def\csname mcresult@cell/full_results_vmax/img_96/VMAX/skill_score/mean\endcsname{0.10}
\expandafter\def\csname mcresult@cell/full_results_vmax/img_96/VMAX/skill_score/sd\endcsname{0.04}
\expandafter\def\csname mcresult@cell/full_results_vmax/img_disc_120/VMAX/r2/mean\endcsname{0.05}
\expandafter\def\csname mcresult@cell/full_results_vmax/img_disc_120/VMAX/r2/sd\endcsname{0.18}
\expandafter\def\csname mcresult@cell/full_results_vmax/img_disc_120/VMAX/rmse/mean\endcsname{28.40}
\expandafter\def\csname mcresult@cell/full_results_vmax/img_disc_120/VMAX/rmse/sd\endcsname{2.81}
\expandafter\def\csname mcresult@cell/full_results_vmax/img_disc_120/VMAX/skill_score/mean\endcsname{0.03}
\expandafter\def\csname mcresult@cell/full_results_vmax/img_disc_120/VMAX/skill_score/sd\endcsname{0.10}
\expandafter\def\csname mcresult@cell/full_results_vmax/img_disc_24/VMAX/r2/mean\endcsname{0.47}
\expandafter\def\csname mcresult@cell/full_results_vmax/img_disc_24/VMAX/r2/sd\endcsname{0.03}
\expandafter\def\csname mcresult@cell/full_results_vmax/img_disc_24/VMAX/rmse/mean\endcsname{13.84}
\expandafter\def\csname mcresult@cell/full_results_vmax/img_disc_24/VMAX/rmse/sd\endcsname{0.36}
\expandafter\def\csname mcresult@cell/full_results_vmax/img_disc_24/VMAX/skill_score/mean\endcsname{0.27}
\expandafter\def\csname mcresult@cell/full_results_vmax/img_disc_24/VMAX/skill_score/sd\endcsname{0.02}
\expandafter\def\csname mcresult@cell/full_results_vmax/img_disc_48/VMAX/r2/mean\endcsname{0.52}
\expandafter\def\csname mcresult@cell/full_results_vmax/img_disc_48/VMAX/r2/sd\endcsname{0.03}
\expandafter\def\csname mcresult@cell/full_results_vmax/img_disc_48/VMAX/rmse/mean\endcsname{19.78}
\expandafter\def\csname mcresult@cell/full_results_vmax/img_disc_48/VMAX/rmse/sd\endcsname{0.54}
\expandafter\def\csname mcresult@cell/full_results_vmax/img_disc_48/VMAX/skill_score/mean\endcsname{0.31}
\expandafter\def\csname mcresult@cell/full_results_vmax/img_disc_48/VMAX/skill_score/sd\endcsname{0.02}
\expandafter\def\csname mcresult@cell/full_results_vmax/img_disc_72/VMAX/r2/mean\endcsname{0.39}
\expandafter\def\csname mcresult@cell/full_results_vmax/img_disc_72/VMAX/r2/sd\endcsname{0.10}
\expandafter\def\csname mcresult@cell/full_results_vmax/img_disc_72/VMAX/rmse/mean\endcsname{23.68}
\expandafter\def\csname mcresult@cell/full_results_vmax/img_disc_72/VMAX/rmse/sd\endcsname{1.86}
\expandafter\def\csname mcresult@cell/full_results_vmax/img_disc_72/VMAX/skill_score/mean\endcsname{0.22}
\expandafter\def\csname mcresult@cell/full_results_vmax/img_disc_72/VMAX/skill_score/sd\endcsname{0.06}
\expandafter\def\csname mcresult@cell/full_results_vmax/img_disc_96/VMAX/r2/mean\endcsname{0.08}
\expandafter\def\csname mcresult@cell/full_results_vmax/img_disc_96/VMAX/r2/sd\endcsname{0.19}
\expandafter\def\csname mcresult@cell/full_results_vmax/img_disc_96/VMAX/rmse/mean\endcsname{27.98}
\expandafter\def\csname mcresult@cell/full_results_vmax/img_disc_96/VMAX/rmse/sd\endcsname{2.93}
\expandafter\def\csname mcresult@cell/full_results_vmax/img_disc_96/VMAX/skill_score/mean\endcsname{0.05}
\expandafter\def\csname mcresult@cell/full_results_vmax/img_disc_96/VMAX/skill_score/sd\endcsname{0.10}
\expandafter\def\csname mcresult@cell/full_results_vmax/tab_120/VMAX/r2/mean\endcsname{0.21}
\expandafter\def\csname mcresult@cell/full_results_vmax/tab_120/VMAX/r2/sd\endcsname{0.03}
\expandafter\def\csname mcresult@cell/full_results_vmax/tab_120/VMAX/rmse/mean\endcsname{26.11}
\expandafter\def\csname mcresult@cell/full_results_vmax/tab_120/VMAX/rmse/sd\endcsname{0.43}
\expandafter\def\csname mcresult@cell/full_results_vmax/tab_120/VMAX/skill_score/mean\endcsname{0.11}
\expandafter\def\csname mcresult@cell/full_results_vmax/tab_120/VMAX/skill_score/sd\endcsname{0.01}
\expandafter\def\csname mcresult@cell/full_results_vmax/tab_24/VMAX/r2/mean\endcsname{0.44}
\expandafter\def\csname mcresult@cell/full_results_vmax/tab_24/VMAX/r2/sd\endcsname{0.01}
\expandafter\def\csname mcresult@cell/full_results_vmax/tab_24/VMAX/rmse/mean\endcsname{14.22}
\expandafter\def\csname mcresult@cell/full_results_vmax/tab_24/VMAX/rmse/sd\endcsname{0.12}
\expandafter\def\csname mcresult@cell/full_results_vmax/tab_24/VMAX/skill_score/mean\endcsname{0.25}
\expandafter\def\csname mcresult@cell/full_results_vmax/tab_24/VMAX/skill_score/sd\endcsname{0.01}
\expandafter\def\csname mcresult@cell/full_results_vmax/tab_48/VMAX/r2/mean\endcsname{0.47}
\expandafter\def\csname mcresult@cell/full_results_vmax/tab_48/VMAX/r2/sd\endcsname{0.02}
\expandafter\def\csname mcresult@cell/full_results_vmax/tab_48/VMAX/rmse/mean\endcsname{20.78}
\expandafter\def\csname mcresult@cell/full_results_vmax/tab_48/VMAX/rmse/sd\endcsname{0.38}
\expandafter\def\csname mcresult@cell/full_results_vmax/tab_48/VMAX/skill_score/mean\endcsname{0.27}
\expandafter\def\csname mcresult@cell/full_results_vmax/tab_48/VMAX/skill_score/sd\endcsname{0.01}
\expandafter\def\csname mcresult@cell/full_results_vmax/tab_72/VMAX/r2/mean\endcsname{0.39}
\expandafter\def\csname mcresult@cell/full_results_vmax/tab_72/VMAX/r2/sd\endcsname{0.02}
\expandafter\def\csname mcresult@cell/full_results_vmax/tab_72/VMAX/rmse/mean\endcsname{23.78}
\expandafter\def\csname mcresult@cell/full_results_vmax/tab_72/VMAX/rmse/sd\endcsname{0.47}
\expandafter\def\csname mcresult@cell/full_results_vmax/tab_72/VMAX/skill_score/mean\endcsname{0.22}
\expandafter\def\csname mcresult@cell/full_results_vmax/tab_72/VMAX/skill_score/sd\endcsname{0.02}
\expandafter\def\csname mcresult@cell/full_results_vmax/tab_96/VMAX/r2/mean\endcsname{0.27}
\expandafter\def\csname mcresult@cell/full_results_vmax/tab_96/VMAX/r2/sd\endcsname{0.03}
\expandafter\def\csname mcresult@cell/full_results_vmax/tab_96/VMAX/rmse/mean\endcsname{24.97}
\expandafter\def\csname mcresult@cell/full_results_vmax/tab_96/VMAX/rmse/sd\endcsname{0.59}
\expandafter\def\csname mcresult@cell/full_results_vmax/tab_96/VMAX/skill_score/mean\endcsname{0.15}
\expandafter\def\csname mcresult@cell/full_results_vmax/tab_96/VMAX/skill_score/sd\endcsname{0.02}
\expandafter\def\csname mcresult@cell/full_results_vmax/tab_disc_120/VMAX/r2/mean\endcsname{0.06}
\expandafter\def\csname mcresult@cell/full_results_vmax/tab_disc_120/VMAX/r2/sd\endcsname{0.07}
\expandafter\def\csname mcresult@cell/full_results_vmax/tab_disc_120/VMAX/rmse/mean\endcsname{28.43}
\expandafter\def\csname mcresult@cell/full_results_vmax/tab_disc_120/VMAX/rmse/sd\endcsname{1.13}
\expandafter\def\csname mcresult@cell/full_results_vmax/tab_disc_120/VMAX/skill_score/mean\endcsname{0.03}
\expandafter\def\csname mcresult@cell/full_results_vmax/tab_disc_120/VMAX/skill_score/sd\endcsname{0.04}
\expandafter\def\csname mcresult@cell/full_results_vmax/tab_disc_24/VMAX/r2/mean\endcsname{0.54}
\expandafter\def\csname mcresult@cell/full_results_vmax/tab_disc_24/VMAX/r2/sd\endcsname{0.04}
\expandafter\def\csname mcresult@cell/full_results_vmax/tab_disc_24/VMAX/rmse/mean\endcsname{12.92}
\expandafter\def\csname mcresult@cell/full_results_vmax/tab_disc_24/VMAX/rmse/sd\endcsname{0.50}
\expandafter\def\csname mcresult@cell/full_results_vmax/tab_disc_24/VMAX/skill_score/mean\endcsname{0.32}
\expandafter\def\csname mcresult@cell/full_results_vmax/tab_disc_24/VMAX/skill_score/sd\endcsname{0.03}
\expandafter\def\csname mcresult@cell/full_results_vmax/tab_disc_48/VMAX/r2/mean\endcsname{0.58}
\expandafter\def\csname mcresult@cell/full_results_vmax/tab_disc_48/VMAX/r2/sd\endcsname{0.02}
\expandafter\def\csname mcresult@cell/full_results_vmax/tab_disc_48/VMAX/rmse/mean\endcsname{18.43}
\expandafter\def\csname mcresult@cell/full_results_vmax/tab_disc_48/VMAX/rmse/sd\endcsname{0.36}
\expandafter\def\csname mcresult@cell/full_results_vmax/tab_disc_48/VMAX/skill_score/mean\endcsname{0.35}
\expandafter\def\csname mcresult@cell/full_results_vmax/tab_disc_48/VMAX/skill_score/sd\endcsname{0.01}
\expandafter\def\csname mcresult@cell/full_results_vmax/tab_disc_72/VMAX/r2/mean\endcsname{0.46}
\expandafter\def\csname mcresult@cell/full_results_vmax/tab_disc_72/VMAX/r2/sd\endcsname{0.06}
\expandafter\def\csname mcresult@cell/full_results_vmax/tab_disc_72/VMAX/rmse/mean\endcsname{22.39}
\expandafter\def\csname mcresult@cell/full_results_vmax/tab_disc_72/VMAX/rmse/sd\endcsname{1.22}
\expandafter\def\csname mcresult@cell/full_results_vmax/tab_disc_72/VMAX/skill_score/mean\endcsname{0.26}
\expandafter\def\csname mcresult@cell/full_results_vmax/tab_disc_72/VMAX/skill_score/sd\endcsname{0.04}
\expandafter\def\csname mcresult@cell/full_results_vmax/tab_disc_96/VMAX/r2/mean\endcsname{0.24}
\expandafter\def\csname mcresult@cell/full_results_vmax/tab_disc_96/VMAX/r2/sd\endcsname{0.07}
\expandafter\def\csname mcresult@cell/full_results_vmax/tab_disc_96/VMAX/rmse/mean\endcsname{25.56}
\expandafter\def\csname mcresult@cell/full_results_vmax/tab_disc_96/VMAX/rmse/sd\endcsname{1.22}
\expandafter\def\csname mcresult@cell/full_results_vmax/tab_disc_96/VMAX/skill_score/mean\endcsname{0.13}
\expandafter\def\csname mcresult@cell/full_results_vmax/tab_disc_96/VMAX/skill_score/sd\endcsname{0.04}
\expandafter\def\csname mcresult@cell/full_results_vmax/tab_img_120/VMAX/r2/mean\endcsname{0.08}
\expandafter\def\csname mcresult@cell/full_results_vmax/tab_img_120/VMAX/r2/sd\endcsname{0.09}
\expandafter\def\csname mcresult@cell/full_results_vmax/tab_img_120/VMAX/rmse/mean\endcsname{28.02}
\expandafter\def\csname mcresult@cell/full_results_vmax/tab_img_120/VMAX/rmse/sd\endcsname{1.40}
\expandafter\def\csname mcresult@cell/full_results_vmax/tab_img_120/VMAX/skill_score/mean\endcsname{0.04}
\expandafter\def\csname mcresult@cell/full_results_vmax/tab_img_120/VMAX/skill_score/sd\endcsname{0.05}
\expandafter\def\csname mcresult@cell/full_results_vmax/tab_img_24/VMAX/r2/mean\endcsname{0.46}
\expandafter\def\csname mcresult@cell/full_results_vmax/tab_img_24/VMAX/r2/sd\endcsname{0.02}
\expandafter\def\csname mcresult@cell/full_results_vmax/tab_img_24/VMAX/rmse/mean\endcsname{13.98}
\expandafter\def\csname mcresult@cell/full_results_vmax/tab_img_24/VMAX/rmse/sd\endcsname{0.30}
\expandafter\def\csname mcresult@cell/full_results_vmax/tab_img_24/VMAX/skill_score/mean\endcsname{0.27}
\expandafter\def\csname mcresult@cell/full_results_vmax/tab_img_24/VMAX/skill_score/sd\endcsname{0.02}
\expandafter\def\csname mcresult@cell/full_results_vmax/tab_img_48/VMAX/r2/mean\endcsname{0.45}
\expandafter\def\csname mcresult@cell/full_results_vmax/tab_img_48/VMAX/r2/sd\endcsname{0.02}
\expandafter\def\csname mcresult@cell/full_results_vmax/tab_img_48/VMAX/rmse/mean\endcsname{21.14}
\expandafter\def\csname mcresult@cell/full_results_vmax/tab_img_48/VMAX/rmse/sd\endcsname{0.34}
\expandafter\def\csname mcresult@cell/full_results_vmax/tab_img_48/VMAX/skill_score/mean\endcsname{0.26}
\expandafter\def\csname mcresult@cell/full_results_vmax/tab_img_48/VMAX/skill_score/sd\endcsname{0.01}
\expandafter\def\csname mcresult@cell/full_results_vmax/tab_img_72/VMAX/r2/mean\endcsname{0.41}
\expandafter\def\csname mcresult@cell/full_results_vmax/tab_img_72/VMAX/r2/sd\endcsname{0.03}
\expandafter\def\csname mcresult@cell/full_results_vmax/tab_img_72/VMAX/rmse/mean\endcsname{23.37}
\expandafter\def\csname mcresult@cell/full_results_vmax/tab_img_72/VMAX/rmse/sd\endcsname{0.64}
\expandafter\def\csname mcresult@cell/full_results_vmax/tab_img_72/VMAX/skill_score/mean\endcsname{0.23}
\expandafter\def\csname mcresult@cell/full_results_vmax/tab_img_72/VMAX/skill_score/sd\endcsname{0.02}
\expandafter\def\csname mcresult@cell/full_results_vmax/tab_img_96/VMAX/r2/mean\endcsname{0.28}
\expandafter\def\csname mcresult@cell/full_results_vmax/tab_img_96/VMAX/r2/sd\endcsname{0.08}
\expandafter\def\csname mcresult@cell/full_results_vmax/tab_img_96/VMAX/rmse/mean\endcsname{24.81}
\expandafter\def\csname mcresult@cell/full_results_vmax/tab_img_96/VMAX/rmse/sd\endcsname{1.38}
\expandafter\def\csname mcresult@cell/full_results_vmax/tab_img_96/VMAX/skill_score/mean\endcsname{0.15}
\expandafter\def\csname mcresult@cell/full_results_vmax/tab_img_96/VMAX/skill_score/sd\endcsname{0.05}
\expandafter\def\csname mcresult@cell/hurricast_ablation_24h/CNN_Tab_Text_emb/MSLP/mae/mean\endcsname{6.43}
\expandafter\def\csname mcresult@cell/hurricast_ablation_24h/CNN_Tab_Text_emb/MSLP/mae/sd\endcsname{0.18}
\expandafter\def\csname mcresult@cell/hurricast_ablation_24h/CNN_Tab_Text_emb/MSLP/r2/mean\endcsname{0.47}
\expandafter\def\csname mcresult@cell/hurricast_ablation_24h/CNN_Tab_Text_emb/MSLP/r2/sd\endcsname{0.02}
\expandafter\def\csname mcresult@cell/hurricast_ablation_24h/CNN_Tab_Text_emb/VMAX/dir_acc/mean\endcsname{0.664}
\expandafter\def\csname mcresult@cell/hurricast_ablation_24h/CNN_Tab_Text_emb/VMAX/dir_acc/sd\endcsname{0.015}
\expandafter\def\csname mcresult@cell/hurricast_ablation_24h/CNN_Tab_Text_emb/VMAX/mae/mean\endcsname{9.71}
\expandafter\def\csname mcresult@cell/hurricast_ablation_24h/CNN_Tab_Text_emb/VMAX/mae/sd\endcsname{0.27}
\expandafter\def\csname mcresult@cell/hurricast_ablation_24h/CNN_Tab_Text_emb/VMAX/r2/mean\endcsname{0.45}
\expandafter\def\csname mcresult@cell/hurricast_ablation_24h/CNN_Tab_Text_emb/VMAX/r2/sd\endcsname{0.02}
\expandafter\def\csname mcresult@cell/hurricast_ablation_24h/CNN_Tab_emb/MSLP/mae/mean\endcsname{7.05}
\expandafter\def\csname mcresult@cell/hurricast_ablation_24h/CNN_Tab_emb/MSLP/mae/sd\endcsname{0.10}
\expandafter\def\csname mcresult@cell/hurricast_ablation_24h/CNN_Tab_emb/MSLP/r2/mean\endcsname{0.37}
\expandafter\def\csname mcresult@cell/hurricast_ablation_24h/CNN_Tab_emb/MSLP/r2/sd\endcsname{0.03}
\expandafter\def\csname mcresult@cell/hurricast_ablation_24h/CNN_Tab_emb/VMAX/dir_acc/mean\endcsname{0.648}
\expandafter\def\csname mcresult@cell/hurricast_ablation_24h/CNN_Tab_emb/VMAX/dir_acc/sd\endcsname{0.004}
\expandafter\def\csname mcresult@cell/hurricast_ablation_24h/CNN_Tab_emb/VMAX/mae/mean\endcsname{10.56}
\expandafter\def\csname mcresult@cell/hurricast_ablation_24h/CNN_Tab_emb/VMAX/mae/sd\endcsname{0.11}
\expandafter\def\csname mcresult@cell/hurricast_ablation_24h/CNN_Tab_emb/VMAX/r2/mean\endcsname{0.36}
\expandafter\def\csname mcresult@cell/hurricast_ablation_24h/CNN_Tab_emb/VMAX/r2/sd\endcsname{0.01}
\expandafter\def\csname mcresult@cell/hurricast_ablation_24h/CNN_Text_emb/MSLP/mae/mean\endcsname{7.00}
\expandafter\def\csname mcresult@cell/hurricast_ablation_24h/CNN_Text_emb/MSLP/mae/sd\endcsname{0.16}
\expandafter\def\csname mcresult@cell/hurricast_ablation_24h/CNN_Text_emb/MSLP/r2/mean\endcsname{0.36}
\expandafter\def\csname mcresult@cell/hurricast_ablation_24h/CNN_Text_emb/MSLP/r2/sd\endcsname{0.02}
\expandafter\def\csname mcresult@cell/hurricast_ablation_24h/CNN_Text_emb/VMAX/dir_acc/mean\endcsname{0.641}
\expandafter\def\csname mcresult@cell/hurricast_ablation_24h/CNN_Text_emb/VMAX/dir_acc/sd\endcsname{0.021}
\expandafter\def\csname mcresult@cell/hurricast_ablation_24h/CNN_Text_emb/VMAX/mae/mean\endcsname{10.42}
\expandafter\def\csname mcresult@cell/hurricast_ablation_24h/CNN_Text_emb/VMAX/mae/sd\endcsname{0.19}
\expandafter\def\csname mcresult@cell/hurricast_ablation_24h/CNN_Text_emb/VMAX/r2/mean\endcsname{0.36}
\expandafter\def\csname mcresult@cell/hurricast_ablation_24h/CNN_Text_emb/VMAX/r2/sd\endcsname{0.02}
\expandafter\def\csname mcresult@cell/hurricast_ablation_24h/CNN_only_emb/MSLP/mae/mean\endcsname{8.21}
\expandafter\def\csname mcresult@cell/hurricast_ablation_24h/CNN_only_emb/MSLP/mae/sd\endcsname{0.28}
\expandafter\def\csname mcresult@cell/hurricast_ablation_24h/CNN_only_emb/MSLP/r2/mean\endcsname{0.16}
\expandafter\def\csname mcresult@cell/hurricast_ablation_24h/CNN_only_emb/MSLP/r2/sd\endcsname{0.04}
\expandafter\def\csname mcresult@cell/hurricast_ablation_24h/CNN_only_emb/VMAX/dir_acc/mean\endcsname{0.591}
\expandafter\def\csname mcresult@cell/hurricast_ablation_24h/CNN_only_emb/VMAX/dir_acc/sd\endcsname{0.013}
\expandafter\def\csname mcresult@cell/hurricast_ablation_24h/CNN_only_emb/VMAX/mae/mean\endcsname{11.91}
\expandafter\def\csname mcresult@cell/hurricast_ablation_24h/CNN_only_emb/VMAX/mae/sd\endcsname{0.30}
\expandafter\def\csname mcresult@cell/hurricast_ablation_24h/CNN_only_emb/VMAX/r2/mean\endcsname{0.18}
\expandafter\def\csname mcresult@cell/hurricast_ablation_24h/CNN_only_emb/VMAX/r2/sd\endcsname{0.03}
\expandafter\def\csname mcresult@cell/hurricast_ablation_24h/NN_direct/MSLP/mae/mean\endcsname{6.63}
\expandafter\def\csname mcresult@cell/hurricast_ablation_24h/NN_direct/MSLP/mae/sd\endcsname{0.13}
\expandafter\def\csname mcresult@cell/hurricast_ablation_24h/NN_direct/MSLP/r2/mean\endcsname{0.45}
\expandafter\def\csname mcresult@cell/hurricast_ablation_24h/NN_direct/MSLP/r2/sd\endcsname{0.01}
\expandafter\def\csname mcresult@cell/hurricast_ablation_24h/NN_direct/VMAX/dir_acc/mean\endcsname{0.656}
\expandafter\def\csname mcresult@cell/hurricast_ablation_24h/NN_direct/VMAX/dir_acc/sd\endcsname{0.015}
\expandafter\def\csname mcresult@cell/hurricast_ablation_24h/NN_direct/VMAX/mae/mean\endcsname{9.92}
\expandafter\def\csname mcresult@cell/hurricast_ablation_24h/NN_direct/VMAX/mae/sd\endcsname{0.27}
\expandafter\def\csname mcresult@cell/hurricast_ablation_24h/NN_direct/VMAX/r2/mean\endcsname{0.44}
\expandafter\def\csname mcresult@cell/hurricast_ablation_24h/NN_direct/VMAX/r2/sd\endcsname{0.02}
\expandafter\def\csname mcresult@cell/hurricast_ablation_24h/XGB_ALL/MSLP/mae/mean\endcsname{6.41}
\expandafter\def\csname mcresult@cell/hurricast_ablation_24h/XGB_ALL/MSLP/mae/sd\endcsname{0.17}
\expandafter\def\csname mcresult@cell/hurricast_ablation_24h/XGB_ALL/MSLP/r2/mean\endcsname{0.47}
\expandafter\def\csname mcresult@cell/hurricast_ablation_24h/XGB_ALL/MSLP/r2/sd\endcsname{0.02}
\expandafter\def\csname mcresult@cell/hurricast_ablation_24h/XGB_ALL/VMAX/dir_acc/mean\endcsname{0.666}
\expandafter\def\csname mcresult@cell/hurricast_ablation_24h/XGB_ALL/VMAX/dir_acc/sd\endcsname{0.010}
\expandafter\def\csname mcresult@cell/hurricast_ablation_24h/XGB_ALL/VMAX/mae/mean\endcsname{9.68}
\expandafter\def\csname mcresult@cell/hurricast_ablation_24h/XGB_ALL/VMAX/mae/sd\endcsname{0.21}
\expandafter\def\csname mcresult@cell/hurricast_ablation_24h/XGB_ALL/VMAX/r2/mean\endcsname{0.46}
\expandafter\def\csname mcresult@cell/hurricast_ablation_24h/XGB_ALL/VMAX/r2/sd\endcsname{0.02}
\expandafter\def\csname mcresult@cell/leadtime_diagnostics/120/joint_mse/mean\endcsname{811.70}
\expandafter\def\csname mcresult@cell/leadtime_diagnostics/120/joint_mse/sd\endcsname{83.60}
\expandafter\def\csname mcresult@cell/leadtime_diagnostics/120/joint_r2/mean\endcsname{0.06}
\expandafter\def\csname mcresult@cell/leadtime_diagnostics/120/joint_r2/sd\endcsname{0.10}
\expandafter\def\csname mcresult@cell/leadtime_diagnostics/120/variance_mslp_ddof0/mean\endcsname{477.61}
\expandafter\def\csname mcresult@cell/leadtime_diagnostics/120/variance_vmax_ddof0/mean\endcsname{859.05}
\expandafter\def\csname mcresult@cell/leadtime_diagnostics/24/joint_mse/mean\endcsname{167.29}
\expandafter\def\csname mcresult@cell/leadtime_diagnostics/24/joint_mse/sd\endcsname{8.43}
\expandafter\def\csname mcresult@cell/leadtime_diagnostics/24/joint_r2/mean\endcsname{0.54}
\expandafter\def\csname mcresult@cell/leadtime_diagnostics/24/joint_r2/sd\endcsname{0.02}
\expandafter\def\csname mcresult@cell/leadtime_diagnostics/24/variance_mslp_ddof0/mean\endcsname{169.93}
\expandafter\def\csname mcresult@cell/leadtime_diagnostics/24/variance_vmax_ddof0/mean\endcsname{362.24}
\expandafter\def\csname mcresult@cell/leadtime_diagnostics/48/joint_mse/mean\endcsname{362.97}
\expandafter\def\csname mcresult@cell/leadtime_diagnostics/48/joint_mse/sd\endcsname{12.51}
\expandafter\def\csname mcresult@cell/leadtime_diagnostics/48/joint_r2/mean\endcsname{0.55}
\expandafter\def\csname mcresult@cell/leadtime_diagnostics/48/joint_r2/sd\endcsname{0.02}
\expandafter\def\csname mcresult@cell/leadtime_diagnostics/48/variance_mslp_ddof0/mean\endcsname{388.25}
\expandafter\def\csname mcresult@cell/leadtime_diagnostics/48/variance_vmax_ddof0/mean\endcsname{811.35}
\expandafter\def\csname mcresult@cell/leadtime_diagnostics/72/joint_mse/mean\endcsname{518.43}
\expandafter\def\csname mcresult@cell/leadtime_diagnostics/72/joint_mse/sd\endcsname{25.92}
\expandafter\def\csname mcresult@cell/leadtime_diagnostics/72/joint_r2/mean\endcsname{0.44}
\expandafter\def\csname mcresult@cell/leadtime_diagnostics/72/joint_r2/sd\endcsname{0.03}
\expandafter\def\csname mcresult@cell/leadtime_diagnostics/72/variance_mslp_ddof0/mean\endcsname{456.85}
\expandafter\def\csname mcresult@cell/leadtime_diagnostics/72/variance_vmax_ddof0/mean\endcsname{925.55}
\expandafter\def\csname mcresult@cell/leadtime_diagnostics/96/joint_mse/mean\endcsname{683.29}
\expandafter\def\csname mcresult@cell/leadtime_diagnostics/96/joint_mse/sd\endcsname{53.18}
\expandafter\def\csname mcresult@cell/leadtime_diagnostics/96/joint_r2/mean\endcsname{0.20}
\expandafter\def\csname mcresult@cell/leadtime_diagnostics/96/joint_r2/sd\endcsname{0.06}
\expandafter\def\csname mcresult@cell/leadtime_diagnostics/96/variance_mslp_ddof0/mean\endcsname{446.34}
\expandafter\def\csname mcresult@cell/leadtime_diagnostics/96/variance_vmax_ddof0/mean\endcsname{859.42}
\expandafter\def\csname mcresult@cell/main_results_24h/disc/MSLP/r2/mean\endcsname{0.46}
\expandafter\def\csname mcresult@cell/main_results_24h/disc/MSLP/r2/sd\endcsname{0.03}
\expandafter\def\csname mcresult@cell/main_results_24h/disc/VMAX/r2/mean\endcsname{0.50}
\expandafter\def\csname mcresult@cell/main_results_24h/disc/VMAX/r2/sd\endcsname{0.01}
\expandafter\def\csname mcresult@cell/main_results_24h/full/MSLP/r2/mean\endcsname{0.51}
\expandafter\def\csname mcresult@cell/main_results_24h/full/MSLP/r2/sd\endcsname{0.01}
\expandafter\def\csname mcresult@cell/main_results_24h/full/VMAX/r2/mean\endcsname{0.53}
\expandafter\def\csname mcresult@cell/main_results_24h/full/VMAX/r2/sd\endcsname{0.01}
\expandafter\def\csname mcresult@cell/main_results_24h/img/MSLP/r2/mean\endcsname{0.20}
\expandafter\def\csname mcresult@cell/main_results_24h/img/MSLP/r2/sd\endcsname{0.02}
\expandafter\def\csname mcresult@cell/main_results_24h/img/VMAX/r2/mean\endcsname{0.22}
\expandafter\def\csname mcresult@cell/main_results_24h/img/VMAX/r2/sd\endcsname{0.02}
\expandafter\def\csname mcresult@cell/main_results_24h/img_disc/MSLP/r2/mean\endcsname{0.48}
\expandafter\def\csname mcresult@cell/main_results_24h/img_disc/MSLP/r2/sd\endcsname{0.02}
\expandafter\def\csname mcresult@cell/main_results_24h/img_disc/VMAX/r2/mean\endcsname{0.50}
\expandafter\def\csname mcresult@cell/main_results_24h/img_disc/VMAX/r2/sd\endcsname{0.01}
\expandafter\def\csname mcresult@cell/main_results_24h/tab/MSLP/r2/mean\endcsname{0.44}
\expandafter\def\csname mcresult@cell/main_results_24h/tab/MSLP/r2/sd\endcsname{0.01}
\expandafter\def\csname mcresult@cell/main_results_24h/tab/VMAX/r2/mean\endcsname{0.44}
\expandafter\def\csname mcresult@cell/main_results_24h/tab/VMAX/r2/sd\endcsname{0.01}
\expandafter\def\csname mcresult@cell/main_results_24h/tab_disc/MSLP/r2/mean\endcsname{0.51}
\expandafter\def\csname mcresult@cell/main_results_24h/tab_disc/MSLP/r2/sd\endcsname{0.01}
\expandafter\def\csname mcresult@cell/main_results_24h/tab_disc/VMAX/r2/mean\endcsname{0.53}
\expandafter\def\csname mcresult@cell/main_results_24h/tab_disc/VMAX/r2/sd\endcsname{0.01}
\expandafter\def\csname mcresult@cell/main_results_24h/tab_img/MSLP/r2/mean\endcsname{0.45}
\expandafter\def\csname mcresult@cell/main_results_24h/tab_img/MSLP/r2/sd\endcsname{0.01}
\expandafter\def\csname mcresult@cell/main_results_24h/tab_img/VMAX/r2/mean\endcsname{0.45}
\expandafter\def\csname mcresult@cell/main_results_24h/tab_img/VMAX/r2/sd\endcsname{0.01}
\expandafter\def\csname mcresult@cell/missingness_sensitivity/all/left_maeMSLP/mean\endcsname{6.43}
\expandafter\def\csname mcresult@cell/missingness_sensitivity/all/left_maeVMAX/mean\endcsname{9.89}
\expandafter\def\csname mcresult@cell/missingness_sensitivity/all/mae_gain_left_to_rightMSLP/mean\endcsname{0.57}
\expandafter\def\csname mcresult@cell/missingness_sensitivity/all/mae_gain_left_to_rightVMAX/mean\endcsname{0.99}
\expandafter\def\csname mcresult@cell/missingness_sensitivity/all/right_maeMSLP/mean\endcsname{5.87}
\expandafter\def\csname mcresult@cell/missingness_sensitivity/all/right_maeVMAX/mean\endcsname{8.90}
\expandafter\def\csname mcresult@cell/missingness_sensitivity/allMSLP/storm_ci_high/mean\endcsname{1.05}
\expandafter\def\csname mcresult@cell/missingness_sensitivity/allMSLP/storm_ci_low/mean\endcsname{0.07}
\expandafter\def\csname mcresult@cell/missingness_sensitivity/allVMAX/storm_ci_high/mean\endcsname{1.70}
\expandafter\def\csname mcresult@cell/missingness_sensitivity/allVMAX/storm_ci_low/mean\endcsname{0.32}
\expandafter\def\csname mcresult@cell/missingness_sensitivity/at_or_after_first_landfall/left_maeMSLP/mean\endcsname{9.27}
\expandafter\def\csname mcresult@cell/missingness_sensitivity/at_or_after_first_landfall/left_maeVMAX/mean\endcsname{13.92}
\expandafter\def\csname mcresult@cell/missingness_sensitivity/at_or_after_first_landfall/mae_gain_left_to_rightMSLP/mean\endcsname{2.05}
\expandafter\def\csname mcresult@cell/missingness_sensitivity/at_or_after_first_landfall/mae_gain_left_to_rightVMAX/mean\endcsname{3.59}
\expandafter\def\csname mcresult@cell/missingness_sensitivity/at_or_after_first_landfall/right_maeMSLP/mean\endcsname{7.22}
\expandafter\def\csname mcresult@cell/missingness_sensitivity/at_or_after_first_landfall/right_maeVMAX/mean\endcsname{10.33}
\expandafter\def\csname mcresult@cell/missingness_sensitivity/at_or_after_first_landfallMSLP/storm_ci_high/mean\endcsname{5.95}
\expandafter\def\csname mcresult@cell/missingness_sensitivity/at_or_after_first_landfallMSLP/storm_ci_low/mean\endcsname{-0.22}
\expandafter\def\csname mcresult@cell/missingness_sensitivity/at_or_after_first_landfallVMAX/storm_ci_high/mean\endcsname{9.97}
\expandafter\def\csname mcresult@cell/missingness_sensitivity/at_or_after_first_landfallVMAX/storm_ci_low/mean\endcsname{-0.02}
\expandafter\def\csname mcresult@cell/missingness_sensitivity/discussion_missing/left_maeMSLP/mean\endcsname{3.43}
\expandafter\def\csname mcresult@cell/missingness_sensitivity/discussion_missing/left_maeVMAX/mean\endcsname{5.94}
\expandafter\def\csname mcresult@cell/missingness_sensitivity/discussion_missing/mae_gain_left_to_rightMSLP/mean\endcsname{-0.54}
\expandafter\def\csname mcresult@cell/missingness_sensitivity/discussion_missing/mae_gain_left_to_rightVMAX/mean\endcsname{-0.95}
\expandafter\def\csname mcresult@cell/missingness_sensitivity/discussion_missing/right_maeMSLP/mean\endcsname{3.97}
\expandafter\def\csname mcresult@cell/missingness_sensitivity/discussion_missing/right_maeVMAX/mean\endcsname{6.90}
\expandafter\def\csname mcresult@cell/missingness_sensitivity/discussion_missingMSLP/storm_ci_high/mean\endcsname{0.40}
\expandafter\def\csname mcresult@cell/missingness_sensitivity/discussion_missingMSLP/storm_ci_low/mean\endcsname{-1.65}
\expandafter\def\csname mcresult@cell/missingness_sensitivity/discussion_missingVMAX/storm_ci_high/mean\endcsname{0.62}
\expandafter\def\csname mcresult@cell/missingness_sensitivity/discussion_missingVMAX/storm_ci_low/mean\endcsname{-2.64}
\expandafter\def\csname mcresult@cell/missingness_sensitivity/discussion_present/left_maeMSLP/mean\endcsname{6.55}
\expandafter\def\csname mcresult@cell/missingness_sensitivity/discussion_present/left_maeVMAX/mean\endcsname{10.05}
\expandafter\def\csname mcresult@cell/missingness_sensitivity/discussion_present/mae_gain_left_to_rightMSLP/mean\endcsname{0.61}
\expandafter\def\csname mcresult@cell/missingness_sensitivity/discussion_present/mae_gain_left_to_rightVMAX/mean\endcsname{1.07}
\expandafter\def\csname mcresult@cell/missingness_sensitivity/discussion_present/right_maeMSLP/mean\endcsname{5.94}
\expandafter\def\csname mcresult@cell/missingness_sensitivity/discussion_present/right_maeVMAX/mean\endcsname{8.98}
\expandafter\def\csname mcresult@cell/missingness_sensitivity/discussion_presentMSLP/storm_ci_high/mean\endcsname{1.12}
\expandafter\def\csname mcresult@cell/missingness_sensitivity/discussion_presentMSLP/storm_ci_low/mean\endcsname{0.10}
\expandafter\def\csname mcresult@cell/missingness_sensitivity/discussion_presentVMAX/storm_ci_high/mean\endcsname{1.82}
\expandafter\def\csname mcresult@cell/missingness_sensitivity/discussion_presentVMAX/storm_ci_low/mean\endcsname{0.37}
\expandafter\def\csname mcresult@cell/missingness_sensitivity/pre_landfall/left_maeMSLP/mean\endcsname{6.13}
\expandafter\def\csname mcresult@cell/missingness_sensitivity/pre_landfall/left_maeVMAX/mean\endcsname{9.46}
\expandafter\def\csname mcresult@cell/missingness_sensitivity/pre_landfall/mae_gain_left_to_rightMSLP/mean\endcsname{0.41}
\expandafter\def\csname mcresult@cell/missingness_sensitivity/pre_landfall/mae_gain_left_to_rightVMAX/mean\endcsname{0.71}
\expandafter\def\csname mcresult@cell/missingness_sensitivity/pre_landfall/right_maeMSLP/mean\endcsname{5.72}
\expandafter\def\csname mcresult@cell/missingness_sensitivity/pre_landfall/right_maeVMAX/mean\endcsname{8.74}
\expandafter\def\csname mcresult@cell/missingness_sensitivity/pre_landfallMSLP/storm_ci_high/mean\endcsname{0.87}
\expandafter\def\csname mcresult@cell/missingness_sensitivity/pre_landfallMSLP/storm_ci_low/mean\endcsname{-0.06}
\expandafter\def\csname mcresult@cell/missingness_sensitivity/pre_landfallVMAX/storm_ci_high/mean\endcsname{1.34}
\expandafter\def\csname mcresult@cell/missingness_sensitivity/pre_landfallVMAX/storm_ci_low/mean\endcsname{0.12}
\expandafter\def\csname mcresult@cell/missingness_sensitivity/pre_landfall_discussion_present/left_maeMSLP/mean\endcsname{6.25}
\expandafter\def\csname mcresult@cell/missingness_sensitivity/pre_landfall_discussion_present/left_maeVMAX/mean\endcsname{9.62}
\expandafter\def\csname mcresult@cell/missingness_sensitivity/pre_landfall_discussion_present/mae_gain_left_to_rightMSLP/mean\endcsname{0.45}
\expandafter\def\csname mcresult@cell/missingness_sensitivity/pre_landfall_discussion_present/mae_gain_left_to_rightVMAX/mean\endcsname{0.79}
\expandafter\def\csname mcresult@cell/missingness_sensitivity/pre_landfall_discussion_present/right_maeMSLP/mean\endcsname{5.80}
\expandafter\def\csname mcresult@cell/missingness_sensitivity/pre_landfall_discussion_present/right_maeVMAX/mean\endcsname{8.83}
\expandafter\def\csname mcresult@cell/missingness_sensitivity/pre_landfall_discussion_presentMSLP/storm_ci_high/mean\endcsname{0.94}
\expandafter\def\csname mcresult@cell/missingness_sensitivity/pre_landfall_discussion_presentMSLP/storm_ci_low/mean\endcsname{-0.04}
\expandafter\def\csname mcresult@cell/missingness_sensitivity/pre_landfall_discussion_presentVMAX/storm_ci_high/mean\endcsname{1.44}
\expandafter\def\csname mcresult@cell/missingness_sensitivity/pre_landfall_discussion_presentVMAX/storm_ci_low/mean\endcsname{0.15}
\expandafter\def\csname mcresult@cell/nn_ablation_24h/disc/MSLP/mae/mean\endcsname{6.56}
\expandafter\def\csname mcresult@cell/nn_ablation_24h/disc/MSLP/mae/sd\endcsname{0.17}
\expandafter\def\csname mcresult@cell/nn_ablation_24h/disc/MSLP/r2/mean\endcsname{0.46}
\expandafter\def\csname mcresult@cell/nn_ablation_24h/disc/MSLP/r2/sd\endcsname{0.03}
\expandafter\def\csname mcresult@cell/nn_ablation_24h/disc/VMAX/mae/mean\endcsname{9.90}
\expandafter\def\csname mcresult@cell/nn_ablation_24h/disc/VMAX/mae/sd\endcsname{0.20}
\expandafter\def\csname mcresult@cell/nn_ablation_24h/disc/VMAX/r2/mean\endcsname{0.50}
\expandafter\def\csname mcresult@cell/nn_ablation_24h/disc/VMAX/r2/sd\endcsname{0.01}
\expandafter\def\csname mcresult@cell/nn_ablation_24h/full/MSLP/mae/mean\endcsname{6.23}
\expandafter\def\csname mcresult@cell/nn_ablation_24h/full/MSLP/mae/sd\endcsname{0.08}
\expandafter\def\csname mcresult@cell/nn_ablation_24h/full/MSLP/r2/mean\endcsname{0.51}
\expandafter\def\csname mcresult@cell/nn_ablation_24h/full/MSLP/r2/sd\endcsname{0.01}
\expandafter\def\csname mcresult@cell/nn_ablation_24h/full/VMAX/mae/mean\endcsname{9.45}
\expandafter\def\csname mcresult@cell/nn_ablation_24h/full/VMAX/mae/sd\endcsname{0.13}
\expandafter\def\csname mcresult@cell/nn_ablation_24h/full/VMAX/r2/mean\endcsname{0.53}
\expandafter\def\csname mcresult@cell/nn_ablation_24h/full/VMAX/r2/sd\endcsname{0.01}
\expandafter\def\csname mcresult@cell/nn_ablation_24h/img/MSLP/mae/mean\endcsname{8.15}
\expandafter\def\csname mcresult@cell/nn_ablation_24h/img/MSLP/mae/sd\endcsname{0.13}
\expandafter\def\csname mcresult@cell/nn_ablation_24h/img/MSLP/r2/mean\endcsname{0.20}
\expandafter\def\csname mcresult@cell/nn_ablation_24h/img/MSLP/r2/sd\endcsname{0.02}
\expandafter\def\csname mcresult@cell/nn_ablation_24h/img/VMAX/mae/mean\endcsname{12.40}
\expandafter\def\csname mcresult@cell/nn_ablation_24h/img/VMAX/mae/sd\endcsname{0.16}
\expandafter\def\csname mcresult@cell/nn_ablation_24h/img/VMAX/r2/mean\endcsname{0.22}
\expandafter\def\csname mcresult@cell/nn_ablation_24h/img/VMAX/r2/sd\endcsname{0.02}
\expandafter\def\csname mcresult@cell/nn_ablation_24h/img_disc/MSLP/mae/mean\endcsname{6.45}
\expandafter\def\csname mcresult@cell/nn_ablation_24h/img_disc/MSLP/mae/sd\endcsname{0.18}
\expandafter\def\csname mcresult@cell/nn_ablation_24h/img_disc/MSLP/r2/mean\endcsname{0.48}
\expandafter\def\csname mcresult@cell/nn_ablation_24h/img_disc/MSLP/r2/sd\endcsname{0.02}
\expandafter\def\csname mcresult@cell/nn_ablation_24h/img_disc/VMAX/mae/mean\endcsname{9.74}
\expandafter\def\csname mcresult@cell/nn_ablation_24h/img_disc/VMAX/mae/sd\endcsname{0.27}
\expandafter\def\csname mcresult@cell/nn_ablation_24h/img_disc/VMAX/r2/mean\endcsname{0.50}
\expandafter\def\csname mcresult@cell/nn_ablation_24h/img_disc/VMAX/r2/sd\endcsname{0.01}
\expandafter\def\csname mcresult@cell/nn_ablation_24h/tab/MSLP/mae/mean\endcsname{6.60}
\expandafter\def\csname mcresult@cell/nn_ablation_24h/tab/MSLP/mae/sd\endcsname{0.08}
\expandafter\def\csname mcresult@cell/nn_ablation_24h/tab/MSLP/r2/mean\endcsname{0.44}
\expandafter\def\csname mcresult@cell/nn_ablation_24h/tab/MSLP/r2/sd\endcsname{0.01}
\expandafter\def\csname mcresult@cell/nn_ablation_24h/tab/VMAX/mae/mean\endcsname{10.13}
\expandafter\def\csname mcresult@cell/nn_ablation_24h/tab/VMAX/mae/sd\endcsname{0.05}
\expandafter\def\csname mcresult@cell/nn_ablation_24h/tab/VMAX/r2/mean\endcsname{0.44}
\expandafter\def\csname mcresult@cell/nn_ablation_24h/tab/VMAX/r2/sd\endcsname{0.01}
\expandafter\def\csname mcresult@cell/nn_ablation_24h/tab_disc/MSLP/mae/mean\endcsname{6.23}
\expandafter\def\csname mcresult@cell/nn_ablation_24h/tab_disc/MSLP/mae/sd\endcsname{0.08}
\expandafter\def\csname mcresult@cell/nn_ablation_24h/tab_disc/MSLP/r2/mean\endcsname{0.51}
\expandafter\def\csname mcresult@cell/nn_ablation_24h/tab_disc/MSLP/r2/sd\endcsname{0.01}
\expandafter\def\csname mcresult@cell/nn_ablation_24h/tab_disc/VMAX/mae/mean\endcsname{9.46}
\expandafter\def\csname mcresult@cell/nn_ablation_24h/tab_disc/VMAX/mae/sd\endcsname{0.16}
\expandafter\def\csname mcresult@cell/nn_ablation_24h/tab_disc/VMAX/r2/mean\endcsname{0.53}
\expandafter\def\csname mcresult@cell/nn_ablation_24h/tab_disc/VMAX/r2/sd\endcsname{0.01}
\expandafter\def\csname mcresult@cell/nn_ablation_24h/tab_img/MSLP/mae/mean\endcsname{6.72}
\expandafter\def\csname mcresult@cell/nn_ablation_24h/tab_img/MSLP/mae/sd\endcsname{0.05}
\expandafter\def\csname mcresult@cell/nn_ablation_24h/tab_img/MSLP/r2/mean\endcsname{0.45}
\expandafter\def\csname mcresult@cell/nn_ablation_24h/tab_img/MSLP/r2/sd\endcsname{0.01}
\expandafter\def\csname mcresult@cell/nn_ablation_24h/tab_img/VMAX/mae/mean\endcsname{10.27}
\expandafter\def\csname mcresult@cell/nn_ablation_24h/tab_img/VMAX/mae/sd\endcsname{0.06}
\expandafter\def\csname mcresult@cell/nn_ablation_24h/tab_img/VMAX/r2/mean\endcsname{0.45}
\expandafter\def\csname mcresult@cell/nn_ablation_24h/tab_img/VMAX/r2/sd\endcsname{0.01}
\expandafter\def\csname mcresult@cell/nn_ensemble_24h/disc/MSLPmae/mean\endcsname{6.16}
\expandafter\def\csname mcresult@cell/nn_ensemble_24h/disc/MSLPr2/mean\endcsname{0.51}
\expandafter\def\csname mcresult@cell/nn_ensemble_24h/disc/VMAXmae/mean\endcsname{9.28}
\expandafter\def\csname mcresult@cell/nn_ensemble_24h/disc/VMAXr2/mean\endcsname{0.55}
\expandafter\def\csname mcresult@cell/nn_ensemble_24h/full/MSLPmae/mean\endcsname{5.88}
\expandafter\def\csname mcresult@cell/nn_ensemble_24h/full/MSLPr2/mean\endcsname{0.55}
\expandafter\def\csname mcresult@cell/nn_ensemble_24h/full/VMAXmae/mean\endcsname{8.92}
\expandafter\def\csname mcresult@cell/nn_ensemble_24h/full/VMAXr2/mean\endcsname{0.57}
\expandafter\def\csname mcresult@cell/nn_ensemble_24h/img/MSLPmae/mean\endcsname{7.85}
\expandafter\def\csname mcresult@cell/nn_ensemble_24h/img/MSLPr2/mean\endcsname{0.25}
\expandafter\def\csname mcresult@cell/nn_ensemble_24h/img/VMAXmae/mean\endcsname{11.99}
\expandafter\def\csname mcresult@cell/nn_ensemble_24h/img/VMAXr2/mean\endcsname{0.27}
\expandafter\def\csname mcresult@cell/nn_ensemble_24h/img_disc/MSLPmae/mean\endcsname{6.01}
\expandafter\def\csname mcresult@cell/nn_ensemble_24h/img_disc/MSLPr2/mean\endcsname{0.53}
\expandafter\def\csname mcresult@cell/nn_ensemble_24h/img_disc/VMAXmae/mean\endcsname{9.11}
\expandafter\def\csname mcresult@cell/nn_ensemble_24h/img_disc/VMAXr2/mean\endcsname{0.55}
\expandafter\def\csname mcresult@cell/nn_ensemble_24h/tab/MSLPmae/mean\endcsname{6.43}
\expandafter\def\csname mcresult@cell/nn_ensemble_24h/tab/MSLPr2/mean\endcsname{0.46}
\expandafter\def\csname mcresult@cell/nn_ensemble_24h/tab/VMAXmae/mean\endcsname{9.89}
\expandafter\def\csname mcresult@cell/nn_ensemble_24h/tab/VMAXr2/mean\endcsname{0.46}
\expandafter\def\csname mcresult@cell/nn_ensemble_24h/tab_disc/MSLPmae/mean\endcsname{5.87}
\expandafter\def\csname mcresult@cell/nn_ensemble_24h/tab_disc/MSLPr2/mean\endcsname{0.55}
\expandafter\def\csname mcresult@cell/nn_ensemble_24h/tab_disc/VMAXmae/mean\endcsname{8.90}
\expandafter\def\csname mcresult@cell/nn_ensemble_24h/tab_disc/VMAXr2/mean\endcsname{0.57}
\expandafter\def\csname mcresult@cell/nn_ensemble_24h/tab_img/MSLPmae/mean\endcsname{6.35}
\expandafter\def\csname mcresult@cell/nn_ensemble_24h/tab_img/MSLPr2/mean\endcsname{0.49}
\expandafter\def\csname mcresult@cell/nn_ensemble_24h/tab_img/VMAXmae/mean\endcsname{9.84}
\expandafter\def\csname mcresult@cell/nn_ensemble_24h/tab_img/VMAXr2/mean\endcsname{0.49}
\expandafter\def\csname mcresult@cell/operational_baseline/DSHP/TD_mae/mean\endcsname{8.90}
\expandafter\def\csname mcresult@cell/operational_baseline/DSHP/TD_minus_aid_mae/mean\endcsname{-0.33}
\expandafter\def\csname mcresult@cell/operational_baseline/DSHP/TID_mae/mean\endcsname{8.93}
\expandafter\def\csname mcresult@cell/operational_baseline/DSHP/T_mae/mean\endcsname{9.91}
\expandafter\def\csname mcresult@cell/operational_baseline/DSHP/aid_mae/mean\endcsname{9.23}
\expandafter\def\csname mcresult@cell/operational_baseline/DSHP/storm_ci_high/mean\endcsname{0.17}
\expandafter\def\csname mcresult@cell/operational_baseline/DSHP/storm_ci_low/mean\endcsname{-0.86}
\expandafter\def\csname mcresult@cell/operational_baseline/HF3A/TD_mae/mean\endcsname{8.28}
\expandafter\def\csname mcresult@cell/operational_baseline/HF3A/TD_minus_aid_mae/mean\endcsname{-0.57}
\expandafter\def\csname mcresult@cell/operational_baseline/HF3A/TID_mae/mean\endcsname{8.26}
\expandafter\def\csname mcresult@cell/operational_baseline/HF3A/T_mae/mean\endcsname{11.77}
\expandafter\def\csname mcresult@cell/operational_baseline/HF3A/aid_mae/mean\endcsname{8.85}
\expandafter\def\csname mcresult@cell/operational_baseline/HF3A/storm_ci_high/mean\endcsname{2.35}
\expandafter\def\csname mcresult@cell/operational_baseline/HF3A/storm_ci_low/mean\endcsname{-3.35}
\expandafter\def\csname mcresult@cell/operational_baseline/HFAI/TD_mae/mean\endcsname{9.01}
\expandafter\def\csname mcresult@cell/operational_baseline/HFAI/TD_minus_aid_mae/mean\endcsname{-0.58}
\expandafter\def\csname mcresult@cell/operational_baseline/HFAI/TID_mae/mean\endcsname{9.00}
\expandafter\def\csname mcresult@cell/operational_baseline/HFAI/T_mae/mean\endcsname{9.75}
\expandafter\def\csname mcresult@cell/operational_baseline/HFAI/aid_mae/mean\endcsname{9.59}
\expandafter\def\csname mcresult@cell/operational_baseline/HFAI/storm_ci_high/mean\endcsname{0.58}
\expandafter\def\csname mcresult@cell/operational_baseline/HFAI/storm_ci_low/mean\endcsname{-1.77}
\expandafter\def\csname mcresult@cell/operational_baseline/HFBI/TD_mae/mean\endcsname{9.01}
\expandafter\def\csname mcresult@cell/operational_baseline/HFBI/TD_minus_aid_mae/mean\endcsname{-0.52}
\expandafter\def\csname mcresult@cell/operational_baseline/HFBI/TID_mae/mean\endcsname{9.00}
\expandafter\def\csname mcresult@cell/operational_baseline/HFBI/T_mae/mean\endcsname{9.75}
\expandafter\def\csname mcresult@cell/operational_baseline/HFBI/aid_mae/mean\endcsname{9.53}
\expandafter\def\csname mcresult@cell/operational_baseline/HFBI/storm_ci_high/mean\endcsname{0.37}
\expandafter\def\csname mcresult@cell/operational_baseline/HFBI/storm_ci_low/mean\endcsname{-1.40}
\expandafter\def\csname mcresult@cell/operational_baseline/HFSA/TD_mae/mean\endcsname{9.01}
\expandafter\def\csname mcresult@cell/operational_baseline/HFSA/TD_minus_aid_mae/mean\endcsname{-1.68}
\expandafter\def\csname mcresult@cell/operational_baseline/HFSA/TID_mae/mean\endcsname{9.00}
\expandafter\def\csname mcresult@cell/operational_baseline/HFSA/T_mae/mean\endcsname{9.73}
\expandafter\def\csname mcresult@cell/operational_baseline/HFSA/aid_mae/mean\endcsname{10.69}
\expandafter\def\csname mcresult@cell/operational_baseline/HFSA/storm_ci_high/mean\endcsname{-0.72}
\expandafter\def\csname mcresult@cell/operational_baseline/HFSA/storm_ci_low/mean\endcsname{-2.66}
\expandafter\def\csname mcresult@cell/operational_baseline/HFSB/TD_mae/mean\endcsname{9.01}
\expandafter\def\csname mcresult@cell/operational_baseline/HFSB/TD_minus_aid_mae/mean\endcsname{-1.99}
\expandafter\def\csname mcresult@cell/operational_baseline/HFSB/TID_mae/mean\endcsname{9.00}
\expandafter\def\csname mcresult@cell/operational_baseline/HFSB/T_mae/mean\endcsname{9.73}
\expandafter\def\csname mcresult@cell/operational_baseline/HFSB/aid_mae/mean\endcsname{11.00}
\expandafter\def\csname mcresult@cell/operational_baseline/HFSB/storm_ci_high/mean\endcsname{-1.22}
\expandafter\def\csname mcresult@cell/operational_baseline/HFSB/storm_ci_low/mean\endcsname{-2.69}
\expandafter\def\csname mcresult@cell/operational_baseline/HWFI/TD_mae/mean\endcsname{8.90}
\expandafter\def\csname mcresult@cell/operational_baseline/HWFI/TD_minus_aid_mae/mean\endcsname{0.06}
\expandafter\def\csname mcresult@cell/operational_baseline/HWFI/TID_mae/mean\endcsname{8.92}
\expandafter\def\csname mcresult@cell/operational_baseline/HWFI/T_mae/mean\endcsname{9.98}
\expandafter\def\csname mcresult@cell/operational_baseline/HWFI/aid_mae/mean\endcsname{8.85}
\expandafter\def\csname mcresult@cell/operational_baseline/HWFI/storm_ci_high/mean\endcsname{0.72}
\expandafter\def\csname mcresult@cell/operational_baseline/HWFI/storm_ci_low/mean\endcsname{-0.63}
\expandafter\def\csname mcresult@cell/operational_baseline/HWRF/TD_mae/mean\endcsname{8.97}
\expandafter\def\csname mcresult@cell/operational_baseline/HWRF/TD_minus_aid_mae/mean\endcsname{-0.71}
\expandafter\def\csname mcresult@cell/operational_baseline/HWRF/TID_mae/mean\endcsname{9.00}
\expandafter\def\csname mcresult@cell/operational_baseline/HWRF/T_mae/mean\endcsname{9.98}
\expandafter\def\csname mcresult@cell/operational_baseline/HWRF/aid_mae/mean\endcsname{9.67}
\expandafter\def\csname mcresult@cell/operational_baseline/HWRF/storm_ci_high/mean\endcsname{0.20}
\expandafter\def\csname mcresult@cell/operational_baseline/HWRF/storm_ci_low/mean\endcsname{-1.55}
\expandafter\def\csname mcresult@cell/operational_baseline/LGEM/TD_mae/mean\endcsname{8.90}
\expandafter\def\csname mcresult@cell/operational_baseline/LGEM/TD_minus_aid_mae/mean\endcsname{-0.34}
\expandafter\def\csname mcresult@cell/operational_baseline/LGEM/TID_mae/mean\endcsname{8.93}
\expandafter\def\csname mcresult@cell/operational_baseline/LGEM/T_mae/mean\endcsname{9.91}
\expandafter\def\csname mcresult@cell/operational_baseline/LGEM/aid_mae/mean\endcsname{9.24}
\expandafter\def\csname mcresult@cell/operational_baseline/LGEM/storm_ci_high/mean\endcsname{0.22}
\expandafter\def\csname mcresult@cell/operational_baseline/LGEM/storm_ci_low/mean\endcsname{-0.95}
\expandafter\def\csname mcresult@cell/operational_baseline/OFCI/TD_mae/mean\endcsname{9.09}
\expandafter\def\csname mcresult@cell/operational_baseline/OFCI/TD_minus_aid_mae/mean\endcsname{0.66}
\expandafter\def\csname mcresult@cell/operational_baseline/OFCI/TID_mae/mean\endcsname{9.12}
\expandafter\def\csname mcresult@cell/operational_baseline/OFCI/T_mae/mean\endcsname{10.16}
\expandafter\def\csname mcresult@cell/operational_baseline/OFCI/aid_mae/mean\endcsname{8.43}
\expandafter\def\csname mcresult@cell/operational_baseline/OFCI/storm_ci_high/mean\endcsname{1.11}
\expandafter\def\csname mcresult@cell/operational_baseline/OFCI/storm_ci_low/mean\endcsname{0.17}
\expandafter\def\csname mcresult@cell/operational_baseline/OFCL/TD_mae/mean\endcsname{8.97}
\expandafter\def\csname mcresult@cell/operational_baseline/OFCL/TD_minus_aid_mae/mean\endcsname{1.28}
\expandafter\def\csname mcresult@cell/operational_baseline/OFCL/TID_mae/mean\endcsname{8.99}
\expandafter\def\csname mcresult@cell/operational_baseline/OFCL/T_mae/mean\endcsname{10.07}
\expandafter\def\csname mcresult@cell/operational_baseline/OFCL/aid_mae/mean\endcsname{7.69}
\expandafter\def\csname mcresult@cell/operational_baseline/OFCL/storm_ci_high/mean\endcsname{1.82}
\expandafter\def\csname mcresult@cell/operational_baseline/OFCL/storm_ci_low/mean\endcsname{0.71}
\expandafter\def\csname mcresult@cell/per_horizon_vs_joint/disc_fph_120/VMAX/mae/mean\endcsname{23.42}
\expandafter\def\csname mcresult@cell/per_horizon_vs_joint/disc_fph_120/VMAX/mae/sd\endcsname{2.03}
\expandafter\def\csname mcresult@cell/per_horizon_vs_joint/disc_fph_120/VMAX/r2/mean\endcsname{0.00}
\expandafter\def\csname mcresult@cell/per_horizon_vs_joint/disc_fph_120/VMAX/r2/sd\endcsname{0.15}
\expandafter\def\csname mcresult@cell/per_horizon_vs_joint/disc_fph_24/VMAX/mae/mean\endcsname{9.49}
\expandafter\def\csname mcresult@cell/per_horizon_vs_joint/disc_fph_24/VMAX/mae/sd\endcsname{0.15}
\expandafter\def\csname mcresult@cell/per_horizon_vs_joint/disc_fph_24/VMAX/r2/mean\endcsname{0.53}
\expandafter\def\csname mcresult@cell/per_horizon_vs_joint/disc_fph_24/VMAX/r2/sd\endcsname{0.01}
\expandafter\def\csname mcresult@cell/per_horizon_vs_joint/disc_fph_48/VMAX/mae/mean\endcsname{14.32}
\expandafter\def\csname mcresult@cell/per_horizon_vs_joint/disc_fph_48/VMAX/mae/sd\endcsname{0.17}
\expandafter\def\csname mcresult@cell/per_horizon_vs_joint/disc_fph_48/VMAX/r2/mean\endcsname{0.54}
\expandafter\def\csname mcresult@cell/per_horizon_vs_joint/disc_fph_48/VMAX/r2/sd\endcsname{0.01}
\expandafter\def\csname mcresult@cell/per_horizon_vs_joint/disc_fph_72/VMAX/mae/mean\endcsname{17.63}
\expandafter\def\csname mcresult@cell/per_horizon_vs_joint/disc_fph_72/VMAX/mae/sd\endcsname{0.80}
\expandafter\def\csname mcresult@cell/per_horizon_vs_joint/disc_fph_72/VMAX/r2/mean\endcsname{0.42}
\expandafter\def\csname mcresult@cell/per_horizon_vs_joint/disc_fph_72/VMAX/r2/sd\endcsname{0.06}
\expandafter\def\csname mcresult@cell/per_horizon_vs_joint/disc_fph_96/VMAX/mae/mean\endcsname{20.80}
\expandafter\def\csname mcresult@cell/per_horizon_vs_joint/disc_fph_96/VMAX/mae/sd\endcsname{1.60}
\expandafter\def\csname mcresult@cell/per_horizon_vs_joint/disc_fph_96/VMAX/r2/mean\endcsname{0.19}
\expandafter\def\csname mcresult@cell/per_horizon_vs_joint/disc_fph_96/VMAX/r2/sd\endcsname{0.11}
\expandafter\def\csname mcresult@cell/per_horizon_vs_joint/disc_mhj_120/VMAX/mae/mean\endcsname{20.67}
\expandafter\def\csname mcresult@cell/per_horizon_vs_joint/disc_mhj_120/VMAX/mae/sd\endcsname{0.43}
\expandafter\def\csname mcresult@cell/per_horizon_vs_joint/disc_mhj_120/VMAX/r2/mean\endcsname{0.10}
\expandafter\def\csname mcresult@cell/per_horizon_vs_joint/disc_mhj_120/VMAX/r2/sd\endcsname{0.05}
\expandafter\def\csname mcresult@cell/per_horizon_vs_joint/disc_mhj_24/VMAX/mae/mean\endcsname{9.78}
\expandafter\def\csname mcresult@cell/per_horizon_vs_joint/disc_mhj_24/VMAX/mae/sd\endcsname{0.25}
\expandafter\def\csname mcresult@cell/per_horizon_vs_joint/disc_mhj_24/VMAX/r2/mean\endcsname{0.51}
\expandafter\def\csname mcresult@cell/per_horizon_vs_joint/disc_mhj_24/VMAX/r2/sd\endcsname{0.02}
\expandafter\def\csname mcresult@cell/per_horizon_vs_joint/disc_mhj_48/VMAX/mae/mean\endcsname{14.35}
\expandafter\def\csname mcresult@cell/per_horizon_vs_joint/disc_mhj_48/VMAX/mae/sd\endcsname{0.27}
\expandafter\def\csname mcresult@cell/per_horizon_vs_joint/disc_mhj_48/VMAX/r2/mean\endcsname{0.53}
\expandafter\def\csname mcresult@cell/per_horizon_vs_joint/disc_mhj_48/VMAX/r2/sd\endcsname{0.02}
\expandafter\def\csname mcresult@cell/per_horizon_vs_joint/disc_mhj_72/VMAX/mae/mean\endcsname{17.19}
\expandafter\def\csname mcresult@cell/per_horizon_vs_joint/disc_mhj_72/VMAX/mae/sd\endcsname{0.13}
\expandafter\def\csname mcresult@cell/per_horizon_vs_joint/disc_mhj_72/VMAX/r2/mean\endcsname{0.43}
\expandafter\def\csname mcresult@cell/per_horizon_vs_joint/disc_mhj_72/VMAX/r2/sd\endcsname{0.02}
\expandafter\def\csname mcresult@cell/per_horizon_vs_joint/disc_mhj_96/VMAX/mae/mean\endcsname{19.39}
\expandafter\def\csname mcresult@cell/per_horizon_vs_joint/disc_mhj_96/VMAX/mae/sd\endcsname{0.45}
\expandafter\def\csname mcresult@cell/per_horizon_vs_joint/disc_mhj_96/VMAX/r2/mean\endcsname{0.24}
\expandafter\def\csname mcresult@cell/per_horizon_vs_joint/disc_mhj_96/VMAX/r2/sd\endcsname{0.03}
\expandafter\def\csname mcresult@cell/per_horizon_vs_joint/full_fph_120/VMAX/mae/mean\endcsname{23.65}
\expandafter\def\csname mcresult@cell/per_horizon_vs_joint/full_fph_120/VMAX/mae/sd\endcsname{2.26}
\expandafter\def\csname mcresult@cell/per_horizon_vs_joint/full_fph_120/VMAX/r2/mean\endcsname{-0.00}
\expandafter\def\csname mcresult@cell/per_horizon_vs_joint/full_fph_120/VMAX/r2/sd\endcsname{0.13}
\expandafter\def\csname mcresult@cell/per_horizon_vs_joint/full_fph_24/VMAX/mae/mean\endcsname{9.15}
\expandafter\def\csname mcresult@cell/per_horizon_vs_joint/full_fph_24/VMAX/mae/sd\endcsname{0.19}
\expandafter\def\csname mcresult@cell/per_horizon_vs_joint/full_fph_24/VMAX/r2/mean\endcsname{0.55}
\expandafter\def\csname mcresult@cell/per_horizon_vs_joint/full_fph_24/VMAX/r2/sd\endcsname{0.02}
\expandafter\def\csname mcresult@cell/per_horizon_vs_joint/full_fph_48/VMAX/mae/mean\endcsname{13.65}
\expandafter\def\csname mcresult@cell/per_horizon_vs_joint/full_fph_48/VMAX/mae/sd\endcsname{0.22}
\expandafter\def\csname mcresult@cell/per_horizon_vs_joint/full_fph_48/VMAX/r2/mean\endcsname{0.57}
\expandafter\def\csname mcresult@cell/per_horizon_vs_joint/full_fph_48/VMAX/r2/sd\endcsname{0.01}
\expandafter\def\csname mcresult@cell/per_horizon_vs_joint/full_fph_72/VMAX/mae/mean\endcsname{17.11}
\expandafter\def\csname mcresult@cell/per_horizon_vs_joint/full_fph_72/VMAX/mae/sd\endcsname{0.61}
\expandafter\def\csname mcresult@cell/per_horizon_vs_joint/full_fph_72/VMAX/r2/mean\endcsname{0.45}
\expandafter\def\csname mcresult@cell/per_horizon_vs_joint/full_fph_72/VMAX/r2/sd\endcsname{0.04}
\expandafter\def\csname mcresult@cell/per_horizon_vs_joint/full_fph_96/VMAX/mae/mean\endcsname{20.43}
\expandafter\def\csname mcresult@cell/per_horizon_vs_joint/full_fph_96/VMAX/mae/sd\endcsname{0.81}
\expandafter\def\csname mcresult@cell/per_horizon_vs_joint/full_fph_96/VMAX/r2/mean\endcsname{0.19}
\expandafter\def\csname mcresult@cell/per_horizon_vs_joint/full_fph_96/VMAX/r2/sd\endcsname{0.08}
\expandafter\def\csname mcresult@cell/per_horizon_vs_joint/full_mhj_120/VMAX/mae/mean\endcsname{20.61}
\expandafter\def\csname mcresult@cell/per_horizon_vs_joint/full_mhj_120/VMAX/mae/sd\endcsname{1.17}
\expandafter\def\csname mcresult@cell/per_horizon_vs_joint/full_mhj_120/VMAX/r2/mean\endcsname{0.09}
\expandafter\def\csname mcresult@cell/per_horizon_vs_joint/full_mhj_120/VMAX/r2/sd\endcsname{0.13}
\expandafter\def\csname mcresult@cell/per_horizon_vs_joint/full_mhj_24/VMAX/mae/mean\endcsname{9.24}
\expandafter\def\csname mcresult@cell/per_horizon_vs_joint/full_mhj_24/VMAX/mae/sd\endcsname{0.22}
\expandafter\def\csname mcresult@cell/per_horizon_vs_joint/full_mhj_24/VMAX/r2/mean\endcsname{0.54}
\expandafter\def\csname mcresult@cell/per_horizon_vs_joint/full_mhj_24/VMAX/r2/sd\endcsname{0.03}
\expandafter\def\csname mcresult@cell/per_horizon_vs_joint/full_mhj_48/VMAX/mae/mean\endcsname{13.66}
\expandafter\def\csname mcresult@cell/per_horizon_vs_joint/full_mhj_48/VMAX/mae/sd\endcsname{0.48}
\expandafter\def\csname mcresult@cell/per_horizon_vs_joint/full_mhj_48/VMAX/r2/mean\endcsname{0.56}
\expandafter\def\csname mcresult@cell/per_horizon_vs_joint/full_mhj_48/VMAX/r2/sd\endcsname{0.03}
\expandafter\def\csname mcresult@cell/per_horizon_vs_joint/full_mhj_72/VMAX/mae/mean\endcsname{16.88}
\expandafter\def\csname mcresult@cell/per_horizon_vs_joint/full_mhj_72/VMAX/mae/sd\endcsname{0.76}
\expandafter\def\csname mcresult@cell/per_horizon_vs_joint/full_mhj_72/VMAX/r2/mean\endcsname{0.41}
\expandafter\def\csname mcresult@cell/per_horizon_vs_joint/full_mhj_72/VMAX/r2/sd\endcsname{0.05}
\expandafter\def\csname mcresult@cell/per_horizon_vs_joint/full_mhj_96/VMAX/mae/mean\endcsname{19.58}
\expandafter\def\csname mcresult@cell/per_horizon_vs_joint/full_mhj_96/VMAX/mae/sd\endcsname{1.34}
\expandafter\def\csname mcresult@cell/per_horizon_vs_joint/full_mhj_96/VMAX/r2/mean\endcsname{0.19}
\expandafter\def\csname mcresult@cell/per_horizon_vs_joint/full_mhj_96/VMAX/r2/sd\endcsname{0.10}
\expandafter\def\csname mcresult@cell/per_horizon_vs_joint/img_disc_fph_120/VMAX/mae/mean\endcsname{23.13}
\expandafter\def\csname mcresult@cell/per_horizon_vs_joint/img_disc_fph_120/VMAX/mae/sd\endcsname{2.57}
\expandafter\def\csname mcresult@cell/per_horizon_vs_joint/img_disc_fph_120/VMAX/r2/mean\endcsname{0.05}
\expandafter\def\csname mcresult@cell/per_horizon_vs_joint/img_disc_fph_120/VMAX/r2/sd\endcsname{0.18}
\expandafter\def\csname mcresult@cell/per_horizon_vs_joint/img_disc_fph_24/VMAX/mae/mean\endcsname{10.06}
\expandafter\def\csname mcresult@cell/per_horizon_vs_joint/img_disc_fph_24/VMAX/mae/sd\endcsname{0.39}
\expandafter\def\csname mcresult@cell/per_horizon_vs_joint/img_disc_fph_24/VMAX/r2/mean\endcsname{0.47}
\expandafter\def\csname mcresult@cell/per_horizon_vs_joint/img_disc_fph_24/VMAX/r2/sd\endcsname{0.03}
\expandafter\def\csname mcresult@cell/per_horizon_vs_joint/img_disc_fph_48/VMAX/mae/mean\endcsname{14.38}
\expandafter\def\csname mcresult@cell/per_horizon_vs_joint/img_disc_fph_48/VMAX/mae/sd\endcsname{0.54}
\expandafter\def\csname mcresult@cell/per_horizon_vs_joint/img_disc_fph_48/VMAX/r2/mean\endcsname{0.52}
\expandafter\def\csname mcresult@cell/per_horizon_vs_joint/img_disc_fph_48/VMAX/r2/sd\endcsname{0.03}
\expandafter\def\csname mcresult@cell/per_horizon_vs_joint/img_disc_fph_72/VMAX/mae/mean\endcsname{18.08}
\expandafter\def\csname mcresult@cell/per_horizon_vs_joint/img_disc_fph_72/VMAX/mae/sd\endcsname{1.48}
\expandafter\def\csname mcresult@cell/per_horizon_vs_joint/img_disc_fph_72/VMAX/r2/mean\endcsname{0.39}
\expandafter\def\csname mcresult@cell/per_horizon_vs_joint/img_disc_fph_72/VMAX/r2/sd\endcsname{0.10}
\expandafter\def\csname mcresult@cell/per_horizon_vs_joint/img_disc_fph_96/VMAX/mae/mean\endcsname{21.62}
\expandafter\def\csname mcresult@cell/per_horizon_vs_joint/img_disc_fph_96/VMAX/mae/sd\endcsname{2.66}
\expandafter\def\csname mcresult@cell/per_horizon_vs_joint/img_disc_fph_96/VMAX/r2/mean\endcsname{0.08}
\expandafter\def\csname mcresult@cell/per_horizon_vs_joint/img_disc_fph_96/VMAX/r2/sd\endcsname{0.19}
\expandafter\def\csname mcresult@cell/per_horizon_vs_joint/img_disc_mhj_120/VMAX/mae/mean\endcsname{20.84}
\expandafter\def\csname mcresult@cell/per_horizon_vs_joint/img_disc_mhj_120/VMAX/mae/sd\endcsname{1.06}
\expandafter\def\csname mcresult@cell/per_horizon_vs_joint/img_disc_mhj_120/VMAX/r2/mean\endcsname{0.09}
\expandafter\def\csname mcresult@cell/per_horizon_vs_joint/img_disc_mhj_120/VMAX/r2/sd\endcsname{0.11}
\expandafter\def\csname mcresult@cell/per_horizon_vs_joint/img_disc_mhj_24/VMAX/mae/mean\endcsname{9.66}
\expandafter\def\csname mcresult@cell/per_horizon_vs_joint/img_disc_mhj_24/VMAX/mae/sd\endcsname{0.12}
\expandafter\def\csname mcresult@cell/per_horizon_vs_joint/img_disc_mhj_24/VMAX/r2/mean\endcsname{0.52}
\expandafter\def\csname mcresult@cell/per_horizon_vs_joint/img_disc_mhj_24/VMAX/r2/sd\endcsname{0.01}
\expandafter\def\csname mcresult@cell/per_horizon_vs_joint/img_disc_mhj_48/VMAX/mae/mean\endcsname{14.81}
\expandafter\def\csname mcresult@cell/per_horizon_vs_joint/img_disc_mhj_48/VMAX/mae/sd\endcsname{0.35}
\expandafter\def\csname mcresult@cell/per_horizon_vs_joint/img_disc_mhj_48/VMAX/r2/mean\endcsname{0.50}
\expandafter\def\csname mcresult@cell/per_horizon_vs_joint/img_disc_mhj_48/VMAX/r2/sd\endcsname{0.02}
\expandafter\def\csname mcresult@cell/per_horizon_vs_joint/img_disc_mhj_72/VMAX/mae/mean\endcsname{17.55}
\expandafter\def\csname mcresult@cell/per_horizon_vs_joint/img_disc_mhj_72/VMAX/mae/sd\endcsname{0.18}
\expandafter\def\csname mcresult@cell/per_horizon_vs_joint/img_disc_mhj_72/VMAX/r2/mean\endcsname{0.42}
\expandafter\def\csname mcresult@cell/per_horizon_vs_joint/img_disc_mhj_72/VMAX/r2/sd\endcsname{0.03}
\expandafter\def\csname mcresult@cell/per_horizon_vs_joint/img_disc_mhj_96/VMAX/mae/mean\endcsname{19.62}
\expandafter\def\csname mcresult@cell/per_horizon_vs_joint/img_disc_mhj_96/VMAX/mae/sd\endcsname{0.65}
\expandafter\def\csname mcresult@cell/per_horizon_vs_joint/img_disc_mhj_96/VMAX/r2/mean\endcsname{0.22}
\expandafter\def\csname mcresult@cell/per_horizon_vs_joint/img_disc_mhj_96/VMAX/r2/sd\endcsname{0.06}
\expandafter\def\csname mcresult@cell/per_horizon_vs_joint/img_fph_120/VMAX/mae/mean\endcsname{20.76}
\expandafter\def\csname mcresult@cell/per_horizon_vs_joint/img_fph_120/VMAX/mae/sd\endcsname{0.62}
\expandafter\def\csname mcresult@cell/per_horizon_vs_joint/img_fph_120/VMAX/r2/mean\endcsname{0.15}
\expandafter\def\csname mcresult@cell/per_horizon_vs_joint/img_fph_120/VMAX/r2/sd\endcsname{0.04}
\expandafter\def\csname mcresult@cell/per_horizon_vs_joint/img_fph_24/VMAX/mae/mean\endcsname{12.43}
\expandafter\def\csname mcresult@cell/per_horizon_vs_joint/img_fph_24/VMAX/mae/sd\endcsname{0.26}
\expandafter\def\csname mcresult@cell/per_horizon_vs_joint/img_fph_24/VMAX/r2/mean\endcsname{0.21}
\expandafter\def\csname mcresult@cell/per_horizon_vs_joint/img_fph_24/VMAX/r2/sd\endcsname{0.03}
\expandafter\def\csname mcresult@cell/per_horizon_vs_joint/img_fph_48/VMAX/mae/mean\endcsname{18.97}
\expandafter\def\csname mcresult@cell/per_horizon_vs_joint/img_fph_48/VMAX/mae/sd\endcsname{0.37}
\expandafter\def\csname mcresult@cell/per_horizon_vs_joint/img_fph_48/VMAX/r2/mean\endcsname{0.16}
\expandafter\def\csname mcresult@cell/per_horizon_vs_joint/img_fph_48/VMAX/r2/sd\endcsname{0.06}
\expandafter\def\csname mcresult@cell/per_horizon_vs_joint/img_fph_72/VMAX/mae/mean\endcsname{20.43}
\expandafter\def\csname mcresult@cell/per_horizon_vs_joint/img_fph_72/VMAX/mae/sd\endcsname{0.75}
\expandafter\def\csname mcresult@cell/per_horizon_vs_joint/img_fph_72/VMAX/r2/mean\endcsname{0.24}
\expandafter\def\csname mcresult@cell/per_horizon_vs_joint/img_fph_72/VMAX/r2/sd\endcsname{0.06}
\expandafter\def\csname mcresult@cell/per_horizon_vs_joint/img_fph_96/VMAX/mae/mean\endcsname{21.00}
\expandafter\def\csname mcresult@cell/per_horizon_vs_joint/img_fph_96/VMAX/mae/sd\endcsname{0.94}
\expandafter\def\csname mcresult@cell/per_horizon_vs_joint/img_fph_96/VMAX/r2/mean\endcsname{0.18}
\expandafter\def\csname mcresult@cell/per_horizon_vs_joint/img_fph_96/VMAX/r2/sd\endcsname{0.07}
\expandafter\def\csname mcresult@cell/per_horizon_vs_joint/img_mhj_120/VMAX/mae/mean\endcsname{21.31}
\expandafter\def\csname mcresult@cell/per_horizon_vs_joint/img_mhj_120/VMAX/mae/sd\endcsname{2.87}
\expandafter\def\csname mcresult@cell/per_horizon_vs_joint/img_mhj_120/VMAX/r2/mean\endcsname{-0.46}
\expandafter\def\csname mcresult@cell/per_horizon_vs_joint/img_mhj_120/VMAX/r2/sd\endcsname{1.44}
\expandafter\def\csname mcresult@cell/per_horizon_vs_joint/img_mhj_24/VMAX/mae/mean\endcsname{12.60}
\expandafter\def\csname mcresult@cell/per_horizon_vs_joint/img_mhj_24/VMAX/mae/sd\endcsname{0.66}
\expandafter\def\csname mcresult@cell/per_horizon_vs_joint/img_mhj_24/VMAX/r2/mean\endcsname{0.09}
\expandafter\def\csname mcresult@cell/per_horizon_vs_joint/img_mhj_24/VMAX/r2/sd\endcsname{0.23}
\expandafter\def\csname mcresult@cell/per_horizon_vs_joint/img_mhj_48/VMAX/mae/mean\endcsname{18.94}
\expandafter\def\csname mcresult@cell/per_horizon_vs_joint/img_mhj_48/VMAX/mae/sd\endcsname{1.14}
\expandafter\def\csname mcresult@cell/per_horizon_vs_joint/img_mhj_48/VMAX/r2/mean\endcsname{0.12}
\expandafter\def\csname mcresult@cell/per_horizon_vs_joint/img_mhj_48/VMAX/r2/sd\endcsname{0.29}
\expandafter\def\csname mcresult@cell/per_horizon_vs_joint/img_mhj_72/VMAX/mae/mean\endcsname{20.91}
\expandafter\def\csname mcresult@cell/per_horizon_vs_joint/img_mhj_72/VMAX/mae/sd\endcsname{1.28}
\expandafter\def\csname mcresult@cell/per_horizon_vs_joint/img_mhj_72/VMAX/r2/mean\endcsname{0.08}
\expandafter\def\csname mcresult@cell/per_horizon_vs_joint/img_mhj_72/VMAX/r2/sd\endcsname{0.38}
\expandafter\def\csname mcresult@cell/per_horizon_vs_joint/img_mhj_96/VMAX/mae/mean\endcsname{20.95}
\expandafter\def\csname mcresult@cell/per_horizon_vs_joint/img_mhj_96/VMAX/mae/sd\endcsname{1.90}
\expandafter\def\csname mcresult@cell/per_horizon_vs_joint/img_mhj_96/VMAX/r2/mean\endcsname{-0.06}
\expandafter\def\csname mcresult@cell/per_horizon_vs_joint/img_mhj_96/VMAX/r2/sd\endcsname{0.65}
\expandafter\def\csname mcresult@cell/per_horizon_vs_joint/tab_disc_fph_120/VMAX/mae/mean\endcsname{22.51}
\expandafter\def\csname mcresult@cell/per_horizon_vs_joint/tab_disc_fph_120/VMAX/mae/sd\endcsname{1.49}
\expandafter\def\csname mcresult@cell/per_horizon_vs_joint/tab_disc_fph_120/VMAX/r2/mean\endcsname{0.06}
\expandafter\def\csname mcresult@cell/per_horizon_vs_joint/tab_disc_fph_120/VMAX/r2/sd\endcsname{0.07}
\expandafter\def\csname mcresult@cell/per_horizon_vs_joint/tab_disc_fph_24/VMAX/mae/mean\endcsname{9.26}
\expandafter\def\csname mcresult@cell/per_horizon_vs_joint/tab_disc_fph_24/VMAX/mae/sd\endcsname{0.29}
\expandafter\def\csname mcresult@cell/per_horizon_vs_joint/tab_disc_fph_24/VMAX/r2/mean\endcsname{0.54}
\expandafter\def\csname mcresult@cell/per_horizon_vs_joint/tab_disc_fph_24/VMAX/r2/sd\endcsname{0.04}
\expandafter\def\csname mcresult@cell/per_horizon_vs_joint/tab_disc_fph_48/VMAX/mae/mean\endcsname{13.44}
\expandafter\def\csname mcresult@cell/per_horizon_vs_joint/tab_disc_fph_48/VMAX/mae/sd\endcsname{0.23}
\expandafter\def\csname mcresult@cell/per_horizon_vs_joint/tab_disc_fph_48/VMAX/r2/mean\endcsname{0.58}
\expandafter\def\csname mcresult@cell/per_horizon_vs_joint/tab_disc_fph_48/VMAX/r2/sd\endcsname{0.02}
\expandafter\def\csname mcresult@cell/per_horizon_vs_joint/tab_disc_fph_72/VMAX/mae/mean\endcsname{16.93}
\expandafter\def\csname mcresult@cell/per_horizon_vs_joint/tab_disc_fph_72/VMAX/mae/sd\endcsname{0.67}
\expandafter\def\csname mcresult@cell/per_horizon_vs_joint/tab_disc_fph_72/VMAX/r2/mean\endcsname{0.46}
\expandafter\def\csname mcresult@cell/per_horizon_vs_joint/tab_disc_fph_72/VMAX/r2/sd\endcsname{0.06}
\expandafter\def\csname mcresult@cell/per_horizon_vs_joint/tab_disc_fph_96/VMAX/mae/mean\endcsname{19.76}
\expandafter\def\csname mcresult@cell/per_horizon_vs_joint/tab_disc_fph_96/VMAX/mae/sd\endcsname{1.22}
\expandafter\def\csname mcresult@cell/per_horizon_vs_joint/tab_disc_fph_96/VMAX/r2/mean\endcsname{0.24}
\expandafter\def\csname mcresult@cell/per_horizon_vs_joint/tab_disc_fph_96/VMAX/r2/sd\endcsname{0.07}
\expandafter\def\csname mcresult@cell/per_horizon_vs_joint/tab_disc_mhj_120/VMAX/mae/mean\endcsname{21.29}
\expandafter\def\csname mcresult@cell/per_horizon_vs_joint/tab_disc_mhj_120/VMAX/mae/sd\endcsname{0.66}
\expandafter\def\csname mcresult@cell/per_horizon_vs_joint/tab_disc_mhj_120/VMAX/r2/mean\endcsname{0.06}
\expandafter\def\csname mcresult@cell/per_horizon_vs_joint/tab_disc_mhj_120/VMAX/r2/sd\endcsname{0.10}
\expandafter\def\csname mcresult@cell/per_horizon_vs_joint/tab_disc_mhj_24/VMAX/mae/mean\endcsname{9.40}
\expandafter\def\csname mcresult@cell/per_horizon_vs_joint/tab_disc_mhj_24/VMAX/mae/sd\endcsname{0.30}
\expandafter\def\csname mcresult@cell/per_horizon_vs_joint/tab_disc_mhj_24/VMAX/r2/mean\endcsname{0.54}
\expandafter\def\csname mcresult@cell/per_horizon_vs_joint/tab_disc_mhj_24/VMAX/r2/sd\endcsname{0.02}
\expandafter\def\csname mcresult@cell/per_horizon_vs_joint/tab_disc_mhj_48/VMAX/mae/mean\endcsname{13.84}
\expandafter\def\csname mcresult@cell/per_horizon_vs_joint/tab_disc_mhj_48/VMAX/mae/sd\endcsname{0.28}
\expandafter\def\csname mcresult@cell/per_horizon_vs_joint/tab_disc_mhj_48/VMAX/r2/mean\endcsname{0.55}
\expandafter\def\csname mcresult@cell/per_horizon_vs_joint/tab_disc_mhj_48/VMAX/r2/sd\endcsname{0.02}
\expandafter\def\csname mcresult@cell/per_horizon_vs_joint/tab_disc_mhj_72/VMAX/mae/mean\endcsname{16.74}
\expandafter\def\csname mcresult@cell/per_horizon_vs_joint/tab_disc_mhj_72/VMAX/mae/sd\endcsname{0.41}
\expandafter\def\csname mcresult@cell/per_horizon_vs_joint/tab_disc_mhj_72/VMAX/r2/mean\endcsname{0.44}
\expandafter\def\csname mcresult@cell/per_horizon_vs_joint/tab_disc_mhj_72/VMAX/r2/sd\endcsname{0.03}
\expandafter\def\csname mcresult@cell/per_horizon_vs_joint/tab_disc_mhj_96/VMAX/mae/mean\endcsname{19.55}
\expandafter\def\csname mcresult@cell/per_horizon_vs_joint/tab_disc_mhj_96/VMAX/mae/sd\endcsname{0.47}
\expandafter\def\csname mcresult@cell/per_horizon_vs_joint/tab_disc_mhj_96/VMAX/r2/mean\endcsname{0.20}
\expandafter\def\csname mcresult@cell/per_horizon_vs_joint/tab_disc_mhj_96/VMAX/r2/sd\endcsname{0.06}
\expandafter\def\csname mcresult@cell/per_horizon_vs_joint/tab_fph_120/VMAX/mae/mean\endcsname{20.17}
\expandafter\def\csname mcresult@cell/per_horizon_vs_joint/tab_fph_120/VMAX/mae/sd\endcsname{0.09}
\expandafter\def\csname mcresult@cell/per_horizon_vs_joint/tab_fph_120/VMAX/r2/mean\endcsname{0.21}
\expandafter\def\csname mcresult@cell/per_horizon_vs_joint/tab_fph_120/VMAX/r2/sd\endcsname{0.03}
\expandafter\def\csname mcresult@cell/per_horizon_vs_joint/tab_fph_24/VMAX/mae/mean\endcsname{10.11}
\expandafter\def\csname mcresult@cell/per_horizon_vs_joint/tab_fph_24/VMAX/mae/sd\endcsname{0.06}
\expandafter\def\csname mcresult@cell/per_horizon_vs_joint/tab_fph_24/VMAX/r2/mean\endcsname{0.44}
\expandafter\def\csname mcresult@cell/per_horizon_vs_joint/tab_fph_24/VMAX/r2/sd\endcsname{0.01}
\expandafter\def\csname mcresult@cell/per_horizon_vs_joint/tab_fph_48/VMAX/mae/mean\endcsname{15.32}
\expandafter\def\csname mcresult@cell/per_horizon_vs_joint/tab_fph_48/VMAX/mae/sd\endcsname{0.38}
\expandafter\def\csname mcresult@cell/per_horizon_vs_joint/tab_fph_48/VMAX/r2/mean\endcsname{0.47}
\expandafter\def\csname mcresult@cell/per_horizon_vs_joint/tab_fph_48/VMAX/r2/sd\endcsname{0.02}
\expandafter\def\csname mcresult@cell/per_horizon_vs_joint/tab_fph_72/VMAX/mae/mean\endcsname{18.77}
\expandafter\def\csname mcresult@cell/per_horizon_vs_joint/tab_fph_72/VMAX/mae/sd\endcsname{0.58}
\expandafter\def\csname mcresult@cell/per_horizon_vs_joint/tab_fph_72/VMAX/r2/mean\endcsname{0.39}
\expandafter\def\csname mcresult@cell/per_horizon_vs_joint/tab_fph_72/VMAX/r2/sd\endcsname{0.02}
\expandafter\def\csname mcresult@cell/per_horizon_vs_joint/tab_fph_96/VMAX/mae/mean\endcsname{19.81}
\expandafter\def\csname mcresult@cell/per_horizon_vs_joint/tab_fph_96/VMAX/mae/sd\endcsname{0.42}
\expandafter\def\csname mcresult@cell/per_horizon_vs_joint/tab_fph_96/VMAX/r2/mean\endcsname{0.27}
\expandafter\def\csname mcresult@cell/per_horizon_vs_joint/tab_fph_96/VMAX/r2/sd\endcsname{0.03}
\expandafter\def\csname mcresult@cell/per_horizon_vs_joint/tab_img_fph_120/VMAX/mae/mean\endcsname{21.28}
\expandafter\def\csname mcresult@cell/per_horizon_vs_joint/tab_img_fph_120/VMAX/mae/sd\endcsname{0.87}
\expandafter\def\csname mcresult@cell/per_horizon_vs_joint/tab_img_fph_120/VMAX/r2/mean\endcsname{0.08}
\expandafter\def\csname mcresult@cell/per_horizon_vs_joint/tab_img_fph_120/VMAX/r2/sd\endcsname{0.09}
\expandafter\def\csname mcresult@cell/per_horizon_vs_joint/tab_img_fph_24/VMAX/mae/mean\endcsname{10.09}
\expandafter\def\csname mcresult@cell/per_horizon_vs_joint/tab_img_fph_24/VMAX/mae/sd\endcsname{0.11}
\expandafter\def\csname mcresult@cell/per_horizon_vs_joint/tab_img_fph_24/VMAX/r2/mean\endcsname{0.46}
\expandafter\def\csname mcresult@cell/per_horizon_vs_joint/tab_img_fph_24/VMAX/r2/sd\endcsname{0.02}
\expandafter\def\csname mcresult@cell/per_horizon_vs_joint/tab_img_fph_48/VMAX/mae/mean\endcsname{15.49}
\expandafter\def\csname mcresult@cell/per_horizon_vs_joint/tab_img_fph_48/VMAX/mae/sd\endcsname{0.54}
\expandafter\def\csname mcresult@cell/per_horizon_vs_joint/tab_img_fph_48/VMAX/r2/mean\endcsname{0.45}
\expandafter\def\csname mcresult@cell/per_horizon_vs_joint/tab_img_fph_48/VMAX/r2/sd\endcsname{0.02}
\expandafter\def\csname mcresult@cell/per_horizon_vs_joint/tab_img_fph_72/VMAX/mae/mean\endcsname{18.34}
\expandafter\def\csname mcresult@cell/per_horizon_vs_joint/tab_img_fph_72/VMAX/mae/sd\endcsname{0.44}
\expandafter\def\csname mcresult@cell/per_horizon_vs_joint/tab_img_fph_72/VMAX/r2/mean\endcsname{0.41}
\expandafter\def\csname mcresult@cell/per_horizon_vs_joint/tab_img_fph_72/VMAX/r2/sd\endcsname{0.03}
\expandafter\def\csname mcresult@cell/per_horizon_vs_joint/tab_img_fph_96/VMAX/mae/mean\endcsname{19.55}
\expandafter\def\csname mcresult@cell/per_horizon_vs_joint/tab_img_fph_96/VMAX/mae/sd\endcsname{0.95}
\expandafter\def\csname mcresult@cell/per_horizon_vs_joint/tab_img_fph_96/VMAX/r2/mean\endcsname{0.28}
\expandafter\def\csname mcresult@cell/per_horizon_vs_joint/tab_img_fph_96/VMAX/r2/sd\endcsname{0.08}
\expandafter\def\csname mcresult@cell/per_horizon_vs_joint/tab_img_mhj_120/VMAX/mae/mean\endcsname{20.22}
\expandafter\def\csname mcresult@cell/per_horizon_vs_joint/tab_img_mhj_120/VMAX/mae/sd\endcsname{0.53}
\expandafter\def\csname mcresult@cell/per_horizon_vs_joint/tab_img_mhj_120/VMAX/r2/mean\endcsname{0.20}
\expandafter\def\csname mcresult@cell/per_horizon_vs_joint/tab_img_mhj_120/VMAX/r2/sd\endcsname{0.04}
\expandafter\def\csname mcresult@cell/per_horizon_vs_joint/tab_img_mhj_24/VMAX/mae/mean\endcsname{10.44}
\expandafter\def\csname mcresult@cell/per_horizon_vs_joint/tab_img_mhj_24/VMAX/mae/sd\endcsname{0.09}
\expandafter\def\csname mcresult@cell/per_horizon_vs_joint/tab_img_mhj_24/VMAX/r2/mean\endcsname{0.41}
\expandafter\def\csname mcresult@cell/per_horizon_vs_joint/tab_img_mhj_24/VMAX/r2/sd\endcsname{0.01}
\expandafter\def\csname mcresult@cell/per_horizon_vs_joint/tab_img_mhj_48/VMAX/mae/mean\endcsname{15.50}
\expandafter\def\csname mcresult@cell/per_horizon_vs_joint/tab_img_mhj_48/VMAX/mae/sd\endcsname{0.30}
\expandafter\def\csname mcresult@cell/per_horizon_vs_joint/tab_img_mhj_48/VMAX/r2/mean\endcsname{0.46}
\expandafter\def\csname mcresult@cell/per_horizon_vs_joint/tab_img_mhj_48/VMAX/r2/sd\endcsname{0.01}
\expandafter\def\csname mcresult@cell/per_horizon_vs_joint/tab_img_mhj_72/VMAX/mae/mean\endcsname{17.99}
\expandafter\def\csname mcresult@cell/per_horizon_vs_joint/tab_img_mhj_72/VMAX/mae/sd\endcsname{0.48}
\expandafter\def\csname mcresult@cell/per_horizon_vs_joint/tab_img_mhj_72/VMAX/r2/mean\endcsname{0.42}
\expandafter\def\csname mcresult@cell/per_horizon_vs_joint/tab_img_mhj_72/VMAX/r2/sd\endcsname{0.02}
\expandafter\def\csname mcresult@cell/per_horizon_vs_joint/tab_img_mhj_96/VMAX/mae/mean\endcsname{19.19}
\expandafter\def\csname mcresult@cell/per_horizon_vs_joint/tab_img_mhj_96/VMAX/mae/sd\endcsname{0.65}
\expandafter\def\csname mcresult@cell/per_horizon_vs_joint/tab_img_mhj_96/VMAX/r2/mean\endcsname{0.30}
\expandafter\def\csname mcresult@cell/per_horizon_vs_joint/tab_img_mhj_96/VMAX/r2/sd\endcsname{0.04}
\expandafter\def\csname mcresult@cell/per_horizon_vs_joint/tab_mhj_120/VMAX/mae/mean\endcsname{20.23}
\expandafter\def\csname mcresult@cell/per_horizon_vs_joint/tab_mhj_120/VMAX/mae/sd\endcsname{0.51}
\expandafter\def\csname mcresult@cell/per_horizon_vs_joint/tab_mhj_120/VMAX/r2/mean\endcsname{0.19}
\expandafter\def\csname mcresult@cell/per_horizon_vs_joint/tab_mhj_120/VMAX/r2/sd\endcsname{0.04}
\expandafter\def\csname mcresult@cell/per_horizon_vs_joint/tab_mhj_24/VMAX/mae/mean\endcsname{10.32}
\expandafter\def\csname mcresult@cell/per_horizon_vs_joint/tab_mhj_24/VMAX/mae/sd\endcsname{0.10}
\expandafter\def\csname mcresult@cell/per_horizon_vs_joint/tab_mhj_24/VMAX/r2/mean\endcsname{0.41}
\expandafter\def\csname mcresult@cell/per_horizon_vs_joint/tab_mhj_24/VMAX/r2/sd\endcsname{0.01}
\expandafter\def\csname mcresult@cell/per_horizon_vs_joint/tab_mhj_48/VMAX/mae/mean\endcsname{15.27}
\expandafter\def\csname mcresult@cell/per_horizon_vs_joint/tab_mhj_48/VMAX/mae/sd\endcsname{0.30}
\expandafter\def\csname mcresult@cell/per_horizon_vs_joint/tab_mhj_48/VMAX/r2/mean\endcsname{0.47}
\expandafter\def\csname mcresult@cell/per_horizon_vs_joint/tab_mhj_48/VMAX/r2/sd\endcsname{0.01}
\expandafter\def\csname mcresult@cell/per_horizon_vs_joint/tab_mhj_72/VMAX/mae/mean\endcsname{18.05}
\expandafter\def\csname mcresult@cell/per_horizon_vs_joint/tab_mhj_72/VMAX/mae/sd\endcsname{0.20}
\expandafter\def\csname mcresult@cell/per_horizon_vs_joint/tab_mhj_72/VMAX/r2/mean\endcsname{0.42}
\expandafter\def\csname mcresult@cell/per_horizon_vs_joint/tab_mhj_72/VMAX/r2/sd\endcsname{0.01}
\expandafter\def\csname mcresult@cell/per_horizon_vs_joint/tab_mhj_96/VMAX/mae/mean\endcsname{19.32}
\expandafter\def\csname mcresult@cell/per_horizon_vs_joint/tab_mhj_96/VMAX/mae/sd\endcsname{0.37}
\expandafter\def\csname mcresult@cell/per_horizon_vs_joint/tab_mhj_96/VMAX/r2/mean\endcsname{0.30}
\expandafter\def\csname mcresult@cell/per_horizon_vs_joint/tab_mhj_96/VMAX/r2/sd\endcsname{0.03}
\expandafter\def\csname mcresult@cell/physical_baselines/120_exact_h_hour_trend/maeMSLP/mean\endcsname{20.38}
\expandafter\def\csname mcresult@cell/physical_baselines/120_exact_h_hour_trend/maeVMAX/mean\endcsname{27.67}
\expandafter\def\csname mcresult@cell/physical_baselines/120_exact_h_hour_trend/r2MSLP/mean\endcsname{-0.68}
\expandafter\def\csname mcresult@cell/physical_baselines/120_exact_h_hour_trend/r2VMAX/mean\endcsname{-0.56}
\expandafter\def\csname mcresult@cell/physical_baselines/120_exact_h_hour_trend/rmseMSLP/mean\endcsname{28.31}
\expandafter\def\csname mcresult@cell/physical_baselines/120_exact_h_hour_trend/rmseVMAX/mean\endcsname{36.57}
\expandafter\def\csname mcresult@cell/physical_baselines/120_persistence/maeMSLP/mean\endcsname{18.94}
\expandafter\def\csname mcresult@cell/physical_baselines/120_persistence/maeVMAX/mean\endcsname{25.68}
\expandafter\def\csname mcresult@cell/physical_baselines/120_persistence/r2MSLP/mean\endcsname{-0.46}
\expandafter\def\csname mcresult@cell/physical_baselines/120_persistence/r2VMAX/mean\endcsname{-0.32}
\expandafter\def\csname mcresult@cell/physical_baselines/120_persistence/rmseMSLP/mean\endcsname{26.39}
\expandafter\def\csname mcresult@cell/physical_baselines/120_persistence/rmseVMAX/mean\endcsname{33.72}
\expandafter\def\csname mcresult@cell/physical_baselines/120_train_mean_climatology/maeMSLP/mean\endcsname{17.18}
\expandafter\def\csname mcresult@cell/physical_baselines/120_train_mean_climatology/maeVMAX/mean\endcsname{24.34}
\expandafter\def\csname mcresult@cell/physical_baselines/120_train_mean_climatology/r2MSLP/mean\endcsname{-0.18}
\expandafter\def\csname mcresult@cell/physical_baselines/120_train_mean_climatology/r2VMAX/mean\endcsname{-0.19}
\expandafter\def\csname mcresult@cell/physical_baselines/120_train_mean_climatology/rmseMSLP/mean\endcsname{23.75}
\expandafter\def\csname mcresult@cell/physical_baselines/120_train_mean_climatology/rmseVMAX/mean\endcsname{32.03}
\expandafter\def\csname mcresult@cell/physical_baselines/24_exact_h_hour_trend/maeMSLP/mean\endcsname{10.32}
\expandafter\def\csname mcresult@cell/physical_baselines/24_exact_h_hour_trend/maeVMAX/mean\endcsname{16.12}
\expandafter\def\csname mcresult@cell/physical_baselines/24_exact_h_hour_trend/r2MSLP/mean\endcsname{-0.60}
\expandafter\def\csname mcresult@cell/physical_baselines/24_exact_h_hour_trend/r2VMAX/mean\endcsname{-0.54}
\expandafter\def\csname mcresult@cell/physical_baselines/24_exact_h_hour_trend/rmseMSLP/mean\endcsname{16.51}
\expandafter\def\csname mcresult@cell/physical_baselines/24_exact_h_hour_trend/rmseVMAX/mean\endcsname{23.63}
\expandafter\def\csname mcresult@cell/physical_baselines/24_persistence/maeMSLP/mean\endcsname{8.86}
\expandafter\def\csname mcresult@cell/physical_baselines/24_persistence/maeVMAX/mean\endcsname{13.78}
\expandafter\def\csname mcresult@cell/physical_baselines/24_persistence/r2MSLP/mean\endcsname{-0.04}
\expandafter\def\csname mcresult@cell/physical_baselines/24_persistence/r2VMAX/mean\endcsname{-0.03}
\expandafter\def\csname mcresult@cell/physical_baselines/24_persistence/rmseMSLP/mean\endcsname{13.29}
\expandafter\def\csname mcresult@cell/physical_baselines/24_persistence/rmseVMAX/mean\endcsname{19.33}
\expandafter\def\csname mcresult@cell/physical_baselines/24_train_mean_climatology/maeMSLP/mean\endcsname{8.73}
\expandafter\def\csname mcresult@cell/physical_baselines/24_train_mean_climatology/maeVMAX/mean\endcsname{13.76}
\expandafter\def\csname mcresult@cell/physical_baselines/24_train_mean_climatology/r2MSLP/mean\endcsname{-0.01}
\expandafter\def\csname mcresult@cell/physical_baselines/24_train_mean_climatology/r2VMAX/mean\endcsname{-0.01}
\expandafter\def\csname mcresult@cell/physical_baselines/24_train_mean_climatology/rmseMSLP/mean\endcsname{13.11}
\expandafter\def\csname mcresult@cell/physical_baselines/24_train_mean_climatology/rmseVMAX/mean\endcsname{19.15}
\expandafter\def\csname mcresult@cell/physical_baselines/48_exact_h_hour_trend/maeMSLP/mean\endcsname{18.80}
\expandafter\def\csname mcresult@cell/physical_baselines/48_exact_h_hour_trend/maeVMAX/mean\endcsname{28.30}
\expandafter\def\csname mcresult@cell/physical_baselines/48_exact_h_hour_trend/r2MSLP/mean\endcsname{-1.05}
\expandafter\def\csname mcresult@cell/physical_baselines/48_exact_h_hour_trend/r2VMAX/mean\endcsname{-1.01}
\expandafter\def\csname mcresult@cell/physical_baselines/48_exact_h_hour_trend/rmseMSLP/mean\endcsname{28.20}
\expandafter\def\csname mcresult@cell/physical_baselines/48_exact_h_hour_trend/rmseVMAX/mean\endcsname{40.36}
\expandafter\def\csname mcresult@cell/physical_baselines/48_persistence/maeMSLP/mean\endcsname{14.46}
\expandafter\def\csname mcresult@cell/physical_baselines/48_persistence/maeVMAX/mean\endcsname{21.49}
\expandafter\def\csname mcresult@cell/physical_baselines/48_persistence/r2MSLP/mean\endcsname{-0.08}
\expandafter\def\csname mcresult@cell/physical_baselines/48_persistence/r2VMAX/mean\endcsname{-0.06}
\expandafter\def\csname mcresult@cell/physical_baselines/48_persistence/rmseMSLP/mean\endcsname{20.49}
\expandafter\def\csname mcresult@cell/physical_baselines/48_persistence/rmseVMAX/mean\endcsname{29.31}
\expandafter\def\csname mcresult@cell/physical_baselines/48_train_mean_climatology/maeMSLP/mean\endcsname{14.05}
\expandafter\def\csname mcresult@cell/physical_baselines/48_train_mean_climatology/maeVMAX/mean\endcsname{21.16}
\expandafter\def\csname mcresult@cell/physical_baselines/48_train_mean_climatology/r2MSLP/mean\endcsname{-0.02}
\expandafter\def\csname mcresult@cell/physical_baselines/48_train_mean_climatology/r2VMAX/mean\endcsname{-0.02}
\expandafter\def\csname mcresult@cell/physical_baselines/48_train_mean_climatology/rmseMSLP/mean\endcsname{19.88}
\expandafter\def\csname mcresult@cell/physical_baselines/48_train_mean_climatology/rmseVMAX/mean\endcsname{28.76}
\expandafter\def\csname mcresult@cell/physical_baselines/72_exact_h_hour_trend/maeMSLP/mean\endcsname{19.86}
\expandafter\def\csname mcresult@cell/physical_baselines/72_exact_h_hour_trend/maeVMAX/mean\endcsname{29.01}
\expandafter\def\csname mcresult@cell/physical_baselines/72_exact_h_hour_trend/r2MSLP/mean\endcsname{-0.71}
\expandafter\def\csname mcresult@cell/physical_baselines/72_exact_h_hour_trend/r2VMAX/mean\endcsname{-0.69}
\expandafter\def\csname mcresult@cell/physical_baselines/72_exact_h_hour_trend/rmseMSLP/mean\endcsname{27.92}
\expandafter\def\csname mcresult@cell/physical_baselines/72_exact_h_hour_trend/rmseVMAX/mean\endcsname{39.57}
\expandafter\def\csname mcresult@cell/physical_baselines/72_persistence/maeMSLP/mean\endcsname{16.62}
\expandafter\def\csname mcresult@cell/physical_baselines/72_persistence/maeVMAX/mean\endcsname{23.79}
\expandafter\def\csname mcresult@cell/physical_baselines/72_persistence/r2MSLP/mean\endcsname{-0.14}
\expandafter\def\csname mcresult@cell/physical_baselines/72_persistence/r2VMAX/mean\endcsname{-0.09}
\expandafter\def\csname mcresult@cell/physical_baselines/72_persistence/rmseMSLP/mean\endcsname{22.82}
\expandafter\def\csname mcresult@cell/physical_baselines/72_persistence/rmseVMAX/mean\endcsname{31.82}
\expandafter\def\csname mcresult@cell/physical_baselines/72_train_mean_climatology/maeMSLP/mean\endcsname{15.89}
\expandafter\def\csname mcresult@cell/physical_baselines/72_train_mean_climatology/maeVMAX/mean\endcsname{23.18}
\expandafter\def\csname mcresult@cell/physical_baselines/72_train_mean_climatology/r2MSLP/mean\endcsname{-0.03}
\expandafter\def\csname mcresult@cell/physical_baselines/72_train_mean_climatology/r2VMAX/mean\endcsname{-0.03}
\expandafter\def\csname mcresult@cell/physical_baselines/72_train_mean_climatology/rmseMSLP/mean\endcsname{21.72}
\expandafter\def\csname mcresult@cell/physical_baselines/72_train_mean_climatology/rmseVMAX/mean\endcsname{30.93}
\expandafter\def\csname mcresult@cell/physical_baselines/96_exact_h_hour_trend/maeMSLP/mean\endcsname{19.41}
\expandafter\def\csname mcresult@cell/physical_baselines/96_exact_h_hour_trend/maeVMAX/mean\endcsname{27.22}
\expandafter\def\csname mcresult@cell/physical_baselines/96_exact_h_hour_trend/r2MSLP/mean\endcsname{-0.52}
\expandafter\def\csname mcresult@cell/physical_baselines/96_exact_h_hour_trend/r2VMAX/mean\endcsname{-0.43}
\expandafter\def\csname mcresult@cell/physical_baselines/96_exact_h_hour_trend/rmseMSLP/mean\endcsname{26.03}
\expandafter\def\csname mcresult@cell/physical_baselines/96_exact_h_hour_trend/rmseVMAX/mean\endcsname{35.09}
\expandafter\def\csname mcresult@cell/physical_baselines/96_persistence/maeMSLP/mean\endcsname{17.43}
\expandafter\def\csname mcresult@cell/physical_baselines/96_persistence/maeVMAX/mean\endcsname{24.33}
\expandafter\def\csname mcresult@cell/physical_baselines/96_persistence/r2MSLP/mean\endcsname{-0.27}
\expandafter\def\csname mcresult@cell/physical_baselines/96_persistence/r2VMAX/mean\endcsname{-0.18}
\expandafter\def\csname mcresult@cell/physical_baselines/96_persistence/rmseMSLP/mean\endcsname{23.78}
\expandafter\def\csname mcresult@cell/physical_baselines/96_persistence/rmseVMAX/mean\endcsname{31.79}
\expandafter\def\csname mcresult@cell/physical_baselines/96_train_mean_climatology/maeMSLP/mean\endcsname{16.15}
\expandafter\def\csname mcresult@cell/physical_baselines/96_train_mean_climatology/maeVMAX/mean\endcsname{23.46}
\expandafter\def\csname mcresult@cell/physical_baselines/96_train_mean_climatology/r2MSLP/mean\endcsname{-0.09}
\expandafter\def\csname mcresult@cell/physical_baselines/96_train_mean_climatology/r2VMAX/mean\endcsname{-0.09}
\expandafter\def\csname mcresult@cell/physical_baselines/96_train_mean_climatology/rmseMSLP/mean\endcsname{22.03}
\expandafter\def\csname mcresult@cell/physical_baselines/96_train_mean_climatology/rmseVMAX/mean\endcsname{30.62}
\expandafter\def\csname mcresult@cell/strat_gain/discussion_present/left_maeMSLP/mean\endcsname{6.55}
\expandafter\def\csname mcresult@cell/strat_gain/discussion_present/left_maeVMAX/mean\endcsname{10.05}
\expandafter\def\csname mcresult@cell/strat_gain/discussion_present/mae_gain_left_to_rightMSLP/mean\endcsname{0.61}
\expandafter\def\csname mcresult@cell/strat_gain/discussion_present/mae_gain_left_to_rightVMAX/mean\endcsname{1.07}
\expandafter\def\csname mcresult@cell/strat_gain/discussion_present/right_maeMSLP/mean\endcsname{5.94}
\expandafter\def\csname mcresult@cell/strat_gain/discussion_present/right_maeVMAX/mean\endcsname{8.98}
\expandafter\def\csname mcresult@cell/strat_gain/discussion_presentMSLP/storm_ci_high/mean\endcsname{1.12}
\expandafter\def\csname mcresult@cell/strat_gain/discussion_presentMSLP/storm_ci_low/mean\endcsname{0.10}
\expandafter\def\csname mcresult@cell/strat_gain/discussion_presentVMAX/storm_ci_high/mean\endcsname{1.82}
\expandafter\def\csname mcresult@cell/strat_gain/discussion_presentVMAX/storm_ci_low/mean\endcsname{0.37}
\expandafter\def\csname mcresult@cell/strat_gain/regex_not_opinionated_matched/left_maeMSLP/mean\endcsname{6.55}
\expandafter\def\csname mcresult@cell/strat_gain/regex_not_opinionated_matched/left_maeVMAX/mean\endcsname{10.04}
\expandafter\def\csname mcresult@cell/strat_gain/regex_not_opinionated_matched/mae_gain_left_to_rightMSLP/mean\endcsname{0.62}
\expandafter\def\csname mcresult@cell/strat_gain/regex_not_opinionated_matched/mae_gain_left_to_rightVMAX/mean\endcsname{1.10}
\expandafter\def\csname mcresult@cell/strat_gain/regex_not_opinionated_matched/right_maeMSLP/mean\endcsname{5.93}
\expandafter\def\csname mcresult@cell/strat_gain/regex_not_opinionated_matched/right_maeVMAX/mean\endcsname{8.94}
\expandafter\def\csname mcresult@cell/strat_gain/regex_not_opinionated_matchedMSLP/storm_ci_high/mean\endcsname{1.14}
\expandafter\def\csname mcresult@cell/strat_gain/regex_not_opinionated_matchedMSLP/storm_ci_low/mean\endcsname{0.10}
\expandafter\def\csname mcresult@cell/strat_gain/regex_not_opinionated_matchedVMAX/storm_ci_high/mean\endcsname{1.85}
\expandafter\def\csname mcresult@cell/strat_gain/regex_not_opinionated_matchedVMAX/storm_ci_low/mean\endcsname{0.39}
\expandafter\def\csname mcresult@cell/strat_gain/regex_opinionated_matched/left_maeMSLP/mean\endcsname{6.86}
\expandafter\def\csname mcresult@cell/strat_gain/regex_opinionated_matched/left_maeVMAX/mean\endcsname{10.81}
\expandafter\def\csname mcresult@cell/strat_gain/regex_opinionated_matched/mae_gain_left_to_rightMSLP/mean\endcsname{-0.12}
\expandafter\def\csname mcresult@cell/strat_gain/regex_opinionated_matched/mae_gain_left_to_rightVMAX/mean\endcsname{-0.65}
\expandafter\def\csname mcresult@cell/strat_gain/regex_opinionated_matched/right_maeMSLP/mean\endcsname{6.98}
\expandafter\def\csname mcresult@cell/strat_gain/regex_opinionated_matched/right_maeVMAX/mean\endcsname{11.46}
\expandafter\def\csname mcresult@cell/strat_gain/regex_opinionated_matchedMSLP/storm_ci_high/mean\endcsname{2.49}
\expandafter\def\csname mcresult@cell/strat_gain/regex_opinionated_matchedMSLP/storm_ci_low/mean\endcsname{-2.64}
\expandafter\def\csname mcresult@cell/strat_gain/regex_opinionated_matchedVMAX/storm_ci_high/mean\endcsname{3.45}
\expandafter\def\csname mcresult@cell/strat_gain/regex_opinionated_matchedVMAX/storm_ci_low/mean\endcsname{-4.51}
\expandafter\def\csname mcresult@cell/temporal_common_subset/temporal_commone2eTD_aligned/MSLP/mae/mean\endcsname{6.33}
\expandafter\def\csname mcresult@cell/temporal_common_subset/temporal_commone2eTD_aligned/MSLP/mae/sd\endcsname{0.12}
\expandafter\def\csname mcresult@cell/temporal_common_subset/temporal_commone2eTD_aligned/MSLP/r2/mean\endcsname{0.50}
\expandafter\def\csname mcresult@cell/temporal_common_subset/temporal_commone2eTD_aligned/MSLP/r2/sd\endcsname{0.02}
\expandafter\def\csname mcresult@cell/temporal_common_subset/temporal_commone2eTD_aligned/VMAX/mae/mean\endcsname{9.47}
\expandafter\def\csname mcresult@cell/temporal_common_subset/temporal_commone2eTD_aligned/VMAX/mae/sd\endcsname{0.20}
\expandafter\def\csname mcresult@cell/temporal_common_subset/temporal_commone2eTD_aligned/VMAX/r2/mean\endcsname{0.49}
\expandafter\def\csname mcresult@cell/temporal_common_subset/temporal_commone2eTD_aligned/VMAX/r2/sd\endcsname{0.02}
\expandafter\def\csname mcresult@cell/temporal_common_subset/temporal_commone2eTID_aligned/MSLP/mae/mean\endcsname{6.31}
\expandafter\def\csname mcresult@cell/temporal_common_subset/temporal_commone2eTID_aligned/MSLP/mae/sd\endcsname{0.06}
\expandafter\def\csname mcresult@cell/temporal_common_subset/temporal_commone2eTID_aligned/MSLP/r2/mean\endcsname{0.50}
\expandafter\def\csname mcresult@cell/temporal_common_subset/temporal_commone2eTID_aligned/MSLP/r2/sd\endcsname{0.01}
\expandafter\def\csname mcresult@cell/temporal_common_subset/temporal_commone2eTID_aligned/VMAX/mae/mean\endcsname{9.45}
\expandafter\def\csname mcresult@cell/temporal_common_subset/temporal_commone2eTID_aligned/VMAX/mae/sd\endcsname{0.11}
\expandafter\def\csname mcresult@cell/temporal_common_subset/temporal_commone2eTID_aligned/VMAX/r2/mean\endcsname{0.49}
\expandafter\def\csname mcresult@cell/temporal_common_subset/temporal_commone2eTID_aligned/VMAX/r2/sd\endcsname{0.01}
\expandafter\def\csname mcresult@cell/temporal_common_subset/temporal_commone2eT_reference/MSLP/mae/mean\endcsname{6.70}
\expandafter\def\csname mcresult@cell/temporal_common_subset/temporal_commone2eT_reference/MSLP/mae/sd\endcsname{0.12}
\expandafter\def\csname mcresult@cell/temporal_common_subset/temporal_commone2eT_reference/MSLP/r2/mean\endcsname{0.44}
\expandafter\def\csname mcresult@cell/temporal_common_subset/temporal_commone2eT_reference/MSLP/r2/sd\endcsname{0.02}
\expandafter\def\csname mcresult@cell/temporal_common_subset/temporal_commone2eT_reference/VMAX/mae/mean\endcsname{10.03}
\expandafter\def\csname mcresult@cell/temporal_common_subset/temporal_commone2eT_reference/VMAX/mae/sd\endcsname{0.13}
\expandafter\def\csname mcresult@cell/temporal_common_subset/temporal_commone2eT_reference/VMAX/r2/mean\endcsname{0.41}
\expandafter\def\csname mcresult@cell/temporal_common_subset/temporal_commone2eT_reference/VMAX/r2/sd\endcsname{0.01}
\expandafter\def\csname mcresult@cell/temporal_common_subset/temporal_commonhybridXGB_Tab/MSLP/mae/mean\endcsname{6.74}
\expandafter\def\csname mcresult@cell/temporal_common_subset/temporal_commonhybridXGB_Tab/MSLP/mae/sd\endcsname{0.02}
\expandafter\def\csname mcresult@cell/temporal_common_subset/temporal_commonhybridXGB_Tab/MSLP/r2/mean\endcsname{0.43}
\expandafter\def\csname mcresult@cell/temporal_common_subset/temporal_commonhybridXGB_Tab/MSLP/r2/sd\endcsname{0.00}
\expandafter\def\csname mcresult@cell/temporal_common_subset/temporal_commonhybridXGB_Tab/VMAX/mae/mean\endcsname{9.84}
\expandafter\def\csname mcresult@cell/temporal_common_subset/temporal_commonhybridXGB_Tab/VMAX/mae/sd\endcsname{0.07}
\expandafter\def\csname mcresult@cell/temporal_common_subset/temporal_commonhybridXGB_Tab/VMAX/r2/mean\endcsname{0.42}
\expandafter\def\csname mcresult@cell/temporal_common_subset/temporal_commonhybridXGB_Tab/VMAX/r2/sd\endcsname{0.01}
\expandafter\def\csname mcresult@cell/temporal_common_subset/temporal_commonhybridXGB_Tab_Img_Text/MSLP/mae/mean\endcsname{6.63}
\expandafter\def\csname mcresult@cell/temporal_common_subset/temporal_commonhybridXGB_Tab_Img_Text/MSLP/mae/sd\endcsname{0.19}
\expandafter\def\csname mcresult@cell/temporal_common_subset/temporal_commonhybridXGB_Tab_Img_Text/MSLP/r2/mean\endcsname{0.48}
\expandafter\def\csname mcresult@cell/temporal_common_subset/temporal_commonhybridXGB_Tab_Img_Text/MSLP/r2/sd\endcsname{0.03}
\expandafter\def\csname mcresult@cell/temporal_common_subset/temporal_commonhybridXGB_Tab_Img_Text/VMAX/mae/mean\endcsname{9.89}
\expandafter\def\csname mcresult@cell/temporal_common_subset/temporal_commonhybridXGB_Tab_Img_Text/VMAX/mae/sd\endcsname{0.19}
\expandafter\def\csname mcresult@cell/temporal_common_subset/temporal_commonhybridXGB_Tab_Img_Text/VMAX/r2/mean\endcsname{0.46}
\expandafter\def\csname mcresult@cell/temporal_common_subset/temporal_commonhybridXGB_Tab_Img_Text/VMAX/r2/sd\endcsname{0.02}
\expandafter\def\csname mcresult@cell/temporal_common_subset/temporal_commonhybridXGB_Tab_Text/MSLP/mae/mean\endcsname{6.56}
\expandafter\def\csname mcresult@cell/temporal_common_subset/temporal_commonhybridXGB_Tab_Text/MSLP/mae/sd\endcsname{0.11}
\expandafter\def\csname mcresult@cell/temporal_common_subset/temporal_commonhybridXGB_Tab_Text/MSLP/r2/mean\endcsname{0.49}
\expandafter\def\csname mcresult@cell/temporal_common_subset/temporal_commonhybridXGB_Tab_Text/MSLP/r2/sd\endcsname{0.01}
\expandafter\def\csname mcresult@cell/temporal_common_subset/temporal_commonhybridXGB_Tab_Text/VMAX/mae/mean\endcsname{9.55}
\expandafter\def\csname mcresult@cell/temporal_common_subset/temporal_commonhybridXGB_Tab_Text/VMAX/mae/sd\endcsname{0.10}
\expandafter\def\csname mcresult@cell/temporal_common_subset/temporal_commonhybridXGB_Tab_Text/VMAX/r2/mean\endcsname{0.47}
\expandafter\def\csname mcresult@cell/temporal_common_subset/temporal_commonhybridXGB_Tab_Text/VMAX/r2/sd\endcsname{0.01}
\expandafter\def\csname mcresult@cell/temporal_common_subset/temporal_commontemporalNN_direct/MSLP/mae/mean\endcsname{6.63}
\expandafter\def\csname mcresult@cell/temporal_common_subset/temporal_commontemporalNN_direct/MSLP/mae/sd\endcsname{0.13}
\expandafter\def\csname mcresult@cell/temporal_common_subset/temporal_commontemporalNN_direct/MSLP/r2/mean\endcsname{0.45}
\expandafter\def\csname mcresult@cell/temporal_common_subset/temporal_commontemporalNN_direct/MSLP/r2/sd\endcsname{0.01}
\expandafter\def\csname mcresult@cell/temporal_common_subset/temporal_commontemporalNN_direct/VMAX/mae/mean\endcsname{9.92}
\expandafter\def\csname mcresult@cell/temporal_common_subset/temporal_commontemporalNN_direct/VMAX/mae/sd\endcsname{0.27}
\expandafter\def\csname mcresult@cell/temporal_common_subset/temporal_commontemporalNN_direct/VMAX/r2/mean\endcsname{0.44}
\expandafter\def\csname mcresult@cell/temporal_common_subset/temporal_commontemporalNN_direct/VMAX/r2/sd\endcsname{0.02}
\expandafter\def\csname mcresult@cell/temporal_common_subset/temporal_commontemporalXGB_ALL/MSLP/mae/mean\endcsname{6.41}
\expandafter\def\csname mcresult@cell/temporal_common_subset/temporal_commontemporalXGB_ALL/MSLP/mae/sd\endcsname{0.17}
\expandafter\def\csname mcresult@cell/temporal_common_subset/temporal_commontemporalXGB_ALL/MSLP/r2/mean\endcsname{0.47}
\expandafter\def\csname mcresult@cell/temporal_common_subset/temporal_commontemporalXGB_ALL/MSLP/r2/sd\endcsname{0.02}
\expandafter\def\csname mcresult@cell/temporal_common_subset/temporal_commontemporalXGB_ALL/VMAX/mae/mean\endcsname{9.68}
\expandafter\def\csname mcresult@cell/temporal_common_subset/temporal_commontemporalXGB_ALL/VMAX/mae/sd\endcsname{0.21}
\expandafter\def\csname mcresult@cell/temporal_common_subset/temporal_commontemporalXGB_ALL/VMAX/r2/mean\endcsname{0.46}
\expandafter\def\csname mcresult@cell/temporal_common_subset/temporal_commontemporalXGB_ALL/VMAX/r2/sd\endcsname{0.02}
\expandafter\def\csname mcresult@cell/temporal_common_subset/temporal_commontemporalXGB_emb_only/MSLP/mae/mean\endcsname{6.43}
\expandafter\def\csname mcresult@cell/temporal_common_subset/temporal_commontemporalXGB_emb_only/MSLP/mae/sd\endcsname{0.18}
\expandafter\def\csname mcresult@cell/temporal_common_subset/temporal_commontemporalXGB_emb_only/MSLP/r2/mean\endcsname{0.47}
\expandafter\def\csname mcresult@cell/temporal_common_subset/temporal_commontemporalXGB_emb_only/MSLP/r2/sd\endcsname{0.02}
\expandafter\def\csname mcresult@cell/temporal_common_subset/temporal_commontemporalXGB_emb_only/VMAX/mae/mean\endcsname{9.71}
\expandafter\def\csname mcresult@cell/temporal_common_subset/temporal_commontemporalXGB_emb_only/VMAX/mae/sd\endcsname{0.27}
\expandafter\def\csname mcresult@cell/temporal_common_subset/temporal_commontemporalXGB_emb_only/VMAX/r2/mean\endcsname{0.45}
\expandafter\def\csname mcresult@cell/temporal_common_subset/temporal_commontemporalXGB_emb_only/VMAX/r2/sd\endcsname{0.02}
\expandafter\def\csname mcresult@cell/temporal_common_subset/temporal_commontemporal_ablationCNN_Tab_Text_emb/MSLP/mae/mean\endcsname{6.43}
\expandafter\def\csname mcresult@cell/temporal_common_subset/temporal_commontemporal_ablationCNN_Tab_Text_emb/MSLP/mae/sd\endcsname{0.18}
\expandafter\def\csname mcresult@cell/temporal_common_subset/temporal_commontemporal_ablationCNN_Tab_Text_emb/MSLP/r2/mean\endcsname{0.47}
\expandafter\def\csname mcresult@cell/temporal_common_subset/temporal_commontemporal_ablationCNN_Tab_Text_emb/MSLP/r2/sd\endcsname{0.02}
\expandafter\def\csname mcresult@cell/temporal_common_subset/temporal_commontemporal_ablationCNN_Tab_Text_emb/VMAX/mae/mean\endcsname{9.71}
\expandafter\def\csname mcresult@cell/temporal_common_subset/temporal_commontemporal_ablationCNN_Tab_Text_emb/VMAX/mae/sd\endcsname{0.27}
\expandafter\def\csname mcresult@cell/temporal_common_subset/temporal_commontemporal_ablationCNN_Tab_Text_emb/VMAX/r2/mean\endcsname{0.45}
\expandafter\def\csname mcresult@cell/temporal_common_subset/temporal_commontemporal_ablationCNN_Tab_Text_emb/VMAX/r2/sd\endcsname{0.02}
\expandafter\def\csname mcresult@cell/temporal_common_subset/temporal_commontemporal_ablationCNN_Tab_emb/MSLP/mae/mean\endcsname{7.05}
\expandafter\def\csname mcresult@cell/temporal_common_subset/temporal_commontemporal_ablationCNN_Tab_emb/MSLP/mae/sd\endcsname{0.10}
\expandafter\def\csname mcresult@cell/temporal_common_subset/temporal_commontemporal_ablationCNN_Tab_emb/MSLP/r2/mean\endcsname{0.37}
\expandafter\def\csname mcresult@cell/temporal_common_subset/temporal_commontemporal_ablationCNN_Tab_emb/MSLP/r2/sd\endcsname{0.03}
\expandafter\def\csname mcresult@cell/temporal_common_subset/temporal_commontemporal_ablationCNN_Tab_emb/VMAX/mae/mean\endcsname{10.56}
\expandafter\def\csname mcresult@cell/temporal_common_subset/temporal_commontemporal_ablationCNN_Tab_emb/VMAX/mae/sd\endcsname{0.11}
\expandafter\def\csname mcresult@cell/temporal_common_subset/temporal_commontemporal_ablationCNN_Tab_emb/VMAX/r2/mean\endcsname{0.36}
\expandafter\def\csname mcresult@cell/temporal_common_subset/temporal_commontemporal_ablationCNN_Tab_emb/VMAX/r2/sd\endcsname{0.01}
\expandafter\def\csname mcresult@cell/temporal_common_subset/temporal_matched_commone2eD_reference/MSLP/mae/mean\endcsname{6.65}
\expandafter\def\csname mcresult@cell/temporal_common_subset/temporal_matched_commone2eD_reference/MSLP/mae/sd\endcsname{0.15}
\expandafter\def\csname mcresult@cell/temporal_common_subset/temporal_matched_commone2eD_reference/MSLP/r2/mean\endcsname{0.44}
\expandafter\def\csname mcresult@cell/temporal_common_subset/temporal_matched_commone2eD_reference/MSLP/r2/sd\endcsname{0.02}
\expandafter\def\csname mcresult@cell/temporal_common_subset/temporal_matched_commone2eD_reference/VMAX/mae/mean\endcsname{9.82}
\expandafter\def\csname mcresult@cell/temporal_common_subset/temporal_matched_commone2eD_reference/VMAX/mae/sd\endcsname{0.19}
\expandafter\def\csname mcresult@cell/temporal_common_subset/temporal_matched_commone2eD_reference/VMAX/r2/mean\endcsname{0.46}
\expandafter\def\csname mcresult@cell/temporal_common_subset/temporal_matched_commone2eD_reference/VMAX/r2/sd\endcsname{0.02}
\expandafter\def\csname mcresult@cell/temporal_common_subset/temporal_matched_commone2eTD_aligned/MSLP/mae/mean\endcsname{6.35}
\expandafter\def\csname mcresult@cell/temporal_common_subset/temporal_matched_commone2eTD_aligned/MSLP/mae/sd\endcsname{0.12}
\expandafter\def\csname mcresult@cell/temporal_common_subset/temporal_matched_commone2eTD_aligned/MSLP/r2/mean\endcsname{0.50}
\expandafter\def\csname mcresult@cell/temporal_common_subset/temporal_matched_commone2eTD_aligned/MSLP/r2/sd\endcsname{0.02}
\expandafter\def\csname mcresult@cell/temporal_common_subset/temporal_matched_commone2eTD_aligned/VMAX/mae/mean\endcsname{9.52}
\expandafter\def\csname mcresult@cell/temporal_common_subset/temporal_matched_commone2eTD_aligned/VMAX/mae/sd\endcsname{0.20}
\expandafter\def\csname mcresult@cell/temporal_common_subset/temporal_matched_commone2eTD_aligned/VMAX/r2/mean\endcsname{0.50}
\expandafter\def\csname mcresult@cell/temporal_common_subset/temporal_matched_commone2eTD_aligned/VMAX/r2/sd\endcsname{0.02}
\expandafter\def\csname mcresult@cell/temporal_common_subset/temporal_matched_commone2eTID_aligned/MSLP/mae/mean\endcsname{6.34}
\expandafter\def\csname mcresult@cell/temporal_common_subset/temporal_matched_commone2eTID_aligned/MSLP/mae/sd\endcsname{0.06}
\expandafter\def\csname mcresult@cell/temporal_common_subset/temporal_matched_commone2eTID_aligned/MSLP/r2/mean\endcsname{0.50}
\expandafter\def\csname mcresult@cell/temporal_common_subset/temporal_matched_commone2eTID_aligned/MSLP/r2/sd\endcsname{0.01}
\expandafter\def\csname mcresult@cell/temporal_common_subset/temporal_matched_commone2eTID_aligned/VMAX/mae/mean\endcsname{9.51}
\expandafter\def\csname mcresult@cell/temporal_common_subset/temporal_matched_commone2eTID_aligned/VMAX/mae/sd\endcsname{0.11}
\expandafter\def\csname mcresult@cell/temporal_common_subset/temporal_matched_commone2eTID_aligned/VMAX/r2/mean\endcsname{0.49}
\expandafter\def\csname mcresult@cell/temporal_common_subset/temporal_matched_commone2eTID_aligned/VMAX/r2/sd\endcsname{0.01}
\expandafter\def\csname mcresult@cell/temporal_common_subset/temporal_matched_commone2eT_reference/MSLP/mae/mean\endcsname{6.73}
\expandafter\def\csname mcresult@cell/temporal_common_subset/temporal_matched_commone2eT_reference/MSLP/mae/sd\endcsname{0.13}
\expandafter\def\csname mcresult@cell/temporal_common_subset/temporal_matched_commone2eT_reference/MSLP/r2/mean\endcsname{0.44}
\expandafter\def\csname mcresult@cell/temporal_common_subset/temporal_matched_commone2eT_reference/MSLP/r2/sd\endcsname{0.02}
\expandafter\def\csname mcresult@cell/temporal_common_subset/temporal_matched_commone2eT_reference/VMAX/mae/mean\endcsname{10.12}
\expandafter\def\csname mcresult@cell/temporal_common_subset/temporal_matched_commone2eT_reference/VMAX/mae/sd\endcsname{0.13}
\expandafter\def\csname mcresult@cell/temporal_common_subset/temporal_matched_commone2eT_reference/VMAX/r2/mean\endcsname{0.41}
\expandafter\def\csname mcresult@cell/temporal_common_subset/temporal_matched_commone2eT_reference/VMAX/r2/sd\endcsname{0.01}
\expandafter\def\csname mcresult@cell/temporal_common_subset/temporal_matched_commonhybridXGB_Tab/MSLP/mae/mean\endcsname{6.79}
\expandafter\def\csname mcresult@cell/temporal_common_subset/temporal_matched_commonhybridXGB_Tab/MSLP/mae/sd\endcsname{0.02}
\expandafter\def\csname mcresult@cell/temporal_common_subset/temporal_matched_commonhybridXGB_Tab/MSLP/r2/mean\endcsname{0.43}
\expandafter\def\csname mcresult@cell/temporal_common_subset/temporal_matched_commonhybridXGB_Tab/MSLP/r2/sd\endcsname{0.00}
\expandafter\def\csname mcresult@cell/temporal_common_subset/temporal_matched_commonhybridXGB_Tab/VMAX/mae/mean\endcsname{9.95}
\expandafter\def\csname mcresult@cell/temporal_common_subset/temporal_matched_commonhybridXGB_Tab/VMAX/mae/sd\endcsname{0.07}
\expandafter\def\csname mcresult@cell/temporal_common_subset/temporal_matched_commonhybridXGB_Tab/VMAX/r2/mean\endcsname{0.42}
\expandafter\def\csname mcresult@cell/temporal_common_subset/temporal_matched_commonhybridXGB_Tab/VMAX/r2/sd\endcsname{0.01}
\expandafter\def\csname mcresult@cell/temporal_common_subset/temporal_matched_commonhybridXGB_Tab_Img_Text/MSLP/mae/mean\endcsname{6.69}
\expandafter\def\csname mcresult@cell/temporal_common_subset/temporal_matched_commonhybridXGB_Tab_Img_Text/MSLP/mae/sd\endcsname{0.19}
\expandafter\def\csname mcresult@cell/temporal_common_subset/temporal_matched_commonhybridXGB_Tab_Img_Text/MSLP/r2/mean\endcsname{0.48}
\expandafter\def\csname mcresult@cell/temporal_common_subset/temporal_matched_commonhybridXGB_Tab_Img_Text/MSLP/r2/sd\endcsname{0.03}
\expandafter\def\csname mcresult@cell/temporal_common_subset/temporal_matched_commonhybridXGB_Tab_Img_Text/VMAX/mae/mean\endcsname{9.98}
\expandafter\def\csname mcresult@cell/temporal_common_subset/temporal_matched_commonhybridXGB_Tab_Img_Text/VMAX/mae/sd\endcsname{0.20}
\expandafter\def\csname mcresult@cell/temporal_common_subset/temporal_matched_commonhybridXGB_Tab_Img_Text/VMAX/r2/mean\endcsname{0.46}
\expandafter\def\csname mcresult@cell/temporal_common_subset/temporal_matched_commonhybridXGB_Tab_Img_Text/VMAX/r2/sd\endcsname{0.03}
\expandafter\def\csname mcresult@cell/temporal_common_subset/temporal_matched_commonhybridXGB_Tab_Text/MSLP/mae/mean\endcsname{6.62}
\expandafter\def\csname mcresult@cell/temporal_common_subset/temporal_matched_commonhybridXGB_Tab_Text/MSLP/mae/sd\endcsname{0.12}
\expandafter\def\csname mcresult@cell/temporal_common_subset/temporal_matched_commonhybridXGB_Tab_Text/MSLP/r2/mean\endcsname{0.49}
\expandafter\def\csname mcresult@cell/temporal_common_subset/temporal_matched_commonhybridXGB_Tab_Text/MSLP/r2/sd\endcsname{0.01}
\expandafter\def\csname mcresult@cell/temporal_common_subset/temporal_matched_commonhybridXGB_Tab_Text/VMAX/mae/mean\endcsname{9.64}
\expandafter\def\csname mcresult@cell/temporal_common_subset/temporal_matched_commonhybridXGB_Tab_Text/VMAX/mae/sd\endcsname{0.10}
\expandafter\def\csname mcresult@cell/temporal_common_subset/temporal_matched_commonhybridXGB_Tab_Text/VMAX/r2/mean\endcsname{0.47}
\expandafter\def\csname mcresult@cell/temporal_common_subset/temporal_matched_commonhybridXGB_Tab_Text/VMAX/r2/sd\endcsname{0.01}
\expandafter\def\csname mcresult@cell/temporal_common_subset/temporal_matched_commontemporalNN_direct/MSLP/mae/mean\endcsname{6.66}
\expandafter\def\csname mcresult@cell/temporal_common_subset/temporal_matched_commontemporalNN_direct/MSLP/mae/sd\endcsname{0.12}
\expandafter\def\csname mcresult@cell/temporal_common_subset/temporal_matched_commontemporalNN_direct/MSLP/r2/mean\endcsname{0.45}
\expandafter\def\csname mcresult@cell/temporal_common_subset/temporal_matched_commontemporalNN_direct/MSLP/r2/sd\endcsname{0.01}
\expandafter\def\csname mcresult@cell/temporal_common_subset/temporal_matched_commontemporalNN_direct/VMAX/mae/mean\endcsname{9.98}
\expandafter\def\csname mcresult@cell/temporal_common_subset/temporal_matched_commontemporalNN_direct/VMAX/mae/sd\endcsname{0.26}
\expandafter\def\csname mcresult@cell/temporal_common_subset/temporal_matched_commontemporalNN_direct/VMAX/r2/mean\endcsname{0.44}
\expandafter\def\csname mcresult@cell/temporal_common_subset/temporal_matched_commontemporalNN_direct/VMAX/r2/sd\endcsname{0.02}
\expandafter\def\csname mcresult@cell/temporal_common_subset/temporal_matched_commontemporalXGB_ALL/MSLP/mae/mean\endcsname{6.45}
\expandafter\def\csname mcresult@cell/temporal_common_subset/temporal_matched_commontemporalXGB_ALL/MSLP/mae/sd\endcsname{0.17}
\expandafter\def\csname mcresult@cell/temporal_common_subset/temporal_matched_commontemporalXGB_ALL/MSLP/r2/mean\endcsname{0.47}
\expandafter\def\csname mcresult@cell/temporal_common_subset/temporal_matched_commontemporalXGB_ALL/MSLP/r2/sd\endcsname{0.02}
\expandafter\def\csname mcresult@cell/temporal_common_subset/temporal_matched_commontemporalXGB_ALL/VMAX/mae/mean\endcsname{9.76}
\expandafter\def\csname mcresult@cell/temporal_common_subset/temporal_matched_commontemporalXGB_ALL/VMAX/mae/sd\endcsname{0.21}
\expandafter\def\csname mcresult@cell/temporal_common_subset/temporal_matched_commontemporalXGB_ALL/VMAX/r2/mean\endcsname{0.46}
\expandafter\def\csname mcresult@cell/temporal_common_subset/temporal_matched_commontemporalXGB_ALL/VMAX/r2/sd\endcsname{0.02}
\expandafter\def\csname mcresult@cell/temporal_common_subset/temporal_matched_commontemporalXGB_emb_only/MSLP/mae/mean\endcsname{6.46}
\expandafter\def\csname mcresult@cell/temporal_common_subset/temporal_matched_commontemporalXGB_emb_only/MSLP/mae/sd\endcsname{0.18}
\expandafter\def\csname mcresult@cell/temporal_common_subset/temporal_matched_commontemporalXGB_emb_only/MSLP/r2/mean\endcsname{0.47}
\expandafter\def\csname mcresult@cell/temporal_common_subset/temporal_matched_commontemporalXGB_emb_only/MSLP/r2/sd\endcsname{0.02}
\expandafter\def\csname mcresult@cell/temporal_common_subset/temporal_matched_commontemporalXGB_emb_only/VMAX/mae/mean\endcsname{9.78}
\expandafter\def\csname mcresult@cell/temporal_common_subset/temporal_matched_commontemporalXGB_emb_only/VMAX/mae/sd\endcsname{0.26}
\expandafter\def\csname mcresult@cell/temporal_common_subset/temporal_matched_commontemporalXGB_emb_only/VMAX/r2/mean\endcsname{0.45}
\expandafter\def\csname mcresult@cell/temporal_common_subset/temporal_matched_commontemporalXGB_emb_only/VMAX/r2/sd\endcsname{0.02}
\expandafter\def\csname mcresult@cell/temporal_common_subset/temporal_matched_commontemporal_ablationCNN_Tab_Text_emb/MSLP/mae/mean\endcsname{6.46}
\expandafter\def\csname mcresult@cell/temporal_common_subset/temporal_matched_commontemporal_ablationCNN_Tab_Text_emb/MSLP/mae/sd\endcsname{0.18}
\expandafter\def\csname mcresult@cell/temporal_common_subset/temporal_matched_commontemporal_ablationCNN_Tab_Text_emb/MSLP/r2/mean\endcsname{0.47}
\expandafter\def\csname mcresult@cell/temporal_common_subset/temporal_matched_commontemporal_ablationCNN_Tab_Text_emb/MSLP/r2/sd\endcsname{0.02}
\expandafter\def\csname mcresult@cell/temporal_common_subset/temporal_matched_commontemporal_ablationCNN_Tab_Text_emb/VMAX/mae/mean\endcsname{9.78}
\expandafter\def\csname mcresult@cell/temporal_common_subset/temporal_matched_commontemporal_ablationCNN_Tab_Text_emb/VMAX/mae/sd\endcsname{0.26}
\expandafter\def\csname mcresult@cell/temporal_common_subset/temporal_matched_commontemporal_ablationCNN_Tab_Text_emb/VMAX/r2/mean\endcsname{0.45}
\expandafter\def\csname mcresult@cell/temporal_common_subset/temporal_matched_commontemporal_ablationCNN_Tab_Text_emb/VMAX/r2/sd\endcsname{0.02}
\expandafter\def\csname mcresult@cell/temporal_common_subset/temporal_matched_commontemporal_ablationCNN_Tab_emb/MSLP/mae/mean\endcsname{7.07}
\expandafter\def\csname mcresult@cell/temporal_common_subset/temporal_matched_commontemporal_ablationCNN_Tab_emb/MSLP/mae/sd\endcsname{0.11}
\expandafter\def\csname mcresult@cell/temporal_common_subset/temporal_matched_commontemporal_ablationCNN_Tab_emb/MSLP/r2/mean\endcsname{0.38}
\expandafter\def\csname mcresult@cell/temporal_common_subset/temporal_matched_commontemporal_ablationCNN_Tab_emb/MSLP/r2/sd\endcsname{0.03}
\expandafter\def\csname mcresult@cell/temporal_common_subset/temporal_matched_commontemporal_ablationCNN_Tab_emb/VMAX/mae/mean\endcsname{10.63}
\expandafter\def\csname mcresult@cell/temporal_common_subset/temporal_matched_commontemporal_ablationCNN_Tab_emb/VMAX/mae/sd\endcsname{0.11}
\expandafter\def\csname mcresult@cell/temporal_common_subset/temporal_matched_commontemporal_ablationCNN_Tab_emb/VMAX/r2/mean\endcsname{0.36}
\expandafter\def\csname mcresult@cell/temporal_common_subset/temporal_matched_commontemporal_ablationCNN_Tab_emb/VMAX/r2/sd\endcsname{0.01}
\expandafter\def\csname mcresult@comparison/ensemble/discussion/wind_mae_reduction_percent\endcsname{10.0}
\expandafter\def\csname mcresult@comparison/primary/discussion/wind_mae_reduction_percent\endcsname{6.6}
\expandafter\def\csname mcresult@dataset/atlantic_storms\endcsname{282}
\expandafter\def\csname mcresult@dataset/discussion_3h_percent\endcsname{98}
\expandafter\def\csname mcresult@dataset/discussion_delay_max_hours\endcsname{7}
\expandafter\def\csname mcresult@dataset/discussion_delay_min_hours\endcsname{3}
\expandafter\def\csname mcresult@dataset/fully_matched_observations\endcsname{6{,}303}
\expandafter\def\csname mcresult@dataset/fully_matched_storms\endcsname{308}
\expandafter\def\csname mcresult@dataset/matched\endcsname{12{,}704}
\expandafter\def\csname mcresult@dataset/matched_percent\endcsname{91.5}
\expandafter\def\csname mcresult@dataset/missing_twac\endcsname{74}
\expandafter\def\csname mcresult@dataset/observations\endcsname{13{,}885}
\expandafter\def\csname mcresult@dataset/pacific_storms\endcsname{355}
\expandafter\def\csname mcresult@dataset/partially_matched_observations\endcsname{7{,}582}
\expandafter\def\csname mcresult@dataset/partially_matched_reports\endcsname{6{,}401}
\expandafter\def\csname mcresult@dataset/partially_matched_storms\endcsname{329}
\expandafter\def\csname mcresult@dataset/storms\endcsname{637}
\expandafter\def\csname mcresult@dataset/unmatched\endcsname{1{,}181}
\expandafter\def\csname mcresult@dataset/unmatched_percent\endcsname{8.5}
\expandafter\def\csname mcresult@era5/n_observations\endcsname{11{,}021}
\expandafter\def\csname mcresult@era5/n_storms\endcsname{495}
\expandafter\def\csname mcresult@interval/AGATHA/T+D/coverage_percent\endcsname{73}
\expandafter\def\csname mcresult@interval/AGATHA/T+D/mae\endcsname{14.4}
\expandafter\def\csname mcresult@interval/AGATHA/T+D/median_width\endcsname{52.5}
\expandafter\def\csname mcresult@interval/AGATHA/T+I+D/coverage_percent\endcsname{91}
\expandafter\def\csname mcresult@interval/AGATHA/T+I+D/mae\endcsname{14.3}
\expandafter\def\csname mcresult@interval/AGATHA/T+I+D/median_width\endcsname{53.9}
\expandafter\def\csname mcresult@interval/AGATHA/T/coverage_percent\endcsname{82}
\expandafter\def\csname mcresult@interval/AGATHA/T/mae\endcsname{16.9}
\expandafter\def\csname mcresult@interval/AGATHA/T/median_width\endcsname{66.2}
\expandafter\def\csname mcresult@interval/CALVIN/T+D/coverage_percent\endcsname{96}
\expandafter\def\csname mcresult@interval/CALVIN/T+D/mae\endcsname{8.6}
\expandafter\def\csname mcresult@interval/CALVIN/T+D/median_width\endcsname{39.6}
\expandafter\def\csname mcresult@interval/CALVIN/T+I+D/coverage_percent\endcsname{100}
\expandafter\def\csname mcresult@interval/CALVIN/T+I+D/mae\endcsname{9.1}
\expandafter\def\csname mcresult@interval/CALVIN/T+I+D/median_width\endcsname{41.9}
\expandafter\def\csname mcresult@interval/CALVIN/T/coverage_percent\endcsname{89}
\expandafter\def\csname mcresult@interval/CALVIN/T/mae\endcsname{12.5}
\expandafter\def\csname mcresult@interval/CALVIN/T/median_width\endcsname{42.7}
\expandafter\def\csname mcresult@interval/DARBY/T+D/coverage_percent\endcsname{81}
\expandafter\def\csname mcresult@interval/DARBY/T+D/mae\endcsname{14.8}
\expandafter\def\csname mcresult@interval/DARBY/T+D/median_width\endcsname{47.8}
\expandafter\def\csname mcresult@interval/DARBY/T+I+D/coverage_percent\endcsname{74}
\expandafter\def\csname mcresult@interval/DARBY/T+I+D/mae\endcsname{15.7}
\expandafter\def\csname mcresult@interval/DARBY/T+I+D/median_width\endcsname{50.9}
\expandafter\def\csname mcresult@interval/DARBY/T/coverage_percent\endcsname{81}
\expandafter\def\csname mcresult@interval/DARBY/T/mae\endcsname{13.7}
\expandafter\def\csname mcresult@interval/DARBY/T/median_width\endcsname{42.2}
\expandafter\def\csname mcresult@interval/DORA/T+D/coverage_percent\endcsname{85}
\expandafter\def\csname mcresult@interval/DORA/T+D/mae\endcsname{13.8}
\expandafter\def\csname mcresult@interval/DORA/T+D/median_width\endcsname{50.8}
\expandafter\def\csname mcresult@interval/DORA/T+I+D/coverage_percent\endcsname{85}
\expandafter\def\csname mcresult@interval/DORA/T+I+D/mae\endcsname{14.0}
\expandafter\def\csname mcresult@interval/DORA/T+I+D/median_width\endcsname{51.8}
\expandafter\def\csname mcresult@interval/DORA/T/coverage_percent\endcsname{80}
\expandafter\def\csname mcresult@interval/DORA/T/mae\endcsname{15.3}
\expandafter\def\csname mcresult@interval/DORA/T/median_width\endcsname{52.6}
\expandafter\def\csname mcresult@interval/FIONA/T+D/coverage_percent\endcsname{97}
\expandafter\def\csname mcresult@interval/FIONA/T+D/mae\endcsname{6.9}
\expandafter\def\csname mcresult@interval/FIONA/T+D/median_width\endcsname{48.8}
\expandafter\def\csname mcresult@interval/FIONA/T+I+D/coverage_percent\endcsname{100}
\expandafter\def\csname mcresult@interval/FIONA/T+I+D/mae\endcsname{6.9}
\expandafter\def\csname mcresult@interval/FIONA/T+I+D/median_width\endcsname{50.1}
\expandafter\def\csname mcresult@interval/FIONA/T/coverage_percent\endcsname{89}
\expandafter\def\csname mcresult@interval/FIONA/T/mae\endcsname{10.7}
\expandafter\def\csname mcresult@interval/FIONA/T/median_width\endcsname{50.7}
\expandafter\def\csname mcresult@interval/HILARY/T+D/coverage_percent\endcsname{73}
\expandafter\def\csname mcresult@interval/HILARY/T+D/mae\endcsname{15.7}
\expandafter\def\csname mcresult@interval/HILARY/T+D/median_width\endcsname{66.2}
\expandafter\def\csname mcresult@interval/HILARY/T+I+D/coverage_percent\endcsname{73}
\expandafter\def\csname mcresult@interval/HILARY/T+I+D/mae\endcsname{15.5}
\expandafter\def\csname mcresult@interval/HILARY/T+I+D/median_width\endcsname{77.2}
\expandafter\def\csname mcresult@interval/HILARY/T/coverage_percent\endcsname{87}
\expandafter\def\csname mcresult@interval/HILARY/T/mae\endcsname{15.0}
\expandafter\def\csname mcresult@interval/HILARY/T/median_width\endcsname{57.5}
\expandafter\def\csname mcresult@interval/IAN/T+D/coverage_percent\endcsname{74}
\expandafter\def\csname mcresult@interval/IAN/T+D/mae\endcsname{15.8}
\expandafter\def\csname mcresult@interval/IAN/T+D/median_width\endcsname{40.6}
\expandafter\def\csname mcresult@interval/IAN/T+I+D/coverage_percent\endcsname{70}
\expandafter\def\csname mcresult@interval/IAN/T+I+D/mae\endcsname{15.9}
\expandafter\def\csname mcresult@interval/IAN/T+I+D/median_width\endcsname{43.8}
\expandafter\def\csname mcresult@interval/IAN/T/coverage_percent\endcsname{85}
\expandafter\def\csname mcresult@interval/IAN/T/mae\endcsname{19.1}
\expandafter\def\csname mcresult@interval/IAN/T/median_width\endcsname{52.7}
\expandafter\def\csname mcresult@interval/IDALIA/T+D/coverage_percent\endcsname{94}
\expandafter\def\csname mcresult@interval/IDALIA/T+D/mae\endcsname{11.5}
\expandafter\def\csname mcresult@interval/IDALIA/T+D/median_width\endcsname{43.2}
\expandafter\def\csname mcresult@interval/IDALIA/T+I+D/coverage_percent\endcsname{88}
\expandafter\def\csname mcresult@interval/IDALIA/T+I+D/mae\endcsname{11.5}
\expandafter\def\csname mcresult@interval/IDALIA/T+I+D/median_width\endcsname{41.2}
\expandafter\def\csname mcresult@interval/IDALIA/T/coverage_percent\endcsname{88}
\expandafter\def\csname mcresult@interval/IDALIA/T/mae\endcsname{14.2}
\expandafter\def\csname mcresult@interval/IDALIA/T/median_width\endcsname{40.5}
\expandafter\def\csname mcresult@interval/JULIA/T+D/coverage_percent\endcsname{100}
\expandafter\def\csname mcresult@interval/JULIA/T+D/mae\endcsname{7.3}
\expandafter\def\csname mcresult@interval/JULIA/T+D/median_width\endcsname{49.6}
\expandafter\def\csname mcresult@interval/JULIA/T+I+D/coverage_percent\endcsname{100}
\expandafter\def\csname mcresult@interval/JULIA/T+I+D/mae\endcsname{7.4}
\expandafter\def\csname mcresult@interval/JULIA/T+I+D/median_width\endcsname{43.9}
\expandafter\def\csname mcresult@interval/JULIA/T/coverage_percent\endcsname{64}
\expandafter\def\csname mcresult@interval/JULIA/T/mae\endcsname{20.9}
\expandafter\def\csname mcresult@interval/JULIA/T/median_width\endcsname{44.5}
\expandafter\def\csname mcresult@interval/KARL/T+D/coverage_percent\endcsname{100}
\expandafter\def\csname mcresult@interval/KARL/T+D/mae\endcsname{4.0}
\expandafter\def\csname mcresult@interval/KARL/T+D/median_width\endcsname{49.9}
\expandafter\def\csname mcresult@interval/KARL/T+I+D/coverage_percent\endcsname{100}
\expandafter\def\csname mcresult@interval/KARL/T+I+D/mae\endcsname{4.4}
\expandafter\def\csname mcresult@interval/KARL/T+I+D/median_width\endcsname{50.3}
\expandafter\def\csname mcresult@interval/KARL/T/coverage_percent\endcsname{100}
\expandafter\def\csname mcresult@interval/KARL/T/mae\endcsname{3.6}
\expandafter\def\csname mcresult@interval/KARL/T/median_width\endcsname{33.2}
\expandafter\def\csname mcresult@interval/LEE/T+D/coverage_percent\endcsname{82}
\expandafter\def\csname mcresult@interval/LEE/T+D/mae\endcsname{12.0}
\expandafter\def\csname mcresult@interval/LEE/T+D/median_width\endcsname{38.2}
\expandafter\def\csname mcresult@interval/LEE/T+I+D/coverage_percent\endcsname{82}
\expandafter\def\csname mcresult@interval/LEE/T+I+D/mae\endcsname{11.8}
\expandafter\def\csname mcresult@interval/LEE/T+I+D/median_width\endcsname{38.4}
\expandafter\def\csname mcresult@interval/LEE/T/coverage_percent\endcsname{95}
\expandafter\def\csname mcresult@interval/LEE/T/mae\endcsname{11.9}
\expandafter\def\csname mcresult@interval/LEE/T/median_width\endcsname{56.2}
\expandafter\def\csname mcresult@interval/MADELINE/T+D/coverage_percent\endcsname{82}
\expandafter\def\csname mcresult@interval/MADELINE/T+D/mae\endcsname{7.5}
\expandafter\def\csname mcresult@interval/MADELINE/T+D/median_width\endcsname{34.5}
\expandafter\def\csname mcresult@interval/MADELINE/T+I+D/coverage_percent\endcsname{82}
\expandafter\def\csname mcresult@interval/MADELINE/T+I+D/mae\endcsname{7.4}
\expandafter\def\csname mcresult@interval/MADELINE/T+I+D/median_width\endcsname{29.0}
\expandafter\def\csname mcresult@interval/MADELINE/T/coverage_percent\endcsname{100}
\expandafter\def\csname mcresult@interval/MADELINE/T/mae\endcsname{4.8}
\expandafter\def\csname mcresult@interval/MADELINE/T/median_width\endcsname{37.4}
\expandafter\def\csname mcresult@interval/OTIS/T+D/coverage_percent\endcsname{70}
\expandafter\def\csname mcresult@interval/OTIS/T+D/mae\endcsname{26.1}
\expandafter\def\csname mcresult@interval/OTIS/T+D/median_width\endcsname{40.2}
\expandafter\def\csname mcresult@interval/OTIS/T+I+D/coverage_percent\endcsname{70}
\expandafter\def\csname mcresult@interval/OTIS/T+I+D/mae\endcsname{26.7}
\expandafter\def\csname mcresult@interval/OTIS/T+I+D/median_width\endcsname{40.2}
\expandafter\def\csname mcresult@interval/OTIS/T/coverage_percent\endcsname{40}
\expandafter\def\csname mcresult@interval/OTIS/T/mae\endcsname{32.0}
\expandafter\def\csname mcresult@interval/OTIS/T/median_width\endcsname{35.8}
\expandafter\def\csname mcresult@interval/PHILIPPE/T+D/coverage_percent\endcsname{100}
\expandafter\def\csname mcresult@interval/PHILIPPE/T+D/mae\endcsname{3.6}
\expandafter\def\csname mcresult@interval/PHILIPPE/T+D/median_width\endcsname{34.7}
\expandafter\def\csname mcresult@interval/PHILIPPE/T+I+D/coverage_percent\endcsname{100}
\expandafter\def\csname mcresult@interval/PHILIPPE/T+I+D/mae\endcsname{3.6}
\expandafter\def\csname mcresult@interval/PHILIPPE/T+I+D/median_width\endcsname{33.3}
\expandafter\def\csname mcresult@interval/PHILIPPE/T/coverage_percent\endcsname{100}
\expandafter\def\csname mcresult@interval/PHILIPPE/T/mae\endcsname{5.5}
\expandafter\def\csname mcresult@interval/PHILIPPE/T/median_width\endcsname{37.0}
\expandafter\def\csname mcresult@map/D_IAN_010.202209251500_VMAX_24h/prediction_delta_abs\endcsname{+26.1}
\expandafter\def\csname mcresult@map/D_IAN_010.202209251500_VMAX_24h/true_delta_abs\endcsname{+30}
\expandafter\def\csname mcresult@map/D_IAN_010.202209251500_VMAX_48h/prediction_delta_abs\endcsname{+43.0}
\expandafter\def\csname mcresult@map/D_IAN_010.202209251500_VMAX_48h/true_delta_abs\endcsname{+60}
\expandafter\def\csname mcresult@map/D_LEE_009.202309071500_VMAX_72h/prediction_delta_abs\endcsname{+42.6}
\expandafter\def\csname mcresult@map/D_LEE_009.202309071500_VMAX_72h/true_delta_abs\endcsname{+10}
\expandafter\def\csname mcresult@map/TD_AGATHA_009.202205300300_VMAX_24h/prediction_delta_abs\endcsname{-0.5}
\expandafter\def\csname mcresult@map/TD_AGATHA_009.202205300300_VMAX_24h/true_delta_abs\endcsname{-25}
\expandafter\def\csname mcresult@map/TD_CALVIN_020.202307161500_VMAX_24h/prediction_delta_abs\endcsname{-21.5}
\expandafter\def\csname mcresult@map/TD_CALVIN_020.202307161500_VMAX_24h/true_delta_abs\endcsname{-20}
\expandafter\def\csname mcresult@map/TD_LEE_009.202309071500_VMAX_24h/prediction_delta_abs\endcsname{+15.4}
\expandafter\def\csname mcresult@map/TD_LEE_009.202309071500_VMAX_24h/true_delta_abs\endcsname{+50}
\expandafter\def\csname mcresult@map/TD_MADELINE_007.202209190900_VMAX_24h/prediction_delta_abs\endcsname{-7.1}
\expandafter\def\csname mcresult@map/TD_MADELINE_007.202209190900_VMAX_24h/true_delta_abs\endcsname{-15}
\expandafter\def\csname mcresult@map/TD_OTIS_008.202310240900_VMAX_24h/prediction_delta_abs\endcsname{+24.0}
\expandafter\def\csname mcresult@map/TD_OTIS_008.202310240900_VMAX_24h/true_delta_abs\endcsname{+90}
\expandafter\def\csname mcresult@operational/DSHP/n_obs\endcsname{1{,}122}
\expandafter\def\csname mcresult@operational/DSHP/n_storms\endcsname{66}
\expandafter\def\csname mcresult@operational/HF3A/n_obs\endcsname{158}
\expandafter\def\csname mcresult@operational/HF3A/n_storms\endcsname{13}
\expandafter\def\csname mcresult@operational/HFAI/n_obs\endcsname{641}
\expandafter\def\csname mcresult@operational/HFAI/n_storms\endcsname{35}
\expandafter\def\csname mcresult@operational/HFBI/n_obs\endcsname{641}
\expandafter\def\csname mcresult@operational/HFBI/n_storms\endcsname{35}
\expandafter\def\csname mcresult@operational/HFSA/n_obs\endcsname{646}
\expandafter\def\csname mcresult@operational/HFSA/n_storms\endcsname{35}
\expandafter\def\csname mcresult@operational/HFSB/n_obs\endcsname{646}
\expandafter\def\csname mcresult@operational/HFSB/n_storms\endcsname{35}
\expandafter\def\csname mcresult@operational/HWFI/n_obs\endcsname{1{,}066}
\expandafter\def\csname mcresult@operational/HWFI/n_storms\endcsname{64}
\expandafter\def\csname mcresult@operational/HWRF/n_obs\endcsname{1{,}104}
\expandafter\def\csname mcresult@operational/HWRF/n_storms\endcsname{64}
\expandafter\def\csname mcresult@operational/LGEM/n_obs\endcsname{1{,}122}
\expandafter\def\csname mcresult@operational/LGEM/n_storms\endcsname{66}
\expandafter\def\csname mcresult@operational/OFCI/n_obs\endcsname{1{,}018}
\expandafter\def\csname mcresult@operational/OFCI/n_storms\endcsname{64}
\expandafter\def\csname mcresult@operational/OFCL/n_obs\endcsname{1{,}078}
\expandafter\def\csname mcresult@operational/OFCL/n_storms\endcsname{64}
\expandafter\def\csname mcresult@operational_post/HFAI/TD_mae\endcsname{9.30}
\expandafter\def\csname mcresult@operational_post/HFAI/TD_minus_aid_mae\endcsname{-0.76}
\expandafter\def\csname mcresult@operational_post/HFAI/aid_mae\endcsname{10.07}
\expandafter\def\csname mcresult@operational_post/HFAI/excluded_before_common_start\endcsname{36}
\expandafter\def\csname mcresult@operational_post/HFAI/n_obs\endcsname{600}
\expandafter\def\csname mcresult@operational_post/HFAI/n_storms\endcsname{32}
\expandafter\def\csname mcresult@operational_post/HFAI/preimplementation_rows\endcsname{34}
\expandafter\def\csname mcresult@operational_post/HFAI/preimplementation_storms\endcsname{3}
\expandafter\def\csname mcresult@operational_post/HFAI/storm_ci_high\endcsname{0.42}
\expandafter\def\csname mcresult@operational_post/HFAI/storm_ci_low\endcsname{-2.02}
\expandafter\def\csname mcresult@operational_post/HFBI/TD_mae\endcsname{9.30}
\expandafter\def\csname mcresult@operational_post/HFBI/TD_minus_aid_mae\endcsname{-0.59}
\expandafter\def\csname mcresult@operational_post/HFBI/aid_mae\endcsname{9.89}
\expandafter\def\csname mcresult@operational_post/HFBI/excluded_before_common_start\endcsname{36}
\expandafter\def\csname mcresult@operational_post/HFBI/n_obs\endcsname{600}
\expandafter\def\csname mcresult@operational_post/HFBI/n_storms\endcsname{32}
\expandafter\def\csname mcresult@operational_post/HFBI/preimplementation_rows\endcsname{34}
\expandafter\def\csname mcresult@operational_post/HFBI/preimplementation_storms\endcsname{3}
\expandafter\def\csname mcresult@operational_post/HFBI/storm_ci_high\endcsname{0.32}
\expandafter\def\csname mcresult@operational_post/HFBI/storm_ci_low\endcsname{-1.51}
\expandafter\def\csname mcresult@operational_post/HFSA/TD_mae\endcsname{9.31}
\expandafter\def\csname mcresult@operational_post/HFSA/TD_minus_aid_mae\endcsname{-1.82}
\expandafter\def\csname mcresult@operational_post/HFSA/aid_mae\endcsname{11.12}
\expandafter\def\csname mcresult@operational_post/HFSA/excluded_before_common_start\endcsname{37}
\expandafter\def\csname mcresult@operational_post/HFSA/n_obs\endcsname{604}
\expandafter\def\csname mcresult@operational_post/HFSA/n_storms\endcsname{32}
\expandafter\def\csname mcresult@operational_post/HFSA/preimplementation_rows\endcsname{35}
\expandafter\def\csname mcresult@operational_post/HFSA/preimplementation_storms\endcsname{3}
\expandafter\def\csname mcresult@operational_post/HFSA/storm_ci_high\endcsname{-0.82}
\expandafter\def\csname mcresult@operational_post/HFSA/storm_ci_low\endcsname{-2.83}
\expandafter\def\csname mcresult@operational_post/HFSB/TD_mae\endcsname{9.31}
\expandafter\def\csname mcresult@operational_post/HFSB/TD_minus_aid_mae\endcsname{-2.04}
\expandafter\def\csname mcresult@operational_post/HFSB/aid_mae\endcsname{11.35}
\expandafter\def\csname mcresult@operational_post/HFSB/excluded_before_common_start\endcsname{37}
\expandafter\def\csname mcresult@operational_post/HFSB/n_obs\endcsname{604}
\expandafter\def\csname mcresult@operational_post/HFSB/n_storms\endcsname{32}
\expandafter\def\csname mcresult@operational_post/HFSB/preimplementation_rows\endcsname{35}
\expandafter\def\csname mcresult@operational_post/HFSB/preimplementation_storms\endcsname{3}
\expandafter\def\csname mcresult@operational_post/HFSB/storm_ci_high\endcsname{-1.22}
\expandafter\def\csname mcresult@operational_post/HFSB/storm_ci_low\endcsname{-2.81}
\expandafter\def\csname mcresult@primary/n_obs\endcsname{1{,}150}
\expandafter\def\csname mcresult@primary/n_storms\endcsname{66}
\expandafter\def\csname mcresult@primary/pressure_gain/left_mae\endcsname{6.43}
\expandafter\def\csname mcresult@primary/pressure_gain/mae_gain_left_to_right\endcsname{0.57}
\expandafter\def\csname mcresult@primary/pressure_gain/percent\endcsname{8.8}
\expandafter\def\csname mcresult@primary/pressure_gain/right_mae\endcsname{5.87}
\expandafter\def\csname mcresult@primary/pressure_gain/storm_ci_high\endcsname{1.05}
\expandafter\def\csname mcresult@primary/pressure_gain/storm_ci_low\endcsname{0.07}
\expandafter\def\csname mcresult@primary/wind_gain/left_mae\endcsname{9.89}
\expandafter\def\csname mcresult@primary/wind_gain/mae_gain_left_to_right\endcsname{0.99}
\expandafter\def\csname mcresult@primary/wind_gain/percent\endcsname{10.0}
\expandafter\def\csname mcresult@primary/wind_gain/right_mae\endcsname{8.90}
\expandafter\def\csname mcresult@primary/wind_gain/storm_ci_high\endcsname{1.70}
\expandafter\def\csname mcresult@primary/wind_gain/storm_ci_low\endcsname{0.32}
\expandafter\def\csname mcresult@probe/perturb/baseline/delta_mae\endcsname{+0.00}
\expandafter\def\csname mcresult@probe/perturb/baseline/delta_mae_sd\endcsname{0.00}
\expandafter\def\csname mcresult@probe/perturb/baseline/delta_r2\endcsname{+0.000}
\expandafter\def\csname mcresult@probe/perturb/baseline/delta_r2_sd\endcsname{0.000}
\expandafter\def\csname mcresult@probe/perturb/baseline/mae\endcsname{9.90}
\expandafter\def\csname mcresult@probe/perturb/baseline/mae_sd\endcsname{0.20}
\expandafter\def\csname mcresult@probe/perturb/baseline/r2\endcsname{0.500}
\expandafter\def\csname mcresult@probe/perturb/baseline/r2_sd\endcsname{0.015}
\expandafter\def\csname mcresult@probe/perturb/drop_intensity_vocab/delta_mae\endcsname{+2.97}
\expandafter\def\csname mcresult@probe/perturb/drop_intensity_vocab/delta_mae_sd\endcsname{0.25}
\expandafter\def\csname mcresult@probe/perturb/drop_intensity_vocab/delta_r2\endcsname{-0.364}
\expandafter\def\csname mcresult@probe/perturb/drop_intensity_vocab/delta_r2_sd\endcsname{0.022}
\expandafter\def\csname mcresult@probe/perturb/drop_intensity_vocab/mae\endcsname{12.87}
\expandafter\def\csname mcresult@probe/perturb/drop_intensity_vocab/mae_sd\endcsname{0.29}
\expandafter\def\csname mcresult@probe/perturb/drop_intensity_vocab/r2\endcsname{0.136}
\expandafter\def\csname mcresult@probe/perturb/drop_intensity_vocab/r2_sd\endcsname{0.035}
\expandafter\def\csname mcresult@probe/perturb/drop_outlook/delta_mae\endcsname{+2.24}
\expandafter\def\csname mcresult@probe/perturb/drop_outlook/delta_mae_sd\endcsname{0.20}
\expandafter\def\csname mcresult@probe/perturb/drop_outlook/delta_r2\endcsname{-0.251}
\expandafter\def\csname mcresult@probe/perturb/drop_outlook/delta_r2_sd\endcsname{0.019}
\expandafter\def\csname mcresult@probe/perturb/drop_outlook/mae\endcsname{12.14}
\expandafter\def\csname mcresult@probe/perturb/drop_outlook/mae_sd\endcsname{0.28}
\expandafter\def\csname mcresult@probe/perturb/drop_outlook/r2\endcsname{0.249}
\expandafter\def\csname mcresult@probe/perturb/drop_outlook/r2_sd\endcsname{0.030}
\expandafter\def\csname mcresult@probe/perturb/dropout_realizations\endcsname{5}
\expandafter\def\csname mcresult@probe/perturb/keep_only_outlook/delta_mae\endcsname{+0.73}
\expandafter\def\csname mcresult@probe/perturb/keep_only_outlook/delta_mae_sd\endcsname{0.34}
\expandafter\def\csname mcresult@probe/perturb/keep_only_outlook/delta_r2\endcsname{-0.091}
\expandafter\def\csname mcresult@probe/perturb/keep_only_outlook/delta_r2_sd\endcsname{0.039}
\expandafter\def\csname mcresult@probe/perturb/keep_only_outlook/mae\endcsname{10.63}
\expandafter\def\csname mcresult@probe/perturb/keep_only_outlook/mae_sd\endcsname{0.47}
\expandafter\def\csname mcresult@probe/perturb/keep_only_outlook/r2\endcsname{0.408}
\expandafter\def\csname mcresult@probe/perturb/keep_only_outlook/r2_sd\endcsname{0.050}
\expandafter\def\csname mcresult@probe/perturb/mask_numbers+drop_outlook+mask_storm_name/delta_mae\endcsname{+2.25}
\expandafter\def\csname mcresult@probe/perturb/mask_numbers+drop_outlook+mask_storm_name/delta_mae_sd\endcsname{0.22}
\expandafter\def\csname mcresult@probe/perturb/mask_numbers+drop_outlook+mask_storm_name/delta_r2\endcsname{-0.255}
\expandafter\def\csname mcresult@probe/perturb/mask_numbers+drop_outlook+mask_storm_name/delta_r2_sd\endcsname{0.025}
\expandafter\def\csname mcresult@probe/perturb/mask_numbers+drop_outlook+mask_storm_name/mae\endcsname{12.14}
\expandafter\def\csname mcresult@probe/perturb/mask_numbers+drop_outlook+mask_storm_name/mae_sd\endcsname{0.32}
\expandafter\def\csname mcresult@probe/perturb/mask_numbers+drop_outlook+mask_storm_name/r2\endcsname{0.245}
\expandafter\def\csname mcresult@probe/perturb/mask_numbers+drop_outlook+mask_storm_name/r2_sd\endcsname{0.035}
\expandafter\def\csname mcresult@probe/perturb/mask_numbers/delta_mae\endcsname{+0.20}
\expandafter\def\csname mcresult@probe/perturb/mask_numbers/delta_mae_sd\endcsname{0.20}
\expandafter\def\csname mcresult@probe/perturb/mask_numbers/delta_r2\endcsname{-0.014}
\expandafter\def\csname mcresult@probe/perturb/mask_numbers/delta_r2_sd\endcsname{0.009}
\expandafter\def\csname mcresult@probe/perturb/mask_numbers/mae\endcsname{10.10}
\expandafter\def\csname mcresult@probe/perturb/mask_numbers/mae_sd\endcsname{0.32}
\expandafter\def\csname mcresult@probe/perturb/mask_numbers/r2\endcsname{0.486}
\expandafter\def\csname mcresult@probe/perturb/mask_numbers/r2_sd\endcsname{0.020}
\expandafter\def\csname mcresult@probe/perturb/mask_storm_name/delta_mae\endcsname{-0.12}
\expandafter\def\csname mcresult@probe/perturb/mask_storm_name/delta_mae_sd\endcsname{0.17}
\expandafter\def\csname mcresult@probe/perturb/mask_storm_name/delta_r2\endcsname{+0.009}
\expandafter\def\csname mcresult@probe/perturb/mask_storm_name/delta_r2_sd\endcsname{0.016}
\expandafter\def\csname mcresult@probe/perturb/mask_storm_name/mae\endcsname{9.77}
\expandafter\def\csname mcresult@probe/perturb/mask_storm_name/mae_sd\endcsname{0.20}
\expandafter\def\csname mcresult@probe/perturb/mask_storm_name/r2\endcsname{0.509}
\expandafter\def\csname mcresult@probe/perturb/mask_storm_name/r2_sd\endcsname{0.017}
\expandafter\def\csname mcresult@probe/perturb/random_dropout_50/delta_mae\endcsname{+1.61}
\expandafter\def\csname mcresult@probe/perturb/random_dropout_50/delta_mae_sd\endcsname{0.25}
\expandafter\def\csname mcresult@probe/perturb/random_dropout_50/delta_r2\endcsname{-0.195}
\expandafter\def\csname mcresult@probe/perturb/random_dropout_50/delta_r2_sd\endcsname{0.028}
\expandafter\def\csname mcresult@probe/perturb/random_dropout_50/mae\endcsname{11.51}
\expandafter\def\csname mcresult@probe/perturb/random_dropout_50/mae_sd\endcsname{0.36}
\expandafter\def\csname mcresult@probe/perturb/random_dropout_50/r2\endcsname{0.305}
\expandafter\def\csname mcresult@probe/perturb/random_dropout_50/r2_sd\endcsname{0.039}
\expandafter\def\csname mcresult@probe/perturb/training_seeds\endcsname{5}
\expandafter\def\csname mcresult@probe/tfidf/MSLP/alpha\endcsname{1.0}
\expandafter\def\csname mcresult@probe/tfidf/MSLP/mae\endcsname{7.8}
\expandafter\def\csname mcresult@probe/tfidf/MSLP/neural_mean_r2\endcsname{0.46}
\expandafter\def\csname mcresult@probe/tfidf/MSLP/r2\endcsname{0.35}
\expandafter\def\csname mcresult@probe/tfidf/VMAX/alpha\endcsname{1.0}
\expandafter\def\csname mcresult@probe/tfidf/VMAX/mae\endcsname{11.1}
\expandafter\def\csname mcresult@probe/tfidf/VMAX/neural_mean_r2\endcsname{0.50}
\expandafter\def\csname mcresult@probe/tfidf/VMAX/r2\endcsname{0.41}
\expandafter\def\csname mcresult@reproduction/cpu/entries\endcsname{351}
\expandafter\def\csname mcresult@reproduction/cpu/replayed\endcsname{246}
\expandafter\def\csname mcresult@reproduction/cpu/rounded_metric_changes\endcsname{16}
\expandafter\def\csname mcresult@reproduction/cpu/strict_fail\endcsname{165}
\expandafter\def\csname mcresult@reproduction/cpu/strict_pass\endcsname{81}
\expandafter\def\csname mcresult@ri/aci/coverage_mean\endcsname{0.910}
\expandafter\def\csname mcresult@ri/aci/coverage_mean_forest_sd\endcsname{0.000}
\expandafter\def\csname mcresult@ri/aci/interval_score\endcsname{58.62}
\expandafter\def\csname mcresult@ri/aci/interval_score_forest_sd\endcsname{0.00}
\expandafter\def\csname mcresult@ri/aci/ri_coverage\endcsname{0.507}
\expandafter\def\csname mcresult@ri/aci/ri_coverage_forest_sd\endcsname{0.000}
\expandafter\def\csname mcresult@ri/aci/ri_interval_score\endcsname{172.33}
\expandafter\def\csname mcresult@ri/aci/ri_interval_score_forest_sd\endcsname{0.00}
\expandafter\def\csname mcresult@ri/appearances\endcsname{3{,}651}
\expandafter\def\csname mcresult@ri/block/coverage_mean\endcsname{0.920}
\expandafter\def\csname mcresult@ri/block/coverage_mean_forest_sd\endcsname{0.000}
\expandafter\def\csname mcresult@ri/block/interval_score\endcsname{57.51}
\expandafter\def\csname mcresult@ri/block/interval_score_forest_sd\endcsname{0.00}
\expandafter\def\csname mcresult@ri/block/ri_coverage\endcsname{0.554}
\expandafter\def\csname mcresult@ri/block/ri_coverage_forest_sd\endcsname{0.000}
\expandafter\def\csname mcresult@ri/block/ri_interval_score\endcsname{166.59}
\expandafter\def\csname mcresult@ri/block/ri_interval_score_forest_sd\endcsname{0.00}
\expandafter\def\csname mcresult@ri/block_norm/coverage_mean\endcsname{0.917}
\expandafter\def\csname mcresult@ri/block_norm/coverage_mean_forest_sd\endcsname{0.000}
\expandafter\def\csname mcresult@ri/block_norm/interval_score\endcsname{57.34}
\expandafter\def\csname mcresult@ri/block_norm/interval_score_forest_sd\endcsname{0.00}
\expandafter\def\csname mcresult@ri/block_norm/ri_coverage\endcsname{0.560}
\expandafter\def\csname mcresult@ri/block_norm/ri_coverage_forest_sd\endcsname{0.000}
\expandafter\def\csname mcresult@ri/block_norm/ri_interval_score\endcsname{154.27}
\expandafter\def\csname mcresult@ri/block_norm/ri_interval_score_forest_sd\endcsname{0.00}
\expandafter\def\csname mcresult@ri/clear_naive/coverage_mean\endcsname{0.895}
\expandafter\def\csname mcresult@ri/clear_naive/coverage_mean_forest_sd\endcsname{0.001}
\expandafter\def\csname mcresult@ri/clear_naive/interval_score\endcsname{54.40}
\expandafter\def\csname mcresult@ri/clear_naive/interval_score_forest_sd\endcsname{0.06}
\expandafter\def\csname mcresult@ri/clear_naive/ri_coverage\endcsname{0.585}
\expandafter\def\csname mcresult@ri/clear_naive/ri_coverage_forest_sd\endcsname{0.002}
\expandafter\def\csname mcresult@ri/clear_naive/ri_interval_score\endcsname{152.55}
\expandafter\def\csname mcresult@ri/clear_naive/ri_interval_score_forest_sd\endcsname{0.33}
\expandafter\def\csname mcresult@ri/clear_storm_block/coverage_mean\endcsname{0.913}
\expandafter\def\csname mcresult@ri/clear_storm_block/coverage_mean_forest_sd\endcsname{0.001}
\expandafter\def\csname mcresult@ri/clear_storm_block/interval_score\endcsname{54.83}
\expandafter\def\csname mcresult@ri/clear_storm_block/interval_score_forest_sd\endcsname{0.09}
\expandafter\def\csname mcresult@ri/clear_storm_block/ri_coverage\endcsname{0.618}
\expandafter\def\csname mcresult@ri/clear_storm_block/ri_coverage_forest_sd\endcsname{0.002}
\expandafter\def\csname mcresult@ri/clear_storm_block/ri_interval_score\endcsname{143.39}
\expandafter\def\csname mcresult@ri/clear_storm_block/ri_interval_score_forest_sd\endcsname{0.44}
\expandafter\def\csname mcresult@ri/clear_storm_block_cf/coverage_mean\endcsname{0.923}
\expandafter\def\csname mcresult@ri/clear_storm_block_cf/coverage_mean_forest_sd\endcsname{0.000}
\expandafter\def\csname mcresult@ri/clear_storm_block_cf/interval_score\endcsname{53.49}
\expandafter\def\csname mcresult@ri/clear_storm_block_cf/interval_score_forest_sd\endcsname{0.03}
\expandafter\def\csname mcresult@ri/clear_storm_block_cf/ri_coverage\endcsname{0.632}
\expandafter\def\csname mcresult@ri/clear_storm_block_cf/ri_coverage_forest_sd\endcsname{0.002}
\expandafter\def\csname mcresult@ri/clear_storm_block_cf/ri_interval_score\endcsname{137.78}
\expandafter\def\csname mcresult@ri/clear_storm_block_cf/ri_interval_score_forest_sd\endcsname{0.48}
\expandafter\def\csname mcresult@ri/enbpi_proxy/coverage_mean\endcsname{0.891}
\expandafter\def\csname mcresult@ri/enbpi_proxy/coverage_mean_forest_sd\endcsname{0.000}
\expandafter\def\csname mcresult@ri/enbpi_proxy/interval_score\endcsname{58.28}
\expandafter\def\csname mcresult@ri/enbpi_proxy/interval_score_forest_sd\endcsname{0.00}
\expandafter\def\csname mcresult@ri/enbpi_proxy/ri_coverage\endcsname{0.498}
\expandafter\def\csname mcresult@ri/enbpi_proxy/ri_coverage_forest_sd\endcsname{0.000}
\expandafter\def\csname mcresult@ri/enbpi_proxy/ri_interval_score\endcsname{182.28}
\expandafter\def\csname mcresult@ri/enbpi_proxy/ri_interval_score_forest_sd\endcsname{0.00}
\expandafter\def\csname mcresult@ri/naive_iid/coverage_mean\endcsname{0.905}
\expandafter\def\csname mcresult@ri/naive_iid/coverage_mean_forest_sd\endcsname{0.000}
\expandafter\def\csname mcresult@ri/naive_iid/interval_score\endcsname{57.01}
\expandafter\def\csname mcresult@ri/naive_iid/interval_score_forest_sd\endcsname{0.00}
\expandafter\def\csname mcresult@ri/naive_iid/ri_coverage\endcsname{0.517}
\expandafter\def\csname mcresult@ri/naive_iid/ri_coverage_forest_sd\endcsname{0.000}
\expandafter\def\csname mcresult@ri/naive_iid/ri_interval_score\endcsname{175.96}
\expandafter\def\csname mcresult@ri/naive_iid/ri_interval_score_forest_sd\endcsname{0.00}
\expandafter\def\csname mcresult@ri/unique\endcsname{97}
\expandafter\def\csname mcresult@seed/abl/disc/MSLP/24/dir_acc_mean\endcsname{0.738}
\expandafter\def\csname mcresult@seed/abl/disc/MSLP/24/dir_acc_sd\endcsname{0.017}
\expandafter\def\csname mcresult@seed/abl/disc/MSLP/24/mae_mean\endcsname{6.56}
\expandafter\def\csname mcresult@seed/abl/disc/MSLP/24/mae_sd\endcsname{0.17}
\expandafter\def\csname mcresult@seed/abl/disc/MSLP/24/r2_mean\endcsname{0.46}
\expandafter\def\csname mcresult@seed/abl/disc/MSLP/24/r2_sd\endcsname{0.03}
\expandafter\def\csname mcresult@seed/abl/disc/MSLP/24/rmse_mean\endcsname{9.73}
\expandafter\def\csname mcresult@seed/abl/disc/MSLP/24/rmse_sd\endcsname{0.23}
\expandafter\def\csname mcresult@seed/abl/disc/MSLP/24/skill_score_mean\endcsname{0.27}
\expandafter\def\csname mcresult@seed/abl/disc/MSLP/24/skill_score_sd\endcsname{0.02}
\expandafter\def\csname mcresult@seed/abl/disc/VMAX/24/dir_acc_mean\endcsname{0.701}
\expandafter\def\csname mcresult@seed/abl/disc/VMAX/24/dir_acc_sd\endcsname{0.014}
\expandafter\def\csname mcresult@seed/abl/disc/VMAX/24/mae_mean\endcsname{9.90}
\expandafter\def\csname mcresult@seed/abl/disc/VMAX/24/mae_sd\endcsname{0.20}
\expandafter\def\csname mcresult@seed/abl/disc/VMAX/24/r2_mean\endcsname{0.50}
\expandafter\def\csname mcresult@seed/abl/disc/VMAX/24/r2_sd\endcsname{0.01}
\expandafter\def\csname mcresult@seed/abl/disc/VMAX/24/rmse_mean\endcsname{13.63}
\expandafter\def\csname mcresult@seed/abl/disc/VMAX/24/rmse_sd\endcsname{0.20}
\expandafter\def\csname mcresult@seed/abl/disc/VMAX/24/skill_score_mean\endcsname{0.29}
\expandafter\def\csname mcresult@seed/abl/disc/VMAX/24/skill_score_sd\endcsname{0.01}
\expandafter\def\csname mcresult@seed/abl/full/MSLP/24/dir_acc_mean\endcsname{0.752}
\expandafter\def\csname mcresult@seed/abl/full/MSLP/24/dir_acc_sd\endcsname{0.010}
\expandafter\def\csname mcresult@seed/abl/full/MSLP/24/mae_mean\endcsname{6.23}
\expandafter\def\csname mcresult@seed/abl/full/MSLP/24/mae_sd\endcsname{0.08}
\expandafter\def\csname mcresult@seed/abl/full/MSLP/24/r2_mean\endcsname{0.51}
\expandafter\def\csname mcresult@seed/abl/full/MSLP/24/r2_sd\endcsname{0.01}
\expandafter\def\csname mcresult@seed/abl/full/MSLP/24/rmse_mean\endcsname{9.12}
\expandafter\def\csname mcresult@seed/abl/full/MSLP/24/rmse_sd\endcsname{0.07}
\expandafter\def\csname mcresult@seed/abl/full/MSLP/24/skill_score_mean\endcsname{0.30}
\expandafter\def\csname mcresult@seed/abl/full/MSLP/24/skill_score_sd\endcsname{0.01}
\expandafter\def\csname mcresult@seed/abl/full/VMAX/24/dir_acc_mean\endcsname{0.724}
\expandafter\def\csname mcresult@seed/abl/full/VMAX/24/dir_acc_sd\endcsname{0.006}
\expandafter\def\csname mcresult@seed/abl/full/VMAX/24/mae_mean\endcsname{9.45}
\expandafter\def\csname mcresult@seed/abl/full/VMAX/24/mae_sd\endcsname{0.13}
\expandafter\def\csname mcresult@seed/abl/full/VMAX/24/r2_mean\endcsname{0.53}
\expandafter\def\csname mcresult@seed/abl/full/VMAX/24/r2_sd\endcsname{0.01}
\expandafter\def\csname mcresult@seed/abl/full/VMAX/24/rmse_mean\endcsname{13.07}
\expandafter\def\csname mcresult@seed/abl/full/VMAX/24/rmse_sd\endcsname{0.14}
\expandafter\def\csname mcresult@seed/abl/full/VMAX/24/skill_score_mean\endcsname{0.31}
\expandafter\def\csname mcresult@seed/abl/full/VMAX/24/skill_score_sd\endcsname{0.01}
\expandafter\def\csname mcresult@seed/abl/img+disc/MSLP/24/dir_acc_mean\endcsname{0.736}
\expandafter\def\csname mcresult@seed/abl/img+disc/MSLP/24/dir_acc_sd\endcsname{0.019}
\expandafter\def\csname mcresult@seed/abl/img+disc/MSLP/24/mae_mean\endcsname{6.45}
\expandafter\def\csname mcresult@seed/abl/img+disc/MSLP/24/mae_sd\endcsname{0.18}
\expandafter\def\csname mcresult@seed/abl/img+disc/MSLP/24/r2_mean\endcsname{0.48}
\expandafter\def\csname mcresult@seed/abl/img+disc/MSLP/24/r2_sd\endcsname{0.02}
\expandafter\def\csname mcresult@seed/abl/img+disc/MSLP/24/rmse_mean\endcsname{9.43}
\expandafter\def\csname mcresult@seed/abl/img+disc/MSLP/24/rmse_sd\endcsname{0.18}
\expandafter\def\csname mcresult@seed/abl/img+disc/MSLP/24/skill_score_mean\endcsname{0.28}
\expandafter\def\csname mcresult@seed/abl/img+disc/MSLP/24/skill_score_sd\endcsname{0.01}
\expandafter\def\csname mcresult@seed/abl/img+disc/VMAX/24/dir_acc_mean\endcsname{0.709}
\expandafter\def\csname mcresult@seed/abl/img+disc/VMAX/24/dir_acc_sd\endcsname{0.006}
\expandafter\def\csname mcresult@seed/abl/img+disc/VMAX/24/mae_mean\endcsname{9.74}
\expandafter\def\csname mcresult@seed/abl/img+disc/VMAX/24/mae_sd\endcsname{0.27}
\expandafter\def\csname mcresult@seed/abl/img+disc/VMAX/24/r2_mean\endcsname{0.50}
\expandafter\def\csname mcresult@seed/abl/img+disc/VMAX/24/r2_sd\endcsname{0.01}
\expandafter\def\csname mcresult@seed/abl/img+disc/VMAX/24/rmse_mean\endcsname{13.41}
\expandafter\def\csname mcresult@seed/abl/img+disc/VMAX/24/rmse_sd\endcsname{0.19}
\expandafter\def\csname mcresult@seed/abl/img+disc/VMAX/24/skill_score_mean\endcsname{0.30}
\expandafter\def\csname mcresult@seed/abl/img+disc/VMAX/24/skill_score_sd\endcsname{0.01}
\expandafter\def\csname mcresult@seed/abl/img/MSLP/24/dir_acc_mean\endcsname{0.665}
\expandafter\def\csname mcresult@seed/abl/img/MSLP/24/dir_acc_sd\endcsname{0.009}
\expandafter\def\csname mcresult@seed/abl/img/MSLP/24/mae_mean\endcsname{8.15}
\expandafter\def\csname mcresult@seed/abl/img/MSLP/24/mae_sd\endcsname{0.13}
\expandafter\def\csname mcresult@seed/abl/img/MSLP/24/r2_mean\endcsname{0.20}
\expandafter\def\csname mcresult@seed/abl/img/MSLP/24/r2_sd\endcsname{0.02}
\expandafter\def\csname mcresult@seed/abl/img/MSLP/24/rmse_mean\endcsname{11.65}
\expandafter\def\csname mcresult@seed/abl/img/MSLP/24/rmse_sd\endcsname{0.13}
\expandafter\def\csname mcresult@seed/abl/img/MSLP/24/skill_score_mean\endcsname{0.11}
\expandafter\def\csname mcresult@seed/abl/img/MSLP/24/skill_score_sd\endcsname{0.01}
\expandafter\def\csname mcresult@seed/abl/img/VMAX/24/dir_acc_mean\endcsname{0.616}
\expandafter\def\csname mcresult@seed/abl/img/VMAX/24/dir_acc_sd\endcsname{0.007}
\expandafter\def\csname mcresult@seed/abl/img/VMAX/24/mae_mean\endcsname{12.40}
\expandafter\def\csname mcresult@seed/abl/img/VMAX/24/mae_sd\endcsname{0.16}
\expandafter\def\csname mcresult@seed/abl/img/VMAX/24/r2_mean\endcsname{0.22}
\expandafter\def\csname mcresult@seed/abl/img/VMAX/24/r2_sd\endcsname{0.02}
\expandafter\def\csname mcresult@seed/abl/img/VMAX/24/rmse_mean\endcsname{16.77}
\expandafter\def\csname mcresult@seed/abl/img/VMAX/24/rmse_sd\endcsname{0.19}
\expandafter\def\csname mcresult@seed/abl/img/VMAX/24/skill_score_mean\endcsname{0.12}
\expandafter\def\csname mcresult@seed/abl/img/VMAX/24/skill_score_sd\endcsname{0.01}
\expandafter\def\csname mcresult@seed/abl/tab+disc/MSLP/24/dir_acc_mean\endcsname{0.752}
\expandafter\def\csname mcresult@seed/abl/tab+disc/MSLP/24/dir_acc_sd\endcsname{0.013}
\expandafter\def\csname mcresult@seed/abl/tab+disc/MSLP/24/mae_mean\endcsname{6.23}
\expandafter\def\csname mcresult@seed/abl/tab+disc/MSLP/24/mae_sd\endcsname{0.08}
\expandafter\def\csname mcresult@seed/abl/tab+disc/MSLP/24/r2_mean\endcsname{0.51}
\expandafter\def\csname mcresult@seed/abl/tab+disc/MSLP/24/r2_sd\endcsname{0.01}
\expandafter\def\csname mcresult@seed/abl/tab+disc/MSLP/24/rmse_mean\endcsname{9.14}
\expandafter\def\csname mcresult@seed/abl/tab+disc/MSLP/24/rmse_sd\endcsname{0.11}
\expandafter\def\csname mcresult@seed/abl/tab+disc/MSLP/24/skill_score_mean\endcsname{0.30}
\expandafter\def\csname mcresult@seed/abl/tab+disc/MSLP/24/skill_score_sd\endcsname{0.01}
\expandafter\def\csname mcresult@seed/abl/tab+disc/VMAX/24/dir_acc_mean\endcsname{0.723}
\expandafter\def\csname mcresult@seed/abl/tab+disc/VMAX/24/dir_acc_sd\endcsname{0.009}
\expandafter\def\csname mcresult@seed/abl/tab+disc/VMAX/24/mae_mean\endcsname{9.46}
\expandafter\def\csname mcresult@seed/abl/tab+disc/VMAX/24/mae_sd\endcsname{0.16}
\expandafter\def\csname mcresult@seed/abl/tab+disc/VMAX/24/r2_mean\endcsname{0.53}
\expandafter\def\csname mcresult@seed/abl/tab+disc/VMAX/24/r2_sd\endcsname{0.01}
\expandafter\def\csname mcresult@seed/abl/tab+disc/VMAX/24/rmse_mean\endcsname{13.04}
\expandafter\def\csname mcresult@seed/abl/tab+disc/VMAX/24/rmse_sd\endcsname{0.16}
\expandafter\def\csname mcresult@seed/abl/tab+disc/VMAX/24/skill_score_mean\endcsname{0.31}
\expandafter\def\csname mcresult@seed/abl/tab+disc/VMAX/24/skill_score_sd\endcsname{0.01}
\expandafter\def\csname mcresult@seed/abl/tab+img/MSLP/24/dir_acc_mean\endcsname{0.724}
\expandafter\def\csname mcresult@seed/abl/tab+img/MSLP/24/dir_acc_sd\endcsname{0.005}
\expandafter\def\csname mcresult@seed/abl/tab+img/MSLP/24/mae_mean\endcsname{6.72}
\expandafter\def\csname mcresult@seed/abl/tab+img/MSLP/24/mae_sd\endcsname{0.05}
\expandafter\def\csname mcresult@seed/abl/tab+img/MSLP/24/r2_mean\endcsname{0.45}
\expandafter\def\csname mcresult@seed/abl/tab+img/MSLP/24/r2_sd\endcsname{0.01}
\expandafter\def\csname mcresult@seed/abl/tab+img/MSLP/24/rmse_mean\endcsname{9.71}
\expandafter\def\csname mcresult@seed/abl/tab+img/MSLP/24/rmse_sd\endcsname{0.08}
\expandafter\def\csname mcresult@seed/abl/tab+img/MSLP/24/skill_score_mean\endcsname{0.26}
\expandafter\def\csname mcresult@seed/abl/tab+img/MSLP/24/skill_score_sd\endcsname{0.01}
\expandafter\def\csname mcresult@seed/abl/tab+img/VMAX/24/dir_acc_mean\endcsname{0.685}
\expandafter\def\csname mcresult@seed/abl/tab+img/VMAX/24/dir_acc_sd\endcsname{0.009}
\expandafter\def\csname mcresult@seed/abl/tab+img/VMAX/24/mae_mean\endcsname{10.27}
\expandafter\def\csname mcresult@seed/abl/tab+img/VMAX/24/mae_sd\endcsname{0.06}
\expandafter\def\csname mcresult@seed/abl/tab+img/VMAX/24/r2_mean\endcsname{0.45}
\expandafter\def\csname mcresult@seed/abl/tab+img/VMAX/24/r2_sd\endcsname{0.01}
\expandafter\def\csname mcresult@seed/abl/tab+img/VMAX/24/rmse_mean\endcsname{14.11}
\expandafter\def\csname mcresult@seed/abl/tab+img/VMAX/24/rmse_sd\endcsname{0.11}
\expandafter\def\csname mcresult@seed/abl/tab+img/VMAX/24/skill_score_mean\endcsname{0.26}
\expandafter\def\csname mcresult@seed/abl/tab+img/VMAX/24/skill_score_sd\endcsname{0.01}
\expandafter\def\csname mcresult@seed/abl/tab/MSLP/24/dir_acc_mean\endcsname{0.726}
\expandafter\def\csname mcresult@seed/abl/tab/MSLP/24/dir_acc_sd\endcsname{0.007}
\expandafter\def\csname mcresult@seed/abl/tab/MSLP/24/mae_mean\endcsname{6.60}
\expandafter\def\csname mcresult@seed/abl/tab/MSLP/24/mae_sd\endcsname{0.08}
\expandafter\def\csname mcresult@seed/abl/tab/MSLP/24/r2_mean\endcsname{0.44}
\expandafter\def\csname mcresult@seed/abl/tab/MSLP/24/r2_sd\endcsname{0.01}
\expandafter\def\csname mcresult@seed/abl/tab/MSLP/24/rmse_mean\endcsname{9.76}
\expandafter\def\csname mcresult@seed/abl/tab/MSLP/24/rmse_sd\endcsname{0.07}
\expandafter\def\csname mcresult@seed/abl/tab/MSLP/24/skill_score_mean\endcsname{0.25}
\expandafter\def\csname mcresult@seed/abl/tab/MSLP/24/skill_score_sd\endcsname{0.01}
\expandafter\def\csname mcresult@seed/abl/tab/VMAX/24/dir_acc_mean\endcsname{0.684}
\expandafter\def\csname mcresult@seed/abl/tab/VMAX/24/dir_acc_sd\endcsname{0.006}
\expandafter\def\csname mcresult@seed/abl/tab/VMAX/24/mae_mean\endcsname{10.13}
\expandafter\def\csname mcresult@seed/abl/tab/VMAX/24/mae_sd\endcsname{0.05}
\expandafter\def\csname mcresult@seed/abl/tab/VMAX/24/r2_mean\endcsname{0.44}
\expandafter\def\csname mcresult@seed/abl/tab/VMAX/24/r2_sd\endcsname{0.01}
\expandafter\def\csname mcresult@seed/abl/tab/VMAX/24/rmse_mean\endcsname{14.27}
\expandafter\def\csname mcresult@seed/abl/tab/VMAX/24/rmse_sd\endcsname{0.13}
\expandafter\def\csname mcresult@seed/abl/tab/VMAX/24/skill_score_mean\endcsname{0.25}
\expandafter\def\csname mcresult@seed/abl/tab/VMAX/24/skill_score_sd\endcsname{0.01}
\expandafter\def\csname mcresult@seed/fph/disc/MSLP/120/dir_acc_mean\endcsname{0.740}
\expandafter\def\csname mcresult@seed/fph/disc/MSLP/120/dir_acc_sd\endcsname{0.013}
\expandafter\def\csname mcresult@seed/fph/disc/MSLP/120/mae_mean\endcsname{17.19}
\expandafter\def\csname mcresult@seed/fph/disc/MSLP/120/mae_sd\endcsname{1.44}
\expandafter\def\csname mcresult@seed/fph/disc/MSLP/120/r2_mean\endcsname{0.00}
\expandafter\def\csname mcresult@seed/fph/disc/MSLP/120/r2_sd\endcsname{0.15}
\expandafter\def\csname mcresult@seed/fph/disc/MSLP/120/rmse_mean\endcsname{22.00}
\expandafter\def\csname mcresult@seed/fph/disc/MSLP/120/rmse_sd\endcsname{1.56}
\expandafter\def\csname mcresult@seed/fph/disc/MSLP/120/skill_score_mean\endcsname{0.00}
\expandafter\def\csname mcresult@seed/fph/disc/MSLP/120/skill_score_sd\endcsname{0.07}
\expandafter\def\csname mcresult@seed/fph/disc/MSLP/24/dir_acc_mean\endcsname{0.749}
\expandafter\def\csname mcresult@seed/fph/disc/MSLP/24/dir_acc_sd\endcsname{0.004}
\expandafter\def\csname mcresult@seed/fph/disc/MSLP/24/mae_mean\endcsname{6.35}
\expandafter\def\csname mcresult@seed/fph/disc/MSLP/24/mae_sd\endcsname{0.17}
\expandafter\def\csname mcresult@seed/fph/disc/MSLP/24/r2_mean\endcsname{0.48}
\expandafter\def\csname mcresult@seed/fph/disc/MSLP/24/r2_sd\endcsname{0.02}
\expandafter\def\csname mcresult@seed/fph/disc/MSLP/24/rmse_mean\endcsname{9.51}
\expandafter\def\csname mcresult@seed/fph/disc/MSLP/24/rmse_sd\endcsname{0.16}
\expandafter\def\csname mcresult@seed/fph/disc/MSLP/24/skill_score_mean\endcsname{0.28}
\expandafter\def\csname mcresult@seed/fph/disc/MSLP/24/skill_score_sd\endcsname{0.01}
\expandafter\def\csname mcresult@seed/fph/disc/MSLP/48/dir_acc_mean\endcsname{0.752}
\expandafter\def\csname mcresult@seed/fph/disc/MSLP/48/dir_acc_sd\endcsname{0.012}
\expandafter\def\csname mcresult@seed/fph/disc/MSLP/48/mae_mean\endcsname{10.26}
\expandafter\def\csname mcresult@seed/fph/disc/MSLP/48/mae_sd\endcsname{0.19}
\expandafter\def\csname mcresult@seed/fph/disc/MSLP/48/r2_mean\endcsname{0.51}
\expandafter\def\csname mcresult@seed/fph/disc/MSLP/48/r2_sd\endcsname{0.01}
\expandafter\def\csname mcresult@seed/fph/disc/MSLP/48/rmse_mean\endcsname{13.87}
\expandafter\def\csname mcresult@seed/fph/disc/MSLP/48/rmse_sd\endcsname{0.20}
\expandafter\def\csname mcresult@seed/fph/disc/MSLP/48/skill_score_mean\endcsname{0.30}
\expandafter\def\csname mcresult@seed/fph/disc/MSLP/48/skill_score_sd\endcsname{0.01}
\expandafter\def\csname mcresult@seed/fph/disc/MSLP/72/dir_acc_mean\endcsname{0.722}
\expandafter\def\csname mcresult@seed/fph/disc/MSLP/72/dir_acc_sd\endcsname{0.012}
\expandafter\def\csname mcresult@seed/fph/disc/MSLP/72/mae_mean\endcsname{12.94}
\expandafter\def\csname mcresult@seed/fph/disc/MSLP/72/mae_sd\endcsname{0.46}
\expandafter\def\csname mcresult@seed/fph/disc/MSLP/72/r2_mean\endcsname{0.35}
\expandafter\def\csname mcresult@seed/fph/disc/MSLP/72/r2_sd\endcsname{0.06}
\expandafter\def\csname mcresult@seed/fph/disc/MSLP/72/rmse_mean\endcsname{17.36}
\expandafter\def\csname mcresult@seed/fph/disc/MSLP/72/rmse_sd\endcsname{0.80}
\expandafter\def\csname mcresult@seed/fph/disc/MSLP/72/skill_score_mean\endcsname{0.19}
\expandafter\def\csname mcresult@seed/fph/disc/MSLP/72/skill_score_sd\endcsname{0.04}
\expandafter\def\csname mcresult@seed/fph/disc/MSLP/96/dir_acc_mean\endcsname{0.711}
\expandafter\def\csname mcresult@seed/fph/disc/MSLP/96/dir_acc_sd\endcsname{0.006}
\expandafter\def\csname mcresult@seed/fph/disc/MSLP/96/mae_mean\endcsname{15.26}
\expandafter\def\csname mcresult@seed/fph/disc/MSLP/96/mae_sd\endcsname{1.34}
\expandafter\def\csname mcresult@seed/fph/disc/MSLP/96/r2_mean\endcsname{0.14}
\expandafter\def\csname mcresult@seed/fph/disc/MSLP/96/r2_sd\endcsname{0.14}
\expandafter\def\csname mcresult@seed/fph/disc/MSLP/96/rmse_mean\endcsname{19.73}
\expandafter\def\csname mcresult@seed/fph/disc/MSLP/96/rmse_sd\endcsname{1.57}
\expandafter\def\csname mcresult@seed/fph/disc/MSLP/96/skill_score_mean\endcsname{0.07}
\expandafter\def\csname mcresult@seed/fph/disc/MSLP/96/skill_score_sd\endcsname{0.07}
\expandafter\def\csname mcresult@seed/fph/disc/VMAX/120/dir_acc_mean\endcsname{0.718}
\expandafter\def\csname mcresult@seed/fph/disc/VMAX/120/dir_acc_sd\endcsname{0.016}
\expandafter\def\csname mcresult@seed/fph/disc/VMAX/120/mae_mean\endcsname{23.42}
\expandafter\def\csname mcresult@seed/fph/disc/VMAX/120/mae_sd\endcsname{2.03}
\expandafter\def\csname mcresult@seed/fph/disc/VMAX/120/r2_mean\endcsname{0.00}
\expandafter\def\csname mcresult@seed/fph/disc/VMAX/120/r2_sd\endcsname{0.15}
\expandafter\def\csname mcresult@seed/fph/disc/VMAX/120/rmse_mean\endcsname{29.33}
\expandafter\def\csname mcresult@seed/fph/disc/VMAX/120/rmse_sd\endcsname{2.15}
\expandafter\def\csname mcresult@seed/fph/disc/VMAX/120/skill_score_mean\endcsname{0.00}
\expandafter\def\csname mcresult@seed/fph/disc/VMAX/120/skill_score_sd\endcsname{0.07}
\expandafter\def\csname mcresult@seed/fph/disc/VMAX/24/dir_acc_mean\endcsname{0.718}
\expandafter\def\csname mcresult@seed/fph/disc/VMAX/24/dir_acc_sd\endcsname{0.005}
\expandafter\def\csname mcresult@seed/fph/disc/VMAX/24/mae_mean\endcsname{9.49}
\expandafter\def\csname mcresult@seed/fph/disc/VMAX/24/mae_sd\endcsname{0.15}
\expandafter\def\csname mcresult@seed/fph/disc/VMAX/24/r2_mean\endcsname{0.53}
\expandafter\def\csname mcresult@seed/fph/disc/VMAX/24/r2_sd\endcsname{0.01}
\expandafter\def\csname mcresult@seed/fph/disc/VMAX/24/rmse_mean\endcsname{13.27}
\expandafter\def\csname mcresult@seed/fph/disc/VMAX/24/rmse_sd\endcsname{0.17}
\expandafter\def\csname mcresult@seed/fph/disc/VMAX/24/skill_score_mean\endcsname{0.31}
\expandafter\def\csname mcresult@seed/fph/disc/VMAX/24/skill_score_sd\endcsname{0.01}
\expandafter\def\csname mcresult@seed/fph/disc/VMAX/48/dir_acc_mean\endcsname{0.734}
\expandafter\def\csname mcresult@seed/fph/disc/VMAX/48/dir_acc_sd\endcsname{0.009}
\expandafter\def\csname mcresult@seed/fph/disc/VMAX/48/mae_mean\endcsname{14.32}
\expandafter\def\csname mcresult@seed/fph/disc/VMAX/48/mae_sd\endcsname{0.17}
\expandafter\def\csname mcresult@seed/fph/disc/VMAX/48/r2_mean\endcsname{0.54}
\expandafter\def\csname mcresult@seed/fph/disc/VMAX/48/r2_sd\endcsname{0.01}
\expandafter\def\csname mcresult@seed/fph/disc/VMAX/48/rmse_mean\endcsname{19.23}
\expandafter\def\csname mcresult@seed/fph/disc/VMAX/48/rmse_sd\endcsname{0.25}
\expandafter\def\csname mcresult@seed/fph/disc/VMAX/48/skill_score_mean\endcsname{0.32}
\expandafter\def\csname mcresult@seed/fph/disc/VMAX/48/skill_score_sd\endcsname{0.01}
\expandafter\def\csname mcresult@seed/fph/disc/VMAX/72/dir_acc_mean\endcsname{0.712}
\expandafter\def\csname mcresult@seed/fph/disc/VMAX/72/dir_acc_sd\endcsname{0.012}
\expandafter\def\csname mcresult@seed/fph/disc/VMAX/72/mae_mean\endcsname{17.63}
\expandafter\def\csname mcresult@seed/fph/disc/VMAX/72/mae_sd\endcsname{0.80}
\expandafter\def\csname mcresult@seed/fph/disc/VMAX/72/r2_mean\endcsname{0.42}
\expandafter\def\csname mcresult@seed/fph/disc/VMAX/72/r2_sd\endcsname{0.06}
\expandafter\def\csname mcresult@seed/fph/disc/VMAX/72/rmse_mean\endcsname{23.19}
\expandafter\def\csname mcresult@seed/fph/disc/VMAX/72/rmse_sd\endcsname{1.12}
\expandafter\def\csname mcresult@seed/fph/disc/VMAX/72/skill_score_mean\endcsname{0.24}
\expandafter\def\csname mcresult@seed/fph/disc/VMAX/72/skill_score_sd\endcsname{0.04}
\expandafter\def\csname mcresult@seed/fph/disc/VMAX/96/dir_acc_mean\endcsname{0.686}
\expandafter\def\csname mcresult@seed/fph/disc/VMAX/96/dir_acc_sd\endcsname{0.004}
\expandafter\def\csname mcresult@seed/fph/disc/VMAX/96/mae_mean\endcsname{20.80}
\expandafter\def\csname mcresult@seed/fph/disc/VMAX/96/mae_sd\endcsname{1.60}
\expandafter\def\csname mcresult@seed/fph/disc/VMAX/96/r2_mean\endcsname{0.19}
\expandafter\def\csname mcresult@seed/fph/disc/VMAX/96/r2_sd\endcsname{0.11}
\expandafter\def\csname mcresult@seed/fph/disc/VMAX/96/rmse_mean\endcsname{26.29}
\expandafter\def\csname mcresult@seed/fph/disc/VMAX/96/rmse_sd\endcsname{1.76}
\expandafter\def\csname mcresult@seed/fph/disc/VMAX/96/skill_score_mean\endcsname{0.10}
\expandafter\def\csname mcresult@seed/fph/disc/VMAX/96/skill_score_sd\endcsname{0.06}
\expandafter\def\csname mcresult@seed/fph/full/MSLP/120/dir_acc_mean\endcsname{0.738}
\expandafter\def\csname mcresult@seed/fph/full/MSLP/120/dir_acc_sd\endcsname{0.014}
\expandafter\def\csname mcresult@seed/fph/full/MSLP/120/mae_mean\endcsname{17.24}
\expandafter\def\csname mcresult@seed/fph/full/MSLP/120/mae_sd\endcsname{1.03}
\expandafter\def\csname mcresult@seed/fph/full/MSLP/120/r2_mean\endcsname{0.01}
\expandafter\def\csname mcresult@seed/fph/full/MSLP/120/r2_sd\endcsname{0.10}
\expandafter\def\csname mcresult@seed/fph/full/MSLP/120/rmse_mean\endcsname{21.74}
\expandafter\def\csname mcresult@seed/fph/full/MSLP/120/rmse_sd\endcsname{1.09}
\expandafter\def\csname mcresult@seed/fph/full/MSLP/120/skill_score_mean\endcsname{0.01}
\expandafter\def\csname mcresult@seed/fph/full/MSLP/120/skill_score_sd\endcsname{0.05}
\expandafter\def\csname mcresult@seed/fph/full/MSLP/24/dir_acc_mean\endcsname{0.758}
\expandafter\def\csname mcresult@seed/fph/full/MSLP/24/dir_acc_sd\endcsname{0.013}
\expandafter\def\csname mcresult@seed/fph/full/MSLP/24/mae_mean\endcsname{6.12}
\expandafter\def\csname mcresult@seed/fph/full/MSLP/24/mae_sd\endcsname{0.18}
\expandafter\def\csname mcresult@seed/fph/full/MSLP/24/r2_mean\endcsname{0.52}
\expandafter\def\csname mcresult@seed/fph/full/MSLP/24/r2_sd\endcsname{0.03}
\expandafter\def\csname mcresult@seed/fph/full/MSLP/24/rmse_mean\endcsname{9.03}
\expandafter\def\csname mcresult@seed/fph/full/MSLP/24/rmse_sd\endcsname{0.31}
\expandafter\def\csname mcresult@seed/fph/full/MSLP/24/skill_score_mean\endcsname{0.31}
\expandafter\def\csname mcresult@seed/fph/full/MSLP/24/skill_score_sd\endcsname{0.02}
\expandafter\def\csname mcresult@seed/fph/full/MSLP/48/dir_acc_mean\endcsname{0.782}
\expandafter\def\csname mcresult@seed/fph/full/MSLP/48/dir_acc_sd\endcsname{0.007}
\expandafter\def\csname mcresult@seed/fph/full/MSLP/48/mae_mean\endcsname{9.67}
\expandafter\def\csname mcresult@seed/fph/full/MSLP/48/mae_sd\endcsname{0.27}
\expandafter\def\csname mcresult@seed/fph/full/MSLP/48/r2_mean\endcsname{0.54}
\expandafter\def\csname mcresult@seed/fph/full/MSLP/48/r2_sd\endcsname{0.02}
\expandafter\def\csname mcresult@seed/fph/full/MSLP/48/rmse_mean\endcsname{13.34}
\expandafter\def\csname mcresult@seed/fph/full/MSLP/48/rmse_sd\endcsname{0.26}
\expandafter\def\csname mcresult@seed/fph/full/MSLP/48/skill_score_mean\endcsname{0.32}
\expandafter\def\csname mcresult@seed/fph/full/MSLP/48/skill_score_sd\endcsname{0.01}
\expandafter\def\csname mcresult@seed/fph/full/MSLP/72/dir_acc_mean\endcsname{0.740}
\expandafter\def\csname mcresult@seed/fph/full/MSLP/72/dir_acc_sd\endcsname{0.007}
\expandafter\def\csname mcresult@seed/fph/full/MSLP/72/mae_mean\endcsname{12.44}
\expandafter\def\csname mcresult@seed/fph/full/MSLP/72/mae_sd\endcsname{0.49}
\expandafter\def\csname mcresult@seed/fph/full/MSLP/72/r2_mean\endcsname{0.39}
\expandafter\def\csname mcresult@seed/fph/full/MSLP/72/r2_sd\endcsname{0.06}
\expandafter\def\csname mcresult@seed/fph/full/MSLP/72/rmse_mean\endcsname{16.66}
\expandafter\def\csname mcresult@seed/fph/full/MSLP/72/rmse_sd\endcsname{0.89}
\expandafter\def\csname mcresult@seed/fph/full/MSLP/72/skill_score_mean\endcsname{0.22}
\expandafter\def\csname mcresult@seed/fph/full/MSLP/72/skill_score_sd\endcsname{0.04}
\expandafter\def\csname mcresult@seed/fph/full/MSLP/96/dir_acc_mean\endcsname{0.736}
\expandafter\def\csname mcresult@seed/fph/full/MSLP/96/dir_acc_sd\endcsname{0.008}
\expandafter\def\csname mcresult@seed/fph/full/MSLP/96/mae_mean\endcsname{14.49}
\expandafter\def\csname mcresult@seed/fph/full/MSLP/96/mae_sd\endcsname{0.73}
\expandafter\def\csname mcresult@seed/fph/full/MSLP/96/r2_mean\endcsname{0.18}
\expandafter\def\csname mcresult@seed/fph/full/MSLP/96/r2_sd\endcsname{0.08}
\expandafter\def\csname mcresult@seed/fph/full/MSLP/96/rmse_mean\endcsname{19.12}
\expandafter\def\csname mcresult@seed/fph/full/MSLP/96/rmse_sd\endcsname{0.94}
\expandafter\def\csname mcresult@seed/fph/full/MSLP/96/skill_score_mean\endcsname{0.10}
\expandafter\def\csname mcresult@seed/fph/full/MSLP/96/skill_score_sd\endcsname{0.04}
\expandafter\def\csname mcresult@seed/fph/full/VMAX/120/dir_acc_mean\endcsname{0.724}
\expandafter\def\csname mcresult@seed/fph/full/VMAX/120/dir_acc_sd\endcsname{0.015}
\expandafter\def\csname mcresult@seed/fph/full/VMAX/120/mae_mean\endcsname{23.65}
\expandafter\def\csname mcresult@seed/fph/full/VMAX/120/mae_sd\endcsname{2.26}
\expandafter\def\csname mcresult@seed/fph/full/VMAX/120/r2_mean\endcsname{-0.00}
\expandafter\def\csname mcresult@seed/fph/full/VMAX/120/r2_sd\endcsname{0.13}
\expandafter\def\csname mcresult@seed/fph/full/VMAX/120/rmse_mean\endcsname{29.29}
\expandafter\def\csname mcresult@seed/fph/full/VMAX/120/rmse_sd\endcsname{1.94}
\expandafter\def\csname mcresult@seed/fph/full/VMAX/120/skill_score_mean\endcsname{0.00}
\expandafter\def\csname mcresult@seed/fph/full/VMAX/120/skill_score_sd\endcsname{0.07}
\expandafter\def\csname mcresult@seed/fph/full/VMAX/24/dir_acc_mean\endcsname{0.735}
\expandafter\def\csname mcresult@seed/fph/full/VMAX/24/dir_acc_sd\endcsname{0.004}
\expandafter\def\csname mcresult@seed/fph/full/VMAX/24/mae_mean\endcsname{9.15}
\expandafter\def\csname mcresult@seed/fph/full/VMAX/24/mae_sd\endcsname{0.19}
\expandafter\def\csname mcresult@seed/fph/full/VMAX/24/r2_mean\endcsname{0.55}
\expandafter\def\csname mcresult@seed/fph/full/VMAX/24/r2_sd\endcsname{0.02}
\expandafter\def\csname mcresult@seed/fph/full/VMAX/24/rmse_mean\endcsname{12.77}
\expandafter\def\csname mcresult@seed/fph/full/VMAX/24/rmse_sd\endcsname{0.35}
\expandafter\def\csname mcresult@seed/fph/full/VMAX/24/skill_score_mean\endcsname{0.33}
\expandafter\def\csname mcresult@seed/fph/full/VMAX/24/skill_score_sd\endcsname{0.02}
\expandafter\def\csname mcresult@seed/fph/full/VMAX/48/dir_acc_mean\endcsname{0.762}
\expandafter\def\csname mcresult@seed/fph/full/VMAX/48/dir_acc_sd\endcsname{0.007}
\expandafter\def\csname mcresult@seed/fph/full/VMAX/48/mae_mean\endcsname{13.65}
\expandafter\def\csname mcresult@seed/fph/full/VMAX/48/mae_sd\endcsname{0.22}
\expandafter\def\csname mcresult@seed/fph/full/VMAX/48/r2_mean\endcsname{0.57}
\expandafter\def\csname mcresult@seed/fph/full/VMAX/48/r2_sd\endcsname{0.01}
\expandafter\def\csname mcresult@seed/fph/full/VMAX/48/rmse_mean\endcsname{18.72}
\expandafter\def\csname mcresult@seed/fph/full/VMAX/48/rmse_sd\endcsname{0.26}
\expandafter\def\csname mcresult@seed/fph/full/VMAX/48/skill_score_mean\endcsname{0.34}
\expandafter\def\csname mcresult@seed/fph/full/VMAX/48/skill_score_sd\endcsname{0.01}
\expandafter\def\csname mcresult@seed/fph/full/VMAX/72/dir_acc_mean\endcsname{0.721}
\expandafter\def\csname mcresult@seed/fph/full/VMAX/72/dir_acc_sd\endcsname{0.005}
\expandafter\def\csname mcresult@seed/fph/full/VMAX/72/mae_mean\endcsname{17.11}
\expandafter\def\csname mcresult@seed/fph/full/VMAX/72/mae_sd\endcsname{0.61}
\expandafter\def\csname mcresult@seed/fph/full/VMAX/72/r2_mean\endcsname{0.45}
\expandafter\def\csname mcresult@seed/fph/full/VMAX/72/r2_sd\endcsname{0.04}
\expandafter\def\csname mcresult@seed/fph/full/VMAX/72/rmse_mean\endcsname{22.51}
\expandafter\def\csname mcresult@seed/fph/full/VMAX/72/rmse_sd\endcsname{0.94}
\expandafter\def\csname mcresult@seed/fph/full/VMAX/72/skill_score_mean\endcsname{0.26}
\expandafter\def\csname mcresult@seed/fph/full/VMAX/72/skill_score_sd\endcsname{0.03}
\expandafter\def\csname mcresult@seed/fph/full/VMAX/96/dir_acc_mean\endcsname{0.707}
\expandafter\def\csname mcresult@seed/fph/full/VMAX/96/dir_acc_sd\endcsname{0.018}
\expandafter\def\csname mcresult@seed/fph/full/VMAX/96/mae_mean\endcsname{20.43}
\expandafter\def\csname mcresult@seed/fph/full/VMAX/96/mae_sd\endcsname{0.81}
\expandafter\def\csname mcresult@seed/fph/full/VMAX/96/r2_mean\endcsname{0.19}
\expandafter\def\csname mcresult@seed/fph/full/VMAX/96/r2_sd\endcsname{0.08}
\expandafter\def\csname mcresult@seed/fph/full/VMAX/96/rmse_mean\endcsname{26.27}
\expandafter\def\csname mcresult@seed/fph/full/VMAX/96/rmse_sd\endcsname{1.37}
\expandafter\def\csname mcresult@seed/fph/full/VMAX/96/skill_score_mean\endcsname{0.10}
\expandafter\def\csname mcresult@seed/fph/full/VMAX/96/skill_score_sd\endcsname{0.05}
\expandafter\def\csname mcresult@seed/fph/img+disc/MSLP/120/dir_acc_mean\endcsname{0.750}
\expandafter\def\csname mcresult@seed/fph/img+disc/MSLP/120/dir_acc_sd\endcsname{0.009}
\expandafter\def\csname mcresult@seed/fph/img+disc/MSLP/120/mae_mean\endcsname{17.18}
\expandafter\def\csname mcresult@seed/fph/img+disc/MSLP/120/mae_sd\endcsname{1.20}
\expandafter\def\csname mcresult@seed/fph/img+disc/MSLP/120/r2_mean\endcsname{0.02}
\expandafter\def\csname mcresult@seed/fph/img+disc/MSLP/120/r2_sd\endcsname{0.15}
\expandafter\def\csname mcresult@seed/fph/img+disc/MSLP/120/rmse_mean\endcsname{21.61}
\expandafter\def\csname mcresult@seed/fph/img+disc/MSLP/120/rmse_sd\endcsname{1.72}
\expandafter\def\csname mcresult@seed/fph/img+disc/MSLP/120/skill_score_mean\endcsname{0.01}
\expandafter\def\csname mcresult@seed/fph/img+disc/MSLP/120/skill_score_sd\endcsname{0.08}
\expandafter\def\csname mcresult@seed/fph/img+disc/MSLP/24/dir_acc_mean\endcsname{0.740}
\expandafter\def\csname mcresult@seed/fph/img+disc/MSLP/24/dir_acc_sd\endcsname{0.013}
\expandafter\def\csname mcresult@seed/fph/img+disc/MSLP/24/mae_mean\endcsname{6.49}
\expandafter\def\csname mcresult@seed/fph/img+disc/MSLP/24/mae_sd\endcsname{0.26}
\expandafter\def\csname mcresult@seed/fph/img+disc/MSLP/24/r2_mean\endcsname{0.47}
\expandafter\def\csname mcresult@seed/fph/img+disc/MSLP/24/r2_sd\endcsname{0.03}
\expandafter\def\csname mcresult@seed/fph/img+disc/MSLP/24/rmse_mean\endcsname{9.51}
\expandafter\def\csname mcresult@seed/fph/img+disc/MSLP/24/rmse_sd\endcsname{0.23}
\expandafter\def\csname mcresult@seed/fph/img+disc/MSLP/24/skill_score_mean\endcsname{0.27}
\expandafter\def\csname mcresult@seed/fph/img+disc/MSLP/24/skill_score_sd\endcsname{0.02}
\expandafter\def\csname mcresult@seed/fph/img+disc/MSLP/48/dir_acc_mean\endcsname{0.743}
\expandafter\def\csname mcresult@seed/fph/img+disc/MSLP/48/dir_acc_sd\endcsname{0.026}
\expandafter\def\csname mcresult@seed/fph/img+disc/MSLP/48/mae_mean\endcsname{10.11}
\expandafter\def\csname mcresult@seed/fph/img+disc/MSLP/48/mae_sd\endcsname{0.15}
\expandafter\def\csname mcresult@seed/fph/img+disc/MSLP/48/r2_mean\endcsname{0.49}
\expandafter\def\csname mcresult@seed/fph/img+disc/MSLP/48/r2_sd\endcsname{0.02}
\expandafter\def\csname mcresult@seed/fph/img+disc/MSLP/48/rmse_mean\endcsname{14.06}
\expandafter\def\csname mcresult@seed/fph/img+disc/MSLP/48/rmse_sd\endcsname{0.31}
\expandafter\def\csname mcresult@seed/fph/img+disc/MSLP/48/skill_score_mean\endcsname{0.29}
\expandafter\def\csname mcresult@seed/fph/img+disc/MSLP/48/skill_score_sd\endcsname{0.02}
\expandafter\def\csname mcresult@seed/fph/img+disc/MSLP/72/dir_acc_mean\endcsname{0.711}
\expandafter\def\csname mcresult@seed/fph/img+disc/MSLP/72/dir_acc_sd\endcsname{0.018}
\expandafter\def\csname mcresult@seed/fph/img+disc/MSLP/72/mae_mean\endcsname{12.93}
\expandafter\def\csname mcresult@seed/fph/img+disc/MSLP/72/mae_sd\endcsname{0.59}
\expandafter\def\csname mcresult@seed/fph/img+disc/MSLP/72/r2_mean\endcsname{0.36}
\expandafter\def\csname mcresult@seed/fph/img+disc/MSLP/72/r2_sd\endcsname{0.05}
\expandafter\def\csname mcresult@seed/fph/img+disc/MSLP/72/rmse_mean\endcsname{17.09}
\expandafter\def\csname mcresult@seed/fph/img+disc/MSLP/72/rmse_sd\endcsname{0.66}
\expandafter\def\csname mcresult@seed/fph/img+disc/MSLP/72/skill_score_mean\endcsname{0.20}
\expandafter\def\csname mcresult@seed/fph/img+disc/MSLP/72/skill_score_sd\endcsname{0.03}
\expandafter\def\csname mcresult@seed/fph/img+disc/MSLP/96/dir_acc_mean\endcsname{0.696}
\expandafter\def\csname mcresult@seed/fph/img+disc/MSLP/96/dir_acc_sd\endcsname{0.034}
\expandafter\def\csname mcresult@seed/fph/img+disc/MSLP/96/mae_mean\endcsname{15.61}
\expandafter\def\csname mcresult@seed/fph/img+disc/MSLP/96/mae_sd\endcsname{1.76}
\expandafter\def\csname mcresult@seed/fph/img+disc/MSLP/96/r2_mean\endcsname{0.03}
\expandafter\def\csname mcresult@seed/fph/img+disc/MSLP/96/r2_sd\endcsname{0.22}
\expandafter\def\csname mcresult@seed/fph/img+disc/MSLP/96/rmse_mean\endcsname{20.72}
\expandafter\def\csname mcresult@seed/fph/img+disc/MSLP/96/rmse_sd\endcsname{2.34}
\expandafter\def\csname mcresult@seed/fph/img+disc/MSLP/96/skill_score_mean\endcsname{0.02}
\expandafter\def\csname mcresult@seed/fph/img+disc/MSLP/96/skill_score_sd\endcsname{0.11}
\expandafter\def\csname mcresult@seed/fph/img+disc/VMAX/120/dir_acc_mean\endcsname{0.728}
\expandafter\def\csname mcresult@seed/fph/img+disc/VMAX/120/dir_acc_sd\endcsname{0.017}
\expandafter\def\csname mcresult@seed/fph/img+disc/VMAX/120/mae_mean\endcsname{23.13}
\expandafter\def\csname mcresult@seed/fph/img+disc/VMAX/120/mae_sd\endcsname{2.57}
\expandafter\def\csname mcresult@seed/fph/img+disc/VMAX/120/r2_mean\endcsname{0.05}
\expandafter\def\csname mcresult@seed/fph/img+disc/VMAX/120/r2_sd\endcsname{0.18}
\expandafter\def\csname mcresult@seed/fph/img+disc/VMAX/120/rmse_mean\endcsname{28.40}
\expandafter\def\csname mcresult@seed/fph/img+disc/VMAX/120/rmse_sd\endcsname{2.81}
\expandafter\def\csname mcresult@seed/fph/img+disc/VMAX/120/skill_score_mean\endcsname{0.03}
\expandafter\def\csname mcresult@seed/fph/img+disc/VMAX/120/skill_score_sd\endcsname{0.10}
\expandafter\def\csname mcresult@seed/fph/img+disc/VMAX/24/dir_acc_mean\endcsname{0.685}
\expandafter\def\csname mcresult@seed/fph/img+disc/VMAX/24/dir_acc_sd\endcsname{0.027}
\expandafter\def\csname mcresult@seed/fph/img+disc/VMAX/24/mae_mean\endcsname{10.06}
\expandafter\def\csname mcresult@seed/fph/img+disc/VMAX/24/mae_sd\endcsname{0.39}
\expandafter\def\csname mcresult@seed/fph/img+disc/VMAX/24/r2_mean\endcsname{0.47}
\expandafter\def\csname mcresult@seed/fph/img+disc/VMAX/24/r2_sd\endcsname{0.03}
\expandafter\def\csname mcresult@seed/fph/img+disc/VMAX/24/rmse_mean\endcsname{13.84}
\expandafter\def\csname mcresult@seed/fph/img+disc/VMAX/24/rmse_sd\endcsname{0.36}
\expandafter\def\csname mcresult@seed/fph/img+disc/VMAX/24/skill_score_mean\endcsname{0.27}
\expandafter\def\csname mcresult@seed/fph/img+disc/VMAX/24/skill_score_sd\endcsname{0.02}
\expandafter\def\csname mcresult@seed/fph/img+disc/VMAX/48/dir_acc_mean\endcsname{0.728}
\expandafter\def\csname mcresult@seed/fph/img+disc/VMAX/48/dir_acc_sd\endcsname{0.027}
\expandafter\def\csname mcresult@seed/fph/img+disc/VMAX/48/mae_mean\endcsname{14.38}
\expandafter\def\csname mcresult@seed/fph/img+disc/VMAX/48/mae_sd\endcsname{0.54}
\expandafter\def\csname mcresult@seed/fph/img+disc/VMAX/48/r2_mean\endcsname{0.52}
\expandafter\def\csname mcresult@seed/fph/img+disc/VMAX/48/r2_sd\endcsname{0.03}
\expandafter\def\csname mcresult@seed/fph/img+disc/VMAX/48/rmse_mean\endcsname{19.78}
\expandafter\def\csname mcresult@seed/fph/img+disc/VMAX/48/rmse_sd\endcsname{0.54}
\expandafter\def\csname mcresult@seed/fph/img+disc/VMAX/48/skill_score_mean\endcsname{0.31}
\expandafter\def\csname mcresult@seed/fph/img+disc/VMAX/48/skill_score_sd\endcsname{0.02}
\expandafter\def\csname mcresult@seed/fph/img+disc/VMAX/72/dir_acc_mean\endcsname{0.700}
\expandafter\def\csname mcresult@seed/fph/img+disc/VMAX/72/dir_acc_sd\endcsname{0.032}
\expandafter\def\csname mcresult@seed/fph/img+disc/VMAX/72/mae_mean\endcsname{18.08}
\expandafter\def\csname mcresult@seed/fph/img+disc/VMAX/72/mae_sd\endcsname{1.48}
\expandafter\def\csname mcresult@seed/fph/img+disc/VMAX/72/r2_mean\endcsname{0.39}
\expandafter\def\csname mcresult@seed/fph/img+disc/VMAX/72/r2_sd\endcsname{0.10}
\expandafter\def\csname mcresult@seed/fph/img+disc/VMAX/72/rmse_mean\endcsname{23.68}
\expandafter\def\csname mcresult@seed/fph/img+disc/VMAX/72/rmse_sd\endcsname{1.86}
\expandafter\def\csname mcresult@seed/fph/img+disc/VMAX/72/skill_score_mean\endcsname{0.22}
\expandafter\def\csname mcresult@seed/fph/img+disc/VMAX/72/skill_score_sd\endcsname{0.06}
\expandafter\def\csname mcresult@seed/fph/img+disc/VMAX/96/dir_acc_mean\endcsname{0.695}
\expandafter\def\csname mcresult@seed/fph/img+disc/VMAX/96/dir_acc_sd\endcsname{0.016}
\expandafter\def\csname mcresult@seed/fph/img+disc/VMAX/96/mae_mean\endcsname{21.62}
\expandafter\def\csname mcresult@seed/fph/img+disc/VMAX/96/mae_sd\endcsname{2.66}
\expandafter\def\csname mcresult@seed/fph/img+disc/VMAX/96/r2_mean\endcsname{0.08}
\expandafter\def\csname mcresult@seed/fph/img+disc/VMAX/96/r2_sd\endcsname{0.19}
\expandafter\def\csname mcresult@seed/fph/img+disc/VMAX/96/rmse_mean\endcsname{27.98}
\expandafter\def\csname mcresult@seed/fph/img+disc/VMAX/96/rmse_sd\endcsname{2.93}
\expandafter\def\csname mcresult@seed/fph/img+disc/VMAX/96/skill_score_mean\endcsname{0.05}
\expandafter\def\csname mcresult@seed/fph/img+disc/VMAX/96/skill_score_sd\endcsname{0.10}
\expandafter\def\csname mcresult@seed/fph/img/MSLP/120/dir_acc_mean\endcsname{0.757}
\expandafter\def\csname mcresult@seed/fph/img/MSLP/120/dir_acc_sd\endcsname{0.017}
\expandafter\def\csname mcresult@seed/fph/img/MSLP/120/mae_mean\endcsname{16.36}
\expandafter\def\csname mcresult@seed/fph/img/MSLP/120/mae_sd\endcsname{0.28}
\expandafter\def\csname mcresult@seed/fph/img/MSLP/120/r2_mean\endcsname{0.05}
\expandafter\def\csname mcresult@seed/fph/img/MSLP/120/r2_sd\endcsname{0.03}
\expandafter\def\csname mcresult@seed/fph/img/MSLP/120/rmse_mean\endcsname{21.33}
\expandafter\def\csname mcresult@seed/fph/img/MSLP/120/rmse_sd\endcsname{0.33}
\expandafter\def\csname mcresult@seed/fph/img/MSLP/120/skill_score_mean\endcsname{0.02}
\expandafter\def\csname mcresult@seed/fph/img/MSLP/120/skill_score_sd\endcsname{0.01}
\expandafter\def\csname mcresult@seed/fph/img/MSLP/24/dir_acc_mean\endcsname{0.649}
\expandafter\def\csname mcresult@seed/fph/img/MSLP/24/dir_acc_sd\endcsname{0.012}
\expandafter\def\csname mcresult@seed/fph/img/MSLP/24/mae_mean\endcsname{8.18}
\expandafter\def\csname mcresult@seed/fph/img/MSLP/24/mae_sd\endcsname{0.16}
\expandafter\def\csname mcresult@seed/fph/img/MSLP/24/r2_mean\endcsname{0.18}
\expandafter\def\csname mcresult@seed/fph/img/MSLP/24/r2_sd\endcsname{0.03}
\expandafter\def\csname mcresult@seed/fph/img/MSLP/24/rmse_mean\endcsname{11.82}
\expandafter\def\csname mcresult@seed/fph/img/MSLP/24/rmse_sd\endcsname{0.22}
\expandafter\def\csname mcresult@seed/fph/img/MSLP/24/skill_score_mean\endcsname{0.09}
\expandafter\def\csname mcresult@seed/fph/img/MSLP/24/skill_score_sd\endcsname{0.02}
\expandafter\def\csname mcresult@seed/fph/img/MSLP/48/dir_acc_mean\endcsname{0.702}
\expandafter\def\csname mcresult@seed/fph/img/MSLP/48/dir_acc_sd\endcsname{0.013}
\expandafter\def\csname mcresult@seed/fph/img/MSLP/48/mae_mean\endcsname{13.31}
\expandafter\def\csname mcresult@seed/fph/img/MSLP/48/mae_sd\endcsname{0.21}
\expandafter\def\csname mcresult@seed/fph/img/MSLP/48/r2_mean\endcsname{0.12}
\expandafter\def\csname mcresult@seed/fph/img/MSLP/48/r2_sd\endcsname{0.06}
\expandafter\def\csname mcresult@seed/fph/img/MSLP/48/rmse_mean\endcsname{18.49}
\expandafter\def\csname mcresult@seed/fph/img/MSLP/48/rmse_sd\endcsname{0.65}
\expandafter\def\csname mcresult@seed/fph/img/MSLP/48/skill_score_mean\endcsname{0.06}
\expandafter\def\csname mcresult@seed/fph/img/MSLP/48/skill_score_sd\endcsname{0.03}
\expandafter\def\csname mcresult@seed/fph/img/MSLP/72/dir_acc_mean\endcsname{0.692}
\expandafter\def\csname mcresult@seed/fph/img/MSLP/72/dir_acc_sd\endcsname{0.030}
\expandafter\def\csname mcresult@seed/fph/img/MSLP/72/mae_mean\endcsname{15.13}
\expandafter\def\csname mcresult@seed/fph/img/MSLP/72/mae_sd\endcsname{0.52}
\expandafter\def\csname mcresult@seed/fph/img/MSLP/72/r2_mean\endcsname{0.15}
\expandafter\def\csname mcresult@seed/fph/img/MSLP/72/r2_sd\endcsname{0.08}
\expandafter\def\csname mcresult@seed/fph/img/MSLP/72/rmse_mean\endcsname{19.67}
\expandafter\def\csname mcresult@seed/fph/img/MSLP/72/rmse_sd\endcsname{0.89}
\expandafter\def\csname mcresult@seed/fph/img/MSLP/72/skill_score_mean\endcsname{0.08}
\expandafter\def\csname mcresult@seed/fph/img/MSLP/72/skill_score_sd\endcsname{0.04}
\expandafter\def\csname mcresult@seed/fph/img/MSLP/96/dir_acc_mean\endcsname{0.702}
\expandafter\def\csname mcresult@seed/fph/img/MSLP/96/dir_acc_sd\endcsname{0.027}
\expandafter\def\csname mcresult@seed/fph/img/MSLP/96/mae_mean\endcsname{15.67}
\expandafter\def\csname mcresult@seed/fph/img/MSLP/96/mae_sd\endcsname{0.61}
\expandafter\def\csname mcresult@seed/fph/img/MSLP/96/r2_mean\endcsname{0.07}
\expandafter\def\csname mcresult@seed/fph/img/MSLP/96/r2_sd\endcsname{0.09}
\expandafter\def\csname mcresult@seed/fph/img/MSLP/96/rmse_mean\endcsname{20.30}
\expandafter\def\csname mcresult@seed/fph/img/MSLP/96/rmse_sd\endcsname{0.94}
\expandafter\def\csname mcresult@seed/fph/img/MSLP/96/skill_score_mean\endcsname{0.04}
\expandafter\def\csname mcresult@seed/fph/img/MSLP/96/skill_score_sd\endcsname{0.04}
\expandafter\def\csname mcresult@seed/fph/img/VMAX/120/dir_acc_mean\endcsname{0.720}
\expandafter\def\csname mcresult@seed/fph/img/VMAX/120/dir_acc_sd\endcsname{0.017}
\expandafter\def\csname mcresult@seed/fph/img/VMAX/120/mae_mean\endcsname{20.76}
\expandafter\def\csname mcresult@seed/fph/img/VMAX/120/mae_sd\endcsname{0.62}
\expandafter\def\csname mcresult@seed/fph/img/VMAX/120/r2_mean\endcsname{0.15}
\expandafter\def\csname mcresult@seed/fph/img/VMAX/120/r2_sd\endcsname{0.04}
\expandafter\def\csname mcresult@seed/fph/img/VMAX/120/rmse_mean\endcsname{26.95}
\expandafter\def\csname mcresult@seed/fph/img/VMAX/120/rmse_sd\endcsname{0.62}
\expandafter\def\csname mcresult@seed/fph/img/VMAX/120/skill_score_mean\endcsname{0.08}
\expandafter\def\csname mcresult@seed/fph/img/VMAX/120/skill_score_sd\endcsname{0.02}
\expandafter\def\csname mcresult@seed/fph/img/VMAX/24/dir_acc_mean\endcsname{0.612}
\expandafter\def\csname mcresult@seed/fph/img/VMAX/24/dir_acc_sd\endcsname{0.016}
\expandafter\def\csname mcresult@seed/fph/img/VMAX/24/mae_mean\endcsname{12.43}
\expandafter\def\csname mcresult@seed/fph/img/VMAX/24/mae_sd\endcsname{0.26}
\expandafter\def\csname mcresult@seed/fph/img/VMAX/24/r2_mean\endcsname{0.21}
\expandafter\def\csname mcresult@seed/fph/img/VMAX/24/r2_sd\endcsname{0.03}
\expandafter\def\csname mcresult@seed/fph/img/VMAX/24/rmse_mean\endcsname{16.93}
\expandafter\def\csname mcresult@seed/fph/img/VMAX/24/rmse_sd\endcsname{0.29}
\expandafter\def\csname mcresult@seed/fph/img/VMAX/24/skill_score_mean\endcsname{0.11}
\expandafter\def\csname mcresult@seed/fph/img/VMAX/24/skill_score_sd\endcsname{0.02}
\expandafter\def\csname mcresult@seed/fph/img/VMAX/48/dir_acc_mean\endcsname{0.674}
\expandafter\def\csname mcresult@seed/fph/img/VMAX/48/dir_acc_sd\endcsname{0.011}
\expandafter\def\csname mcresult@seed/fph/img/VMAX/48/mae_mean\endcsname{18.97}
\expandafter\def\csname mcresult@seed/fph/img/VMAX/48/mae_sd\endcsname{0.37}
\expandafter\def\csname mcresult@seed/fph/img/VMAX/48/r2_mean\endcsname{0.16}
\expandafter\def\csname mcresult@seed/fph/img/VMAX/48/r2_sd\endcsname{0.06}
\expandafter\def\csname mcresult@seed/fph/img/VMAX/48/rmse_mean\endcsname{26.03}
\expandafter\def\csname mcresult@seed/fph/img/VMAX/48/rmse_sd\endcsname{0.95}
\expandafter\def\csname mcresult@seed/fph/img/VMAX/48/skill_score_mean\endcsname{0.09}
\expandafter\def\csname mcresult@seed/fph/img/VMAX/48/skill_score_sd\endcsname{0.03}
\expandafter\def\csname mcresult@seed/fph/img/VMAX/72/dir_acc_mean\endcsname{0.664}
\expandafter\def\csname mcresult@seed/fph/img/VMAX/72/dir_acc_sd\endcsname{0.017}
\expandafter\def\csname mcresult@seed/fph/img/VMAX/72/mae_mean\endcsname{20.43}
\expandafter\def\csname mcresult@seed/fph/img/VMAX/72/mae_sd\endcsname{0.75}
\expandafter\def\csname mcresult@seed/fph/img/VMAX/72/r2_mean\endcsname{0.24}
\expandafter\def\csname mcresult@seed/fph/img/VMAX/72/r2_sd\endcsname{0.06}
\expandafter\def\csname mcresult@seed/fph/img/VMAX/72/rmse_mean\endcsname{26.44}
\expandafter\def\csname mcresult@seed/fph/img/VMAX/72/rmse_sd\endcsname{1.04}
\expandafter\def\csname mcresult@seed/fph/img/VMAX/72/skill_score_mean\endcsname{0.13}
\expandafter\def\csname mcresult@seed/fph/img/VMAX/72/skill_score_sd\endcsname{0.03}
\expandafter\def\csname mcresult@seed/fph/img/VMAX/96/dir_acc_mean\endcsname{0.680}
\expandafter\def\csname mcresult@seed/fph/img/VMAX/96/dir_acc_sd\endcsname{0.022}
\expandafter\def\csname mcresult@seed/fph/img/VMAX/96/mae_mean\endcsname{21.00}
\expandafter\def\csname mcresult@seed/fph/img/VMAX/96/mae_sd\endcsname{0.94}
\expandafter\def\csname mcresult@seed/fph/img/VMAX/96/r2_mean\endcsname{0.18}
\expandafter\def\csname mcresult@seed/fph/img/VMAX/96/r2_sd\endcsname{0.07}
\expandafter\def\csname mcresult@seed/fph/img/VMAX/96/rmse_mean\endcsname{26.50}
\expandafter\def\csname mcresult@seed/fph/img/VMAX/96/rmse_sd\endcsname{1.16}
\expandafter\def\csname mcresult@seed/fph/img/VMAX/96/skill_score_mean\endcsname{0.10}
\expandafter\def\csname mcresult@seed/fph/img/VMAX/96/skill_score_sd\endcsname{0.04}
\expandafter\def\csname mcresult@seed/fph/tab+disc/MSLP/120/dir_acc_mean\endcsname{0.754}
\expandafter\def\csname mcresult@seed/fph/tab+disc/MSLP/120/dir_acc_sd\endcsname{0.011}
\expandafter\def\csname mcresult@seed/fph/tab+disc/MSLP/120/mae_mean\endcsname{16.72}
\expandafter\def\csname mcresult@seed/fph/tab+disc/MSLP/120/mae_sd\endcsname{1.10}
\expandafter\def\csname mcresult@seed/fph/tab+disc/MSLP/120/r2_mean\endcsname{0.06}
\expandafter\def\csname mcresult@seed/fph/tab+disc/MSLP/120/r2_sd\endcsname{0.07}
\expandafter\def\csname mcresult@seed/fph/tab+disc/MSLP/120/rmse_mean\endcsname{21.17}
\expandafter\def\csname mcresult@seed/fph/tab+disc/MSLP/120/rmse_sd\endcsname{0.74}
\expandafter\def\csname mcresult@seed/fph/tab+disc/MSLP/120/skill_score_mean\endcsname{0.03}
\expandafter\def\csname mcresult@seed/fph/tab+disc/MSLP/120/skill_score_sd\endcsname{0.03}
\expandafter\def\csname mcresult@seed/fph/tab+disc/MSLP/24/dir_acc_mean\endcsname{0.739}
\expandafter\def\csname mcresult@seed/fph/tab+disc/MSLP/24/dir_acc_sd\endcsname{0.024}
\expandafter\def\csname mcresult@seed/fph/tab+disc/MSLP/24/mae_mean\endcsname{6.28}
\expandafter\def\csname mcresult@seed/fph/tab+disc/MSLP/24/mae_sd\endcsname{0.27}
\expandafter\def\csname mcresult@seed/fph/tab+disc/MSLP/24/r2_mean\endcsname{0.50}
\expandafter\def\csname mcresult@seed/fph/tab+disc/MSLP/24/r2_sd\endcsname{0.04}
\expandafter\def\csname mcresult@seed/fph/tab+disc/MSLP/24/rmse_mean\endcsname{9.21}
\expandafter\def\csname mcresult@seed/fph/tab+disc/MSLP/24/rmse_sd\endcsname{0.36}
\expandafter\def\csname mcresult@seed/fph/tab+disc/MSLP/24/skill_score_mean\endcsname{0.29}
\expandafter\def\csname mcresult@seed/fph/tab+disc/MSLP/24/skill_score_sd\endcsname{0.03}
\expandafter\def\csname mcresult@seed/fph/tab+disc/MSLP/48/dir_acc_mean\endcsname{0.781}
\expandafter\def\csname mcresult@seed/fph/tab+disc/MSLP/48/dir_acc_sd\endcsname{0.006}
\expandafter\def\csname mcresult@seed/fph/tab+disc/MSLP/48/mae_mean\endcsname{9.58}
\expandafter\def\csname mcresult@seed/fph/tab+disc/MSLP/48/mae_sd\endcsname{0.30}
\expandafter\def\csname mcresult@seed/fph/tab+disc/MSLP/48/r2_mean\endcsname{0.55}
\expandafter\def\csname mcresult@seed/fph/tab+disc/MSLP/48/r2_sd\endcsname{0.02}
\expandafter\def\csname mcresult@seed/fph/tab+disc/MSLP/48/rmse_mean\endcsname{13.15}
\expandafter\def\csname mcresult@seed/fph/tab+disc/MSLP/48/rmse_sd\endcsname{0.32}
\expandafter\def\csname mcresult@seed/fph/tab+disc/MSLP/48/skill_score_mean\endcsname{0.33}
\expandafter\def\csname mcresult@seed/fph/tab+disc/MSLP/48/skill_score_sd\endcsname{0.02}
\expandafter\def\csname mcresult@seed/fph/tab+disc/MSLP/72/dir_acc_mean\endcsname{0.743}
\expandafter\def\csname mcresult@seed/fph/tab+disc/MSLP/72/dir_acc_sd\endcsname{0.007}
\expandafter\def\csname mcresult@seed/fph/tab+disc/MSLP/72/mae_mean\endcsname{12.50}
\expandafter\def\csname mcresult@seed/fph/tab+disc/MSLP/72/mae_sd\endcsname{0.32}
\expandafter\def\csname mcresult@seed/fph/tab+disc/MSLP/72/r2_mean\endcsname{0.39}
\expandafter\def\csname mcresult@seed/fph/tab+disc/MSLP/72/r2_sd\endcsname{0.06}
\expandafter\def\csname mcresult@seed/fph/tab+disc/MSLP/72/rmse_mean\endcsname{16.66}
\expandafter\def\csname mcresult@seed/fph/tab+disc/MSLP/72/rmse_sd\endcsname{0.80}
\expandafter\def\csname mcresult@seed/fph/tab+disc/MSLP/72/skill_score_mean\endcsname{0.22}
\expandafter\def\csname mcresult@seed/fph/tab+disc/MSLP/72/skill_score_sd\endcsname{0.04}
\expandafter\def\csname mcresult@seed/fph/tab+disc/MSLP/96/dir_acc_mean\endcsname{0.729}
\expandafter\def\csname mcresult@seed/fph/tab+disc/MSLP/96/dir_acc_sd\endcsname{0.012}
\expandafter\def\csname mcresult@seed/fph/tab+disc/MSLP/96/mae_mean\endcsname{14.49}
\expandafter\def\csname mcresult@seed/fph/tab+disc/MSLP/96/mae_sd\endcsname{0.80}
\expandafter\def\csname mcresult@seed/fph/tab+disc/MSLP/96/r2_mean\endcsname{0.19}
\expandafter\def\csname mcresult@seed/fph/tab+disc/MSLP/96/r2_sd\endcsname{0.09}
\expandafter\def\csname mcresult@seed/fph/tab+disc/MSLP/96/rmse_mean\endcsname{19.02}
\expandafter\def\csname mcresult@seed/fph/tab+disc/MSLP/96/rmse_sd\endcsname{1.08}
\expandafter\def\csname mcresult@seed/fph/tab+disc/MSLP/96/skill_score_mean\endcsname{0.10}
\expandafter\def\csname mcresult@seed/fph/tab+disc/MSLP/96/skill_score_sd\endcsname{0.05}
\expandafter\def\csname mcresult@seed/fph/tab+disc/VMAX/120/dir_acc_mean\endcsname{0.731}
\expandafter\def\csname mcresult@seed/fph/tab+disc/VMAX/120/dir_acc_sd\endcsname{0.019}
\expandafter\def\csname mcresult@seed/fph/tab+disc/VMAX/120/mae_mean\endcsname{22.51}
\expandafter\def\csname mcresult@seed/fph/tab+disc/VMAX/120/mae_sd\endcsname{1.49}
\expandafter\def\csname mcresult@seed/fph/tab+disc/VMAX/120/r2_mean\endcsname{0.06}
\expandafter\def\csname mcresult@seed/fph/tab+disc/VMAX/120/r2_sd\endcsname{0.07}
\expandafter\def\csname mcresult@seed/fph/tab+disc/VMAX/120/rmse_mean\endcsname{28.43}
\expandafter\def\csname mcresult@seed/fph/tab+disc/VMAX/120/rmse_sd\endcsname{1.13}
\expandafter\def\csname mcresult@seed/fph/tab+disc/VMAX/120/skill_score_mean\endcsname{0.03}
\expandafter\def\csname mcresult@seed/fph/tab+disc/VMAX/120/skill_score_sd\endcsname{0.04}
\expandafter\def\csname mcresult@seed/fph/tab+disc/VMAX/24/dir_acc_mean\endcsname{0.720}
\expandafter\def\csname mcresult@seed/fph/tab+disc/VMAX/24/dir_acc_sd\endcsname{0.014}
\expandafter\def\csname mcresult@seed/fph/tab+disc/VMAX/24/mae_mean\endcsname{9.26}
\expandafter\def\csname mcresult@seed/fph/tab+disc/VMAX/24/mae_sd\endcsname{0.29}
\expandafter\def\csname mcresult@seed/fph/tab+disc/VMAX/24/r2_mean\endcsname{0.54}
\expandafter\def\csname mcresult@seed/fph/tab+disc/VMAX/24/r2_sd\endcsname{0.04}
\expandafter\def\csname mcresult@seed/fph/tab+disc/VMAX/24/rmse_mean\endcsname{12.92}
\expandafter\def\csname mcresult@seed/fph/tab+disc/VMAX/24/rmse_sd\endcsname{0.50}
\expandafter\def\csname mcresult@seed/fph/tab+disc/VMAX/24/skill_score_mean\endcsname{0.32}
\expandafter\def\csname mcresult@seed/fph/tab+disc/VMAX/24/skill_score_sd\endcsname{0.03}
\expandafter\def\csname mcresult@seed/fph/tab+disc/VMAX/48/dir_acc_mean\endcsname{0.762}
\expandafter\def\csname mcresult@seed/fph/tab+disc/VMAX/48/dir_acc_sd\endcsname{0.006}
\expandafter\def\csname mcresult@seed/fph/tab+disc/VMAX/48/mae_mean\endcsname{13.44}
\expandafter\def\csname mcresult@seed/fph/tab+disc/VMAX/48/mae_sd\endcsname{0.23}
\expandafter\def\csname mcresult@seed/fph/tab+disc/VMAX/48/r2_mean\endcsname{0.58}
\expandafter\def\csname mcresult@seed/fph/tab+disc/VMAX/48/r2_sd\endcsname{0.02}
\expandafter\def\csname mcresult@seed/fph/tab+disc/VMAX/48/rmse_mean\endcsname{18.43}
\expandafter\def\csname mcresult@seed/fph/tab+disc/VMAX/48/rmse_sd\endcsname{0.36}
\expandafter\def\csname mcresult@seed/fph/tab+disc/VMAX/48/skill_score_mean\endcsname{0.35}
\expandafter\def\csname mcresult@seed/fph/tab+disc/VMAX/48/skill_score_sd\endcsname{0.01}
\expandafter\def\csname mcresult@seed/fph/tab+disc/VMAX/72/dir_acc_mean\endcsname{0.729}
\expandafter\def\csname mcresult@seed/fph/tab+disc/VMAX/72/dir_acc_sd\endcsname{0.003}
\expandafter\def\csname mcresult@seed/fph/tab+disc/VMAX/72/mae_mean\endcsname{16.93}
\expandafter\def\csname mcresult@seed/fph/tab+disc/VMAX/72/mae_sd\endcsname{0.67}
\expandafter\def\csname mcresult@seed/fph/tab+disc/VMAX/72/r2_mean\endcsname{0.46}
\expandafter\def\csname mcresult@seed/fph/tab+disc/VMAX/72/r2_sd\endcsname{0.06}
\expandafter\def\csname mcresult@seed/fph/tab+disc/VMAX/72/rmse_mean\endcsname{22.39}
\expandafter\def\csname mcresult@seed/fph/tab+disc/VMAX/72/rmse_sd\endcsname{1.22}
\expandafter\def\csname mcresult@seed/fph/tab+disc/VMAX/72/skill_score_mean\endcsname{0.26}
\expandafter\def\csname mcresult@seed/fph/tab+disc/VMAX/72/skill_score_sd\endcsname{0.04}
\expandafter\def\csname mcresult@seed/fph/tab+disc/VMAX/96/dir_acc_mean\endcsname{0.706}
\expandafter\def\csname mcresult@seed/fph/tab+disc/VMAX/96/dir_acc_sd\endcsname{0.011}
\expandafter\def\csname mcresult@seed/fph/tab+disc/VMAX/96/mae_mean\endcsname{19.76}
\expandafter\def\csname mcresult@seed/fph/tab+disc/VMAX/96/mae_sd\endcsname{1.22}
\expandafter\def\csname mcresult@seed/fph/tab+disc/VMAX/96/r2_mean\endcsname{0.24}
\expandafter\def\csname mcresult@seed/fph/tab+disc/VMAX/96/r2_sd\endcsname{0.07}
\expandafter\def\csname mcresult@seed/fph/tab+disc/VMAX/96/rmse_mean\endcsname{25.56}
\expandafter\def\csname mcresult@seed/fph/tab+disc/VMAX/96/rmse_sd\endcsname{1.22}
\expandafter\def\csname mcresult@seed/fph/tab+disc/VMAX/96/skill_score_mean\endcsname{0.13}
\expandafter\def\csname mcresult@seed/fph/tab+disc/VMAX/96/skill_score_sd\endcsname{0.04}
\expandafter\def\csname mcresult@seed/fph/tab+img/MSLP/120/dir_acc_mean\endcsname{0.743}
\expandafter\def\csname mcresult@seed/fph/tab+img/MSLP/120/dir_acc_sd\endcsname{0.018}
\expandafter\def\csname mcresult@seed/fph/tab+img/MSLP/120/mae_mean\endcsname{16.64}
\expandafter\def\csname mcresult@seed/fph/tab+img/MSLP/120/mae_sd\endcsname{0.75}
\expandafter\def\csname mcresult@seed/fph/tab+img/MSLP/120/r2_mean\endcsname{0.05}
\expandafter\def\csname mcresult@seed/fph/tab+img/MSLP/120/r2_sd\endcsname{0.10}
\expandafter\def\csname mcresult@seed/fph/tab+img/MSLP/120/rmse_mean\endcsname{21.28}
\expandafter\def\csname mcresult@seed/fph/tab+img/MSLP/120/rmse_sd\endcsname{1.08}
\expandafter\def\csname mcresult@seed/fph/tab+img/MSLP/120/skill_score_mean\endcsname{0.03}
\expandafter\def\csname mcresult@seed/fph/tab+img/MSLP/120/skill_score_sd\endcsname{0.05}
\expandafter\def\csname mcresult@seed/fph/tab+img/MSLP/24/dir_acc_mean\endcsname{0.725}
\expandafter\def\csname mcresult@seed/fph/tab+img/MSLP/24/dir_acc_sd\endcsname{0.006}
\expandafter\def\csname mcresult@seed/fph/tab+img/MSLP/24/mae_mean\endcsname{6.57}
\expandafter\def\csname mcresult@seed/fph/tab+img/MSLP/24/mae_sd\endcsname{0.11}
\expandafter\def\csname mcresult@seed/fph/tab+img/MSLP/24/r2_mean\endcsname{0.46}
\expandafter\def\csname mcresult@seed/fph/tab+img/MSLP/24/r2_sd\endcsname{0.03}
\expandafter\def\csname mcresult@seed/fph/tab+img/MSLP/24/rmse_mean\endcsname{9.58}
\expandafter\def\csname mcresult@seed/fph/tab+img/MSLP/24/rmse_sd\endcsname{0.22}
\expandafter\def\csname mcresult@seed/fph/tab+img/MSLP/24/skill_score_mean\endcsname{0.27}
\expandafter\def\csname mcresult@seed/fph/tab+img/MSLP/24/skill_score_sd\endcsname{0.02}
\expandafter\def\csname mcresult@seed/fph/tab+img/MSLP/48/dir_acc_mean\endcsname{0.774}
\expandafter\def\csname mcresult@seed/fph/tab+img/MSLP/48/dir_acc_sd\endcsname{0.004}
\expandafter\def\csname mcresult@seed/fph/tab+img/MSLP/48/mae_mean\endcsname{10.80}
\expandafter\def\csname mcresult@seed/fph/tab+img/MSLP/48/mae_sd\endcsname{0.31}
\expandafter\def\csname mcresult@seed/fph/tab+img/MSLP/48/r2_mean\endcsname{0.42}
\expandafter\def\csname mcresult@seed/fph/tab+img/MSLP/48/r2_sd\endcsname{0.03}
\expandafter\def\csname mcresult@seed/fph/tab+img/MSLP/48/rmse_mean\endcsname{15.01}
\expandafter\def\csname mcresult@seed/fph/tab+img/MSLP/48/rmse_sd\endcsname{0.40}
\expandafter\def\csname mcresult@seed/fph/tab+img/MSLP/48/skill_score_mean\endcsname{0.24}
\expandafter\def\csname mcresult@seed/fph/tab+img/MSLP/48/skill_score_sd\endcsname{0.02}
\expandafter\def\csname mcresult@seed/fph/tab+img/MSLP/72/dir_acc_mean\endcsname{0.736}
\expandafter\def\csname mcresult@seed/fph/tab+img/MSLP/72/dir_acc_sd\endcsname{0.039}
\expandafter\def\csname mcresult@seed/fph/tab+img/MSLP/72/mae_mean\endcsname{13.95}
\expandafter\def\csname mcresult@seed/fph/tab+img/MSLP/72/mae_sd\endcsname{0.42}
\expandafter\def\csname mcresult@seed/fph/tab+img/MSLP/72/r2_mean\endcsname{0.31}
\expandafter\def\csname mcresult@seed/fph/tab+img/MSLP/72/r2_sd\endcsname{0.10}
\expandafter\def\csname mcresult@seed/fph/tab+img/MSLP/72/rmse_mean\endcsname{17.69}
\expandafter\def\csname mcresult@seed/fph/tab+img/MSLP/72/rmse_sd\endcsname{1.21}
\expandafter\def\csname mcresult@seed/fph/tab+img/MSLP/72/skill_score_mean\endcsname{0.17}
\expandafter\def\csname mcresult@seed/fph/tab+img/MSLP/72/skill_score_sd\endcsname{0.06}
\expandafter\def\csname mcresult@seed/fph/tab+img/MSLP/96/dir_acc_mean\endcsname{0.744}
\expandafter\def\csname mcresult@seed/fph/tab+img/MSLP/96/dir_acc_sd\endcsname{0.009}
\expandafter\def\csname mcresult@seed/fph/tab+img/MSLP/96/mae_mean\endcsname{14.82}
\expandafter\def\csname mcresult@seed/fph/tab+img/MSLP/96/mae_sd\endcsname{0.86}
\expandafter\def\csname mcresult@seed/fph/tab+img/MSLP/96/r2_mean\endcsname{0.20}
\expandafter\def\csname mcresult@seed/fph/tab+img/MSLP/96/r2_sd\endcsname{0.06}
\expandafter\def\csname mcresult@seed/fph/tab+img/MSLP/96/rmse_mean\endcsname{18.83}
\expandafter\def\csname mcresult@seed/fph/tab+img/MSLP/96/rmse_sd\endcsname{0.75}
\expandafter\def\csname mcresult@seed/fph/tab+img/MSLP/96/skill_score_mean\endcsname{0.11}
\expandafter\def\csname mcresult@seed/fph/tab+img/MSLP/96/skill_score_sd\endcsname{0.04}
\expandafter\def\csname mcresult@seed/fph/tab+img/VMAX/120/dir_acc_mean\endcsname{0.728}
\expandafter\def\csname mcresult@seed/fph/tab+img/VMAX/120/dir_acc_sd\endcsname{0.021}
\expandafter\def\csname mcresult@seed/fph/tab+img/VMAX/120/mae_mean\endcsname{21.28}
\expandafter\def\csname mcresult@seed/fph/tab+img/VMAX/120/mae_sd\endcsname{0.87}
\expandafter\def\csname mcresult@seed/fph/tab+img/VMAX/120/r2_mean\endcsname{0.08}
\expandafter\def\csname mcresult@seed/fph/tab+img/VMAX/120/r2_sd\endcsname{0.09}
\expandafter\def\csname mcresult@seed/fph/tab+img/VMAX/120/rmse_mean\endcsname{28.02}
\expandafter\def\csname mcresult@seed/fph/tab+img/VMAX/120/rmse_sd\endcsname{1.40}
\expandafter\def\csname mcresult@seed/fph/tab+img/VMAX/120/skill_score_mean\endcsname{0.04}
\expandafter\def\csname mcresult@seed/fph/tab+img/VMAX/120/skill_score_sd\endcsname{0.05}
\expandafter\def\csname mcresult@seed/fph/tab+img/VMAX/24/dir_acc_mean\endcsname{0.685}
\expandafter\def\csname mcresult@seed/fph/tab+img/VMAX/24/dir_acc_sd\endcsname{0.008}
\expandafter\def\csname mcresult@seed/fph/tab+img/VMAX/24/mae_mean\endcsname{10.09}
\expandafter\def\csname mcresult@seed/fph/tab+img/VMAX/24/mae_sd\endcsname{0.11}
\expandafter\def\csname mcresult@seed/fph/tab+img/VMAX/24/r2_mean\endcsname{0.46}
\expandafter\def\csname mcresult@seed/fph/tab+img/VMAX/24/r2_sd\endcsname{0.02}
\expandafter\def\csname mcresult@seed/fph/tab+img/VMAX/24/rmse_mean\endcsname{13.98}
\expandafter\def\csname mcresult@seed/fph/tab+img/VMAX/24/rmse_sd\endcsname{0.30}
\expandafter\def\csname mcresult@seed/fph/tab+img/VMAX/24/skill_score_mean\endcsname{0.27}
\expandafter\def\csname mcresult@seed/fph/tab+img/VMAX/24/skill_score_sd\endcsname{0.02}
\expandafter\def\csname mcresult@seed/fph/tab+img/VMAX/48/dir_acc_mean\endcsname{0.744}
\expandafter\def\csname mcresult@seed/fph/tab+img/VMAX/48/dir_acc_sd\endcsname{0.008}
\expandafter\def\csname mcresult@seed/fph/tab+img/VMAX/48/mae_mean\endcsname{15.49}
\expandafter\def\csname mcresult@seed/fph/tab+img/VMAX/48/mae_sd\endcsname{0.54}
\expandafter\def\csname mcresult@seed/fph/tab+img/VMAX/48/r2_mean\endcsname{0.45}
\expandafter\def\csname mcresult@seed/fph/tab+img/VMAX/48/r2_sd\endcsname{0.02}
\expandafter\def\csname mcresult@seed/fph/tab+img/VMAX/48/rmse_mean\endcsname{21.14}
\expandafter\def\csname mcresult@seed/fph/tab+img/VMAX/48/rmse_sd\endcsname{0.34}
\expandafter\def\csname mcresult@seed/fph/tab+img/VMAX/48/skill_score_mean\endcsname{0.26}
\expandafter\def\csname mcresult@seed/fph/tab+img/VMAX/48/skill_score_sd\endcsname{0.01}
\expandafter\def\csname mcresult@seed/fph/tab+img/VMAX/72/dir_acc_mean\endcsname{0.715}
\expandafter\def\csname mcresult@seed/fph/tab+img/VMAX/72/dir_acc_sd\endcsname{0.019}
\expandafter\def\csname mcresult@seed/fph/tab+img/VMAX/72/mae_mean\endcsname{18.34}
\expandafter\def\csname mcresult@seed/fph/tab+img/VMAX/72/mae_sd\endcsname{0.44}
\expandafter\def\csname mcresult@seed/fph/tab+img/VMAX/72/r2_mean\endcsname{0.41}
\expandafter\def\csname mcresult@seed/fph/tab+img/VMAX/72/r2_sd\endcsname{0.03}
\expandafter\def\csname mcresult@seed/fph/tab+img/VMAX/72/rmse_mean\endcsname{23.37}
\expandafter\def\csname mcresult@seed/fph/tab+img/VMAX/72/rmse_sd\endcsname{0.64}
\expandafter\def\csname mcresult@seed/fph/tab+img/VMAX/72/skill_score_mean\endcsname{0.23}
\expandafter\def\csname mcresult@seed/fph/tab+img/VMAX/72/skill_score_sd\endcsname{0.02}
\expandafter\def\csname mcresult@seed/fph/tab+img/VMAX/96/dir_acc_mean\endcsname{0.729}
\expandafter\def\csname mcresult@seed/fph/tab+img/VMAX/96/dir_acc_sd\endcsname{0.008}
\expandafter\def\csname mcresult@seed/fph/tab+img/VMAX/96/mae_mean\endcsname{19.55}
\expandafter\def\csname mcresult@seed/fph/tab+img/VMAX/96/mae_sd\endcsname{0.95}
\expandafter\def\csname mcresult@seed/fph/tab+img/VMAX/96/r2_mean\endcsname{0.28}
\expandafter\def\csname mcresult@seed/fph/tab+img/VMAX/96/r2_sd\endcsname{0.08}
\expandafter\def\csname mcresult@seed/fph/tab+img/VMAX/96/rmse_mean\endcsname{24.81}
\expandafter\def\csname mcresult@seed/fph/tab+img/VMAX/96/rmse_sd\endcsname{1.38}
\expandafter\def\csname mcresult@seed/fph/tab+img/VMAX/96/skill_score_mean\endcsname{0.15}
\expandafter\def\csname mcresult@seed/fph/tab+img/VMAX/96/skill_score_sd\endcsname{0.05}
\expandafter\def\csname mcresult@seed/fph/tab/MSLP/120/dir_acc_mean\endcsname{0.771}
\expandafter\def\csname mcresult@seed/fph/tab/MSLP/120/dir_acc_sd\endcsname{0.008}
\expandafter\def\csname mcresult@seed/fph/tab/MSLP/120/mae_mean\endcsname{15.91}
\expandafter\def\csname mcresult@seed/fph/tab/MSLP/120/mae_sd\endcsname{0.27}
\expandafter\def\csname mcresult@seed/fph/tab/MSLP/120/r2_mean\endcsname{0.12}
\expandafter\def\csname mcresult@seed/fph/tab/MSLP/120/r2_sd\endcsname{0.05}
\expandafter\def\csname mcresult@seed/fph/tab/MSLP/120/rmse_mean\endcsname{20.55}
\expandafter\def\csname mcresult@seed/fph/tab/MSLP/120/rmse_sd\endcsname{0.59}
\expandafter\def\csname mcresult@seed/fph/tab/MSLP/120/skill_score_mean\endcsname{0.06}
\expandafter\def\csname mcresult@seed/fph/tab/MSLP/120/skill_score_sd\endcsname{0.03}
\expandafter\def\csname mcresult@seed/fph/tab/MSLP/24/dir_acc_mean\endcsname{0.728}
\expandafter\def\csname mcresult@seed/fph/tab/MSLP/24/dir_acc_sd\endcsname{0.006}
\expandafter\def\csname mcresult@seed/fph/tab/MSLP/24/mae_mean\endcsname{6.59}
\expandafter\def\csname mcresult@seed/fph/tab/MSLP/24/mae_sd\endcsname{0.08}
\expandafter\def\csname mcresult@seed/fph/tab/MSLP/24/r2_mean\endcsname{0.44}
\expandafter\def\csname mcresult@seed/fph/tab/MSLP/24/r2_sd\endcsname{0.01}
\expandafter\def\csname mcresult@seed/fph/tab/MSLP/24/rmse_mean\endcsname{9.74}
\expandafter\def\csname mcresult@seed/fph/tab/MSLP/24/rmse_sd\endcsname{0.09}
\expandafter\def\csname mcresult@seed/fph/tab/MSLP/24/skill_score_mean\endcsname{0.25}
\expandafter\def\csname mcresult@seed/fph/tab/MSLP/24/skill_score_sd\endcsname{0.01}
\expandafter\def\csname mcresult@seed/fph/tab/MSLP/48/dir_acc_mean\endcsname{0.778}
\expandafter\def\csname mcresult@seed/fph/tab/MSLP/48/dir_acc_sd\endcsname{0.008}
\expandafter\def\csname mcresult@seed/fph/tab/MSLP/48/mae_mean\endcsname{10.77}
\expandafter\def\csname mcresult@seed/fph/tab/MSLP/48/mae_sd\endcsname{0.20}
\expandafter\def\csname mcresult@seed/fph/tab/MSLP/48/r2_mean\endcsname{0.44}
\expandafter\def\csname mcresult@seed/fph/tab/MSLP/48/r2_sd\endcsname{0.02}
\expandafter\def\csname mcresult@seed/fph/tab/MSLP/48/rmse_mean\endcsname{14.80}
\expandafter\def\csname mcresult@seed/fph/tab/MSLP/48/rmse_sd\endcsname{0.24}
\expandafter\def\csname mcresult@seed/fph/tab/MSLP/48/skill_score_mean\endcsname{0.25}
\expandafter\def\csname mcresult@seed/fph/tab/MSLP/48/skill_score_sd\endcsname{0.01}
\expandafter\def\csname mcresult@seed/fph/tab/MSLP/72/dir_acc_mean\endcsname{0.748}
\expandafter\def\csname mcresult@seed/fph/tab/MSLP/72/dir_acc_sd\endcsname{0.013}
\expandafter\def\csname mcresult@seed/fph/tab/MSLP/72/mae_mean\endcsname{13.82}
\expandafter\def\csname mcresult@seed/fph/tab/MSLP/72/mae_sd\endcsname{0.42}
\expandafter\def\csname mcresult@seed/fph/tab/MSLP/72/r2_mean\endcsname{0.33}
\expandafter\def\csname mcresult@seed/fph/tab/MSLP/72/r2_sd\endcsname{0.03}
\expandafter\def\csname mcresult@seed/fph/tab/MSLP/72/rmse_mean\endcsname{17.52}
\expandafter\def\csname mcresult@seed/fph/tab/MSLP/72/rmse_sd\endcsname{0.39}
\expandafter\def\csname mcresult@seed/fph/tab/MSLP/72/skill_score_mean\endcsname{0.18}
\expandafter\def\csname mcresult@seed/fph/tab/MSLP/72/skill_score_sd\endcsname{0.02}
\expandafter\def\csname mcresult@seed/fph/tab/MSLP/96/dir_acc_mean\endcsname{0.733}
\expandafter\def\csname mcresult@seed/fph/tab/MSLP/96/dir_acc_sd\endcsname{0.006}
\expandafter\def\csname mcresult@seed/fph/tab/MSLP/96/mae_mean\endcsname{15.03}
\expandafter\def\csname mcresult@seed/fph/tab/MSLP/96/mae_sd\endcsname{0.41}
\expandafter\def\csname mcresult@seed/fph/tab/MSLP/96/r2_mean\endcsname{0.18}
\expandafter\def\csname mcresult@seed/fph/tab/MSLP/96/r2_sd\endcsname{0.05}
\expandafter\def\csname mcresult@seed/fph/tab/MSLP/96/rmse_mean\endcsname{19.09}
\expandafter\def\csname mcresult@seed/fph/tab/MSLP/96/rmse_sd\endcsname{0.57}
\expandafter\def\csname mcresult@seed/fph/tab/MSLP/96/skill_score_mean\endcsname{0.10}
\expandafter\def\csname mcresult@seed/fph/tab/MSLP/96/skill_score_sd\endcsname{0.03}
\expandafter\def\csname mcresult@seed/fph/tab/VMAX/120/dir_acc_mean\endcsname{0.756}
\expandafter\def\csname mcresult@seed/fph/tab/VMAX/120/dir_acc_sd\endcsname{0.012}
\expandafter\def\csname mcresult@seed/fph/tab/VMAX/120/mae_mean\endcsname{20.17}
\expandafter\def\csname mcresult@seed/fph/tab/VMAX/120/mae_sd\endcsname{0.09}
\expandafter\def\csname mcresult@seed/fph/tab/VMAX/120/r2_mean\endcsname{0.21}
\expandafter\def\csname mcresult@seed/fph/tab/VMAX/120/r2_sd\endcsname{0.03}
\expandafter\def\csname mcresult@seed/fph/tab/VMAX/120/rmse_mean\endcsname{26.11}
\expandafter\def\csname mcresult@seed/fph/tab/VMAX/120/rmse_sd\endcsname{0.43}
\expandafter\def\csname mcresult@seed/fph/tab/VMAX/120/skill_score_mean\endcsname{0.11}
\expandafter\def\csname mcresult@seed/fph/tab/VMAX/120/skill_score_sd\endcsname{0.01}
\expandafter\def\csname mcresult@seed/fph/tab/VMAX/24/dir_acc_mean\endcsname{0.685}
\expandafter\def\csname mcresult@seed/fph/tab/VMAX/24/dir_acc_sd\endcsname{0.005}
\expandafter\def\csname mcresult@seed/fph/tab/VMAX/24/mae_mean\endcsname{10.11}
\expandafter\def\csname mcresult@seed/fph/tab/VMAX/24/mae_sd\endcsname{0.06}
\expandafter\def\csname mcresult@seed/fph/tab/VMAX/24/r2_mean\endcsname{0.44}
\expandafter\def\csname mcresult@seed/fph/tab/VMAX/24/r2_sd\endcsname{0.01}
\expandafter\def\csname mcresult@seed/fph/tab/VMAX/24/rmse_mean\endcsname{14.22}
\expandafter\def\csname mcresult@seed/fph/tab/VMAX/24/rmse_sd\endcsname{0.12}
\expandafter\def\csname mcresult@seed/fph/tab/VMAX/24/skill_score_mean\endcsname{0.25}
\expandafter\def\csname mcresult@seed/fph/tab/VMAX/24/skill_score_sd\endcsname{0.01}
\expandafter\def\csname mcresult@seed/fph/tab/VMAX/48/dir_acc_mean\endcsname{0.753}
\expandafter\def\csname mcresult@seed/fph/tab/VMAX/48/dir_acc_sd\endcsname{0.003}
\expandafter\def\csname mcresult@seed/fph/tab/VMAX/48/mae_mean\endcsname{15.32}
\expandafter\def\csname mcresult@seed/fph/tab/VMAX/48/mae_sd\endcsname{0.38}
\expandafter\def\csname mcresult@seed/fph/tab/VMAX/48/r2_mean\endcsname{0.47}
\expandafter\def\csname mcresult@seed/fph/tab/VMAX/48/r2_sd\endcsname{0.02}
\expandafter\def\csname mcresult@seed/fph/tab/VMAX/48/rmse_mean\endcsname{20.78}
\expandafter\def\csname mcresult@seed/fph/tab/VMAX/48/rmse_sd\endcsname{0.38}
\expandafter\def\csname mcresult@seed/fph/tab/VMAX/48/skill_score_mean\endcsname{0.27}
\expandafter\def\csname mcresult@seed/fph/tab/VMAX/48/skill_score_sd\endcsname{0.01}
\expandafter\def\csname mcresult@seed/fph/tab/VMAX/72/dir_acc_mean\endcsname{0.714}
\expandafter\def\csname mcresult@seed/fph/tab/VMAX/72/dir_acc_sd\endcsname{0.010}
\expandafter\def\csname mcresult@seed/fph/tab/VMAX/72/mae_mean\endcsname{18.77}
\expandafter\def\csname mcresult@seed/fph/tab/VMAX/72/mae_sd\endcsname{0.58}
\expandafter\def\csname mcresult@seed/fph/tab/VMAX/72/r2_mean\endcsname{0.39}
\expandafter\def\csname mcresult@seed/fph/tab/VMAX/72/r2_sd\endcsname{0.02}
\expandafter\def\csname mcresult@seed/fph/tab/VMAX/72/rmse_mean\endcsname{23.78}
\expandafter\def\csname mcresult@seed/fph/tab/VMAX/72/rmse_sd\endcsname{0.47}
\expandafter\def\csname mcresult@seed/fph/tab/VMAX/72/skill_score_mean\endcsname{0.22}
\expandafter\def\csname mcresult@seed/fph/tab/VMAX/72/skill_score_sd\endcsname{0.02}
\expandafter\def\csname mcresult@seed/fph/tab/VMAX/96/dir_acc_mean\endcsname{0.722}
\expandafter\def\csname mcresult@seed/fph/tab/VMAX/96/dir_acc_sd\endcsname{0.009}
\expandafter\def\csname mcresult@seed/fph/tab/VMAX/96/mae_mean\endcsname{19.81}
\expandafter\def\csname mcresult@seed/fph/tab/VMAX/96/mae_sd\endcsname{0.42}
\expandafter\def\csname mcresult@seed/fph/tab/VMAX/96/r2_mean\endcsname{0.27}
\expandafter\def\csname mcresult@seed/fph/tab/VMAX/96/r2_sd\endcsname{0.03}
\expandafter\def\csname mcresult@seed/fph/tab/VMAX/96/rmse_mean\endcsname{24.97}
\expandafter\def\csname mcresult@seed/fph/tab/VMAX/96/rmse_sd\endcsname{0.59}
\expandafter\def\csname mcresult@seed/fph/tab/VMAX/96/skill_score_mean\endcsname{0.15}
\expandafter\def\csname mcresult@seed/fph/tab/VMAX/96/skill_score_sd\endcsname{0.02}
\expandafter\def\csname mcresult@seed/mhj/disc/MSLP/120/dir_acc_mean\endcsname{0.743}
\expandafter\def\csname mcresult@seed/mhj/disc/MSLP/120/dir_acc_sd\endcsname{0.014}
\expandafter\def\csname mcresult@seed/mhj/disc/MSLP/120/mae_mean\endcsname{15.73}
\expandafter\def\csname mcresult@seed/mhj/disc/MSLP/120/mae_sd\endcsname{0.64}
\expandafter\def\csname mcresult@seed/mhj/disc/MSLP/120/r2_mean\endcsname{0.04}
\expandafter\def\csname mcresult@seed/mhj/disc/MSLP/120/r2_sd\endcsname{0.10}
\expandafter\def\csname mcresult@seed/mhj/disc/MSLP/120/rmse_mean\endcsname{21.61}
\expandafter\def\csname mcresult@seed/mhj/disc/MSLP/120/rmse_sd\endcsname{1.08}
\expandafter\def\csname mcresult@seed/mhj/disc/MSLP/120/skill_score_mean\endcsname{0.02}
\expandafter\def\csname mcresult@seed/mhj/disc/MSLP/120/skill_score_sd\endcsname{0.05}
\expandafter\def\csname mcresult@seed/mhj/disc/MSLP/24/dir_acc_mean\endcsname{0.751}
\expandafter\def\csname mcresult@seed/mhj/disc/MSLP/24/dir_acc_sd\endcsname{0.009}
\expandafter\def\csname mcresult@seed/mhj/disc/MSLP/24/mae_mean\endcsname{6.49}
\expandafter\def\csname mcresult@seed/mhj/disc/MSLP/24/mae_sd\endcsname{0.21}
\expandafter\def\csname mcresult@seed/mhj/disc/MSLP/24/r2_mean\endcsname{0.47}
\expandafter\def\csname mcresult@seed/mhj/disc/MSLP/24/r2_sd\endcsname{0.02}
\expandafter\def\csname mcresult@seed/mhj/disc/MSLP/24/rmse_mean\endcsname{9.62}
\expandafter\def\csname mcresult@seed/mhj/disc/MSLP/24/rmse_sd\endcsname{0.14}
\expandafter\def\csname mcresult@seed/mhj/disc/MSLP/24/skill_score_mean\endcsname{0.27}
\expandafter\def\csname mcresult@seed/mhj/disc/MSLP/24/skill_score_sd\endcsname{0.01}
\expandafter\def\csname mcresult@seed/mhj/disc/MSLP/48/dir_acc_mean\endcsname{0.765}
\expandafter\def\csname mcresult@seed/mhj/disc/MSLP/48/dir_acc_sd\endcsname{0.010}
\expandafter\def\csname mcresult@seed/mhj/disc/MSLP/48/mae_mean\endcsname{10.16}
\expandafter\def\csname mcresult@seed/mhj/disc/MSLP/48/mae_sd\endcsname{0.21}
\expandafter\def\csname mcresult@seed/mhj/disc/MSLP/48/r2_mean\endcsname{0.49}
\expandafter\def\csname mcresult@seed/mhj/disc/MSLP/48/r2_sd\endcsname{0.02}
\expandafter\def\csname mcresult@seed/mhj/disc/MSLP/48/rmse_mean\endcsname{14.15}
\expandafter\def\csname mcresult@seed/mhj/disc/MSLP/48/rmse_sd\endcsname{0.28}
\expandafter\def\csname mcresult@seed/mhj/disc/MSLP/48/skill_score_mean\endcsname{0.28}
\expandafter\def\csname mcresult@seed/mhj/disc/MSLP/48/skill_score_sd\endcsname{0.01}
\expandafter\def\csname mcresult@seed/mhj/disc/MSLP/72/dir_acc_mean\endcsname{0.734}
\expandafter\def\csname mcresult@seed/mhj/disc/MSLP/72/dir_acc_sd\endcsname{0.013}
\expandafter\def\csname mcresult@seed/mhj/disc/MSLP/72/mae_mean\endcsname{12.55}
\expandafter\def\csname mcresult@seed/mhj/disc/MSLP/72/mae_sd\endcsname{0.27}
\expandafter\def\csname mcresult@seed/mhj/disc/MSLP/72/r2_mean\endcsname{0.37}
\expandafter\def\csname mcresult@seed/mhj/disc/MSLP/72/r2_sd\endcsname{0.03}
\expandafter\def\csname mcresult@seed/mhj/disc/MSLP/72/rmse_mean\endcsname{17.11}
\expandafter\def\csname mcresult@seed/mhj/disc/MSLP/72/rmse_sd\endcsname{0.42}
\expandafter\def\csname mcresult@seed/mhj/disc/MSLP/72/skill_score_mean\endcsname{0.21}
\expandafter\def\csname mcresult@seed/mhj/disc/MSLP/72/skill_score_sd\endcsname{0.02}
\expandafter\def\csname mcresult@seed/mhj/disc/MSLP/96/dir_acc_mean\endcsname{0.721}
\expandafter\def\csname mcresult@seed/mhj/disc/MSLP/96/dir_acc_sd\endcsname{0.014}
\expandafter\def\csname mcresult@seed/mhj/disc/MSLP/96/mae_mean\endcsname{14.12}
\expandafter\def\csname mcresult@seed/mhj/disc/MSLP/96/mae_sd\endcsname{0.52}
\expandafter\def\csname mcresult@seed/mhj/disc/MSLP/96/r2_mean\endcsname{0.18}
\expandafter\def\csname mcresult@seed/mhj/disc/MSLP/96/r2_sd\endcsname{0.07}
\expandafter\def\csname mcresult@seed/mhj/disc/MSLP/96/rmse_mean\endcsname{19.26}
\expandafter\def\csname mcresult@seed/mhj/disc/MSLP/96/rmse_sd\endcsname{0.80}
\expandafter\def\csname mcresult@seed/mhj/disc/MSLP/96/skill_score_mean\endcsname{0.10}
\expandafter\def\csname mcresult@seed/mhj/disc/MSLP/96/skill_score_sd\endcsname{0.04}
\expandafter\def\csname mcresult@seed/mhj/disc/VMAX/120/dir_acc_mean\endcsname{0.723}
\expandafter\def\csname mcresult@seed/mhj/disc/VMAX/120/dir_acc_sd\endcsname{0.012}
\expandafter\def\csname mcresult@seed/mhj/disc/VMAX/120/mae_mean\endcsname{20.67}
\expandafter\def\csname mcresult@seed/mhj/disc/VMAX/120/mae_sd\endcsname{0.43}
\expandafter\def\csname mcresult@seed/mhj/disc/VMAX/120/r2_mean\endcsname{0.10}
\expandafter\def\csname mcresult@seed/mhj/disc/VMAX/120/r2_sd\endcsname{0.05}
\expandafter\def\csname mcresult@seed/mhj/disc/VMAX/120/rmse_mean\endcsname{27.85}
\expandafter\def\csname mcresult@seed/mhj/disc/VMAX/120/rmse_sd\endcsname{0.78}
\expandafter\def\csname mcresult@seed/mhj/disc/VMAX/120/skill_score_mean\endcsname{0.05}
\expandafter\def\csname mcresult@seed/mhj/disc/VMAX/120/skill_score_sd\endcsname{0.03}
\expandafter\def\csname mcresult@seed/mhj/disc/VMAX/24/dir_acc_mean\endcsname{0.715}
\expandafter\def\csname mcresult@seed/mhj/disc/VMAX/24/dir_acc_sd\endcsname{0.006}
\expandafter\def\csname mcresult@seed/mhj/disc/VMAX/24/mae_mean\endcsname{9.78}
\expandafter\def\csname mcresult@seed/mhj/disc/VMAX/24/mae_sd\endcsname{0.25}
\expandafter\def\csname mcresult@seed/mhj/disc/VMAX/24/r2_mean\endcsname{0.51}
\expandafter\def\csname mcresult@seed/mhj/disc/VMAX/24/r2_sd\endcsname{0.02}
\expandafter\def\csname mcresult@seed/mhj/disc/VMAX/24/rmse_mean\endcsname{13.53}
\expandafter\def\csname mcresult@seed/mhj/disc/VMAX/24/rmse_sd\endcsname{0.22}
\expandafter\def\csname mcresult@seed/mhj/disc/VMAX/24/skill_score_mean\endcsname{0.30}
\expandafter\def\csname mcresult@seed/mhj/disc/VMAX/24/skill_score_sd\endcsname{0.01}
\expandafter\def\csname mcresult@seed/mhj/disc/VMAX/48/dir_acc_mean\endcsname{0.747}
\expandafter\def\csname mcresult@seed/mhj/disc/VMAX/48/dir_acc_sd\endcsname{0.004}
\expandafter\def\csname mcresult@seed/mhj/disc/VMAX/48/mae_mean\endcsname{14.35}
\expandafter\def\csname mcresult@seed/mhj/disc/VMAX/48/mae_sd\endcsname{0.27}
\expandafter\def\csname mcresult@seed/mhj/disc/VMAX/48/r2_mean\endcsname{0.53}
\expandafter\def\csname mcresult@seed/mhj/disc/VMAX/48/r2_sd\endcsname{0.02}
\expandafter\def\csname mcresult@seed/mhj/disc/VMAX/48/rmse_mean\endcsname{19.55}
\expandafter\def\csname mcresult@seed/mhj/disc/VMAX/48/rmse_sd\endcsname{0.45}
\expandafter\def\csname mcresult@seed/mhj/disc/VMAX/48/skill_score_mean\endcsname{0.31}
\expandafter\def\csname mcresult@seed/mhj/disc/VMAX/48/skill_score_sd\endcsname{0.02}
\expandafter\def\csname mcresult@seed/mhj/disc/VMAX/72/dir_acc_mean\endcsname{0.724}
\expandafter\def\csname mcresult@seed/mhj/disc/VMAX/72/dir_acc_sd\endcsname{0.006}
\expandafter\def\csname mcresult@seed/mhj/disc/VMAX/72/mae_mean\endcsname{17.19}
\expandafter\def\csname mcresult@seed/mhj/disc/VMAX/72/mae_sd\endcsname{0.13}
\expandafter\def\csname mcresult@seed/mhj/disc/VMAX/72/r2_mean\endcsname{0.43}
\expandafter\def\csname mcresult@seed/mhj/disc/VMAX/72/r2_sd\endcsname{0.02}
\expandafter\def\csname mcresult@seed/mhj/disc/VMAX/72/rmse_mean\endcsname{23.03}
\expandafter\def\csname mcresult@seed/mhj/disc/VMAX/72/rmse_sd\endcsname{0.36}
\expandafter\def\csname mcresult@seed/mhj/disc/VMAX/72/skill_score_mean\endcsname{0.24}
\expandafter\def\csname mcresult@seed/mhj/disc/VMAX/72/skill_score_sd\endcsname{0.01}
\expandafter\def\csname mcresult@seed/mhj/disc/VMAX/96/dir_acc_mean\endcsname{0.705}
\expandafter\def\csname mcresult@seed/mhj/disc/VMAX/96/dir_acc_sd\endcsname{0.012}
\expandafter\def\csname mcresult@seed/mhj/disc/VMAX/96/mae_mean\endcsname{19.39}
\expandafter\def\csname mcresult@seed/mhj/disc/VMAX/96/mae_sd\endcsname{0.45}
\expandafter\def\csname mcresult@seed/mhj/disc/VMAX/96/r2_mean\endcsname{0.24}
\expandafter\def\csname mcresult@seed/mhj/disc/VMAX/96/r2_sd\endcsname{0.03}
\expandafter\def\csname mcresult@seed/mhj/disc/VMAX/96/rmse_mean\endcsname{25.53}
\expandafter\def\csname mcresult@seed/mhj/disc/VMAX/96/rmse_sd\endcsname{0.59}
\expandafter\def\csname mcresult@seed/mhj/disc/VMAX/96/skill_score_mean\endcsname{0.13}
\expandafter\def\csname mcresult@seed/mhj/disc/VMAX/96/skill_score_sd\endcsname{0.02}
\expandafter\def\csname mcresult@seed/mhj/full/MSLP/120/dir_acc_mean\endcsname{0.766}
\expandafter\def\csname mcresult@seed/mhj/full/MSLP/120/dir_acc_sd\endcsname{0.014}
\expandafter\def\csname mcresult@seed/mhj/full/MSLP/120/mae_mean\endcsname{15.59}
\expandafter\def\csname mcresult@seed/mhj/full/MSLP/120/mae_sd\endcsname{0.98}
\expandafter\def\csname mcresult@seed/mhj/full/MSLP/120/r2_mean\endcsname{0.06}
\expandafter\def\csname mcresult@seed/mhj/full/MSLP/120/r2_sd\endcsname{0.15}
\expandafter\def\csname mcresult@seed/mhj/full/MSLP/120/rmse_mean\endcsname{21.17}
\expandafter\def\csname mcresult@seed/mhj/full/MSLP/120/rmse_sd\endcsname{1.71}
\expandafter\def\csname mcresult@seed/mhj/full/MSLP/120/skill_score_mean\endcsname{0.03}
\expandafter\def\csname mcresult@seed/mhj/full/MSLP/120/skill_score_sd\endcsname{0.08}
\expandafter\def\csname mcresult@seed/mhj/full/MSLP/24/dir_acc_mean\endcsname{0.757}
\expandafter\def\csname mcresult@seed/mhj/full/MSLP/24/dir_acc_sd\endcsname{0.010}
\expandafter\def\csname mcresult@seed/mhj/full/MSLP/24/mae_mean\endcsname{6.11}
\expandafter\def\csname mcresult@seed/mhj/full/MSLP/24/mae_sd\endcsname{0.25}
\expandafter\def\csname mcresult@seed/mhj/full/MSLP/24/r2_mean\endcsname{0.52}
\expandafter\def\csname mcresult@seed/mhj/full/MSLP/24/r2_sd\endcsname{0.04}
\expandafter\def\csname mcresult@seed/mhj/full/MSLP/24/rmse_mean\endcsname{9.04}
\expandafter\def\csname mcresult@seed/mhj/full/MSLP/24/rmse_sd\endcsname{0.37}
\expandafter\def\csname mcresult@seed/mhj/full/MSLP/24/skill_score_mean\endcsname{0.31}
\expandafter\def\csname mcresult@seed/mhj/full/MSLP/24/skill_score_sd\endcsname{0.03}
\expandafter\def\csname mcresult@seed/mhj/full/MSLP/48/dir_acc_mean\endcsname{0.796}
\expandafter\def\csname mcresult@seed/mhj/full/MSLP/48/dir_acc_sd\endcsname{0.010}
\expandafter\def\csname mcresult@seed/mhj/full/MSLP/48/mae_mean\endcsname{9.73}
\expandafter\def\csname mcresult@seed/mhj/full/MSLP/48/mae_sd\endcsname{0.34}
\expandafter\def\csname mcresult@seed/mhj/full/MSLP/48/r2_mean\endcsname{0.52}
\expandafter\def\csname mcresult@seed/mhj/full/MSLP/48/r2_sd\endcsname{0.03}
\expandafter\def\csname mcresult@seed/mhj/full/MSLP/48/rmse_mean\endcsname{13.71}
\expandafter\def\csname mcresult@seed/mhj/full/MSLP/48/rmse_sd\endcsname{0.45}
\expandafter\def\csname mcresult@seed/mhj/full/MSLP/48/skill_score_mean\endcsname{0.30}
\expandafter\def\csname mcresult@seed/mhj/full/MSLP/48/skill_score_sd\endcsname{0.02}
\expandafter\def\csname mcresult@seed/mhj/full/MSLP/72/dir_acc_mean\endcsname{0.768}
\expandafter\def\csname mcresult@seed/mhj/full/MSLP/72/dir_acc_sd\endcsname{0.021}
\expandafter\def\csname mcresult@seed/mhj/full/MSLP/72/mae_mean\endcsname{12.19}
\expandafter\def\csname mcresult@seed/mhj/full/MSLP/72/mae_sd\endcsname{0.58}
\expandafter\def\csname mcresult@seed/mhj/full/MSLP/72/r2_mean\endcsname{0.38}
\expandafter\def\csname mcresult@seed/mhj/full/MSLP/72/r2_sd\endcsname{0.06}
\expandafter\def\csname mcresult@seed/mhj/full/MSLP/72/rmse_mean\endcsname{16.84}
\expandafter\def\csname mcresult@seed/mhj/full/MSLP/72/rmse_sd\endcsname{0.84}
\expandafter\def\csname mcresult@seed/mhj/full/MSLP/72/skill_score_mean\endcsname{0.21}
\expandafter\def\csname mcresult@seed/mhj/full/MSLP/72/skill_score_sd\endcsname{0.04}
\expandafter\def\csname mcresult@seed/mhj/full/MSLP/96/dir_acc_mean\endcsname{0.747}
\expandafter\def\csname mcresult@seed/mhj/full/MSLP/96/dir_acc_sd\endcsname{0.020}
\expandafter\def\csname mcresult@seed/mhj/full/MSLP/96/mae_mean\endcsname{14.28}
\expandafter\def\csname mcresult@seed/mhj/full/MSLP/96/mae_sd\endcsname{0.99}
\expandafter\def\csname mcresult@seed/mhj/full/MSLP/96/r2_mean\endcsname{0.16}
\expandafter\def\csname mcresult@seed/mhj/full/MSLP/96/r2_sd\endcsname{0.11}
\expandafter\def\csname mcresult@seed/mhj/full/MSLP/96/rmse_mean\endcsname{19.38}
\expandafter\def\csname mcresult@seed/mhj/full/MSLP/96/rmse_sd\endcsname{1.30}
\expandafter\def\csname mcresult@seed/mhj/full/MSLP/96/skill_score_mean\endcsname{0.08}
\expandafter\def\csname mcresult@seed/mhj/full/MSLP/96/skill_score_sd\endcsname{0.06}
\expandafter\def\csname mcresult@seed/mhj/full/VMAX/120/dir_acc_mean\endcsname{0.748}
\expandafter\def\csname mcresult@seed/mhj/full/VMAX/120/dir_acc_sd\endcsname{0.016}
\expandafter\def\csname mcresult@seed/mhj/full/VMAX/120/mae_mean\endcsname{20.61}
\expandafter\def\csname mcresult@seed/mhj/full/VMAX/120/mae_sd\endcsname{1.17}
\expandafter\def\csname mcresult@seed/mhj/full/VMAX/120/r2_mean\endcsname{0.09}
\expandafter\def\csname mcresult@seed/mhj/full/VMAX/120/r2_sd\endcsname{0.13}
\expandafter\def\csname mcresult@seed/mhj/full/VMAX/120/rmse_mean\endcsname{27.90}
\expandafter\def\csname mcresult@seed/mhj/full/VMAX/120/rmse_sd\endcsname{2.05}
\expandafter\def\csname mcresult@seed/mhj/full/VMAX/120/skill_score_mean\endcsname{0.05}
\expandafter\def\csname mcresult@seed/mhj/full/VMAX/120/skill_score_sd\endcsname{0.07}
\expandafter\def\csname mcresult@seed/mhj/full/VMAX/24/dir_acc_mean\endcsname{0.728}
\expandafter\def\csname mcresult@seed/mhj/full/VMAX/24/dir_acc_sd\endcsname{0.003}
\expandafter\def\csname mcresult@seed/mhj/full/VMAX/24/mae_mean\endcsname{9.24}
\expandafter\def\csname mcresult@seed/mhj/full/VMAX/24/mae_sd\endcsname{0.22}
\expandafter\def\csname mcresult@seed/mhj/full/VMAX/24/r2_mean\endcsname{0.54}
\expandafter\def\csname mcresult@seed/mhj/full/VMAX/24/r2_sd\endcsname{0.03}
\expandafter\def\csname mcresult@seed/mhj/full/VMAX/24/rmse_mean\endcsname{12.87}
\expandafter\def\csname mcresult@seed/mhj/full/VMAX/24/rmse_sd\endcsname{0.44}
\expandafter\def\csname mcresult@seed/mhj/full/VMAX/24/skill_score_mean\endcsname{0.32}
\expandafter\def\csname mcresult@seed/mhj/full/VMAX/24/skill_score_sd\endcsname{0.02}
\expandafter\def\csname mcresult@seed/mhj/full/VMAX/48/dir_acc_mean\endcsname{0.772}
\expandafter\def\csname mcresult@seed/mhj/full/VMAX/48/dir_acc_sd\endcsname{0.010}
\expandafter\def\csname mcresult@seed/mhj/full/VMAX/48/mae_mean\endcsname{13.66}
\expandafter\def\csname mcresult@seed/mhj/full/VMAX/48/mae_sd\endcsname{0.48}
\expandafter\def\csname mcresult@seed/mhj/full/VMAX/48/r2_mean\endcsname{0.56}
\expandafter\def\csname mcresult@seed/mhj/full/VMAX/48/r2_sd\endcsname{0.03}
\expandafter\def\csname mcresult@seed/mhj/full/VMAX/48/rmse_mean\endcsname{18.99}
\expandafter\def\csname mcresult@seed/mhj/full/VMAX/48/rmse_sd\endcsname{0.54}
\expandafter\def\csname mcresult@seed/mhj/full/VMAX/48/skill_score_mean\endcsname{0.33}
\expandafter\def\csname mcresult@seed/mhj/full/VMAX/48/skill_score_sd\endcsname{0.02}
\expandafter\def\csname mcresult@seed/mhj/full/VMAX/72/dir_acc_mean\endcsname{0.734}
\expandafter\def\csname mcresult@seed/mhj/full/VMAX/72/dir_acc_sd\endcsname{0.017}
\expandafter\def\csname mcresult@seed/mhj/full/VMAX/72/mae_mean\endcsname{16.88}
\expandafter\def\csname mcresult@seed/mhj/full/VMAX/72/mae_sd\endcsname{0.76}
\expandafter\def\csname mcresult@seed/mhj/full/VMAX/72/r2_mean\endcsname{0.41}
\expandafter\def\csname mcresult@seed/mhj/full/VMAX/72/r2_sd\endcsname{0.05}
\expandafter\def\csname mcresult@seed/mhj/full/VMAX/72/rmse_mean\endcsname{23.28}
\expandafter\def\csname mcresult@seed/mhj/full/VMAX/72/rmse_sd\endcsname{1.01}
\expandafter\def\csname mcresult@seed/mhj/full/VMAX/72/skill_score_mean\endcsname{0.23}
\expandafter\def\csname mcresult@seed/mhj/full/VMAX/72/skill_score_sd\endcsname{0.03}
\expandafter\def\csname mcresult@seed/mhj/full/VMAX/96/dir_acc_mean\endcsname{0.730}
\expandafter\def\csname mcresult@seed/mhj/full/VMAX/96/dir_acc_sd\endcsname{0.019}
\expandafter\def\csname mcresult@seed/mhj/full/VMAX/96/mae_mean\endcsname{19.58}
\expandafter\def\csname mcresult@seed/mhj/full/VMAX/96/mae_sd\endcsname{1.34}
\expandafter\def\csname mcresult@seed/mhj/full/VMAX/96/r2_mean\endcsname{0.19}
\expandafter\def\csname mcresult@seed/mhj/full/VMAX/96/r2_sd\endcsname{0.10}
\expandafter\def\csname mcresult@seed/mhj/full/VMAX/96/rmse_mean\endcsname{26.33}
\expandafter\def\csname mcresult@seed/mhj/full/VMAX/96/rmse_sd\endcsname{1.58}
\expandafter\def\csname mcresult@seed/mhj/full/VMAX/96/skill_score_mean\endcsname{0.10}
\expandafter\def\csname mcresult@seed/mhj/full/VMAX/96/skill_score_sd\endcsname{0.05}
\expandafter\def\csname mcresult@seed/mhj/img+disc/MSLP/120/dir_acc_mean\endcsname{0.739}
\expandafter\def\csname mcresult@seed/mhj/img+disc/MSLP/120/dir_acc_sd\endcsname{0.010}
\expandafter\def\csname mcresult@seed/mhj/img+disc/MSLP/120/mae_mean\endcsname{16.12}
\expandafter\def\csname mcresult@seed/mhj/img+disc/MSLP/120/mae_sd\endcsname{0.71}
\expandafter\def\csname mcresult@seed/mhj/img+disc/MSLP/120/r2_mean\endcsname{0.01}
\expandafter\def\csname mcresult@seed/mhj/img+disc/MSLP/120/r2_sd\endcsname{0.10}
\expandafter\def\csname mcresult@seed/mhj/img+disc/MSLP/120/rmse_mean\endcsname{21.74}
\expandafter\def\csname mcresult@seed/mhj/img+disc/MSLP/120/rmse_sd\endcsname{1.15}
\expandafter\def\csname mcresult@seed/mhj/img+disc/MSLP/120/skill_score_mean\endcsname{0.01}
\expandafter\def\csname mcresult@seed/mhj/img+disc/MSLP/120/skill_score_sd\endcsname{0.05}
\expandafter\def\csname mcresult@seed/mhj/img+disc/MSLP/24/dir_acc_mean\endcsname{0.736}
\expandafter\def\csname mcresult@seed/mhj/img+disc/MSLP/24/dir_acc_sd\endcsname{0.015}
\expandafter\def\csname mcresult@seed/mhj/img+disc/MSLP/24/mae_mean\endcsname{6.43}
\expandafter\def\csname mcresult@seed/mhj/img+disc/MSLP/24/mae_sd\endcsname{0.16}
\expandafter\def\csname mcresult@seed/mhj/img+disc/MSLP/24/r2_mean\endcsname{0.48}
\expandafter\def\csname mcresult@seed/mhj/img+disc/MSLP/24/r2_sd\endcsname{0.01}
\expandafter\def\csname mcresult@seed/mhj/img+disc/MSLP/24/rmse_mean\endcsname{9.41}
\expandafter\def\csname mcresult@seed/mhj/img+disc/MSLP/24/rmse_sd\endcsname{0.13}
\expandafter\def\csname mcresult@seed/mhj/img+disc/MSLP/24/skill_score_mean\endcsname{0.28}
\expandafter\def\csname mcresult@seed/mhj/img+disc/MSLP/24/skill_score_sd\endcsname{0.01}
\expandafter\def\csname mcresult@seed/mhj/img+disc/MSLP/48/dir_acc_mean\endcsname{0.768}
\expandafter\def\csname mcresult@seed/mhj/img+disc/MSLP/48/dir_acc_sd\endcsname{0.008}
\expandafter\def\csname mcresult@seed/mhj/img+disc/MSLP/48/mae_mean\endcsname{10.28}
\expandafter\def\csname mcresult@seed/mhj/img+disc/MSLP/48/mae_sd\endcsname{0.17}
\expandafter\def\csname mcresult@seed/mhj/img+disc/MSLP/48/r2_mean\endcsname{0.47}
\expandafter\def\csname mcresult@seed/mhj/img+disc/MSLP/48/r2_sd\endcsname{0.02}
\expandafter\def\csname mcresult@seed/mhj/img+disc/MSLP/48/rmse_mean\endcsname{14.36}
\expandafter\def\csname mcresult@seed/mhj/img+disc/MSLP/48/rmse_sd\endcsname{0.25}
\expandafter\def\csname mcresult@seed/mhj/img+disc/MSLP/48/skill_score_mean\endcsname{0.27}
\expandafter\def\csname mcresult@seed/mhj/img+disc/MSLP/48/skill_score_sd\endcsname{0.01}
\expandafter\def\csname mcresult@seed/mhj/img+disc/MSLP/72/dir_acc_mean\endcsname{0.738}
\expandafter\def\csname mcresult@seed/mhj/img+disc/MSLP/72/dir_acc_sd\endcsname{0.011}
\expandafter\def\csname mcresult@seed/mhj/img+disc/MSLP/72/mae_mean\endcsname{13.10}
\expandafter\def\csname mcresult@seed/mhj/img+disc/MSLP/72/mae_sd\endcsname{0.25}
\expandafter\def\csname mcresult@seed/mhj/img+disc/MSLP/72/r2_mean\endcsname{0.32}
\expandafter\def\csname mcresult@seed/mhj/img+disc/MSLP/72/r2_sd\endcsname{0.04}
\expandafter\def\csname mcresult@seed/mhj/img+disc/MSLP/72/rmse_mean\endcsname{17.56}
\expandafter\def\csname mcresult@seed/mhj/img+disc/MSLP/72/rmse_sd\endcsname{0.52}
\expandafter\def\csname mcresult@seed/mhj/img+disc/MSLP/72/skill_score_mean\endcsname{0.18}
\expandafter\def\csname mcresult@seed/mhj/img+disc/MSLP/72/skill_score_sd\endcsname{0.02}
\expandafter\def\csname mcresult@seed/mhj/img+disc/MSLP/96/dir_acc_mean\endcsname{0.721}
\expandafter\def\csname mcresult@seed/mhj/img+disc/MSLP/96/dir_acc_sd\endcsname{0.009}
\expandafter\def\csname mcresult@seed/mhj/img+disc/MSLP/96/mae_mean\endcsname{14.66}
\expandafter\def\csname mcresult@seed/mhj/img+disc/MSLP/96/mae_sd\endcsname{0.54}
\expandafter\def\csname mcresult@seed/mhj/img+disc/MSLP/96/r2_mean\endcsname{0.15}
\expandafter\def\csname mcresult@seed/mhj/img+disc/MSLP/96/r2_sd\endcsname{0.09}
\expandafter\def\csname mcresult@seed/mhj/img+disc/MSLP/96/rmse_mean\endcsname{19.48}
\expandafter\def\csname mcresult@seed/mhj/img+disc/MSLP/96/rmse_sd\endcsname{1.02}
\expandafter\def\csname mcresult@seed/mhj/img+disc/MSLP/96/skill_score_mean\endcsname{0.08}
\expandafter\def\csname mcresult@seed/mhj/img+disc/MSLP/96/skill_score_sd\endcsname{0.05}
\expandafter\def\csname mcresult@seed/mhj/img+disc/VMAX/120/dir_acc_mean\endcsname{0.736}
\expandafter\def\csname mcresult@seed/mhj/img+disc/VMAX/120/dir_acc_sd\endcsname{0.018}
\expandafter\def\csname mcresult@seed/mhj/img+disc/VMAX/120/mae_mean\endcsname{20.84}
\expandafter\def\csname mcresult@seed/mhj/img+disc/VMAX/120/mae_sd\endcsname{1.06}
\expandafter\def\csname mcresult@seed/mhj/img+disc/VMAX/120/r2_mean\endcsname{0.09}
\expandafter\def\csname mcresult@seed/mhj/img+disc/VMAX/120/r2_sd\endcsname{0.11}
\expandafter\def\csname mcresult@seed/mhj/img+disc/VMAX/120/rmse_mean\endcsname{27.87}
\expandafter\def\csname mcresult@seed/mhj/img+disc/VMAX/120/rmse_sd\endcsname{1.71}
\expandafter\def\csname mcresult@seed/mhj/img+disc/VMAX/120/skill_score_mean\endcsname{0.05}
\expandafter\def\csname mcresult@seed/mhj/img+disc/VMAX/120/skill_score_sd\endcsname{0.06}
\expandafter\def\csname mcresult@seed/mhj/img+disc/VMAX/24/dir_acc_mean\endcsname{0.708}
\expandafter\def\csname mcresult@seed/mhj/img+disc/VMAX/24/dir_acc_sd\endcsname{0.008}
\expandafter\def\csname mcresult@seed/mhj/img+disc/VMAX/24/mae_mean\endcsname{9.66}
\expandafter\def\csname mcresult@seed/mhj/img+disc/VMAX/24/mae_sd\endcsname{0.12}
\expandafter\def\csname mcresult@seed/mhj/img+disc/VMAX/24/r2_mean\endcsname{0.52}
\expandafter\def\csname mcresult@seed/mhj/img+disc/VMAX/24/r2_sd\endcsname{0.01}
\expandafter\def\csname mcresult@seed/mhj/img+disc/VMAX/24/rmse_mean\endcsname{13.25}
\expandafter\def\csname mcresult@seed/mhj/img+disc/VMAX/24/rmse_sd\endcsname{0.08}
\expandafter\def\csname mcresult@seed/mhj/img+disc/VMAX/24/skill_score_mean\endcsname{0.30}
\expandafter\def\csname mcresult@seed/mhj/img+disc/VMAX/24/skill_score_sd\endcsname{0.00}
\expandafter\def\csname mcresult@seed/mhj/img+disc/VMAX/48/dir_acc_mean\endcsname{0.748}
\expandafter\def\csname mcresult@seed/mhj/img+disc/VMAX/48/dir_acc_sd\endcsname{0.016}
\expandafter\def\csname mcresult@seed/mhj/img+disc/VMAX/48/mae_mean\endcsname{14.81}
\expandafter\def\csname mcresult@seed/mhj/img+disc/VMAX/48/mae_sd\endcsname{0.35}
\expandafter\def\csname mcresult@seed/mhj/img+disc/VMAX/48/r2_mean\endcsname{0.50}
\expandafter\def\csname mcresult@seed/mhj/img+disc/VMAX/48/r2_sd\endcsname{0.02}
\expandafter\def\csname mcresult@seed/mhj/img+disc/VMAX/48/rmse_mean\endcsname{20.07}
\expandafter\def\csname mcresult@seed/mhj/img+disc/VMAX/48/rmse_sd\endcsname{0.44}
\expandafter\def\csname mcresult@seed/mhj/img+disc/VMAX/48/skill_score_mean\endcsname{0.30}
\expandafter\def\csname mcresult@seed/mhj/img+disc/VMAX/48/skill_score_sd\endcsname{0.02}
\expandafter\def\csname mcresult@seed/mhj/img+disc/VMAX/72/dir_acc_mean\endcsname{0.719}
\expandafter\def\csname mcresult@seed/mhj/img+disc/VMAX/72/dir_acc_sd\endcsname{0.010}
\expandafter\def\csname mcresult@seed/mhj/img+disc/VMAX/72/mae_mean\endcsname{17.55}
\expandafter\def\csname mcresult@seed/mhj/img+disc/VMAX/72/mae_sd\endcsname{0.18}
\expandafter\def\csname mcresult@seed/mhj/img+disc/VMAX/72/r2_mean\endcsname{0.42}
\expandafter\def\csname mcresult@seed/mhj/img+disc/VMAX/72/r2_sd\endcsname{0.03}
\expandafter\def\csname mcresult@seed/mhj/img+disc/VMAX/72/rmse_mean\endcsname{23.25}
\expandafter\def\csname mcresult@seed/mhj/img+disc/VMAX/72/rmse_sd\endcsname{0.51}
\expandafter\def\csname mcresult@seed/mhj/img+disc/VMAX/72/skill_score_mean\endcsname{0.24}
\expandafter\def\csname mcresult@seed/mhj/img+disc/VMAX/72/skill_score_sd\endcsname{0.02}
\expandafter\def\csname mcresult@seed/mhj/img+disc/VMAX/96/dir_acc_mean\endcsname{0.713}
\expandafter\def\csname mcresult@seed/mhj/img+disc/VMAX/96/dir_acc_sd\endcsname{0.011}
\expandafter\def\csname mcresult@seed/mhj/img+disc/VMAX/96/mae_mean\endcsname{19.62}
\expandafter\def\csname mcresult@seed/mhj/img+disc/VMAX/96/mae_sd\endcsname{0.65}
\expandafter\def\csname mcresult@seed/mhj/img+disc/VMAX/96/r2_mean\endcsname{0.22}
\expandafter\def\csname mcresult@seed/mhj/img+disc/VMAX/96/r2_sd\endcsname{0.06}
\expandafter\def\csname mcresult@seed/mhj/img+disc/VMAX/96/rmse_mean\endcsname{25.84}
\expandafter\def\csname mcresult@seed/mhj/img+disc/VMAX/96/rmse_sd\endcsname{1.02}
\expandafter\def\csname mcresult@seed/mhj/img+disc/VMAX/96/skill_score_mean\endcsname{0.12}
\expandafter\def\csname mcresult@seed/mhj/img+disc/VMAX/96/skill_score_sd\endcsname{0.03}
\expandafter\def\csname mcresult@seed/mhj/img/MSLP/120/dir_acc_mean\endcsname{0.764}
\expandafter\def\csname mcresult@seed/mhj/img/MSLP/120/dir_acc_sd\endcsname{0.022}
\expandafter\def\csname mcresult@seed/mhj/img/MSLP/120/mae_mean\endcsname{16.84}
\expandafter\def\csname mcresult@seed/mhj/img/MSLP/120/mae_sd\endcsname{1.99}
\expandafter\def\csname mcresult@seed/mhj/img/MSLP/120/r2_mean\endcsname{-0.45}
\expandafter\def\csname mcresult@seed/mhj/img/MSLP/120/r2_sd\endcsname{1.13}
\expandafter\def\csname mcresult@seed/mhj/img/MSLP/120/rmse_mean\endcsname{25.13}
\expandafter\def\csname mcresult@seed/mhj/img/MSLP/120/rmse_sd\endcsname{8.75}
\expandafter\def\csname mcresult@seed/mhj/img/MSLP/120/skill_score_mean\endcsname{-0.15}
\expandafter\def\csname mcresult@seed/mhj/img/MSLP/120/skill_score_sd\endcsname{0.40}
\expandafter\def\csname mcresult@seed/mhj/img/MSLP/24/dir_acc_mean\endcsname{0.652}
\expandafter\def\csname mcresult@seed/mhj/img/MSLP/24/dir_acc_sd\endcsname{0.008}
\expandafter\def\csname mcresult@seed/mhj/img/MSLP/24/mae_mean\endcsname{8.32}
\expandafter\def\csname mcresult@seed/mhj/img/MSLP/24/mae_sd\endcsname{0.51}
\expandafter\def\csname mcresult@seed/mhj/img/MSLP/24/r2_mean\endcsname{0.05}
\expandafter\def\csname mcresult@seed/mhj/img/MSLP/24/r2_sd\endcsname{0.26}
\expandafter\def\csname mcresult@seed/mhj/img/MSLP/24/rmse_mean\endcsname{12.65}
\expandafter\def\csname mcresult@seed/mhj/img/MSLP/24/rmse_sd\endcsname{1.63}
\expandafter\def\csname mcresult@seed/mhj/img/MSLP/24/skill_score_mean\endcsname{0.03}
\expandafter\def\csname mcresult@seed/mhj/img/MSLP/24/skill_score_sd\endcsname{0.13}
\expandafter\def\csname mcresult@seed/mhj/img/MSLP/48/dir_acc_mean\endcsname{0.694}
\expandafter\def\csname mcresult@seed/mhj/img/MSLP/48/dir_acc_sd\endcsname{0.012}
\expandafter\def\csname mcresult@seed/mhj/img/MSLP/48/mae_mean\endcsname{13.42}
\expandafter\def\csname mcresult@seed/mhj/img/MSLP/48/mae_sd\endcsname{0.86}
\expandafter\def\csname mcresult@seed/mhj/img/MSLP/48/r2_mean\endcsname{0.04}
\expandafter\def\csname mcresult@seed/mhj/img/MSLP/48/r2_sd\endcsname{0.35}
\expandafter\def\csname mcresult@seed/mhj/img/MSLP/48/rmse_mean\endcsname{19.08}
\expandafter\def\csname mcresult@seed/mhj/img/MSLP/48/rmse_sd\endcsname{3.18}
\expandafter\def\csname mcresult@seed/mhj/img/MSLP/48/skill_score_mean\endcsname{0.03}
\expandafter\def\csname mcresult@seed/mhj/img/MSLP/48/skill_score_sd\endcsname{0.16}
\expandafter\def\csname mcresult@seed/mhj/img/MSLP/72/dir_acc_mean\endcsname{0.689}
\expandafter\def\csname mcresult@seed/mhj/img/MSLP/72/dir_acc_sd\endcsname{0.014}
\expandafter\def\csname mcresult@seed/mhj/img/MSLP/72/mae_mean\endcsname{15.42}
\expandafter\def\csname mcresult@seed/mhj/img/MSLP/72/mae_sd\endcsname{0.94}
\expandafter\def\csname mcresult@seed/mhj/img/MSLP/72/r2_mean\endcsname{-0.03}
\expandafter\def\csname mcresult@seed/mhj/img/MSLP/72/r2_sd\endcsname{0.43}
\expandafter\def\csname mcresult@seed/mhj/img/MSLP/72/rmse_mean\endcsname{21.35}
\expandafter\def\csname mcresult@seed/mhj/img/MSLP/72/rmse_sd\endcsname{4.10}
\expandafter\def\csname mcresult@seed/mhj/img/MSLP/72/skill_score_mean\endcsname{0.00}
\expandafter\def\csname mcresult@seed/mhj/img/MSLP/72/skill_score_sd\endcsname{0.19}
\expandafter\def\csname mcresult@seed/mhj/img/MSLP/96/dir_acc_mean\endcsname{0.720}
\expandafter\def\csname mcresult@seed/mhj/img/MSLP/96/dir_acc_sd\endcsname{0.020}
\expandafter\def\csname mcresult@seed/mhj/img/MSLP/96/mae_mean\endcsname{15.71}
\expandafter\def\csname mcresult@seed/mhj/img/MSLP/96/mae_sd\endcsname{1.27}
\expandafter\def\csname mcresult@seed/mhj/img/MSLP/96/r2_mean\endcsname{-0.15}
\expandafter\def\csname mcresult@seed/mhj/img/MSLP/96/r2_sd\endcsname{0.61}
\expandafter\def\csname mcresult@seed/mhj/img/MSLP/96/rmse_mean\endcsname{22.13}
\expandafter\def\csname mcresult@seed/mhj/img/MSLP/96/rmse_sd\endcsname{5.30}
\expandafter\def\csname mcresult@seed/mhj/img/MSLP/96/skill_score_mean\endcsname{-0.05}
\expandafter\def\csname mcresult@seed/mhj/img/MSLP/96/skill_score_sd\endcsname{0.25}
\expandafter\def\csname mcresult@seed/mhj/img/VMAX/120/dir_acc_mean\endcsname{0.735}
\expandafter\def\csname mcresult@seed/mhj/img/VMAX/120/dir_acc_sd\endcsname{0.026}
\expandafter\def\csname mcresult@seed/mhj/img/VMAX/120/mae_mean\endcsname{21.31}
\expandafter\def\csname mcresult@seed/mhj/img/VMAX/120/mae_sd\endcsname{2.87}
\expandafter\def\csname mcresult@seed/mhj/img/VMAX/120/r2_mean\endcsname{-0.46}
\expandafter\def\csname mcresult@seed/mhj/img/VMAX/120/r2_sd\endcsname{1.44}
\expandafter\def\csname mcresult@seed/mhj/img/VMAX/120/rmse_mean\endcsname{33.00}
\expandafter\def\csname mcresult@seed/mhj/img/VMAX/120/rmse_sd\endcsname{14.49}
\expandafter\def\csname mcresult@seed/mhj/img/VMAX/120/skill_score_mean\endcsname{-0.13}
\expandafter\def\csname mcresult@seed/mhj/img/VMAX/120/skill_score_sd\endcsname{0.49}
\expandafter\def\csname mcresult@seed/mhj/img/VMAX/24/dir_acc_mean\endcsname{0.617}
\expandafter\def\csname mcresult@seed/mhj/img/VMAX/24/dir_acc_sd\endcsname{0.009}
\expandafter\def\csname mcresult@seed/mhj/img/VMAX/24/mae_mean\endcsname{12.60}
\expandafter\def\csname mcresult@seed/mhj/img/VMAX/24/mae_sd\endcsname{0.66}
\expandafter\def\csname mcresult@seed/mhj/img/VMAX/24/r2_mean\endcsname{0.09}
\expandafter\def\csname mcresult@seed/mhj/img/VMAX/24/r2_sd\endcsname{0.23}
\expandafter\def\csname mcresult@seed/mhj/img/VMAX/24/rmse_mean\endcsname{18.03}
\expandafter\def\csname mcresult@seed/mhj/img/VMAX/24/rmse_sd\endcsname{2.14}
\expandafter\def\csname mcresult@seed/mhj/img/VMAX/24/skill_score_mean\endcsname{0.05}
\expandafter\def\csname mcresult@seed/mhj/img/VMAX/24/skill_score_sd\endcsname{0.11}
\expandafter\def\csname mcresult@seed/mhj/img/VMAX/48/dir_acc_mean\endcsname{0.673}
\expandafter\def\csname mcresult@seed/mhj/img/VMAX/48/dir_acc_sd\endcsname{0.011}
\expandafter\def\csname mcresult@seed/mhj/img/VMAX/48/mae_mean\endcsname{18.94}
\expandafter\def\csname mcresult@seed/mhj/img/VMAX/48/mae_sd\endcsname{1.14}
\expandafter\def\csname mcresult@seed/mhj/img/VMAX/48/r2_mean\endcsname{0.12}
\expandafter\def\csname mcresult@seed/mhj/img/VMAX/48/r2_sd\endcsname{0.29}
\expandafter\def\csname mcresult@seed/mhj/img/VMAX/48/rmse_mean\endcsname{26.48}
\expandafter\def\csname mcresult@seed/mhj/img/VMAX/48/rmse_sd\endcsname{4.08}
\expandafter\def\csname mcresult@seed/mhj/img/VMAX/48/skill_score_mean\endcsname{0.07}
\expandafter\def\csname mcresult@seed/mhj/img/VMAX/48/skill_score_sd\endcsname{0.14}
\expandafter\def\csname mcresult@seed/mhj/img/VMAX/72/dir_acc_mean\endcsname{0.669}
\expandafter\def\csname mcresult@seed/mhj/img/VMAX/72/dir_acc_sd\endcsname{0.015}
\expandafter\def\csname mcresult@seed/mhj/img/VMAX/72/mae_mean\endcsname{20.91}
\expandafter\def\csname mcresult@seed/mhj/img/VMAX/72/mae_sd\endcsname{1.28}
\expandafter\def\csname mcresult@seed/mhj/img/VMAX/72/r2_mean\endcsname{0.08}
\expandafter\def\csname mcresult@seed/mhj/img/VMAX/72/r2_sd\endcsname{0.38}
\expandafter\def\csname mcresult@seed/mhj/img/VMAX/72/rmse_mean\endcsname{28.72}
\expandafter\def\csname mcresult@seed/mhj/img/VMAX/72/rmse_sd\endcsname{5.47}
\expandafter\def\csname mcresult@seed/mhj/img/VMAX/72/skill_score_mean\endcsname{0.06}
\expandafter\def\csname mcresult@seed/mhj/img/VMAX/72/skill_score_sd\endcsname{0.18}
\expandafter\def\csname mcresult@seed/mhj/img/VMAX/96/dir_acc_mean\endcsname{0.692}
\expandafter\def\csname mcresult@seed/mhj/img/VMAX/96/dir_acc_sd\endcsname{0.016}
\expandafter\def\csname mcresult@seed/mhj/img/VMAX/96/mae_mean\endcsname{20.95}
\expandafter\def\csname mcresult@seed/mhj/img/VMAX/96/mae_sd\endcsname{1.90}
\expandafter\def\csname mcresult@seed/mhj/img/VMAX/96/r2_mean\endcsname{-0.06}
\expandafter\def\csname mcresult@seed/mhj/img/VMAX/96/r2_sd\endcsname{0.65}
\expandafter\def\csname mcresult@seed/mhj/img/VMAX/96/rmse_mean\endcsname{29.28}
\expandafter\def\csname mcresult@seed/mhj/img/VMAX/96/rmse_sd\endcsname{8.08}
\expandafter\def\csname mcresult@seed/mhj/img/VMAX/96/skill_score_mean\endcsname{0.00}
\expandafter\def\csname mcresult@seed/mhj/img/VMAX/96/skill_score_sd\endcsname{0.28}
\expandafter\def\csname mcresult@seed/mhj/tab+disc/MSLP/120/dir_acc_mean\endcsname{0.756}
\expandafter\def\csname mcresult@seed/mhj/tab+disc/MSLP/120/dir_acc_sd\endcsname{0.019}
\expandafter\def\csname mcresult@seed/mhj/tab+disc/MSLP/120/mae_mean\endcsname{16.01}
\expandafter\def\csname mcresult@seed/mhj/tab+disc/MSLP/120/mae_sd\endcsname{0.74}
\expandafter\def\csname mcresult@seed/mhj/tab+disc/MSLP/120/r2_mean\endcsname{-0.00}
\expandafter\def\csname mcresult@seed/mhj/tab+disc/MSLP/120/r2_sd\endcsname{0.14}
\expandafter\def\csname mcresult@seed/mhj/tab+disc/MSLP/120/rmse_mean\endcsname{21.83}
\expandafter\def\csname mcresult@seed/mhj/tab+disc/MSLP/120/rmse_sd\endcsname{1.58}
\expandafter\def\csname mcresult@seed/mhj/tab+disc/MSLP/120/skill_score_mean\endcsname{0.00}
\expandafter\def\csname mcresult@seed/mhj/tab+disc/MSLP/120/skill_score_sd\endcsname{0.07}
\expandafter\def\csname mcresult@seed/mhj/tab+disc/MSLP/24/dir_acc_mean\endcsname{0.763}
\expandafter\def\csname mcresult@seed/mhj/tab+disc/MSLP/24/dir_acc_sd\endcsname{0.007}
\expandafter\def\csname mcresult@seed/mhj/tab+disc/MSLP/24/mae_mean\endcsname{6.19}
\expandafter\def\csname mcresult@seed/mhj/tab+disc/MSLP/24/mae_sd\endcsname{0.18}
\expandafter\def\csname mcresult@seed/mhj/tab+disc/MSLP/24/r2_mean\endcsname{0.51}
\expandafter\def\csname mcresult@seed/mhj/tab+disc/MSLP/24/r2_sd\endcsname{0.02}
\expandafter\def\csname mcresult@seed/mhj/tab+disc/MSLP/24/rmse_mean\endcsname{9.15}
\expandafter\def\csname mcresult@seed/mhj/tab+disc/MSLP/24/rmse_sd\endcsname{0.16}
\expandafter\def\csname mcresult@seed/mhj/tab+disc/MSLP/24/skill_score_mean\endcsname{0.30}
\expandafter\def\csname mcresult@seed/mhj/tab+disc/MSLP/24/skill_score_sd\endcsname{0.01}
\expandafter\def\csname mcresult@seed/mhj/tab+disc/MSLP/48/dir_acc_mean\endcsname{0.781}
\expandafter\def\csname mcresult@seed/mhj/tab+disc/MSLP/48/dir_acc_sd\endcsname{0.006}
\expandafter\def\csname mcresult@seed/mhj/tab+disc/MSLP/48/mae_mean\endcsname{9.89}
\expandafter\def\csname mcresult@seed/mhj/tab+disc/MSLP/48/mae_sd\endcsname{0.18}
\expandafter\def\csname mcresult@seed/mhj/tab+disc/MSLP/48/r2_mean\endcsname{0.51}
\expandafter\def\csname mcresult@seed/mhj/tab+disc/MSLP/48/r2_sd\endcsname{0.01}
\expandafter\def\csname mcresult@seed/mhj/tab+disc/MSLP/48/rmse_mean\endcsname{13.80}
\expandafter\def\csname mcresult@seed/mhj/tab+disc/MSLP/48/rmse_sd\endcsname{0.21}
\expandafter\def\csname mcresult@seed/mhj/tab+disc/MSLP/48/skill_score_mean\endcsname{0.30}
\expandafter\def\csname mcresult@seed/mhj/tab+disc/MSLP/48/skill_score_sd\endcsname{0.01}
\expandafter\def\csname mcresult@seed/mhj/tab+disc/MSLP/72/dir_acc_mean\endcsname{0.756}
\expandafter\def\csname mcresult@seed/mhj/tab+disc/MSLP/72/dir_acc_sd\endcsname{0.006}
\expandafter\def\csname mcresult@seed/mhj/tab+disc/MSLP/72/mae_mean\endcsname{12.38}
\expandafter\def\csname mcresult@seed/mhj/tab+disc/MSLP/72/mae_sd\endcsname{0.35}
\expandafter\def\csname mcresult@seed/mhj/tab+disc/MSLP/72/r2_mean\endcsname{0.37}
\expandafter\def\csname mcresult@seed/mhj/tab+disc/MSLP/72/r2_sd\endcsname{0.04}
\expandafter\def\csname mcresult@seed/mhj/tab+disc/MSLP/72/rmse_mean\endcsname{17.00}
\expandafter\def\csname mcresult@seed/mhj/tab+disc/MSLP/72/rmse_sd\endcsname{0.61}
\expandafter\def\csname mcresult@seed/mhj/tab+disc/MSLP/72/skill_score_mean\endcsname{0.20}
\expandafter\def\csname mcresult@seed/mhj/tab+disc/MSLP/72/skill_score_sd\endcsname{0.03}
\expandafter\def\csname mcresult@seed/mhj/tab+disc/MSLP/96/dir_acc_mean\endcsname{0.737}
\expandafter\def\csname mcresult@seed/mhj/tab+disc/MSLP/96/dir_acc_sd\endcsname{0.010}
\expandafter\def\csname mcresult@seed/mhj/tab+disc/MSLP/96/mae_mean\endcsname{14.23}
\expandafter\def\csname mcresult@seed/mhj/tab+disc/MSLP/96/mae_sd\endcsname{0.44}
\expandafter\def\csname mcresult@seed/mhj/tab+disc/MSLP/96/r2_mean\endcsname{0.15}
\expandafter\def\csname mcresult@seed/mhj/tab+disc/MSLP/96/r2_sd\endcsname{0.08}
\expandafter\def\csname mcresult@seed/mhj/tab+disc/MSLP/96/rmse_mean\endcsname{19.45}
\expandafter\def\csname mcresult@seed/mhj/tab+disc/MSLP/96/rmse_sd\endcsname{0.89}
\expandafter\def\csname mcresult@seed/mhj/tab+disc/MSLP/96/skill_score_mean\endcsname{0.08}
\expandafter\def\csname mcresult@seed/mhj/tab+disc/MSLP/96/skill_score_sd\endcsname{0.04}
\expandafter\def\csname mcresult@seed/mhj/tab+disc/VMAX/120/dir_acc_mean\endcsname{0.743}
\expandafter\def\csname mcresult@seed/mhj/tab+disc/VMAX/120/dir_acc_sd\endcsname{0.018}
\expandafter\def\csname mcresult@seed/mhj/tab+disc/VMAX/120/mae_mean\endcsname{21.29}
\expandafter\def\csname mcresult@seed/mhj/tab+disc/VMAX/120/mae_sd\endcsname{0.66}
\expandafter\def\csname mcresult@seed/mhj/tab+disc/VMAX/120/r2_mean\endcsname{0.06}
\expandafter\def\csname mcresult@seed/mhj/tab+disc/VMAX/120/r2_sd\endcsname{0.10}
\expandafter\def\csname mcresult@seed/mhj/tab+disc/VMAX/120/rmse_mean\endcsname{28.46}
\expandafter\def\csname mcresult@seed/mhj/tab+disc/VMAX/120/rmse_sd\endcsname{1.48}
\expandafter\def\csname mcresult@seed/mhj/tab+disc/VMAX/120/skill_score_mean\endcsname{0.03}
\expandafter\def\csname mcresult@seed/mhj/tab+disc/VMAX/120/skill_score_sd\endcsname{0.05}
\expandafter\def\csname mcresult@seed/mhj/tab+disc/VMAX/24/dir_acc_mean\endcsname{0.724}
\expandafter\def\csname mcresult@seed/mhj/tab+disc/VMAX/24/dir_acc_sd\endcsname{0.003}
\expandafter\def\csname mcresult@seed/mhj/tab+disc/VMAX/24/mae_mean\endcsname{9.40}
\expandafter\def\csname mcresult@seed/mhj/tab+disc/VMAX/24/mae_sd\endcsname{0.30}
\expandafter\def\csname mcresult@seed/mhj/tab+disc/VMAX/24/r2_mean\endcsname{0.54}
\expandafter\def\csname mcresult@seed/mhj/tab+disc/VMAX/24/r2_sd\endcsname{0.02}
\expandafter\def\csname mcresult@seed/mhj/tab+disc/VMAX/24/rmse_mean\endcsname{12.93}
\expandafter\def\csname mcresult@seed/mhj/tab+disc/VMAX/24/rmse_sd\endcsname{0.32}
\expandafter\def\csname mcresult@seed/mhj/tab+disc/VMAX/24/skill_score_mean\endcsname{0.32}
\expandafter\def\csname mcresult@seed/mhj/tab+disc/VMAX/24/skill_score_sd\endcsname{0.02}
\expandafter\def\csname mcresult@seed/mhj/tab+disc/VMAX/48/dir_acc_mean\endcsname{0.767}
\expandafter\def\csname mcresult@seed/mhj/tab+disc/VMAX/48/dir_acc_sd\endcsname{0.007}
\expandafter\def\csname mcresult@seed/mhj/tab+disc/VMAX/48/mae_mean\endcsname{13.84}
\expandafter\def\csname mcresult@seed/mhj/tab+disc/VMAX/48/mae_sd\endcsname{0.28}
\expandafter\def\csname mcresult@seed/mhj/tab+disc/VMAX/48/r2_mean\endcsname{0.55}
\expandafter\def\csname mcresult@seed/mhj/tab+disc/VMAX/48/r2_sd\endcsname{0.02}
\expandafter\def\csname mcresult@seed/mhj/tab+disc/VMAX/48/rmse_mean\endcsname{19.05}
\expandafter\def\csname mcresult@seed/mhj/tab+disc/VMAX/48/rmse_sd\endcsname{0.33}
\expandafter\def\csname mcresult@seed/mhj/tab+disc/VMAX/48/skill_score_mean\endcsname{0.33}
\expandafter\def\csname mcresult@seed/mhj/tab+disc/VMAX/48/skill_score_sd\endcsname{0.01}
\expandafter\def\csname mcresult@seed/mhj/tab+disc/VMAX/72/dir_acc_mean\endcsname{0.738}
\expandafter\def\csname mcresult@seed/mhj/tab+disc/VMAX/72/dir_acc_sd\endcsname{0.005}
\expandafter\def\csname mcresult@seed/mhj/tab+disc/VMAX/72/mae_mean\endcsname{16.74}
\expandafter\def\csname mcresult@seed/mhj/tab+disc/VMAX/72/mae_sd\endcsname{0.41}
\expandafter\def\csname mcresult@seed/mhj/tab+disc/VMAX/72/r2_mean\endcsname{0.44}
\expandafter\def\csname mcresult@seed/mhj/tab+disc/VMAX/72/r2_sd\endcsname{0.03}
\expandafter\def\csname mcresult@seed/mhj/tab+disc/VMAX/72/rmse_mean\endcsname{22.76}
\expandafter\def\csname mcresult@seed/mhj/tab+disc/VMAX/72/rmse_sd\endcsname{0.58}
\expandafter\def\csname mcresult@seed/mhj/tab+disc/VMAX/72/skill_score_mean\endcsname{0.25}
\expandafter\def\csname mcresult@seed/mhj/tab+disc/VMAX/72/skill_score_sd\endcsname{0.02}
\expandafter\def\csname mcresult@seed/mhj/tab+disc/VMAX/96/dir_acc_mean\endcsname{0.733}
\expandafter\def\csname mcresult@seed/mhj/tab+disc/VMAX/96/dir_acc_sd\endcsname{0.007}
\expandafter\def\csname mcresult@seed/mhj/tab+disc/VMAX/96/mae_mean\endcsname{19.55}
\expandafter\def\csname mcresult@seed/mhj/tab+disc/VMAX/96/mae_sd\endcsname{0.47}
\expandafter\def\csname mcresult@seed/mhj/tab+disc/VMAX/96/r2_mean\endcsname{0.20}
\expandafter\def\csname mcresult@seed/mhj/tab+disc/VMAX/96/r2_sd\endcsname{0.06}
\expandafter\def\csname mcresult@seed/mhj/tab+disc/VMAX/96/rmse_mean\endcsname{26.12}
\expandafter\def\csname mcresult@seed/mhj/tab+disc/VMAX/96/rmse_sd\endcsname{1.02}
\expandafter\def\csname mcresult@seed/mhj/tab+disc/VMAX/96/skill_score_mean\endcsname{0.11}
\expandafter\def\csname mcresult@seed/mhj/tab+disc/VMAX/96/skill_score_sd\endcsname{0.03}
\expandafter\def\csname mcresult@seed/mhj/tab+img/MSLP/120/dir_acc_mean\endcsname{0.768}
\expandafter\def\csname mcresult@seed/mhj/tab+img/MSLP/120/dir_acc_sd\endcsname{0.005}
\expandafter\def\csname mcresult@seed/mhj/tab+img/MSLP/120/mae_mean\endcsname{16.00}
\expandafter\def\csname mcresult@seed/mhj/tab+img/MSLP/120/mae_sd\endcsname{0.40}
\expandafter\def\csname mcresult@seed/mhj/tab+img/MSLP/120/r2_mean\endcsname{0.11}
\expandafter\def\csname mcresult@seed/mhj/tab+img/MSLP/120/r2_sd\endcsname{0.03}
\expandafter\def\csname mcresult@seed/mhj/tab+img/MSLP/120/rmse_mean\endcsname{20.60}
\expandafter\def\csname mcresult@seed/mhj/tab+img/MSLP/120/rmse_sd\endcsname{0.37}
\expandafter\def\csname mcresult@seed/mhj/tab+img/MSLP/120/skill_score_mean\endcsname{0.06}
\expandafter\def\csname mcresult@seed/mhj/tab+img/MSLP/120/skill_score_sd\endcsname{0.02}
\expandafter\def\csname mcresult@seed/mhj/tab+img/MSLP/24/dir_acc_mean\endcsname{0.713}
\expandafter\def\csname mcresult@seed/mhj/tab+img/MSLP/24/dir_acc_sd\endcsname{0.012}
\expandafter\def\csname mcresult@seed/mhj/tab+img/MSLP/24/mae_mean\endcsname{6.80}
\expandafter\def\csname mcresult@seed/mhj/tab+img/MSLP/24/mae_sd\endcsname{0.10}
\expandafter\def\csname mcresult@seed/mhj/tab+img/MSLP/24/r2_mean\endcsname{0.41}
\expandafter\def\csname mcresult@seed/mhj/tab+img/MSLP/24/r2_sd\endcsname{0.02}
\expandafter\def\csname mcresult@seed/mhj/tab+img/MSLP/24/rmse_mean\endcsname{9.99}
\expandafter\def\csname mcresult@seed/mhj/tab+img/MSLP/24/rmse_sd\endcsname{0.17}
\expandafter\def\csname mcresult@seed/mhj/tab+img/MSLP/24/skill_score_mean\endcsname{0.23}
\expandafter\def\csname mcresult@seed/mhj/tab+img/MSLP/24/skill_score_sd\endcsname{0.01}
\expandafter\def\csname mcresult@seed/mhj/tab+img/MSLP/48/dir_acc_mean\endcsname{0.767}
\expandafter\def\csname mcresult@seed/mhj/tab+img/MSLP/48/dir_acc_sd\endcsname{0.015}
\expandafter\def\csname mcresult@seed/mhj/tab+img/MSLP/48/mae_mean\endcsname{10.96}
\expandafter\def\csname mcresult@seed/mhj/tab+img/MSLP/48/mae_sd\endcsname{0.38}
\expandafter\def\csname mcresult@seed/mhj/tab+img/MSLP/48/r2_mean\endcsname{0.42}
\expandafter\def\csname mcresult@seed/mhj/tab+img/MSLP/48/r2_sd\endcsname{0.02}
\expandafter\def\csname mcresult@seed/mhj/tab+img/MSLP/48/rmse_mean\endcsname{15.00}
\expandafter\def\csname mcresult@seed/mhj/tab+img/MSLP/48/rmse_sd\endcsname{0.25}
\expandafter\def\csname mcresult@seed/mhj/tab+img/MSLP/48/skill_score_mean\endcsname{0.24}
\expandafter\def\csname mcresult@seed/mhj/tab+img/MSLP/48/skill_score_sd\endcsname{0.01}
\expandafter\def\csname mcresult@seed/mhj/tab+img/MSLP/72/dir_acc_mean\endcsname{0.750}
\expandafter\def\csname mcresult@seed/mhj/tab+img/MSLP/72/dir_acc_sd\endcsname{0.016}
\expandafter\def\csname mcresult@seed/mhj/tab+img/MSLP/72/mae_mean\endcsname{13.30}
\expandafter\def\csname mcresult@seed/mhj/tab+img/MSLP/72/mae_sd\endcsname{0.40}
\expandafter\def\csname mcresult@seed/mhj/tab+img/MSLP/72/r2_mean\endcsname{0.37}
\expandafter\def\csname mcresult@seed/mhj/tab+img/MSLP/72/r2_sd\endcsname{0.02}
\expandafter\def\csname mcresult@seed/mhj/tab+img/MSLP/72/rmse_mean\endcsname{16.98}
\expandafter\def\csname mcresult@seed/mhj/tab+img/MSLP/72/rmse_sd\endcsname{0.31}
\expandafter\def\csname mcresult@seed/mhj/tab+img/MSLP/72/skill_score_mean\endcsname{0.21}
\expandafter\def\csname mcresult@seed/mhj/tab+img/MSLP/72/skill_score_sd\endcsname{0.01}
\expandafter\def\csname mcresult@seed/mhj/tab+img/MSLP/96/dir_acc_mean\endcsname{0.751}
\expandafter\def\csname mcresult@seed/mhj/tab+img/MSLP/96/dir_acc_sd\endcsname{0.011}
\expandafter\def\csname mcresult@seed/mhj/tab+img/MSLP/96/mae_mean\endcsname{14.21}
\expandafter\def\csname mcresult@seed/mhj/tab+img/MSLP/96/mae_sd\endcsname{0.34}
\expandafter\def\csname mcresult@seed/mhj/tab+img/MSLP/96/r2_mean\endcsname{0.26}
\expandafter\def\csname mcresult@seed/mhj/tab+img/MSLP/96/r2_sd\endcsname{0.02}
\expandafter\def\csname mcresult@seed/mhj/tab+img/MSLP/96/rmse_mean\endcsname{18.21}
\expandafter\def\csname mcresult@seed/mhj/tab+img/MSLP/96/rmse_sd\endcsname{0.25}
\expandafter\def\csname mcresult@seed/mhj/tab+img/MSLP/96/skill_score_mean\endcsname{0.14}
\expandafter\def\csname mcresult@seed/mhj/tab+img/MSLP/96/skill_score_sd\endcsname{0.01}
\expandafter\def\csname mcresult@seed/mhj/tab+img/VMAX/120/dir_acc_mean\endcsname{0.761}
\expandafter\def\csname mcresult@seed/mhj/tab+img/VMAX/120/dir_acc_sd\endcsname{0.009}
\expandafter\def\csname mcresult@seed/mhj/tab+img/VMAX/120/mae_mean\endcsname{20.22}
\expandafter\def\csname mcresult@seed/mhj/tab+img/VMAX/120/mae_sd\endcsname{0.53}
\expandafter\def\csname mcresult@seed/mhj/tab+img/VMAX/120/r2_mean\endcsname{0.20}
\expandafter\def\csname mcresult@seed/mhj/tab+img/VMAX/120/r2_sd\endcsname{0.04}
\expandafter\def\csname mcresult@seed/mhj/tab+img/VMAX/120/rmse_mean\endcsname{26.14}
\expandafter\def\csname mcresult@seed/mhj/tab+img/VMAX/120/rmse_sd\endcsname{0.65}
\expandafter\def\csname mcresult@seed/mhj/tab+img/VMAX/120/skill_score_mean\endcsname{0.11}
\expandafter\def\csname mcresult@seed/mhj/tab+img/VMAX/120/skill_score_sd\endcsname{0.02}
\expandafter\def\csname mcresult@seed/mhj/tab+img/VMAX/24/dir_acc_mean\endcsname{0.684}
\expandafter\def\csname mcresult@seed/mhj/tab+img/VMAX/24/dir_acc_sd\endcsname{0.011}
\expandafter\def\csname mcresult@seed/mhj/tab+img/VMAX/24/mae_mean\endcsname{10.44}
\expandafter\def\csname mcresult@seed/mhj/tab+img/VMAX/24/mae_sd\endcsname{0.09}
\expandafter\def\csname mcresult@seed/mhj/tab+img/VMAX/24/r2_mean\endcsname{0.41}
\expandafter\def\csname mcresult@seed/mhj/tab+img/VMAX/24/r2_sd\endcsname{0.01}
\expandafter\def\csname mcresult@seed/mhj/tab+img/VMAX/24/rmse_mean\endcsname{14.56}
\expandafter\def\csname mcresult@seed/mhj/tab+img/VMAX/24/rmse_sd\endcsname{0.14}
\expandafter\def\csname mcresult@seed/mhj/tab+img/VMAX/24/skill_score_mean\endcsname{0.24}
\expandafter\def\csname mcresult@seed/mhj/tab+img/VMAX/24/skill_score_sd\endcsname{0.01}
\expandafter\def\csname mcresult@seed/mhj/tab+img/VMAX/48/dir_acc_mean\endcsname{0.741}
\expandafter\def\csname mcresult@seed/mhj/tab+img/VMAX/48/dir_acc_sd\endcsname{0.008}
\expandafter\def\csname mcresult@seed/mhj/tab+img/VMAX/48/mae_mean\endcsname{15.50}
\expandafter\def\csname mcresult@seed/mhj/tab+img/VMAX/48/mae_sd\endcsname{0.30}
\expandafter\def\csname mcresult@seed/mhj/tab+img/VMAX/48/r2_mean\endcsname{0.46}
\expandafter\def\csname mcresult@seed/mhj/tab+img/VMAX/48/r2_sd\endcsname{0.01}
\expandafter\def\csname mcresult@seed/mhj/tab+img/VMAX/48/rmse_mean\endcsname{20.96}
\expandafter\def\csname mcresult@seed/mhj/tab+img/VMAX/48/rmse_sd\endcsname{0.23}
\expandafter\def\csname mcresult@seed/mhj/tab+img/VMAX/48/skill_score_mean\endcsname{0.26}
\expandafter\def\csname mcresult@seed/mhj/tab+img/VMAX/48/skill_score_sd\endcsname{0.01}
\expandafter\def\csname mcresult@seed/mhj/tab+img/VMAX/72/dir_acc_mean\endcsname{0.724}
\expandafter\def\csname mcresult@seed/mhj/tab+img/VMAX/72/dir_acc_sd\endcsname{0.009}
\expandafter\def\csname mcresult@seed/mhj/tab+img/VMAX/72/mae_mean\endcsname{17.99}
\expandafter\def\csname mcresult@seed/mhj/tab+img/VMAX/72/mae_sd\endcsname{0.48}
\expandafter\def\csname mcresult@seed/mhj/tab+img/VMAX/72/r2_mean\endcsname{0.42}
\expandafter\def\csname mcresult@seed/mhj/tab+img/VMAX/72/r2_sd\endcsname{0.02}
\expandafter\def\csname mcresult@seed/mhj/tab+img/VMAX/72/rmse_mean\endcsname{23.20}
\expandafter\def\csname mcresult@seed/mhj/tab+img/VMAX/72/rmse_sd\endcsname{0.44}
\expandafter\def\csname mcresult@seed/mhj/tab+img/VMAX/72/skill_score_mean\endcsname{0.24}
\expandafter\def\csname mcresult@seed/mhj/tab+img/VMAX/72/skill_score_sd\endcsname{0.01}
\expandafter\def\csname mcresult@seed/mhj/tab+img/VMAX/96/dir_acc_mean\endcsname{0.720}
\expandafter\def\csname mcresult@seed/mhj/tab+img/VMAX/96/dir_acc_sd\endcsname{0.010}
\expandafter\def\csname mcresult@seed/mhj/tab+img/VMAX/96/mae_mean\endcsname{19.19}
\expandafter\def\csname mcresult@seed/mhj/tab+img/VMAX/96/mae_sd\endcsname{0.65}
\expandafter\def\csname mcresult@seed/mhj/tab+img/VMAX/96/r2_mean\endcsname{0.30}
\expandafter\def\csname mcresult@seed/mhj/tab+img/VMAX/96/r2_sd\endcsname{0.04}
\expandafter\def\csname mcresult@seed/mhj/tab+img/VMAX/96/rmse_mean\endcsname{24.46}
\expandafter\def\csname mcresult@seed/mhj/tab+img/VMAX/96/rmse_sd\endcsname{0.71}
\expandafter\def\csname mcresult@seed/mhj/tab+img/VMAX/96/skill_score_mean\endcsname{0.17}
\expandafter\def\csname mcresult@seed/mhj/tab+img/VMAX/96/skill_score_sd\endcsname{0.02}
\expandafter\def\csname mcresult@seed/mhj/tab/MSLP/120/dir_acc_mean\endcsname{0.776}
\expandafter\def\csname mcresult@seed/mhj/tab/MSLP/120/dir_acc_sd\endcsname{0.013}
\expandafter\def\csname mcresult@seed/mhj/tab/MSLP/120/mae_mean\endcsname{16.07}
\expandafter\def\csname mcresult@seed/mhj/tab/MSLP/120/mae_sd\endcsname{0.44}
\expandafter\def\csname mcresult@seed/mhj/tab/MSLP/120/r2_mean\endcsname{0.10}
\expandafter\def\csname mcresult@seed/mhj/tab/MSLP/120/r2_sd\endcsname{0.04}
\expandafter\def\csname mcresult@seed/mhj/tab/MSLP/120/rmse_mean\endcsname{20.75}
\expandafter\def\csname mcresult@seed/mhj/tab/MSLP/120/rmse_sd\endcsname{0.50}
\expandafter\def\csname mcresult@seed/mhj/tab/MSLP/120/skill_score_mean\endcsname{0.05}
\expandafter\def\csname mcresult@seed/mhj/tab/MSLP/120/skill_score_sd\endcsname{0.02}
\expandafter\def\csname mcresult@seed/mhj/tab/MSLP/24/dir_acc_mean\endcsname{0.714}
\expandafter\def\csname mcresult@seed/mhj/tab/MSLP/24/dir_acc_sd\endcsname{0.012}
\expandafter\def\csname mcresult@seed/mhj/tab/MSLP/24/mae_mean\endcsname{6.73}
\expandafter\def\csname mcresult@seed/mhj/tab/MSLP/24/mae_sd\endcsname{0.10}
\expandafter\def\csname mcresult@seed/mhj/tab/MSLP/24/r2_mean\endcsname{0.41}
\expandafter\def\csname mcresult@seed/mhj/tab/MSLP/24/r2_sd\endcsname{0.02}
\expandafter\def\csname mcresult@seed/mhj/tab/MSLP/24/rmse_mean\endcsname{10.01}
\expandafter\def\csname mcresult@seed/mhj/tab/MSLP/24/rmse_sd\endcsname{0.17}
\expandafter\def\csname mcresult@seed/mhj/tab/MSLP/24/skill_score_mean\endcsname{0.23}
\expandafter\def\csname mcresult@seed/mhj/tab/MSLP/24/skill_score_sd\endcsname{0.01}
\expandafter\def\csname mcresult@seed/mhj/tab/MSLP/48/dir_acc_mean\endcsname{0.779}
\expandafter\def\csname mcresult@seed/mhj/tab/MSLP/48/dir_acc_sd\endcsname{0.007}
\expandafter\def\csname mcresult@seed/mhj/tab/MSLP/48/mae_mean\endcsname{10.68}
\expandafter\def\csname mcresult@seed/mhj/tab/MSLP/48/mae_sd\endcsname{0.12}
\expandafter\def\csname mcresult@seed/mhj/tab/MSLP/48/r2_mean\endcsname{0.44}
\expandafter\def\csname mcresult@seed/mhj/tab/MSLP/48/r2_sd\endcsname{0.01}
\expandafter\def\csname mcresult@seed/mhj/tab/MSLP/48/rmse_mean\endcsname{14.75}
\expandafter\def\csname mcresult@seed/mhj/tab/MSLP/48/rmse_sd\endcsname{0.19}
\expandafter\def\csname mcresult@seed/mhj/tab/MSLP/48/skill_score_mean\endcsname{0.25}
\expandafter\def\csname mcresult@seed/mhj/tab/MSLP/48/skill_score_sd\endcsname{0.01}
\expandafter\def\csname mcresult@seed/mhj/tab/MSLP/72/dir_acc_mean\endcsname{0.756}
\expandafter\def\csname mcresult@seed/mhj/tab/MSLP/72/dir_acc_sd\endcsname{0.005}
\expandafter\def\csname mcresult@seed/mhj/tab/MSLP/72/mae_mean\endcsname{13.23}
\expandafter\def\csname mcresult@seed/mhj/tab/MSLP/72/mae_sd\endcsname{0.15}
\expandafter\def\csname mcresult@seed/mhj/tab/MSLP/72/r2_mean\endcsname{0.37}
\expandafter\def\csname mcresult@seed/mhj/tab/MSLP/72/r2_sd\endcsname{0.02}
\expandafter\def\csname mcresult@seed/mhj/tab/MSLP/72/rmse_mean\endcsname{16.94}
\expandafter\def\csname mcresult@seed/mhj/tab/MSLP/72/rmse_sd\endcsname{0.26}
\expandafter\def\csname mcresult@seed/mhj/tab/MSLP/72/skill_score_mean\endcsname{0.21}
\expandafter\def\csname mcresult@seed/mhj/tab/MSLP/72/skill_score_sd\endcsname{0.01}
\expandafter\def\csname mcresult@seed/mhj/tab/MSLP/96/dir_acc_mean\endcsname{0.748}
\expandafter\def\csname mcresult@seed/mhj/tab/MSLP/96/dir_acc_sd\endcsname{0.010}
\expandafter\def\csname mcresult@seed/mhj/tab/MSLP/96/mae_mean\endcsname{14.33}
\expandafter\def\csname mcresult@seed/mhj/tab/MSLP/96/mae_sd\endcsname{0.34}
\expandafter\def\csname mcresult@seed/mhj/tab/MSLP/96/r2_mean\endcsname{0.24}
\expandafter\def\csname mcresult@seed/mhj/tab/MSLP/96/r2_sd\endcsname{0.04}
\expandafter\def\csname mcresult@seed/mhj/tab/MSLP/96/rmse_mean\endcsname{18.39}
\expandafter\def\csname mcresult@seed/mhj/tab/MSLP/96/rmse_sd\endcsname{0.48}
\expandafter\def\csname mcresult@seed/mhj/tab/MSLP/96/skill_score_mean\endcsname{0.13}
\expandafter\def\csname mcresult@seed/mhj/tab/MSLP/96/skill_score_sd\endcsname{0.02}
\expandafter\def\csname mcresult@seed/mhj/tab/VMAX/120/dir_acc_mean\endcsname{0.759}
\expandafter\def\csname mcresult@seed/mhj/tab/VMAX/120/dir_acc_sd\endcsname{0.009}
\expandafter\def\csname mcresult@seed/mhj/tab/VMAX/120/mae_mean\endcsname{20.23}
\expandafter\def\csname mcresult@seed/mhj/tab/VMAX/120/mae_sd\endcsname{0.51}
\expandafter\def\csname mcresult@seed/mhj/tab/VMAX/120/r2_mean\endcsname{0.19}
\expandafter\def\csname mcresult@seed/mhj/tab/VMAX/120/r2_sd\endcsname{0.04}
\expandafter\def\csname mcresult@seed/mhj/tab/VMAX/120/rmse_mean\endcsname{26.43}
\expandafter\def\csname mcresult@seed/mhj/tab/VMAX/120/rmse_sd\endcsname{0.64}
\expandafter\def\csname mcresult@seed/mhj/tab/VMAX/120/skill_score_mean\endcsname{0.10}
\expandafter\def\csname mcresult@seed/mhj/tab/VMAX/120/skill_score_sd\endcsname{0.02}
\expandafter\def\csname mcresult@seed/mhj/tab/VMAX/24/dir_acc_mean\endcsname{0.682}
\expandafter\def\csname mcresult@seed/mhj/tab/VMAX/24/dir_acc_sd\endcsname{0.013}
\expandafter\def\csname mcresult@seed/mhj/tab/VMAX/24/mae_mean\endcsname{10.32}
\expandafter\def\csname mcresult@seed/mhj/tab/VMAX/24/mae_sd\endcsname{0.10}
\expandafter\def\csname mcresult@seed/mhj/tab/VMAX/24/r2_mean\endcsname{0.41}
\expandafter\def\csname mcresult@seed/mhj/tab/VMAX/24/r2_sd\endcsname{0.01}
\expandafter\def\csname mcresult@seed/mhj/tab/VMAX/24/rmse_mean\endcsname{14.62}
\expandafter\def\csname mcresult@seed/mhj/tab/VMAX/24/rmse_sd\endcsname{0.15}
\expandafter\def\csname mcresult@seed/mhj/tab/VMAX/24/skill_score_mean\endcsname{0.23}
\expandafter\def\csname mcresult@seed/mhj/tab/VMAX/24/skill_score_sd\endcsname{0.01}
\expandafter\def\csname mcresult@seed/mhj/tab/VMAX/48/dir_acc_mean\endcsname{0.747}
\expandafter\def\csname mcresult@seed/mhj/tab/VMAX/48/dir_acc_sd\endcsname{0.014}
\expandafter\def\csname mcresult@seed/mhj/tab/VMAX/48/mae_mean\endcsname{15.27}
\expandafter\def\csname mcresult@seed/mhj/tab/VMAX/48/mae_sd\endcsname{0.30}
\expandafter\def\csname mcresult@seed/mhj/tab/VMAX/48/r2_mean\endcsname{0.47}
\expandafter\def\csname mcresult@seed/mhj/tab/VMAX/48/r2_sd\endcsname{0.01}
\expandafter\def\csname mcresult@seed/mhj/tab/VMAX/48/rmse_mean\endcsname{20.79}
\expandafter\def\csname mcresult@seed/mhj/tab/VMAX/48/rmse_sd\endcsname{0.24}
\expandafter\def\csname mcresult@seed/mhj/tab/VMAX/48/skill_score_mean\endcsname{0.27}
\expandafter\def\csname mcresult@seed/mhj/tab/VMAX/48/skill_score_sd\endcsname{0.01}
\expandafter\def\csname mcresult@seed/mhj/tab/VMAX/72/dir_acc_mean\endcsname{0.720}
\expandafter\def\csname mcresult@seed/mhj/tab/VMAX/72/dir_acc_sd\endcsname{0.004}
\expandafter\def\csname mcresult@seed/mhj/tab/VMAX/72/mae_mean\endcsname{18.05}
\expandafter\def\csname mcresult@seed/mhj/tab/VMAX/72/mae_sd\endcsname{0.20}
\expandafter\def\csname mcresult@seed/mhj/tab/VMAX/72/r2_mean\endcsname{0.42}
\expandafter\def\csname mcresult@seed/mhj/tab/VMAX/72/r2_sd\endcsname{0.01}
\expandafter\def\csname mcresult@seed/mhj/tab/VMAX/72/rmse_mean\endcsname{23.13}
\expandafter\def\csname mcresult@seed/mhj/tab/VMAX/72/rmse_sd\endcsname{0.27}
\expandafter\def\csname mcresult@seed/mhj/tab/VMAX/72/skill_score_mean\endcsname{0.24}
\expandafter\def\csname mcresult@seed/mhj/tab/VMAX/72/skill_score_sd\endcsname{0.01}
\expandafter\def\csname mcresult@seed/mhj/tab/VMAX/96/dir_acc_mean\endcsname{0.719}
\expandafter\def\csname mcresult@seed/mhj/tab/VMAX/96/dir_acc_sd\endcsname{0.012}
\expandafter\def\csname mcresult@seed/mhj/tab/VMAX/96/mae_mean\endcsname{19.32}
\expandafter\def\csname mcresult@seed/mhj/tab/VMAX/96/mae_sd\endcsname{0.37}
\expandafter\def\csname mcresult@seed/mhj/tab/VMAX/96/r2_mean\endcsname{0.30}
\expandafter\def\csname mcresult@seed/mhj/tab/VMAX/96/r2_sd\endcsname{0.03}
\expandafter\def\csname mcresult@seed/mhj/tab/VMAX/96/rmse_mean\endcsname{24.49}
\expandafter\def\csname mcresult@seed/mhj/tab/VMAX/96/rmse_sd\endcsname{0.57}
\expandafter\def\csname mcresult@seed/mhj/tab/VMAX/96/skill_score_mean\endcsname{0.16}
\expandafter\def\csname mcresult@seed/mhj/tab/VMAX/96/skill_score_sd\endcsname{0.02}
\expandafter\def\csname mcresult@subjectivity/audit_agreement_n\endcsname{34}
\expandafter\def\csname mcresult@subjectivity/audit_agreement_percent\endcsname{85}
\expandafter\def\csname mcresult@subjectivity/audit_balanced_agree_n\endcsname{13}
\expandafter\def\csname mcresult@subjectivity/audit_balanced_agree_percent\endcsname{100}
\expandafter\def\csname mcresult@subjectivity/audit_balanced_n\endcsname{13}
\expandafter\def\csname mcresult@subjectivity/audit_disagreement_n\endcsname{6}
\expandafter\def\csname mcresult@subjectivity/audit_factual_agree_n\endcsname{7}
\expandafter\def\csname mcresult@subjectivity/audit_factual_agree_percent\endcsname{54}
\expandafter\def\csname mcresult@subjectivity/audit_factual_n\endcsname{13}
\expandafter\def\csname mcresult@subjectivity/audit_n\endcsname{40}
\expandafter\def\csname mcresult@subjectivity/audit_opinionated_agree_n\endcsname{14}
\expandafter\def\csname mcresult@subjectivity/audit_opinionated_agree_percent\endcsname{100}
\expandafter\def\csname mcresult@subjectivity/audit_opinionated_n\endcsname{14}
\expandafter\def\csname mcresult@subjectivity/benchmark_balanced_n\endcsname{196}
\expandafter\def\csname mcresult@subjectivity/benchmark_balanced_percent\endcsname{78.4}
\expandafter\def\csname mcresult@subjectivity/benchmark_factual_n\endcsname{40}
\expandafter\def\csname mcresult@subjectivity/benchmark_factual_percent\endcsname{16.0}
\expandafter\def\csname mcresult@subjectivity/benchmark_n\endcsname{250}
\expandafter\def\csname mcresult@subjectivity/benchmark_opinionated_n\endcsname{14}
\expandafter\def\csname mcresult@subjectivity/benchmark_opinionated_percent\endcsname{5.6}
\expandafter\def\csname mcresult@subjectivity/benchmark_regex_false_negatives\endcsname{0}
\expandafter\def\csname mcresult@subjectivity/benchmark_regex_false_positives\endcsname{0}
\expandafter\def\csname mcresult@subjectivity/benchmark_regex_hits\endcsname{14}
\expandafter\def\csname mcresult@subjectivity/corpus_n\endcsname{12{,}704}
\expandafter\def\csname mcresult@subjectivity/hybrid_balanced_correct_n\endcsname{194}
\expandafter\def\csname mcresult@subjectivity/hybrid_balanced_recall_percent\endcsname{99.0}
\expandafter\def\csname mcresult@subjectivity/hybrid_exact_n\endcsname{227}
\expandafter\def\csname mcresult@subjectivity/hybrid_exact_percent\endcsname{90.8}
\expandafter\def\csname mcresult@subjectivity/hybrid_factual_correct_n\endcsname{19}
\expandafter\def\csname mcresult@subjectivity/hybrid_factual_recall_percent\endcsname{47.5}
\expandafter\def\csname mcresult@subjectivity/hybrid_opinionated_correct_n\endcsname{14}
\expandafter\def\csname mcresult@subjectivity/hybrid_opinionated_recall_percent\endcsname{100.0}
\expandafter\def\csname mcresult@subjectivity/raw_balanced_correct_n\endcsname{194}
\expandafter\def\csname mcresult@subjectivity/raw_balanced_recall_percent\endcsname{99.0}
\expandafter\def\csname mcresult@subjectivity/raw_exact_n\endcsname{223}
\expandafter\def\csname mcresult@subjectivity/raw_exact_percent\endcsname{89.2}
\expandafter\def\csname mcresult@subjectivity/raw_factual_correct_n\endcsname{19}
\expandafter\def\csname mcresult@subjectivity/raw_factual_recall_percent\endcsname{47.5}
\expandafter\def\csname mcresult@subjectivity/raw_opinionated_correct_n\endcsname{10}
\expandafter\def\csname mcresult@subjectivity/raw_opinionated_recall_percent\endcsname{71.4}
\expandafter\def\csname mcresult@subjectivity/regex_hits\endcsname{813}
\expandafter\def\csname mcresult@subjectivity/regex_hits_percent\endcsname{6.40}
\expandafter\def\csname mcresult@temporal/ablation_text_gain/MSLP/mae_gain\endcsname{0.46}
\expandafter\def\csname mcresult@temporal/ablation_text_gain/MSLP/storm_ci_high\endcsname{0.98}
\expandafter\def\csname mcresult@temporal/ablation_text_gain/MSLP/storm_ci_low\endcsname{-0.04}
\expandafter\def\csname mcresult@temporal/ablation_text_gain/VMAX/mae_gain\endcsname{0.59}
\expandafter\def\csname mcresult@temporal/ablation_text_gain/VMAX/storm_ci_high\endcsname{1.33}
\expandafter\def\csname mcresult@temporal/ablation_text_gain/VMAX/storm_ci_low\endcsname{-0.11}
\expandafter\def\csname mcresult@temporal/common/n_obs\endcsname{834}
\expandafter\def\csname mcresult@temporal/common/n_storms\endcsname{55}
\expandafter\def\csname mcresult@temporal/matched/n_obs\endcsname{820}
\expandafter\def\csname mcresult@temporal/matched/n_storms\endcsname{55}
\expandafter\def\csname mcresult@uq/benchmark/mslp/aci/coverage_mean\endcsname{0.912}
\expandafter\def\csname mcresult@uq/benchmark/mslp/aci/coverage_mean_forest_sd\endcsname{0.000}
\expandafter\def\csname mcresult@uq/benchmark/mslp/aci/coverage_sd\endcsname{0.023}
\expandafter\def\csname mcresult@uq/benchmark/mslp/aci/coverage_sd_forest_sd\endcsname{0.000}
\expandafter\def\csname mcresult@uq/benchmark/mslp/aci/crps\endcsname{3.97}
\expandafter\def\csname mcresult@uq/benchmark/mslp/aci/crps_forest_sd\endcsname{0.00}
\expandafter\def\csname mcresult@uq/benchmark/mslp/aci/interval_score\endcsname{43.85}
\expandafter\def\csname mcresult@uq/benchmark/mslp/aci/interval_score_forest_sd\endcsname{0.00}
\expandafter\def\csname mcresult@uq/benchmark/mslp/aci/mean_width\endcsname{29.82}
\expandafter\def\csname mcresult@uq/benchmark/mslp/aci/mean_width_forest_sd\endcsname{0.00}
\expandafter\def\csname mcresult@uq/benchmark/mslp/aci/median_width\endcsname{25.76}
\expandafter\def\csname mcresult@uq/benchmark/mslp/aci/median_width_forest_sd\endcsname{0.00}
\expandafter\def\csname mcresult@uq/benchmark/mslp/block/coverage_mean\endcsname{0.917}
\expandafter\def\csname mcresult@uq/benchmark/mslp/block/coverage_mean_forest_sd\endcsname{0.000}
\expandafter\def\csname mcresult@uq/benchmark/mslp/block/coverage_sd\endcsname{0.034}
\expandafter\def\csname mcresult@uq/benchmark/mslp/block/coverage_sd_forest_sd\endcsname{0.000}
\expandafter\def\csname mcresult@uq/benchmark/mslp/block/crps\endcsname{3.99}
\expandafter\def\csname mcresult@uq/benchmark/mslp/block/crps_forest_sd\endcsname{0.00}
\expandafter\def\csname mcresult@uq/benchmark/mslp/block/interval_score\endcsname{42.77}
\expandafter\def\csname mcresult@uq/benchmark/mslp/block/interval_score_forest_sd\endcsname{0.00}
\expandafter\def\csname mcresult@uq/benchmark/mslp/block/mean_width\endcsname{29.54}
\expandafter\def\csname mcresult@uq/benchmark/mslp/block/mean_width_forest_sd\endcsname{0.00}
\expandafter\def\csname mcresult@uq/benchmark/mslp/block/median_width\endcsname{29.97}
\expandafter\def\csname mcresult@uq/benchmark/mslp/block/median_width_forest_sd\endcsname{0.00}
\expandafter\def\csname mcresult@uq/benchmark/mslp/block_norm/coverage_mean\endcsname{0.916}
\expandafter\def\csname mcresult@uq/benchmark/mslp/block_norm/coverage_mean_forest_sd\endcsname{0.000}
\expandafter\def\csname mcresult@uq/benchmark/mslp/block_norm/coverage_sd\endcsname{0.028}
\expandafter\def\csname mcresult@uq/benchmark/mslp/block_norm/coverage_sd_forest_sd\endcsname{0.000}
\expandafter\def\csname mcresult@uq/benchmark/mslp/block_norm/crps\endcsname{3.91}
\expandafter\def\csname mcresult@uq/benchmark/mslp/block_norm/crps_forest_sd\endcsname{0.00}
\expandafter\def\csname mcresult@uq/benchmark/mslp/block_norm/interval_score\endcsname{40.15}
\expandafter\def\csname mcresult@uq/benchmark/mslp/block_norm/interval_score_forest_sd\endcsname{0.00}
\expandafter\def\csname mcresult@uq/benchmark/mslp/block_norm/mean_width\endcsname{27.49}
\expandafter\def\csname mcresult@uq/benchmark/mslp/block_norm/mean_width_forest_sd\endcsname{0.00}
\expandafter\def\csname mcresult@uq/benchmark/mslp/block_norm/median_width\endcsname{24.43}
\expandafter\def\csname mcresult@uq/benchmark/mslp/block_norm/median_width_forest_sd\endcsname{0.00}
\expandafter\def\csname mcresult@uq/benchmark/mslp/clear_naive/coverage_mean\endcsname{0.897}
\expandafter\def\csname mcresult@uq/benchmark/mslp/clear_naive/coverage_mean_forest_sd\endcsname{0.001}
\expandafter\def\csname mcresult@uq/benchmark/mslp/clear_naive/coverage_sd\endcsname{0.040}
\expandafter\def\csname mcresult@uq/benchmark/mslp/clear_naive/coverage_sd_forest_sd\endcsname{0.001}
\expandafter\def\csname mcresult@uq/benchmark/mslp/clear_naive/crps\endcsname{3.86}
\expandafter\def\csname mcresult@uq/benchmark/mslp/clear_naive/crps_forest_sd\endcsname{0.00}
\expandafter\def\csname mcresult@uq/benchmark/mslp/clear_naive/interval_score\endcsname{37.51}
\expandafter\def\csname mcresult@uq/benchmark/mslp/clear_naive/interval_score_forest_sd\endcsname{0.03}
\expandafter\def\csname mcresult@uq/benchmark/mslp/clear_naive/mean_width\endcsname{24.58}
\expandafter\def\csname mcresult@uq/benchmark/mslp/clear_naive/mean_width_forest_sd\endcsname{0.08}
\expandafter\def\csname mcresult@uq/benchmark/mslp/clear_naive/median_width\endcsname{22.86}
\expandafter\def\csname mcresult@uq/benchmark/mslp/clear_naive/median_width_forest_sd\endcsname{0.07}
\expandafter\def\csname mcresult@uq/benchmark/mslp/clear_storm_block/coverage_mean\endcsname{0.913}
\expandafter\def\csname mcresult@uq/benchmark/mslp/clear_storm_block/coverage_mean_forest_sd\endcsname{0.001}
\expandafter\def\csname mcresult@uq/benchmark/mslp/clear_storm_block/coverage_sd\endcsname{0.039}
\expandafter\def\csname mcresult@uq/benchmark/mslp/clear_storm_block/coverage_sd_forest_sd\endcsname{0.001}
\expandafter\def\csname mcresult@uq/benchmark/mslp/clear_storm_block/crps\endcsname{3.87}
\expandafter\def\csname mcresult@uq/benchmark/mslp/clear_storm_block/crps_forest_sd\endcsname{0.00}
\expandafter\def\csname mcresult@uq/benchmark/mslp/clear_storm_block/interval_score\endcsname{37.72}
\expandafter\def\csname mcresult@uq/benchmark/mslp/clear_storm_block/interval_score_forest_sd\endcsname{0.02}
\expandafter\def\csname mcresult@uq/benchmark/mslp/clear_storm_block/mean_width\endcsname{26.48}
\expandafter\def\csname mcresult@uq/benchmark/mslp/clear_storm_block/mean_width_forest_sd\endcsname{0.05}
\expandafter\def\csname mcresult@uq/benchmark/mslp/clear_storm_block/median_width\endcsname{24.61}
\expandafter\def\csname mcresult@uq/benchmark/mslp/clear_storm_block/median_width_forest_sd\endcsname{0.06}
\expandafter\def\csname mcresult@uq/benchmark/mslp/enbpi_proxy/coverage_mean\endcsname{0.882}
\expandafter\def\csname mcresult@uq/benchmark/mslp/enbpi_proxy/coverage_mean_forest_sd\endcsname{0.000}
\expandafter\def\csname mcresult@uq/benchmark/mslp/enbpi_proxy/coverage_sd\endcsname{0.059}
\expandafter\def\csname mcresult@uq/benchmark/mslp/enbpi_proxy/coverage_sd_forest_sd\endcsname{0.000}
\expandafter\def\csname mcresult@uq/benchmark/mslp/enbpi_proxy/crps\endcsname{4.00}
\expandafter\def\csname mcresult@uq/benchmark/mslp/enbpi_proxy/crps_forest_sd\endcsname{0.00}
\expandafter\def\csname mcresult@uq/benchmark/mslp/enbpi_proxy/interval_score\endcsname{43.30}
\expandafter\def\csname mcresult@uq/benchmark/mslp/enbpi_proxy/interval_score_forest_sd\endcsname{0.00}
\expandafter\def\csname mcresult@uq/benchmark/mslp/enbpi_proxy/mean_width\endcsname{25.82}
\expandafter\def\csname mcresult@uq/benchmark/mslp/enbpi_proxy/mean_width_forest_sd\endcsname{0.00}
\expandafter\def\csname mcresult@uq/benchmark/mslp/enbpi_proxy/median_width\endcsname{23.48}
\expandafter\def\csname mcresult@uq/benchmark/mslp/enbpi_proxy/median_width_forest_sd\endcsname{0.00}
\expandafter\def\csname mcresult@uq/benchmark/mslp/naive_iid/coverage_mean\endcsname{0.902}
\expandafter\def\csname mcresult@uq/benchmark/mslp/naive_iid/coverage_mean_forest_sd\endcsname{0.000}
\expandafter\def\csname mcresult@uq/benchmark/mslp/naive_iid/coverage_sd\endcsname{0.039}
\expandafter\def\csname mcresult@uq/benchmark/mslp/naive_iid/coverage_sd_forest_sd\endcsname{0.000}
\expandafter\def\csname mcresult@uq/benchmark/mslp/naive_iid/crps\endcsname{3.98}
\expandafter\def\csname mcresult@uq/benchmark/mslp/naive_iid/crps_forest_sd\endcsname{0.00}
\expandafter\def\csname mcresult@uq/benchmark/mslp/naive_iid/interval_score\endcsname{42.44}
\expandafter\def\csname mcresult@uq/benchmark/mslp/naive_iid/interval_score_forest_sd\endcsname{0.00}
\expandafter\def\csname mcresult@uq/benchmark/mslp/naive_iid/mean_width\endcsname{26.91}
\expandafter\def\csname mcresult@uq/benchmark/mslp/naive_iid/mean_width_forest_sd\endcsname{0.00}
\expandafter\def\csname mcresult@uq/benchmark/mslp/naive_iid/median_width\endcsname{26.68}
\expandafter\def\csname mcresult@uq/benchmark/mslp/naive_iid/median_width_forest_sd\endcsname{0.00}
\expandafter\def\csname mcresult@uq/benchmark/vmax/aci/coverage_mean\endcsname{0.910}
\expandafter\def\csname mcresult@uq/benchmark/vmax/aci/coverage_mean_forest_sd\endcsname{0.000}
\expandafter\def\csname mcresult@uq/benchmark/vmax/aci/coverage_sd\endcsname{0.027}
\expandafter\def\csname mcresult@uq/benchmark/vmax/aci/coverage_sd_forest_sd\endcsname{0.000}
\expandafter\def\csname mcresult@uq/benchmark/vmax/aci/crps\endcsname{5.87}
\expandafter\def\csname mcresult@uq/benchmark/vmax/aci/crps_forest_sd\endcsname{0.00}
\expandafter\def\csname mcresult@uq/benchmark/vmax/aci/interval_score\endcsname{58.62}
\expandafter\def\csname mcresult@uq/benchmark/vmax/aci/interval_score_forest_sd\endcsname{0.00}
\expandafter\def\csname mcresult@uq/benchmark/vmax/aci/mean_width\endcsname{43.16}
\expandafter\def\csname mcresult@uq/benchmark/vmax/aci/mean_width_forest_sd\endcsname{0.00}
\expandafter\def\csname mcresult@uq/benchmark/vmax/aci/median_width\endcsname{38.61}
\expandafter\def\csname mcresult@uq/benchmark/vmax/aci/median_width_forest_sd\endcsname{0.00}
\expandafter\def\csname mcresult@uq/benchmark/vmax/block/coverage_mean\endcsname{0.920}
\expandafter\def\csname mcresult@uq/benchmark/vmax/block/coverage_mean_forest_sd\endcsname{0.000}
\expandafter\def\csname mcresult@uq/benchmark/vmax/block/coverage_sd\endcsname{0.036}
\expandafter\def\csname mcresult@uq/benchmark/vmax/block/coverage_sd_forest_sd\endcsname{0.000}
\expandafter\def\csname mcresult@uq/benchmark/vmax/block/crps\endcsname{5.88}
\expandafter\def\csname mcresult@uq/benchmark/vmax/block/crps_forest_sd\endcsname{0.00}
\expandafter\def\csname mcresult@uq/benchmark/vmax/block/interval_score\endcsname{57.51}
\expandafter\def\csname mcresult@uq/benchmark/vmax/block/interval_score_forest_sd\endcsname{0.00}
\expandafter\def\csname mcresult@uq/benchmark/vmax/block/mean_width\endcsname{42.16}
\expandafter\def\csname mcresult@uq/benchmark/vmax/block/mean_width_forest_sd\endcsname{0.00}
\expandafter\def\csname mcresult@uq/benchmark/vmax/block/median_width\endcsname{41.80}
\expandafter\def\csname mcresult@uq/benchmark/vmax/block/median_width_forest_sd\endcsname{0.00}
\expandafter\def\csname mcresult@uq/benchmark/vmax/block_norm/coverage_mean\endcsname{0.917}
\expandafter\def\csname mcresult@uq/benchmark/vmax/block_norm/coverage_mean_forest_sd\endcsname{0.000}
\expandafter\def\csname mcresult@uq/benchmark/vmax/block_norm/coverage_sd\endcsname{0.031}
\expandafter\def\csname mcresult@uq/benchmark/vmax/block_norm/coverage_sd_forest_sd\endcsname{0.000}
\expandafter\def\csname mcresult@uq/benchmark/vmax/block_norm/crps\endcsname{5.87}
\expandafter\def\csname mcresult@uq/benchmark/vmax/block_norm/crps_forest_sd\endcsname{0.00}
\expandafter\def\csname mcresult@uq/benchmark/vmax/block_norm/interval_score\endcsname{57.34}
\expandafter\def\csname mcresult@uq/benchmark/vmax/block_norm/interval_score_forest_sd\endcsname{0.00}
\expandafter\def\csname mcresult@uq/benchmark/vmax/block_norm/mean_width\endcsname{42.85}
\expandafter\def\csname mcresult@uq/benchmark/vmax/block_norm/mean_width_forest_sd\endcsname{0.00}
\expandafter\def\csname mcresult@uq/benchmark/vmax/block_norm/median_width\endcsname{39.46}
\expandafter\def\csname mcresult@uq/benchmark/vmax/block_norm/median_width_forest_sd\endcsname{0.00}
\expandafter\def\csname mcresult@uq/benchmark/vmax/clear_naive/coverage_mean\endcsname{0.895}
\expandafter\def\csname mcresult@uq/benchmark/vmax/clear_naive/coverage_mean_forest_sd\endcsname{0.001}
\expandafter\def\csname mcresult@uq/benchmark/vmax/clear_naive/coverage_sd\endcsname{0.043}
\expandafter\def\csname mcresult@uq/benchmark/vmax/clear_naive/coverage_sd_forest_sd\endcsname{0.001}
\expandafter\def\csname mcresult@uq/benchmark/vmax/clear_naive/crps\endcsname{5.80}
\expandafter\def\csname mcresult@uq/benchmark/vmax/clear_naive/crps_forest_sd\endcsname{0.00}
\expandafter\def\csname mcresult@uq/benchmark/vmax/clear_naive/interval_score\endcsname{54.40}
\expandafter\def\csname mcresult@uq/benchmark/vmax/clear_naive/interval_score_forest_sd\endcsname{0.06}
\expandafter\def\csname mcresult@uq/benchmark/vmax/clear_naive/mean_width\endcsname{38.05}
\expandafter\def\csname mcresult@uq/benchmark/vmax/clear_naive/mean_width_forest_sd\endcsname{0.05}
\expandafter\def\csname mcresult@uq/benchmark/vmax/clear_naive/median_width\endcsname{36.32}
\expandafter\def\csname mcresult@uq/benchmark/vmax/clear_naive/median_width_forest_sd\endcsname{0.06}
\expandafter\def\csname mcresult@uq/benchmark/vmax/clear_storm_block/coverage_mean\endcsname{0.913}
\expandafter\def\csname mcresult@uq/benchmark/vmax/clear_storm_block/coverage_mean_forest_sd\endcsname{0.001}
\expandafter\def\csname mcresult@uq/benchmark/vmax/clear_storm_block/coverage_sd\endcsname{0.040}
\expandafter\def\csname mcresult@uq/benchmark/vmax/clear_storm_block/coverage_sd_forest_sd\endcsname{0.000}
\expandafter\def\csname mcresult@uq/benchmark/vmax/clear_storm_block/crps\endcsname{5.82}
\expandafter\def\csname mcresult@uq/benchmark/vmax/clear_storm_block/crps_forest_sd\endcsname{0.00}
\expandafter\def\csname mcresult@uq/benchmark/vmax/clear_storm_block/interval_score\endcsname{54.83}
\expandafter\def\csname mcresult@uq/benchmark/vmax/clear_storm_block/interval_score_forest_sd\endcsname{0.09}
\expandafter\def\csname mcresult@uq/benchmark/vmax/clear_storm_block/mean_width\endcsname{40.95}
\expandafter\def\csname mcresult@uq/benchmark/vmax/clear_storm_block/mean_width_forest_sd\endcsname{0.11}
\expandafter\def\csname mcresult@uq/benchmark/vmax/clear_storm_block/median_width\endcsname{39.05}
\expandafter\def\csname mcresult@uq/benchmark/vmax/clear_storm_block/median_width_forest_sd\endcsname{0.12}
\expandafter\def\csname mcresult@uq/benchmark/vmax/enbpi_proxy/coverage_mean\endcsname{0.891}
\expandafter\def\csname mcresult@uq/benchmark/vmax/enbpi_proxy/coverage_mean_forest_sd\endcsname{0.000}
\expandafter\def\csname mcresult@uq/benchmark/vmax/enbpi_proxy/coverage_sd\endcsname{0.061}
\expandafter\def\csname mcresult@uq/benchmark/vmax/enbpi_proxy/coverage_sd_forest_sd\endcsname{0.000}
\expandafter\def\csname mcresult@uq/benchmark/vmax/enbpi_proxy/crps\endcsname{5.89}
\expandafter\def\csname mcresult@uq/benchmark/vmax/enbpi_proxy/crps_forest_sd\endcsname{0.00}
\expandafter\def\csname mcresult@uq/benchmark/vmax/enbpi_proxy/interval_score\endcsname{58.28}
\expandafter\def\csname mcresult@uq/benchmark/vmax/enbpi_proxy/interval_score_forest_sd\endcsname{0.00}
\expandafter\def\csname mcresult@uq/benchmark/vmax/enbpi_proxy/mean_width\endcsname{38.93}
\expandafter\def\csname mcresult@uq/benchmark/vmax/enbpi_proxy/mean_width_forest_sd\endcsname{0.00}
\expandafter\def\csname mcresult@uq/benchmark/vmax/enbpi_proxy/median_width\endcsname{38.49}
\expandafter\def\csname mcresult@uq/benchmark/vmax/enbpi_proxy/median_width_forest_sd\endcsname{0.00}
\expandafter\def\csname mcresult@uq/benchmark/vmax/naive_iid/coverage_mean\endcsname{0.905}
\expandafter\def\csname mcresult@uq/benchmark/vmax/naive_iid/coverage_mean_forest_sd\endcsname{0.000}
\expandafter\def\csname mcresult@uq/benchmark/vmax/naive_iid/coverage_sd\endcsname{0.039}
\expandafter\def\csname mcresult@uq/benchmark/vmax/naive_iid/coverage_sd_forest_sd\endcsname{0.000}
\expandafter\def\csname mcresult@uq/benchmark/vmax/naive_iid/crps\endcsname{5.87}
\expandafter\def\csname mcresult@uq/benchmark/vmax/naive_iid/crps_forest_sd\endcsname{0.00}
\expandafter\def\csname mcresult@uq/benchmark/vmax/naive_iid/interval_score\endcsname{57.01}
\expandafter\def\csname mcresult@uq/benchmark/vmax/naive_iid/interval_score_forest_sd\endcsname{0.00}
\expandafter\def\csname mcresult@uq/benchmark/vmax/naive_iid/mean_width\endcsname{39.42}
\expandafter\def\csname mcresult@uq/benchmark/vmax/naive_iid/mean_width_forest_sd\endcsname{0.00}
\expandafter\def\csname mcresult@uq/benchmark/vmax/naive_iid/median_width\endcsname{39.13}
\expandafter\def\csname mcresult@uq/benchmark/vmax/naive_iid/median_width_forest_sd\endcsname{0.00}
\expandafter\def\csname mcresult@uq/textgain/mslp/T+D/coverage_mean\endcsname{0.916}
\expandafter\def\csname mcresult@uq/textgain/mslp/T+D/interval_score\endcsname{40.15}
\expandafter\def\csname mcresult@uq/textgain/mslp/T+D/mean_width\endcsname{27.49}
\expandafter\def\csname mcresult@uq/textgain/mslp/T+D/median_width\endcsname{24.43}
\expandafter\def\csname mcresult@uq/textgain/mslp/T/coverage_mean\endcsname{0.918}
\expandafter\def\csname mcresult@uq/textgain/mslp/T/interval_score\endcsname{42.91}
\expandafter\def\csname mcresult@uq/textgain/mslp/T/mean_width\endcsname{29.70}
\expandafter\def\csname mcresult@uq/textgain/mslp/T/median_width\endcsname{26.42}
\expandafter\def\csname mcresult@uq/textgain/vmax/T+D/coverage_mean\endcsname{0.917}
\expandafter\def\csname mcresult@uq/textgain/vmax/T+D/interval_score\endcsname{57.34}
\expandafter\def\csname mcresult@uq/textgain/vmax/T+D/mean_width\endcsname{42.85}
\expandafter\def\csname mcresult@uq/textgain/vmax/T+D/median_width\endcsname{39.46}
\expandafter\def\csname mcresult@uq/textgain/vmax/T/coverage_mean\endcsname{0.917}
\expandafter\def\csname mcresult@uq/textgain/vmax/T/interval_score\endcsname{62.30}
\expandafter\def\csname mcresult@uq/textgain/vmax/T/mean_width\endcsname{45.93}
\expandafter\def\csname mcresult@uq/textgain/vmax/T/median_width\endcsname{41.88}
\expandafter\def\csname mcresult@xai/above_tolerance\endcsname{6}
\expandafter\def\csname mcresult@xai/above_tolerance_residual_max_kt\endcsname{0.7}
\expandafter\def\csname mcresult@xai/above_tolerance_residual_min_kt\endcsname{0.3}
\expandafter\def\csname mcresult@xai/completeness_verified\endcsname{33}
\expandafter\def\csname mcresult@xai/maps\endcsname{39}
\expandafter\def\csname mcresult@xai/step_cap\endcsname{8{,}192}

\usepackage{libertinus}
\usepackage{libertinust1math}
\ifdefined\pdfglyphtounicode
  \pdfglyphtounicode{u1D434}{0041}
  \pdfglyphtounicode{u1D435}{0042}
  \pdfglyphtounicode{u1D436}{0043}
  \pdfglyphtounicode{u1D437}{0044}
  \pdfglyphtounicode{u1D438}{0045}
  \pdfglyphtounicode{u1D439}{0046}
  \pdfglyphtounicode{u1D43A}{0047}
  \pdfglyphtounicode{u1D43B}{0048}
  \pdfglyphtounicode{u1D43C}{0049}
  \pdfglyphtounicode{u1D43D}{004A}
  \pdfglyphtounicode{u1D43E}{004B}
  \pdfglyphtounicode{u1D43F}{004C}
  \pdfglyphtounicode{u1D440}{004D}
  \pdfglyphtounicode{u1D441}{004E}
  \pdfglyphtounicode{u1D442}{004F}
  \pdfglyphtounicode{u1D443}{0050}
  \pdfglyphtounicode{u1D444}{0051}
  \pdfglyphtounicode{u1D445}{0052}
  \pdfglyphtounicode{u1D446}{0053}
  \pdfglyphtounicode{u1D447}{0054}
  \pdfglyphtounicode{u1D448}{0055}
  \pdfglyphtounicode{u1D449}{0056}
  \pdfglyphtounicode{u1D44A}{0057}
  \pdfglyphtounicode{u1D44B}{0058}
  \pdfglyphtounicode{u1D44C}{0059}
  \pdfglyphtounicode{u1D44D}{005A}
  \pdfglyphtounicode{u1D44E}{0061}
  \pdfglyphtounicode{u1D44F}{0062}
  \pdfglyphtounicode{u1D450}{0063}
  \pdfglyphtounicode{u1D451}{0064}
  \pdfglyphtounicode{u1D452}{0065}
  \pdfglyphtounicode{u1D453}{0066}
  \pdfglyphtounicode{u1D454}{0067}
  \pdfglyphtounicode{u1D456}{0069}
  \pdfglyphtounicode{u1D457}{006A}
  \pdfglyphtounicode{u1D458}{006B}
  \pdfglyphtounicode{u1D459}{006C}
  \pdfglyphtounicode{u1D45A}{006D}
  \pdfglyphtounicode{u1D45B}{006E}
  \pdfglyphtounicode{u1D45C}{006F}
  \pdfglyphtounicode{u1D45D}{0070}
  \pdfglyphtounicode{u1D45E}{0071}
  \pdfglyphtounicode{u1D45F}{0072}
  \pdfglyphtounicode{u1D460}{0073}
  \pdfglyphtounicode{u1D461}{0074}
  \pdfglyphtounicode{u1D462}{0075}
  \pdfglyphtounicode{u1D463}{0076}
  \pdfglyphtounicode{u1D464}{0077}
  \pdfglyphtounicode{u1D465}{0078}
  \pdfglyphtounicode{u1D466}{0079}
  \pdfglyphtounicode{u1D467}{007A}
  \pdfglyphtounicode{uni210E}{0068}
  \pdfglyphtounicode{u1D463.alt}{0076}
  \pdfglyphtounicode{u1D6E2}{0391}
  \pdfglyphtounicode{u1D6E3}{0392}
  \pdfglyphtounicode{u1D6E4}{0393}
  \pdfglyphtounicode{u1D6E5}{0394}
  \pdfglyphtounicode{u1D6E6}{0395}
  \pdfglyphtounicode{u1D6E7}{0396}
  \pdfglyphtounicode{u1D6E8}{0397}
  \pdfglyphtounicode{u1D6E9}{0398}
  \pdfglyphtounicode{u1D6EA}{0399}
  \pdfglyphtounicode{u1D6EB}{039A}
  \pdfglyphtounicode{u1D6EC}{039B}
  \pdfglyphtounicode{u1D6ED}{039C}
  \pdfglyphtounicode{u1D6EE}{039D}
  \pdfglyphtounicode{u1D6EF}{039E}
  \pdfglyphtounicode{u1D6F0}{039F}
  \pdfglyphtounicode{u1D6F1}{03A0}
  \pdfglyphtounicode{u1D6F2}{03A1}
  \pdfglyphtounicode{u1D6F3}{03F4}
  \pdfglyphtounicode{u1D6F4}{03A3}
  \pdfglyphtounicode{u1D6F5}{03A4}
  \pdfglyphtounicode{u1D6F6}{03A5}
  \pdfglyphtounicode{u1D6F7}{03A6}
  \pdfglyphtounicode{u1D6F8}{03A7}
  \pdfglyphtounicode{u1D6F9}{03A8}
  \pdfglyphtounicode{u1D6FA}{03A9}
  \pdfglyphtounicode{u1D6FB}{2207}
  \pdfglyphtounicode{u1D6FC}{03B1}
  \pdfglyphtounicode{u1D6FD}{03B2}
  \pdfglyphtounicode{u1D6FE}{03B3}
  \pdfglyphtounicode{u1D6FF}{03B4}
  \pdfglyphtounicode{u1D700}{03B5}
  \pdfglyphtounicode{u1D701}{03B6}
  \pdfglyphtounicode{u1D702}{03B7}
  \pdfglyphtounicode{u1D703}{03B8}
  \pdfglyphtounicode{u1D704}{03B9}
  \pdfglyphtounicode{u1D705}{03BA}
  \pdfglyphtounicode{u1D706}{03BB}
  \pdfglyphtounicode{u1D707}{03BC}
  \pdfglyphtounicode{u1D708}{03BD}
  \pdfglyphtounicode{u1D709}{03BE}
  \pdfglyphtounicode{u1D70A}{03BF}
  \pdfglyphtounicode{u1D70B}{03C0}
  \pdfglyphtounicode{u1D70C}{03C1}
  \pdfglyphtounicode{u1D70D}{03C2}
  \pdfglyphtounicode{u1D70E}{03C3}
  \pdfglyphtounicode{u1D70F}{03C4}
  \pdfglyphtounicode{u1D710}{03C5}
  \pdfglyphtounicode{u1D711}{03C6}
  \pdfglyphtounicode{u1D712}{03C7}
  \pdfglyphtounicode{u1D713}{03C8}
  \pdfglyphtounicode{u1D714}{03C9}
  \pdfglyphtounicode{u1D715}{2202}
  \pdfglyphtounicode{u1D716}{03F5}
  \pdfglyphtounicode{u1D717}{03D1}
  \pdfglyphtounicode{u1D718}{03F0}
  \pdfglyphtounicode{u1D719}{03D5}
  \pdfglyphtounicode{u1D71A}{03F1}
  \pdfglyphtounicode{u1D71B}{03D6}
\fi

\ifdefined\pdfmapline
\pdfmapline{=libertinust1-mathit LibertinusT1Math " libMIenc ReEncodeFont " <[libusMI.enc <fonts/LibertinusT1Math-selection.pfb}
\pdfmapline{=libertinust1-mathit-bold LibertinusT1Math " libBMIenc ReEncodeFont " <[libusBMI.enc <fonts/LibertinusT1Math-selection.pfb}
\pdfmapline{=libertinust1-mathex LibertinusT1Math " libusEXenc ReEncodeFont " <[libusEX.enc <fonts/LibertinusT1Math-selection.pfb}
\pdfmapline{=libertinust1-mathrm LibertinusT1Math " libusMRenc ReEncodeFont " <[libusMR.enc <fonts/LibertinusT1Math-selection.pfb}
\pdfmapline{=libertinust1-mathrm-bold LibertinusT1Math " libusBMRenc ReEncodeFont " <[libusBMR.enc <fonts/LibertinusT1Math-selection.pfb}
\pdfmapline{=libertinust1-mathbb LibertinusT1Math " libusBBenc ReEncodeFont " <[libusBB.enc <fonts/LibertinusT1Math-selection.pfb}
\pdfmapline{=libertinust1-mathsf LibertinusT1Math " libusSFenc ReEncodeFont " <[libusSF.enc <fonts/LibertinusT1Math-selection.pfb}
\pdfmapline{=libertinust1-mathsfi LibertinusT1Math " libusSFIenc ReEncodeFont " <[libusSFI.enc <fonts/LibertinusT1Math-selection.pfb}
\pdfmapline{=libertinust1-mathsfb LibertinusT1Math " libusSFBenc ReEncodeFont " <[libusSFB.enc <fonts/LibertinusT1Math-selection.pfb}
\pdfmapline{=libertinust1-mathsym LibertinusT1Math " libusSCRenc ReEncodeFont " <[libusSYM.enc <fonts/LibertinusT1Math-selection.pfb}
\fi

\hypersetup{hidelinks,pdftitle={Quantifying the added value of human forecaster discussions for tropical cyclone intensity prediction},pdfauthor={Ilia Azizi, Frederick Iat-Hin Tam, Marc-Olivier Boldi, Val\'{e}rie Chavez-Demoulin, Tom Beucler}}
\newif\ifshowabstract
\showabstracttrue
\newif\ifshowkeywords
\showkeywordstrue
\newcommand{\authoremailformat}{first.last@unil.ch}
\begin{document}
\begin{flushleft}
{\fontsize{19}{23}\selectfont\bfseries Quantifying the added value of human forecaster discussions for tropical cyclone intensity prediction\par}
\vspace{0.55em}
{\color{black!75}\rule{\linewidth}{0.45pt}\par}
\vspace{0.85em}
{\small Ilia Azizi\textsuperscript{1,4*} \enspace Frederick Iat-Hin Tam\textsuperscript{2,3,4} \enspace Marc-Olivier Boldi\textsuperscript{1,4} \enspace Val\'{e}rie Chavez-Demoulin\textsuperscript{1,3,4} \enspace Tom Beucler\textsuperscript{2,3,4}\par}
\vspace{0.75em}
{\small
\textsuperscript{1} Department of Operations, HEC Lausanne\\
\textsuperscript{2} Faculty of Geosciences and Environment\\
\textsuperscript{3} Expertise Center for Climate Extremes\\
\textsuperscript{4} University of Lausanne, Switzerland\\
\textsuperscript{*} Corresponding author, email format: \authoremailformat\par}
\end{flushleft}
\ifshowabstract
\noindent\textbf{Abstract}\par\vspace{0.35em}
\begingroup
\noindent
The National Hurricane Center routinely releases Tropical Cyclone Discussions that offer insights into its tropical cyclone forecasts. This study examines the added value of these discussions for intensity forecasting beyond a set of numerical predictors. For \resultvalue{dataset/storms} Atlantic and eastern North Pacific tropical cyclones from the 2002--2023 hurricane seasons, we analyze the discussion text and its combination with 13 hurricane environmental predictors from SHIPS and ERA5 10-meter wind fields, for forecasting intensity changes at lead times of 24 to 120 hours using three modeling methods. In one of the approaches, we employ an end-to-end deep learning framework for modeling the text, which shows that the discussions alone offer comparable value for 24-hour intensity forecasts to the SHIPS predictors. Adding discussion information to the SHIPS predictors in an ensemble of five such models reduces the mean absolute error of 24-hour maximum sustained wind forecasts for 2022--2023 cyclones from $\resultvalue{primary/wind_gain/left_mae}$ to $\resultvalue{primary/wind_gain/right_mae}$\,kt, with a confidence interval that excludes zero when whole storms are resampled. The ERA5 wind fields have a minor effect on mean absolute error in wind predictions, and model rankings differ across predicted variables, lead times and model types. We also provide prediction intervals for these forecasts, whose coverage falls below the 90\% target during rapid intensification. Analysis of the text shows that words related to intensity and intensity change (such as \emph{intensify}, \emph{weaken}, \emph{eyewall} and \emph{category}) are important for the model predictions. This retrospective study using reanalysis data shows the added value of the forecasters' discussions and suggests that multi-modal learning can help understand the predictive contribution of expert human judgment. The reanalysis data differ from real-time analyses, and the discussions are issued at least three hours after the valid time of the other inputs, so these results should not be interpreted operationally.
\par\endgroup\vspace{0.65em}
\fi
\ifshowkeywords
\fi
\section{Introduction}

Tropical cyclone (TC) intensity forecasting remains one of the most challenging problems in operational meteorology due to the complex processes across many scales that govern intensification and the nonlinear interactions between storms and their environment. Traditional approaches have provided valuable foundations for intensity prediction. Statistical-dynamical schemes like the Statistical Hurricane Intensity Prediction Scheme (SHIPS) use regression on environmental predictors, including sea surface temperature and wind shear, to forecast intensity change \citep{DeMaria2005}. The GFDL hurricane model has improved track and intensity forecasts through changes to initialization, model physics and ocean coupling \citep{Bender1993,Bender2007}. Discussions written by weather forecasters for individual storms provide a complementary source of information, combining observations, model guidance and expert interpretation. Such discussions inform critical decisions and assist threat communication for key National Weather Service partners \citep{Morss2022a}.

\begin{figure*}[!t]
\centering
\includegraphics[width=\textwidth]{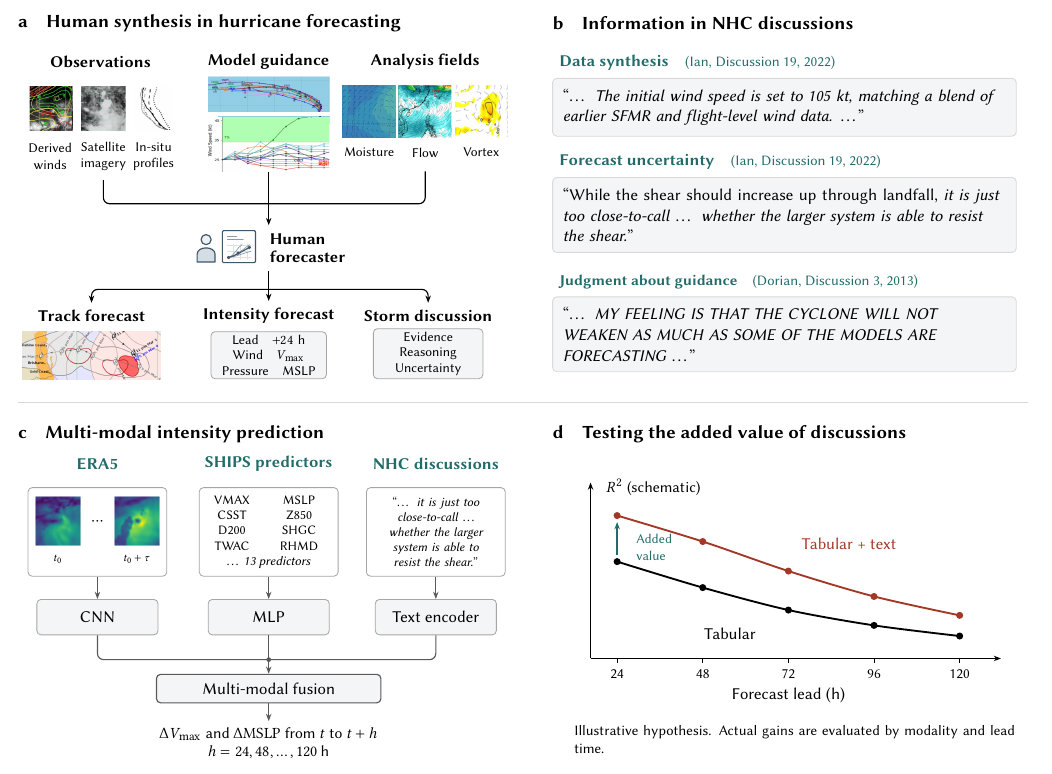}
\caption{Overview of MultiCyclone. (a) Operational forecasters synthesize observations, analysis fields and model guidance. (b) Excerpts from \href{https://www.nhc.noaa.gov/archive/2022/al09/al092022.discus.019.shtml}{Ian Discussion 19} and \href{https://www.nhc.noaa.gov/archive/2013/al04/al042013.discus.003.shtml}{Dorian Discussion 3} illustrate data synthesis, uncertainty and judgment. A typographical error in the Dorian excerpt has been corrected. (c) Two frames of ERA5 wind fields from the input sequence, SHIPS predictors and a shortened NHC excerpt feed the modality encoders for intensity-change prediction. The symbol $\tau$ denotes the elapsed time between the shown frames. (d) The study tests the added value of discussion text. The curves are schematic, not measured results.}
\label{fig:multicyclone_overview}
\end{figure*}

In recent years, machine learning (ML) has offered new ways to extract predictive patterns from large climate datasets. Deep learning models trained on satellite imagery can help locate cyclone centers and classify current intensity \citep{Do2025-yc}. Advances in large language models (LLMs) also open the possibility of integrating textual analyses, such as forecast discussions, with numerical data. LLMs can even act as regressors from examples supplied in a prompt, without additional training \citep{vacareanu2024from}. Integrating heterogeneous forms of information for a shared prediction task, known as multi-modal learning \citep{parcalabescu2021}, has inspired research on TC intensity forecasting using ML and data fusion. Recent works demonstrate diverse architectural approaches. For example, \citet{Boussioux2022} combine features learned from atmospheric reanalysis maps with environmental predictors in gradient-boosted trees, \citet{Wang2025} introduce an attention-based model trained on global TC data, \citet{Zhang2025} use a diffusion model for multi-modal forecasting, and \citet{Gupta2025} study ensemble learning with heterogeneous inputs. We therefore evaluate whether National Hurricane Center (NHC) discussion text provides measurable predictive information when combined with selected tabular predictors and ERA5 wind fields, as summarized in \autoref{fig:multicyclone_overview}. We call this approach MultiCyclone.

\subsection{Contributions}

MultiCyclone combines tabular meteorological predictors, ERA5 reanalysis winds, and storm-specific NHC discussions to predict changes in TC intensity. We consider leads from 24 to 120\,h and focus on the 24\,h lead time, for which the retained data include more storms and cases than at longer leads. The experiments assess the information carried by discussions in this retrospective setting. Our contributions are as follows:

\begin{itemize}
    \item \textbf{Forecaster discussion text as a predictive modality}. We quantify the contribution of storm-specific discussions to retrospective 24\,h intensity prediction. The primary neural models using tabular inputs and discussions achieve lower 24\,h wind MAE and higher wind $R^2$ than the corresponding models using tabular inputs alone. The size and uncertainty of the gain depend on architecture and evaluation population.
    \item \textbf{Framework for heterogeneous modalities}. We develop a modular framework using pretrained embeddings and encoders trained for each modality to combine tabular predictors, gridded wind fields, and unstructured text, with fixed modality ablations that show when additional sources help or fail to help.
    \item \textbf{Comparison across forecast leads}. We report fixed modality configurations, reference baselines, comparisons across model families, and diagnostics across forecast leads using Atlantic and eastern North Pacific storms.
    \item \textbf{Analysis of the text signal}. We examine the signal in discussion text through integrated gradients attribution, a TF-IDF lexical baseline, and a set of text perturbations. These analyses highlight sensitivity to vocabulary describing intensity and its evolution, but do not confirm the causal contribution of individual information sources and explicit forecaster judgment.
    \item \textbf{Uncertainty analysis by storm}. We evaluate post-hoc interval calibration when forecast errors are correlated within storm tracks. We compare a quantile based on pooled cases with bootstrap calibration that resamples whole storms, Adaptive Conformal Inference with delayed feedback \citep{gibbs2021adaptive}, an EnbPI-style proxy \citep{Xu2021EnbPI}, and the CLEAR framework \citep{azizi2025clear} using the same splits, with whole storms held out for evaluation (\autoref{sec:uq}). Coverage is assessed empirically, without a finite-sample guarantee under storm dependence.
\end{itemize}

\FloatBarrier
\newcommand{\yes}{{\color{mcEmbedBdr}\cmark}}%
\newcommand{\no}{{\color{mcFusionBdr}\xmark}}%
\newcommand{\pmark}{{\color{mcTemporalBdr}(\cmark)}}%
\begin{table}[!t]
    \raggedright
    \small
    \setlength{\tabcolsep}{4pt}
    \renewcommand{\arraystretch}{1.2}
    \begin{adjustbox}{max width=\linewidth}
    \begin{tabular}{@{}lccccc@{}}
        \toprule
        \textbf{Method} &
        \textbf{Tabular} &
        \textbf{Spatial fields} &
        \textbf{Text} &
        \textbf{Multi-modal} &
        \textbf{Interp.} \\
        \midrule
        SHIPS \citep{DeMaria2005} & \yes & \no & \no & \no & \pmark \\
        GFDL \citep{Bender1993,Bender2007} & \yes & \yes & \no & \no & \no \\
        CNN-based \citep{Chen2019,Combinido2018} & \no & \yes & \no & \no & \no \\
        HurriCast \citep{Boussioux2022} & \yes & \yes & \no & \yes & \pmark \\
        MSCAR \citep{Wang2025} & \yes & \yes & \no & \yes & \no \\
        MTCIE \citep{Song2024MTCIE} & \no & \yes & \no & \no & \no \\
        \textbf{MultiCyclone} & \yes & \yes & \yes & \yes & \yes \\
        \bottomrule
    \end{tabular}
    \end{adjustbox}
    \caption{Comparison of tropical cyclone intensity estimation and forecasting methods. Spatial fields include satellite imagery, gridded reanalysis or numerical model analyses, depending on the study. The input categories summarize different approaches rather than a common architecture. {\color{mcEmbedBdr}\cmark}~= fully supported, {\color{mcTemporalBdr}(\cmark)}~= partially supported, and {\color{mcFusionBdr}\xmark}~= not supported. MultiCyclone combines all three modalities and includes local attribution and analyses of sensitivity to text (Interp.).}
    \label{tab:tc_methods_comparison}
\end{table}

\section{Related Work}

The work most relevant to this study falls into four areas, covering established intensity forecasting, machine learning with numerical and image inputs, meteorological text models, and uncertainty quantification. We review these areas to identify what storm-specific forecast discussions may add to the numerical inputs used here.

\subsection{Traditional TC Intensity Prediction Models}
Early intensity forecasting relied on statistical and statistical-dynamical approaches. For example, SHIPS employs multiple regression with environmental predictors such as sea surface temperature, vertical wind shear, and storm structure parameters to forecast intensity changes \citep{DeMaria2005}. Successive versions incorporating new predictors and updated datasets showed significant skill out to 72 hours in the Atlantic compared to climatology and persistence baselines, while dynamical forecast systems based on numerical weather prediction models have evolved substantially over recent decades. The GFDL hurricane model lineage, beginning with pure atmospheric dynamical cores and progressing through ocean coupling, has demonstrated the value of improved physics parameterizations and increased resolution \citep{Bender1993, Bender2007}. Modern operational systems like HWRF and HAFS continue this evolution, incorporating advanced vortex initialization schemes and high-resolution nested grids. Together, these traditional methods provide essential baselines and remain operationally important, but the substantial cost of running these models motivates exploration of data-driven models that can more flexibly incorporate diverse observations, akin to the use of data assimilation in dynamical forecast systems.

\subsection{Machine Learning for TC Forecasting}
Machine learning studies differ in whether they use a single data source or combine complementary modalities. The distinction matters here because our comparisons isolate the contribution of discussion text within a shared evaluation design.

\subsubsection{Uni-modal Approaches}
Machine learning approaches are applied to TC forecasting using tabular or image data. Gradient boosting machines (e.g., XGBoost) trained on environmental variables can capture nonlinear relationships and reduce intensity forecast errors in retrospective tests \citep{Jin2019}. Convolutional neural networks (CNNs) trained directly on satellite, infrared or microwave images of cyclones have achieved intensity estimates with error comparable to or better than traditional methods \citep{Combinido2018, Chen2019}. \citet{Do2025-yc} use geostationary satellite images to jointly locate cyclone centers and classify current intensity. Their task concerns storm monitoring. A separate CNN approach uses reanalysis fields for intensity forecasting \citep{Xu2022}.

\subsubsection{Multi-modal Approaches}

Recent TC studies combine different numerical or spatial inputs. \citet{Gupta2025} study ensembles with heterogeneous datasets, while \citet{Zhang2025} introduce a multi-modal diffusion model. HurriCast \citep{Boussioux2022} encodes atmospheric reanalysis maps and combines the learned representation with tabular predictors in a gradient-boosted tree. These studies motivate testing whether a separately produced source of information, the NHC discussion, adds value to a numerical model.

Other models combine images and environmental data directly. MSCAR uses causal autoregressive attention to forecast intensity from satellite and reanalysis inputs \citep{Wang2025}. TropiCycloneNet introduces a multi-modal dataset and a generative forecast model that combines images with environmental and temporal features \citep{Huang2025}. TCP-Diffusion instead predicts cyclone-associated rainfall from past precipitation and environmental fields \citep{Huang2025tcpdiffusion}. MMSTN jointly forecasts track and intensity \citep{Huang2022MMSTN}, whereas MT-GN estimates current intensity through classification and regression from satellite imagery \citep{Zhao2024MTGN}. The hybrid model of \citet{You2025} combines rainfall images with SHIPS predictors for intensity forecasting.

\subsection{Text Modality}
While most forecasting models focus on numerical or image data, there is increasing exploration of textual information in weather prediction. This includes not only numerical weather bulletins and structured meteorological text, but also expert discussions, news, and public reports. Text enters weather research as a benchmark target, a source of predictive features, or a direct input to a language model. Our focus is the information carried by forecasters' storm-specific prose.

\subsubsection{Weather Language Models and Benchmarks}

Text often enters indirectly, as a benchmark target paired with weather data. \citet{Ma2024WeatherQA} pair plots of mesoscale analyses, surface observations and radar reflectivity with Storm Prediction Center discussions, and document limitations of the tested vision-language models through benchmark tasks and expert review. \citet{shinde2026wxc} present WxC-Bench, a multi-modal benchmark spanning six downstream tasks including both TC monitoring and natural language forecast generation from NOAA Storm Prediction Center reports. \citet{Chen2025ClimateIQA} construct ClimateIQA, a meteorological visual question answering dataset comprising 762,120 instruction samples paired with reanalysis-derived heatmaps for extreme weather anomaly analysis. Similarly, CLLMate \citep{Li2025CLLMate} aligns environmental news articles with reanalysis data and has been used to evaluate 32 multi-modal LLMs on weather and climate event forecasting, revealing both strengths and persistent gaps.

Text is also used directly in some weather-related research tasks. Aquilon \citep{Varambally2025Aquilon} combines language queries with encoded weather fields to study multi-modal weather reasoning. Knowledge-guided prompt learning has also been studied for TC intensity estimation \citep{Li2025KGPL}. Natural language processing has also been used to extract structured information from weather text. For example, \citet{Lai2022NLPFloods} geocode historical storms, floods, and adaptation measures from newspaper reports. However, these tasks differ substantially from the focus of this study, which uses multi-modal learning to extract potential human insights from NHC forecaster discussions.

\subsubsection{LLMs as Regressors}

Recent studies have demonstrated that LLMs can serve as effective regressors through in-context learning, eliminating the need for task-specific fine-tuning. In-context learning for numeric prediction is when pretrained LLMs are prompted with examples to regress outputs. These approaches view the LLM itself as a text-to-text regression engine \citep{Aliakbar2025}. \citet{vacareanu2024from} show that LLMs can equal or outperform traditional regression models on various prediction tasks when provided with appropriate in-context examples. \citet{tang2025llmregressors} study LLM embeddings as features for regression, finding that larger models are not always better. The OmniPred framework extends this concept \citep{Song2024OmniPred}, training language models as universal regressors capable of precise numerical prediction across diverse domains using textual representations of mathematical parameters. A known limitation of LLM regressors is that discrete number tokenization and cross entropy over tokens do not directly encode numerical distance. Continuous numerical representations \citep{Golkar2024xVa} and losses that account for numerical distance \citep{Zausinger2025NTL} address these limitations, and therefore performance depends on the model, representation and task.

\subsubsection{Interpreting LLMs}
We review interpretation methods because a central question here is which parts of a discussion the text models respond to. Interpreting which input features drive predictions in transformer-based models remains an active research challenge, with methods broadly falling into three categories: attention-based, gradient-based, and perturbation-based approaches. Attention weights are often used to inspect which tokens a model attends to. Whether they provide a faithful explanation depends on the question being asked and on checks that connect the weights to model behavior \citep{Wiegreffe2019Attention,Serrano2019Attention,Bastings2020Elephant}. Integrated gradients \citep{Sundararajan2017IG} accumulate gradients along a path from a baseline to the input. The method was developed to satisfy sensitivity and implementation invariance, and its attributions sum to the output change from that baseline. Perturbation-based approaches measure how the output changes when parts of the input are masked or removed. A complementary line of work, mechanistic interpretability, aims to reverse-engineer the internal computations of transformer models by identifying interpretable circuits, features in residual streams, and neuron-level roles \citep{Rai2024MechInterp}. While these techniques offer deeper insight into \textit{how} models process information internally, they remain nascent for domain-specific regression tasks such as meteorological prediction.

\subsection{Uncertainty Quantification}
Intensity forecasts also require an account of predictive uncertainty. We distinguish general interval methods from approaches evaluated specifically for tropical cyclones.

\subsubsection{General Approaches}
Split conformal prediction provides marginal coverage under exchangeability of calibration and test scores \citep{vovk2005algorithmic}. For models with several input types, \citet{azizi2026modalitycalibration} studies calibration based on disagreement among modality predictors or on which inputs are available. This is relevant when discussions are missing, although its exchangeability assumptions do not directly cover dependent cases within storms. Ensemble Batch Prediction Intervals (EnbPI) use bootstrap ensemble residuals to form sequential intervals under assumptions on the time series error process \citep{Xu2021EnbPI}. Those results do not directly apply to our residual proxy based on a fixed ensemble. Functional Generative Networks use learned model perturbations to generate joint ensemble weather forecasts while training against marginal continuous ranked probability scores \citep{Alet2025FGN}. CLEAR \citep{azizi2025clear} calibrates regression intervals that combine aleatoric and epistemic uncertainty, and \citet{azizi2026matchabilityllms} applies a related separation to open-ended answers of language models, with conformal sets calibrated for groups of questions identified before generation. These methods address different prediction and dependence settings.

\subsubsection{Uncertainty Quantification for TC Forecasting}
For TC uncertainty quantification, researchers have studied dynamical ensembles, neural distribution models, and conformal methods. The reliability of dynamical ensemble guidance varies by case and verification setting \citep{bachmann2021validation}. MTCIE predicts both a mean and a standard deviation for intensity from satellite imagery \citep{Song2024MTCIE}, while \citet{fernandez2025predicting} estimate uncertainty in track forecasts with a neural network. Conformal methods have been applied to TC track \citep{Meng2024} and intensity \citep{Chen2025}. Dependence within storms and shifts between seasons make the exchangeability required for split-conformal coverage uncertain here, which motivates the empirical storm-level analysis in Section~\ref{sec:uq}.

\section{Data}
Our regional dataset uses TCBench storm identifiers and forecast structure \citep{gomez2024tcbench}, with the 13 selected predictors and initial intensity read from the SHIPS developmental data at the initial time. These data are compiled after each season from the NHC best track and model analyses. We add ERA5 reanalysis winds and NHC discussions for this study. The discussions are not part of the cited TCBench release. The dataset covers the Atlantic and eastern North Pacific basins from 2002 to 2023. It contains \resultvalue{dataset/storms} storms (\resultvalue{dataset/atlantic_storms} Atlantic and \resultvalue{dataset/pacific_storms} eastern North Pacific) and \resultvalue{dataset/observations} cases at 6\,h intervals, one per storm and synoptic time. The selected SHIPS predictors and ERA5 wind frames cover all these storms. Section~\ref{sec:curation} describes how discussions are matched to cases and how many cases have a match.

\subsection{Modalities}

\paragraph{Tabular Data (SHIPS)}
The SHIPS dataset provides 13 meteorological variables derived for each case and summarized in \autoref{tab:variables}. These predictors are grouped into four categories, comprising storm characteristics (e.g., \texttt{VMAX}, \texttt{MSLP}), environmental fields (e.g., \texttt{CSST}, \texttt{Z850}), thermodynamic variables (e.g., \texttt{EPOS}, \texttt{RHMD}), and dynamic variables (e.g., \texttt{D200}, \texttt{SHGC}). Together, these 13 predictors capture both the internal state of the storm and the environmental conditions influencing its intensity evolution. The pairwise Pearson correlations among these variables are shown in \autoref{app:tab_correlations} (\autoref{fig:tabular_correlations}). As expected, $V_{\max}$ and MSLP are strongly anti-correlated, while most environmental predictors are only weakly correlated with one another.

\begin{table}[!htbp]
\centering
\renewcommand{\arraystretch}{1.1}
\caption{SHIPS predictors. Definitions, averaging radii and stored units follow the \href{https://www.aoml.noaa.gov/ftp/hrd/kaplan/ships/ships_predictor_file_2022.pdf}{NOAA predictor specification}. Model inputs are standardized using the training split.}
\begin{tabularx}{\columnwidth}{lX}
\toprule
\textbf{Variable} & \textbf{Description} \\ 
\midrule
\texttt{VMAX} & Maximum sustained surface wind (kt) \\
\texttt{MSLP} & Minimum sea-level pressure (hPa) \\
\texttt{lat} & Latitude (degrees north) \\
\texttt{lon} & Longitude (degrees east, negative westward) \\
\texttt{CSST} & Climatological sea-surface temperature (0.1$^\circ$C) \\
\texttt{Z850} & 850 hPa vorticity, 0--1000\,km radius ($10^{-7}$\,s$^{-1}$) \\
\texttt{D200} & 200 hPa divergence, 0--1000\,km radius ($10^{-7}$\,s$^{-1}$) \\
\texttt{EPOS} & Mean positive lifted-parcel minus environment $\theta_e$, 200--800\,km radius (0.1\,K) \\
\texttt{RHMD} & Mean 700--500 hPa relative humidity, 200--800\,km radius (\%) \\
\texttt{TWAC} & Mean symmetric 850 hPa tangential wind, 0--600\,km radius (0.1\,m\,s$^{-1}$) \\
\texttt{G200} & Symmetric-vortex temperature perturbation at 200 hPa, 200--800\,km radius (0.1\,K) \\
\texttt{TADV} & Mean 850--700 hPa temperature advection, 0--500\,km radius ($10^{-6}$\,K\,s$^{-1}$) \\
\texttt{SHGC} & Generalized multilevel wind shear, vortex removed, 0--500\,km radius (0.1\,kt) \\
\bottomrule
\end{tabularx}
\label{tab:variables}
\end{table}

\begin{figure}[!htbp]
\centering
\includegraphics[width=\linewidth]{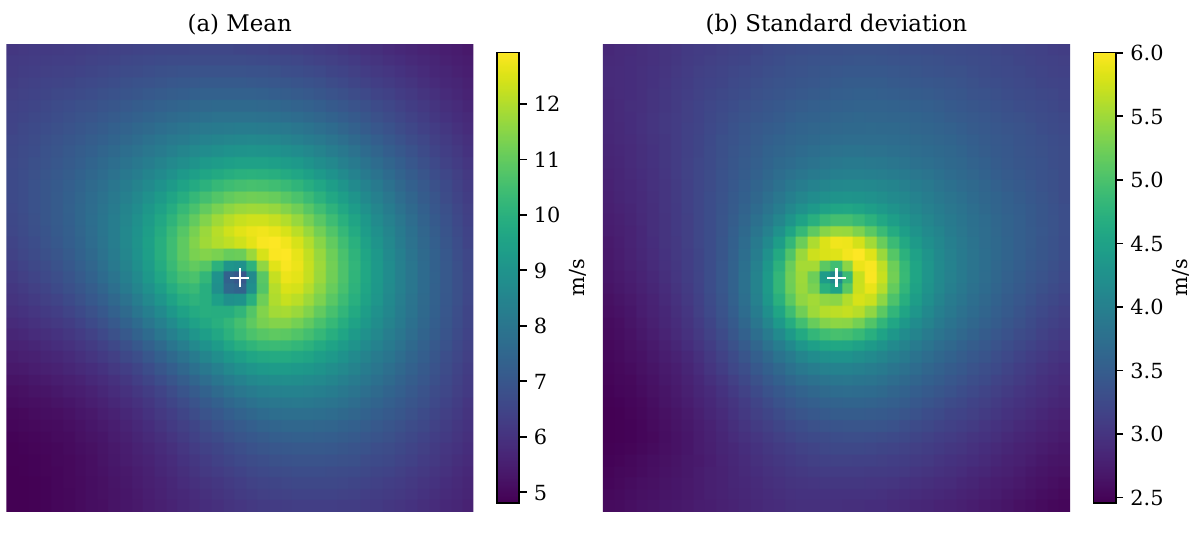}
\caption{Mean (a) and standard deviation (b) of storm-centered ERA5 10\,m wind speed, $\sqrt{u_{10}^2+v_{10}^2}$, across \resultvalue{era5/n_observations} training cases from \resultvalue{era5/n_storms} storms. Each case has equal weight. The cross marks the center of the $41\times41$ grid. Individual zero-valued grid points are retained. No entirely zero-padded frames occur among these cases.}
\label{cyclone:center}
\end{figure}

\paragraph{Reanalysis Data (ERA5)} We use ERA5 reanalysis data \citep{era5_2020} at 0.25\textdegree\ $\times$ 0.25\textdegree\ resolution, extracting a 41$\times$41 grid-point box (about 10\textdegree\ $\times$ 10\textdegree) centered on each storm location. The image input contains the $u$ and $v$ components of the 10\,m wind ($u_{10}$, $v_{10}$). These two fields retain wind direction and speed within the chosen spatial crop. \autoref{sec:magdir_comparison} gives the formulas that link these components to wind speed and direction. We use only these two 10\,m wind components and omit potentially useful fields (e.g., humidity, temperature, and upper-level winds) to avoid redundancy with the SHIPS predictors. The data pipeline handles the ERA5 variable names used before 2022 (10u, 10v) and from 2022 onward (u10, v10). \autoref{cyclone:center} summarizes the storm-centered wind magnitude in the training data. ERA5 is produced retrospectively through atmospheric modeling and data assimilation, so these fields should not be read as real-time analysis inputs.

\paragraph{Textual Data}
The NHC publishes routine Tropical Weather Discussions for the Atlantic basin (TWDAT, \autoref{fig:weather_discussion_example}) and Tropical Cyclone Discussions for individual storms after they form (\autoref{fig:storm_discussion_example}). Early experiments with TWDAT were not promising, so we used the storm-specific discussions instead, without developing a systematic TWDAT ablation. Storm-specific discussions contain observations, guidance summaries, intensity forecasts, and forecaster interpretations. \autoref{fig:storm_discussion_example} shows one explicit expression of judgment, but our experiments do not isolate the value of such statements. The matching rule is described below, and \autoref{app:missingness} examines sensitivity to discussion availability.

\begin{figure}[!htbp]
  \begin{mdframed}[style=DashedBox]
    \footnotesize
      \begin{Verbatim}[breaklines=true]
TROPICAL WEATHER DISCUSSION
NWS NATIONAL HURRICANE CENTER MIAMI FL 
205 PM EDT WED JUL 24 2013
...
GULF OF MEXICO
SURFACE RIDGING IS OVER THE GULF OF MEXICO...
...
CARIBBEAN SEA
A TROPICAL WAVE IS S OF THE CAYMAN ISLANDS WITH NO SIGNIFICANT...
...
ATLANTIC OCEAN
A 1019 MB HIGH IS OVER THE W ATLANTIC NEAR 28N64W...
...
      \end{Verbatim}
  \end{mdframed}
  \caption{Tropical Weather Discussion, 24 July 2013. For illustration only, excluded from reported results. Ellipses mark omissions.
  Source \href{https://www.nhc.noaa.gov/archive/text/TWDAT/2013/TWDAT.201307241749.txt}{NHC (2013)}.}
  \label{fig:weather_discussion_example}
\end{figure}

\begin{figure}[!htbp]
  \begin{mdframed}[style=DashedBox]
    \footnotesize
      \begin{Verbatim}[breaklines=true,commandchars=\\\{\}]
TROPICAL STORM DORIAN DISCUSSION NUMBER   3
NWS NATIONAL HURRICANE CENTER MIAMI FL
500 PM EDT WED JUL 24 2013
...

THE INITIAL MOTION ESTIMATE IS 285/17 KT. DORIAN IS EXPECTED TO REMAIN SOUTH OF AND MOVE WEST-NORTHWESTWARD TO WESTWARD IN TANDEM WITH A STRONG  SUBTROPICAL RIDGE LOCATED WELL TO THE NORTH. THE NHC MODEL GUIDANCE CONTINUES  TO REMAIN TIGHTLY PACKED AND IN EXCELLENT AGREEMENT ON THIS SCENARIO. [...]
GIVEN THAT DORIAN WILL BE MOVING OVER COOLER WATER AT A BRISK 18 KT FOR THE NEXT 24-36 HOURS...\textcolor{blue}{MY FEELING IS THAT THE CYCLONE WILL NOT WEAKEN AS MUCH AS SOME OF THE MODELS ARE FORECASTING}...AND AS A RESULT...THE OFFICIAL TRACK FORECAST LEANS TOWARD THE SLOWER GFS...UKMET...AND CANADIAN MODEL SOLUTIONS.

...
      \end{Verbatim}
  \end{mdframed}
  \caption{Excerpt from Tropical Storm Dorian Discussion 3 (typographical error corrected).
  Source \href{https://www.nhc.noaa.gov/archive/2013/al04/al042013.discus.003.shtml}{National Hurricane Center (2013)}.}
  \label{fig:storm_discussion_example}
\end{figure}

\subsection{Curation}\label{sec:curation}

\paragraph{Temporal Alignment} Since different modalities are recorded at different times, we specify their timing explicitly. For each case at synoptic time $t$, the matching pipeline first searches for storm-specific discussions whose issue times lie within 1\,h of $t{+}6$\,h. Issue times are read from the header printed in each discussion. Routine discussions for synoptic time $t$ are issued near $t{+}3$\,h, so this window can contain only special discussions. The pipeline prefers a discussion later than $t{+}6$\,h within this window and otherwise takes the closest one. If this search has no candidate, it chooses the closest discussion in the archive, taking the earlier one when two are equally close. Most matches arise from this step and are routine discussions issued near $t{+}3$\,h. The pipeline then discards selections outside $[t,t{+}7\,\mathrm{h}]$, without trying another candidate. Consequently, a case can remain unmatched even when another discussion would satisfy the final time window. This matching is held fixed across the model comparisons. In the matched data, $\resultvalue{dataset/discussion_3h_percent}\%$ of issue times are exactly $t{+}3$\,h and all offsets lie between $t{+}\resultvalue{dataset/discussion_delay_min_hours}$\,h and $t{+}\resultvalue{dataset/discussion_delay_max_hours}$\,h. The matches later than $t{+}3$\,h are special discussions. The discussion therefore has a later information set than the time-$t$ numerical inputs. This is a retrospective comparison, not an operational forecast using the same information at issue time (\autoref{sec:limitations}). This matching succeeds for \resultvalue{dataset/matched} cases, or \resultvalue{dataset/matched_percent}\% of the dataset. Each of the \resultvalue{dataset/storms} storms is matched to at least one discussion, because storms without any match were removed when the dataset was built. That step used the earlier file-name times and left out 56 Atlantic storms (1,176 training and validation cases, none in the test seasons), most of whose cases the printed issue times would match. Coverage is uneven across cases within storms. \resultvalue{dataset/fully_matched_storms} storms have a matched discussion for every one of their \resultvalue{dataset/fully_matched_observations} cases. The remaining \resultvalue{dataset/partially_matched_storms} storms span \resultvalue{dataset/partially_matched_observations} cases, of which \resultvalue{dataset/partially_matched_reports} are matched and \resultvalue{dataset/unmatched} are not. An unmatched case means that no discussion was matched in the collected archive under this rule, not necessarily that no discussion was issued. For fused models, an unmatched discussion is represented by a zero text embedding during training and inference. The neural discussion-only model omits unmatched rows, while the hybrid and temporal text-only regressors keep them with a zero vector. We therefore distinguish the full fused-model population from the matched-discussion population when comparing models, and assess the missingness channel separately.

\paragraph{Text Processing} We remove the terminal \texttt{FORECAST POSITIONS AND MAX WINDS} section from storm-specific discussions. This excludes the structured official forecast table, except in two training discussions (\autoref{app:text_preprocessing}), but the remaining narrative may still contain numerical intensities and forecasts. Table removal alone does not establish the absence of information leakage. The end-to-end neural model loader trims the bulletin header to the first datetime line, normalizes whitespace, converts text to lowercase, and tokenizes it with the RoBERTa tokenizer. We retain up to the first 510 content tokens of each cleaned discussion, plus two special tokens. An exploratory inference check using length-weighted averaging of embeddings from chunks covering the full discussion showed no clear improvement. The hybrid and temporal pipelines use the \texttt{Qwen3-Embedding-0.6B} sentence encoder with its own text-cleaning procedure and the same forecast-section cutoff (\autoref{sec:hybrid}, \autoref{sec:temporal}). Both paths use the bounded discussion matching described above.

\paragraph{Normalization} Numerical tabular predictors are standardized using feature statistics from the training set, with the family-specific missing-value handling described in \autoref{app:data_processing}. ERA5 images are normalized independently per channel ($u_{10}$, $v_{10}$), using the means and standard deviations of nonzero frames from training storms. Frames that are entirely zero remain zero. Validation and test inputs reuse these statistics. The neural models standardize each intensity-change target using statistics from the training set for the multi-output MSE loss, and predictions are transformed back to knots or hPa for evaluation. The hybrid XGBoost and Ridge regressors fit the intensity-change targets directly in physical units. For fixed predictions, inversion of a common affine target transformation leaves target-specific $R^2$ unchanged. Scaling targets during neural training nevertheless changes the relative weighting of the two losses.

\section{Methodology}
This section describes our approach to multi-modal tropical cyclone intensity prediction. We begin by formalizing the prediction task, then present three complementary modeling frameworks. These comprise a hybrid architecture combining deep learning encoders with gradient boosting, an end-to-end neural network with learnable fusion, and a temporal extension that models storm evolution over consecutive timesteps. We conclude with training strategies common to all approaches.

\subsection{Problem Formulation}\label{sec:problem}
For each case at synoptic time $t$, we use the available tabular (\textbf{T}), image (\textbf{I}) and discussion text (\textbf{D}) features to predict the \emph{change} in storm intensity after $h$ hours, where $h \in \{24, 48, 72, 96, 120\}$ and cases occur at 6\,h intervals. For the full multi-modal ablation we evaluate all \textbf{uni-modal} (T, I, D), \textbf{bi-modal} (T+I, T+D, I+D), and \textbf{tri-modal} (T+I+D) combinations. The prediction targets are
\begin{equation*}
\Delta V_{\max,h}(t) = V_{\max}(t+h) - V_{\max}(t), \qquad
\Delta \mathrm{MSLP}_{h}(t) = \mathrm{MSLP}(t+h) - \mathrm{MSLP}(t),
\end{equation*}
Here $V_{\max}$ is maximum sustained wind and MSLP is minimum sea-level pressure. Models are trained and scored on these changes. Absolute future intensity is reconstructed by adding the predicted change to the dataset intensity at time $t$. Models containing tabular inputs also receive $V_{\max}(t)$ and $\mathrm{MSLP}(t)$ among their numerical predictors. We emphasize 24\,h results because longer leads have fewer valid future intensities. A forecast initialized when the discussion was issued would require a new timing and verification protocol.

\subsection{Hybrid Approach}\label{sec:hybrid}

\begin{figure}[!htbp]
\centering
\resizebox{\linewidth}{!}{\input{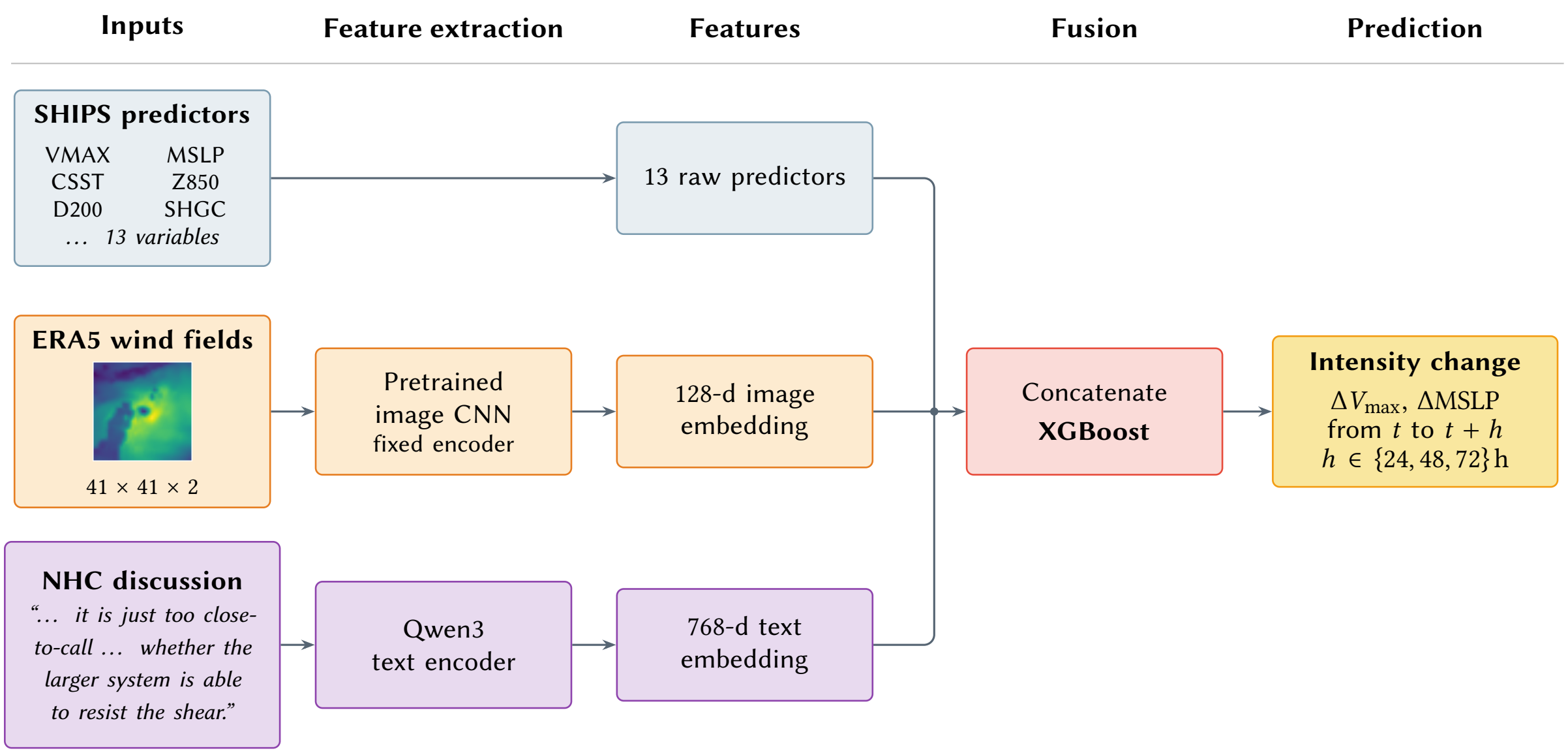}}
\caption{Hybrid architecture. The 13 SHIPS predictors enter XGBoost directly. A CNN pretrained on ERA5 frames from the training years alone, using next-frame prediction and reconstruction, supplies fixed 128-dimensional image embeddings. The frozen Qwen3 sentence encoder supplies 768 retained text dimensions. These features are concatenated for XGBoost prediction. The separate T+I neural model uses both a CNN and a tabular MLP (\autoref{fig:architecture}). That MLP is not part of this hybrid encoder. Outputs give changes in $V_{\max}$ and MSLP over lead $h$. The shortened discussion excerpt is from Ian Discussion 19.}
\label{fig:hybrid_architecture}
\end{figure}

Discussion text, reanalysis fields, and tabular predictors are not jointly available for every case. The hybrid approach in \autoref{fig:hybrid_architecture} trains or supplies separate encoders, then fits a regressor to their fixed-length outputs. Because the encoders are fixed, each combination of inputs only requires refitting the regressor. Raw XGBoost feature importance scores over 13 tabular variables and much larger image and text embedding blocks are not comparable measures of modality contribution. Correlated predictors can also distort feature importance~\citep{Nicodemus2010}. We therefore assess added predictive value by fitting the model with and without a modality and comparing errors on the test seasons, not by raw feature ranks. With correlated modalities this measures the value added beyond the other inputs, not the total information in a modality.

Inspired by~\citet{Boussioux2022}, we pretrain a CNN encoder on ERA5 wind fields with a self-supervised objective combining next-frame embedding prediction and frame reconstruction (\autoref{sec:pretrain_cnn_details}), and use the trained encoder to produce 128-dimensional image embeddings. We repeat the full hybrid training process with five random seeds. In each run, we first train a CNN, then use it to extract image features for the regressor without further updating the CNN. We also explored DINOv2~\citep{oquab2024dinov2}, pretrained on natural images, but did not retain it for the reported experiments (\autoref{sec:encoder_comparison}). For text encoding we use the pretrained \texttt{Qwen3-Embedding-0.6B}~\citep{qwen3embedding} sentence encoder without fine-tuning, retaining the first 768 dimensions of its cached output. An XGBoost gradient boosting model then fuses the raw 13 tabular SHIPS predictors, the image embeddings, and the text embeddings into a single intensity prediction.

\subsection{Deep Learning Approach}\label{sec:e2e}

\begin{figure}[!htbp]
\centering
\resizebox{\linewidth}{!}{\input{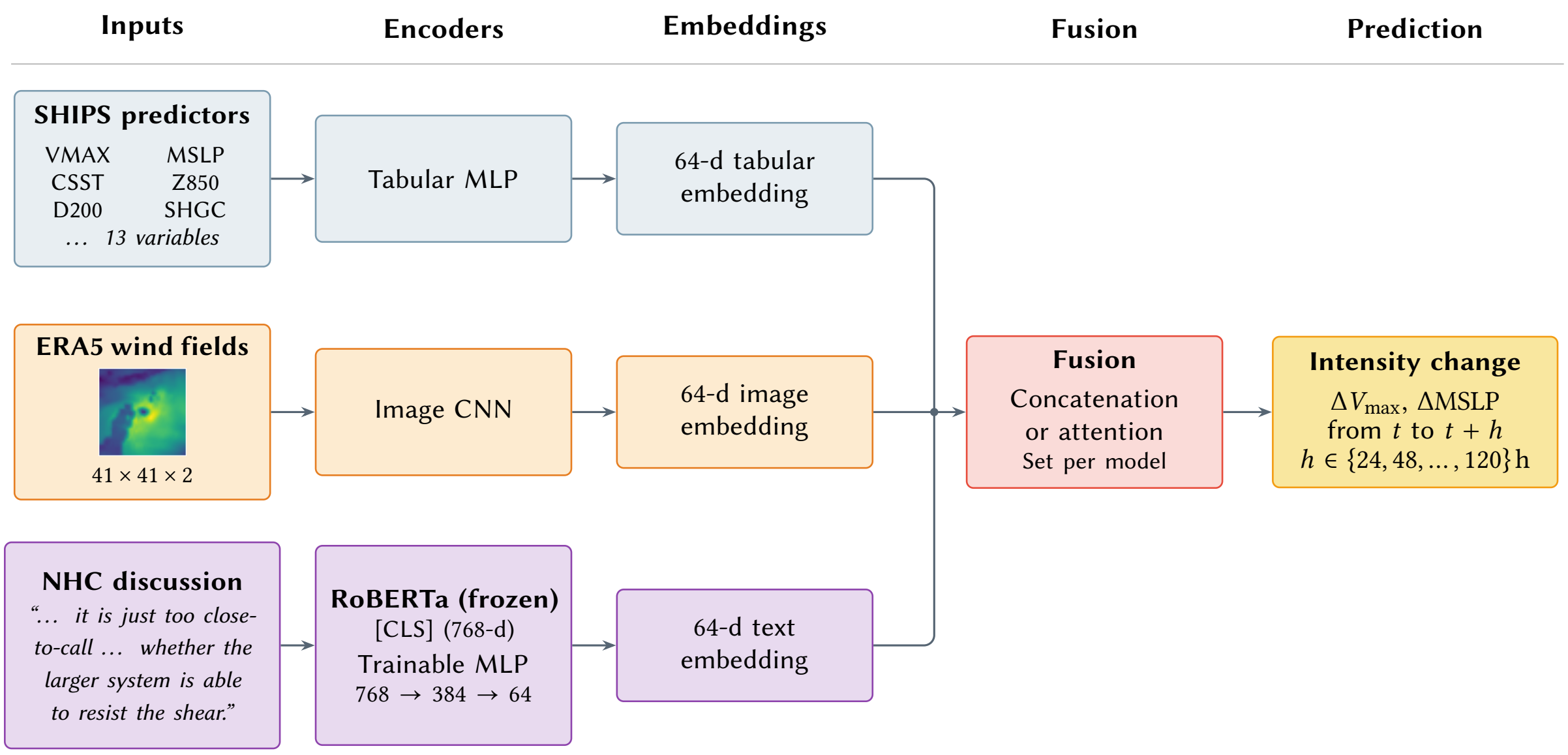}}
\caption{Neural architecture. Separate tabular, image and text branches each produce a 64-dimensional embedding before concatenation (Concat) or multi-head attention (MHA). The T+I bi-modal CNN+MLP configuration uses the image and tabular branches. The text branch is added for T+I+D. The RoBERTa backbone is frozen during fusion training. Its projection is trainable. Outputs give changes in $V_{\max}$ and MSLP over forecast lead $h$. The shortened discussion excerpt is from Ian Discussion 19.}
\label{fig:architecture}
\end{figure}

We also propose an end-to-end neural network approach in \autoref{fig:architecture}, optimizing the trainable modality branches and fusion jointly through backpropagation while using an externally supplied, frozen text backbone. Each modality is encoded into a shared 64-dimensional embedding space $\mathbb{R}^{64}$ before fusion. The \textbf{tabular encoder} processes the 13 SHIPS predictors through a three-layer MLP ($13 \to 128 \to 64 \to 64$) with LeakyReLU activations, layer normalization, and dropout ($p{=}0.3$). The \textbf{image encoder} extracts spatial features from the ERA5 wind fields ($\mathbf{x}_{\text{img}} \in \mathbb{R}^{2 \times 41 \times 41}$) using a CNN with three convolutional blocks (batch normalization, ReLU, max/adaptive-average pooling) of increasing channel depth ($32 \to 64 \to 128$), followed by fully connected layers that compress the spatial representation to 64 dimensions. The \textbf{text encoder} consumes the embedding of the [CLS] classification token from a RoBERTa-base backbone~\citep{roberta2019} ($\mathbb{R}^{768}$), used as a frozen feature extractor, and projects it to $\mathbb{R}^{64}$ through a two-layer network with GELU activation. We compare generic pretrained RoBERTa with forecast-tuned RoBERTa. The forecast-tuned backbone is first trained in the discussion-only (D) model for the same seed and lead time, then frozen during fusion training. Both regimes train the text projection and fusion head. The forecast-tuned regime therefore uses a two-stage procedure. Predictions depend on both the fusion model and its frozen text backbone. Unless stated otherwise, the neural text-fusion models (T+D, I+D and T+I+D) use the forecast-tuned backbone.

We investigate two fusion strategies. \textit{Concatenation fusion} stacks the normalized modality embeddings into a vector of dimension $64M$, where $M$ is the number of configured modalities. \textit{Attention-based fusion} projects each embedding to 256 dimensions, adds a learnable [FUSION] token and modality-position embeddings, and applies a single pre-LN Transformer layer with four attention heads. Layer normalization precedes the attention and feed-forward sublayers. The [FUSION] output supplies the fused representation. This architecture can learn interactions between modalities, but the fitted comparisons do not identify a specific weighting mechanism. A linear head produces $2H$ intensity-change predictions, where $H$ is the number of lead times. Full specifications are given in \autoref{sec:encoder_details}.

\subsection{Temporal Approach}\label{sec:temporal}

\begin{figure}[t]
\centering
\resizebox{\textwidth}{!}{\input{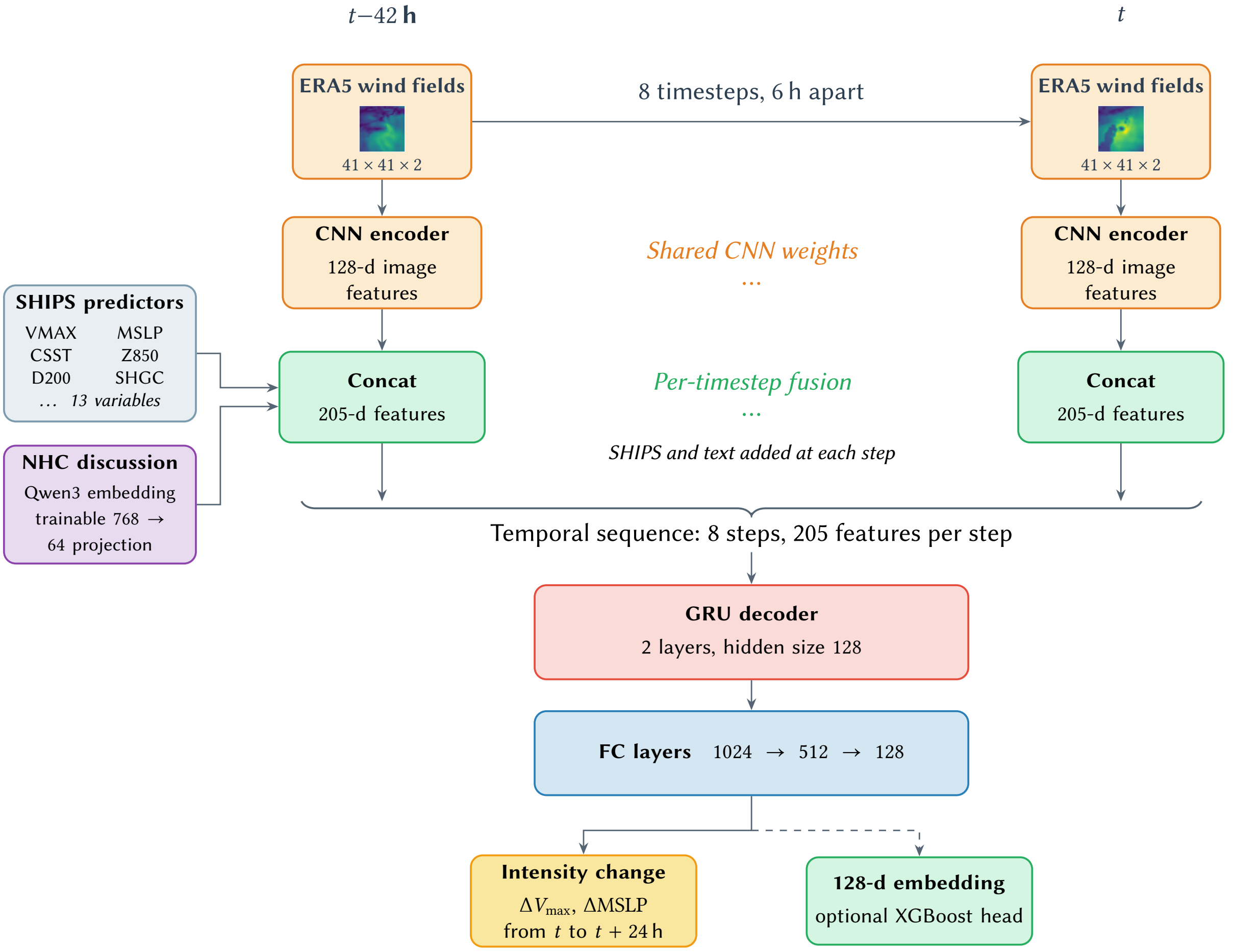}}
\caption{Temporal architecture extending HurriCast~\citep{Boussioux2022} with discussion text. Eight six-hourly ERA5 frames share one CNN, while 13 SHIPS predictors and 64 projected Qwen3 text features join the 128 image features at each time step. The resulting $8\times205$ sequence enters a two-layer GRU and fully connected (FC) layers. The final 128-dimensional representation supports direct 24\,h changes in $V_{\max}$ and MSLP or an optional XGBoost head.}
\label{fig:hurricast_temporal}
\end{figure}

In a standard setting, both the hybrid and end-to-end approaches treat each case independently. To capture the temporal evolution of storm intensity, we extend the HurriCast framework~\citep{Boussioux2022} with multi-modal inputs (\autoref{fig:hurricast_temporal}). A shared CNN encoder processes ERA5 wind fields at each of 8 consecutive 6-hourly timesteps ($t{-}42$\,h to $t$), producing 128-dimensional image embeddings followed by layer normalization. At every timestep, these embeddings are concatenated with the 13 tabular SHIPS predictors and, optionally, the 768 retained dimensions of the pretrained \texttt{Qwen3-Embedding-0.6B}~\citep{qwen3embedding} sentence embedding, projected to 64 dimensions by a trainable linear layer with GELU and layer normalization. The full configuration forms a sequence of shape $8 \times 205$. At each time step, the temporal model uses the discussion matched to the case at the ERA5 frame timestamp. Missing discussions receive zero embeddings. The matched discussion can have a later issue time than that timestamp, following the rule in Section~\ref{sec:curation}.

A 2-layer GRU decoder models the temporal evolution of this sequence, producing 128-dimensional hidden representations that are flattened and projected through fully connected layers. The resulting 128-d embeddings can either be used directly for prediction via a linear head, or serve as features for an XGBoost ensemble, providing a second hybrid pathway complementary to the approach in Section~\ref{sec:hybrid}.

\subsection{Training Strategy}\label{sec:training}

\subsubsection{Loss Function}
We train the end-to-end neural models to minimize the mean squared error (MSE) in predicted wind speed and pressure changes. To compare errors in knots and hPa on a common scale, we divide each error by the standard deviation of its target in the training set. For a training batch $B$ with at least one available target, the loss is
\begin{equation}\label{eq:loss}
\mathcal{L}_B = \frac{1}{N_B}\sum_{j\in\mathcal{O}_B}
 \left(\frac{\Delta\hat y_j-\Delta y_j}{\sigma_j^{\mathrm{train}}}\right)^2,
\end{equation}
where $\Delta y_j$ is an observed wind speed or pressure change and $\Delta\hat y_j$ is its prediction, both in physical units. The scale $\sigma_j^{\mathrm{train}}$ is the standard deviation in the training set for the corresponding variable and forecast lead. It is computed once on the whole training set, not within the batch, so all values of the same variable and lead share one scale. The index $j$ runs over all available target values in the batch, collected in $\mathcal{O}_B$, and $N_B$ is their number. This expression is equivalent to MSE on standardized targets because subtracting the training mean does not change the prediction error. Missing targets contribute neither an error term nor a count in the denominator.

This treatment is useful when one model predicts several forecast leads. For example, a case with a valid 24\,h target can still contribute to training when its 120\,h target is unavailable. We retain a case whenever at least one wind or pressure target is available. The temporal neural models use MSE on complete target pairs after sequence filtering, while the hybrid regressors fit each target in physical units. Validation averaging and model selection details are given in Appendix~\ref{app:training_configuration}.

\subsubsection{Training Details}
We train the neural models with AdamW~\citep{loshchilov2019adamw}, using learning rates specific to each modality, batch sizes and early stopping based on validation loss. The primary 24\,h runs use patience 7 and epoch caps of 200 for T, I and T+I and 500 for D and the models using text fusion. The configuration details and their selection limits are given in Appendix~\ref{app:training_configuration}. Learning rates decrease on a validation plateau (factor $0.5$, patience $3$), and gradient norms are clipped at 1.0. In the 24\,h multi-modal ablations that include text, linear weights are reset with Kaiming normal initialization (Appendix~\ref{app:initialization}). Our temporal split uses 2002--2019 for training, 2020--2021 for validation, and 2022--2023 for testing. We retain rows with at least one valid 24\,h target. The split avoids train--test storm overlap, but it does not remove the later discussion timing or the retrospective nature of ERA5. Numerical inputs and targets are standardized as described in Section~\ref{sec:curation}.

\subsection{Baselines and Evaluation}\label{sec:eval}

Tables of fitted models report means across five training seeds, with sample standard deviations where shown, unless stated otherwise. In comparative tables, bold and underlined values indicate the best and second-best displayed scores. Paired comparisons use identical cases, whereas Table~\ref{tab:best_models} summarizes scores from models evaluated on different sets of cases. Ties at the reported precision share emphasis. A higher ranking alone does not establish a statistically significant difference. For paired comparisons of forecast errors, we obtain 95\% confidence intervals by resampling whole storms (Appendix~\ref{app:subj_stratified_gain}).

We compare our multi-modal approaches against baselines of increasing complexity (\autoref{tab:baselines}). We use three simple reference baselines. \emph{Persistence} predicts zero intensity change. \emph{Trend persistence} extrapolates the observed past change over the same lead time, $\Delta\hat y_{t,h}=y_t-y_{t-h}$. \emph{Climatology} predicts the mean change in the training set for each target and lead time. The trend baseline requires an exact timestamp match within the same storm at $t-h$, rather than a fixed number of preceding rows, and falls back to zero change otherwise. It performs worse than persistence and climatology here (\autoref{tab:physical_baselines}). This result applies to the extrapolation rule and does not show that past intensity changes are uninformative. We then add Ridge regression and XGBoost in three modality configurations, comprising tabular only (T, the 13 SHIPS predictors), discussion only (D, the 768-d \texttt{Qwen3-Embedding-0.6B} embedding truncated from the native 1024-d output), and T+D. A tabular-only neural network additionally provides the neural reference. The multi-modal TC studies reviewed above use different datasets, targets and evaluation periods, so their reported scores are not direct baselines for our experiment.

\begin{table*}[htbp]
\centering
\caption{Classical ML baselines and reference forecasts for 24\,h intensity change. T=Tabular (SHIPS), D=Discussion embeddings. \textbf{Bold} = best, \underline{underlined} = second-best per column. Fitted model scores are mean $\pm$ sample SD across five seeds. Ridge and reference baselines are deterministic.}
\label{tab:baselines}
\small
\renewcommand{\arraystretch}{1.12}
\setlength{\tabcolsep}{5pt}
\fullwidthresulttable{lcccccc}{
\toprule
 & \multicolumn{3}{c}{$V_{\max}$} & \multicolumn{3}{c}{MSLP} \\
\cmidrule(lr){2-4} \cmidrule(lr){5-7}
Model & $R^2$ & RMSE (kt) & MAE (kt) & $R^2$ & RMSE (hPa) & MAE (hPa) \\
\midrule
Persistence & $-0.03$ & $19.33$ & $13.78$ & $-0.04$ & $13.29$ & $8.86$ \\
Trend persistence & $-0.54$ & $23.63$ & $16.12$ & $-0.60$ & $16.51$ & $10.32$ \\
Climatology & $-0.01$ & $19.15$ & $13.76$ & $-0.01$ & $13.11$ & $8.73$ \\
\midrule
Ridge (T) & $0.26$ & $16.36$ & $12.14$ & $0.24$ & $11.36$ & $8.12$ \\
Ridge (D) & $0.40$ & $14.74$ & $10.93$ & $0.36$ & $10.41$ & $7.47$ \\
Ridge (T+D) & $0.45$ & $14.11$ & $10.47$ & $0.41$ & $10.04$ & $7.18$ \\
XGB (T) & $\underline{0.46\pm0.00}$ & $\underline{14.04\pm0.06}$ & $\underline{9.88\pm0.06}$ & $\underline{0.45\pm0.00}$ & $\underline{9.69\pm0.03}$ & $\underline{6.55\pm0.02}$ \\
XGB (D) & $0.39\pm0.01$ & $14.92\pm0.10$ & $10.79\pm0.07$ & $0.35\pm0.01$ & $10.50\pm0.05$ & $7.26\pm0.05$ \\
XGB (T+D) & $\boldsymbol{0.51\pm0.01}$ & $\boldsymbol{13.29\pm0.13}$ & $\boldsymbol{9.49\pm0.08}$ & $\boldsymbol{0.51\pm0.01}$ & $\boldsymbol{9.16\pm0.08}$ & $\boldsymbol{6.36\pm0.08}$ \\
\bottomrule
}
\par\smallskip
{\footnotesize All models use 1{,}150 test cases from 66 storms. Reference forecasts are deterministic.}
\end{table*}

We report $R^2$, RMSE, and MAE. For the primary 24\,h neural ablation, \autoref{tab:nn_ablation} gives the mean and sample standard deviation of each metric over five networks trained from different seeds. Averaging their predictions gives a deep ensemble \citep{lakshminarayanan2017deep}, and we call its scores \emph{ensemble} scores (\autoref{tab:nn_ensemble}). This ensemble of five training runs is distinct from the dynamical-model ensembles of numerical weather prediction. Models trained separately for each forecast lead and models predicting all leads jointly also use five seeds for every modality configuration. Neural D is evaluated only on test cases with a matched discussion, while the other neural models use the full test set (\autoref{tab:neural_populations}), so comparisons with D need a common subset. For 24\,h $V_{\max}$ directional accuracy (DAcc), a forecast is correct when $\operatorname{sign}(\Delta\hat V_{\max,24})=\operatorname{sign}(\Delta V_{\max,24})$, where $\operatorname{sign}$ takes the values $-1$ (weakening), $0$ (steady) and $+1$ (strengthening). We call this rule strict sign agreement because a predicted change counts as steady only when it is exactly zero. Observed classes come from the original targets in physical units, so that inverting the standardization cannot turn an exact zero change into a small nonzero value. \autoref{app:dacc_confusion} gives the ensemble confusion matrices under this rule and a supplemental convention that treats predictions within a fixed $\pm2.5$\,kt band as steady.

\FloatBarrier
\section{Results}

\subsection{Modality Ablation}

At 24, 48 and 72\,h, \autoref{tab:ablation} reports $R^2$ and \autoref{tab:ablation_mae} reports MAE. In the primary 24\,h comparison, adding forecaster discussions to the tabular neural model gives a clear gain. Average MAE for wind change falls from $\resultvalue{seed/abl/tab/VMAX/24/mae_mean}$ to $\resultvalue{seed/abl/tab+disc/VMAX/24/mae_mean}$\,kt, a $\resultvalue{comparison/primary/discussion/wind_mae_reduction_percent}\%$ reduction on the same \resultvalue{primary/n_obs} cases. Wind $R^2$ rises from $\resultvalue{seed/abl/tab/VMAX/24/r2_mean}$ to $\resultvalue{seed/abl/tab+disc/VMAX/24/r2_mean}$. These scores are averages over five training runs, and \autoref{tab:nn_ablation} gives their variation.

The wind fields carry predictive information on their own, with mean $R^2=\resultvalue{seed/abl/img/VMAX/24/r2_mean}$, but adding them to tabular inputs in the neural model changes the mean score only slightly. Adding wind fields to the model using tabular inputs with discussions changes mean wind MAE from $\resultvalue{seed/abl/tab+disc/VMAX/24/mae_mean}$ to $\resultvalue{seed/abl/full/VMAX/24/mae_mean}$\,kt, less than the variation across training runs. This limited contribution is in line with the findings of \citet{Boussioux2022}, whose image-based configurations provided smaller improvements for 24\,h intensity forecasts than for track forecasts in the same basins. Our result concerns the two near-surface wind fields and training configurations used here. Richer images, a different image representation or further tuning on the validation set could still help. The ordering also varies across targets and forecast leads, as the full tables show.

\begin{table*}[!htbp]
\centering
\caption{Modality ablation across lead times ($R^2$). T=Tabular, I=Images, D=Discussions. \textbf{Bold} = best, \underline{underlined} = second-best per column (each approach ranked separately). Both approaches use the same temporal train/test split. A dagger ($\dagger$) marks neural D, which is evaluated on a smaller test population. Displayed values are means of scores from five runs.}
\label{tab:ablation}
\footnotesize
\renewcommand{\arraystretch}{1.1}
\setlength{\tabcolsep}{3.2pt}
\begin{tabularx}{\linewidth}{@{}l*{12}{>{\centering\arraybackslash}X}@{}}
\toprule
 & \multicolumn{6}{c}{\textbf{Hybrid XGB}} & \multicolumn{6}{c}{\textbf{End-to-end NN}} \\
\cmidrule(lr){2-7} \cmidrule(lr){8-13}
 & \multicolumn{2}{c}{24\,h} & \multicolumn{2}{c}{48\,h} & \multicolumn{2}{c}{72\,h} & \multicolumn{2}{c}{24\,h} & \multicolumn{2}{c}{48\,h} & \multicolumn{2}{c}{72\,h} \\
\cmidrule(lr){2-3} \cmidrule(lr){4-5} \cmidrule(lr){6-7} \cmidrule(lr){8-9} \cmidrule(lr){10-11} \cmidrule(lr){12-13}
Modality & $V_{\max}$ & MSLP & $V_{\max}$ & MSLP & $V_{\max}$ & MSLP & $V_{\max}$ & MSLP & $V_{\max}$ & MSLP & $V_{\max}$ & MSLP \\
\midrule
\multicolumn{13}{@{}l}{\textit{Single}} \\
T & $0.46$ & $0.45$ & $\underline{0.49}$ & $0.45$ & $\underline{0.42}$ & $0.36$ & $0.44$ & $0.44$ & $0.47$ & $0.44$ & $0.39$ & $0.33$ \\
I & $0.23$ & $0.19$ & $0.26$ & $0.23$ & $0.27$ & $0.22$ & $0.22$ & $0.20$ & $0.16$ & $0.12$ & $0.24$ & $0.15$ \\
D & $0.39$ & $0.35$ & $0.37$ & $0.31$ & $0.27$ & $0.20$ & $0.50^{\dagger}$ & $0.46^{\dagger}$ & $0.54^{\dagger}$ & $0.51^{\dagger}$ & $0.42^{\dagger}$ & $0.35^{\dagger}$ \\
\midrule
\multicolumn{13}{@{}l}{\textit{Dual}} \\
T+I & $0.40$ & $0.40$ & $0.43$ & $0.42$ & $\underline{0.42}$ & $\underline{0.39}$ & $0.45$ & $0.45$ & $0.45$ & $0.42$ & $0.41$ & $0.31$ \\
T+D & $\boldsymbol{0.51}$ & $\boldsymbol{0.51}$ & $\boldsymbol{0.51}$ & $\boldsymbol{0.48}$ & $\underline{0.42}$ & $0.37$ & $\boldsymbol{0.53}$ & $\boldsymbol{0.51}$ & $\boldsymbol{0.58}$ & $\boldsymbol{0.55}$ & $\boldsymbol{0.46}$ & $\boldsymbol{0.39}$ \\
I+D & $0.40$ & $0.39$ & $0.42$ & $0.37$ & $0.35$ & $0.28$ & $\underline{0.50}$ & $\underline{0.48}$ & $0.52$ & $0.49$ & $0.39$ & $\underline{0.36}$ \\
\midrule
\multicolumn{13}{@{}l}{\textit{Triple}} \\
T+I+D & $\underline{0.49}$ & $\underline{0.49}$ & $\underline{0.49}$ & $\underline{0.47}$ & $\boldsymbol{0.44}$ & $\boldsymbol{0.40}$ & $\boldsymbol{0.53}$ & $\boldsymbol{0.51}$ & $\underline{0.57}$ & $\underline{0.54}$ & $\underline{0.45}$ & $\boldsymbol{0.39}$ \\
\bottomrule
\end{tabularx}
\par\smallskip
{\footnotesize $^{\dagger}$Neural D is evaluated only on test cases with a matched discussion (1{,}105 cases from 64 storms at 24\,h, 860 from 53 at 48\,h and 653 from 47 at 72\,h). The other neural models use 1{,}150 cases from 66 storms, 898 from 53 and 686 from 51 at the same leads. Daggered values are excluded from the ranking. Neural results use the primary ablation at 24\,h and separate models at 48 and 72\,h. Evaluation counts are in Table~\ref{tab:neural_populations}.}
\end{table*}

\begin{table*}[!htbp]
\centering
\caption{Modality ablation across lead times (MAE). T=Tabular, I=Images, D=Discussions. \textbf{Bold} = best, \underline{underlined} = second-best per column (each approach ranked separately). Both approaches use the same temporal train/test split. A dagger ($\dagger$) marks neural D, which is evaluated on a smaller test population. Displayed values are means of scores from five runs.}
\label{tab:ablation_mae}
\footnotesize
\renewcommand{\arraystretch}{1.1}
\setlength{\tabcolsep}{3.2pt}
\begin{tabularx}{\linewidth}{@{}l*{12}{>{\centering\arraybackslash}X}@{}}
\toprule
 & \multicolumn{6}{c}{\textbf{Hybrid XGB}} & \multicolumn{6}{c}{\textbf{End-to-end NN}} \\
\cmidrule(lr){2-7} \cmidrule(lr){8-13}
 & \multicolumn{2}{c}{24\,h} & \multicolumn{2}{c}{48\,h} & \multicolumn{2}{c}{72\,h} & \multicolumn{2}{c}{24\,h} & \multicolumn{2}{c}{48\,h} & \multicolumn{2}{c}{72\,h} \\
\cmidrule(lr){2-3} \cmidrule(lr){4-5} \cmidrule(lr){6-7} \cmidrule(lr){8-9} \cmidrule(lr){10-11} \cmidrule(lr){12-13}
Modality & $V_{\max}$ & MSLP & $V_{\max}$ & MSLP & $V_{\max}$ & MSLP & $V_{\max}$ & MSLP & $V_{\max}$ & MSLP & $V_{\max}$ & MSLP \\
\midrule
\multicolumn{13}{@{}l}{\textit{Single}} \\
T & $9.88$ & $6.55$ & $\underline{15.29}$ & $11.04$ & $18.41$ & $13.57$ & $10.13$ & $6.60$ & $15.32$ & $10.77$ & $18.77$ & $13.82$ \\
I & $12.21$ & $8.10$ & $18.35$ & $12.69$ & $20.21$ & $14.53$ & $12.40$ & $8.15$ & $18.97$ & $13.31$ & $20.43$ & $15.13$ \\
D & $10.79$ & $7.26$ & $17.04$ & $12.19$ & $19.84$ & $14.37$ & $9.90^{\dagger}$ & $6.56^{\dagger}$ & $14.32^{\dagger}$ & $10.26^{\dagger}$ & $17.63^{\dagger}$ & $12.94^{\dagger}$ \\
\midrule
\multicolumn{13}{@{}l}{\textit{Dual}} \\
T+I & $10.55$ & $6.90$ & $15.53$ & $\underline{10.97}$ & $\boldsymbol{17.98}$ & $\boldsymbol{13.08}$ & $10.27$ & $6.72$ & $15.49$ & $10.80$ & $18.34$ & $13.95$ \\
T+D & $\boldsymbol{9.49}$ & $\boldsymbol{6.36}$ & $15.41$ & $11.14$ & $18.55$ & $13.90$ & $\underline{9.46}$ & $\boldsymbol{6.23}$ & $\boldsymbol{13.44}$ & $\boldsymbol{9.58}$ & $\boldsymbol{16.93}$ & $\underline{12.50}$ \\
I+D & $10.72$ & $7.12$ & $16.32$ & $11.70$ & $18.94$ & $13.94$ & $9.74$ & $\underline{6.45}$ & $14.38$ & $10.11$ & $18.08$ & $12.93$ \\
\midrule
\multicolumn{13}{@{}l}{\textit{Triple}} \\
T+I+D & $\underline{9.78}$ & $\underline{6.43}$ & $\boldsymbol{15.27}$ & $\boldsymbol{10.82}$ & $\underline{17.99}$ & $\underline{13.31}$ & $\boldsymbol{9.45}$ & $\boldsymbol{6.23}$ & $\underline{13.65}$ & $\underline{9.67}$ & $\underline{17.11}$ & $\boldsymbol{12.44}$ \\
\bottomrule
\end{tabularx}
\par\smallskip
{\footnotesize $^{\dagger}$Neural D is evaluated only on test cases with a matched discussion (1{,}105 cases from 64 storms at 24\,h, 860 from 53 at 48\,h and 653 from 47 at 72\,h). The other neural models use 1{,}150 cases from 66 storms, 898 from 53 and 686 from 51 at the same leads. Daggered values are excluded from the ranking. Neural results use the primary ablation at 24\,h and separate models at 48 and 72\,h. Evaluation counts are in Table~\ref{tab:neural_populations}. MAE is in kt for $V_{\max}$ and hPa for MSLP.}
\end{table*}

\subsection{Temporal Model Results}

\autoref{tab:hurricast_ablation} presents temporal configurations on \resultvalue{temporal/common/n_obs} cases from \resultvalue{temporal/common/n_storms} test storms that meet the eight-frame sequence requirement. In the modality ablation, XGBoost over T+I+D embeddings has mean wind MAE $\resultvalue{cell/hurricast_ablation_24h/CNN_Tab_Text_emb/VMAX/mae/mean}$\,kt versus $\resultvalue{cell/hurricast_ablation_24h/CNN_Tab_emb/VMAX/mae/mean}$\,kt for T+I. The corresponding paired ensemble gain is $\resultvalue{temporal/ablation_text_gain/VMAX/mae_gain}$\,kt, with a whole-storm 95\% interval of $[\resultvalue{temporal/ablation_text_gain/VMAX/storm_ci_low},\resultvalue{temporal/ablation_text_gain/VMAX/storm_ci_high}]$\,kt that includes zero, as does the pressure interval. These results do not establish a clear improvement from adding discussions to the temporal model. The full temporal model has two prediction variants, labeled T+I+D (GRU) and T+I+D (GRU+XGB) in the table. The first predicts directly from the temporal embedding. The second fits XGBoost to that embedding together with tabular and discussion features from the last time step. In contrast, the modality ablation rows use only the temporal embeddings as XGBoost inputs. Neither comparison isolates the causal contribution of spatial wind structure.

\begin{table*}[htbp]
\centering
\caption{Temporal model modality ablation for 24\,h intensity change. T=Tabular, I=Images, D=Discussions. \textbf{Bold} = best, \underline{underlined} = second-best per column. Entries are mean $\pm$ sample SD across five training seeds.}
\label{tab:hurricast_ablation}
\small
\renewcommand{\arraystretch}{1.12}
\setlength{\tabcolsep}{5pt}
\fullwidthresulttable{lccccc}{
\toprule
 & \multicolumn{3}{c}{$V_{\max}$} & \multicolumn{2}{c}{MSLP} \\
\cmidrule(lr){2-4} \cmidrule(lr){5-6}
Configuration & $R^2$ & MAE (kt) & DAcc & $R^2$ & MAE (hPa) \\
\midrule
\multicolumn{6}{@{}l}{\textit{Modality ablation (GRU+XGB)}} \\
I & $0.18\pm0.03$ & $11.91\pm0.30$ & $0.591\pm0.013$ & $0.16\pm0.04$ & $8.21\pm0.28$ \\
T+I & $0.36\pm0.01$ & $10.56\pm0.11$ & $0.648\pm0.004$ & $0.37\pm0.03$ & $7.05\pm0.10$ \\
I+D & $0.36\pm0.02$ & $10.42\pm0.19$ & $0.641\pm0.021$ & $0.36\pm0.02$ & $7.00\pm0.16$ \\
T+I+D & $\underline{0.45\pm0.02}$ & $\underline{9.71\pm0.27}$ & $\underline{0.664\pm0.015}$ & $\boldsymbol{0.47\pm0.02}$ & $\underline{6.43\pm0.18}$ \\
\midrule
\multicolumn{6}{@{}l}{\textit{Full temporal model}} \\
T+I+D (GRU) & $0.44\pm0.02$ & $9.92\pm0.27$ & $0.656\pm0.015$ & $\underline{0.45\pm0.01}$ & $6.63\pm0.13$ \\
T+I+D (GRU+XGB) & $\boldsymbol{0.46\pm0.02}$ & $\boldsymbol{9.68\pm0.21}$ & $\boldsymbol{0.666\pm0.010}$ & $\boldsymbol{0.47\pm0.02}$ & $\boldsymbol{6.41\pm0.17}$ \\
\bottomrule
}
\par\smallskip
{\footnotesize $n=834$, $s=55$, where $n$ is the number of test cases and $s$ the number of storms. DAcc denotes strict sign agreement for wind change. The upper block trains a temporal encoder for each modality combination. The lower block uses the full encoder, directly or with XGBoost and last-step tabular/discussion features.}
\end{table*}

\subsection{Overall Performance}

\autoref{tab:best_models} places the fitted approaches side by side. In the primary comparison, the neural networks using tabular inputs with discussions have the lowest mean wind and pressure MAE, with or without wind fields. These two models differ by less than the variation across training runs, so the scores do not separate them.

Averaging the five predictions gives a separate ensemble forecast (\autoref{tab:nn_ensemble}). Adding discussions to the tabular ensemble reduces wind MAE by $\resultvalue{comparison/ensemble/discussion/wind_mae_reduction_percent}\%$, from $\resultvalue{cell/nn_ensemble_24h/tab/VMAXmae/mean}$ to $\resultvalue{cell/nn_ensemble_24h/tab_disc/VMAXmae/mean}$\,kt. The paired reduction is $\resultvalue{primary/wind_gain/mae_gain_left_to_right}$\,kt, with a 95\% storm-bootstrap interval of $[\resultvalue{primary/wind_gain/storm_ci_low},\resultvalue{primary/wind_gain/storm_ci_high}]$ (\autoref{tab:missingness_sensitivity}). Pressure MAE also falls, by $\resultvalue{primary/pressure_gain/mae_gain_left_to_right}$\,hPa. Subsequent paired analyses use these ensembles rather than the averages of scores over training runs.

\begin{table*}[!htbp]
\centering
\caption{Comparison of hybrid, neural and temporal models for 24\,h intensity change. Entries are mean $\pm$ sample SD across five runs. Bold and underline mark the best and second-best displayed means in each metric across the table. Evaluation populations differ, so this emphasis is descriptive. Table~\ref{tab:temporal_common_subset} compares identical cases.}
\label{tab:best_models}
\small
\renewcommand{\arraystretch}{1.12}
\setlength{\tabcolsep}{4pt}
\fullwidthresulttable{lccccc}{
\toprule
\multirow{2}{*}{Model} & \multicolumn{2}{c}{$R^2$} & \multicolumn{2}{c}{MAE (kt/hPa)} & \multirow{2}{*}{$V_{\max}$ DAcc} \\
\cmidrule(lr){2-3} \cmidrule(lr){4-5}
 & $V_{\max}$ & MSLP & $V_{\max}$ & MSLP &  \\
\midrule
\multicolumn{6}{@{}l}{\textit{Hybrid ($n=1{,}150$, $s=66$)}} \\
T & $0.46\pm0.00$ & $0.45\pm0.00$ & $9.88\pm0.06$ & $6.55\pm0.02$ & $0.700\pm0.004$ \\
T+D & $\underline{0.51}\pm0.01$ & $\boldsymbol{0.51}\pm0.01$ & $9.49\pm0.08$ & $\underline{6.36}\pm0.08$ & $0.714\pm0.006$ \\
T+I+D & $0.49\pm0.02$ & $\underline{0.49}\pm0.03$ & $9.78\pm0.16$ & $6.43\pm0.18$ & $0.708\pm0.009$ \\
\midrule
\multicolumn{6}{@{}l}{\textit{Neural ($n=1{,}150$, $s=66$)}} \\
T & $0.44\pm0.01$ & $0.44\pm0.01$ & $10.13\pm0.05$ & $6.60\pm0.08$ & $0.684\pm0.006$ \\
T+D & $\boldsymbol{0.53}\pm0.01$ & $\boldsymbol{0.51}\pm0.01$ & $\underline{9.46}\pm0.16$ & $\boldsymbol{6.23}\pm0.08$ & $\underline{0.723}\pm0.009$ \\
T+I+D & $\boldsymbol{0.53}\pm0.01$ & $\boldsymbol{0.51}\pm0.01$ & $\boldsymbol{9.45}\pm0.13$ & $\boldsymbol{6.23}\pm0.08$ & $\boldsymbol{0.724}\pm0.006$ \\
\midrule
\multicolumn{6}{@{}l}{\textit{Temporal ($n=834$, $s=55$)}} \\
T+I & $0.36\pm0.01$ & $0.37\pm0.03$ & $10.56\pm0.11$ & $7.05\pm0.10$ & $0.648\pm0.004$ \\
T+I+D (GRU) & $0.44\pm0.02$ & $0.45\pm0.01$ & $9.92\pm0.27$ & $6.63\pm0.13$ & $0.656\pm0.015$ \\
T+I+D (GRU+XGB) & $0.46\pm0.02$ & $0.47\pm0.02$ & $9.68\pm0.21$ & $6.41\pm0.17$ & $0.666\pm0.010$ \\
\bottomrule
}
\par\smallskip
{\footnotesize $n$ denotes test cases and $s$ test storms. MAE is in kt/hPa for $V_{\max}$/MSLP. DAcc is strict sign agreement for wind change. Temporal models use a smaller population. Comparisons on identical cases are in Table~\ref{tab:temporal_common_subset}.}
\end{table*}

For operational context, \autoref{sec:operational_baseline} benchmarks our 24\,h errors against the official forecast (OFCL) and archived guidance (HWRF, HAFS, DSHP and LGEM), including preoperational HAFS forecasts in the NHC ATCF archives.

\section{Model Analysis}

\subsection{Detecting Subjectivity}
\label{sec:detecting_subjectivity}

To describe explicit markers of personal judgment in the discussions, we classify each NHC discussion into three rhetorical categories. An \textsc{opinionated} discussion makes an explicit first-person subjective statement (``I think'', ``it was hard for me to ignore''). A \textsc{balanced} discussion has no first-person language but sustains interpretive reasoning across physical mechanisms or competing numerical weather prediction models. A \textsc{factual} discussion has neither. We labeled a stratified benchmark with Claude Sonnet~4.6 \citep{anthropic2026sonnet46} as LLM-as-a-judge \citep{zheng2023judging}. Claude Opus~4.6 \citep{anthropic2026opus46} then labeled a subset using the same rubric without seeing the Sonnet labels. On the benchmark, a regular expression (regex) that detects first-person language reproduces the judge's \textsc{opinionated} label, so we use it as a proxy for explicit first-person framing in the full corpus. The two judges disagree more often on the \textsc{balanced}/\textsc{factual} boundary, where reasoning depth varies continuously. An exploratory hybrid classifier combines the regex with a few-shot \citep{brown2020language} \texttt{gemma-4-E2B-it} model \citep{google2025gemma}. Its agreement with one LLM judge is descriptive, not independent human validation. \autoref{app:subjectivity} gives the counts, rubric, regex patterns and audit.

We use these labels only to compare performance across groups of discussions, as the forecasting model takes the cleaned text through its encoder (Section~\ref{sec:e2e}) rather than the assigned classes. On cases with matched discussions, adding text to the tabular ensemble reduces wind MAE by \resultvalue{cell/strat_gain/discussion_present/mae_gain_left_to_rightVMAX/mean}\,kt, with an improvement also seen in the larger group without first-person statements detected by the regex. The gain is therefore not confined to explicit first-person language, although the small group flagged by the regex has negative point gains, with wind and pressure confidence intervals that both include zero (\autoref{app:subj_stratified_gain}). These comparisons describe performance within each group without testing differences between groups or isolating the causal contribution of factual content, interpretation, or judgment.

\subsection{Interpretability}
\label{sec:interpretability}

To inspect how individual trained models respond to the discussion text, we apply Integrated Gradients (IG, \citealp{Sundararajan2017IG}) via the Captum library~\citep{Kokhlikyan2020Captum} to the trained T+D and discussion-only (D) models. IG attributes the model's prediction to individual input embeddings by integrating the gradient of the output with respect to the embedding along a straight-line path from a baseline (here, the pad-token embedding) to the actual token embedding. Token-level attributions are summed within tokenizer-defined word spans and sentences so that attributions can be reported for words or sentences that have a meteorological interpretation. To measure \emph{where} text is associated with improved predictions, we compare absolute error for each case between the tabular-only and T+D ensembles on the test set (Appendix~\ref{app:subj_stratified_gain}) and use IG only to characterize a fixed set of illustrative predictions from individual models. \autoref{app:interpretability} details the attribution pipeline, the IG completeness check, and worked examples on a discussion-only rapid intensification case (Tropical Storm Ian, Discussion 10), a T+D case where discussions improve the prediction and include explicit forecaster judgment (Hurricane Agatha, Discussion 9), and a T+D case where discussions worsen the prediction despite cautious forecaster wording (Tropical Storm Madeline, Discussion 7).

\subsection{Uncertainty Quantification}
\label{sec:uq}

Point forecasts do not convey forecast uncertainty, so we construct 90\% prediction intervals without retraining the forecasting models. Cases within a track are dependent, so each split holds out whole storms. For the primary T+D ensemble, we normalize residuals by the ensemble spread and calibrate the intervals by resampling whole storms. Across the paired splits used for comparison with CLEAR \citep{azizi2025clear}, wind coverage is $\resultvalue{uq/textgain/vmax/T+D/coverage_mean}$ and pressure coverage is $\resultvalue{uq/textgain/mslp/T+D/coverage_mean}$. The median wind width is $\resultvalue{uq/textgain/vmax/T+D/median_width}$\,kt. The wind interval score is $\resultvalue{uq/textgain/vmax/T+D/interval_score}$, versus $\resultvalue{uq/textgain/vmax/T/interval_score}$ for the tabular model under the same construction, where lower is better. These are empirical coverage and score measurements, not guarantees. \autoref{app:uq} reports the online methods, conditional results, and a separate analysis across forecast leads.

Marginal coverage is not sufficient for every storm or intensity regime. On the observed rapid intensification subset, every evaluated interval method falls below the 90\% nominal level (\autoref{fig:uq_ri_conditional}). On Otis, the T+D interval misses the largest changes despite covering most of that storm's other cases (\autoref{app:uq}). The full method comparison, per-lead results, storm panels and conditional analysis are reported in the appendix.

\section{Discussion}

\subsection{Comparing Approaches}

The three model families use different encoders and are not always evaluated on the same cases. On all \resultvalue{primary/n_obs} valid 24\,h test cases, adding discussions to hybrid XGBoost reduces mean wind MAE from $\resultvalue{cell/baselines/XGB_Tab/VMAX/mae/mean}$ to $\resultvalue{cell/baselines/XGB_Tab_Text/VMAX/mae/mean}$\,kt, compared with $\resultvalue{seed/abl/tab+disc/VMAX/24/mae_mean}$\,kt for the neural T+D model.

The temporal model is evaluated on the \resultvalue{temporal/common/n_obs} cases that meet its sequence requirements, so a fair comparison requires evaluating the other models on this same subset. On these cases, neural T+D has mean wind MAE $\resultvalue{cell/temporal_common_subset/temporal_commone2eTD_aligned/VMAX/mae/mean}$\,kt, compared with $\resultvalue{cell/temporal_common_subset/temporal_commontemporalXGB_ALL/VMAX/mae/mean}$\,kt for the full temporal model with XGBoost. \autoref{app:temporal_subset} reports the corresponding pressure and matched-discussion scores, distinguishing variation across training runs from uncertainty in paired ensemble comparisons. Where a confidence interval for the difference includes zero, the comparison remains inconclusive and does not establish that the models are equivalent.

These approaches support different research questions. The hybrid approach permits rapid regressor and modality comparisons, neural fusion jointly optimizes its trainable branches, and temporal models condition on a sequence of past storm states.

\subsection{Impact of Text Sources}

The text results concern storm-specific discussions, which contain observations, guidance summaries and forecaster assessments tailored to a cyclone. On its discussion-matched test population, D reaches a mean wind $R^2$ of $\resultvalue{seed/abl/disc/VMAX/24/r2_mean}$ over five training runs (\autoref{tab:nn_ablation}). A lexical TF-IDF model also predicts intensity change, with wind $R^2=\resultvalue{probe/tfidf/VMAX/r2}$ (\autoref{tab:text_only_tfidf}). Across five trained D models, masking numeric tokens slightly lowers mean wind $R^2$ ($\Delta R^2=\resultvalue{probe/perturb/mask_numbers/delta_r2}$), while removing sentences matched to intensity vocabulary produces a much larger decrease ($\Delta R^2=\resultvalue{probe/perturb/drop_intensity_vocab/delta_r2}$) (\autoref{tab:text_only_perturb}). The latter mask overlaps with an outlook-sentence mask, changes the text length and can expose later tokens after truncation. These probes show sensitivity to the specified wording and quantify variation across five training seeds, but do not isolate a semantic mechanism or rule out the later timing of discussions. A separate check on discussion \emph{availability} is reported in \autoref{app:missingness}.

\subsection{Fusion Strategy}

Prior work motivates studying modality competition and gradient imbalance \citep{huang2022modality,wu2022characterizing,peng2022balanced}, but these experiments do not isolate the effect of initialization or show whether the modalities learn complementary representations during training.

The primary fusion operator is specified for each modality combination and forecast lead. At 24\,h, T+I, T+D and T+I+D use concatenation, while I+D uses multi-head attention. At 48--120\,h, T+I and I+D use attention, while T+D and T+I+D use concatenation. Joint multi-lead models use concatenation. These choices are held fixed when comparing generic and forecast-tuned text backbones, and forecast tuning does not provide a uniform improvement (\autoref{tab:fusion_roles_24h}). Additional configurations assess sensitivity to the fusion operator (Tables~\ref{tab:fusion_roles_24h}--\ref{tab:fusion_roles_alternative_leads}).

\subsection{Error Analysis}

Rapid intensification remains difficult in the evaluated cases. In the discussion-only Ian example (\autoref{app:ig_examples}), the attributed checkpoint predicts $\resultvalue{case/IAN/d_seed1}$\,kt against an observed $\resultvalue{case/IAN/truth_delta}$\,kt change over 24\,h. Other large changes, including the Otis case, remain underpredicted. A successful local example therefore cannot establish reliable performance for rapid intensification, as the conditional interval results in \autoref{app:uq} also show.

\subsection{Limitations}\label{sec:limitations}

\paragraph{Scope.} We study storms in the NHC area of responsibility, the Atlantic and eastern North Pacific basins (2002--2023), at 6\,h resolution, which may miss short rapid intensification episodes and limits generalization to other basins or earlier periods. We address intensity, not track, and do not incorporate the Public Advisory or Forecast/Advisory products as additional text sources. Since ERA5 reanalysis fields and temporally matched forecaster discussions are used as inputs, the study is best read as a retrospective study of predictive information rather than a directly deployable operational system.

\paragraph{Operational readiness.} The forecast errors remain substantial for rapid intensification and landfall decisions. The ATCF comparison (\autoref{sec:operational_baseline}) includes the official NHC forecast, which remains more accurate than T+D on the same forecasts (MAE $\resultvalue{cell/operational_baseline/OFCL/aid_mae/mean}$ vs.\ $\resultvalue{cell/operational_baseline/OFCL/TD_mae/mean}$\,kt), and HWRF, HAFS, LGEM and decay-SHIPS guidance on each aid's available rows. It is not a complete verification of the operational guidance suite, which also includes consensus aids such as IVCN and HCCA. Our reanalysis inputs, best-track-based initial intensity and discussions issued after the synoptic time differ from the information available to those systems. The comparison provides retrospective context and does not establish operational equivalence.

\paragraph{Model comparisons.} Adding a modality does not uniformly improve prediction under the training configurations. The image, tabular and text configurations use different learning rates and batch sizes, and the fusion choices inherited from generic text features need not be optimal for the forecast-tuned backbone. These comparisons therefore assess specified models, not the maximum information available in each modality. Checkpoint and fusion configuration selection used validation results. Additional baseline methods were included after the initial test evaluation, so the set of comparisons was not fixed in advance. The storm-bootstrap intervals condition on the fitted models and do not include uncertainty from tuning or model selection. Evaluation on later seasons with the modeling choices fixed beforehand would provide a further test of generalization. The temporal model also requires a sequence of valid past ERA5 frames, which changes its evaluation population. Comparisons on identical cases (\autoref{app:temporal_subset}) address that population difference without removing every training difference.

\paragraph{Robustness of the text signal.} Storm-specific discussions align to most but not all cases. A missingness-indicator baseline and pre-landfall stratification (\autoref{app:missingness}) assess sensitivity to discussion availability without excluding interactions between missingness and other inputs. The post-landfall point gain is larger but less precise, and the selected SHIPS predictors include latitude and longitude but no explicit variable for distance to land. The newer \texttt{Qwen3-Embedding-0.6B} encoder may have encountered hurricane reporting from the test period during pretraining. The original RoBERTa-base predates those test seasons. The TF-IDF baseline (\autoref{app:text_only_tfidf}) shows that lexical features have predictive value without pretraining a language model. The perturbation probes instead measure the sensitivity of five pretrained discussion-only checkpoints. Neither analysis quantifies the effect of possible Qwen pretraining exposure. The retained narrative can state the official intensity forecast and refer to statistical or consensus guidance, so part of the text signal may be the official forecast itself. Only a small number of test discussions are flagged \textsc{opinionated} by the first-person regex, so a differential effect of explicit forecaster judgment cannot be isolated at the present sample size.

\section{Conclusion}

MultiCyclone evaluates NHC discussions alongside tabular meteorological inputs and ERA5 wind fields for retrospective prediction of tropical cyclone intensity. On the 2022--2023 storms held out from training, neural networks using tabular inputs with discussions reduce average 24\,h wind MAE from $\resultvalue{seed/abl/tab/VMAX/24/mae_mean}$ to $\resultvalue{seed/abl/tab+disc/VMAX/24/mae_mean}$\,kt across five runs. Averaging the predictions of those runs gives an ensemble MAE reduction of $\resultvalue{primary/wind_gain/percent}\%$ on identical cases, with a paired whole-storm interval excluding zero. The benefit varies with architecture, lead and evaluation population, and adding the available wind fields leaves this configuration's mean wind MAE essentially unchanged. Attribution and perturbation analyses identify sensitivity to language about intensity and its evolution without establishing a unique causal explanation. These findings support further study of forecaster discussions beyond the selected numerical inputs. Reanalysis inputs, discussions issued after the valid time of the tabular and image data and forecast statements in their narrative limit operational interpretation. Interval coverage also varies across storms and does not establish a general guarantee. An evaluation on a season not used during model development should therefore test these models with richer operational inputs and data available at forecast issue time, followed by calibration across seasons and evaluation in additional basins.

\section*{Acknowledgments}

We thank the DCSR unit at the University of Lausanne for computational resources, NHC for the discussion archives and ECMWF for the ERA5 reanalysis data.

\section*{Data and code availability}

SHIPS, ERA5, TCBench and IBTrACS data are available from their cited sources, and discussions and ATCF forecast archives from NHC. Code, saved predictions and instructions for reproducing the results will be released with a later version of this preprint. Processed training arrays and available model checkpoints can be requested from the corresponding author.

\section*{List of abbreviations}
\label{sec:abbreviations}

\begin{multicols}{2}
\footnotesize
\begin{description}[style=nextline,leftmargin=!,labelwidth=5em,labelsep=0.6em,labelindent=0pt,itemsep=0.5pt,parsep=0pt,topsep=2pt,font=\normalfont]
\setlength{\itemindent}{0pt}
\item[ACI] Adaptive conformal inference
\item[Aid] ATCF technique identifier for a specific forecast/guidance product (e.g.\ OFCL, HWRF)
\item[ASCAT] Advanced Scatterometer
\item[ATCF] Automated Tropical Cyclone Forecasting system (source of the NHC guidance and best-track files)
\item[BN] Batch normalization
\item[CI] Confidence interval
\item[CLS] Classification token
\item[CNN] Convolutional neural network
\item[CSST] Climatological sea-surface temperature
\item[D] Discussion text (modality and text-only model)
\item[D200] 200~hPa divergence
\item[DAcc] 24\,h $V_{\max}$ directional accuracy
\item[DTOPS] Deterministic to Probabilistic Statistical rapid intensification model
\item[DSHP] SHIPS variant with decay-after-landfall (NHC operational guidance)
\item[ECMWF] European Centre for Medium-Range Weather Forecasts
\item[EnbPI] Ensemble batch prediction intervals
\item[EPOS] Mean positive lifted-parcel--environment $\theta_e$ difference
\item[ERA5] ECMWF Reanalysis v5
\item[FC] Fully connected (layer)
\item[FFN] Feed-forward network
\item[G200] 200~hPa temperature perturbation
\item[GELU] Gaussian error linear unit
\item[GFDL] Geophysical Fluid Dynamics Laboratory (hurricane model)
\item[GFS] Global Forecast System
\item[GRU] Gated recurrent unit
\item[HAFS] Hurricane Analysis and Forecast System
\item[HCCA] HFIP Corrected Consensus Approach
\item[HWRF] Hurricane Weather Research and Forecasting model
\item[I] Image (ERA5, modality)
\item[IBTrACS] International Best Track Archive for Climate Stewardship
\item[IG] Integrated gradients
\item[IVCN] Intensity Variable Consensus (NHC intensity consensus aid)
\item[LGEM] Logistic Growth Equation Model (NHC intensity guidance)
\item[LLM] Large language model
\item[LN] Layer normalization (``pre-LN'' = LN applied before the Transformer sub-blocks)
\item[MAE] Mean absolute error
\item[MHA] Multi-head attention
\item[ML] Machine learning
\item[MLP] Multi-layer perceptron
\item[MSE] Mean squared error
\item[MSLP] Minimum sea-level pressure
\item[NHC] National Hurricane Center
\item[NN] Neural network
\item[NOAA] National Oceanic and Atmospheric Administration
\item[NWS] National Weather Service
\item[OFCL] NHC official forecast (ATCF aid identifier)
\item[$R^2$] Coefficient of determination
\item[ReLU] Rectified linear unit
\item[RHMD] 700--500~hPa relative humidity
\item[RI] Rapid intensification
\item[RMSE] Root-mean-square error
\item[RoBERTa] Robustly optimized BERT pretraining approach (language model)
\item[SAB] Satellite Analysis Branch (NOAA/\allowbreak NESDIS)
\item[SD] Standard deviation
\item[SHGC] Generalized multilevel wind shear, vortex removed
\item[SHIPS] Statistical Hurricane Intensity Prediction Scheme
\item[T] Tabular (SHIPS, modality)
\item[TAFB] Tropical Analysis and Forecast Branch (NOAA/NHC)
\item[T+D] Tabular + discussion
\item[T+I+D] Tabular + image + discussion
\item[TADV] 850--700~hPa temperature advection
\item[TC] Tropical cyclone
\item[TCBench] Tropical-cyclone benchmark dataset \citep{gomez2024tcbench}
\item[TF-IDF] Term frequency-inverse document frequency
\item[TWAC] Average symmetric tangential wind at 850~hPa
\item[TWDAT] Tropical Weather Discussion for the Atlantic (NHC product)
\item[$V_{\max}$] Maximum sustained 1-minute wind
\item[WIS] Weighted interval score
\item[XGBoost] Extreme Gradient Boosting
\item[Z850] 850~hPa vorticity ($10^{-7}$\,s$^{-1}$)
\end{description}
\end{multicols}

\printbibliography

\begin{appendices}
\pretocmd{\section}{\Needspace{12\baselineskip}}{}{}
\ifcsname tableorg\endcsname
  \let\table\tableorg
  \let\endtable\endtableorg
\fi

\section{Text preprocessing examples}
\label{app:text_preprocessing}

The cleaning pipeline described in Section~\ref{sec:curation} removes bulletin metadata and the official NHC table of forecast positions before the text is passed to the tokenizer, then lowercases and normalizes the remainder. Figures~\ref{fig:dorian_preprocessing} and~\ref{fig:dorian_preprocessing_clean} illustrate the preprocessing of a storm-specific discussion. The first shows the raw bulletin with stripped content in \textcolor{red!75!black}{red}; the second shows the processed string. The header is trimmed up to the first datetime line, and everything from \texttt{FORECAST POSITIONS AND MAX WINDS} onward is removed. In two training discussions this heading is misspelled or absent, so their table is retained. Numerical forecasts may still occur in the retained narrative. The retained text is whitespace-collapsed, lowercased, and stripped of non-ASCII and control characters before tokenization.

\par\begingroup
\captionsetup{type=figure,hypcap=false}
\begin{Verbatim}[commandchars=\\\{\},fontsize=\footnotesize,baselinestretch=0.9,frame=single,framesep=8pt,samepage=false,breaklines=true,breakanywhere=false,breaksymbolleft={},breakindent=0pt]
\textcolor{red!75!black}{ZCZC MIATCDAT4 ALL}
\textcolor{red!75!black}{TTAA00 KNHC DDHHMM}

\textcolor{red!75!black}{TROPICAL STORM DORIAN DISCUSSION NUMBER   2}
\textcolor{red!75!black}{NWS NATIONAL HURRICANE CENTER MIAMI FL       AL042013}
1100 AM EDT WED JUL 24 2013

VISIBLE SATELLITE IMAGERY AND RECENT ASCAT SCATTEROMETER WINDS INDICATE THAT THE TROPICAL DEPRESSION HAS BECOME BETTER ORGANIZED AND HAS STRENGTHENED INTO TROPICAL STORM DORIAN. THE INITIAL INTENSITY OF 45 KT IS BASED ON A 38-KT ASCAT SURFACE WIND AT 1111Z... AND VARIOUS OBJECTIVE SATELLITE INTENSITY ESTIMATES OF ABOUT 45 KT.

THE INITIAL MOTION ESTIMATE IS 285/18 KT. DORIAN IS EXPECTED TO BE STEERED WEST-NORTHWESTWARD TO WESTWARD BY A STRONG DEEP-LAYER SUBTROPICAL RIDGE OVER THE CENTRAL AND EASTERN ATLANTIC. THE NHC MODEL GUIDANCE REMAINS TIGHTLY PACKED IN EXCELLENT AGREEMENT ON THIS TRACK SCENARIO...WITH THE ONLY DIFFERENCES BEING SLIGHT VARIATIONS IN THE FORWARD SPEED. THE NHC OFFICIAL FORECAST TRACK IS SIMILAR TO THE PREVIOUS ADVISORY AND IS CLOSE TO THE MULTI-MODEL CONSENSUS TVCN.

FOR THE NEXT 24 HOURS OR SO...THE VERTICAL SHEAR IS FORECAST TO BE VERY LOW WHILE DORIAN REMAINS OVER MARGINALLY WARM SSTS. HOWEVER... DURING THE 24-48 HOUR TIME PERIOD...DORIAN WILL BE MOVING OVER 25C SSTS...WHICH COULD INDUCE SOME SLIGHT WEAKENING. THE CYCLONE IS EXPECTED TO MAINTAIN THE CURRENT FAVORABLE EQUATORIAL INFLOW OF WARM UNSTABLE AIR THAT IS OFFSETTING THE COOLER MORE STABLE AIR BEING INGESTED FROM THE NORTH...AND THAT COULD OFFSET THE EFFECTS OF DORIAN PASSING OVER COOLER WATERS. FOR THAT REASON...THE INTENSITY FORECAST ONLY SHOWS SLIGHT WEAKENING AT 36- AND 48-HOURS. AFTER THAT...THE CYCLONE IS LIKELY TO ENCOUNTER BOTH WARMER WATER AND INCREASING WESTERLY SHEAR...SO THE INTENSITY FORECAST IS HELD STEADY. THE OFFICIAL INTENSITY FORECAST IS SLIGHTLY HIGHER THAN THE PREVIOUS ADVISORY THROUGH THE NEXT 24 HOURS DUE TO THE HIGHER INITIAL INTENSITY...BUT THEN FOLLOWS THE TREND OF CONSENSUS MODELS.

\textcolor{red!75!black}{FORECAST POSITIONS AND MAX WINDS}

\textcolor{red!75!black}{INIT  24/1500Z 14.3N  29.9W   45 KT  50 MPH}
\textcolor{red!75!black}{ 12H  25/0000Z 15.0N  32.6W   50 KT  60 MPH}
\textcolor{red!75!black}{ 24H  25/1200Z 15.8N  36.1W   50 KT  60 MPH}
\textcolor{red!75!black}{ 36H  26/0000Z 16.4N  39.8W   45 KT  50 MPH}
\textcolor{red!75!black}{ 48H  26/1200Z 17.1N  43.4W   40 KT  45 MPH}
\textcolor{red!75!black}{ 72H  27/1200Z 18.3N  51.2W   40 KT  45 MPH}
\textcolor{red!75!black}{ 96H  28/1200Z 19.3N  58.2W   40 KT  45 MPH}
\textcolor{red!75!black}{120H  29/1200Z 20.1N  64.0W   40 KT  45 MPH}

\textcolor{red!75!black}{$$}
\textcolor{red!75!black}{FORECASTER STEWART}

\textcolor{red!75!black}{NNNN}
\end{Verbatim}
  \caption{Raw bulletin for Tropical Storm Dorian, Discussion 2, 24 July 2013 1500 UTC (verbatim reproduction of the public NHC archive, \href{https://www.nhc.noaa.gov/archive/2013/al04/al042013.discus.002.shtml}{NHC AL04 Discussion 2, 2013}). Lines in \textcolor{red!75!black}{red} are stripped by the preprocessing pipeline. This includes the bulletin metadata above the first datetime line, and everything from \texttt{FORECAST POSITIONS AND MAX WINDS} onward (the table of forecast positions, the WMO separator \texttt{\$\$}, the forecaster attribution, and the end-of-transmission marker \texttt{NNNN}).}
  \label{fig:dorian_preprocessing}
\par\endgroup

\par\begingroup
\captionsetup{type=figure,hypcap=false}
\begin{Verbatim}[fontsize=\footnotesize,frame=single,framesep=4pt,breaklines=true,breakanywhere=false,samepage=false]
1100 am edt wed jul 24 2013 visible satellite imagery and recent ascat scatterometer winds indicate that the tropical depression has become better organized and has strengthened into tropical storm dorian. the initial intensity of 45 kt is based on a 38-kt ascat surface wind at 1111z... and various objective satellite intensity estimates of about 45 kt. the initial motion estimate is 285/18 kt. dorian is expected to be steered west-northwestward to westward by a strong deep-layer subtropical ridge over the central and eastern atlantic. the nhc model guidance remains tightly packed in excellent agreement on this track scenario...with the only differences being slight variations in the forward speed. the nhc official forecast track is similar to the previous advisory and is close to the multi-model consensus tvcn. for the next 24 hours or so...the vertical shear is forecast to be very low while dorian remains over marginally warm ssts. however... during the 24-48 hour time period...dorian will be moving over 25c ssts...which could induce some slight weakening. the cyclone is expected to maintain the current favorable equatorial inflow of warm unstable air that is offsetting the cooler more stable air being ingested from the north...and that could offset the effects of dorian passing over cooler waters. for that reason...the intensity forecast only shows slight weakening at 36- and 48-hours. after that...the cyclone is likely to encounter both warmer water and increasing westerly shear...so the intensity forecast is held steady. the official intensity forecast is slightly higher than the previous advisory through the next 24 hours due to the higher initial intensity...but then follows the trend of consensus models.
\end{Verbatim}
  \caption{Processed Dorian discussion used as input to the text encoder, retaining the text from \autoref{fig:dorian_preprocessing} after header trimming and forecast-section removal, with whitespace collapsed to single spaces, lowercased, and non-ASCII characters stripped.}
  \label{fig:dorian_preprocessing_clean}
\par\endgroup

\Needspace{12\baselineskip}
\par\begingroup
\captionsetup{type=figure,hypcap=false}
  \captionsetup{font=small}
\begin{Verbatim}[commandchars=\\\{\},fontsize=\footnotesize,baselinestretch=0.85,frame=single,framesep=8pt,samepage=false,breaklines=true,breakanywhere=false,breaksymbolleft={},breakindent=0pt]
000
AXNT20 KNHC 241749
TWDAT

TROPICAL WEATHER DISCUSSION
NWS NATIONAL HURRICANE CENTER MIAMI FL
205 PM EDT WED JUL 24 2013

TROPICAL WEATHER DISCUSSION FOR NORTH AMERICA...CENTRAL AMERICA...GULF OF MEXICO...CARIBBEAN SEA...NORTHERN SECTIONS OF SOUTH AMERICA...AND ATLANTIC OCEAN TO THE AFRICAN COAST FROM THE EQUATOR TO 32N. THE FOLLOWING INFORMATION IS BASED ON SATELLITE IMAGERY...WEATHER OBSERVATIONS...RADAR...AND METEOROLOGICAL ANALYSIS.

BASED ON 1200 UTC SURFACE ANALYSIS AND SATELLITE IMAGERY THROUGH 1715 UTC.

...SPECIAL FEATURE...
TROPICAL STORM DORIAN IS LOCATED NEAR 14.3N 29.9W ABOUT 356 NM WSW OF THE CAPE VERDE ISLANDS ON 24/1500 UTC. ITS PRESENT MOVEMENT IS WNW AT 18 KT WITH MAXIMUM SUSTAINED WINDS OF 45 KT GUSTING TO 55 KT. MINIMUM CENTRAL PRESSURE IS 1002 MB. [...]

...TROPICAL WAVES...
A TROPICAL WAVE EXTENDS FROM THE CAYMAN ISLANDS AT 19N80W TO N OF PANAMA AT 11N81W MOVING W AT 20 KT. [...]

...THE ITCZ/THE MONSOON TROUGH...
THE MONSOON TROUGH AXIS EXTENDS ACROSS AFRICA INTO THE E TROPICAL ATLC NEAR 13N17W AND CONTINUES TO 14N23W. [...]

...DISCUSSION...

GULF OF MEXICO...
SURFACE RIDGING IS OVER THE GULF OF MEXICO WITH AXIS FROM MIAMI FLORIDA TO CORPUS CHRISTIE TEXAS. 10-15 KT SURFACE WINDS ARE NOTED. [...]

CARIBBEAN SEA...
A TROPICAL WAVE IS S OF THE CAYMAN ISLANDS WITH NO SIGNIFICANT PRECIPITATION. A MONSOON TROUGH EXENDS OVER PANAMA PRODUCING SCATTERED MODERATE CONVECTION FROM 6N-10N BETWEEN 77W-83W. [...]

HISPANIOLA...
PRESENTLY A VERY SMALL AREA OF SCATTERED SHOWERS ARE OVER THE WESTERN DOMINICAN REPUBLIC DUE TO AFTERNOON HEATING. EXPECT MOSTLY FAIR WEATHER OVER THE NEXT 24 HOURS.

ATLANTIC OCEAN...
A 1019 MB HIGH IS OVER THE W ATLANTIC NEAR 28N64W. A SURFACE RIDGE AXIS EXTENDS W TO MIAMI FLORIDA. ISOLATED MODERATE CONVECTION IS OVER THE NW ATLANTIC W OF 70W. [...]

FOR ADDITIONAL INFORMATION PLEASE VISIT
HTTP://WWW.HURRICANES.GOV/MARINE

\$\$
FORMOSA
\end{Verbatim}
  \caption{Selected text from a Tropical Weather Discussion \href{https://www.nhc.noaa.gov/archive/text/TWDAT/2013/TWDAT.201307241749.txt}{from the NHC archive, with WMO header time 1749 UTC on 24 July 2013}. The bulletin itself is headed 205 PM EDT (1805 UTC). Bracketed ellipses mark omitted passages. This basin-wide discussion is included for comparison with storm-specific discussions and was not used in the reported quantitative results.}
  \label{fig:twdat_preprocessing}
\par\endgroup

For the end-to-end RoBERTa models, the processed storm discussion is tokenized with the pretrained tokenizer using a maximum sequence length of 512 tokens. The same storm-discussion routine is applied to training, validation, test, and attribution inputs. The TWDAT example in \autoref{fig:twdat_preprocessing} is illustrative. Attribution details are given in \autoref{app:interpretability}.

\FloatBarrier
\section{Complete Experimental Results}

\subsection{Comparison with Traditional ML Methods}
Tables~\ref{tab:full_results} and~\ref{tab:mslp_full} report the separately trained per-lead neural family, using five seeds for each modality and forecast lead. The 24\,h runs in this family are distinct from the primary 24\,h ablation in \autoref{tab:nn_ablation}. We report these experiment families separately across forecast leads. \autoref{tab:neural_populations} lists case and storm counts for all neural families. \autoref{tab:nn_ablation} reports the mean and sample SD of scores calculated separately for five training runs. \autoref{tab:nn_ensemble} scores the combined forecast obtained by averaging predictions from those same runs. The score of an averaged forecast generally differs from the average score of its individual models. \autoref{sec:leadtime_diagnostics} gives reference baselines and target spread to aid interpretation of changing $R^2$ across lead times.

\begin{table}[H]
\centering
\caption{Per-lead neural models for VMAX change, including 24\,h. Entries are mean $\pm$ sample SD across five training runs. RMSE is in kt. SS is $1-\mathrm{RMSE}/\mathrm{SD}_0(y)$ against the test-mean predictor. Population counts appear in Table~\ref{tab:neural_populations}. Bold/underline identify the best/second-best means within each lead, excluding D's different population.}
\label{tab:full_results}
\small
\renewcommand{\arraystretch}{0.90}
\setlength{\tabcolsep}{4pt}
\begin{adjustbox}{max width=\linewidth}
\begin{tabular}{@{}lcccc@{}}
\toprule
\multirow{2}{*}{Lead} & \multirow{2}{*}{Modality} & \multicolumn{3}{c}{Predictive performance} \\
\cmidrule(lr){3-5}
 &  & $R^2$ & RMSE & SS \\
\midrule
\multirow{7}{*}{24\,h} & T & $0.44\pm0.01$ & $14.22\pm0.12$ & $0.25\pm0.01$ \\
 & I & $0.21\pm0.03$ & $16.93\pm0.29$ & $0.11\pm0.02$ \\
 & D & $0.53\pm0.01$ & $13.27\pm0.17$ & $0.31\pm0.01$ \\
 & T+I & $0.46\pm0.02$ & $13.98\pm0.30$ & $0.27\pm0.02$ \\
 & T+D & $\underline{0.54}\pm0.04$ & $\underline{12.92}\pm0.50$ & $\underline{0.32}\pm0.03$ \\
 & I+D & $0.47\pm0.03$ & $13.84\pm0.36$ & $0.27\pm0.02$ \\
 & T+I+D & $\boldsymbol{0.55}\pm0.02$ & $\boldsymbol{12.77}\pm0.35$ & $\boldsymbol{0.33}\pm0.02$ \\
\midrule
\multirow{7}{*}{48\,h} & T & $0.47\pm0.02$ & $20.78\pm0.38$ & $0.27\pm0.01$ \\
 & I & $0.16\pm0.06$ & $26.03\pm0.95$ & $0.09\pm0.03$ \\
 & D & $0.54\pm0.01$ & $19.23\pm0.25$ & $0.32\pm0.01$ \\
 & T+I & $0.45\pm0.02$ & $21.14\pm0.34$ & $0.26\pm0.01$ \\
 & T+D & $\boldsymbol{0.58}\pm0.02$ & $\boldsymbol{18.43}\pm0.36$ & $\boldsymbol{0.35}\pm0.01$ \\
 & I+D & $0.52\pm0.03$ & $19.78\pm0.54$ & $0.31\pm0.02$ \\
 & T+I+D & $\underline{0.57}\pm0.01$ & $\underline{18.72}\pm0.26$ & $\underline{0.34}\pm0.01$ \\
\midrule
\multirow{7}{*}{72\,h} & T & $0.39\pm0.02$ & $23.78\pm0.47$ & $0.22\pm0.02$ \\
 & I & $0.24\pm0.06$ & $26.44\pm1.04$ & $0.13\pm0.03$ \\
 & D & $0.42\pm0.06$ & $23.19\pm1.12$ & $0.24\pm0.04$ \\
 & T+I & $0.41\pm0.03$ & $23.37\pm0.64$ & $\underline{0.23}\pm0.02$ \\
 & T+D & $\boldsymbol{0.46}\pm0.06$ & $\boldsymbol{22.39}\pm1.22$ & $\boldsymbol{0.26}\pm0.04$ \\
 & I+D & $0.39\pm0.10$ & $23.68\pm1.86$ & $0.22\pm0.06$ \\
 & T+I+D & $\underline{0.45}\pm0.04$ & $\underline{22.51}\pm0.94$ & $\boldsymbol{0.26}\pm0.03$ \\
\midrule
\multirow{7}{*}{96\,h} & T & $\underline{0.27}\pm0.03$ & $\underline{24.97}\pm0.59$ & $\boldsymbol{0.15}\pm0.02$ \\
 & I & $0.18\pm0.07$ & $26.50\pm1.16$ & $0.10\pm0.04$ \\
 & D & $0.19\pm0.11$ & $26.29\pm1.76$ & $0.10\pm0.06$ \\
 & T+I & $\boldsymbol{0.28}\pm0.08$ & $\boldsymbol{24.81}\pm1.38$ & $\boldsymbol{0.15}\pm0.05$ \\
 & T+D & $0.24\pm0.07$ & $25.56\pm1.22$ & $\underline{0.13}\pm0.04$ \\
 & I+D & $0.08\pm0.19$ & $27.98\pm2.93$ & $0.05\pm0.10$ \\
 & T+I+D & $0.19\pm0.08$ & $26.27\pm1.37$ & $0.10\pm0.05$ \\
\midrule
\multirow{7}{*}{120\,h} & T & $\boldsymbol{0.21}\pm0.03$ & $\boldsymbol{26.11}\pm0.43$ & $\boldsymbol{0.11}\pm0.01$ \\
 & I & $\underline{0.15}\pm0.04$ & $\underline{26.95}\pm0.62$ & $\underline{0.08}\pm0.02$ \\
 & D & $0.00\pm0.15$ & $29.33\pm2.15$ & $0.00\pm0.07$ \\
 & T+I & $0.08\pm0.09$ & $28.02\pm1.40$ & $0.04\pm0.05$ \\
 & T+D & $0.06\pm0.07$ & $28.43\pm1.13$ & $0.03\pm0.04$ \\
 & I+D & $0.05\pm0.18$ & $28.40\pm2.81$ & $0.03\pm0.10$ \\
 & T+I+D & $0.00\pm0.13$ & $29.29\pm1.94$ & $0.00\pm0.07$ \\
\bottomrule
\end{tabular}
\end{adjustbox}
\end{table}

\subsection{MSLP Prediction Results}
Table~\ref{tab:mslp_full} reports the corresponding pressure results for the separately trained per-lead neural models. The following tables give the evaluation populations (Table~\ref{tab:neural_populations}) and compare the primary 24\,h models for both targets, using the mean and sample SD of scores across five runs (Table~\ref{tab:nn_ablation}) and the scores of their averaged forecast (Table~\ref{tab:nn_ensemble}).

\begin{table}[H]
\centering
\caption{Per-lead neural models for MSLP change, including 24\,h. Entries are mean $\pm$ sample SD across five training runs. RMSE is in hPa. SS is $1-\mathrm{RMSE}/\mathrm{SD}_0(y)$ against the test-mean predictor. Population counts appear in Table~\ref{tab:neural_populations}. Bold/underline identify the best/second-best means within each lead, excluding D's different population.}
\label{tab:mslp_full}
\small
\renewcommand{\arraystretch}{0.90}
\setlength{\tabcolsep}{4pt}
\begin{adjustbox}{max width=\linewidth}
\begin{tabular}{@{}lcccc@{}}
\toprule
\multirow{2}{*}{Lead} & \multirow{2}{*}{Modality} & \multicolumn{3}{c}{Predictive performance} \\
\cmidrule(lr){3-5}
 &  & $R^2$ & RMSE & SS \\
\midrule
\multirow{7}{*}{24\,h} & T & $0.44\pm0.01$ & $9.74\pm0.09$ & $0.25\pm0.01$ \\
 & I & $0.18\pm0.03$ & $11.82\pm0.22$ & $0.09\pm0.02$ \\
 & D & $0.48\pm0.02$ & $9.51\pm0.16$ & $0.28\pm0.01$ \\
 & T+I & $0.46\pm0.03$ & $9.58\pm0.22$ & $0.27\pm0.02$ \\
 & T+D & $\underline{0.50}\pm0.04$ & $\underline{9.21}\pm0.36$ & $\underline{0.29}\pm0.03$ \\
 & I+D & $0.47\pm0.03$ & $9.51\pm0.23$ & $0.27\pm0.02$ \\
 & T+I+D & $\boldsymbol{0.52}\pm0.03$ & $\boldsymbol{9.03}\pm0.31$ & $\boldsymbol{0.31}\pm0.02$ \\
\midrule
\multirow{7}{*}{48\,h} & T & $0.44\pm0.02$ & $14.80\pm0.24$ & $0.25\pm0.01$ \\
 & I & $0.12\pm0.06$ & $18.49\pm0.65$ & $0.06\pm0.03$ \\
 & D & $0.51\pm0.01$ & $13.87\pm0.20$ & $0.30\pm0.01$ \\
 & T+I & $0.42\pm0.03$ & $15.01\pm0.40$ & $0.24\pm0.02$ \\
 & T+D & $\boldsymbol{0.55}\pm0.02$ & $\boldsymbol{13.15}\pm0.32$ & $\boldsymbol{0.33}\pm0.02$ \\
 & I+D & $0.49\pm0.02$ & $14.06\pm0.31$ & $0.29\pm0.02$ \\
 & T+I+D & $\underline{0.54}\pm0.02$ & $\underline{13.34}\pm0.26$ & $\underline{0.32}\pm0.01$ \\
\midrule
\multirow{7}{*}{72\,h} & T & $0.33\pm0.03$ & $17.52\pm0.39$ & $0.18\pm0.02$ \\
 & I & $0.15\pm0.08$ & $19.67\pm0.89$ & $0.08\pm0.04$ \\
 & D & $0.35\pm0.06$ & $17.36\pm0.80$ & $0.19\pm0.04$ \\
 & T+I & $0.31\pm0.10$ & $17.69\pm1.21$ & $0.17\pm0.06$ \\
 & T+D & $\boldsymbol{0.39}\pm0.06$ & $\boldsymbol{16.66}\pm0.80$ & $\boldsymbol{0.22}\pm0.04$ \\
 & I+D & $\underline{0.36}\pm0.05$ & $\underline{17.09}\pm0.66$ & $\underline{0.20}\pm0.03$ \\
 & T+I+D & $\boldsymbol{0.39}\pm0.06$ & $\boldsymbol{16.66}\pm0.89$ & $\boldsymbol{0.22}\pm0.04$ \\
\midrule
\multirow{7}{*}{96\,h} & T & $0.18\pm0.05$ & $19.09\pm0.57$ & $\underline{0.10}\pm0.03$ \\
 & I & $0.07\pm0.09$ & $20.30\pm0.94$ & $0.04\pm0.04$ \\
 & D & $0.14\pm0.14$ & $19.73\pm1.57$ & $0.07\pm0.07$ \\
 & T+I & $\boldsymbol{0.20}\pm0.06$ & $\boldsymbol{18.83}\pm0.75$ & $\boldsymbol{0.11}\pm0.04$ \\
 & T+D & $\underline{0.19}\pm0.09$ & $\underline{19.02}\pm1.08$ & $\underline{0.10}\pm0.05$ \\
 & I+D & $0.03\pm0.22$ & $20.72\pm2.34$ & $0.02\pm0.11$ \\
 & T+I+D & $0.18\pm0.08$ & $19.12\pm0.94$ & $\underline{0.10}\pm0.04$ \\
\midrule
\multirow{7}{*}{120\,h} & T & $\boldsymbol{0.12}\pm0.05$ & $\boldsymbol{20.55}\pm0.59$ & $\boldsymbol{0.06}\pm0.03$ \\
 & I & $0.05\pm0.03$ & $21.33\pm0.33$ & $0.02\pm0.01$ \\
 & D & $0.00\pm0.15$ & $22.00\pm1.56$ & $0.00\pm0.07$ \\
 & T+I & $0.05\pm0.10$ & $21.28\pm1.08$ & $\underline{0.03}\pm0.05$ \\
 & T+D & $\underline{0.06}\pm0.07$ & $\underline{21.17}\pm0.74$ & $\underline{0.03}\pm0.03$ \\
 & I+D & $0.02\pm0.15$ & $21.61\pm1.72$ & $0.01\pm0.08$ \\
 & T+I+D & $0.01\pm0.10$ & $21.74\pm1.09$ & $0.01\pm0.05$ \\
\bottomrule
\end{tabular}
\end{adjustbox}
\end{table}

\begin{table}[H]
\centering
\caption{Evaluation populations for both targets and all neural families. D requires a matched discussion. Other modalities retain cases with missing discussions.}
\label{tab:neural_populations}
\small
\renewcommand{\arraystretch}{1.12}
\setlength{\tabcolsep}{4pt}
\begin{adjustbox}{max width=\linewidth}
\begin{tabular}{@{}lcccc@{}}
\toprule
\multirow{2}{*}{Lead} & \multicolumn{2}{c}{All modalities except D} & \multicolumn{2}{c}{D only} \\
\cmidrule(lr){2-3} \cmidrule(lr){4-5}
 & Cases & Storms & Cases & Storms \\
\midrule
24\,h & 1{,}150 & 66 & 1{,}105 & 64 \\
48\,h & 898 & 53 & 860 & 53 \\
72\,h & 686 & 51 & 653 & 47 \\
96\,h & 499 & 38 & 480 & 36 \\
120\,h & 360 & 28 & 348 & 27 \\
\bottomrule
\end{tabular}
\end{adjustbox}
\end{table}

\begin{table}[H]
\centering
\caption{Primary 24\,h neural ablation: mean $\pm$ sample SD over five runs. Text-fusion models use the forecast-tuned discussion encoder. Bold and underline mark the best and second-best means, excluding D's different population (Table~\ref{tab:neural_populations}).}
\label{tab:nn_ablation}
\small
\renewcommand{\arraystretch}{1.12}
\setlength{\tabcolsep}{4pt}
\begin{adjustbox}{max width=\linewidth}
\begin{tabular}{@{}lcccc@{}}
\toprule
\multirow{2}{*}{Modality} & \multicolumn{2}{c}{$R^2$} & \multicolumn{2}{c}{MAE (kt/hPa)} \\
\cmidrule(lr){2-3} \cmidrule(lr){4-5}
 & $V_{\max}$ & MSLP & $V_{\max}$ & MSLP \\
\midrule
\multicolumn{5}{@{}l}{\textit{Single}} \\
T & $0.44\pm0.01$ & $0.44\pm0.01$ & $10.13\pm0.05$ & $6.60\pm0.08$ \\
I & $0.22\pm0.02$ & $0.20\pm0.02$ & $12.40\pm0.16$ & $8.15\pm0.13$ \\
D & $0.50\pm0.01$ & $0.46\pm0.03$ & $9.90\pm0.20$ & $6.56\pm0.17$ \\
\midrule
\multicolumn{5}{@{}l}{\textit{Dual}} \\
T+I & $0.45\pm0.01$ & $0.45\pm0.01$ & $10.27\pm0.06$ & $6.72\pm0.05$ \\
T+D & $\boldsymbol{0.53}\pm0.01$ & $\boldsymbol{0.51}\pm0.01$ & $\underline{9.46}\pm0.16$ & $\boldsymbol{6.23}\pm0.08$ \\
I+D & $\underline{0.50}\pm0.01$ & $\underline{0.48}\pm0.02$ & $9.74\pm0.27$ & $\underline{6.45}\pm0.18$ \\
\midrule
\multicolumn{5}{@{}l}{\textit{Triple}} \\
T+I+D & $\boldsymbol{0.53}\pm0.01$ & $\boldsymbol{0.51}\pm0.01$ & $\boldsymbol{9.45}\pm0.13$ & $\boldsymbol{6.23}\pm0.08$ \\
\bottomrule
\end{tabular}
\end{adjustbox}
\end{table}

\begin{table}[H]
\centering
\caption{Primary 24\,h neural ablation after averaging predictions from five runs. Models and highlighting follow Table~\ref{tab:nn_ablation}, with populations in Table~\ref{tab:neural_populations}.}
\label{tab:nn_ensemble}
\small
\renewcommand{\arraystretch}{1.12}
\setlength{\tabcolsep}{4pt}
\fullwidthresulttable{lcccc}{
\toprule
\multirow{2}{*}{Modality} & \multicolumn{2}{c}{$R^2$} & \multicolumn{2}{c}{MAE (kt/hPa)} \\
\cmidrule(lr){2-3} \cmidrule(lr){4-5}
 & $V_{\max}$ & MSLP & $V_{\max}$ & MSLP \\
\midrule
\multicolumn{5}{@{}l}{\textit{Single}} \\
T & $0.46$ & $0.46$ & $9.89$ & $6.43$ \\
I & $0.27$ & $0.25$ & $11.99$ & $7.85$ \\
D & $0.55$ & $0.51$ & $9.28$ & $6.16$ \\
\midrule
\multicolumn{5}{@{}l}{\textit{Dual}} \\
T+I & $0.49$ & $0.49$ & $9.84$ & $6.35$ \\
T+D & $\boldsymbol{0.57}$ & $\boldsymbol{0.55}$ & $\boldsymbol{8.90}$ & $\boldsymbol{5.87}$ \\
I+D & $\underline{0.55}$ & $\underline{0.53}$ & $9.11$ & $6.01$ \\
\midrule
\multicolumn{5}{@{}l}{\textit{Triple}} \\
T+I+D & $\boldsymbol{0.57}$ & $\boldsymbol{0.55}$ & $\underline{8.92}$ & $\underline{5.88}$ \\
\bottomrule
}
\end{table}

\FloatBarrier
\section{Data Processing Pipeline}\label{app:data_processing}

\subsection{Tabular Data Preprocessing}
The tabular input consists of the 13 SHIPS predictors in \autoref{tab:variables}. SHIPS values are extracted at the initial time. Latitude and longitude are expressed in degrees. For each storm, a target at lead time $h$ is formed by subtracting the initial intensity from the intensity at exactly $t+h$ when it is available. Storm identifiers, timestamps, discussion filenames, and columns containing future targets are excluded from the predictor vector. The models use these 13 predictors without additional cyclical time encodings, lagged predictors, interaction terms, or outlier clipping.

The dataset contains \resultvalue{dataset/missing_twac} missing TWAC values. The other 12 predictors contain no missing entries. In the end-to-end loader, missing numerical inputs are replaced by \mbox{-999} before transformation by the feature scaler. The scaler itself is fitted on available training values before this replacement. In the hybrid and temporal loaders, missing inputs are replaced by zero before fitting the scaler. Each fitted scaler is then applied unchanged to validation and test inputs. The chronological split uses cases before 2020 for training, 2020--2021 for validation, and 2022--2023 for testing. Target availability is handled separately from input missingness. Absent future intensities are not imputed. Neural losses use target masks, and per-lead hybrid and temporal datasets require the targets needed by the fitted model.

\subsubsection{Pairwise correlations among tabular predictors}
\label{app:tab_correlations}

\begin{figure}[!htbp]
  \centering
  \includegraphics[width=0.9\linewidth]{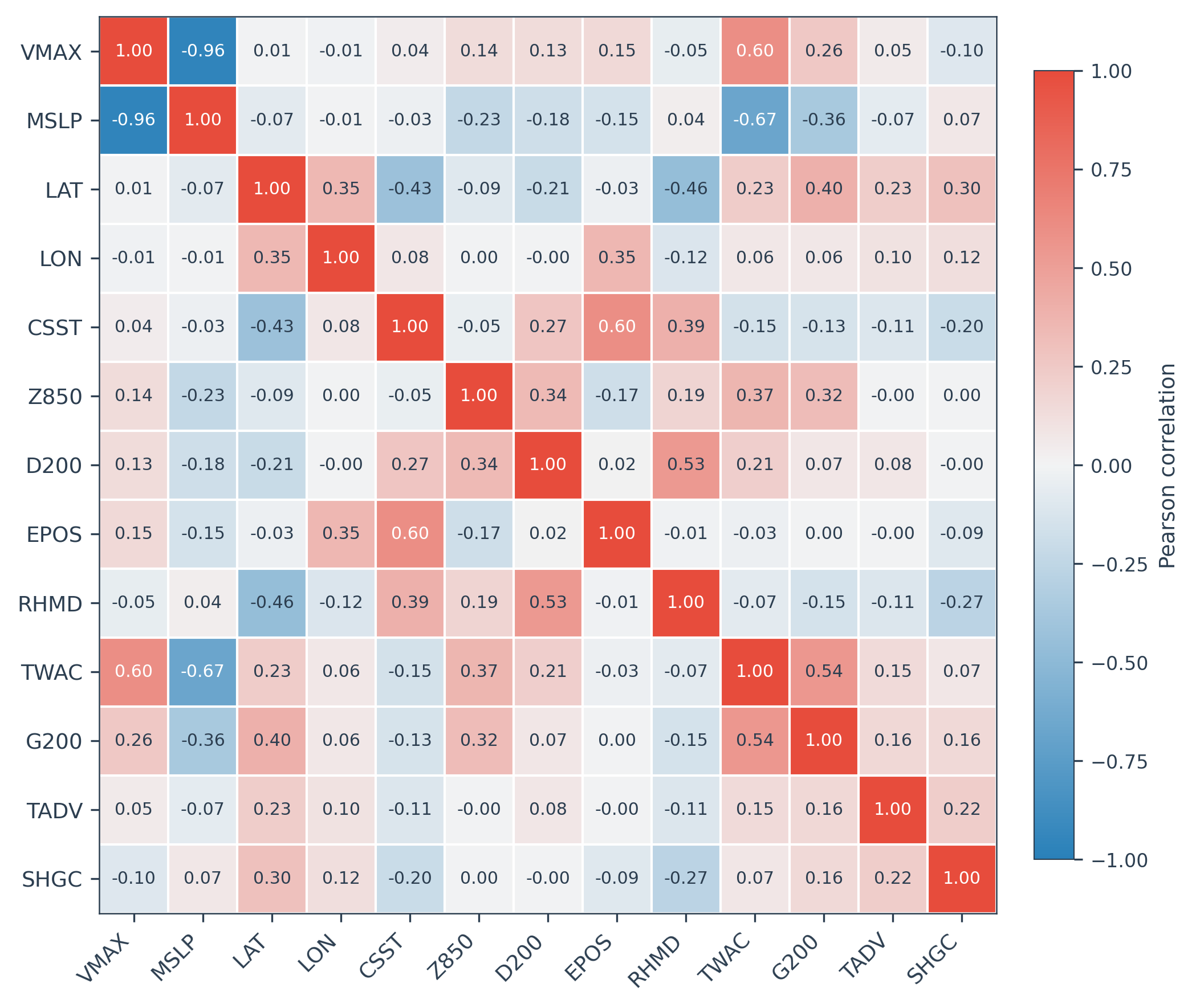}
  \caption{Pairwise Pearson correlations among all 13 SHIPS tabular predictors, using available pairs of training cases.}
  \label{fig:tabular_correlations}
\end{figure}

\autoref{fig:tabular_correlations} reports pairwise Pearson correlations among the SHIPS predictors in \autoref{tab:variables}, computed on the training split. Current wind and pressure are strongly inversely correlated, and the tangential-wind predictor also tracks those current-state variables. Most other environmental predictors have weaker pairwise correlations. Correlation alone does not establish that every predictor is necessary, so we retain the declared input set for comparability across configurations.

\subsection{ERA5 Image Processing}
\begin{enumerate}
\item \textbf{Data Extraction.}
   \begin{itemize}
   \item 41$\times$41 grid centered on storm location
   \item Resolution is 0.25\textdegree\ $\times$ 0.25\textdegree
   \item Variables are $u_{10}$, $v_{10}$ (10m wind components), used as raw two-channel input
   \end{itemize}

\item \textbf{Preprocessing.}
   \begin{itemize}
   \item Channel 0 contains $u_{10}$ (zonal wind component)
   \item Channel 1 contains $v_{10}$ (meridional wind component)
   \item Zero-padded frames preserved as zero (representing no data)
   \item Channel-wise normalization using training statistics (excluding zero-padded frames)
   \end{itemize}
\end{enumerate}

\subsection{Image Representation Comparison}\label{sec:magdir_comparison}
The retained ERA5 inputs use the horizontal wind components $u$ and $v$. They were obtained from aligned magnitude/direction fields through $u=|V|\cos\theta$ and $v=|V|\sin\theta$, where $|V|=\sqrt{u^2+v^2}$ and $\theta=\arctan_2(v,u)$. The coordinate change does not add meteorological information. Direction is undefined at zero wind speed, while both reconstructed components are zero. All reported predictions use the fixed $u/v$ representation. A controlled comparison of predictive performance under the two representations remains future work.

Physically, 10\,m surface winds are the most natural image representation for $V_{\max}$, which is itself a maximum sustained surface wind speed. However, ERA5 strongly underestimates the maximum winds of intense cyclones \citep{Dulac2024ERA5}, so these fields mainly describe the broader circulation. The tabular inputs include a generalized wind-shear measure (\texttt{SHGC}) and upper-level divergence (\texttt{D200}). Their presence does not establish that these predictors contain all useful information from upper-air images. An exploratory comparison with ERA5 850\,hPa $u/v$ wind components, with or without 850\,hPa geopotential height, is left for future work.

\subsection{Image Encoder Choice}\label{sec:encoder_comparison}
For the hybrid image branch we use a CNN trained on ERA5 wind fields. It produces 128-dimensional embeddings using the self-supervised objective described in \autoref{sec:pretrain_cnn_details}. We also explored DINOv2-ViT-S/14~\citep{oquab2024dinov2}, a vision foundation model pretrained on natural images. These exploratory experiments did not provide a useful alternative in our pipeline, so we did not retain DINOv2. All reported hybrid results use the CNN trained on ERA5 wind fields.

\subsection{Hybrid CNN pretraining objective}\label{sec:pretrain_cnn_details}

The hybrid image encoder is trained self-supervised on ERA5 frames alone, without any $V_{\max}$ or MSLP target, so that image embeddings can be produced even for frames that lack matched intensity labels. Optimization uses only frames from 2002--2019 storms, with checkpoint selection on 2020--2021 storms. Pretraining uses learning rate $5\times10^{-4}$, batch size 256 and an epoch cap of 50, repeated independently with five seeds. In each run, the CNN and hybrid regressors use the same random seed. Frames from the 2022--2023 test storms are excluded from pretraining. Two objectives are optimized jointly. (i) A \emph{next-frame prediction} objective uses a small head that reads the encoder's embedding of frame $t$ and predicts the embedding of frame $t{+}1$ of the same storm, using one minus cosine similarity to a detached next-frame embedding as the loss. (ii) A frame-\emph{reconstruction} objective uses a lightweight decoder that reconstructs the input frame from the encoder embedding under a pixelwise MSE loss. The two losses are summed with a weight of $0.5$ on the reconstruction term. After pretraining the predictor and the decoder are discarded. The 128-dimensional encoder output is passed to the XGBoost fusion head together with the 13 raw SHIPS predictors and the text embeddings from the sentence encoder.

The hybrid and temporal CNNs differ from the end-to-end image encoder. They use three $3\times3$ convolutional layers with channel widths $64,64,256$, BatchNorm and ReLU, max pooling after the second and third layers, and average pooling to $4\times4$. Their FC path is $4096\to576\to256\to128$, with Linear--BatchNorm--ReLU in its first two blocks. Each hybrid run uses its own fixed pretrained encoder. The temporal models train their own initially random CNN and GRU.

\subsection{Text Data Processing}
Storm-specific discussions are joined to cases using the bounded matching rule in \autoref{sec:curation}. Matched discussion issue times occur 3--7 hours after the synoptic time in the dataset, usually three hours. Discussions for unmatched cases remain missing. Fused models and the hybrid and temporal text-only regressors use a zero text vector, and the neural discussion-only model omits those cases.

The end-to-end RoBERTa path trims the bulletin header to the first datetime line, removes the \texttt{FORECAST POSITIONS AND MAX WINDS} section and all following lines, removes null/control and non-ASCII characters, collapses whitespace, and lowercases the retained text. Its tokenizer truncates and pads to a total sequence length of 512 tokens, including special tokens, and supplies an attention mask. Narrative numerical predictions may remain in the retained discussion.

The frozen Qwen path uses the same forecast-section cutoff but otherwise preserves the discussion's letter case, bulletin header, and line structure before encoding. The cache stores 1,024-dimensional normalized embeddings. The hybrid and temporal models use the first 768 coordinates, without an additional normalization after this truncation. Missing or empty matched inputs produce zero embeddings. These two text preprocessing paths are distinct.

\FloatBarrier
\section{Encoder Architecture Details}\label{sec:encoder_details}

This appendix provides full mathematical specifications for each encoder and the fusion module summarized in Section~\ref{sec:e2e}.

\subsection{Tabular Encoder}

The tabular encoder processes the 13-dimensional SHIPS predictor vector $\mathbf{x}_{\text{tab}} \in \mathbb{R}^{13}$ through a three-layer MLP with progressively decreasing hidden dimensions.
\begin{equation}
\mathbf{h}^{(1)} = \text{LeakyReLU}_{\alpha}(\mathbf{W}_1^{\text{tab}} \mathbf{x}_{\text{tab}} + \mathbf{b}_1^{\text{tab}}),
\end{equation}
where $\mathbf{W}_1^{\text{tab}} \in \mathbb{R}^{128 \times 13}$, $\mathbf{b}_1^{\text{tab}} \in \mathbb{R}^{128}$, and $\alpha = 0.01$ (default PyTorch negative slope).

The second hidden layer incorporates normalization and regularization.
\begin{align}
\tilde{\mathbf{h}}^{(2)} &= \text{LeakyReLU}_{\alpha}(\mathbf{W}_2^{\text{tab}} \mathbf{h}^{(1)} + \mathbf{b}_2^{\text{tab}}), \nonumber \\
\mathbf{h}^{(2)} &= \text{Dropout}_p(\text{LayerNorm}(\tilde{\mathbf{h}}^{(2)})),
\end{align}
where $\mathbf{W}_2^{\text{tab}} \in \mathbb{R}^{64 \times 128}$ and dropout rate $p=0.3$.

A final linear projection produces the 64-dimensional embedding.
\begin{equation}
\mathbf{h}_{\text{tab}} = \mathbf{W}_3^{\text{tab}} \mathbf{h}^{(2)} + \mathbf{b}_3^{\text{tab}} \in \mathbb{R}^{64},
\end{equation}
where $\mathbf{W}_3^{\text{tab}} \in \mathbb{R}^{64 \times 64}$. Standalone T replaces the final 64-dimensional embedding with $2H$ forecast outputs.

\subsection{Image Encoder}

The image encoder extracts spatial features from ERA5 wind fields $\mathbf{x}_{\text{img}} \in \mathbb{R}^{2 \times 41 \times 41}$, where the two channels represent $u$ and $v$ wind components at 10\,m height. The CNN consists of three convolutional blocks with increasing channel depths. Each block applies convolution, batch normalization, and ReLU activation.
\begin{equation}
\tilde{\mathbf{f}}^{(i)} = \text{ReLU}(\text{BatchNorm}(\text{Conv2d}_i(\mathbf{f}^{(i-1)}))),
\end{equation}
where $\mathbf{f}^{(0)} = \mathbf{x}_{\text{img}}$. The first block uses a $5 \times 5$ kernel with 32 output channels, followed by $2 \times 2$ max pooling ($41 \times 41 \to 20 \times 20$). The second block uses a $3 \times 3$ kernel with 64 channels and another $2 \times 2$ max pooling ($20 \times 20 \to 10 \times 10$). The third block uses a $3 \times 3$ kernel with 128 channels, followed by adaptive average pooling to a fixed $4 \times 4$ spatial grid.
\begin{equation}
\mathbf{f}_{\text{pool}} = \text{AdaptiveAvgPool}_{4 \times 4}(\tilde{\mathbf{f}}^{(3)}) \in \mathbb{R}^{128 \times 4 \times 4}.
\end{equation}

The pooled features pass through three fully connected layers. Each of the first two is followed by ReLU, batch normalization and dropout, in that order. Their dropout probabilities are $p$ and $0.67p$, with $p=0.1$ for primary I and T+I and $p=0.3$ for the other reported image configurations. The final layer is linear.
\begin{equation}
\mathbf{h}_{\text{img}} = \text{FC}(\text{Flatten}(\mathbf{f}_{\text{pool}})) \in \mathbb{R}^{64}, \quad 2048 \to 128 \to 64 \to 64.
\end{equation}

In standalone I, the final output dimension is $2H$ instead of 64.

\subsection{Text Encoder}

The text encoder uses RoBERTa-base~\citep{roberta2019} as a feature extractor. Discussions are tokenized via byte-pair encoding (maximum 512 tokens) and processed through the 12-layer transformer encoder. We extract the [CLS] token hidden state.
\begin{equation}
\mathbf{e}_{\text{CLS}} = RoBERTa(\text{Tokenize}(\mathbf{x}_{\text{text}}))[\text{CLS}] \in \mathbb{R}^{768}.
\end{equation}

A two-layer projection maps the 768-dimensional representation to the common embedding space.
\begin{align}
\mathbf{h}_{\text{text}}^{(1)} &= \text{GELU}(\mathbf{W}_1^{\text{txt}} \mathbf{e}_{\text{CLS}} + \mathbf{b}_1^{\text{txt}}), \\
\mathbf{h}_{\text{text}} &= \mathbf{W}_2^{\text{txt}} \text{Dropout}_{0.1}(\mathbf{h}_{\text{text}}^{(1)}) + \mathbf{b}_2^{\text{txt}} \in \mathbb{R}^{64},
\end{align}
where $\mathbf{W}_1^{\text{txt}} \in \mathbb{R}^{384 \times 768}$, $\mathbf{W}_2^{\text{txt}} \in \mathbb{R}^{64 \times 384}$, with GELU activation and dropout ($p{=}0.1$). In fused models, the backbone is fixed while the projection is trained on precomputed [CLS] features. Generic text uses pretrained RoBERTa weights, while forecast-tuned text uses the separately trained D backbone for the same seed and lead time. Standalone D trains the full pretrained backbone and a $768\to384\to2H$ regression head. Its head uses GELU and dropout with probability 0.1 between the linear layers, with another dropout at its output during training.

\subsection{Multi-modal Fusion}

Before fusion, every active branch output receives layer normalization. The image branch additionally applies a $64\to64$ linear projection, LeakyReLU and another layer normalization. These operations act on learned features, not physical intensity values.

\textbf{Concatenation Fusion.} Modality embeddings are concatenated directly.
\begin{equation}
\mathbf{h}_{\text{stack}} = [\tilde{\mathbf{h}}_{\text{tab}};\, \tilde{\mathbf{h}}_{\text{img}};\, \tilde{\mathbf{h}}_{\text{text}}] \in \mathbb{R}^{192},
\end{equation}
where tildes denote the branch transformations above. With $M$ active modalities the stacked width is $64M$. Global layer normalization gives the fused representation, followed by dropout ($p{=}0.1$) and a linear output layer.

\textbf{Attention-Based Fusion.} Each transformed modality embedding is projected to $d_{\text{model}}=256$ dimensions by a linear layer, GELU and dropout with probability 0.1. A learnable \textsc{[fusion]} token is prepended and modality-position embeddings are added. For $M$ modalities this gives $\mathbf H\in\mathbb R^{(M+1)\times256}$. A single pre-LN Transformer layer uses four heads of width $d_k=64$.
\begin{equation}
\text{Attention}(\mathbf Q,\mathbf K,\mathbf V)=
\text{softmax}\!\left(\frac{\mathbf Q\mathbf K^T}{\sqrt{d_k}}\right)\mathbf V,
\end{equation}
where $\mathbf Q=\mathrm{LN}(\mathbf H)\mathbf W_Q$, $\mathbf K=\mathrm{LN}(\mathbf H)\mathbf W_K$ and $\mathbf V=\mathrm{LN}(\mathbf H)\mathbf W_V$. A residual connection surrounds this sublayer. A second pre-LN sublayer applies a $256\to1024\to256$ feed-forward network with GELU and its own residual connection. Internal and residual dropout probabilities are 0.2. The resulting \textsc{[fusion]} token is layer-normalized, then passed through dropout with probability 0.1 and a linear output head. The architecture supports omitting modality tokens. In the reported fused experiments, an unavailable discussion is represented by a zero input embedding and its token is retained.

The primary 24\,h grid uses concatenation for T+I, T+D and T+I+D, and attention for I+D. Per-lead models retain these choices at 24\,h. At 48--120\,h they use attention for T+I and I+D and concatenation for T+D and T+I+D. All joint multi-lead fusion models use concatenation. The attention diagram illustrates the module rather than the selected fusion for every configuration.

\subsection{Prediction Head}

The fusion module's final linear layer maps the fused representation to standardized intensity-change predictions for the requested lead times.
\begin{equation}
\Delta\hat{\mathbf{y}}_{\mathrm{std}} = \mathbf{W}_{\text{pred}} \mathbf{h}_{\text{fused}} + \mathbf{b}_{\text{pred}} \in \mathbb{R}^{2H},
\end{equation}
where $H$ is the number of lead times ($H{=}5$ for 24--120\,h) and the output lists standardized wind changes for all requested lead times followed by standardized pressure changes for the same lead times. Physical predictions $\Delta\hat{\mathbf{y}}$ are recovered using target means and standard deviations from the training set.

\FloatBarrier
\section{Implementation Details}\label{app:implementation_details}

\subsection{Model Architectures}

\begin{figure}[h]
\centering
\resizebox{\textwidth}{!}{
\begin{tikzpicture}[
  every node/.style={font=\sffamily},
  layer/.style={mcbase, fill=mcEncoder, draw=mcEncoderBdr,
    text width=2.8cm, minimum height=0.9cm, font=\sffamily\scriptsize},
  act/.style={mcbase, fill=mcEmbed!60, draw=mcEmbedBdr!70,
    text width=2.8cm, minimum height=0.9cm, font=\sffamily\scriptsize},
  pool/.style={mcbase, fill=mcTemporal, draw=mcTemporalBdr,
    text width=2.8cm, minimum height=0.9cm, font=\sffamily\scriptsize},
  attn/.style={mcbase, fill=mcFusion, draw=mcFusionBdr,
    text width=2.8cm, minimum height=0.9cm, font=\sffamily\scriptsize},
  io/.style={mcbase, fill=mcInput, draw=mcInputBdr,
    text width=2.8cm, minimum height=0.9cm, font=\sffamily\scriptsize},
]

\def\colA{0}        
\def\colB{10.0}     
\def\colC{4.6}      
\def\colD{15.4}     

\node[mcstage] at (\colA, 0.6) {\textbf{Tabular Encoder}};

\node[io] (ta_in) at (\colA, -0.4) {Input $\mathbb{R}^{13}$};

\node[layer] (ta_fc1) at (\colA, -2.0) {FC + LeakyReLU\\$13 \to 128$};
\node[layer] (ta_fc2) at (\colA, -3.8) {FC + LeakyReLU\\$128 \to 64$};
\node[act] (ta_ln) at (\colA, -5.6) {LayerNorm + Drop(0.3)};
\node[layer] (ta_fc3) at (\colA, -7.4) {FC $64 \to 64$};

\node[io] (ta_out) at (\colA, -10.9) {LayerNorm\\Output $\mathbb{R}^{64}$};

\draw[mcarrow] (ta_in) -- (ta_fc1);
\draw[mcarrow] (ta_fc1) -- (ta_fc2);
\draw[mcarrow] (ta_fc2) -- (ta_ln);
\draw[mcarrow] (ta_ln) -- (ta_fc3);
\draw[mcarrow] (ta_fc3) -- (ta_out);

\node[mcstage] at (\colB, 0.6) {\textbf{CNN Encoder}};

\node[io] (cn_in) at (\colB, -0.4) {Input $41{\times}41{\times}2$};

\fill[mcEncoder!40] (9.55,-2.4) rectangle ++(0.85,0.8);
\fill[mcEncoder!60] (10.4,-2.4) -- ++(0.2,0.2) -- ++(0,0.8) -- ++(-0.2,-0.2) -- cycle;
\fill[mcEncoder!50] (9.55,-1.6) -- ++(0.2,0.2) -- ++(0.85,0) -- ++(-0.2,-0.2) -- cycle;
\draw[thin, mcEncoderBdr] (9.55,-2.4) rectangle ++(0.85,0.8);
\draw[thin, mcEncoderBdr] (10.4,-2.4) -- ++(0.2,0.2) -- ++(0,0.8) -- ++(-0.2,-0.2);
\draw[thin, mcEncoderBdr] (9.55,-1.6) -- ++(0.2,0.2) -- ++(0.85,0) -- ++(-0.2,-0.2);
\fill[mcEncoder!45] (9.65,-3.95) rectangle ++(0.65,0.65);
\fill[mcEncoder!65] (10.3,-3.95) -- ++(0.3,0.3) -- ++(0,0.65) -- ++(-0.3,-0.3) -- cycle;
\fill[mcEncoder!55] (9.65,-3.3) -- ++(0.3,0.3) -- ++(0.65,0) -- ++(-0.3,-0.3) -- cycle;
\draw[thin, mcEncoderBdr] (9.65,-3.95) rectangle ++(0.65,0.65);
\draw[thin, mcEncoderBdr] (10.3,-3.95) -- ++(0.3,0.3) -- ++(0,0.65) -- ++(-0.3,-0.3);
\draw[thin, mcEncoderBdr] (9.65,-3.3) -- ++(0.3,0.3) -- ++(0.65,0) -- ++(-0.3,-0.3);
\fill[mcEncoder!50] (9.65,-5.4) rectangle ++(0.65,0.65);
\fill[mcEncoder!70] (10.3,-5.4) -- ++(0.48,0.48) -- ++(0,0.65) -- ++(-0.48,-0.48) -- cycle;
\fill[mcEncoder!60] (9.65,-4.75) -- ++(0.48,0.48) -- ++(0.65,0) -- ++(-0.48,-0.48) -- cycle;
\draw[thin, mcEncoderBdr] (9.65,-5.4) rectangle ++(0.65,0.65);
\draw[thin, mcEncoderBdr] (10.3,-5.4) -- ++(0.48,0.48) -- ++(0,0.65) -- ++(-0.48,-0.48);
\draw[thin, mcEncoderBdr] (9.65,-4.75) -- ++(0.48,0.48) -- ++(0.65,0) -- ++(-0.48,-0.48);
\fill[mcTemporal!60] (9.8,-6.75) rectangle ++(0.35,0.35);
\fill[mcTemporal!80] (10.15,-6.75) -- ++(0.48,0.48) -- ++(0,0.35) -- ++(-0.48,-0.48) -- cycle;
\fill[mcTemporal!70] (9.8,-6.4) -- ++(0.48,0.48) -- ++(0.35,0) -- ++(-0.48,-0.48) -- cycle;
\draw[thin, mcTemporalBdr] (9.8,-6.75) rectangle ++(0.35,0.35);
\draw[thin, mcTemporalBdr] (10.15,-6.75) -- ++(0.48,0.48) -- ++(0,0.35) -- ++(-0.48,-0.48);
\draw[thin, mcTemporalBdr] (9.8,-6.4) -- ++(0.48,0.48) -- ++(0.35,0) -- ++(-0.48,-0.48);

\node[layer, text width=2.8cm] (cn_fc) at (\colB, -8.3)
  {Flatten + FC blocks\\$2048 \to 128 \to 64 \to 64$};
\node[io] (cn_out) at (\colB, -10.9) {LayerNorm + projection\\Output $\mathbb{R}^{64}$};

\draw[mcarrow] (cn_in.south) -- (\colB,-1.55);
\draw[mcarrow] (\colB,-2.5) -- (\colB,-3.15);
\draw[mcarrow] (\colB,-4.05) -- (\colB,-4.6);
\draw[mcarrow] (\colB,-5.5) -- (\colB,-6.1);
\draw[mcarrow] (\colB,-6.85) -- (cn_fc.north);
\draw[mcarrow] (cn_fc) -- (cn_out);

\node[font=\sffamily\scriptsize, mcLabel, anchor=west] at (10.7,-1.75) {$20{\times}20{\times}32$};
\node[font=\sffamily\scriptsize, mcEncoderBdr, anchor=east, text width=1.6cm, inner sep=0pt, align=left]
  at (9.25,-1.9) {\textit{Conv1+BN+}\\\textit{ReLU+Pool}};
\node[font=\sffamily\scriptsize, mcLabel, anchor=west] at (10.7,-3.45) {$10{\times}10{\times}64$};
\node[font=\sffamily\scriptsize, mcEncoderBdr, anchor=east, text width=1.6cm, inner sep=0pt, align=left]
  at (9.25,-3.6) {\textit{Conv2+BN+}\\\textit{ReLU+Pool}};
\node[font=\sffamily\scriptsize, mcLabel, anchor=west] at (10.7,-4.85) {$10{\times}10{\times}128$};
\node[font=\sffamily\scriptsize, mcEncoderBdr, anchor=east, text width=1.6cm, inner sep=0pt, align=left]
  at (9.25,-5.0) {\textit{Conv3+BN+}\\\textit{ReLU}};
\node[font=\sffamily\scriptsize, mcLabel, anchor=west] at (10.7,-6.3) {$4{\times}4{\times}128$};
\node[font=\sffamily\scriptsize, mcTemporalBdr, anchor=east, text width=1.6cm, inner sep=0pt, align=left]
  at (9.25,-6.4) {\textit{Adaptive}\\\textit{AvgPool}};

\node[mcstage] at (\colC, 0.6) {\textbf{Text Encoder}};

\node[io] (tx_in) at (\colC, -0.4) {RoBERTa [CLS]\\Input $\mathbb{R}^{768}$};

\node[layer] (tx_fc1) at (\colC, -2.0) {FC + GELU\\$768 \to 384$};
\node[act] (tx_dr) at (\colC, -4.8) {Dropout(0.1)};
\node[layer] (tx_fc2) at (\colC, -7.6) {FC $384 \to 64$};

\node[io] (tx_out) at (\colC, -10.9) {LayerNorm\\Output $\mathbb{R}^{64}$};

\draw[mcarrow] (tx_in) -- (tx_fc1);
\draw[mcarrow] (tx_fc1) -- (tx_dr);
\draw[mcarrow] (tx_dr) -- (tx_fc2);
\draw[mcarrow] (tx_fc2) -- (tx_out);

\node[mcstage] at (\colD, 0.6) {\textbf{Attention Fusion}};

\node[io] (fu_in) at (\colD, -0.4) {Project $64\to256$\\Stack $4{\times}256$\\(\textsc{[fusion]} + 3)};

\node[act] (fu_ln1) at (\colD, -1.9) {LayerNorm};
\node[attn] (fu_mha) at (\colD, -3.4) {MHA\\4 heads, $d_k{=}64$};
\node[act] (fu_res1) at (\colD, -4.9) {Drop(0.2) + Residual};
\node[act] (fu_ln2) at (\colD, -6.4) {LayerNorm};
\node[layer] (fu_ffn) at (\colD, -7.9) {FFN (GELU, dropout)\\$256 \to 1024 \to 256$};
\node[act] (fu_res2) at (\colD, -9.4) {Drop(0.2) + Residual};

\node[io] (fu_fused) at (\colD, -10.9) {Transformer output\\$4{\times}256$};

\draw[mcarrow] (fu_in) -- (fu_ln1);
\draw[mcarrow] (fu_ln1) -- (fu_mha);
\draw[mcarrow] (fu_mha) -- (fu_res1);
\draw[mcarrow] (fu_res1) -- (fu_ln2);
\draw[mcarrow] (fu_ln2) -- (fu_ffn);
\draw[mcarrow] (fu_ffn) -- (fu_res2);
\draw[mcarrow] (fu_res2) -- (fu_fused);

\draw[-{Stealth[length=4pt,width=3pt]}, dashed, thick, mcEncoderBdr, rounded corners=4pt]
  (fu_in.east) -- ++(0.7,0) |- (fu_res1.east);
\draw[-{Stealth[length=4pt,width=3pt]}, dashed, thick, mcEncoderBdr, rounded corners=4pt]
  (fu_res1.east) -- ++(0.7,0) |- (fu_res2.east);

\node[mcstage] at (7.7, -12.0) {\textbf{Prediction Head}};

\node[pool, text width=3.2cm] (ph_pool) at (2.5, -13.0) {Extract \textsc{[fusion]} + LN\\$4{\times}256 \to 256$};
\node[layer, text width=2.8cm] (ph_fc) at (7.7, -13.0) {Dropout(0.1) + FC\\$256 \to 2H$};
\node[io, text width=2.8cm] (ph_out) at (12.9, -13.0) {$\Delta\widehat V_{\max}(t+h)$\\$\Delta\widehat{\mathrm{MSLP}}(t+h)$};

\draw[mcarrow] (ph_pool) -- (ph_fc);
\draw[mcarrow] (ph_fc) -- (ph_out);

\begin{scope}[on background layer]
  \fill[black!5, rounded corners=6pt]
    (\colA-1.8, 1.0) rectangle (\colA+1.8, -11.5);
  \fill[black!5, rounded corners=6pt]
    (\colB-2.6, 1.0) rectangle (\colB+2.6, -11.5);
  \fill[black!5, rounded corners=6pt]
    (\colC-1.8, 1.0) rectangle (\colC+1.8, -11.5);
  \fill[black!5, rounded corners=6pt]
    (\colD-1.8, 1.0) rectangle (\colD+1.8, -11.5);
  \fill[mcHead!10, rounded corners=6pt]
    (\colA-1.8, -11.7) rectangle (\colD+1.8, -13.8);
\end{scope}

\end{tikzpicture}}
\caption{Encoder branches and the attention fusion module. The frozen RoBERTa backbone supplies a 768-dimensional [CLS] vector. A separate trainable MLP projects it to 64 dimensions. The tabular MLP and image CNN also produce 64-dimensional embeddings. In the attention variant, a projection maps each normalized branch embedding to 256 dimensions. The three tokens are stacked with a learnable \textsc{[fusion]} token and modality embeddings. One pre-LN Transformer layer uses four attention heads, residual paths, and a $256\to1024\to256$ feed-forward network. The normalized \textsc{[fusion]} token supplies the $256\to2H$ prediction head. The selected primary full model uses concatenation.}
\label{fig:nn_detail}
\end{figure}
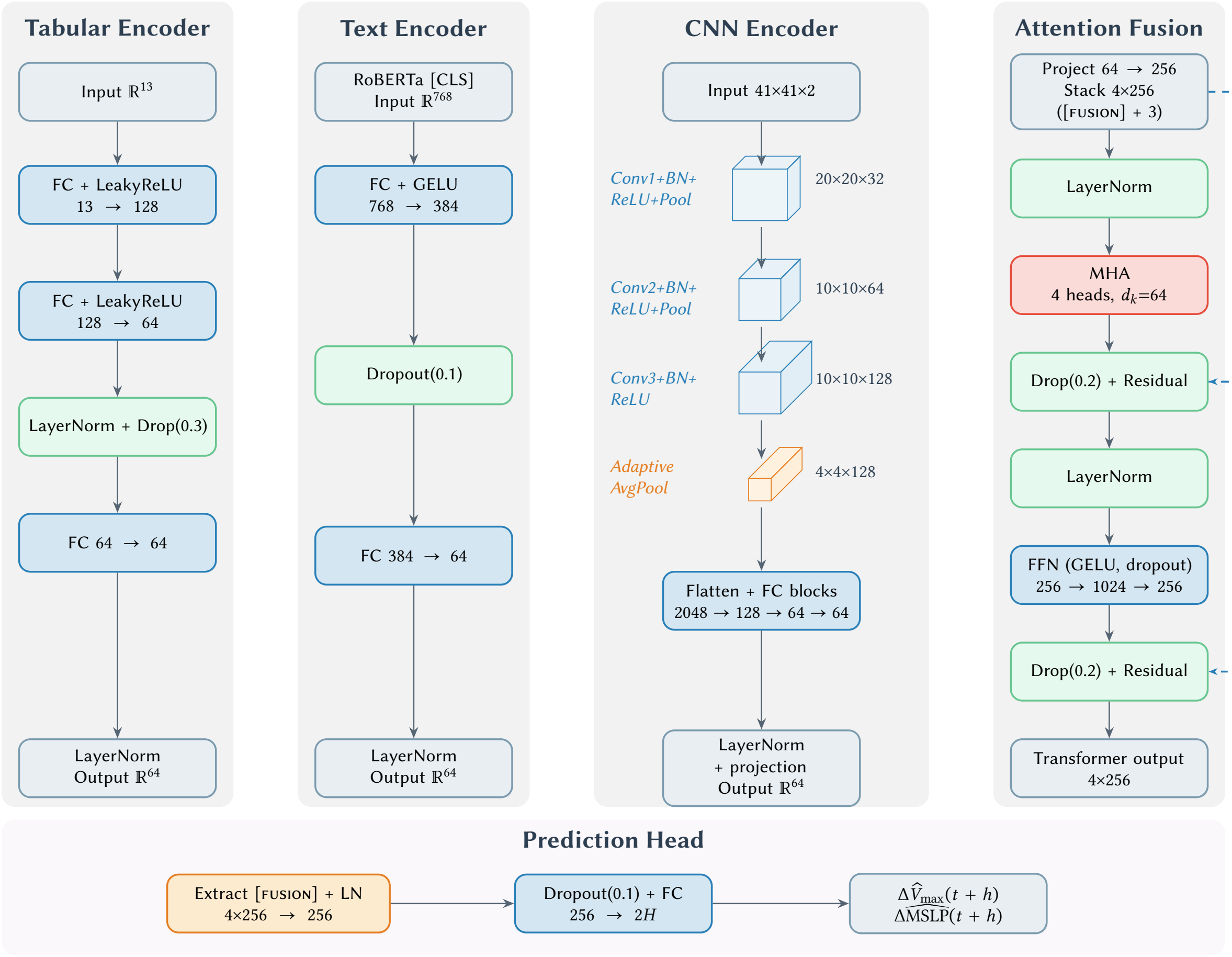

\autoref{fig:nn_detail} provides a detailed view of each encoder and the attention-based fusion module used in the end-to-end MultiCyclone architecture. All encoders project their respective inputs to a shared 64-dimensional embedding space before fusion.

\subsection{Training Configuration}\label{app:training_configuration}
For end-to-end neural models, validation loss is computed for each batch using Eq.~\ref{eq:loss}. The learning-rate scheduler and early stopping use the average weighted by batch size
\[
\mathcal{L}_{\mathrm{val}} =
\frac{\sum_B |B|\mathcal{L}_B}{\sum_B |B|},
\]
where $|B|$ is the number of cases in validation batch $B$. When the number of available targets per case varies, this average can differ from the MSE obtained by pooling all available targets across validation batches. A batch with no available targets has loss zero. Reported evaluation metrics are computed separately for each target and forecast lead over all available cases.

The end-to-end models use AdamW with $(\beta_1,\beta_2)=(0.9,0.999)$. In the primary 24\,h grid, the base learning-rate/batch-size pairs for T, I, D, T+I, T+D, I+D and T+I+D are, respectively, $(5\times10^{-4},32)$, $(5\times10^{-4},32)$, $(2\times10^{-5},16)$, $(5\times10^{-4},32)$, $(10^{-3},32)$, $(5\times10^{-4},32)$ and $(5\times10^{-4},16)$. Weight decay is zero for T, I and T+I and $10^{-2}$ for D and text fusion. The patience for early stopping is 7. The epoch cap is 200 for T, I and T+I and 500 for D and text fusion.

These are the retained, modality-specific settings. Validation loss chooses a stopping epoch and controls the learning-rate scheduler. Generic-text validation comparisons informed some fusion choices, but we did not perform an exhaustive search over architecture, fusion operator, learning rate, or batch size. These comparisons do not establish the best fusion operator for the forecast-tuned text backbone.

The independently trained per-lead models use an epoch cap of 500, patience 5 and weight decay $10^{-2}$. Their base learning-rate/batch-size pairs are $(5\times10^{-4},32)$ for T, $(10^{-3},16)$ for I, $(2\times10^{-5},32)$ for D, $(10^{-3},32)$ for T+D and $(5\times10^{-4},16)$ for T+I+D. T+I and I+D use $5\times10^{-4}$ with batch size 32 at 24\,h and 16 at longer leads. Joint models retain the same epoch cap, patience and weight decay. Their T, I and D rates and batch sizes match the per-lead models. Joint T+I, T+D, I+D and T+I+D use $(10^{-4},32)$, $(10^{-4},16)$, $(5\times10^{-4},32)$ and $(10^{-3},16)$, respectively. In fused models, the image encoder's learning rate is one tenth of the base rate. Standalone I uses the full rate.

These neural runs reduce the learning rate by a factor of $0.5$ on a validation plateau with scheduler patience $3$ and clip gradient norms at 1.0. Temporal CNN--GRU models use AdamW with learning rate $5\times10^{-4}$, batch size 64, weight decay $10^{-3}$, an epoch cap of 100 and patience 10 for early stopping, with the same plateau factor, scheduler patience and clipping norm. Their selection criterion is MSE over complete validation target pairs.

For hybrid and temporal XGBoost heads we fit one regressor per target, using at most 1,000 trees, depth 4, learning rate 0.05, row and column subsampling 0.8, L1/L2 penalties 0.1/1.0, histogram tree construction and 50-round validation early stopping. The hybrid Ridge controls use penalty $\alpha=1$. Ridge regression and the reference baselines are deterministic, so repeated runs give identical predictions rather than independent fits.

\subsection{Initialization}\label{app:initialization}

For the 24\,h multi-modal ablations that include text, linear weights in the constructed model are reset with Kaiming normal initialization and their biases to zero. LayerNorm scales and biases are reset to one and zero. Other module classes, including convolutional and BatchNorm layers, retain their initialization at construction. The external sentence encoder trained in the D model is frozen during fusion training and is not reset. Standalone D starts fine-tuning from its pretrained initialization. The per-lead and joint configurations retain their separately specified initialization settings.

\subsection{Computational Resources}
The neural experiments were run on the University of Lausanne DCSR \textit{Curnagl} high-performance computing cluster as single-GPU NVIDIA A100~40\,GB jobs. Exploratory runs on other GPU architectures are separate from the main comparison. Models are implemented in PyTorch~2.x. The end-to-end neural protocol uses fp32, disables TF32, and requests deterministic operations, including a deterministic replacement for the adaptive-pooling backward pass. Some deterministic checks are configured to warn rather than fail, so these settings do not by themselves establish universal bitwise reproducibility. The reported temporal models use a controlled runtime with deterministic algorithms enforced, cuDNN benchmarking disabled and TF32 disabled. The software settings of the earlier temporal runs were not fully recorded, so exact reproduction of those runs is not established. The supplementary material provides predictions and reproducibility information. Fused text models require both the fusion checkpoint and its external frozen RoBERTa backbone, using generic pretrained weights for generic text, or the corresponding trained D backbone for forecast-tuned text. Transformer attributions are computed on the GPU.

A temporal alignment check compared this timestamp-based construction with an earlier implementation that used preceding cases and could pair an image with an older discussion when synoptic times were missing. Both versions were trained under the same runtime settings, including models without text as controls. The reported temporal results use the timestamp-based construction.

Reproducibility checks distinguish rerunning inference from retraining. Before the discussion issue times were corrected, CPU inference was repeated for \resultvalue{reproduction/cpu/replayed} of \resultvalue{reproduction/cpu/entries} neural and fusion-control entries. At a maximum prediction difference of $5\times10^{-4}$ in physical units, \resultvalue{reproduction/cpu/strict_pass} passed and \resultvalue{reproduction/cpu/strict_fail} failed. For \resultvalue{reproduction/cpu/rounded_metric_changes} replayed runs, at least one metric changed when rounded to two decimals. Comparable averages across complete groups of five seeds retained their rounded values. This agreement does not resolve the failed prediction checks. The CPU runs used batches of 32, including a run originally exported with batches of 16, so this was not solely a hardware comparison. Fresh A100 fits of the primary T, I and T+I models for seed 1 reproduced the accepted checkpoint bytes. After the correction, every model that uses discussion text was refitted with five seeds for each stochastic configuration, including the primary T+D comparison, the fusion and backbone controls, the hybrid and temporal regressors, the text probes and the CLEAR calibration with five forest seeds. All reported scores were recomputed from the exported predictions and the raw targets. Tabular, image and tabular with image models do not use the discussions, so they were retained, as were the five pretrained image encoders of the hybrid models. The replay results above therefore describe the retained models and the earlier text models. The refitted text models are not an exact replay of the earlier fits. For the earlier fusion fits, all eleven seed-1 checkpoints and training result files matched their historical bytes, but predictions exported through the training framework differed from the earlier predictions by more than the stated tolerance. Selected controls using direct export recovered the historical predictions exactly, demonstrating effects of the export policy for those checkpoints. These controls do not explain every CPU discrepancy or establish agreement across all batch sizes. The export settings are recorded for each configuration, including the failed comparisons with earlier predictions. In a later check on the same A100 hardware, the primary 24\,h T, D and T+D models were refitted for all five seeds from the preserved source files, inputs and runtime environment. The T and D checkpoints of this check equal those behind the reported results byte for byte, and its D and T+D predictions are identical to those reported. Exporting six retained D and T+D checkpoints again also returned identical predictions. This check reused the preserved runtime environment, and the executed source files are kept with the code. Some historical checkpoints and software provenance remain unavailable, so full reproduction of every model from a fresh installation is not established.

\FloatBarrier
\Needspace{0.88\textheight}
\section{Per-Lead vs.\ Joint Multi-Lead Prediction}
\label{sec:per_horizon_vs_joint}

\begingroup
\setlength{\intextsep}{2pt}
\begin{table}[!htbp]
\centering
\caption{Per-lead versus joint neural forecasts for $V_{\max}$ change. Both experiment families use five seeds. Entries show individual-seed mean $\pm$ sample SD. Text-fusion rows use forecast-tuned encoders. This descriptive comparison does not isolate training strategy from model configuration. MAE is in kt. D uses matched discussions. Counts are in Table~\ref{tab:neural_populations}. Bold and underline mark the best and second-best displayed means per metric column within each lead (ties included, with D excluded because its test population differs).}
\label{tab:per_horizon_vs_joint}
\small
\renewcommand{\arraystretch}{1.00}
\setlength{\tabcolsep}{4pt}
\begin{adjustbox}{max width=\linewidth}
\begin{tabular}{@{}lccccc@{}}
\toprule
\multirow{2}{*}{Lead} & \multirow{2}{*}{Modality} & \multicolumn{2}{c}{$R^2$} & \multicolumn{2}{c}{MAE (kt)} \\
\cmidrule(lr){3-4} \cmidrule(lr){5-6}
 &  & Per-lead & Joint & Per-lead & Joint \\
\midrule
\multirow{7}{*}{24\,h} & T & $0.44\pm0.01$ & $0.41\pm0.01$ & $10.11\pm0.06$ & $10.32\pm0.10$ \\
 & I & $0.21\pm0.03$ & $0.09\pm0.23$ & $12.43\pm0.26$ & $12.60\pm0.66$ \\
 & D & $0.53\pm0.01$ & $0.51\pm0.02$ & $9.49\pm0.15$ & $9.78\pm0.25$ \\
 & T+I & $0.46\pm0.02$ & $0.41\pm0.01$ & $10.09\pm0.11$ & $10.44\pm0.09$ \\
 & T+D & $\underline{0.54}\pm0.04$ & $\boldsymbol{0.54}\pm0.02$ & $\underline{9.26}\pm0.29$ & $\underline{9.40}\pm0.30$ \\
 & I+D & $0.47\pm0.03$ & $\underline{0.52}\pm0.01$ & $10.06\pm0.39$ & $9.66\pm0.12$ \\
 & T+I+D & $\boldsymbol{0.55}\pm0.02$ & $\boldsymbol{0.54}\pm0.03$ & $\boldsymbol{9.15}\pm0.19$ & $\boldsymbol{9.24}\pm0.22$ \\
\midrule
\multirow{7}{*}{48\,h} & T & $0.47\pm0.02$ & $0.47\pm0.01$ & $15.32\pm0.38$ & $15.27\pm0.30$ \\
 & I & $0.16\pm0.06$ & $0.12\pm0.29$ & $18.97\pm0.37$ & $18.94\pm1.14$ \\
 & D & $0.54\pm0.01$ & $0.53\pm0.02$ & $14.32\pm0.17$ & $14.35\pm0.27$ \\
 & T+I & $0.45\pm0.02$ & $0.46\pm0.01$ & $15.49\pm0.54$ & $15.50\pm0.30$ \\
 & T+D & $\boldsymbol{0.58}\pm0.02$ & $\underline{0.55}\pm0.02$ & $\boldsymbol{13.44}\pm0.23$ & $\underline{13.84}\pm0.28$ \\
 & I+D & $0.52\pm0.03$ & $0.50\pm0.02$ & $14.38\pm0.54$ & $14.81\pm0.35$ \\
 & T+I+D & $\underline{0.57}\pm0.01$ & $\boldsymbol{0.56}\pm0.03$ & $\underline{13.65}\pm0.22$ & $\boldsymbol{13.66}\pm0.48$ \\
\midrule
\multirow{7}{*}{72\,h} & T & $0.39\pm0.02$ & $\underline{0.42}\pm0.01$ & $18.77\pm0.58$ & $18.05\pm0.20$ \\
 & I & $0.24\pm0.06$ & $0.08\pm0.38$ & $20.43\pm0.75$ & $20.91\pm1.28$ \\
 & D & $0.42\pm0.06$ & $0.43\pm0.02$ & $17.63\pm0.80$ & $17.19\pm0.13$ \\
 & T+I & $0.41\pm0.03$ & $\underline{0.42}\pm0.02$ & $18.34\pm0.44$ & $17.99\pm0.48$ \\
 & T+D & $\boldsymbol{0.46}\pm0.06$ & $\boldsymbol{0.44}\pm0.03$ & $\boldsymbol{16.93}\pm0.67$ & $\boldsymbol{16.74}\pm0.41$ \\
 & I+D & $0.39\pm0.10$ & $\underline{0.42}\pm0.03$ & $18.08\pm1.48$ & $17.55\pm0.18$ \\
 & T+I+D & $\underline{0.45}\pm0.04$ & $0.41\pm0.05$ & $\underline{17.11}\pm0.61$ & $\underline{16.88}\pm0.76$ \\
\midrule
\multirow{7}{*}{96\,h} & T & $\underline{0.27}\pm0.03$ & $\boldsymbol{0.30}\pm0.03$ & $19.81\pm0.42$ & $\underline{19.32}\pm0.37$ \\
 & I & $0.18\pm0.07$ & $-0.06\pm0.65$ & $21.00\pm0.94$ & $20.95\pm1.90$ \\
 & D & $0.19\pm0.11$ & $0.24\pm0.03$ & $20.80\pm1.60$ & $19.39\pm0.45$ \\
 & T+I & $\boldsymbol{0.28}\pm0.08$ & $\boldsymbol{0.30}\pm0.04$ & $\boldsymbol{19.55}\pm0.95$ & $\boldsymbol{19.19}\pm0.65$ \\
 & T+D & $0.24\pm0.07$ & $0.20\pm0.06$ & $\underline{19.76}\pm1.22$ & $19.55\pm0.47$ \\
 & I+D & $0.08\pm0.19$ & $\underline{0.22}\pm0.06$ & $21.62\pm2.66$ & $19.62\pm0.65$ \\
 & T+I+D & $0.19\pm0.08$ & $0.19\pm0.10$ & $20.43\pm0.81$ & $19.58\pm1.34$ \\
\midrule
\multirow{7}{*}{120\,h} & T & $\boldsymbol{0.21}\pm0.03$ & $\underline{0.19}\pm0.04$ & $\boldsymbol{20.17}\pm0.09$ & $\underline{20.23}\pm0.51$ \\
 & I & $\underline{0.15}\pm0.04$ & $-0.46\pm1.44$ & $\underline{20.76}\pm0.62$ & $21.31\pm2.87$ \\
 & D & $0.00\pm0.15$ & $0.10\pm0.05$ & $23.42\pm2.03$ & $20.67\pm0.43$ \\
 & T+I & $0.08\pm0.09$ & $\boldsymbol{0.20}\pm0.04$ & $21.28\pm0.87$ & $\boldsymbol{20.22}\pm0.53$ \\
 & T+D & $0.06\pm0.07$ & $0.06\pm0.10$ & $22.51\pm1.49$ & $21.29\pm0.66$ \\
 & I+D & $0.05\pm0.18$ & $0.09\pm0.11$ & $23.13\pm2.57$ & $20.84\pm1.06$ \\
 & T+I+D & $0.00\pm0.13$ & $0.09\pm0.13$ & $23.65\pm2.26$ & $20.61\pm1.17$ \\
\bottomrule
\end{tabular}
\end{adjustbox}
\end{table}

\endgroup
\FloatBarrier

The models in Tables~\ref{tab:full_results}--\ref{tab:mslp_full} are trained separately at each forecast lead, whereas the joint models predict all five leads together. \autoref{tab:per_horizon_vs_joint} compares these approaches across five seeds per configuration, using forecast-tuned text encoders when discussions are combined with other inputs. Joint training with tabular inputs and discussions gives lower mean wind MAE at the three longest leads but not at 24 and 48\,h, with $\resultvalue{cell/per_horizon_vs_joint/tab_disc_mhj_120/VMAX/mae/mean}$ versus $\resultvalue{cell/per_horizon_vs_joint/tab_disc_fph_120/VMAX/mae/mean}$\,kt at 120\,h. This advantage is not consistent across all modality combinations. Sharing an encoder may help when fewer training examples are available at longer leads, although differences in learning rate and batch size mean that this comparison cannot isolate the effect of joint training. Missing targets are masked, allowing a case to contribute at shorter leads even when its targets at longer leads are unavailable. For the joint model using wind fields alone, the poor mean performance at longer leads is dominated by one unstable fit that produced several extreme predictions for a single storm. All five seeds are retained, and this instability cautions against attributing that average solely to joint training.

\FloatBarrier
\section{Why \texorpdfstring{$R^2$}{R2} Can Increase at Longer Lead Times}
\label{sec:leadtime_diagnostics}

Comparisons across forecast leads change both the forecast target and the available cases. \autoref{tab:leadtime_diagnostics} therefore reports train/test counts, target spread, and the same joint T+D configuration at every lead time. \autoref{tab:physical_baselines} reports the reference baseline comparison for both targets. The 24\,h scores in these tables come from the same model family as the longer-lead scores.

\begin{table}[!htbp]
\centering
\caption{Lead-time counts and physical error diagnostics. Test-target population variances are in kt$^2$ ($\Delta V_{\max}$) and hPa$^2$ ($\Delta$MSLP). The joint T+D entries are individual-seed mean $\pm$ sample SD over seeds 1--5 on identical valid targets. Wind MSE is in kt$^2$ and is averaged across the MSEs from the five seeds. All forecast leads use the joint model family. Reference baselines appear separately in Table~\ref{tab:physical_baselines}.}
\label{tab:leadtime_diagnostics}
\small
\renewcommand{\arraystretch}{1.12}
\setlength{\tabcolsep}{4pt}
\begin{adjustbox}{max width=\linewidth}
\begin{tabular}{@{}lcccccc@{}}
\toprule
\multirow{2}{*}{Lead} & \multicolumn{2}{c}{Valid $N/S$} & \multicolumn{2}{c}{Target variance $\mathrm{Var}_0$} & \multicolumn{2}{c}{Joint T+D} \\
\cmidrule(lr){2-3} \cmidrule(lr){4-5} \cmidrule(lr){6-7}
 & Train & Test & $\Delta V$ (kt$^2$) & $\Delta P$ (hPa$^2$) & MSE (kt$^2$) & $R^2$ \\
\midrule
24\,h & 8,954/491 & 1,150/66 & $362.24$ & $169.93$ & $167.29\pm8.43$ & $0.54\pm0.02$ \\
48\,h & 7,016/431 & 898/53 & $811.35$ & $388.25$ & $362.97\pm12.51$ & $0.55\pm0.02$ \\
72\,h & 5,406/365 & 686/51 & $925.55$ & $456.85$ & $518.43\pm25.92$ & $0.44\pm0.03$ \\
96\,h & 4,114/299 & 499/38 & $859.42$ & $446.34$ & $683.29\pm53.18$ & $0.20\pm0.06$ \\
120\,h & 3,082/230 & 360/28 & $859.05$ & $477.61$ & $811.70\pm83.60$ & $0.06\pm0.10$ \\
\bottomrule
\end{tabular}
\end{adjustbox}
\par\smallskip
{\footnotesize $R^2=1-\mathrm{MSE}/\mathrm{Var}_0(y)$ uses the displayed population variance, with denominator $N$. Each target variance uses all valid test cases at that lead. Counts shrink with lead time.}
\end{table}

\begin{table}[!htbp]
\centering
\caption{Reference baselines on all valid test cases. MAE and RMSE units are kt and hPa. Bold and underline mark the best and second-best displayed means per metric column within each lead (ties included).}
\label{tab:physical_baselines}
\small
\renewcommand{\arraystretch}{1.12}
\setlength{\tabcolsep}{4pt}
\begin{adjustbox}{max width=\linewidth}
\begin{tabular}{@{}lcccccccc@{}}
\toprule
\multirow{2}{*}{Lead (h)} & \multirow{2}{*}{Baseline} & \multirow{2}{*}{$N/S$} & \multicolumn{2}{c}{MAE (kt/hPa)} & \multicolumn{2}{c}{RMSE (kt/hPa)} & \multicolumn{2}{c}{$R^2$} \\
\cmidrule(lr){4-5} \cmidrule(lr){6-7} \cmidrule(lr){8-9}
 &  &  & $V_{\max}$ & MSLP & $V_{\max}$ & MSLP & $V_{\max}$ & MSLP \\
\midrule
\multirow{3}{*}{24} & Persistence & 1{,}150/66 & $\underline{13.78}$ & $\underline{8.86}$ & $\underline{19.33}$ & $\underline{13.29}$ & $\underline{-0.03}$ & $\underline{-0.04}$ \\
 & Climatology & 1{,}150/66 & $\boldsymbol{13.76}$ & $\boldsymbol{8.73}$ & $\boldsymbol{19.15}$ & $\boldsymbol{13.11}$ & $\boldsymbol{-0.01}$ & $\boldsymbol{-0.01}$ \\
 & Trend persistence & 1{,}150/66 & $16.12$ & $10.32$ & $23.63$ & $16.51$ & $-0.54$ & $-0.60$ \\
\midrule
\multirow{3}{*}{48} & Persistence & 898/53 & $\underline{21.49}$ & $\underline{14.46}$ & $\underline{29.31}$ & $\underline{20.49}$ & $\underline{-0.06}$ & $\underline{-0.08}$ \\
 & Climatology & 898/53 & $\boldsymbol{21.16}$ & $\boldsymbol{14.05}$ & $\boldsymbol{28.76}$ & $\boldsymbol{19.88}$ & $\boldsymbol{-0.02}$ & $\boldsymbol{-0.02}$ \\
 & Trend persistence & 898/53 & $28.30$ & $18.80$ & $40.36$ & $28.20$ & $-1.01$ & $-1.05$ \\
\midrule
\multirow{3}{*}{72} & Persistence & 686/51 & $\underline{23.79}$ & $\underline{16.62}$ & $\underline{31.82}$ & $\underline{22.82}$ & $\underline{-0.09}$ & $\underline{-0.14}$ \\
 & Climatology & 686/51 & $\boldsymbol{23.18}$ & $\boldsymbol{15.89}$ & $\boldsymbol{30.93}$ & $\boldsymbol{21.72}$ & $\boldsymbol{-0.03}$ & $\boldsymbol{-0.03}$ \\
 & Trend persistence & 686/51 & $29.01$ & $19.86$ & $39.57$ & $27.92$ & $-0.69$ & $-0.71$ \\
\midrule
\multirow{3}{*}{96} & Persistence & 499/38 & $\underline{24.33}$ & $\underline{17.43}$ & $\underline{31.79}$ & $\underline{23.78}$ & $\underline{-0.18}$ & $\underline{-0.27}$ \\
 & Climatology & 499/38 & $\boldsymbol{23.46}$ & $\boldsymbol{16.15}$ & $\boldsymbol{30.62}$ & $\boldsymbol{22.03}$ & $\boldsymbol{-0.09}$ & $\boldsymbol{-0.09}$ \\
 & Trend persistence & 499/38 & $27.22$ & $19.41$ & $35.09$ & $26.03$ & $-0.43$ & $-0.52$ \\
\midrule
\multirow{3}{*}{120} & Persistence & 360/28 & $\underline{25.68}$ & $\underline{18.94}$ & $\underline{33.72}$ & $\underline{26.39}$ & $\underline{-0.32}$ & $\underline{-0.46}$ \\
 & Climatology & 360/28 & $\boldsymbol{24.34}$ & $\boldsymbol{17.18}$ & $\boldsymbol{32.03}$ & $\boldsymbol{23.75}$ & $\boldsymbol{-0.19}$ & $\boldsymbol{-0.18}$ \\
 & Trend persistence & 360/28 & $27.67$ & $20.38$ & $36.57$ & $28.31$ & $-0.56$ & $-0.68$ \\
\bottomrule
\end{tabular}
\end{adjustbox}
\end{table}

For the joint configuration using tabular inputs and discussions, the wind MSE averaged over seeds increases with lead, while the averaged $R^2$ rises slightly between 24 and 48\,h. This is possible because target variance also increases. The identity $R^2=1-\mathrm{MSE}/\mathrm{Var}_0(\Delta V)$ shows why an improving relative score can accompany increasing absolute error. At later leads, target spread changes less while error continues to grow, and $R^2$ declines. \autoref{tab:leadtime_diagnostics} gives the values and the shrinking evaluation population. These lead-specific populations do not provide a controlled test of a physical predictability limit.

Training availability also declines with lead. Discussion-only models use a further matched-discussion subset. Reduced training data is a plausible contributor to long-lead instability, but these runs do not separate that factor from target difficulty, sample composition, architecture or optimization. The joint objective shares an encoder across lead times, but replication across five seeds does not remove differences in configuration and evaluation population. In both experiment families, adding discussion to tabular inputs lowers mean wind MAE at the shorter leads, but this gain does not persist uniformly at the longest leads. The full results are in Tables~\ref{tab:full_results}--\ref{tab:mslp_full} and \ref{tab:per_horizon_vs_joint}.

\FloatBarrier
\section{Comparison with Archived Forecast Guidance}
\label{sec:operational_baseline}

\autoref{tab:operational_baseline} compares 24\,h forecasts from the NHC ATCF archive with the same future wind recorded in our SHIPS-derived dataset used for our models. Forecasts are matched exactly by cyclone, initialization time and lead. Each aid and the neural ensembles are scored on identical rows within that comparison. The sample differs between aids, so the table does not rank all aids on one common storm population. The gap column uses the T+D ensemble, which has the lowest 24\,h wind and pressure MAE among the ensembles scored on the full test set (\autoref{tab:nn_ensemble}). On each aid's matched rows, it also has a lower wind MAE than the T ensemble, and its wind MAE is nearly identical to that of the T+I+D ensemble.

The NHC official forecast (OFCL) has MAE $\resultvalue{cell/operational_baseline/OFCL/aid_mae/mean}$\,kt, compared with $\resultvalue{cell/operational_baseline/OFCL/TD_mae/mean}$\,kt for the ensemble using tabular inputs with discussions, on $\resultvalue{operational/OFCL/n_obs}$ cases from $\resultvalue{operational/OFCL/n_storms}$ storms. The ensemble's MAE is $\resultvalue{cell/operational_baseline/OFCL/TD_minus_aid_mae/mean}$\,kt higher than OFCL\textquotesingle s, with a 95\% confidence interval from resampling whole storms of $[\resultvalue{cell/operational_baseline/OFCL/storm_ci_low/mean},\resultvalue{cell/operational_baseline/OFCL/storm_ci_high/mean}]$. Differences against HWRF and early interpolated HWRF (HWFI) remain unresolved. An interval spanning zero does not establish forecast equivalence.

\href{https://www.nhc.noaa.gov/modelsummary.shtml}{NHC distinguishes early and late guidance} by availability during the forecast cycle. Interpolated aids such as HWFI, HFAI and HFBI adjust an earlier model run to the current initialization. They are separate products from the late HWRF, HFSA and HFSB runs. HF3A is an experimental HAFS parallel from 2022. The 2023 HAFS archive also contains forecasts before implementation, so its full-season rows must not be described as wholly operational.

OFCL is indexed by nominal synoptic initialization time, while routine NHC advisories are issued three hours later. Its 24\,h lead is measured from nominal initialization, following the \href{https://www.nhc.noaa.gov/verification/verify2.shtml}{NHC verification convention}.

\begin{table}[!htbp]
\centering
\caption{24\,h $V_{\max}$ MAE for absolute intensity (kt), with each aid and neural ensemble of five models scored on their identical common rows. A positive T+D-minus-aid gap favors the aid. CIs resample whole storms (10,000 draws, pointwise percentile 95\%).}
\label{tab:operational_baseline}
\small
\renewcommand{\arraystretch}{1.12}
\setlength{\tabcolsep}{4pt}
\begin{adjustbox}{max width=\linewidth}
\begin{tabular}{@{}lccccccc@{}}
\toprule
Forecast aid & $N/S$ & Aid MAE & T & T+D & T+I+D & T+D minus aid & 95\% storm CI \\
\midrule
OFCL & 1{,}078/64 & $7.69$ & $10.07$ & $8.97$ & $8.99$ & $1.28$ & $[0.71, 1.82]$ \\
OFCI & 1{,}018/64 & $8.43$ & $10.16$ & $9.09$ & $9.12$ & $0.66$ & $[0.17, 1.11]$ \\
HWFI & 1{,}066/64 & $8.85$ & $9.98$ & $8.90$ & $8.92$ & $0.06$ & $[-0.63, 0.72]$ \\
HWRF & 1{,}104/64 & $9.67$ & $9.98$ & $8.97$ & $9.00$ & $-0.71$ & $[-1.55, 0.20]$ \\
HFAI & 641/35 & $9.59$ & $9.75$ & $9.01$ & $9.00$ & $-0.58$ & $[-1.77, 0.58]$ \\
HFSA & 646/35 & $10.69$ & $9.73$ & $9.01$ & $9.00$ & $-1.68$ & $[-2.66, -0.72]$ \\
HFBI & 641/35 & $9.53$ & $9.75$ & $9.01$ & $9.00$ & $-0.52$ & $[-1.40, 0.37]$ \\
HFSB & 646/35 & $11.00$ & $9.73$ & $9.01$ & $9.00$ & $-1.99$ & $[-2.69, -1.22]$ \\
HF3A & 158/13 & $8.85$ & $11.77$ & $8.28$ & $8.26$ & $-0.57$ & $[-3.35, 2.35]$ \\
DSHP & 1{,}122/66 & $9.23$ & $9.91$ & $8.90$ & $8.93$ & $-0.33$ & $[-0.86, 0.17]$ \\
LGEM & 1{,}122/66 & $9.24$ & $9.91$ & $8.90$ & $8.93$ & $-0.34$ & $[-0.95, 0.22]$ \\
\bottomrule
\end{tabular}
\end{adjustbox}
\par\smallskip
{\footnotesize An aid is an archived official forecast or model-guidance product; Aid MAE is its mean absolute wind error. $N/S$ gives forecast initializations/storms. Matched by cyclone, initialization, and lead. The verifying value is the dataset future wind at the verifying time. I-suffixed aids are early interpolated guidance; HFSA/HFSB and HFAI/HFBI cover 2023, HF3A is the 2022 experimental parallel. Reanalysis and discussion timing prevent a claim of operational information equivalence. Unresolved intervals do not establish equivalence.}
\end{table}

\paragraph{Sensitivity to the verifying best track and implementation date.}
The b-deck check in \autoref{tab:bdeck_sensitivity} replaces only the future verifying wind. Neural absolute predictions retain the initial intensity of the original dataset plus the fitted change. Exact joins recover $\resultvalue{bdeck/diagnostics/available_future_unique_initializations}$ of $\resultvalue{bdeck/diagnostics/unique_initializations}$ distinct initializations. Future wind differs from the original targets at $\resultvalue{bdeck/diagnostics/changed_future_unique_initializations}$ of them. The Darby initialization without a future b-deck wind is excluded. OFCL remains ahead, while the HWRF and HWFI differences remain unresolved. The Dora archive has an inconsistent internal cyclone identifier. The filename identity is used only for the pinned archive whose $\resultvalue{bdeck/diagnostics/dora_verified_distinct_timestamps}$ timestamp/wind pairs match the \href{https://www.nhc.noaa.gov/data/tcr/EP052023_Dora.pdf}{NHC Tropical Cyclone Report}.

\begin{table}[!htbp]
\centering
\caption{ATCF b-deck verification sensitivity at 24\,h. Each aid and the fixed T+D ensemble of five models use identical rows. Upper block retains every primary-comparison row with the exact b-deck verifying wind. Lower block additionally requires tropical/subtropical status at both the initial and verifying times. Values are absolute wind MAE in kt. Positive gaps favor the aid. $N/S$ gives forecast initializations/storms. Neural forecasts add the predicted change to the initial intensity in the original dataset.}
\label{tab:bdeck_sensitivity}
\small
\renewcommand{\arraystretch}{1.12}
\setlength{\tabcolsep}{4pt}
\begin{adjustbox}{max width=\linewidth}
\begin{tabular}{@{}lccccc@{}}
\toprule
Forecast aid & $N/S$ & Aid MAE & T+D MAE & T+D minus aid & 95\% storm CI \\
\midrule
\multicolumn{6}{@{}l@{}}{\textit{Exact b-deck matches, all statuses}} \\
OFCL & 1,077/64 & $7.68$ & $9.02$ & $1.34$ & $[0.78, 1.88]$ \\
OFCI & 1,017/64 & $8.45$ & $9.15$ & $0.70$ & $[0.21, 1.16]$ \\
HWFI & 1,065/64 & $8.92$ & $8.97$ & $0.05$ & $[-0.64, 0.71]$ \\
HWRF & 1,103/64 & $9.75$ & $9.02$ & $-0.73$ & $[-1.57, 0.19]$ \\
HFAI & 641/35 & $9.72$ & $9.10$ & $-0.61$ & $[-1.79, 0.53]$ \\
HFSA & 646/35 & $10.83$ & $9.10$ & $-1.73$ & $[-2.68, -0.78]$ \\
HFBI & 641/35 & $9.63$ & $9.10$ & $-0.53$ & $[-1.41, 0.36]$ \\
HFSB & 646/35 & $11.06$ & $9.10$ & $-1.96$ & $[-2.65, -1.20]$ \\
HF3A & 158/13 & $8.85$ & $8.28$ & $-0.57$ & $[-3.35, 2.35]$ \\
DSHP & 1,121/66 & $9.26$ & $8.95$ & $-0.30$ & $[-0.84, 0.19]$ \\
LGEM & 1,121/66 & $9.27$ & $8.95$ & $-0.32$ & $[-0.93, 0.24]$ \\
\midrule
\multicolumn{6}{@{}l@{}}{\textit{Tropical/subtropical status at initialization and verification}} \\
OFCL & 1,072/64 & $7.71$ & $9.03$ & $1.33$ & $[0.77, 1.87]$ \\
OFCI & 1,012/64 & $8.48$ & $9.16$ & $0.68$ & $[0.19, 1.14]$ \\
HWFI & 1,061/64 & $8.92$ & $8.97$ & $0.05$ & $[-0.64, 0.72]$ \\
HWRF & 1,098/64 & $9.77$ & $9.03$ & $-0.74$ & $[-1.58, 0.18]$ \\
HFAI & 636/35 & $9.78$ & $9.12$ & $-0.66$ & $[-1.84, 0.49]$ \\
HFSA & 641/35 & $10.86$ & $9.12$ & $-1.74$ & $[-2.70, -0.78]$ \\
HFBI & 636/35 & $9.67$ & $9.12$ & $-0.55$ & $[-1.43, 0.34]$ \\
HFSB & 641/35 & $11.10$ & $9.12$ & $-1.98$ & $[-2.68, -1.21]$ \\
HF3A & 158/13 & $8.85$ & $8.28$ & $-0.57$ & $[-3.35, 2.35]$ \\
DSHP & 1,116/66 & $9.28$ & $8.96$ & $-0.31$ & $[-0.85, 0.18]$ \\
LGEM & 1,116/66 & $9.30$ & $8.96$ & $-0.34$ & $[-0.95, 0.22]$ \\
\bottomrule
\end{tabular}
\end{adjustbox}
\par\smallskip
{\footnotesize Whole storms are resampled jointly for aid and model (10,000 draws. Pointwise percentile 95\% intervals). The lower block applies the NHC storm-status criterion but is still restricted to this study\textquotesingle s test sample. It is not the NHC annual verification population. Reanalysis/discussion timing remains retrospective.}
\end{table}

HAFSv1 was scheduled to enter operations at the 27 June 2023 12 UTC cycle, according to \href{https://www.weather.gov/media/notification/pdf_2023_24/scn23-60_ncep_hurricane_models.pdf}{NWS Service Change Notice 23-60}. The archived HFAI comparison includes $\resultvalue{operational_post/HFAI/preimplementation_rows}$ earlier forecasts from $\resultvalue{operational_post/HFAI/preimplementation_storms}$ storms. \autoref{tab:operational_postimplementation} starts at 28 June 00 UTC to allow for interpolated guidance from a run up to 12\,h old. It uses b-deck wind and requires tropical or subtropical status at both the initial and verifying times, following the \href{https://www.nhc.noaa.gov/verification/verify2.shtml}{NHC verification criterion}. On these $\resultvalue{operational_post/HFAI/n_obs}$ HFAI rows, its MAE is $\resultvalue{operational_post/HFAI/aid_mae}$\,kt versus $\resultvalue{operational_post/HFAI/TD_mae}$\,kt for the ensemble that includes discussions. The storm interval still includes zero. The ensemble has lower error than the late HAFS aids on their corresponding subsets.

\begin{table}[!ht]
\centering
\small
\caption{Postimplementation HAFS v1 archived guidance compared with the fixed primary T+D ensemble on matched 24\,h wind forecasts (kt). Initializations begin at 00 UTC 28 June 2023, twelve hours after the scheduled first operational HAFS cycle. The twelve-hour buffer allows for interpolated guidance from an earlier run. The storm is tropical or subtropical at both initialization and verification in the official b-deck. The model and each aid are scored on identical rows using b-deck future intensity. Neural forecasts add the predicted change to the initial intensity in the original dataset. The gap is T+D MAE minus aid MAE. Here $n/s$ gives forecast initializations/storms. The 95\% percentile intervals resample storms 10{,}000 times. The archive does not establish identical real-time information availability.}
\label{tab:operational_postimplementation}
\begin{tabular*}{\linewidth}{@{\extracolsep{\fill}}lrrrrr@{}}
\toprule
Forecast aid & $n/s$ & Aid MAE & T+D MAE & T+D$-$aid & 95\% storm CI \\
\midrule
HFAI & 600/32 & $10.07$ & $9.30$ & $-0.76$ & $[-2.02, 0.42]$ \\
HFSA & 604/32 & $11.12$ & $9.31$ & $-1.82$ & $[-2.83, -0.82]$ \\
HFBI & 600/32 & $9.89$ & $9.30$ & $-0.59$ & $[-1.51, 0.32]$ \\
HFSB & 604/32 & $11.35$ & $9.31$ & $-2.04$ & $[-2.81, -1.22]$ \\
\bottomrule
\end{tabular*}
\end{table}

All reported intervals are pointwise percentiles from 10{,}000 paired whole-storm resamples, without adjustment for multiple comparisons. These remain retrospective comparisons. Operational guidance had real-time constraints. Our models use the best-track-based initial intensity of the dataset. Discussion inputs are typically issued about 3\,h after the matched synoptic time, and image inputs use ERA5 reanalysis. Neither matching rows nor restricting the implementation date establishes equal information availability or delivery time.

\FloatBarrier
\section{Uncertainty Quantification}
\label{app:uq}

The post-hoc calibration analysis in \autoref{sec:uq} is exploratory relative to the main result for the text modality. The primary analysis uses the T+D ensemble of five models on the full 24\,h test set. A separate analysis across forecast leads evaluates T, T+D and T+I+D models trained independently at each lead. The CLEAR comparison and its conditional-coverage analysis use paired storm splits shared within that comparison but different from the analysis across forecast leads. Neither protocol calibrates on earlier seasons and evaluates on a later season.

\subsection{Dependence and empirical coverage on storm tracks}
\label{app:uq_motivation}

Split conformal prediction \citep{vovk2005algorithmic} guarantees marginal coverage under exchangeability of calibration and test scores. Residuals in these data are dependent within storms, motivating evaluation that holds out whole storms. This dependence alone does not determine the coverage of a particular calibration method. The uncalibrated spread across five seeds gives substantial undercoverage in the analysis across forecast leads. It is distinct from naive i.i.d.\ conformal calibration, which estimates a residual quantile. \autoref{tab:uq_horizon} compares calibrated coverage across the separately trained per-lead families. Coverage varies more across splits at longer leads, when fewer storms are available.

\begin{table}[!htbp]
\centering
\caption{Coverage and interval width of 90\% prediction intervals from 24 to 120\,h. Results use ensembles of five models at each forecast lead (T, forecast-tuned T+D and forecast-tuned T+I+D). Entries equally average the three families\textquotesingle\ mean coverages and median interval widths from 120 splits with whole storms held out. Interval width is the median upper-minus-lower span over all held-out interval appearances within each family. These independently trained models are separate from the primary 24\,h ablation family. Storm-block intervals use empirical bootstrap calibration without a finite-sample coverage guarantee. Bold and underline mark the best and second-best displayed values within each target and lead by lower widths (ties included). Coverage is assessed against the 90\% target alongside interval width.}
\label{tab:uq_horizon}
\small
\renewcommand{\arraystretch}{1.15}
\setlength{\tabcolsep}{4pt}
\begin{tabular*}{\linewidth}{@{\extracolsep{\fill}}lrrrrr@{}}
\toprule
Lead & 24\,h & 48\,h & 72\,h & 96\,h & 120\,h \\
Available storm pool & 66 & 53 & 51 & 38 & 28 \\
\midrule
\multicolumn{6}{l}{\textbf{VMAX (width in kt)}} \\
\multicolumn{6}{l}{Naive i.i.d.\ conformal} \\
\quad Coverage & $0.901$ & $0.895$ & $0.889$ & $0.888$ & $0.895$ \\
\quad Interval width $\downarrow$ & \underline{41.0} & \textbf{60.5} & \textbf{74.0} & \textbf{87.3} & \textbf{100.3} \\
\multicolumn{6}{l}{Storm-block bootstrap} \\
\quad Coverage & $0.918$ & $0.914$ & $0.912$ & $0.914$ & $0.909$ \\
\quad Interval width $\downarrow$ & 44.7 & 66.7 & 85.5 & \underline{102.9} & \underline{109.4} \\
\multicolumn{6}{l}{Ensemble-normalized block} \\
\quad Coverage & $0.914$ & $0.915$ & $0.911$ & $0.915$ & $0.918$ \\
\quad Interval width $\downarrow$ & \textbf{39.6} & \underline{61.6} & \underline{81.2} & 105.2 & 126.1 \\
\addlinespace
\multicolumn{6}{l}{\textbf{MSLP (width in hPa)}} \\
\multicolumn{6}{l}{Naive i.i.d.\ conformal} \\
\quad Coverage & $0.901$ & $0.899$ & $0.886$ & $0.885$ & $0.883$ \\
\quad Interval width $\downarrow$ & \underline{28.8} & \underline{43.6} & \textbf{52.1} & \textbf{64.4} & \textbf{71.9} \\
\multicolumn{6}{l}{Storm-block bootstrap} \\
\quad Coverage & $0.917$ & $0.917$ & $0.908$ & $0.914$ & $0.915$ \\
\quad Interval width $\downarrow$ & 30.9 & 48.2 & 61.0 & 72.7 & \underline{78.5} \\
\multicolumn{6}{l}{Ensemble-normalized block} \\
\quad Coverage & $0.915$ & $0.919$ & $0.907$ & $0.915$ & $0.919$ \\
\quad Interval width $\downarrow$ & \textbf{25.4} & \textbf{42.6} & \underline{56.7} & \underline{70.5} & 92.3 \\
\addlinespace
\bottomrule
\end{tabular*}
\end{table}

\subsection{Method}
\label{app:uq_method}

We use whole storms as resampling blocks, following the storm-block bootstrap used for the MAE-gain intervals in \autoref{app:subj_stratified_gain}. Let $i$ index cases from the calibration storms and $j$ cases from the held-out storms. For each case, $y$ is the observed wind or pressure change, $\hat y$ is the mean prediction of the five-seed ensemble, and $\sigma$ is the standard deviation of its five predictions. The score of calibration case $i$ is $s_i=|y_i-\hat y_i|/w_i$, where the scale $w$ is $1$ for unscaled scores and $\sigma+\epsilon$ for ensemble-normalized scores. The constant $\epsilon$ is fixed at $2$\,kt for wind and $1.5$\,hPa for pressure.

\textbf{Storm-block bootstrap calibration} resamples the calibration storms with replacement $B$ times. In resample $b=1,\dots,B$, it pools the scores of all cases from the drawn storms and computes $\hat q^{(b)}$, their split-conformal quantile at level $1-\alpha$. Here $\alpha=0.1$ is the target miscoverage of the 90\% intervals. The threshold $\hat q^\uparrow$ is the 80th percentile of $\hat q^{(1)},\dots,\hat q^{(B)}$, and the interval for held-out case $j$ is $\hat y_j\pm\hat q^\uparrow w_j$. The tables label the unscaled version \emph{storm-block bootstrap} and the ensemble-normalized version \emph{ensemble-normalized block}. We chose the 80th bootstrap percentile as a moderate upward adjustment for uncertainty in the estimated calibration threshold. Using the 90th percentile would be more conservative and produce wider intervals. The choice is an empirical heuristic, and we do not claim that 80 is optimal. The resulting intervals have no finite-sample coverage guarantee under storm dependence.

\textbf{Adaptive Conformal Inference} (ACI, \citealp{gibbs2021adaptive}) adjusts the miscoverage level at which it issues intervals. Let $\alpha_k$ be this level after $k$ updates. When the outcome of an earlier interval becomes available, ACI sets $\alpha_{k+1}=\alpha_k+\gamma(\alpha-\mathrm{err}_k)$, where $\mathrm{err}_k=1$ if the outcome fell outside that interval and $0$ otherwise. The step size is fixed at $\gamma=0.03$. We use a bounded ACI variant: when the adaptive quantile rank exceeds the calibration sample size, the half-width is capped at the largest absolute calibration residual instead of becoming infinite. We evaluate this variant empirically and do not invoke the original ACI coverage guarantee.

We compare these with \textbf{naive i.i.d.\ conformal} (a case-pooled quantile) and an \textbf{EnbPI-style proxy} (\citealp{Xu2021EnbPI}, using a sliding window of available residuals from a frozen ensemble, without the leave-one-out bootstrap retraining of the original EnbPI protocol). For a lead time $h$, an outcome for initialization time $t$ enters either online method only at $t+h$. Before issuing an interval at time $\tau$, updates use only outcomes satisfying $t+h\leq\tau$, and state is reset for each held-out storm. ACI updates use the miss indicator of the interval actually issued. Both online methods use unscaled absolute residuals. This replay enforces lead-dependent feedback delay, but does not model further operational reporting or best-track revision delays. Batch methods remain calibrated once per split, so online methods receive a different information stream.

Beyond coverage and width, we report the interval (Winkler) score \citep{gneiting2007strictly}, a proper scoring rule combining calibration and sharpness (lower is better). For an interval $[l,u]$ with nominal coverage $1-\alpha$ and observed change $y$,
\[
\mathrm{IS}_\alpha=(u-l)+\frac{2}{\alpha}(l-y)\,\mathbf{1}\{y<l\}+\frac{2}{\alpha}(y-u)\,\mathbf{1}\{y>u\}.
\]

We use $B=200$ bootstrap resamples per calibration in both the 120-split analysis across forecast leads and the 100-split paired CLEAR comparison, and $B=400$ for the leave-one-storm-out illustrative bands. The EnbPI-style proxy retains a sliding window of 200 residual scores.

We additionally benchmark CLEAR \citep{azizi2025clear} using the spread across five seeds to estimate uncertainty across fitted models (epistemic uncertainty), and a random forest quantile regressor on the 13 SHIPS predictors, fitted to the residuals of the ensemble mean, to estimate variability in outcomes given those features (aleatoric uncertainty). The random forest is the fixed choice in this analysis and is refitted with five forest seeds, distinct from the five training seeds of the fixed neural ensemble. The storm splits and all other random choices are held fixed across forest seeds. Each of 100 repetitions partitions the \resultvalue{primary/n_storms} test-period storms into a 60\% calibration pool and a 40\% held-out pool. CLEAR reserves part of the calibration pool for fitting the aleatoric regressor and uses the remainder to calibrate its mixing and scaling parameters. We compare naive CLEAR calibration with a storm-block scaling of the same uncertainty inputs. The cross-fitted variant in the conditional analysis uses a two-fold storm-level swap within the calibration pool. This allocation differs from the conformal baselines\textquotesingle\ use of the full calibration pool, and its effect need not have a fixed direction. All seven main methods share the same held-out folds within each repetition.

Within each forest seed, we average coverage across the held-out splits and pool interval widths before calculating their median and mean, then report the mean of these summaries across the five forest seeds. The other methods do not depend on the forest seed and are shown once. The separate Split SD column averages the SD of coverage across splits within each seed. Because repetitions overlap within one storm pool, split dispersion is descriptive rather than a standard error from independent samples. \autoref{tab:uq_comparison} reports these summaries alongside interval score and a finite-level weighted interval score (WIS).

WIS \citep{bracher2021evaluating} combines interval scores over eleven central levels (10--98\%), weighted by $\alpha/2$, with an absolute error term with weight $0.5$ at the common point forecast from the neural ensemble, and divides the sum by $11.5$. That common point forecast need not be the conditional median of each method\textquotesingle s potentially asymmetric intervals. We therefore interpret this as a score of the supplied intervals and point forecast, without asserting an exact continuous ranked probability score for a fully specified predictive distribution. SEMF also evaluates regression intervals using coverage and normalized width \citep{Azizi2025SEMF}. We retain widths in physical units. Width and score should be read alongside coverage. A narrower band alone need not be better.

\begin{table}[!htbp]
\centering
\caption{Comparison of 90\% prediction intervals for the primary 24\,h T+D ensemble on 100 paired splits of 1,150 cases from 66 storms. CLEAR entries are means over five forest seeds. Widths and scores use kt for wind and hPa for pressure. Bold and underline mark the best and second-best displayed values within each target by lower widths and scores (ties included). Coverage is assessed against the 90\% target alongside interval width.}
\label{tab:uq_comparison}
\small
\renewcommand{\arraystretch}{1.00}
\setlength{\tabcolsep}{4pt}
\begin{adjustbox}{max width=\linewidth}
\begin{tabular}{@{}lrrrrrr@{}}
\toprule
Method & Coverage & Split SD & Median width & Mean width & IS$\downarrow$ & WIS$\downarrow$ \\
\midrule
\multicolumn{7}{@{}l}{\textit{Wind ($V_{\max}$, kt)}} \\
Naive i.i.d.\ conformal & 0.905 & 0.039 & 39.1 & 39.4 & 57.0 & 5.87 \\
EnbPI-style proxy & 0.891 & 0.061 & \underline{38.5} & \underline{38.9} & 58.3 & 5.89 \\
Storm-block bootstrap & 0.920 & 0.036 & 41.8 & 42.2 & 57.5 & 5.88 \\
Ensemble-normalized block & 0.917 & 0.031 & 39.5 & 42.8 & 57.3 & 5.87 \\
ACI (delayed feedback) & 0.910 & 0.027 & 38.6 & 43.2 & 58.6 & 5.87 \\
CLEAR, naive calibration & $0.895$ & 0.043 & $\boldsymbol{36.3}$ & $\boldsymbol{38.1}$ & $\boldsymbol{54.4}$ & $\boldsymbol{5.80}$ \\
CLEAR, storm-block calibration & $0.913$ & 0.040 & $39.1$ & $40.9$ & $\underline{54.8}$ & $\underline{5.82}$ \\
\midrule
\multicolumn{7}{@{}l}{\textit{Pressure (MSLP, hPa)}} \\
Naive i.i.d.\ conformal & 0.902 & 0.039 & 26.7 & 26.9 & 42.4 & 3.98 \\
EnbPI-style proxy & 0.882 & 0.059 & \underline{23.5} & \underline{25.8} & 43.3 & 4.00 \\
Storm-block bootstrap & 0.917 & 0.034 & 30.0 & 29.5 & 42.8 & 3.99 \\
Ensemble-normalized block & 0.916 & 0.028 & 24.4 & 27.5 & 40.2 & 3.91 \\
ACI (delayed feedback) & 0.912 & 0.023 & 25.8 & 29.8 & 43.8 & 3.97 \\
CLEAR, naive calibration & $0.897$ & 0.040 & $\boldsymbol{22.9}$ & $\boldsymbol{24.6}$ & $\boldsymbol{37.5}$ & $\boldsymbol{3.86}$ \\
CLEAR, storm-block calibration & $0.913$ & 0.039 & $24.6$ & $26.5$ & $\underline{37.7}$ & $\underline{3.87}$ \\
\bottomrule
\end{tabular}
\end{adjustbox}
\end{table}

\FloatBarrier

\subsection{Interval performance on individual storms}
\label{app:uq_storms}

\begin{figure}[H]
\centering
\includegraphics[width=\linewidth,height=0.92\textheight,keepaspectratio]{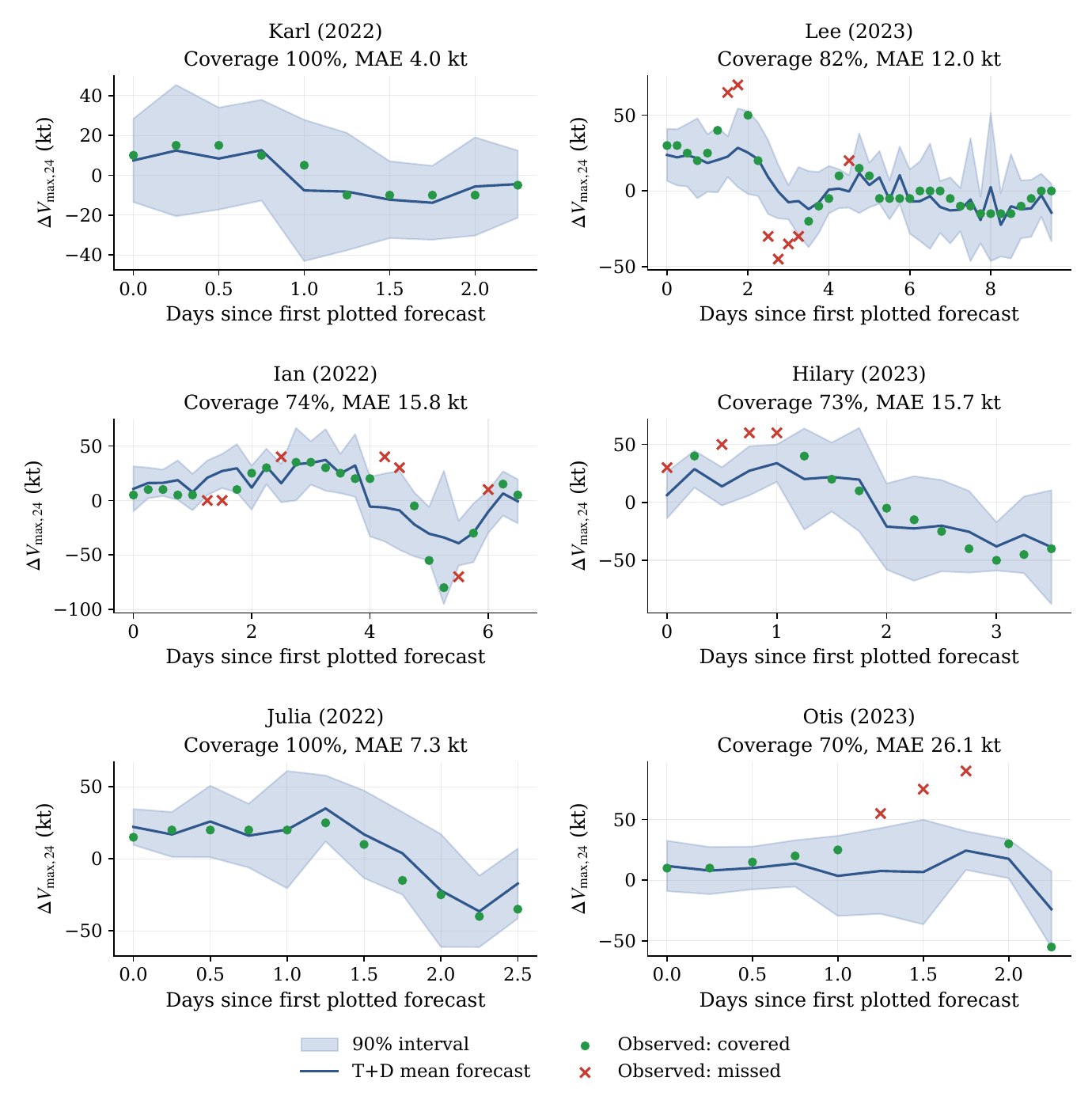}
\caption{Nominal 90\% intervals for the primary 24\,h T+D ensemble of five models. Residuals are normalized by the ensemble spread, and the intervals are calibrated by resampling whole storms. Each storm is excluded from its own calibration set. Green and red markers indicate covered and missed observed changes. Panel titles report the coverage and point MAE of each storm. The horizontal axis uses forecast initialization time, including any gaps. Otis illustrates that near-nominal aggregate coverage does not prevent large misses during rapid intensification.}
\label{fig:uq_storm_panels}
\end{figure}

\autoref{fig:uq_storm_panels} shows 90\% ensemble-normalized storm-block bands for six storms, each excluded from its own calibration set. Their coverage varies across storms. Otis remains a difficult case, with point MAE $\resultvalue{interval/OTIS/T+D/mae}$\,kt and coverage $\resultvalue{interval/OTIS/T+D/coverage_percent}\%$. Its largest rapid intensification changes remain outside the band. The panels show variation that aggregate coverage does not capture.

\begin{figure}[H]
\centering
\includegraphics[width=\linewidth,height=0.56\textheight,keepaspectratio]{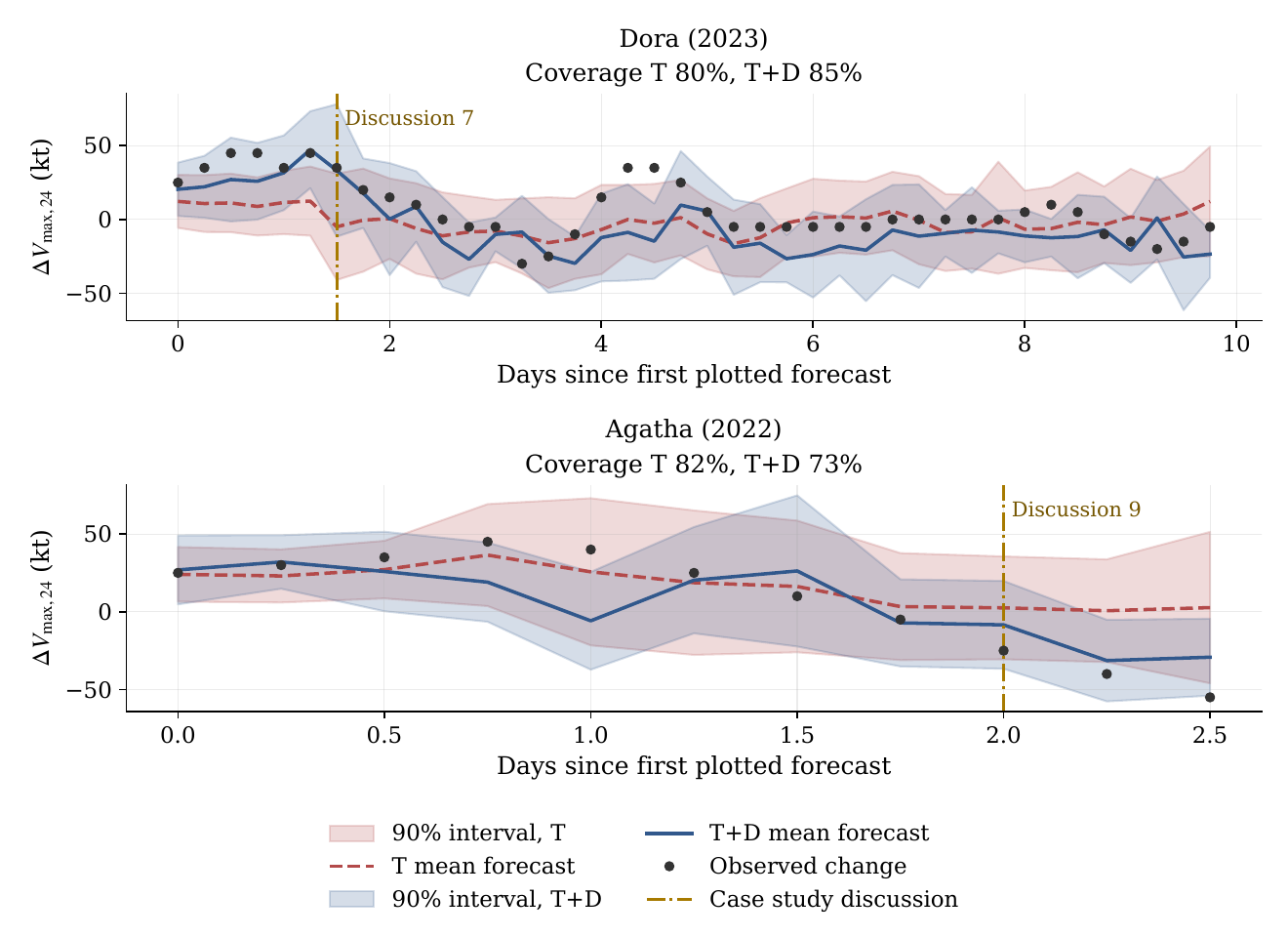}
\caption{Held-out examples for Dora (2023) and Agatha (2022). Dots show observed 24\,h wind changes. Lines show mean forecasts from five neural networks using tabular inputs, with and without discussions. The 90\% intervals use ensemble normalization and exclude the illustrated storm from calibration. Neither storm was used for training. Vertical lines mark the forecast initializations for the fixed interpretability examples associated with Dora Discussion 7 and Agatha Discussion 9. Both discussions were issued three hours after their marked initialization.}
\label{fig:uq_modality_cases}
\end{figure}

\emph{How discussion text affects the intervals.} \autoref{tab:uq_text_gain} compares ensembles using tabular inputs with and without discussions. Both use the same calibration and evaluation storms, with residual scores normalized by ensemble spread and calibrated in whole storm blocks. Adding discussion lowers the wind interval score from \resultvalue{uq/textgain/vmax/T/interval_score} to \resultvalue{uq/textgain/vmax/T+D/interval_score}, with similar marginal coverage (\resultvalue{uq/textgain/vmax/T/coverage_mean} and \resultvalue{uq/textgain/vmax/T+D/coverage_mean}). The pressure score also decreases, as shown in the table. These aggregate comparisons do not imply uniformly narrower intervals at unchanged coverage for individual storms. On Dora, adding discussions lowers the median interval width across the storm and covers more cases. On Agatha, it also lowers the storm\textquotesingle s median width but covers fewer (\autoref{fig:uq_modality_cases}). Adding the ERA5 wind fields changes both centers and widths, with different consequences across storms (\autoref{tab:uq_modality_cases}). All three modality configurations still miss the largest Otis intensification changes.

\begin{table}[H]
\centering
\caption{Primary T and T+D ensembles of five models at 24\,h. The 90\% intervals use ensemble normalization and whole storm blocks on 100 shared splits. Coverage averages the held-out splits. Widths pool all held-out intervals. Widths and interval scores (IS) use kt for wind and hPa for pressure. Lower IS is better. Bold marks the lower width or IS of the two ensembles.}
\label{tab:uq_text_gain}
\small
\renewcommand{\arraystretch}{1.00}
\setlength{\tabcolsep}{4pt}
\fullwidthresulttable{lrrrrrrrr}{
\toprule
 & \multicolumn{4}{c}{$V_{\max}$ (kt)} & \multicolumn{4}{c}{MSLP (hPa)} \\
\cmidrule(lr){2-5}\cmidrule(lr){6-9}
Ensemble & Coverage & Median width & Mean width & IS$\downarrow$ & Coverage & Median width & Mean width & IS$\downarrow$ \\
\midrule
T & 0.917 & 41.9 & 45.9 & 62.3 & 0.918 & 26.4 & 29.7 & 42.9 \\
T+D & 0.917 & \textbf{39.5} & \textbf{42.8} & \textbf{57.3} & 0.916 & \textbf{24.4} & \textbf{27.5} & \textbf{40.2} \\
\bottomrule
}
\end{table}

\FloatBarrier

\autoref{tab:uq_modality_cases} lists the observed changes, ensemble centers and leave-one-storm-out intervals for eleven discussions used as case studies for T, T+D and T+I+D. All three columns use ensembles from the primary 24\,h family. Each forecast predicts the change from its synoptic time to 24\,h later. The named discussion is the one matched to the case at that synoptic time, usually issued three hours afterwards.

\begin{table}[!htbp]
\centering
\caption{Primary 24\,h predictions from ensembles of five models and 90\% ensemble-normalized storm-block intervals for eleven discussions used as case studies, in kt. Each illustrated storm is excluded from calibration, which uses 400 storm resamples and the 80th bootstrap percentile. \cmark\ marks a covered case and \xmark\ marks a miss. These ensemble predictions differ from the individual checkpoints used for attribution.}
\label{tab:uq_modality_cases}
\small
\renewcommand{\arraystretch}{1.15}
\setlength{\tabcolsep}{4pt}
\begin{adjustbox}{max width=\linewidth}
\begin{tabular}{@{}lrrrr@{}}
\toprule
Case & Observed change & T & T+D & T+I+D \\
\midrule
Lee disc.\ 9 & $+50$ & $+27.7$ $[-20,+76]$\,\cmark & $+25.3$ $[-2,+53]$\,\cmark & $+24.2$ $[-10,+58]$\,\cmark \\
Fiona disc.\ 24 & $+20$ & $+2.9$ $[-48,+54]$\,\cmark & $+4.5$ $[-36,+45]$\,\cmark & $+2.6$ $[-38,+43]$\,\cmark \\
Dora disc.\ 7 & $+35$ & $-5.0$ $[-41,+31]$\,\xmark & $+33.2$ $[-12,+78]$\,\cmark & $+31.6$ $[-12,+75]$\,\cmark \\
Otis disc.\ 8 & $+90$ & $+4.7$ $[-29,+38]$\,\xmark & $+24.4$ $[+9,+40]$\,\xmark & $+24.0$ $[+3,+45]$\,\xmark \\
Ian disc.\ 10 & $+30$ & $+14.1$ $[-2,+31]$\,\cmark & $+31.2$ $[+15,+47]$\,\cmark & $+30.7$ $[+9,+52]$\,\cmark \\
Idalia disc.\ 3 & $+25$ & $+10.1$ $[-11,+31]$\,\cmark & $+14.3$ $[+1,+28]$\,\cmark & $+13.5$ $[+3,+24]$\,\xmark \\
Agatha disc.\ 9 & $-25$ & $+2.5$ $[-31,+36]$\,\cmark & $-8.4$ $[-37,+20]$\,\cmark & $-8.7$ $[-41,+23]$\,\cmark \\
Darby disc.\ 24 & $-25$ & $-26.5$ $[-50,-3]$\,\cmark & $-37.8$ $[-69,-7]$\,\cmark & $-43.4$ $[-79,-7]$\,\cmark \\
Madeline disc.\ 7 & $-15$ & $-7.8$ $[-23,+8]$\,\cmark & $-3.4$ $[-26,+19]$\,\cmark & $-5.4$ $[-31,+20]$\,\cmark \\
Philippe disc.\ 35 & $+0$ & $+0.8$ $[-18,+19]$\,\cmark & $+3.0$ $[-16,+22]$\,\cmark & $+1.2$ $[-15,+18]$\,\cmark \\
Calvin disc.\ 20 & $-20$ & $-10.9$ $[-32,+10]$\,\cmark & $-21.3$ $[-38,-5]$\,\cmark & $-20.1$ $[-42,+1]$\,\cmark \\
\bottomrule
\end{tabular}
\end{adjustbox}
\end{table}

\begin{figure}[!tbp]
\centering
\includegraphics[width=\linewidth]{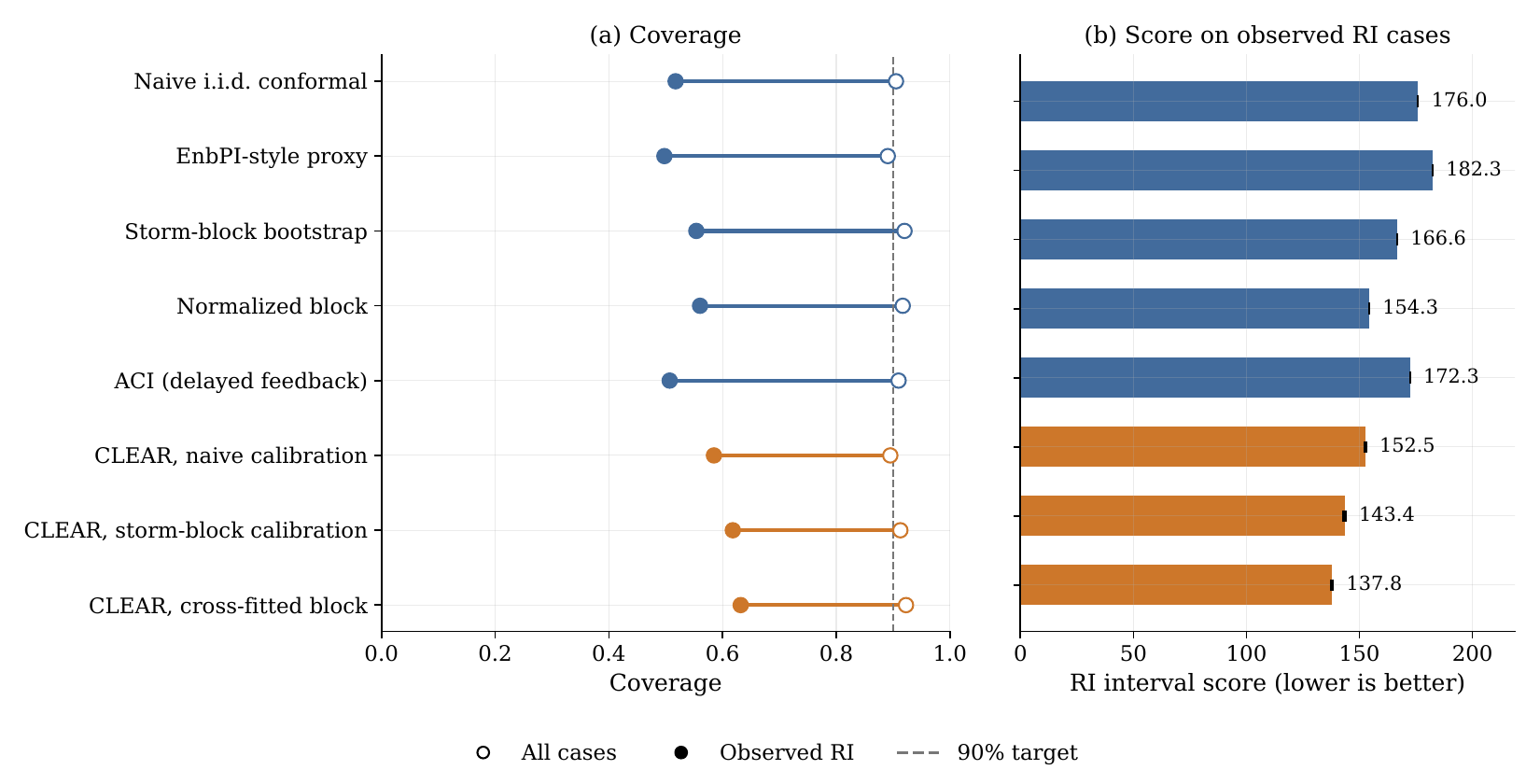}
\caption{Conditional performance on observed rapid intensification (at least 30\,kt in 24\,h) for the primary T+D ensemble. Open and filled markers show marginal and RI coverage, respectively, with a dashed 90\% target. Normalized block denotes the ensemble-normalized block. The right panel shows the RI interval score (lower is better). CLEAR points average five forest seeds, with error bars of one sample SD.}
\label{fig:uq_ri_conditional}
\end{figure}

\emph{Conditional coverage on rapid intensification.} \autoref{fig:uq_ri_conditional} evaluates observed wind rises of at least $+30$\,kt in 24\,h. It contains \resultvalue{ri/unique} unique test cases appearing \resultvalue{ri/appearances} times in held-out folds across the paired splits. These pooled appearances are repeated cases, not independent events. Every evaluated method falls below nominal coverage on this subset, even when marginal coverage is near its target. The figure compares coverage and interval score, which should be read together. This subset diagnostic evaluates calibration on realized rapid intensification. It does not identify those events before they occur or provide a coverage guarantee for future events.

\subsection{Limitations}
\label{app:uq_limitations}

This analysis is exploratory and narrower than the paper\textquotesingle s evaluation of point forecasts. First, the repeated random storm splits reuse the 2022--2023 storm pool. They do not test calibration on earlier seasons followed by deployment on later ones. Seasonal and basin structure can also limit the exchangeability of whole storms. Second, the paired CLEAR comparison is at 24\,h. \autoref{tab:uq_horizon} covers leads from 24 to 120\,h, but only for three of the seven methods. It uses separately trained five-seed ensembles for each lead and its own set of 120 splits. Third, the baseline interval constructions are symmetric around their point forecast and cannot represent residual skew in the way a conditional quantile model might \citep{romano2019cqr}. Fourth, online feedback enters only after the lead time has elapsed, but operational reporting delays and later best-track revisions are not represented. Finally, we have not reimplemented the external literature on uncertainty quantification for TCs (e.g.\ MTCIE, \citealp{Song2024MTCIE}, \citealp{Meng2024,Chen2025}) on this dataset. These results are an empirical calibration comparison, not a claim of state-of-the-art uncertainty quantification or operational coverage guarantees. Future work could test calibration based on input availability and disagreement between modalities \citep{azizi2026modalitycalibration}. Coverage could also be targeted within a group of cases identified before the forecast, such as those at risk of rapid intensification, as was done for groups of questions in language model generation \citep{azizi2026matchabilityllms}. Both would require choices fixed before calibration and an evaluation that respects storm dependence.

\FloatBarrier
\section{Evaluation Sanity Checks}
\label{sec:sanity_checks}

\subsection{Directional-accuracy confusion matrices}
\label{app:dacc_confusion}

\autoref{tab:dacc_confusion} reports strict-sign confusion matrices for predictions from the tabular ensemble and the ensemble that includes discussions on the full test set. The observed classes use the exact physical targets. Neither ensemble emits exactly zero, so both have zero recall for steady cases under this strict convention. The ensemble that includes discussions has higher overall and balanced accuracy in this descriptive comparison. These are ensemble scores, which differ from the averages of scores over seeds in the primary comparison.

As a supplemental check, \autoref{tab:dacc_tolerance} fixes predicted steady to $|\Delta\hat V|\le2.5$\,kt, half the smallest nonzero training change. This threshold is not tuned on the test set and does not guarantee calibration of the predicted class frequencies. The ensemble that includes discussions has higher overall and balanced accuracy under this convention, mainly through improved weakening recall. Steady recall remains low for both models. These tolerance scores are kept separate from strict-sign DAcc.

\begin{table}[H]
\centering
\caption{Confusion matrices for 24\,h wind change from the T and T+D ensembles, each averaging five models on all 1,150 cases (66 storms). Rows are observed classes. Each model's three count columns are predicted strengthening/steady/weakening. Observed labels use the exact signs of the observed changes, including exact zeros. Predicted labels use strict sign, with steady only for exactly zero prediction. These ensemble accuracies differ from mean accuracies across individual seeds.}
\label{tab:dacc_confusion}
\small
\renewcommand{\arraystretch}{1.12}
\setlength{\tabcolsep}{4pt}
\setlength{\aboverulesep}{0pt}
\setlength{\belowrulesep}{0pt}
\begin{adjustbox}{max width=\linewidth}
\begin{tabular}{@{}lccccccccc!{\vrule width 0.8pt}lcc@{}}
\toprule
\multirow{2}{*}{Observed} & \multirow{2}{*}{$N$} & \multicolumn{3}{c}{T predictions} & \multicolumn{3}{c}{T+D predictions} & \multicolumn{2}{c!{\vrule width 0.8pt}}{Recall} & \multirow{2}{*}{Accuracy} & \multirow{2}{*}{T} & \multirow{2}{*}{T+D} \\
\cmidrule{3-10}
 &  & + & 0 & - & + & 0 & - & T & T+D &  &  &  \\
\midrule
Strengthening & 586 & 521 & 0 & 65 & 524 & 0 & 62 & $0.889$ & $0.894$ & \multirow{1.5}{*}{Overall} & \multirow{1.5}{*}{$0.690$} & \multirow{1.5}{*}{$0.737$} \\
Steady & 165 & 104 & 0 & 61 & 103 & 0 & 62 & $0.000$ & $0.000$ &  &  &  \\
Weakening & 399 & 126 & 0 & 273 & 76 & 0 & 323 & $0.684$ & $0.810$ & \multirow{-1.5}{*}{Balanced} & \multirow{-1.5}{*}{$0.524$} & \multirow{-1.5}{*}{$0.568$} \\
\bottomrule
\end{tabular}
\end{adjustbox}
\end{table}

\begin{table}[H]
\centering
\caption{Confusion matrices for 24\,h wind change from the T and T+D ensembles, each averaging five models on all 1,150 cases (66 storms). Rows are observed classes. Each model's three count columns are predicted strengthening/steady/weakening. Observed labels use the exact signs of the observed changes, including exact zeros. Predictions are steady within $\pm2.5$\,kt, half the smallest nonzero training change. This supplemental tolerance convention is distinct from strict-sign headline DAcc and was not optimized on test labels.}
\label{tab:dacc_tolerance}
\small
\renewcommand{\arraystretch}{1.12}
\setlength{\tabcolsep}{4pt}
\setlength{\aboverulesep}{0pt}
\setlength{\belowrulesep}{0pt}
\begin{adjustbox}{max width=\linewidth}
\begin{tabular}{@{}lccccccccc!{\vrule width 0.8pt}lcc@{}}
\toprule
\multirow{2}{*}{Observed} & \multirow{2}{*}{$N$} & \multicolumn{3}{c}{T predictions} & \multicolumn{3}{c}{T+D predictions} & \multicolumn{2}{c!{\vrule width 0.8pt}}{Recall} & \multirow{2}{*}{Accuracy} & \multirow{2}{*}{T} & \multirow{2}{*}{T+D} \\
\cmidrule{3-10}
 &  & + & 0 & - & + & 0 & - & T & T+D &  &  &  \\
\midrule
Strengthening & 586 & 467 & 86 & 33 & 488 & 61 & 37 & $0.797$ & $0.833$ & \multirow{1.5}{*}{Overall} & \multirow{1.5}{*}{$0.655$} & \multirow{1.5}{*}{$0.710$} \\
Steady & 165 & 78 & 48 & 39 & 79 & 42 & 44 & $0.291$ & $0.255$ &  &  &  \\
Weakening & 399 & 91 & 70 & 238 & 53 & 59 & 287 & $0.596$ & $0.719$ & \multirow{-1.5}{*}{Balanced} & \multirow{-1.5}{*}{$0.561$} & \multirow{-1.5}{*}{$0.602$} \\
\bottomrule
\end{tabular}
\end{adjustbox}
\end{table}

\subsection{Discussion availability, storm phase, and missingness}
\label{app:missingness}

Discussion availability could encode storm phase because fused models zero-pad the text branch when no discussion is matched. Three checks assess this concern. First, a baseline that predicts the average change in the training data, separately for cases with and without a matched discussion, performs close to climatology, although this does not exclude interactions with other features. Second, the wind gain associated with discussions persists when every evaluated row has a matched discussion (\autoref{tab:missingness_sensitivity}). Third, it also persists in the pre-landfall subset, while the smaller post-landfall subset has a larger but less precise point estimate. Here landfall is a proxy based on the first zero IBTrACS \citep{Knapp2010IBTrACS} distance-to-land value. Storms with no such record retain all rows in the pre-landfall group. The pressure gain before first landfall remains unresolved. These pointwise subset comparisons do not establish a between-phase interaction or a mechanism. The selected SHIPS inputs lack an explicit distance-to-land predictor, but include latitude and longitude, so geography is not absent from the numerical inputs.

\begin{table}[H]
\centering
\caption{Descriptive sensitivity of the 24\,h MAE gain from T to T+D using ensembles of five models. Positive gain favors T+D. Pre-landfall means strictly before the first IBTrACS track point with DIST2LAND=0. Storms without such a track point retain all cases. This is a center-over-land proxy, not a causal explanation.}
\label{tab:missingness_sensitivity}
\small
\renewcommand{\arraystretch}{1.12}
\setlength{\tabcolsep}{4pt}
\begin{adjustbox}{max width=\linewidth}
\begin{tabular}{@{}lccccc@{}}
\toprule
Stratum & $N/S$ & Target & T MAE & T+D MAE & Gain [95\% CI] \\
\midrule
\multirow{2}{*}{All rows} & 1{,}150/66 & $V_{\max}$ (kt) & $9.89$ & $8.90$ & $0.99$ $[0.32, 1.70]$ \\
 & 1{,}150/66 & MSLP (hPa) & $6.43$ & $5.87$ & $0.57$ $[0.07, 1.05]$ \\
\midrule
\multirow{2}{*}{Discussion present} & 1{,}105/64 & $V_{\max}$ (kt) & $10.05$ & $8.98$ & $1.07$ $[0.37, 1.82]$ \\
 & 1{,}105/64 & MSLP (hPa) & $6.55$ & $5.94$ & $0.61$ $[0.10, 1.12]$ \\
\midrule
\multirow{2}{*}{Discussion missing} & 45/31 & $V_{\max}$ (kt) & $5.94$ & $6.90$ & $-0.95$ $[-2.64, 0.62]$ \\
 & 45/31 & MSLP (hPa) & $3.43$ & $3.97$ & $-0.54$ $[-1.65, 0.40]$ \\
\midrule
\multirow{2}{*}{Pre-landfall} & 1{,}039/64 & $V_{\max}$ (kt) & $9.46$ & $8.74$ & $0.71$ $[0.12, 1.34]$ \\
 & 1{,}039/64 & MSLP (hPa) & $6.13$ & $5.72$ & $0.41$ $[-0.06, 0.87]$ \\
\midrule
\multirow{2}{*}{At/after first landfall} & 111/9 & $V_{\max}$ (kt) & $13.92$ & $10.33$ & $3.59$ $[-0.02, 9.97]$ \\
 & 111/9 & MSLP (hPa) & $9.27$ & $7.22$ & $2.05$ $[-0.22, 5.95]$ \\
\midrule
\multirow{2}{*}{Pre-landfall, matched} & 994/62 & $V_{\max}$ (kt) & $9.62$ & $8.83$ & $0.79$ $[0.15, 1.44]$ \\
 & 994/62 & MSLP (hPa) & $6.25$ & $5.80$ & $0.45$ $[-0.04, 0.94]$ \\
\bottomrule
\end{tabular}
\end{adjustbox}
\par\smallskip
{\footnotesize Wind/pressure MAE units are kt/hPa. All intervals resample whole storms (10,000 draws, percentile 95\%), retain within-storm dependence, and are pointwise without multiplicity adjustment. Stratum differences are not formal interaction tests.}
\end{table}

\subsection{Comparison on identical cases}
\label{app:temporal_subset}

The temporal architecture requires eight consecutive valid, nonzero ERA5 frames ending at the synoptic time. Matching storm identifiers and timestamps gives \resultvalue{temporal/common/n_obs} valid test cases from \resultvalue{temporal/common/n_storms} storms. Requiring a matched discussion leaves \resultvalue{temporal/matched/n_obs} cases. \autoref{tab:temporal_common_subset} recomputes every model on identical row sets and reports means and sample SDs over the five seeds. In both row sets, the neural model that includes discussions has lower mean wind MAE than the full temporal model with XGBoost. The standard deviations describe variation across training runs. They are not confidence intervals for differences between models.

\begin{table}[H]
\centering
\caption{Comparisons of 24\,h forecasts on identical cases within each block. Entries are mean $\pm$ sample SD across five seeds. MAE is in kt/hPa. Bold and underline identify the best and second-best means within each block. Temporal ablation encoders are trained separately from the main temporal encoder. The T+I+D ablation has the main encoder's configuration and reproduces the predictions of the XGB emb row exactly.}
\label{tab:temporal_common_subset}
\small
\renewcommand{\arraystretch}{1.12}
\setlength{\tabcolsep}{4pt}
\begin{adjustbox}{max width=\linewidth}
\begin{tabular}{@{}lcccc@{}}
\toprule
\multirow{2}{*}{Model} & \multicolumn{2}{c}{$R^2$} & \multicolumn{2}{c}{MAE (kt/hPa)} \\
\cmidrule(lr){2-3} \cmidrule(lr){4-5}
 & $V_{\max}$ & MSLP & $V_{\max}$ & MSLP \\
\midrule
\multicolumn{5}{@{}l}{\textit{All eligible sequences (834 cases, 55 storms)}} \\
Neural T+D & $\boldsymbol{0.49}\pm0.02$ & $\boldsymbol{0.50}\pm0.02$ & $\underline{9.47}\pm0.20$ & $\underline{6.33}\pm0.12$ \\
Neural T+I+D & $\boldsymbol{0.49}\pm0.01$ & $\boldsymbol{0.50}\pm0.01$ & $\boldsymbol{9.45}\pm0.11$ & $\boldsymbol{6.31}\pm0.06$ \\
Neural T & $0.41\pm0.01$ & $0.44\pm0.02$ & $10.03\pm0.13$ & $6.70\pm0.12$ \\
Hybrid T & $0.42\pm0.01$ & $0.43\pm0.00$ & $9.84\pm0.07$ & $6.74\pm0.02$ \\
Hybrid T+I+D & $0.46\pm0.02$ & $0.48\pm0.03$ & $9.89\pm0.19$ & $6.63\pm0.19$ \\
Hybrid T+D & $\underline{0.47}\pm0.01$ & $\underline{0.49}\pm0.01$ & $9.55\pm0.10$ & $6.56\pm0.11$ \\
Temporal direct GRU & $0.44\pm0.02$ & $0.45\pm0.01$ & $9.92\pm0.27$ & $6.63\pm0.13$ \\
Temporal T+I+D (GRU+XGB) & $0.46\pm0.02$ & $0.47\pm0.02$ & $9.68\pm0.21$ & $6.41\pm0.17$ \\
Temporal XGB emb & $0.45\pm0.02$ & $0.47\pm0.02$ & $9.71\pm0.27$ & $6.43\pm0.18$ \\
Temporal T+I ablation & $0.36\pm0.01$ & $0.37\pm0.03$ & $10.56\pm0.11$ & $7.05\pm0.10$ \\
Temporal T+I+D ablation & $0.45\pm0.02$ & $0.47\pm0.02$ & $9.71\pm0.27$ & $6.43\pm0.18$ \\
\midrule
\multicolumn{5}{@{}l}{\textit{Sequences with matched discussions (820 cases, 55 storms)}} \\
Neural D & $0.46\pm0.02$ & $0.44\pm0.02$ & $9.82\pm0.19$ & $6.65\pm0.15$ \\
Neural T+D & $\boldsymbol{0.50}\pm0.02$ & $\boldsymbol{0.50}\pm0.02$ & $\underline{9.52}\pm0.20$ & $\underline{6.35}\pm0.12$ \\
Neural T+I+D & $\underline{0.49}\pm0.01$ & $\boldsymbol{0.50}\pm0.01$ & $\boldsymbol{9.51}\pm0.11$ & $\boldsymbol{6.34}\pm0.06$ \\
Neural T & $0.41\pm0.01$ & $0.44\pm0.02$ & $10.12\pm0.13$ & $6.73\pm0.13$ \\
Hybrid T & $0.42\pm0.01$ & $0.43\pm0.00$ & $9.95\pm0.07$ & $6.79\pm0.02$ \\
Hybrid T+I+D & $0.46\pm0.03$ & $0.48\pm0.03$ & $9.98\pm0.20$ & $6.69\pm0.19$ \\
Hybrid T+D & $0.47\pm0.01$ & $\underline{0.49}\pm0.01$ & $9.64\pm0.10$ & $6.62\pm0.12$ \\
Temporal direct GRU & $0.44\pm0.02$ & $0.45\pm0.01$ & $9.98\pm0.26$ & $6.66\pm0.12$ \\
Temporal T+I+D (GRU+XGB) & $0.46\pm0.02$ & $0.47\pm0.02$ & $9.76\pm0.21$ & $6.45\pm0.17$ \\
Temporal XGB emb & $0.45\pm0.02$ & $0.47\pm0.02$ & $9.78\pm0.26$ & $6.46\pm0.18$ \\
Temporal T+I ablation & $0.36\pm0.01$ & $0.38\pm0.03$ & $10.63\pm0.11$ & $7.07\pm0.11$ \\
Temporal T+I+D ablation & $0.45\pm0.02$ & $0.47\pm0.02$ & $9.78\pm0.26$ & $6.46\pm0.18$ \\
\bottomrule
\end{tabular}
\end{adjustbox}
\end{table}

In paired comparisons of fixed ensembles, adding discussions to tabular inputs lowers the neural wind MAE on these cases, but on this smaller population the interval narrowly includes zero. The corresponding pressure interval also spans zero. This neural ensemble has lower wind MAE than XGBoost using the main temporal encoder's embedding alone, with an interval that spans zero. Intervals for adding discussions to the temporal ablation and hybrid models span zero as well. These pointwise intervals use whole storm resampling and pooled MAE, with fitted models held fixed. They are not adjusted for multiple comparisons and do not establish a universal architecture ranking.

\FloatBarrier
\noindent\begin{minipage}{\linewidth}
\section{Text Backbone and Fusion Sensitivity}
\label{sec:fusion_comparison}

Tables~\ref{tab:fusion_roles_24h}--\ref{tab:fusion_roles_alternative_leads} compare two text backbones and two fusion operators on identical cases within each modality and forecast lead. Generic RoBERTa supplies features without cyclone-specific fine-tuning. Forecast-tuned RoBERTa is first trained to predict intensity change from discussions alone, matching the seed and forecast lead of the fused model. Its backbone is then frozen while the text projection, other modality branches and fusion head are trained. The tables name the text backbone and fusion operator explicitly. T+I has no text backbone. The primary 24\,h models and models trained separately at each forecast lead are separate experiment families.

\begin{table}[H]
\centering
\caption{Sensitivity to the text backbone and fusion operator for the primary 24\,h models. Entries are mean $\pm$ sample SD across five seeds. MAE units are kt/hPa. Backbone definitions and training protocols are given in the text. Bold and underline indicate the best and second-best means within each lead.}
\label{tab:fusion_roles_24h}
\small
\renewcommand{\arraystretch}{1.0}
\setlength{\tabcolsep}{4pt}
\begin{adjustbox}{max width=\linewidth}
\begin{tabular}{@{}lccccccc@{}}
\toprule
\multirow{2}{*}{Lead} & \multirow{2}{*}{Modality} & \multirow{2}{*}{Text backbone} & \multirow{2}{*}{Fusion} & \multicolumn{2}{c}{$R^2$} & \multicolumn{2}{c}{MAE (kt/hPa)} \\
\cmidrule(lr){5-6} \cmidrule(lr){7-8}
 &  &  &  & $V_{\max}$ & MSLP & $V_{\max}$ & MSLP \\
\midrule
\multirow{10}{*}{24\,h} & T+I & No text & Concat & $0.45\pm0.01$ & $0.45\pm0.01$ & $10.27\pm0.06$ & $6.72\pm0.05$ \\
 & T+I & No text & Attention & $0.43\pm0.03$ & $0.41\pm0.05$ & $10.30\pm0.19$ & $6.81\pm0.35$ \\
\cmidrule(lr){2-8}
 & T+D & Generic & Concat & $\boldsymbol{0.54}\pm0.01$ & $\boldsymbol{0.53}\pm0.01$ & $\boldsymbol{9.26}\pm0.08$ & $\boldsymbol{6.12}\pm0.07$ \\
 & T+D & Forecast-tuned & Concat & $\underline{0.53}\pm0.01$ & $\underline{0.51}\pm0.01$ & $9.46\pm0.16$ & $6.23\pm0.08$ \\
 & T+D & Generic & Attention & $0.44\pm0.03$ & $0.43\pm0.02$ & $10.25\pm0.46$ & $6.64\pm0.20$ \\
\cmidrule(lr){2-8}
 & I+D & Generic & Attention & $0.42\pm0.02$ & $0.38\pm0.02$ & $10.87\pm0.30$ & $7.24\pm0.17$ \\
 & I+D & Forecast-tuned & Attention & $0.50\pm0.01$ & $0.48\pm0.02$ & $9.74\pm0.27$ & $6.45\pm0.18$ \\
 & I+D & Generic & Concat & $0.39\pm0.02$ & $0.35\pm0.01$ & $11.03\pm0.14$ & $7.42\pm0.07$ \\
\cmidrule(lr){2-8}
 & T+I+D & Generic & Concat & $\underline{0.53}\pm0.02$ & $\boldsymbol{0.53}\pm0.02$ & $9.46\pm0.31$ & $\underline{6.17}\pm0.13$ \\
 & T+I+D & Forecast-tuned & Concat & $\underline{0.53}\pm0.01$ & $\underline{0.51}\pm0.01$ & $\underline{9.45}\pm0.13$ & $6.23\pm0.08$ \\
\bottomrule
\end{tabular}
\end{adjustbox}
\par\smallskip
{\footnotesize Configurations within a modality and lead use the same valid-target population (Table~\ref{tab:neural_populations}). The primary and per-lead families are separate experiments. Configurations can differ in tuned learning rate and training batch size, so these comparisons do not isolate the effect of the backbone or fusion operator. These descriptive configuration comparisons do not establish a universally optimal backbone or fusion operator.}
\end{table}

\end{minipage}

The primary 24\,h forecast-tuned T+D model with concatenation reaches mean wind MAE $\resultvalue{seed/abl/tab+disc/VMAX/24/mae_mean}$\,kt. The generic-text concatenation control has lower mean wind MAE, while its attention counterpart has higher error (\autoref{tab:fusion_roles_24h}). The primary configuration was fixed before these comparisons and is not reselected from test scores. At 120\,h, generic-text concatenation has lower mean wind MAE than forecast-tuned concatenation for T+D (\autoref{tab:fusion_roles_per_horizon}). The I+D controls at that lead also favor generic text, but they differ in fusion operator or inference batch size, so they do not isolate the quality of the text backbone. Every displayed stochastic configuration uses five seeds. Configurations can still differ in learning rate and training batch size, and the recorded inference policies differ between some configurations. Forecast tuning and attention do not provide a uniform improvement across the tested configurations.

\begin{table}[H]
\centering
\caption{Sensitivity to the text backbone and fusion operator for separately trained lead-time models. Entries are mean $\pm$ sample SD across five seeds. MAE units are kt/hPa. Backbone definitions and training protocols are given in the text. Bold and underline indicate the best and second-best means within each lead.}
\label{tab:fusion_roles_per_horizon}
\small
\renewcommand{\arraystretch}{1.0}
\setlength{\tabcolsep}{4pt}
\begin{adjustbox}{max width=\linewidth}
\begin{tabular}{@{}lccccccc@{}}
\toprule
\multirow{2}{*}{Lead} & \multirow{2}{*}{Modality} & \multirow{2}{*}{Text backbone} & \multirow{2}{*}{Fusion} & \multicolumn{2}{c}{$R^2$} & \multicolumn{2}{c}{MAE (kt/hPa)} \\
\cmidrule(lr){5-6} \cmidrule(lr){7-8}
 &  &  &  & $V_{\max}$ & MSLP & $V_{\max}$ & MSLP \\
\midrule
\multirow{5}{*}{24\,h} & T+D & Generic & Concat & $0.53\pm0.01$ & $\underline{0.51}\pm0.02$ & $9.30\pm0.19$ & $\underline{6.23}\pm0.13$ \\
 & T+D & Forecast-tuned & Concat & $\underline{0.54}\pm0.04$ & $0.50\pm0.04$ & $\underline{9.26}\pm0.29$ & $6.28\pm0.27$ \\
 & T+D & Generic & Attention & $0.43\pm0.06$ & $0.43\pm0.05$ & $10.33\pm0.48$ & $6.74\pm0.30$ \\
\cmidrule(lr){2-8}
 & T+I+D & Generic & Concat & $0.52\pm0.02$ & $\underline{0.51}\pm0.01$ & $9.58\pm0.19$ & $6.25\pm0.07$ \\
 & T+I+D & Forecast-tuned & Concat & $\boldsymbol{0.55}\pm0.02$ & $\boldsymbol{0.52}\pm0.03$ & $\boldsymbol{9.15}\pm0.19$ & $\boldsymbol{6.12}\pm0.18$ \\
\midrule
\multirow{4}{*}{48\,h} & T+D & Generic & Concat & $0.52\pm0.02$ & $0.49\pm0.02$ & $14.56\pm0.36$ & $10.32\pm0.28$ \\
 & T+D & Forecast-tuned & Concat & $\boldsymbol{0.58}\pm0.02$ & $\boldsymbol{0.55}\pm0.02$ & $\boldsymbol{13.44}\pm0.23$ & $\boldsymbol{9.58}\pm0.30$ \\
\cmidrule(lr){2-8}
 & T+I+D & Generic & Concat & $0.51\pm0.01$ & $0.46\pm0.02$ & $14.79\pm0.12$ & $10.50\pm0.33$ \\
 & T+I+D & Forecast-tuned & Concat & $\underline{0.57}\pm0.01$ & $\underline{0.54}\pm0.02$ & $\underline{13.65}\pm0.22$ & $\underline{9.67}\pm0.27$ \\
\midrule
\multirow{4}{*}{72\,h} & T+D & Generic & Concat & $0.44\pm0.02$ & $0.36\pm0.04$ & $17.95\pm0.52$ & $13.33\pm0.47$ \\
 & T+D & Forecast-tuned & Concat & $\boldsymbol{0.46}\pm0.06$ & $\underline{0.39}\pm0.06$ & $\boldsymbol{16.93}\pm0.67$ & $\underline{12.50}\pm0.32$ \\
\cmidrule(lr){2-8}
 & T+I+D & Generic & Concat & $\boldsymbol{0.46}\pm0.03$ & $\boldsymbol{0.41}\pm0.03$ & $\underline{16.94}\pm0.36$ & $12.88\pm0.61$ \\
 & T+I+D & Forecast-tuned & Concat & $\underline{0.45}\pm0.04$ & $\underline{0.39}\pm0.06$ & $17.11\pm0.61$ & $\boldsymbol{12.44}\pm0.49$ \\
\midrule
\multirow{4}{*}{96\,h} & T+D & Generic & Concat & $\boldsymbol{0.31}\pm0.04$ & $\boldsymbol{0.23}\pm0.02$ & $\boldsymbol{19.25}\pm0.31$ & $14.72\pm0.40$ \\
 & T+D & Forecast-tuned & Concat & $0.24\pm0.07$ & $\underline{0.19}\pm0.09$ & $\underline{19.76}\pm1.22$ & $\underline{14.49}\pm0.80$ \\
\cmidrule(lr){2-8}
 & T+I+D & Generic & Concat & $\underline{0.28}\pm0.12$ & $\boldsymbol{0.23}\pm0.11$ & $\boldsymbol{19.25}\pm1.05$ & $\boldsymbol{14.36}\pm0.91$ \\
 & T+I+D & Forecast-tuned & Concat & $0.19\pm0.08$ & $0.18\pm0.08$ & $20.43\pm0.81$ & $\underline{14.49}\pm0.73$ \\
\midrule
\multirow{4}{*}{120\,h} & T+D & Generic & Concat & $\boldsymbol{0.22}\pm0.05$ & $\boldsymbol{0.12}\pm0.07$ & $\boldsymbol{20.27}\pm0.85$ & $\boldsymbol{15.71}\pm0.53$ \\
 & T+D & Forecast-tuned & Concat & $0.06\pm0.07$ & $0.06\pm0.07$ & $22.51\pm1.49$ & $16.72\pm1.10$ \\
\cmidrule(lr){2-8}
 & T+I+D & Generic & Concat & $\underline{0.17}\pm0.09$ & $\underline{0.09}\pm0.10$ & $\underline{20.66}\pm0.76$ & $\underline{16.32}\pm0.61$ \\
 & T+I+D & Forecast-tuned & Concat & $0.00\pm0.13$ & $0.01\pm0.10$ & $23.65\pm2.26$ & $17.24\pm1.03$ \\
\bottomrule
\end{tabular}
\end{adjustbox}
\par\smallskip
{\footnotesize Configurations within a modality and lead use the same valid-target population (Table~\ref{tab:neural_populations}). The primary and per-lead families are separate experiments. Configurations can differ in tuned learning rate and training batch size, so these comparisons do not isolate the effect of the backbone or fusion operator. These descriptive configuration comparisons do not establish a universally optimal backbone or fusion operator.}
\end{table}

Fusion type and candidate learning rates and batch sizes were compared using 2020--2021 validation loss with generic frozen text features. The 48\,h fusion choices were carried to separately trained 72--120\,h models, which were not independently swept for fusion type. Joint multi-lead models were evaluated with concatenation only. The comparison of text backbones keeps the primary fusion choices fixed and does not establish their optimality for forecast-tuned text. These experiments assess sensitivity to the text backbone and fusion settings rather than selecting replacement configurations from test scores.

\begin{table}[H]
\centering
\caption{Sensitivity to the text backbone and fusion operator (separately trained lead-time models). Generic uses pretrained RoBERTa. Forecast-tuned uses RoBERTa trained on discussions for the same seed and forecast lead. Both remain frozen during fusion training. T+I has no text backbone. Concat denotes concatenation and Attention denotes multi-head attention. Entries are mean $\pm$ sample SD across five training runs. MAE units are kt/hPa. Bold and underline mark the best and second-best displayed means per metric column within each lead (ties included).}
\label{tab:fusion_roles_alternative_leads}
\small
\renewcommand{\arraystretch}{1.12}
\setlength{\tabcolsep}{4pt}
\begin{adjustbox}{max width=\linewidth}
\begin{tabular}{@{}lcccccccc@{}}
\toprule
\multirow{2}{*}{Lead} & \multirow{2}{*}{Modality} & \multirow{2}{*}{Text backbone} & \multirow{2}{*}{Fusion} & \multirow{2}{*}{Seeds} & \multicolumn{2}{c}{$R^2$} & \multicolumn{2}{c}{MAE (kt/hPa)} \\
\cmidrule(lr){6-7} \cmidrule(lr){8-9}
 &  &  &  &  & $V_{\max}$ & MSLP & $V_{\max}$ & MSLP \\
\midrule
\multirow{5}{*}{24\,h} & T+I & No text & Concat & 5 & $\underline{0.46}\pm0.02$ & $\underline{0.46}\pm0.03$ & $\underline{10.09}\pm0.11$ & $\underline{6.57}\pm0.11$ \\
 & T+I & No text & Attention & 5 & $0.45\pm0.01$ & $0.44\pm0.03$ & $10.14\pm0.17$ & $6.70\pm0.08$ \\
\cmidrule(lr){2-9}
 & I+D & Generic & Attention & 5 & $0.35\pm0.03$ & $0.32\pm0.04$ & $11.38\pm0.43$ & $7.58\pm0.30$ \\
 & I+D & Forecast-tuned & Attention & 5 & $\boldsymbol{0.47}\pm0.03$ & $\boldsymbol{0.47}\pm0.03$ & $\boldsymbol{10.06}\pm0.39$ & $\boldsymbol{6.49}\pm0.26$ \\
 & I+D & Generic & Concat & 5 & $0.36\pm0.02$ & $0.33\pm0.02$ & $11.20\pm0.26$ & $7.44\pm0.22$ \\
\midrule
\multirow{4}{*}{48\,h} & T+I & No text & Attention & 5 & $\underline{0.45}\pm0.02$ & $\underline{0.42}\pm0.03$ & $\underline{15.49}\pm0.54$ & $\underline{10.80}\pm0.31$ \\
\cmidrule(lr){2-9}
 & I+D & Generic & Attention & 5 & $0.23\pm0.08$ & $0.23\pm0.07$ & $18.48\pm1.02$ & $12.72\pm0.69$ \\
 & I+D & Forecast-tuned & Attention & 5 & $\boldsymbol{0.52}\pm0.03$ & $\boldsymbol{0.49}\pm0.02$ & $\boldsymbol{14.38}\pm0.54$ & $\boldsymbol{10.11}\pm0.15$ \\
 & I+D & Generic & Concat & 5 & $0.35\pm0.05$ & $0.30\pm0.06$ & $17.12\pm0.54$ & $12.17\pm0.33$ \\
\midrule
\multirow{4}{*}{72\,h} & T+I & No text & Attention & 5 & $\boldsymbol{0.41}\pm0.03$ & $\underline{0.31}\pm0.10$ & $\underline{18.34}\pm0.44$ & $\underline{13.95}\pm0.42$ \\
\cmidrule(lr){2-9}
 & I+D & Generic & Attention & 5 & $0.19\pm0.03$ & $0.14\pm0.04$ & $21.30\pm0.66$ & $15.39\pm0.25$ \\
 & I+D & Forecast-tuned & Attention & 5 & $\underline{0.39}\pm0.10$ & $\boldsymbol{0.36}\pm0.05$ & $\boldsymbol{18.08}\pm1.48$ & $\boldsymbol{12.93}\pm0.59$ \\
 & I+D & Generic & Concat & 5 & $0.32\pm0.04$ & $0.23\pm0.08$ & $19.71\pm0.78$ & $14.62\pm0.64$ \\
\midrule
\multirow{4}{*}{96\,h} & T+I & No text & Attention & 5 & $\boldsymbol{0.28}\pm0.08$ & $\boldsymbol{0.20}\pm0.06$ & $\boldsymbol{19.55}\pm0.95$ & $\boldsymbol{14.82}\pm0.86$ \\
\cmidrule(lr){2-9}
 & I+D & Generic & Attention & 5 & $\underline{0.20}\pm0.09$ & $0.08\pm0.09$ & $\underline{20.72}\pm1.37$ & $15.62\pm0.89$ \\
 & I+D & Forecast-tuned & Attention & 5 & $0.08\pm0.19$ & $0.03\pm0.22$ & $21.62\pm2.66$ & $15.61\pm1.76$ \\
 & I+D & Generic & Concat & 5 & $0.11\pm0.19$ & $\underline{0.11}\pm0.05$ & $21.80\pm2.06$ & $\underline{15.47}\pm0.45$ \\
\midrule
\multirow{4}{*}{120\,h} & T+I & No text & Attention & 5 & $0.08\pm0.09$ & $\underline{0.05}\pm0.10$ & $\underline{21.28}\pm0.87$ & $\underline{16.64}\pm0.75$ \\
\cmidrule(lr){2-9}
 & I+D & Generic & Attention & 5 & $\underline{0.10}\pm0.16$ & $-0.12\pm0.19$ & $21.55\pm1.99$ & $17.74\pm1.34$ \\
 & I+D & Forecast-tuned & Attention & 5 & $0.05\pm0.18$ & $0.02\pm0.15$ & $23.13\pm2.57$ & $17.18\pm1.20$ \\
 & I+D & Generic & Concat & 5 & $\boldsymbol{0.12}\pm0.20$ & $\boldsymbol{0.07}\pm0.13$ & $\boldsymbol{21.12}\pm1.85$ & $\boldsymbol{16.19}\pm0.93$ \\
\bottomrule
\end{tabular}
\end{adjustbox}
\par\smallskip
{\footnotesize Configurations within a modality and lead use the same valid-target population (Table~\ref{tab:neural_populations}). The primary and per-lead families are separate experiments. Configurations can differ in tuned learning rate and training batch size, so these comparisons do not isolate the effect of the backbone or fusion operator. The T+I baselines use inference batch size 16 and their alternative attention control uses 32. The fresh generic I+D controls also use 32. Universal prediction parity across these batch sizes has not been established. These descriptive configuration comparisons do not establish a universally optimal backbone or fusion operator.}
\end{table}

\FloatBarrier
\section{Rubric, Pipeline and Validation for Subjectivity Classification}
\label{app:subjectivity}

This appendix documents the three-class rubric, the uncertain \textsc{balanced}/\textsc{factual} boundary, the classification pipeline, and corpus counts. The labels are \textsc{opinionated}, \textsc{balanced}, and \textsc{factual}. Claude Sonnet~4.6 and Claude Opus~4.6 \citep{anthropic2026sonnet46,anthropic2026opus46} supply reference labels under the rubric using an LLM-as-a-judge procedure \citep{zheng2023judging}. The compact Gemma-4-E2B-it instruction-tuned model \citep{google2025gemma} is evaluated as a candidate classifier with a three-example prompt \citep{brown2020language}. The reference labels depend on the rubric and on the judges that apply it. We use them to evaluate the classifier and do not treat them as a measure of forecaster intent.

\subsection{Class definitions and example phrases}
\label{app:subj_rubric}

\paragraph{\textsc{opinionated}.} The discussion contains at least one explicit first-person subjective statement. In the NHC corpus this manifests as a standalone first-person pronoun (``I'', ``my'', ``me'') introducing a stance, decision, preference, or admission. The following representative phrase families occur in the raw corpus. We verified each listed form.
\begin{itemize}
    \item \textbf{Opinion markers.} ``I think'', ``I believe'', ``I suspect'', ``my feeling is''.
    \item \textbf{Preferences.} ``I prefer'', ``I would prefer'', ``I am going to stay with''.
    \item \textbf{Decisions / commitments.} ``I have kept'', ``I have opted to'', ``I have decided''.
    \item \textbf{Uncertainty / admissions.} ``I am not confident'', ``I am not ready to'', ``I do not know''.
    \item \textbf{Disagreement with guidance.} ``it was hard for me to ignore'', ``I would rather'', ``I am not quite ready to believe''.
\end{itemize}

On the stratified \resultvalue{subjectivity/benchmark_n}-discussion benchmark labeled by Claude Sonnet~4.6 \citep{anthropic2026sonnet46}, every \textsc{opinionated} file contained a standalone first-person token, and every file with such a token was labeled \textsc{opinionated}. The regex therefore reproduces this benchmark label without errors. We apply it to the full corpus as a fixed marker of explicit first-person framing. This agreement does not establish a universal NHC writing rule or measure judgment expressed without first-person language.

\paragraph{\textsc{balanced} (most common).} No first-person markers, but the forecaster provides sustained interpretive reasoning. Typical signals include the following. Quoted wording in this list and the factual list below is illustrative, not a verbatim corpus quotation.
\begin{itemize}
    \item Weighing competing numerical weather prediction model clusters (e.g.\ ``the GFS and UKMET differ in the outer periods; the ECMWF is closer to observations'').
    \item Explaining physical mechanisms across multiple sentences (shear evolution, sea surface temperature gradient, baroclinic energy, dry-air entrainment, steering-flow transitions).
    \item Raising conditional scenarios (``if the cyclone stays over warm water'').
    \item Discussing forecast uncertainty with competing evidence.
    \item Typically a multi-paragraph narrative containing more than one reasoning chain.
\end{itemize}

\paragraph{\textsc{factual} (least common).} No first-person markers, and minimal sustained reasoning. Typical triggers include the following.
\begin{itemize}
    \item Short bulletins (typically fewer than five sentences of narrative).
    \item Final / dissipation / extratropical-transition discussions that do not add new analysis.
    \item Initial / upgrade discussions that simply state observations and intensity.
    \item Discussions deferring to prior reasoning (``forecast is a compromise of guidance'', ``no change from previous'').
    \item Single-mechanism restatements without elaboration (e.g.\ ``cool water plus shear will continue the weakening'' and nothing else).
    \item Post-landfall inland updates that merely recite observed winds and rainfall.
\end{itemize}

\subsection{Worked examples}
\label{app:subj_examples}

The excerpts below come from three discussions, one per class, that received the same label from both LLM judges in the comparison described in Appendix~\ref{app:subj_fuzzy}. Capitalization and punctuation are adapted for readability, omitted passages are marked, and bold type is added for emphasis. These excerpts are distinct from the model-input cleaning described in Appendix~\ref{app:subj_parity}.

\paragraph{\textsc{opinionated}: Tropical Storm Earl, Discussion 13, 5 August 2016.}
\begin{quote}\small
``This was also confirmed by this morning's radiosondes from the area. \textbf{I prefer to keep the initial intensity generously at 35 kt} at this time, and then adjust it, if necessary, when an Air Force reconnaissance plane checks the system later today.''
\end{quote}
The forecaster writes ``I prefer to keep'', an explicit first-person statement of preference, so the discussion is \textsc{opinionated}. Both judges and the regex pre-filter assign this label.

\paragraph{\textsc{balanced}: Hurricane Frances, Discussion 42, 4 September 2004.}
\begin{quote}\small
``Environmental westerly shear is still present, although the inner core convection has perhaps become a little more symmetric. Global models continue to forecast a lessening of this shear over the next 24 hours, and so there will be an opportunity for a little strengthening before landfall, particularly if the forward speed increases a bit. [\ldots] A mid-level high is located over southeast Georgia. This is expected to nudge Frances back to the west-northwest prior to landfall and across the Florida peninsula. Most of the model guidance has shifted a little to the left after 24 hours in response to stronger ridging over the southeastern states.''
\end{quote}
The discussion contains no first-person language, which rules out \textsc{opinionated}. Over several sentences, the forecaster discusses three physical drivers (shear, ridging, steering) and explicitly weighs the model guidance. This sustained interpretive reasoning makes the discussion \textsc{balanced}.

\paragraph{\textsc{factual}: Tropical Depression Mindy, Discussion 14, 14 October 2003, 0300 UTC.}
\begin{quote}\small
``Satellite imagery this evening indicates Mindy is a mere swirl of low clouds with the exposed center becoming increasingly separated from the deep frontal-looking convection to the east. Mindy is degenerating into a non-convective remnant low and this will be the last advisory unless regeneration occurs. Given the strong westerly shear and dry environment, this seems rather unlikely.''
\end{quote}
This short final discussion has no first-person language, and its reasoning is minimal. It briefly notes that shear and dry air make regeneration unlikely. The judges classified it as \textsc{factual}.

\subsection{The fuzzy boundary and inter-rater reliability}
\label{app:subj_fuzzy}

The \textsc{balanced}/\textsc{factual} boundary is subjective and is the main source of disagreement between the judges. We assessed this by comparing labels for a stratified sample of \resultvalue{subjectivity/audit_n} files drawn from a \resultvalue{subjectivity/benchmark_n}-file benchmark. Two LLM judges from the Anthropic Claude family \citep{anthropic2026sonnet46,anthropic2026opus46} (Claude Sonnet~4.6 and Claude Opus~4.6) used the same rubric without seeing each other's labels, using an LLM-as-a-judge approach~\citep{zheng2023judging}. Agreement by class is as follows.
\begin{itemize}
    \item \textsc{opinionated}. $\resultvalue{subjectivity/audit_opinionated_agree_n}/\resultvalue{subjectivity/audit_opinionated_n}$ files agreed.
    \item \textsc{balanced}. $\resultvalue{subjectivity/audit_balanced_agree_n}/\resultvalue{subjectivity/audit_balanced_n}$ files agreed.
    \item \textsc{factual}. $\resultvalue{subjectivity/audit_factual_agree_n}/\resultvalue{subjectivity/audit_factual_n}$ files agreed. The disagreements were reclassified by the second judge as \textsc{balanced}.
\end{itemize}
Overall agreement on the stratified audit was $\resultvalue{subjectivity/audit_agreement_n}/\resultvalue{subjectivity/audit_n}$, with disagreements confined to discussions labeled \textsc{factual} by the reference judge. Differences in the class frequencies of separate labeling chunks cannot establish annotator consistency because the chunks contain different discussions.

The disagreement suggests that reasoning depth varies continuously across the \textsc{balanced}/\textsc{factual} boundary. It does not establish an accuracy ceiling for future classifiers or human annotators. Both judges agreed on all audited \textsc{opinionated} and \textsc{balanced} files, but these are small stratified samples. We report agreement with the stated reference labels, especially for the less stable \textsc{factual} class, rather than treating any judge as an objective measure of forecaster intent.

\subsection{Classification procedure and agreement}
\label{app:subj_pipeline}

We use a two-stage hybrid pipeline.
\begin{itemize}
    \item \emph{Stage 1 (regex pre-filter for \textsc{opinionated}).} Scan the cleaned discussion text for standalone first-person tokens with a lookaround regex equivalent to \verb!(?<![A-Z/])(I|MY|ME)(?![A-Z.])! under case-insensitive matching. The lookbehind excludes a preceding letter or slash (so that ``SSM/I'' does not match), and the lookahead excludes a trailing letter or period (so that ``i.e.'' and longer words do not match). Any match yields the label \textsc{opinionated}. Sweeping the full corpus yields \resultvalue{subjectivity/regex_hits} matches among \resultvalue{subjectivity/corpus_n} discussions. On the Sonnet-labeled benchmark it recovers every \textsc{opinionated} reference label with no false positives.
    \item \emph{Stage 2 (three-way classifier for the remaining discussions).} For files not flagged by the Stage 1 regex, Gemma-4-E2B-it \citep{google2025gemma} assigns a rubric label using a three-example prompt \citep{brown2020language}. The regex pre-filter preserves recall on Sonnet-flagged \textsc{opinionated} files by construction. The hybrid agrees with \resultvalue{subjectivity/hybrid_exact_n} of \resultvalue{subjectivity/benchmark_n} reference labels, or \resultvalue{subjectivity/hybrid_exact_percent}\%. Agreement is weaker for \textsc{factual}, whose reference boundary is ambiguous.
\end{itemize}
Agreement with one LLM judge is not independent human validation. The separate comparison measures agreement between two related models on a small sample, rather than a confidence band for classifier accuracy. The subjectivity labels remain descriptive of this corpus. Forecasting models ingest the cleaned text directly and do not use these discrete labels.

\subsection{Consistency with forecasting text preprocessing}
\label{app:subj_parity}

The subjectivity classifier and forecasting data loader apply matching content-cleaning rules before model-specific tokenization and truncation. The classifier preserves letter case for LLM prompting, while the neural pipeline lowercases its input. Both apply the following steps.
\begin{itemize}
    \item Bulletin headers are trimmed to the first datetime line.
    \item The \texttt{FORECAST POSITIONS AND MAX WINDS} table is stripped to exclude the structured official forecast table. Narrative forecasts and current-intensity statements remain.
    \item Whitespace is collapsed.
    \item Non-ASCII characters and control characters are removed.
\end{itemize}
The classifier and forecasting loader are intended to apply the same content-cleaning rules apart from letter case. A comparison of their cleaned outputs checks this consistency when the source discussions are available. This checks cleaning of the full discussion, not equality of the retained model inputs. The forecasting encoders subsequently tokenize and truncate the text to 512 tokens for RoBERTa, so a full-discussion rhetorical label need not describe the visible token window. Its triggering phrase may lie outside that window.

\subsection{Corpus-wide counts and interpretation}
\label{app:subj_counts}

The archived list of discussions flagged by the first-person regex contains \resultvalue{subjectivity/regex_hits} discussions out of \resultvalue{subjectivity/corpus_n}, or \resultvalue{subjectivity/regex_hits_percent}\%. This count describes the discussions matched to cases rather than a sample of discussions. The stratified benchmark was drawn from an earlier matched corpus of 11{,}633 discussions and was not designed to reproduce the corpus class distribution.

The subjectivity labels describe the discussion corpus and are not inputs to the forecasting models. The first-person rule is deterministic, but agreement between language models does not establish human validity or usefulness for forecasting. Using either the binary indicator or the three-way label as a model feature would require a separate evaluation.

\FloatBarrier
\section{Integrated gradients for discussion text}
\label{app:interpretability}

This appendix supports Section~\ref{sec:interpretability}. We document the Integrated Gradients (IG) pipeline, numerical checks, and a fixed set of eleven illustrative test cases. The three worked examples are Tropical Storm Ian Discussion~10 (discussion-only D), Hurricane Agatha Discussion~9 (T+D), and Tropical Storm Madeline Discussion~7 (T+D). Agatha (EP01 2022) and Madeline (EP14 2022) are eastern North Pacific storms from the same NHC discussion corpus. Their T$\to$T+D gains have opposite signs, as shown in the case table below.

\subsection{Scope of the attribution analysis}
\label{app:ig_scope}

IG characterizes a single checkpoint's response to a particular discussion relative to a specified embedding baseline. Population-level evidence for the predictive value of discussion text instead comes from the paired comparisons in Appendix~\ref{app:subj_stratified_gain}. The eleven illustrative cases were selected during exploratory analysis. The case table reports their predictions and ranks by improvement in absolute error. Five additional fixed cases illustrate gains and losses, including two cases flagged by the regex for first-person language. These added cases were selected after inspecting test predictions and are explicitly descriptive. Neither set is a representative sample or an independent test of whether a particular forecaster phrase improves prediction. Token-level attributions also do not aggregate directly across documents with different lengths, vocabularies, and token frequencies.

\subsection{Attribution method}
\label{app:ig_method}

We attribute the standardized wind change output to the input discussion embeddings. For T+D, the tabular representation is evaluated once and held fixed along the IG path. Its text encoder was first trained in the corresponding D model and then frozen during fusion training. The projection, modality normalization, and fusion layers are those of the evaluated T+D checkpoint. For D, the encoder is the trained discussion-only encoder. Before attribution, the evaluated model must reproduce its saved seed-1 prediction within $0.005$\,kt.

The baseline replaces every input token embedding, including special-token positions, with the pad-token embedding while retaining the original attention mask. It is a specified reference in the embedding space, not a literal absence of text. Integration starts from a fixed step count for each map, 128 for most, and the resolution is doubled until the completeness criterion below is met or \resultvalue{xai/step_cap} steps are reached. Maps that remain above the tolerance at that limit are marked as such. The limit was raised from 2048 steps after a first run, with the tolerance unchanged. Attributions are in standardized target units. Positive values increase predicted intensity change relative to the baseline. Predictions in captions and prose are converted to knots with the checkpoint's training-split statistics. As in training, discussions are truncated to 512 tokens. Maps display the retained input, which can end before the full discussion. Map headers label each discussion by its advisory number, which is the same number, and give the initial intensity as the current wind. Subtokens are aggregated using fast-tokenizer word identifiers and character offsets. Phrase scores sum the scores of constituent words and repeated occurrences. Overlapping phrases therefore must not be summed as independent contributions. Maps use seed~1, whereas population comparisons and the case table use predictions averaged over five seeds.

\subsection{Completeness check}
\label{app:ig_completeness}

We check the IG identity $\sum_i\mathrm{IG}_i(x)=f(x)-f(x')$ numerically. A map passes only if the absolute residual is at most $0.01+0.01|f(x)-f(x')|$ standardized units and the forward-prediction check also passes. Of the \resultvalue{xai/maps} maps, \resultvalue{xai/completeness_verified} meet this criterion. The other \resultvalue{xai/above_tolerance}, all from 24\,h models, remain above the tolerance at \resultvalue{xai/step_cap} steps, with residuals between \resultvalue{xai/above_tolerance_residual_min_kt} and \resultvalue{xai/above_tolerance_residual_max_kt}\,kt in physical units. They include the Agatha map below, which states its residual. Integration steps and residuals are recorded alongside the attribution maps. The absolute component of the tolerance matters when the input-minus-baseline output is small, as it is for these maps. Completeness is evaluated over all token positions, including special-token scores omitted from the displayed word maps. The sum of visible words therefore need not equal the full attribution sum. Numerical completeness supports attribution accounting for this baseline and checkpoint. It does not establish causal or meteorological validity of individual phrases.

Sentence-final periods can receive large attributions in these maps. \citet{clark2019what} found attention heads focused on punctuation and delimiter tokens in BERT, but that observation does not explain the IG scores of our RoBERTa checkpoints or establish the importance of a whole sentence. Punctuation-only words are excluded from ranked lists, while full-text maps retain their raw scores. Phrase panels display the fifteen eligible trigrams with largest absolute scores after punctuation, stopword and short-token filtering, ordered by signed score. A panel need not contain equal numbers of positive and negative phrases.

\subsection{Worked examples}
\label{app:ig_examples}

Ian illustrates an accurate discussion-only prediction on an observed \resultvalue{case/IAN/truth_delta}\,kt 24\,h intensification target. Agatha and Madeline are test cases flagged by the first-person \textsc{opinionated} regex in Appendix~\ref{app:subj_pipeline}. That flag is a textual heuristic, not a measure of forecaster skill. They illustrate opposite signs of ensemble gain. The case table reports their predictions and gain ranks.

\autoref{tab:ig_cases_tabular} separates the predictions from the ensemble of five models from the seed-1 predictions being explained. Ian's D checkpoint approaches the observed intensification. The Madeline T+D checkpoint predicts weakening and the Agatha T+D checkpoint predicts almost no change, while their ensemble gains relative to T have opposite signs. These are comparisons between separately trained models and estimators, not a decomposition of the ensemble forecast into modality contributions.

\begin{figure}[H]
  \centering
  \includegraphics[width=\linewidth,height=0.88\textheight,keepaspectratio]{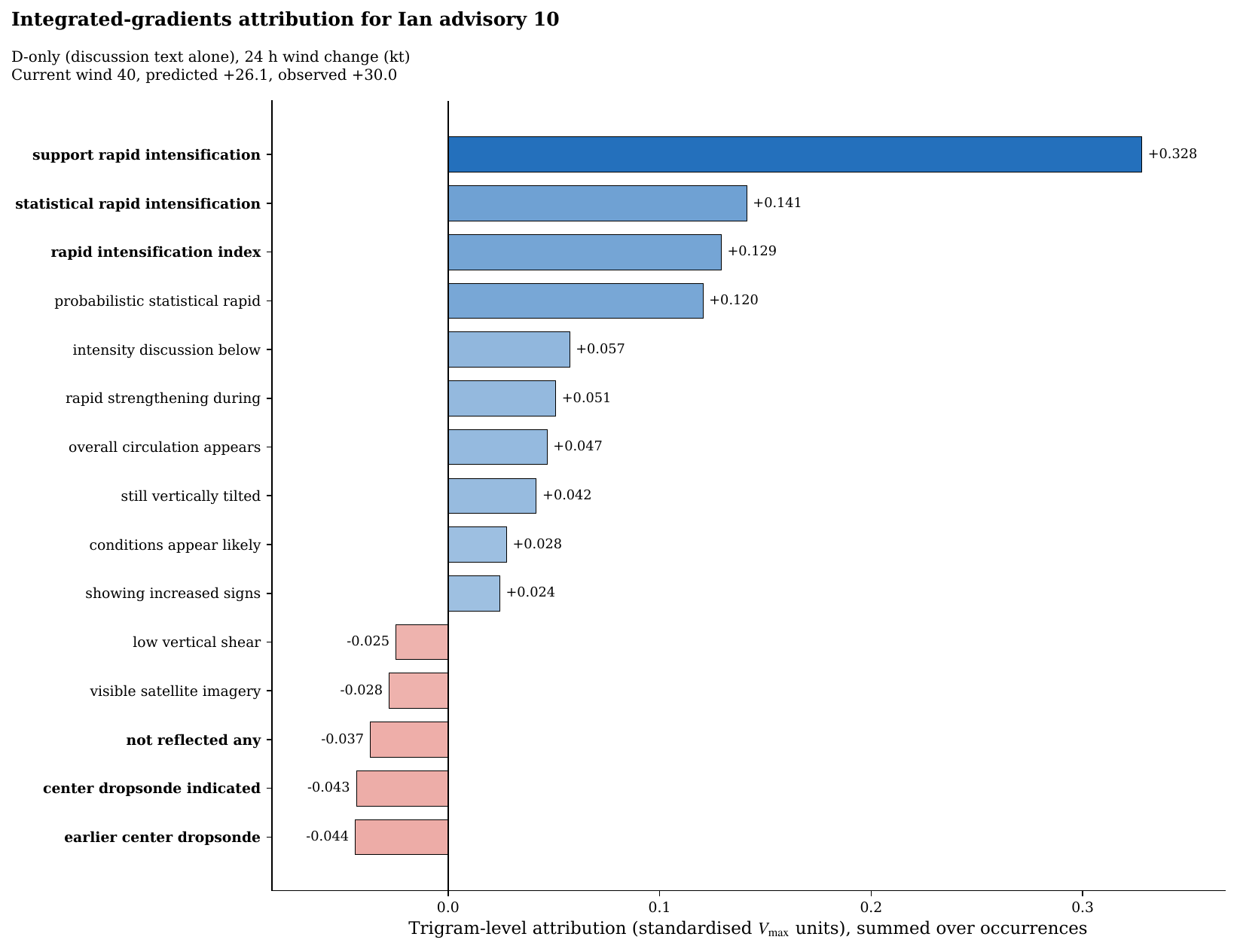}
  \caption{Fifteen eligible trigrams with the largest absolute attribution for Ian Discussion 10 using D seed~1, ordered by signed score. Overlapping rapid intensification phrases dominate the positive side. Scores are conditional on the specified checkpoint and baseline, and overlapping windows are not independent contributions.}
  \label{fig:ig_ian10_phrases}
\end{figure}

\begin{figure}[H]
  \centering
  \includegraphics[width=\linewidth,height=0.88\textheight,keepaspectratio]{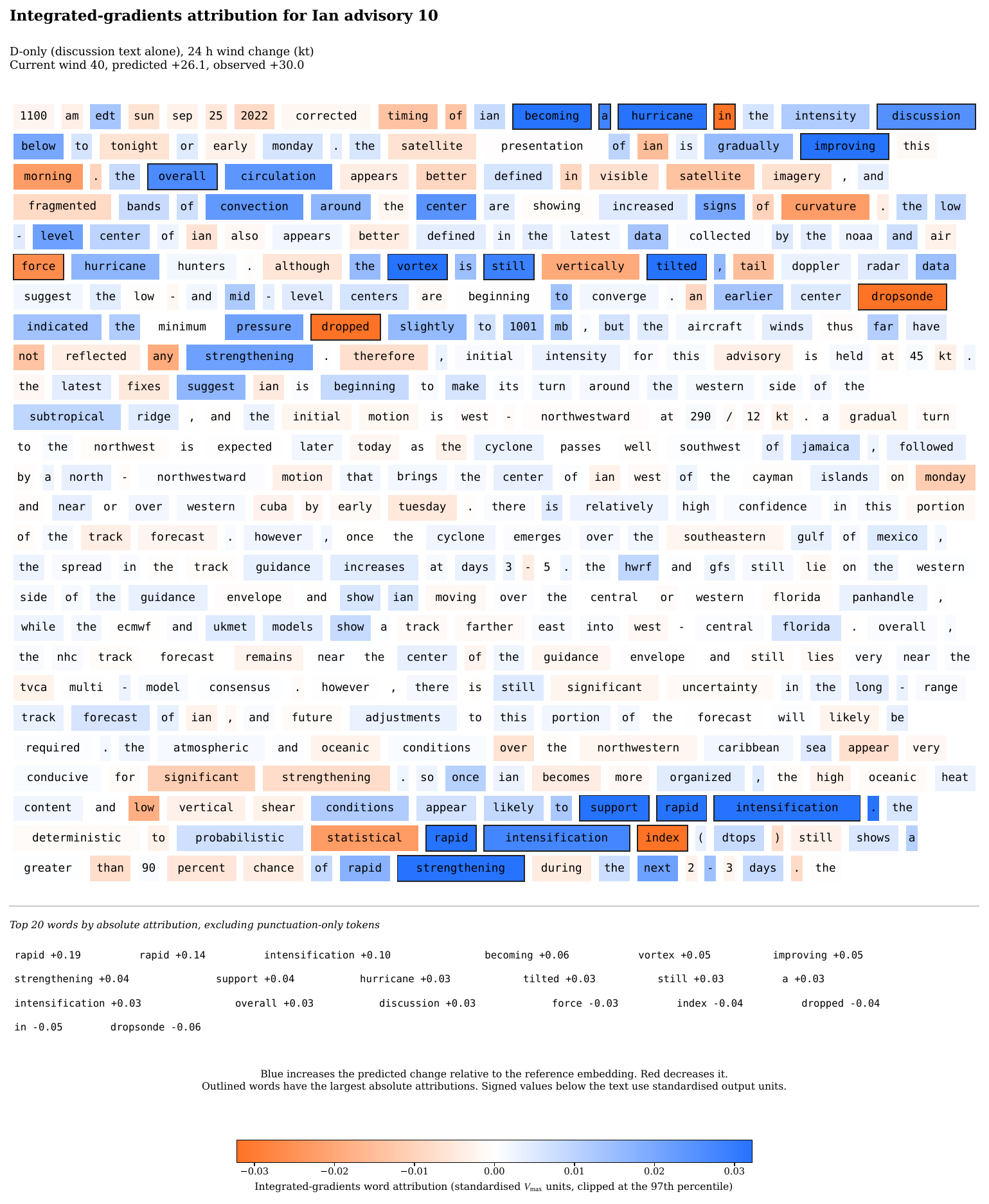}
  \caption{Attribution over the retained discussion input from Ian Discussion 10 using D seed~1 (up to 512 tokens). Blue increases predicted intensity change relative to the pad-embedding baseline. Red decreases it. The scale is clipped at the 97th percentile of absolute attribution. The prediction is $\resultvalue{case/IAN/d_seed1}$\,kt, against $\resultvalue{case/IAN/truth_delta}$ observed.}
  \label{fig:ig_ian10_fulltext}
\end{figure}

\emph{Tropical Storm Ian, Discussion 10 (D model, initial intensity \resultvalue{case/IAN/current_vmax}\,kt).} The discussion-only checkpoint predicts \resultvalue{case/IAN/d_seed1}\,kt on a \resultvalue{case/IAN/truth_delta}\,kt target. The positive trigrams in \autoref{fig:ig_ian10_phrases}, including ``support rapid intensification'' and ``statistical rapid intensification'', overlap the passage citing rapid intensification guidance. The largest negative trigrams refer to an earlier center dropsonde. The full-text map in \autoref{fig:ig_ian10_fulltext} shows where these scores occur in the discussion. This example is consistent with the checkpoint using forecast-relevant language, without identifying the causal contribution of that language or demonstrating general RI skill.

\begin{table}[H]
\centering
\small
\caption{Tabular context and predictions for the three worked examples. The upper block gives raw SHIPS values. CSST is in tenths of degrees Celsius (297 means $29.7^\circ$C). The lower block gives observed 24\,h intensity change and predictions from T and T+D ensembles of five models. The IG row reports the individual model being explained: D seed 1 for Ian and T+D seed 1 for Agatha and Madeline. All predicted intensity changes are in kt.}
\label{tab:ig_cases_tabular}
\setlength{\tabcolsep}{9pt}
\begin{tabular*}{\linewidth}{@{\extracolsep{\fill}}lrrr@{}}
\toprule
 & Ian disc.\ 10 & Agatha disc.\ 9 & Madeline disc.\ 7 \\
\midrule
VMAX & $40$ & $95$ & $50$ \\
MSLP & $1003$ & $964$ & $993$ \\
CSST & $290$ & $297$ & $290$ \\
SHGC & $168$ & $139$ & $260$ \\
RHMD & $71$ & $70$ & $81$ \\
TWAC & $76$ & $88$ & $90$ \\
\midrule
$\Delta V_{\rm obs}$ & $+30.0$ & $-25.0$ & $-15.0$ \\
$\hat\Delta V_{\rm T}$ & $+14.1$ & $+2.5$ & $-7.8$ \\
$\hat\Delta V_{\rm T+D}$ & $+31.2$ & $-8.4$ & $-3.4$ \\
$\hat\Delta V_{\rm IG}$ & $+26.1$ & $-0.5$ & $-7.1$ \\
\bottomrule
\end{tabular*}
\end{table}

\clearpage

\begin{figure}[H]
  \centering
  \includegraphics[width=\linewidth,height=0.88\textheight,keepaspectratio]{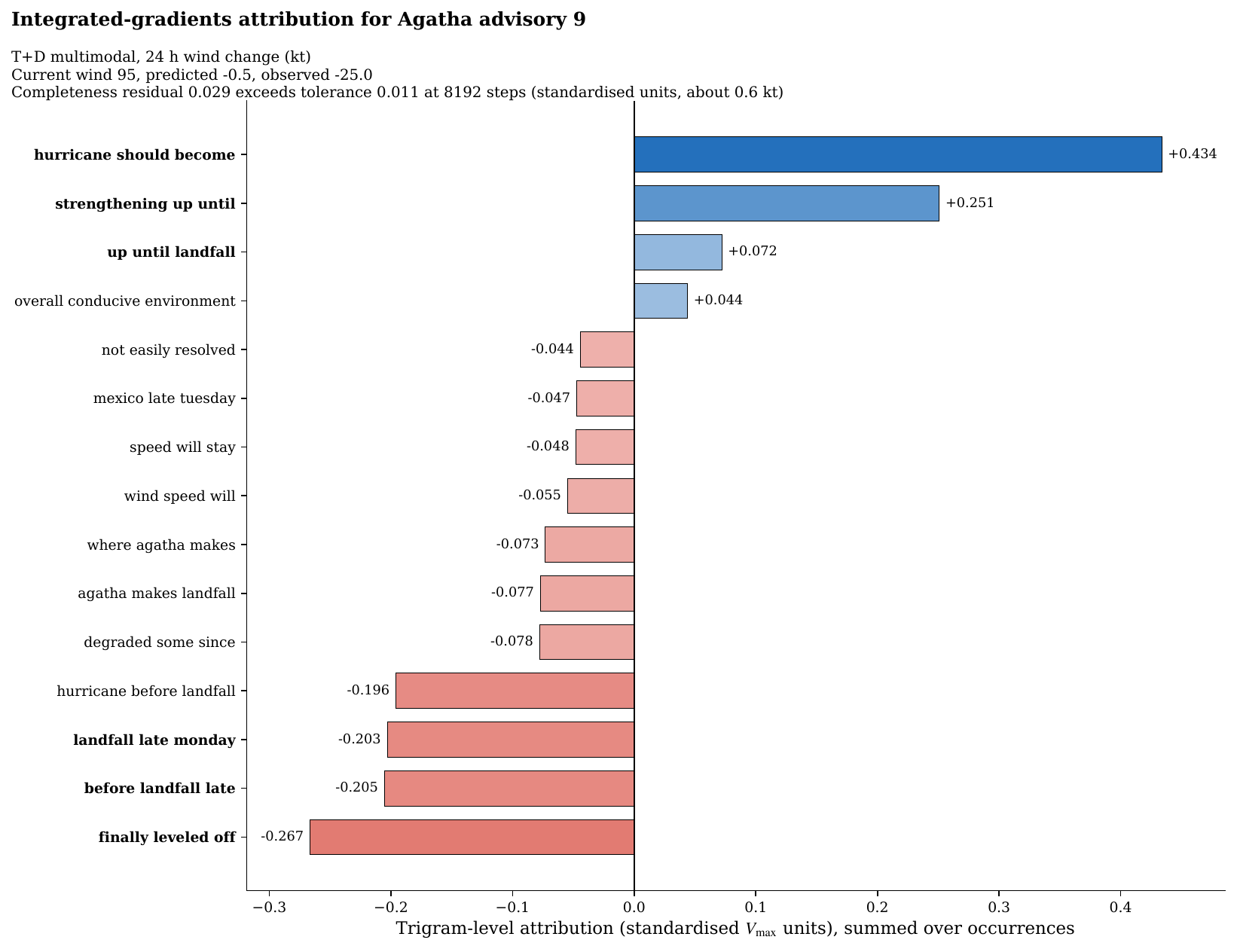}
  \caption{Agatha Discussion 9, T+D seed~1. The checkpoint predicts $\resultvalue{case/AGATHA/td_seed1}$\,kt against $\resultvalue{case/AGATHA/truth_delta}$ observed. Strengthening language contributes positively relative to the reference embedding, while several landfall-related windows contribute negatively. The completeness residual of this map exceeds the tolerance, as its header states. These local scores do not establish a causal effect of the forecaster's explicit judgment.}
  \label{fig:ig_agatha9_phrases}
\end{figure}

\emph{Hurricane Agatha, Discussion 9 (T+D, initial intensity \resultvalue{case/AGATHA/current_vmax}\,kt).} The discussion includes a first-person override of weakening numerical guidance, but it also anticipates landfall and rapid weakening afterward. The T+D checkpoint predicts \resultvalue{case/AGATHA/td_seed1}\,kt against \resultvalue{case/AGATHA/truth_delta}\,kt observed. The strongest positive trigrams in \autoref{fig:ig_agatha9_phrases} are ``hurricane should become'' and ``strengthening up until''. Negative trigrams include ``finally leveled off'', ``before landfall late'', and ``agatha makes landfall''. The ranked mixed-phrase data also contain negative attribution for ``rapid weakening''. The ensemble of five models moves toward weakening and gains \resultvalue{case/AGATHA/gain}\,kt relative to T, while the attributed checkpoint predicts almost no change. This map remains above the completeness tolerance (Appendix~\ref{app:ig_completeness}), so its scores carry that additional caveat. These attributions do not isolate the effect of the forecaster's first-person override from the discussion's landfall and timing information. Context-dependent signs, such as negative attribution for ``very warm waters'', further caution against reading these scores as physical rules.

\begin{figure}[H]
  \centering
  \includegraphics[width=\linewidth,height=0.88\textheight,keepaspectratio]{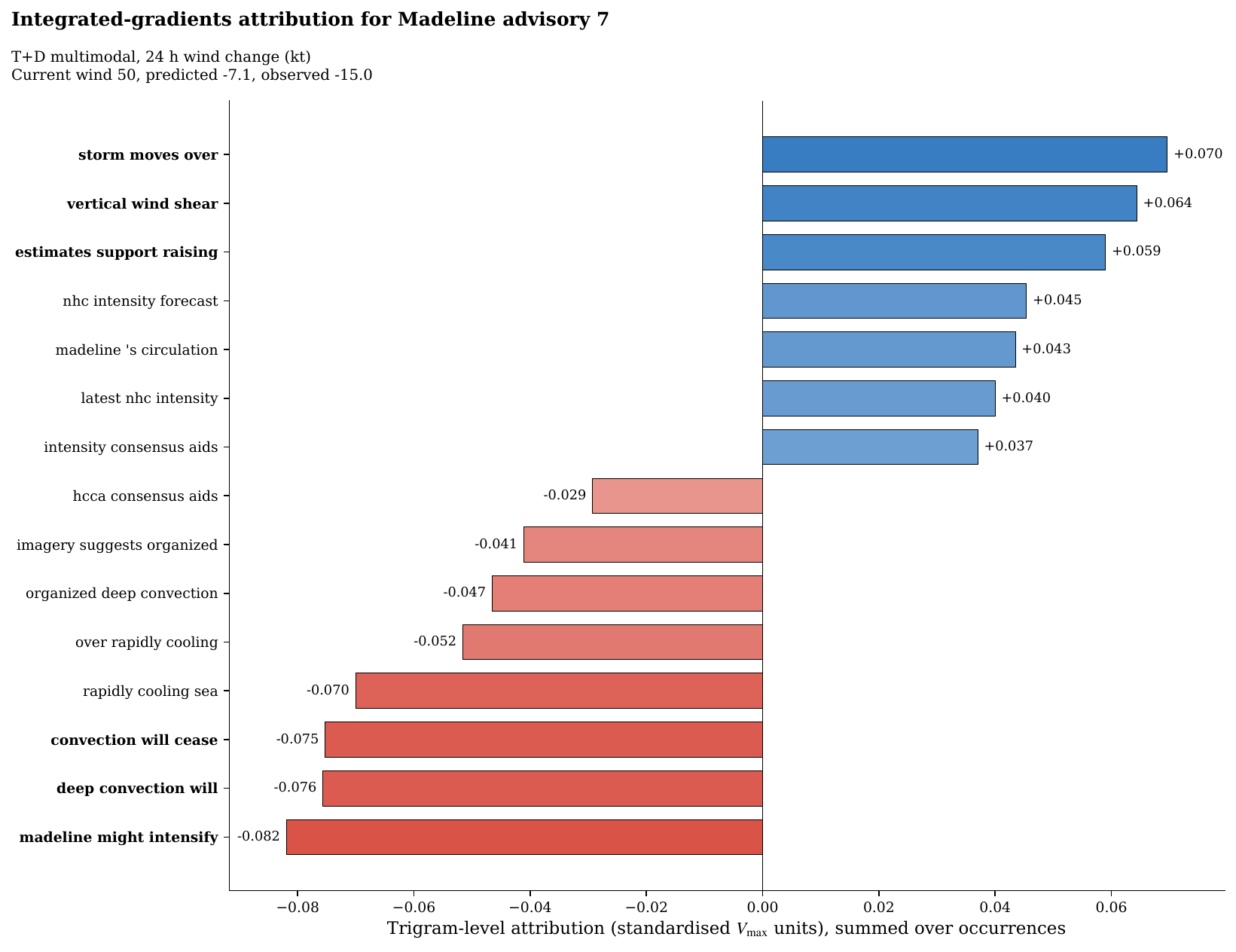}
  \caption{Madeline Discussion 7, T+D seed~1. Eight of the fifteen displayed trigrams have negative attribution, including possible brief intensification, loss of convection, and cooling-water language. The checkpoint predicts $\resultvalue{case/MADELINE/td_seed1}$\,kt, compared with $\resultvalue{case/MADELINE/truth_delta}$ observed.}
  \label{fig:ig_madeline7_phrases}
\end{figure}

\emph{Tropical Storm Madeline, Discussion 7 (T+D, initial intensity \resultvalue{case/MADELINE/current_vmax}\,kt).} The discussion allows possible brief intensification but describes increasing shear, cooling sea-surface temperatures, and eventual loss of deep convection. The checkpoint predicts \resultvalue{case/MADELINE/td_seed1}\,kt against \resultvalue{case/MADELINE/truth_delta}\,kt observed. In \autoref{fig:ig_madeline7_phrases}, ``madeline might intensify'', ``deep convection will'' and ``convection will cease'' have the strongest negative scores. Cooling-water phrases are also negative. ``Vertical wind shear'' has a positive score in this particular context, illustrating why a phrase's attribution sign is not a general physical relationship. The ensemble underpredicts weakening and performs worse than T for this case, as shown in \autoref{tab:ig_cases_tabular}. The displayed attribution does not establish that positive current-state descriptors outweigh the hedge, nor that this single-checkpoint explanation accounts for the ensemble error.

The three cases illustrate checkpoint sensitivity to language about forecasts, timing and current conditions. They support local inspection of model behavior, while the predictive value of the discussion modality and any benefit of explicit forecaster judgment must be assessed separately through controlled comparisons.

\subsection{Attribution gallery across cases and lead times}
\label{app:ig_gallery}

\autoref{tab:ig_gallery_cases} presents predictions and gain ranks for the eleven illustrative cases using the T+D ensemble. Calvin is a case with a positive gain, Darby is a case with a negative gain, and Fiona has gain rank \resultvalue{case/FIONA/gain_rank} among \resultvalue{primary/n_obs} cases. All maps satisfy the forward-prediction criterion, and \resultvalue{xai/completeness_verified} of \resultvalue{xai/maps} satisfy the completeness criterion (Appendix~\ref{app:ig_completeness}). Map headers show predictions from seed 1, while the table reports ensemble means. Figures~\ref{fig:gallery_lee24}--\ref{fig:gallery_calvin24} show the full-text maps for Lee, Otis and Calvin, with the checkpoint and its prediction given in each caption.

\begin{table}[!htbp]
\centering
\small
\caption[Attribution case roster]{Eleven illustrative cases selected during exploratory analysis. Predictions are means of five seeds from the 24\,h ablation family. Gain is the T absolute error minus the T+D absolute error (kt). Rank is descending gain among all 1,150 test rows, and is descriptive, not used to reselect these historical cases. Maps use seed 1, whose prediction may differ from the ensemble. Initialization times precede matched discussion issuance by three hours for these cases. All displayed values are in kt.}
\label{tab:ig_gallery_cases}
\setlength{\tabcolsep}{4pt}
\begin{tabular*}{\textwidth}{@{\extracolsep{\fill}}llrrrrrr@{}}
\toprule
Case & Init (UTC) & $V(t)$ & $\Delta V$ & $\hat\Delta V_{\rm T}$ & $\hat\Delta V_{\rm T+D}$ & Gain & Rank \\
\midrule
Dora disc.\ 7 & 23-08-02 06Z & 80 & $+35.0$ & $-5.0$ & $+33.2$ & $+38.2$ & 3 \\
Otis disc.\ 8 & 23-10-24 06Z & 55 & $+90.0$ & $+4.7$ & $+24.4$ & $+19.7$ & 25 \\
Ian disc.\ 10 & 22-09-25 12Z & 40 & $+30.0$ & $+14.1$ & $+31.2$ & $+14.7$ & 62 \\
Agatha disc.\ 9 & 22-05-30 00Z & 95 & $-25.0$ & $+2.5$ & $-8.4$ & $+10.9$ & 104 \\
Calvin disc.\ 20 & 23-07-16 12Z & 65 & $-20.0$ & $-10.9$ & $-21.3$ & $+7.7$ & 186 \\
Idalia disc.\ 3 & 23-08-27 06Z & 30 & $+25.0$ & $+10.1$ & $+14.3$ & $+4.2$ & 334 \\
Fiona disc.\ 24 & 22-09-20 06Z & 100 & $+20.0$ & $+2.9$ & $+4.5$ & $+1.7$ & 505 \\
Philippe disc.\ 35 & 23-10-02 00Z & 45 & $+0.0$ & $+0.8$ & $+3.0$ & $-2.2$ & 778 \\
Lee disc.\ 9 & 23-09-07 12Z & 85 & $+50.0$ & $+27.7$ & $+25.3$ & $-2.4$ & 792 \\
Madeline disc.\ 7 & 22-09-19 06Z & 50 & $-15.0$ & $-7.8$ & $-3.4$ & $-4.4$ & 914 \\
Darby disc.\ 24 & 22-07-15 06Z & 65 & $-25.0$ & $-26.5$ & $-37.8$ & $-11.3$ & 1093 \\
\bottomrule
\end{tabular*}
\end{table}

\emph{Large gains, losses and a case with a modest gain.} \autoref{tab:ig_additional_cases} adds further T+D cases at 24\,h. Julia and Agatha show large positive ensemble gains, while Darby and Idalia show large losses. Agatha Discussion 5, flagged by the first-person regex like Darby Discussion 10, also loses accuracy. Idalia Discussion 3 (\autoref{tab:ig_gallery_cases}) has a modest gain. These fixed cases were selected after inspecting predictions and are not representative of the test population. The colored maps explain seed~1, while the prediction comparisons and ranks use ensembles of five models. Raw discussions, cleaned input, predictions and convergence records accompany the supplement. The number of integration steps needed varies across maps and is recorded with each one.

\begin{table}[!htbp]
\centering
\small
\caption{Five fixed post-hoc 24\,h illustrations of gains and losses from adding discussion text. The cases were selected from the test period and are not independent confirmatory evidence. Predictions use ensembles of five models. Rank is descending T-to-T+D gain in wind MAE among all 1,150 test cases. IG uses the T+D seed-1 checkpoint, shown separately from the ensemble. Values are intensity changes or error gains in kt. Each case is fixed by storm and synoptic time. The Darby case was selected with Discussion 8 and is matched to Discussion 10 under the corrected discussion times. Ranks describe the current comparison, not the original selection. All five discussions are issued three hours after initialization.}
\label{tab:ig_additional_cases}
\setlength{\tabcolsep}{4pt}
\begin{tabular*}{\textwidth}{@{\extracolsep{\fill}}lrrrrrr@{}}
\toprule
Case & $\Delta V$ & $\hat\Delta V_{\rm T}$ & $\hat\Delta V_{\rm T+D}$ & T+D seed 1 & Gain & Rank \\
\midrule
Julia disc.\ 12 & $-40.0$ & $+19.0$ & $-36.5$ & $-39.0$ & $+55.6$ & 1 \\
Agatha disc.\ 10 & $-40.0$ & $+0.7$ & $-31.4$ & $-27.8$ & $+32.1$ & 7 \\
Darby disc.\ 10 & $-15.0$ & $-14.1$ & $+2.1$ & $+6.9$ & $-16.1$ & 1130 \\
Idalia disc.\ 14 & $-40.0$ & $-20.9$ & $+9.3$ & $+14.8$ & $-30.2$ & 1147 \\
Agatha disc.\ 5 & $+40.0$ & $+25.7$ & $-5.8$ & $-16.3$ & $-31.5$ & 1149 \\
\bottomrule
\end{tabular*}
\end{table}

\emph{Large intensification targets.} The ensemble that includes discussions approaches the observed intensification for Dora more closely than the tabular ensemble. For Lee, both ensembles predict about half of the observed change and adding discussion slightly increases the error. Otis remains a substantial miss for both, despite a positive gain from discussion. The Otis discussion anticipates strengthening before landfall, cites a chance of rapid strengthening during the next 24 hours, and places weakening after landfall. Its T+D checkpoint attributes positively to the mention of rapid strengthening, but its prediction still captures only a fraction of the observed change. The numerical predictions are in \autoref{tab:ig_gallery_cases}.

\emph{Cases with smaller or negative gains.} Calvin improves with discussion, while Philippe and Darby lose accuracy. The tabular ensemble already approaches Darby's observed weakening, and adding discussion moves its prediction farther away. These examples show that gains associated with discussions vary across cases, including when the tabular baseline is already accurate.

\emph{Progression across forecast leads.} The same Ian, Lee, and Idalia discussions are also examined with separately trained 48\,h and 72\,h checkpoints. These are not successive forecasts from a single joint model. The inline D maps for Ian at 48\,h and Lee at 72\,h satisfy the numerical criteria. The selected predictions vary across lead times, and the captions of Figures~\ref{fig:gallery_ian_48h} and~\ref{fig:gallery_lee_72h} report the 48\,h prediction for Ian and the 72\,h prediction for Lee. They do not establish a general learned mechanism for anticipating or limiting rapid intensification.

\newcommand{\gallerymap}[3]{%
\begin{figure}[!htbp]
  \centering
  \includegraphics[width=\linewidth,height=0.92\textheight,keepaspectratio]{assets/images/interpretability/#1}
  \caption{#3}
  \label{#2}
\end{figure}}

\gallerymap{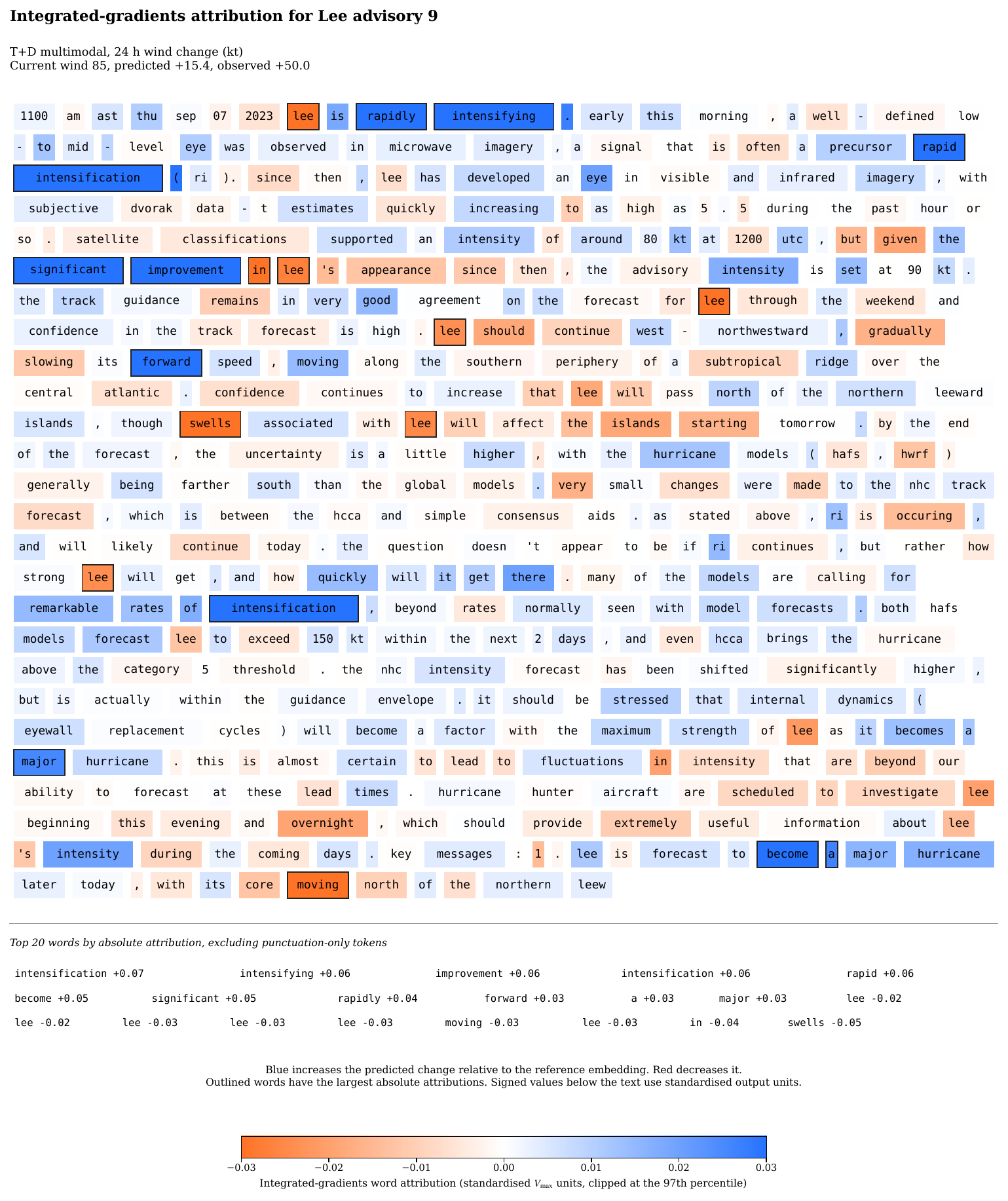}{fig:gallery_lee24}{Lee Discussion 9, T+D seed~1, 24\,h. The attributed checkpoint predicts $\resultvalue{case/LEE/td_seed1}$\,kt against $\resultvalue{case/LEE/truth_delta}$ observed. The ensemble of five models predicts $\resultvalue{case/LEE/td_ensemble}$\,kt, and its gain over T is $\resultvalue{case/LEE/gain}$\,kt.}
\gallerymap{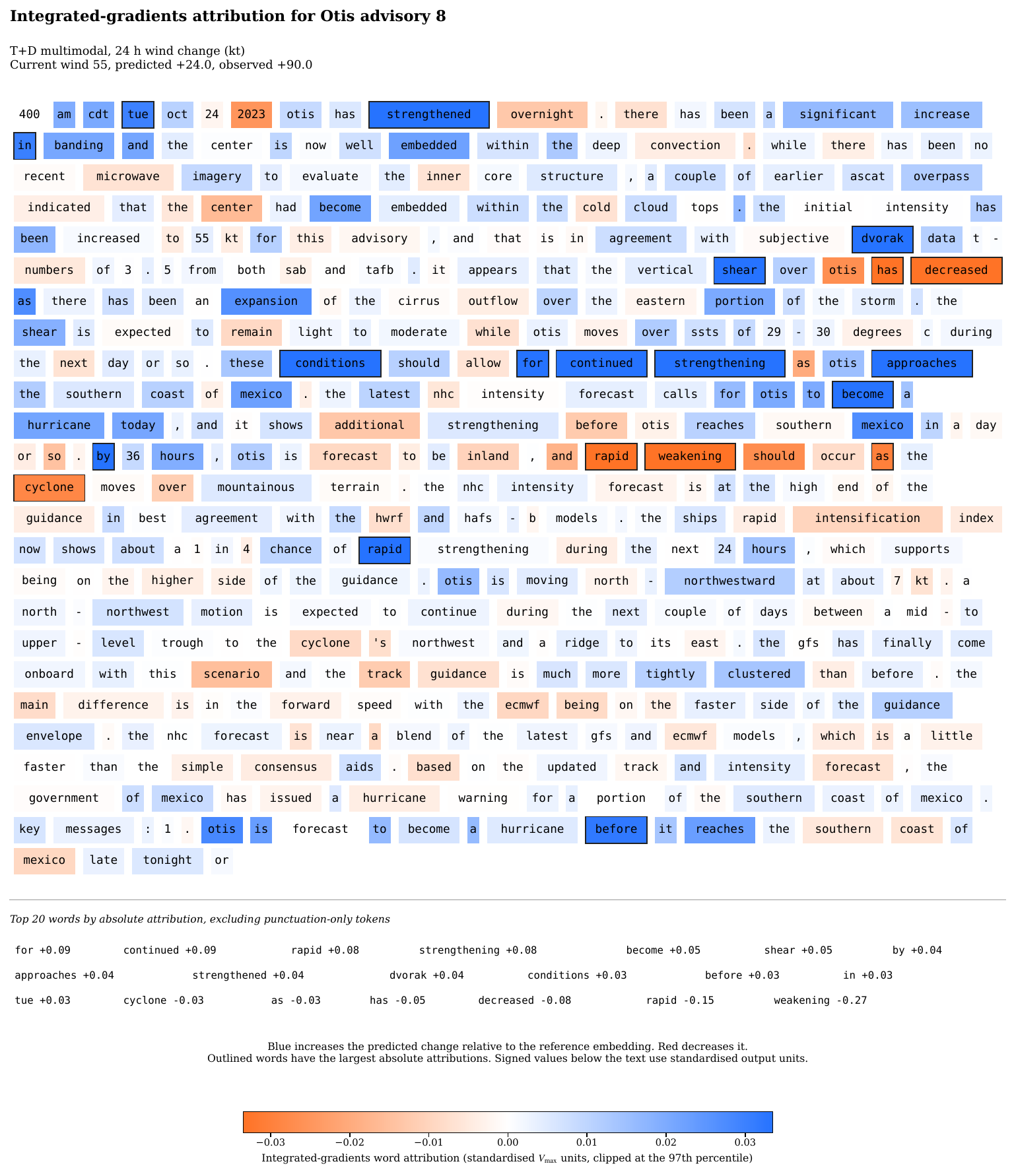}{fig:gallery_otis24}{Otis Discussion 8, T+D seed~1, 24\,h. The discussion anticipates strengthening and cites rapid intensification guidance, but the checkpoint predicts only $\resultvalue{case/OTIS/td_seed1}$\,kt against $\resultvalue{case/OTIS/truth_delta}$ observed.}
\gallerymap{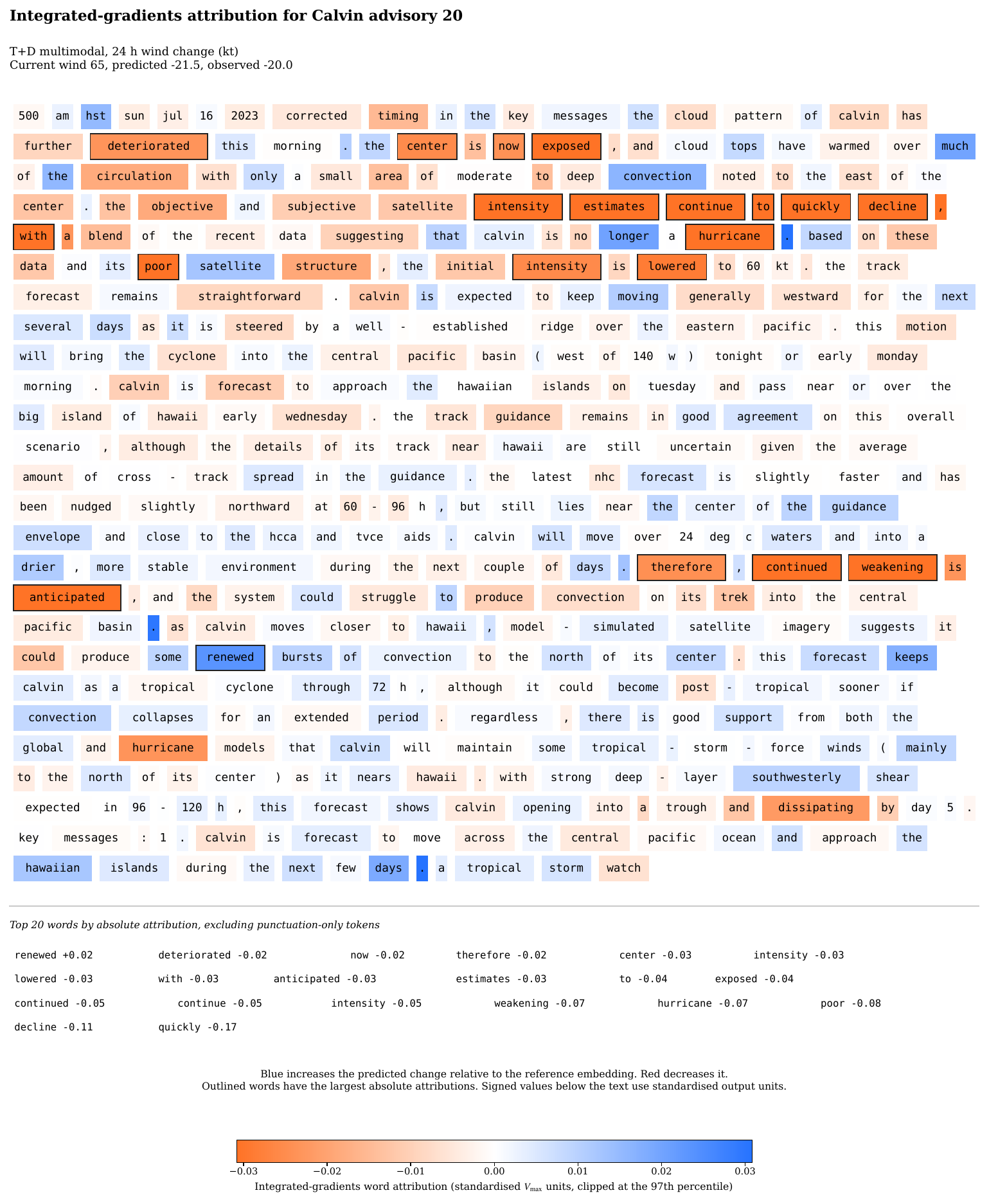}{fig:gallery_calvin24}{Calvin Discussion 20, T+D seed~1, 24\,h. The checkpoint predicts $\resultvalue{case/CALVIN/td_seed1}$\,kt against $\resultvalue{case/CALVIN/truth_delta}$ observed. The T+D ensemble improves on T in this case ($\resultvalue{case/CALVIN/gain}$\,kt relative to T).}
\gallerymap{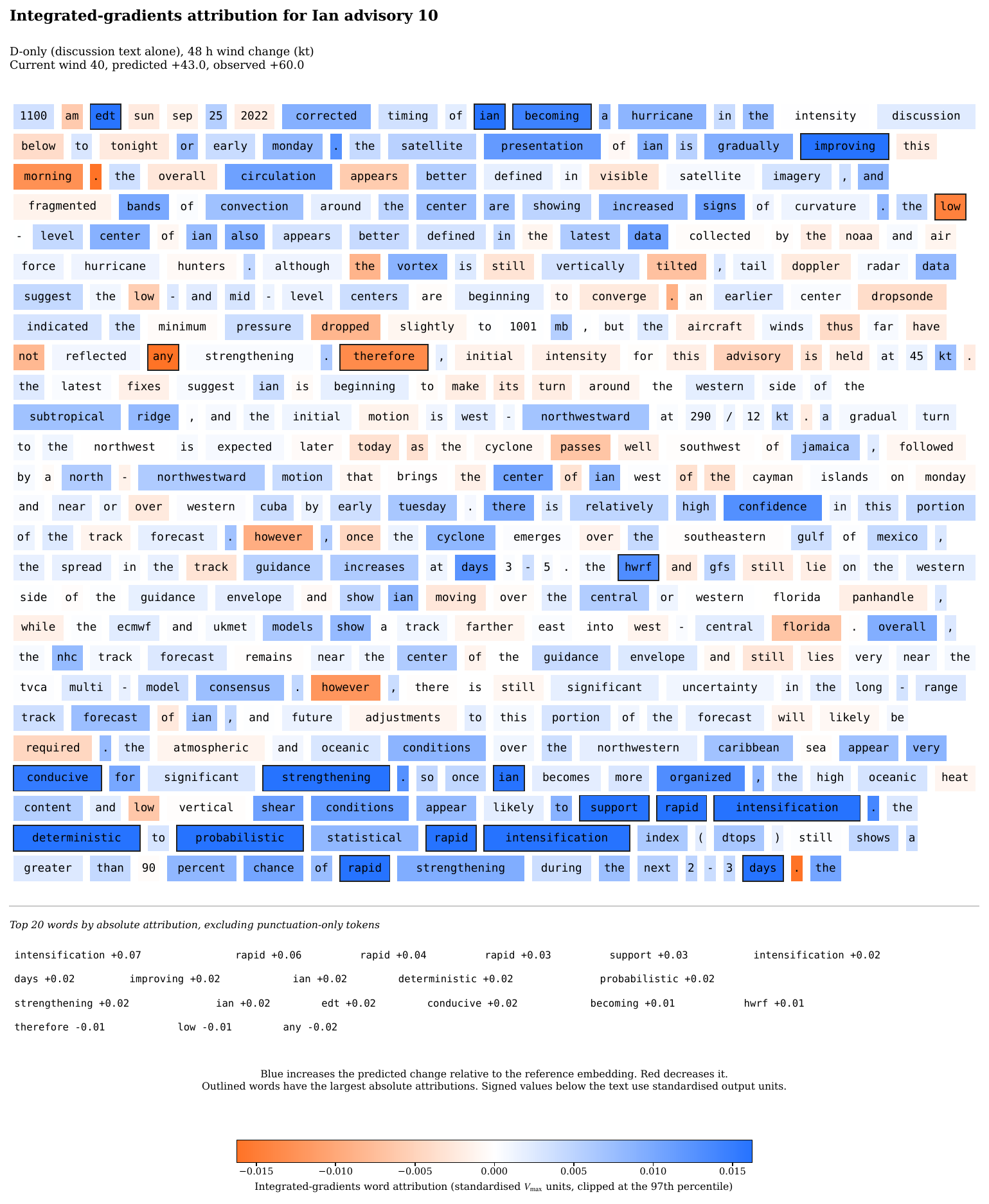}{fig:gallery_ian_48h}{Ian Discussion 10, per-lead D seed~1, 48\,h. The checkpoint predicts $\resultvalue{map/D_IAN_010.202209251500_VMAX_48h/prediction_delta_abs}$\,kt against $\resultvalue{map/D_IAN_010.202209251500_VMAX_48h/true_delta_abs}$ observed. The map passes the stated completeness check.}
\gallerymap{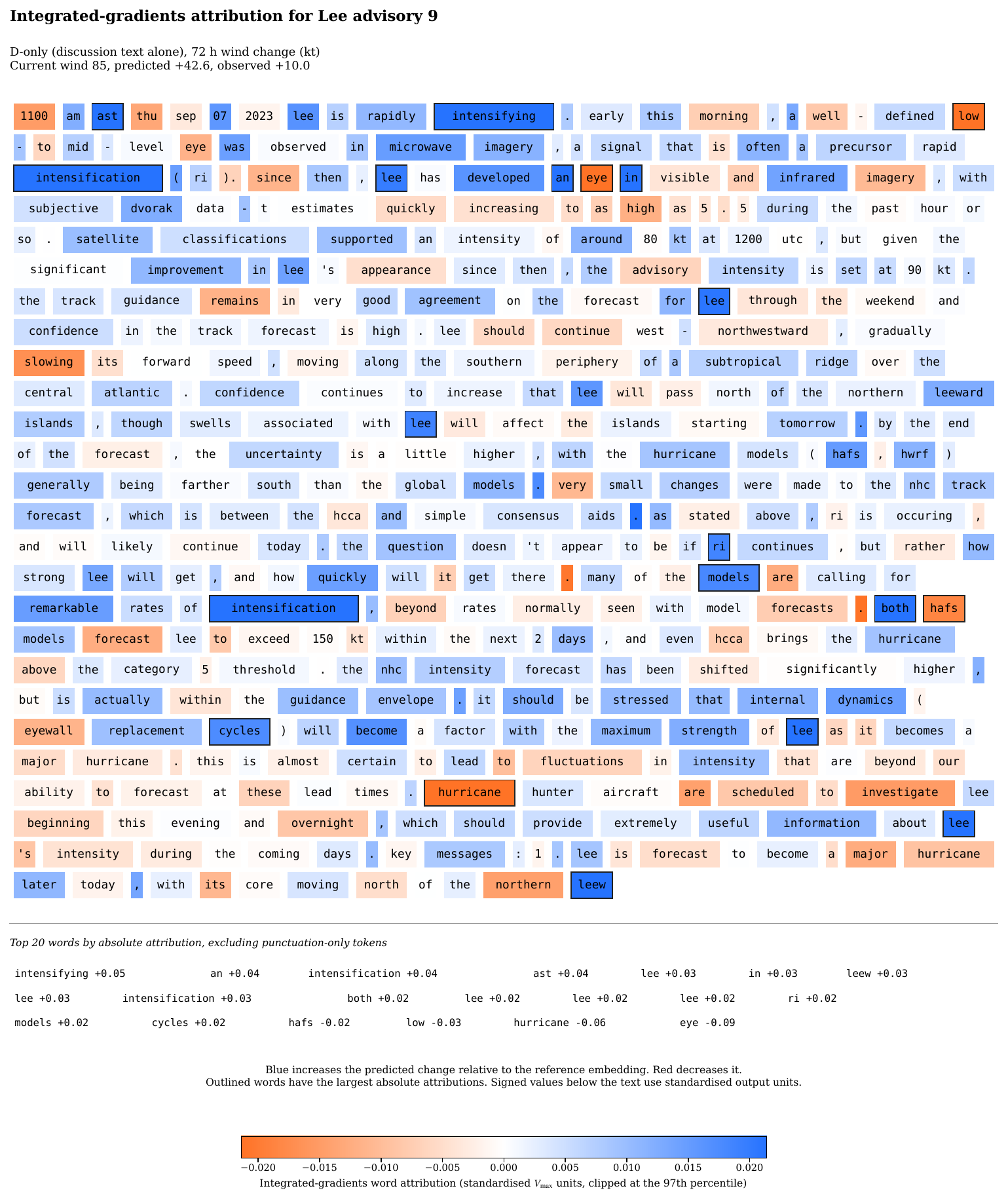}{fig:gallery_lee_72h}{Lee Discussion 9, per-lead D seed~1, 72\,h. The checkpoint predicts $\resultvalue{map/D_LEE_009.202309071500_VMAX_72h/prediction_delta_abs}$\,kt against $\resultvalue{map/D_LEE_009.202309071500_VMAX_72h/true_delta_abs}$ observed. This is a separately trained lead-specific model.}

\FloatBarrier
\section{Decomposing the predictive signal in the discussion-only model}
\label{app:text_only_probes}

The neural comparison across five seeds in \autoref{tab:nn_ablation} reports wind $R^2$ of \resultvalue{seed/abl/disc/VMAX/24/r2_mean} for discussion-only D and \resultvalue{seed/abl/tab/VMAX/24/r2_mean} for the tabular model with 13 SHIPS predictors. These models are evaluated on different test populations, so this is not a paired comparison. We examine D using a bag-of-words baseline and perturbations of the text supplied to all five fixed standalone D checkpoints in the reference set. Their unperturbed predictions agree with independently exported reference predictions to numerical precision. We report mean and sample SD across these training seeds. For random sentence dropout, five mask realizations are scored within each checkpoint before averaging those scores and aggregating across checkpoints.

\subsection{Bag-of-words baseline}
\label{app:text_only_tfidf}

As a lexical baseline we trained a Ridge regressor on a 1--2-gram TF-IDF representation of the cleaned discussion text (20{,}000-feature cap, sublinear term-frequency weighting, document-frequency bounds $[3, 0.95N]$, and regularization $\alpha$ chosen from $\{0.1, 0.3, 1, 3, 10, 30, 100\}$). The temporal train, validation and test splits follow the same convention as the neural models, and the vocabulary is fitted on training discussions only. Alpha is selected by case-level leave-one-out cross-validation within the training split, without grouping by storm. The temporal validation split is evaluated separately and does not select alpha. Both targets select the value shown in \autoref{tab:text_only_tfidf}. TF-IDF uses full cleaned documents, whereas neural inference truncates each input to 512 tokenizer tokens, so the protocols for tuning and input length differ.

\begin{table}[!ht]
\centering
\small
\caption{TF-IDF + Ridge text-only baseline at 24\,h on 1,105 matched-discussion cases from 64 test storms. The D column reports mean individual-seed $R^2$ over the five standalone neural models in \autoref{tab:nn_ablation}. The perturbation table uses these same five trained models. Alpha selection and input-length handling differ between the lexical and neural models.}
\label{tab:text_only_tfidf}
\begin{tabular*}{\linewidth}{@{\extracolsep{\fill}}lcccc@{}}
\toprule
Target & Selected $\alpha$ & Test $R^2$ & Test MAE & D mean $R^2$ \\
\midrule
$\Delta V_{\max}$ (24\,h) & 1.0 & $0.41$ & 11.1 kt & $0.50$ \\
$\Delta$MSLP (24\,h) & 1.0 & $0.35$ & 7.8 hPa & $0.46$ \\
\bottomrule
\end{tabular*}
\end{table}

The lexical baseline attains wind $R^2=\resultvalue{probe/tfidf/VMAX/r2}$, compared with a neural mean across five seeds of \resultvalue{probe/tfidf/VMAX/neural_mean_r2}. This demonstrates predictive utility of lexical features without measuring the fraction of information attributable to lexical or semantic mechanisms. The Ridge coefficients describe associations in the fitted representation. Terms with large positive coefficients include ``strengthening'', ``intensification'', and ``rapid intensification'', while large negative coefficients include ``inland'', ``weakening'', and ``landfall''. Their coefficients do not establish memorization, leakage, or a causal contribution to the separate neural model.

\subsection{Perturbations of the input text}
\label{app:text_only_perturbations}

Each perturbation is applied to the cleaned discussion text before re-tokenization and truncation to 512 tokens. We then run each of the five fixed standalone D checkpoints on the matched-discussion test set with a valid 24\,h target. All perturbations are deterministic except the random-dropout reference, which uses five seeded mask realizations per checkpoint. Scores for those five masks are averaged within each checkpoint before the mean and sample SD across five checkpoints are calculated. Sentence deletion can change which later passages enter the model's input window.

\begingroup
\let\small\footnotesize
\setlength{\tabcolsep}{3pt}
\begin{table}[!ht]
\centering
\small
\caption{Input perturbations of five standalone discussion models on 1{,}105 matched-discussion test cases from 64 storms at 24\,h. Values are means $\pm$ sample SD across five training seeds, using raw observed wind changes. Negative $\Delta R^2$ indicates deterioration from each checkpoint's unperturbed baseline. The five realizations of random sentence dropout are averaged within each checkpoint before seed aggregation. Dropout removes each sentence with probability $0.5$. It does not isolate effects of specific content.}
\label{tab:text_only_perturb}
\begin{tabular*}{\linewidth}{@{\extracolsep{\fill}}lrrr@{}}
\toprule
Perturbation & Wind $R^2$ & $\Delta R^2$ & MAE (kt) \\
\midrule
Baseline (no perturbation) & $0.500 \pm 0.015$ & --- & $9.90 \pm 0.20$ \\
Mask storm name & $0.509 \pm 0.017$ & $+0.009 \pm 0.016$ & $9.77 \pm 0.20$ \\
Mask numbers & $0.486 \pm 0.020$ & $-0.014 \pm 0.009$ & $10.10 \pm 0.32$ \\
Keep only outlook sentences & $0.408 \pm 0.050$ & $-0.091 \pm 0.039$ & $10.63 \pm 0.47$ \\
Drop outlook sentences & $0.249 \pm 0.030$ & $-0.251 \pm 0.019$ & $12.14 \pm 0.28$ \\
Random $50\%$ sentence dropout & $0.305 \pm 0.039$ & $-0.195 \pm 0.028$ & $11.51 \pm 0.36$ \\
Mask numbers + mask storm name + drop outlook & $0.245 \pm 0.035$ & $-0.255 \pm 0.025$ & $12.14 \pm 0.32$ \\
Drop intensity-vocabulary sentences & $0.136 \pm 0.035$ & $-0.364 \pm 0.022$ & $12.87 \pm 0.29$ \\
\bottomrule
\end{tabular*}
\end{table}

\endgroup

The perturbations fall into two categories. Masking replaces specified text with a placeholder, covering the storm name with its possessive, or a regex-matched numeric token together with an optional recognized trailing unit (e.g.\ kt, knots, mb, hPa, mph, km, nm, m/s, mi, ft, in, deg, $^\circ$, and single-letter compass or temperature abbreviations). This rule excludes numbers embedded in alphanumeric identifiers and does not mask every standalone unit or arbitrary compound expression. Sentence-level removals delete every regex-defined sentence containing at least one word from a short vocabulary comprising outlook terms such as \emph{expected}, \emph{will}, \emph{likely}, \emph{could}, \emph{may}, \emph{continued}, and \emph{over the next}, or intensity terms such as \emph{strengthen}, \emph{weaken}, \emph{intensify}, \emph{major hurricane}, \emph{category}, \emph{hurricane}, \emph{tropical storm}, \emph{post-tropical}, \emph{extratropical}, and \emph{eyewall}. The composite applies numeric masking, storm-name masking, and outlook-sentence removal in that order.

\autoref{tab:text_only_perturb} measures the five checkpoints\textquotesingle\ sensitivity to numbers, storm names, and sentence removal. We define $\Delta R^2$ as the perturbed score minus the unperturbed score, so negative values indicate worse performance and positive values indicate better performance. Masking numbers changes $R^2$ by \resultvalue{probe/perturb/mask_numbers/delta_r2}. Masking storm names changes it by \resultvalue{probe/perturb/mask_storm_name/delta_r2}. Removing outlook sentences changes $R^2$ by \resultvalue{probe/perturb/drop_outlook/delta_r2}, compared with \resultvalue{probe/perturb/random_dropout_50/delta_r2} for the seeded random-dropout reference. The largest observed change follows removal of intensity-vocabulary sentences at \resultvalue{probe/perturb/drop_intensity_vocab/delta_r2}. The sentence masks overlap and differ in how many words and positions they remove. Deletion also changes context, length, and truncation, while the random reference averages only five mask realizations per checkpoint. These contrasts therefore do not isolate semantic channels or determine whether outlook prose supplies a timing advantage. The composite probe is close to outlook removal alone, consistent with overlapping interventions but not proof of independent information channels.

\subsection{Synthesis}
\label{app:text_only_synthesis}

The neural model's wind $R^2$ exceeds the lexical baseline. This difference could reflect representations, optimization, handling of input length, or sensitivity to contextual patterns such as \emph{further strengthening is expected}, \emph{has become a major hurricane}, and \emph{rapid intensification likely}. The comparison does not separate these explanations. Across the five fixed checkpoints, the perturbations show a larger mean deterioration after removing intensity-vocabulary sentences than after masking numeric tokens or storm names. Together, these results demonstrate predictive utility of lexical features and sensitivity to the specified interventions. They do not establish a unique learned mechanism, guarantee the same pattern for other training configurations or text interventions, or rule out all forms of information leakage.

\subsection{Stratified utility of the text modality on the 24\,h test set}
\label{app:subj_stratified_gain}

We compare predictions from the fixed tabular ensemble and the ensemble that includes discussions on the held-out 2022--2023 cases with matched discussions. Rows without a discussion are excluded from this analysis. These fixed checkpoints were not newly selected by this stratified comparison. Each retained case is flagged \textsc{opinionated} when its matched discussion belongs to the deterministic corpus list produced by the regex for first-person language (Appendix~\ref{app:subj_pipeline}). For each stratum and target, gain is tabular MAE minus MAE with discussions. We resample whole storms with replacement 10{,}000 times, retaining all rows of each sampled storm and calculating the pooled, case-weighted paired gain. The reported 95\% intervals are pointwise percentile intervals, conditional on the fitted ensembles. They do not incorporate retraining uncertainty or adjust for multiple strata.

\begin{table}[!htbp]
\centering
\caption{24\,h gain from T to T+D using ensembles of five models on matched discussions, stratified by the fixed first-person regex flag. The flag is a proxy based on wording, not a validated measure of expert judgment. The flagged group is small, so its estimated gain should be read with its uncertainty interval.}
\label{tab:strat_gain}
\small
\renewcommand{\arraystretch}{1.12}
\setlength{\tabcolsep}{4pt}
\begin{adjustbox}{max width=\linewidth}
\begin{tabular}{@{}lccccc@{}}
\toprule
Stratum & $N/S$ & Target & T MAE & T+D MAE & Gain [95\% CI] \\
\midrule
\multirow{2}{*}{All matched} & 1{,}105/64 & $V_{\max}$ (kt) & $10.05$ & $8.98$ & $1.07$ $[0.37, 1.82]$ \\
 & 1{,}105/64 & MSLP (hPa) & $6.55$ & $5.94$ & $0.61$ $[0.10, 1.12]$ \\
\midrule
\multirow{2}{*}{Regex present} & 17/13 & $V_{\max}$ (kt) & $10.81$ & $11.46$ & $-0.65$ $[-4.51, 3.45]$ \\
 & 17/13 & MSLP (hPa) & $6.86$ & $6.98$ & $-0.12$ $[-2.64, 2.49]$ \\
\midrule
\multirow{2}{*}{Regex absent} & 1{,}088/64 & $V_{\max}$ (kt) & $10.04$ & $8.94$ & $1.10$ $[0.39, 1.85]$ \\
 & 1{,}088/64 & MSLP (hPa) & $6.55$ & $5.93$ & $0.62$ $[0.10, 1.14]$ \\
\bottomrule
\end{tabular}
\end{adjustbox}
\par\smallskip
{\footnotesize Wind/pressure MAE units are kt/hPa. All intervals resample whole storms (10,000 draws, percentile 95\%), retain within-storm dependence, and are pointwise without multiplicity adjustment. Stratum differences are not formal interaction tests.}
\end{table}

On the full discussion-matched subset, the wind MAE reduction is \resultvalue{cell/strat_gain/discussion_present/mae_gain_left_to_rightVMAX/mean}\,kt, and the pressure reduction is \resultvalue{cell/strat_gain/discussion_present/mae_gain_left_to_rightMSLP/mean}\,hPa. Their intervals are shown in \autoref{tab:strat_gain}. The non-\textsc{opinionated} stratum retains a wind gain of \resultvalue{cell/strat_gain/regex_not_opinionated_matched/mae_gain_left_to_rightVMAX/mean}\,kt. In the small flagged stratum, the wind gain is \resultvalue{cell/strat_gain/regex_opinionated_matched/mae_gain_left_to_rightVMAX/mean}\,kt with a storm-resampled interval of $[\resultvalue{cell/strat_gain/regex_opinionated_matchedVMAX/storm_ci_low/mean},\resultvalue{cell/strat_gain/regex_opinionated_matchedVMAX/storm_ci_high/mean}]$\,kt. Its interval spans zero, as does the pressure interval in the table. The larger stratum not flagged by the regex supports the overall association between discussion inputs and lower error. The flagged stratum supports neither a beneficial nor harmful judgment-specific effect. Separate confidence intervals for these strata do not test their difference, and the regex partition does not isolate the causal value of a sentence type.

Few cases in this held-out analysis carry the first-person flag, so a reliable estimate of a differential effect would need a larger, independently defined holdout. The flag is a reproducible marker of first-person language but not a complete measure of forecaster judgment, because interpretive reasoning can occur without a first-person pronoun. The stratified gains and their intervals use only this fixed regex definition, and labels from another classifier would define a different analysis.

\end{appendices}
\end{document}